# DeformAr: Rethinking NER Evaluation through Component Analysis and Visual Analytics

*A Comparative Case Study of Arabic and English NER*

Ahmed Mostafa Younes

*A Thesis submitted for the degree of Doctor of Philosophy*

School of Engineering and Informatics

University of Sussex

November 2025

*Supervisors:*

Julie Weeds

David Weir

بسم الله الرحمن الرحيم
الحمد لله رب العالمين

# ACKNOWLEDGEMENTS

I would like to express my gratitude to my father for supporting me with everything he could throughout this journey. None of this would have been possible without his support. I am also grateful to my mother for the sacrifices she made so that I could pursue and fulfil my dreams.

I would like to thank my PhD supervisors, Julie Weeds and David Weir, for their guidance, support, and inspiration. It has been a challenging journey, but they continuously pushed me forward. I wouldn't have made it this far without their support.

A very special thank you to my wife — she has stood beside me every step of the way and has witnessed this journey in its entirety. Her support and patience have meant more than words can express.

I would also like to thank my thesis committee, John Carroll and Jeremy Reffin, for their valuable feedback and insightful suggestions. My thanks also go to the Text Analytics Group Lab. Although my time there in person was brief, I am pretty sure it would have been a wonderful experience.

I am also grateful to the team at CASM Technology. It has been a pleasure to work in such a positive and supportive environment, which has helped shape me into a better researcher and data scientist. A special thanks to Chris, whose collaboration had a massive impact on my work. I would also like to thank Jack, Nestor, and Andy for their help, advice, and invaluable insights — especially regarding their own PhD experiences.

I would like to thank my PhD supervisors, Julie Weeds and David Weir, for their guidance, support, and inspiration. It has been a challenging journey, but they continuously pushed me forward. I wouldn't have made it this far without their support.

A very special thank you to my wife — she has stood beside me every step of the way and has witnessed this journey in its entirety. Her support and patience have meant more than words can express.

I would also like to thank my thesis committee, John Carroll and Jeremy Reffin, for their valuable feedback and insightful suggestions. My thanks also go to the Text Analytics Group Lab. Although my time there in person was brief, I am pretty sure it would have been a wonderful experience.

I am also grateful to the team at CASM Technology. It has been a pleasure to work in such a positive and supportive environment, which has helped shape me into a better researcher and data scientist. A special thanks to Chris, whose collaboration had a massive impact on my work. I would also like to thank Jack, Nestor, and Andy for their help, advice, and invaluable insights — especially regarding their own PhD experiences.

# Declaration

I, Ahmed Mostafa Younes, hereby declare that this thesis has not been and will not be, submitted in whole or in part to another university for the award of any other degree.

*Brighton,*

*November 2025*

Ahmed Mostafa Younes

# Abstract

Transformer models have significantly advanced Natural Language Processing (NLP), demonstrating strong performance in English. However, their effectiveness in Arabic particularly for Named Entity Recognition (NER) remains limited, even when Arabic models are pre-trained on larger datasets. This performance gap may be attributed to multiple factors, including tokenisation, dataset quality, and annotation inconsistencies. While previous studies have examined these issues individually, analysing them in isolation makes it difficult to understand how they interact and jointly affect system behaviour and performance.

This thesis introduces DeformAr (Debugging and Evaluation Framework for Transformer-based NER Systems), a framework designed to investigate the performance discrepancy between Arabic and English NER systems and explore the factors behind this gap. DeformAr integrates a data extraction library and an interactive dashboard supporting two modes of evaluation: cross-component analysis and behavioural analysis. The framework divides each language into dataset and model components and examines interactions between them. During the feature examination phase, we select subcomponents for analysis based on their expected contribution to performance variation.

The analysis proceeds in two stages. First, cross-component analysis provides systematic and behavioural diagnostic measures across both data and model subcomponents, addressing the "what," "how," and parts of the "why" behind observed performance discrepancies. Building on these findings, the second stage applies behavioural analysis by combining interpretability techniques with token-level metrics, interactive visualisations, and representation space analysis. Together, these stages enable a component-aware diagnostic process that not only detects model behaviours but also explains them by linking them to underlying representational patterns and data-related factors. To our knowledge, this is the first Arabic-specific, component-based interpretability tool offering a novel resource for advancing model analysis in under-resourced languages.

**Keywords:** Arabic NLP, Named Entity Recognition, Transformer Models, Evaluation Framework, Interpretability, Tokenisation, Cross-lingual Analysis, Behavioural Diagnostics, Representation Space Analysis, DeformAr.

# Contents

# List of Abbreviations

| | |
|---|---|
| AraBERT | Arabic Bidirectional Encoder Representations from Transformers |
| BERT | Bidirectional Encoder Representations from Transformers |
| NER | Named Entity Recognition |
| NLP | Natural Language Processing |
| PLM | Pretrained Language Model |
| TP | True Positive |
| TN | True Negative |
| FN | False Negative |
| FP | False Positive |

# Introduction 1

Transformer-based models have significantly improved performance in many Natural Language Processing (NLP) tasks, especially in English (Peng et al., 2023; Brown et al., 2020; Patwardhan et al., 2023; Cai et al., 2023; Morsali et al., 2023; Devlin et al., 2019; Wang et al., 2018). However, their performance in other languages, such as Arabic, remains behind. In the task of Named Entity Recognition (NER), for instance, the English version of BERT has achieved F1 scores above 92.4 (Devlin et al., 2019), while AraBERTv01[1] has reached only 84.2 (Antoun et al., 2020)—despite being pretrained on a larger dataset (23GB for AraBERTv01 vs. 16GB for BERT-base) (Antoun et al., 2020; Liu et al., 2019).

This raises a key question: why does Arabic show weaker performance despite larger training data, and what factors contribute to this gap? This thesis investigates this discrepancy by comparing Arabic and English NER systems built on transformer architectures—specifically BERT, with a focus on diagnostic and interpretability challenges. While Arabic NLP has gained attention, systematic evaluation—particularly for NER—remains limited.

In this thesis, interpretability refers to the ability to explain or present how a model processes inputs and makes decisions in a way that humans can understand—by showing how internal mechanisms lead to specific outputs (Saha et al., 2022; Zini and Awad, 2023; Singh et al., 2024). Most multilingual studies rely on high-level performance metrics such as precision, recall, and F1-score, offering little insight into how specific model or data characteristics shape those outcomes (Ruder et al., 2022). English NLP benefits from mature interpretability tools (Sun et al., 2021; Ruder et al., 2022), but Arabic remains under-explored—both in system performance evaluation and in interpretability tools for diagnostic analysis. NER, in particular, lacks interpretability work in general (Sun et al., 2021; Jain et al., 2022).

[1] This thesis uses AraBERTv02, a newer and larger version trained on 77GB of data, for all experiments.

Reproducibility is another challenge in NER research. Reported results often vary, and reproducing them is difficult due to inconsistent preprocessing, undocumented scorer behaviour, and unclear evaluation procedures (Lignos and Kamyab, 2020; Palen-Michel et al., 2021; Cuevas Villarmin et al., 2024). These issues, while not language-specific, further show that standard metrics alone are insufficient for understanding or diagnosing why NER performance varies across languages.

To address these issues, this thesis introduces DeformAr (Debugging Transformer Models for Arabic-English NER), a modular diagnostic and evaluation framework for comparative NER analysis. DeformAr supports structured comparison across languages using both metric-based and interpretability approaches. The name reflects its focus on debugging transformer models, with "Ar" highlighting its emphasis on Arabic-language evaluation. It enables component-level analysis of model behaviours and dataset characteristics through cross-component and behavioural analysis. While language-agnostic in design, in this thesis we apply DeformAr as a case study on Arabic and English to identify and explain the performance gap. Unlike existing tools, DeformAr supports multi-level debugging, comparative error analysis, and interpretability—all within a single extensible interface.

## 1.1 Rationale for Focusing on Arabic and Sequence Labelling

Given the observed performance gap between Arabic and English in transformer-based models, this thesis adopts a comparative case study approach to examine the underlying causes of these discrepancies in the context of sequence labelling tasks. The goal is not only to analyse language-specific factors but also to demonstrate how a diagnostic framework like DeformAr can support structured, cross-lingual evaluation. This section outlines the rationale for focusing on Arabic as the primary language of interest—using

English as a baseline—and for selecting sequence labelling as the task domain.

### 1.1.1 Why Arabic vs English

In the field of NLP, English has been the default language for algorithm development and evaluation (Bender, 2019). Recently, the field has seen significant advances in methodologies such as transformers (Vaswani et al., 2017), leading to improved understanding of English processing—particularly in terms of model performance and interpretability (Clark et al., 2019). However, this Anglocentric perspective raises questions about the universality of transformer-based methods, often leaving other languages less explored and understood (Hovy and Prabhumoye, 2021; Bender, 2019). For instance, Bender (2011) showed that traditional n-gram models perform poorly for morphologically rich languages due to data sparsity challenges. Bender (2019) also reported that NLP conference reviewers often conflate "state of the art" with results on English datasets, undervaluing work that doesn't benchmark against English. This issue is further supported by Ruder et al. (2022), who found that nearly 70% of the papers they analysed evaluate exclusively on English, highlighting the lack of linguistic diversity in NLP. In terms of interpretability, Ruder et al. (2022) noted that most BlackBox NLP papers focus on English, while Lertvittayakumjorn and Toni (2021) found that the majority of explanation-based debugging studies rely solely on English datasets.

Recognising the dominance of English in NLP research, it is important to acknowledge growing efforts to expand language coverage. There is increasing recognition of the need to evaluate and interpret transformer models across languages. Arabic, with its distinct linguistic features and status as a high-resource language (Bender, 2019), is well suited for such comparative research. This is particularly relevant given the earlier-noted anomaly: Arabic models often underperform despite being trained on more data. Additionally, despite available resources, Arabic remains under-represented in interpretability research (Ruder et al., 2022; Lertvittayakumjorn and Toni, 2021). These factors make Arabic an informative case for evaluating how well current NLP methods generalise beyond English. This thesis focuses specifically on Modern Standard Arabic (MSA), as it represents the most standardised form of the language in terms of syntax, morphology, and semantics. For further detail on the language variety, see Chapter 2.

Focusing on Arabic—with English as a comparative baseline—can help test the limits of current models and contribute to advancing Arabic NLP. Comparing the two highlights how transformer models handle structurally different languages and reveals areas where further development is needed. It also offers insight into how language influences interpretability—challenging the assumption that findings and tools developed for English generalise easily to other languages. This comparative approach supports more equitable development of NLP methods across linguistic contexts.

### 1.1.2 Why Sequence Labelling

According to Jurafsky and Martin (2008), sequence labelling refers to the task of assigning a label to each element in a sequence of words, where each label corresponds to a word in the input sequence. A common example of sequence labelling is NER, which involves identifying spans of text that represent proper names and classifying them into entity types such as names of people, organisations, and locations. These tasks have been used extensively in downstream applications such as information retrieval, machine translation, and question answering (He et al., 2020).

As previously discussed, there is a gap in interpretability research within Arabic NLP, However, this gap is not limited to language; it extends across various NLP tasks, particularly sequence labelling. For instance, Sun et al. (2021) observed that most interpretability methods have been developed for text classification, which raises questions about their applicability to tasks such as NER. Interpretability in sequence labelling involves examining the model's behaviour at the level of individual tokens, unlike text classification, which focuses on broader representations such as CLS token or averaged token embeddings. This difference suggests that studying sequence labelling can offer more fine-grained insights into how models understand context and token-level semantics. Such analysis can help explain performance differences across languages and model configurations.

## 1.2 Contributing Factors and Their Interactions in NER Performance

Performance in NER systems is influenced by multiple interdependent factors. Broadly, these can be grouped into three categories: model architecture, language complexity, and training procedure. Each has been studied in prior work, but typically in isolation. For example, subword tokenisers like BPE or WordPiece have been shown to misalign with Arabic morphology, leading to fragmented representations and degraded performance (Bostrom and Durrett, 2020; Kiyono et al., 2023; Alkaoud and Syed, 2020). Other work has investigated how linguistic features—such as flexible word order, rich morphology, and lack of capitalisation—pose challenges for sentence segmentation and named entity identification in Arabic (Farghaly and Shaalan, 2009; Darwish et al., 2021; Shaalan et al., 2018). Training decisions such as annotation schemes, class imbalance, and label quality also affect performance, especially in under-resourced settings. Inconsistent or noisy annotations, in particular, can introduce instability and lead models to overfit to artefacts in the data rather than generalisable patterns (Ratinov and Roth, 2009; Stanislawek et al., 2019; Rücker and Akbik, 2023).

While these studies offer valuable insights, they often assess individual factors in isolation, overlooking how these elements interact in practice. For instance, tokenisation challenges intersect with linguistic structure, and annotation inconsistency can introduce noise that imposes a glass ceiling on model performance (Stanislawek et al., 2019). This noise may stem from data characteristics such as ambiguity or sparsity, or from system configurations—such as model architecture, fine-tuning strategies, or tokenisation choices—that influence how the model learns from the data. This isolated investigation of contributing factors highlights the need for a more integrated evaluation approach. Rather than revisiting each factor exhaustively, this thesis focuses on their interplay—examining how these components jointly shape system behaviour across languages, and how diagnostic tools can help uncover such interactions. A more detailed review of the supporting literature is provided in Chapter 2, while the methodological limitations of existing evaluation practices are discussed in the next section.

## 1.3 Limitations of Existing Evaluation Approaches

NER research can be broadly grouped according to its primary analytical focus. One strand is data- or model-centric: these studies investigate how specific components—such as tokenisation, annotation schemes, or architectural choices—affect performance, typically by isolating individual variables (Bostrom and Durrett, 2020; Rust et al., 2021). A second strand is methodology-centric, applying tools such as probing, attention visualisation, or interpretability techniques, but often without a guiding hypothesis related to linguistic or task-specific behaviour (Jain et al., 2022; Lee et al., 2023a). A third, and often underdeveloped, dimension concerns language. As Ruder et al. (2022) note, most NLP studies follow a "square one" setup—optimising accuracy on English—while extensions beyond this often explore only a single axis of variation. For example, multilingual studies typically focus on accuracy alone, while interpretability or fairness research is usually restricted to English. As a result, language-specific insights are rarely explored in relation to data or model behaviours.

This thesis addresses that gap by proposing a multi-dimensional diagnostic framework that explicitly links model behaviour, data characteristics, and language variation. DeformAr enables structured, comparative evaluation across components and across languages, combining metrics and interpretability to uncover the underlying causes of performance differences. This supports more transparent, language-aware system analysis and helps move beyond one-dimensional evaluations.

## 1.4 Towards Multi-Dimensional Evaluation

To investigate the performance discrepancy between Arabic and English in transformer-based NER systems—and to address the methodological limitations outlined in the previous section—this thesis proposes a modular diagnostic framework: **DeformAr** (Debugging Transformer Models for Arabic-English NER). DeformAr supports both metric-based and interpretability-driven analysis, enabling systematic comparison across model and data components. While it is designed to be extensible and language-agnostic, this thesis applies it as a comparative case study focused on Arabic and English.

The framework builds on the idea that NER performance is shaped by two main components: the data and the model. Each component can be analysed independently—for example, by examining annotation patterns or inspecting model outputs—but they also interact in complex ways that influence overall behaviour. For instance, ambiguous tokens and class imbalance in the data can skew model predictions, while tokenisation affects how input is segmented and represented.

DeformAr operates through a three-phase workflow: extraction, evaluation, and exploration. In the extraction phase, the framework decomposes both the model and data into subcomponents. The data component includes lexical characteristics (e.g., ambiguity, frequency), entity distribution, and annotation consistency. The model component consists of the pretrained encoder and the output layer, which can be examined independently or in relation to the data. This enables analysis of model–data interactions, such as how annotation inconsistency affect representation quality or how annotation imbalance influences prediction bias. All subcomponents are quantified using system-level metrics and token-level behavioural measures, which are detailed further in Chapter 3.

The extracted subcomponents are then visualised and analysed through an interactive dashboard, which supports two complementary modes of evaluation. The first is cross-component evaluation, which presents structured comparisons across model and data subcomponents. This includes metrics drawn from data characteristics (e.g., ambiguity rate, class distribution), model performance (e.g., prediction confidence, class-level performance), and behavioural patterns (e.g., tokenisation rate, annotation ambiguity or inconsistency). These views support direct comparisons between Arabic and English across three dimensions—data characteristics, model behaviour, and performance metrics—and help establish relationships between them.

The second mode is behavioural analysis, which connects data and model components through token-level metrics, representation space analysis, and interactive visualisation techniques. These views enable instance-level analysis and highlight how specific properties—such as annotation ambiguity or tokenisation rate—relate to model confidence, representation quality, and prediction patterns. By explicitly linking characteristics in the data (e.g., ambiguity, class imbalance, entity frequency) to observable patterns in the model's outputs (e.g., low-confidence predictions, misclassifications) and representation space (e.g., overlapping or poorly separated clusters), this mode provides a direct, interpretable explanation for cross-lingual performance differences. It addresses a key gap in existing work, where interpretability is rarely used to establish concrete causal links between data properties and model behaviour.

Together, these features form the basis of the thesis's multi-layered, multi-dimensional analytical approach. They support the research aims and questions outlined in the next section and guide the structure of the overall investigation.

## 1.5 Research Aim and Questions

The primary aim of this thesis is to investigate the performance discrepancy between Arabic and English in BERT-based NER systems. To achieve this, we develop **DeformAr**, a modular diagnostic framework for NER evaluation. DeformAr enables comparative analysis across model and data components by combining cross-component evaluation with behavioural analysis. It focuses on identifying how key contributing factors—such as tokenisation, morphological complexity, and annotation inconsistency (introduced in Section 1.2)—influence model behaviour and performance across languages.

Guided by this aim, the following research questions are addressed:

1. **RQ0:** How can model and data components be decomposed and represented in a way that supports interpretable, comparative analysis across languages?
2. **RQ1:** What data-specific factors—such as annotation patterns, tokenisation, or lexical variation—influence model performance across the two languages?
3. **RQ2:** How do model-specific behaviours—such as learned representations, prediction confidence, and loss—differ between Arabic and English, and how are these linked to performance discrepancies?
4. **RQ3:** How does model performance differ between Arabic and English in NER, and what patterns emerge when considering standard evaluation metrics, annotation schemes, and scoring behaviours?
5. **RQ4:** How do interactions between data characteristics and model behaviours influence performance across languages?

6. **RQ5:** How can interpretability techniques, visual analytics, and behavioural analysis be used to (i) connect model behaviour to dataset characteristics, (ii) explore their interactions, and (iii) link these insights back to the underlying contributing factors discussed in Section 1.2?

Each research question is addressed through a structured set of analyses, distributed across the core chapters. RQ0 is tackled in Chapter 3, and encompasses both the conceptual and technical foundations of the framework. The conceptual layer defines how NER systems can be decomposed into interpretable model and data subcomponents. The technical layer focuses on the design of the system pipeline, along with diagnostic metrics and extraction methods that quantify subcomponent behaviours at both the system and token levels.

RQs 1–3 involve cross-lingual comparisons of individual components across three analytical layers: data characteristics, such as lexical and structural properties (e.g., ambiguity, tokenisation behaviour, annotation patterns); model behaviours (e.g., loss distributions, prediction confidence, and representation separability); and standard evaluation metrics.

RQ4 examines how these components interact across layers. For example, sparse and lexically diverse tags like MISC in Arabic often exhibit high overlap with the O label (data layer), leading to exclusion errors where correct entities are mislabelled as non-entities (performance layer), often with low prediction uncertainty (model behaviour layer). In contrast, LOC tags, which tend to have simpler spans but substantial lexical overlap with ORG, frequently result in entity confusion. PER tags, on the other hand, often span multiple words and are more prone to boundary errors. These interaction patterns are tracked across components and mapped to individual entity spans, forming a cross-component view that helps explain cross-lingual performance differences.

RQ5 is addressed through a combination of behavioural analysis, interpretability techniques, and visual analytics that integrate token-level metrics, representation space projections, and contextual views. This question builds on the findings from Chapter 4 and shifts the focus toward deeper, example-driven exploration. For instance, representation overlap is often found to correspond with tokenisation patterns. Through visual analysis, we are able to distinguish whether ambiguity arises from linguistic variation or tokenisation artefacts, and examine how each type influences model confidence, loss, and embedding quality. Together, these views support both quantitative and exploratory analysis, helping to establish causal links between data characteristics and model behaviour.

## 1.6 Thesis Structure

This thesis is organised around a multi-stage investigation into the performance discrepancy between Arabic and English in BERT-based NER systems. The analysis consist of three phases: system decomposition (RQ0), cross-lingual component analysis (RQs 1–3), and interaction-focused exploration (RQs 4–5).

Rather than presenting all background and related work in a single dedicated chapter, this thesis distributes them across chapters where they best support the argument. Chapter 2 provides broad context—including linguistic, modelling, and task-specific foundations—as well as related work relevant to NER, interpretability, and multilingual evaluation, helping to motivate the research questions. More targeted theoretical background, such as definitions of evaluation metrics, is introduced in Chapter 4 to support interpretation of the results. Technical background appears in Chapter 5, where less common techniques adopted or combined in this thesis—such as clustering alignment and attention-based diagnostics—are introduced to support the visual analysis. Related work is also integrated within each chapter to position the analysis within the literature and highlight the specific contributions in each chapter.

The chapters are structured as follows:

- **Chapter 1 — Introduction**
  Outlines the motivation and research gap in Arabic–English NER performance, introduces DeformAr as a diagnostic framework, and presents the research questions guiding the thesis.
- **Chapter 2 — Background and Related Work**
  Reviews linguistic properties of Arabic, transformer-based modelling, and key challenges in NER. Summarises related work on evaluation, interpretability, and multilingual NLP.

- **Chapter 3 — Methodology**
  Defines the architecture of the DeformAr framework and details its three-phase workflow: extraction, evaluation, and exploration. It introduces the system decomposition strategy and the design of component-level metrics, explaining how the pipeline captures interactions between data and model subcomponents. The chapter also outlines how DeformAr is applied as a comparative case study of Arabic and English, forming the basis for the analyses in subsequent chapters. This chapter addresses RQ0.

- **Chapter 4 — Cross-Component Analysis**
  Represents the first analytical stage of the DeformAr dashboard. This chapter applies DeformAr to compare Arabic and English systems across data and model components using standard metrics, error analysis, and representation-based measures. In addition to isolated comparisons, it synthesises insights across layers—grouped by entity types such as MISC—to examine how characteristics in the data layer (e.g. ambiguity) relate to model behaviour (e.g. prediction confidence) and representation structure. This chapter addresses RQs 1–4.

- **Chapter 5 — Behavioural Analysis**
  Represents the second analytical stage of the DeformAr dashboard, extending the cross-component findings through interpretability, visual analytics, and behavioural metrics. It examines how data subcomponents—such as annotation ambiguity and tokenisation variation—manifest in model behaviours like prediction loss, confidence, and representation shifts. These relationships are explored through an interactive visualisations that links data and model metrics, enabling example-based and integrated analysis. This supports language-specific explanations for observed performance differences. This chapter addresses RQ5.

- **Chapter 6 — Discussion and Contributions**
  Synthesises the key findings, highlighting the main sources of the Arabic–English performance gap and the diagnostic insights enabled by DeformAr. It formalises the novelty of the framework's integrated, cross-component evaluation approach and reflects on its limitations. Finally, it considers implications for future work, including how the framework could be adapted to other tasks (e.g., POS tagging), model types (e.g., LLMs), and languages beyond Arabic and English.

## 1.7 Thesis Contributions

This thesis makes four primary contributions: (1) A conceptual diagnostic framework for structured NER evaluation; (2) A modular toolkit for data exploration, extraction, fine-tuning, and diagnostic metric computation; (3) An interactive visual analytics dashboard built on top of the toolkit; and (4) Empirical insights into the Arabic–English performance gap in NER. It also results in an ongoing research outputs.

**Conceptual Framework** This thesis introduces a multi-layered evaluation framework for structured NER, which reconceptualises evaluation as a component-based diagnostic process linking data properties, model behaviours, and performance outcomes. The framework decomposes NER systems into model and data subcomponents, enabling both isolated and cross-component analysis of their interactions. These subcomponents are defined by segmenting the fine-tuning pipeline into discrete steps and examining the transformations applied to model and data components at each stage. Each is then quantified using system-level and token-level metrics. Conceptually, the approach is task- and language-agnostic, and can be extended to support additional subcomponents, metrics, or interaction types. This contribution is demonstrated in Chapter 3, addressing the conceptual layer of RQ0.

**Modular Toolkit** To support the evaluation framework, this thesis presents a modular open-source toolkit for data extraction, fine-tuning, and metric computation. The extraction module—available at deformar-extraction—serves as the backend of the framework, transforming system outputs into a structured, language-agnostic format. We extend existing tools such as seqeval by incorporating a novel span-level error typology and introducing a richer set of multi-level (system- and token-level) behavioural metrics that capture the behaviour of various subcomponents. To our knowledge, this is the first evaluation toolkit developed specifically for Arabic NER with built-in interpretability support. The toolkit is designed to be reusable and extensible, allowing researchers to adapt it for other tasks and languages. A full description of the toolkit's design, architecture, and role in the framework pipeline is provided in Chapter 3, addressing the technical layer of RQ0.

**DeformAR Dashboard** Built on top of the toolkit, the dashboard provides a user-facing interface for diagnostic evaluation and interpretability. It integrates structured, cross-component views—such as comparative breakdowns of data characteristics and model behaviours across languages—with interactive, token-level exploration. Users can inspect how factors like ambiguity, tokenisation, and tagging errors relate to prediction loss, confidence, and embedding structure. These coordinated views allow users to trace model outputs back to specific data properties, enabling both coarse-grained component-level comparisons and fine-grained analysis of individual examples at the token level, supporting targeted debugging and explanation. This contribution is demonstrated in Chapters 4 and 5, addressing RQs 1–5. The dashboard is available at deformar-dashboard.

**Empirical Insights** Applying the framework to Arabic and English NER systems reveals novel insights into how dataset properties interact with model behaviour—and how this interaction differs across languages. The analysis uncovers systematic annotation inconsistencies in Arabic—such as IOB2 misalignment and sentence-initial tagging errors—which lead to confidently incorrect predictions. Cross-component analysis identifies distinct error patterns for each language and links them to underlying data characteristics. It also shows that Arabic models often learn semantically coherent embeddings that misalign with the prediction layer. The analysis further quantifies how tokenisation reshapes the lexical structure of the data—affecting ambiguity and type distributions—and demonstrates how evaluation settings (e.g., scorer configurations) can mask or introduce errors, especially in Arabic, where annotation inconsistencies exist in both the predictions and the gold standard. These findings demonstrate how multi-level diagnostics uncover language-specific failure modes that standard evaluation methods often overlook.

**Ongoing Work** This research has produced two papers that will be made available on *arXiv*: one focuses on Arabic-specific findings from the framework, while the other is a system demonstration of the diagnostic toolkit. Several follow-up studies are also in progress, including: (i) an empirical analysis of annotation scheme inconsistencies in multilingual NER benchmarks; (ii) a comparative study on the effects of tokenisation strategies on lexical structure and NER performance; (iii) an investigation into how model scale and pretraining data influence cross-lingual NER behaviour; (iv) a study proposing attention-based token-level behavioural metrics as interpretable signals of model decision-making; and (v) an evaluation of how class imbalance across languages affects model performance and error distributions.

# 2 Background and Related Work

This chapter provides the foundational background necessary to understand the methodology and experiments presented in this thesis. It introduces the core components involved in building and analysing Named Entity Recognition (NER) systems based on transformer models. The chapter begins by introducing the architecture of transformer-based language models, key linguistic properties of Arabic, and an overview of the NER task.

The remainder of the chapter reviews relevant literature that examines how each of these components—model architecture, language characteristics, and task configuration—affects NER system performance. It provides an overview of the various factors that contribute to performance variation across languages, including architectural design bias (such as tokenisation), morphological mismatch, challenges arising from Arabic's linguistic complexity, and issues related to data quality and evaluation.

Following this, the chapter surveys existing evaluation and diagnostic approaches applied to NLP and NER systems, identifying common limitations—particularly in the context of NER and Arabic. It then introduces interpretability as a complementary approach for analysing system behaviour, reviewing key taxonomies and techniques used in the literature. The chapter concludes by motivating the need for a more integrated, multi-dimensional diagnostic approach introduced in Chapter 3.

## 2.1 Transformers

Transformer-based models have significantly improved performance in many Natural Language Processing (NLP) tasks (Peng et al., 2023; Brown et al., 2020; Patwardhan et al., 2023; Cai et al., 2023; Morsali et al., 2023; Devlin et al., 2019; Wang et al., 2018). The *Transformer* architecture was first introduced by Vaswani et al. (2017) as a general encoder-decoder model built on attention mechanisms, designed to generate contextual representations. Since its introduction, the transformer architecture has evolved into three main variants: encoder-decoder models such as T5 (Raffel et al., 2020) and BART (Lewis et al., 2020), decoder-only models such as GPT-2 (Radford et al., 2019), GPT-3 (Brown et al., 2020), and RWKV (Peng et al., 2023), and encoder-only models such as BERT (Devlin et al., 2019), RoBERTa (Liu et al., 2019), and ALBERT (Lan et al., 2019).

In this thesis, we focus on encoder-only transformer architectures, specifically BERT and its monolingual variants. BERT (Bidirectional Encoder Representations from Transformers) is an encoder-only transformer model introduced by Devlin et al. (2019). BERT models are available in multiple configurations, varying in both size and language coverage. While multilingual versions such as mBERT (Devlin et al., 2019) are trained on cross-lingual corpora, this work focuses on monolingual models to better isolate language-specific behaviour. Specifically, we use the original English BERT and the Arabic AraBERTv02 (Antoun et al., 2020), both pretrained on data drawn from their respective linguistic domains.

Each model is available in two architectural configurations. BERT-base consists of 12 layers, 768 hidden units, 12 attention heads, and approximately 110 million parameters (Devlin et al., 2019). BERT-large has 24 layers, 1024 hidden units, 16 attention heads, and around 340 million parameters (Devlin et al., 2019). Both versions are limited to a maximum input sequence length of 512 tokens. AraBERT follows the same architectural configurations (Antoun et al., 2020), and we use the *base model* throughout the thesis for both languages. These models also differ in certain language-specific aspects. For example, English BERT is available in both cased and uncased versions, with the uncased model normalizing words to lowercase during pretraining. In this work, we use the *bert-base-cased* variant, which retains capitalisation—a useful feature for NER tasks, where casing can provide important signals for identifying proper nouns (Devlin et al., 2019).

AraBERT, on the other hand, comes in multiple versions: v1, v2, v01, and v02 [1]. The v01 and v1 models were trained on approximately 23 GB of text (around 77 million sentences and 2 billion words), while v02 and v2 were trained on a much larger 77 GB corpus (approximately 8.6 billion words) [2]. In addition to differences in the pre-training data size, the preprocessing pipeline also varies across versions. Notably, the v2 model uses Farasa segmentation, but it has been shown to underperform on NER tasks compared to v02 (Antoun et al., 2020). As a result, we use the *AraBERTv02-base* model in all Arabic experiments. Table 2.1 summarizes the key configurations and pretraining data sizes for the BERT-base and AraBERTv02-base models used in this thesis.

Table 2.1: Architectural configurations for the selected base models used in this thesis. AraBERTv02 is used for Arabic, and BERT-base is used for English.

| **Feature** | **BERT-base** | **AraBERTv02-base** |
|---|---|---|
| Pretraining Data Size | 16GB | 77GB |
| Number of Layers | 12 | 12 |
| Hidden Size | 768 | 768 |
| Number of Attention Heads | 12 | 12 |
| Number of Parameters | 110M | 110M |

### 2.1.1 BERT Architecture

In this section, we describe the architecture of BERT originally introduced by Devlin et al. (2019), focusing on the core components shared across both English and Arabic monolingual variants. The discussion is organized into three stages: input components, contextualised encoder layers, and output representations.

**Input Components** BERT's input pipeline begins with a tokeniser and an embedding layer. The tokeniser converts raw input text into a sequence of integer token IDs using the WordPiece algorithm (discussed further in Section 2.1.3). Each token ID maps to a vocabulary entry defined by the tokeniser and is subsequently associated with a learned vector representation in the embedding layer. The embedding layer is similar to traditional word embedding layers but operates at the subword level. Figure 2.1 illustrates this process: raw text is first tokenised into subword units, then mapped to token IDs, which are used to retrieve dense vector representations from the embedding layer.

[1] Some of these versions were not publicly released at the time of the original publication but are available on the official GitHub repository: https://github.com/aub-mind/arabert

[2] There is an ongoing debate about how to compare the size of pretraining corpora, especially in multilingual settings (Arnett et al., 2024). Some papers use raw file size (in GB), while others compare the number of words or sentences. However, sentence-based comparisons can be misleading due to variation in sentence definition across languages. Similarly, word counts depend on morphological structure of the language. In this work, we use the total corpus size (in GB) as a proxy for scale, noting that AraBERTv02 was trained on approximately 8.6 billion words compared to BERT's 3.3 billion words. It is unlikely that the larger size is due to the encoding alone.

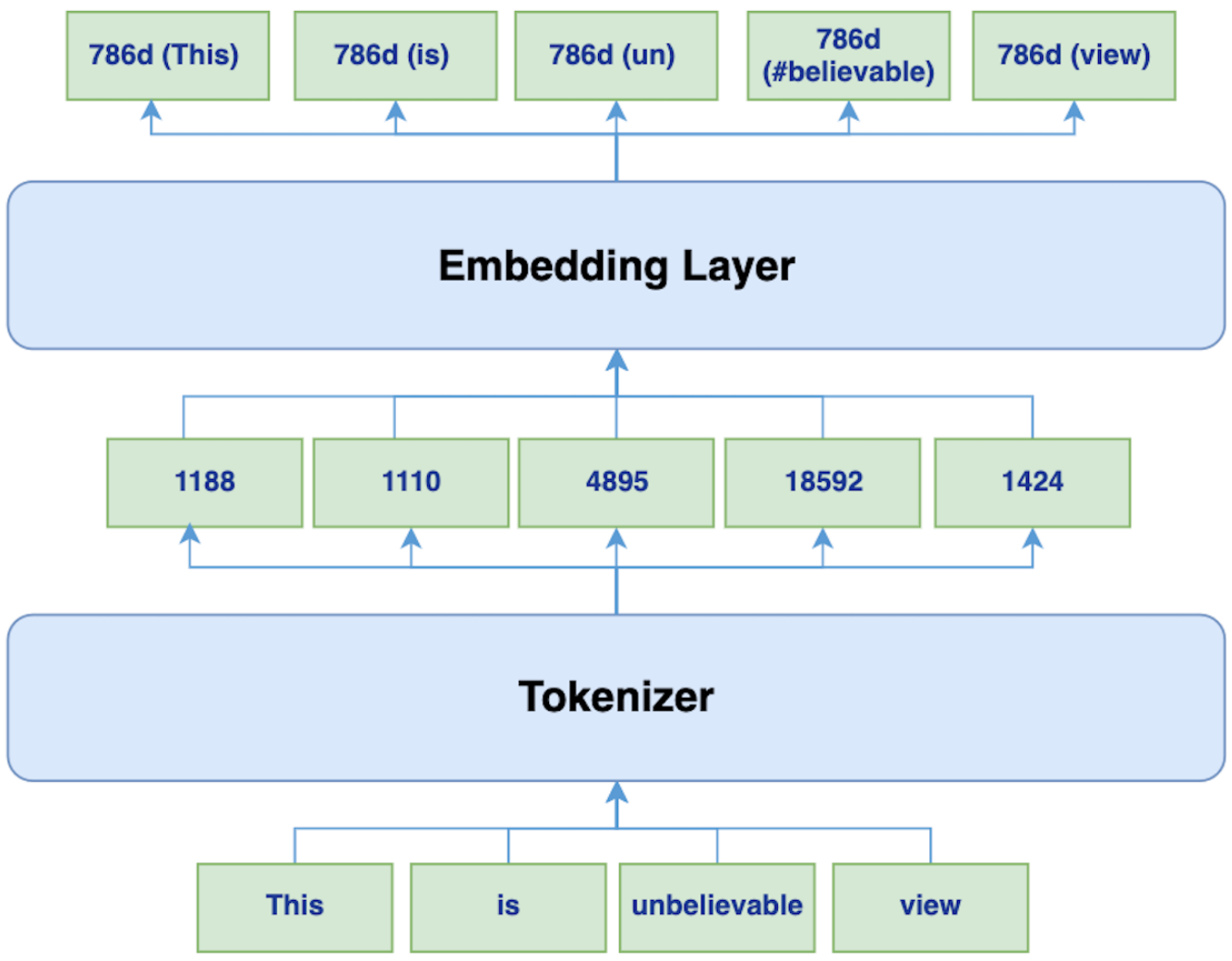

Figure 2.1: Overview of BERT's tokenisation and embedding process. Raw text is first tokenised into subword units and converted into token IDs, which are then mapped to vector embeddings in the model's input layer.

The embedding layer of BERT consists of three components: token embeddings, positional embeddings, and segment embeddings. These are summed to produce the final input representation for each token. BERT uses learned positional embeddings, which assign a unique embedding vector to each position in the input sequence. These embeddings encode the position of each token, enabling the model to capture word order, A special classification token ([CLS]) is added at the beginning of every sequence, and a separator token ([SEP]) is used to mark sentence boundaries or segments. Figure 2.2 illustrates the structure of the BERT embedding layer. The sum of these three components—token, position, and token type embeddings—forms the input representation that is passed into BERT's encoder layers.

**Encoder Layers** The encoder layers are responsible for transforming the input embeddings into contextual representations. In the base model, this is achieved through

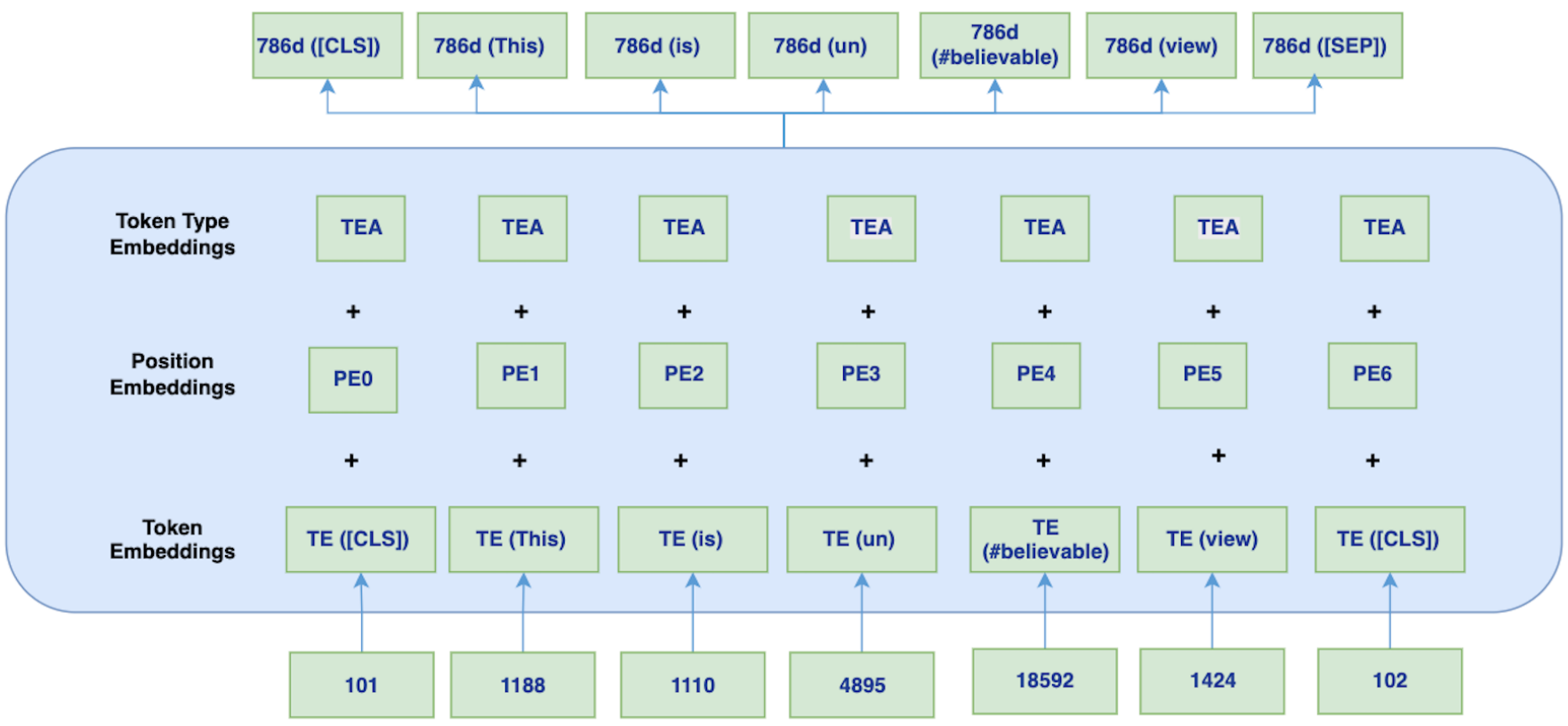

Figure 2.2: Overview of BERT's embedding layer. Each token ID is mapped to a token embedding, position embedding, and token type embedding. These vectors are summed and normalized before being passed to the encoder. Adapted from Devlin et al. (2019).

12 stacked transformer encoder layers. Throughout this thesis, we refer to layers closer to the input as *lower layers*, and layers closer to the output as *higher layers*; these terms are used interchangeably in the literature[3]. Each layer updates the representation of every token by attending to all other tokens in the sequence, enabling the model to build a contextual bidirectional understanding of context. These updates are computed using a combination of multi-head self-attention and feed-forward sublayers. A more detailed explanation of the encoder structure and self-attention mechanism is provided in Section 2.1.2. Figure 2.3 provides an overview of the encoder layers. Each token's representation is updated as it passes through the encoder layers, generating various outputs discussed next.

**Output Representations** BERT produces two main types of outputs: token-level hidden states and attention scores. The hidden states are computed at every encoder layer, with each layer's output serving as the input to the next. This layer-wise structure allows the model to accumulate and build a contextualized representation for each token, integrating information from the entire input sequence. In the base model, these hidden state vectors consist of 768 dimensions. The final hidden states—those produced by the last encoder layer—are typically used for downstream tasks (Devlin et al., 2019). In addition to hidden states, BERT also outputs attention scores at each layer and for each attention head. These scores indicate how much each token attends to every other token in the sequence. Both hidden states and attention maps are frequently used in interpretability analyses and are discussed further in Section 2.6.

[3] Multiple terminologies are used in the literature to refer to transformer layers, including "early"/"late", "bottom"/"top", and "shallow"/"deep". These differences often reflect whether the naming is based on diagram layout (e.g., input at the top), processing depth, or analogy to abstraction levels. In this thesis, we adopt the terms *lower* and *higher* layers, which reflect both position (closer to input or output), processing depth and abstraction.

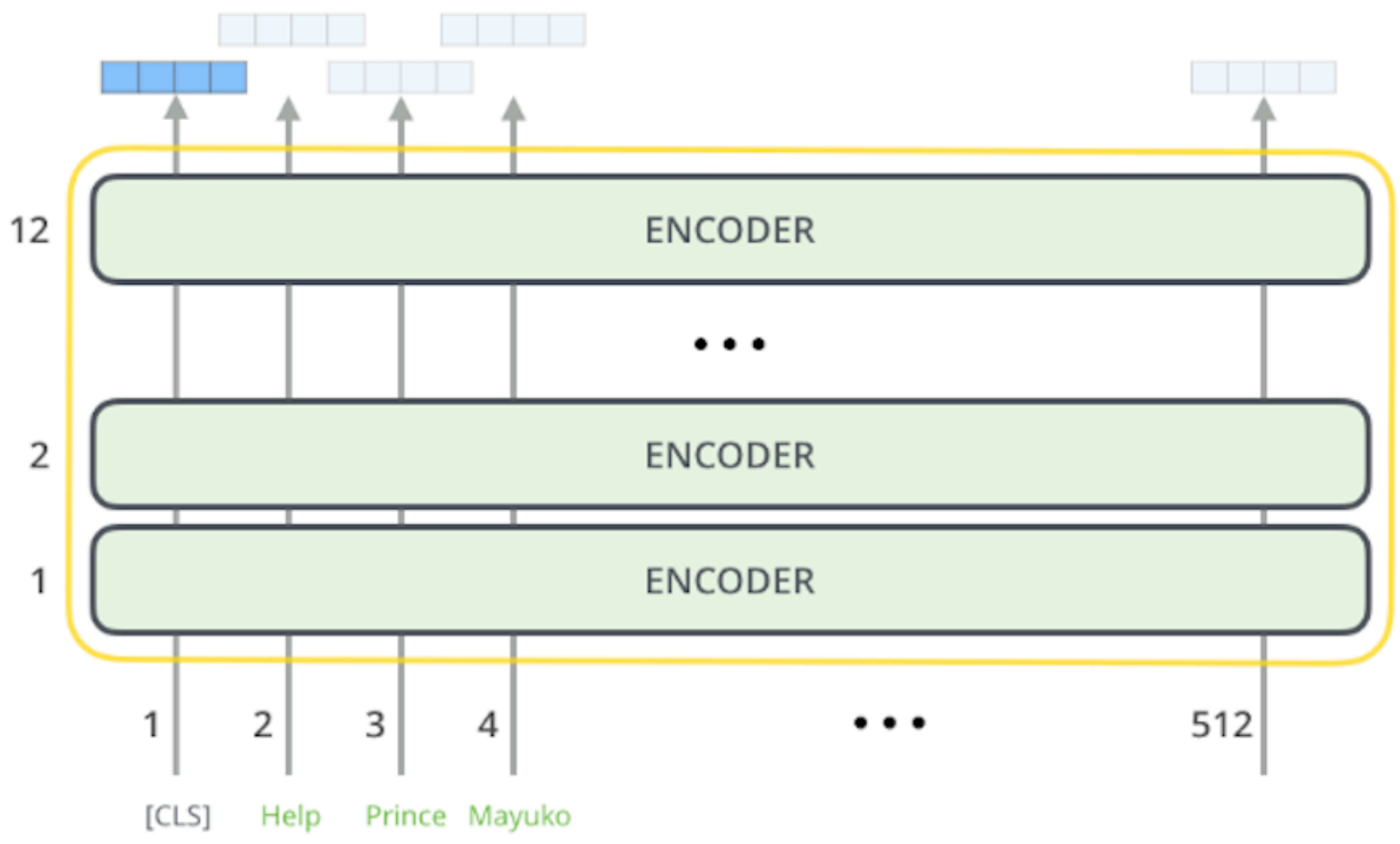

Figure 2.3: Simplified view of the BERT encoder stack. Each token is processed across 12 encoder layers (in the base model), allowing its representation to be contextualized with respect to all other tokens in the sequence. Reproduced from (Alammar, 2018).

### 2.1.2 Encoder Layers and the Attention Mechanism

Each encoder layer in BERT consists of two main components: a multi-head self-attention mechanism and a feed-forward network, followed by layer normalization. The self-attention mechanism is the key feature that enables transformer models to capture contextual relationships between tokens by allowing each token to attend to all others in the sequence. In contrast, the feed-forward layer applies a fixed transformation to the attended representations, projecting them into a new space before passing the result to the next layer in the stack. In the base model, each encoder layer contains 12 attention heads—hence the term multi-head attention—allowing the model to capture different types of relationships between tokens by running several independent attention heads in parallel. Each head has its own set of learned projection matrices, allowing it to attend to different parts of the input sequence and extract distinct patterns of contextual information (Vaswani et al., 2017). Figure 2.4 presents the components of a BERT encoder layer.

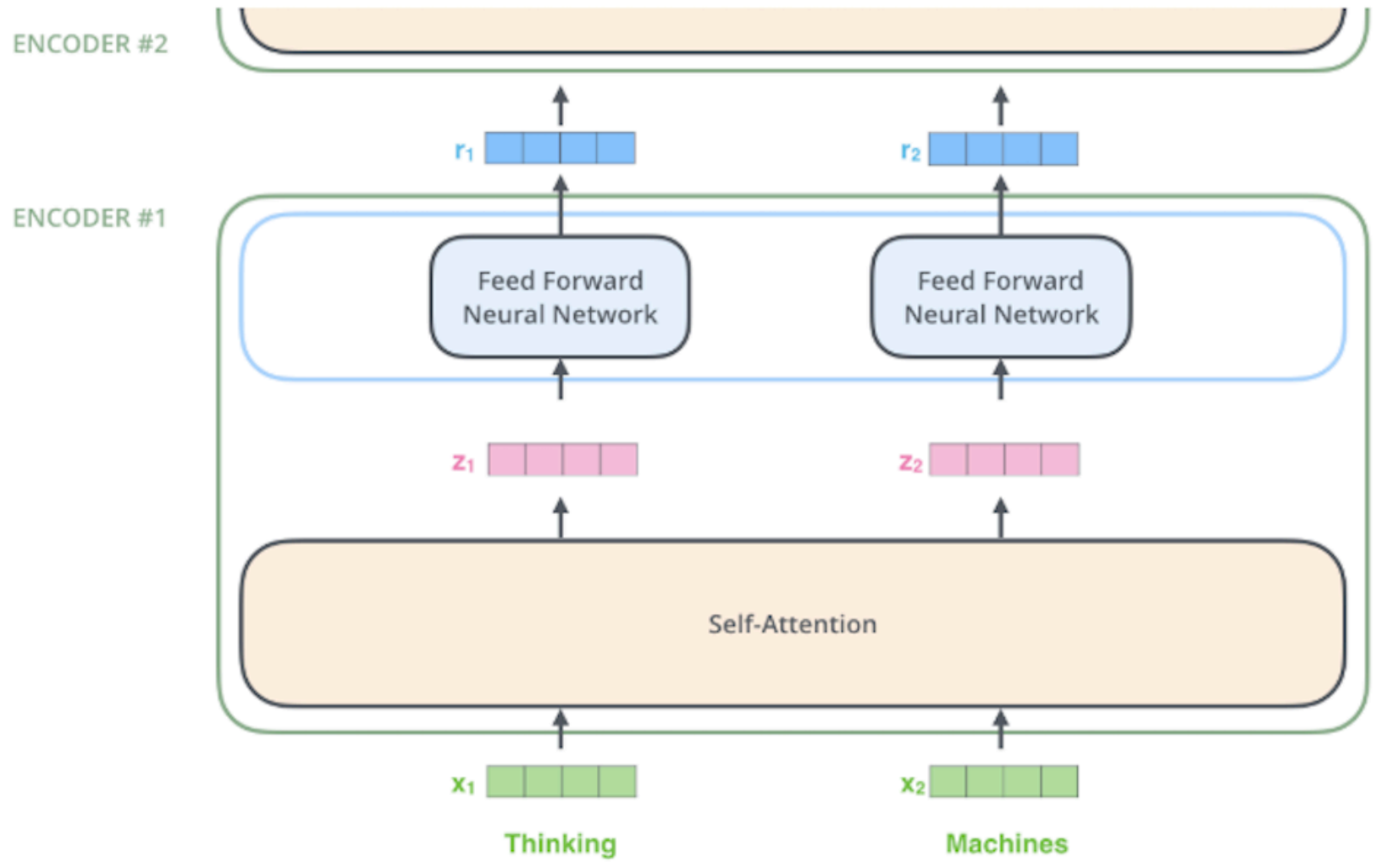

Figure 2.4: Overview of a BERT encoder block. Each token representation first passes through a multi-head self-attention mechanism, followed by a feed-forward neural network. This process is repeated across all encoder layers to build contextualized token representations. Reproduced from (Alammar, 2018).

Within each head, BERT uses a self-attention mechanism, where every token attends to all others in the input sequence, including itself. To enable this, the model computes three vectors for each token: a query ($\mathbf{q}$), a key ($\mathbf{k}$), and a value ($\mathbf{v}$). These vectors are generated by multiplying the token's embedding $\mathbf{x}$ by three separate, learned projection matrices: $\mathbf{W}_Q$, $\mathbf{W}_K$, and $\mathbf{W}_V$. These matrices are part of the model's parameters and are trained during both pretraining and fine-tuning. They learn to extract relevant contextual relationships through the attention mechanism. This is similar to the token embedding matrix, which is also a learned parameter: it maps each vocabulary item to a fixed vector representation prior to any contextualization. As shown in Figure 2.5, each token embedding is projected into its corresponding query, key, and value vectors, which are then used to compute attention. This projection allows the model to generate distinct representations for each head.

The actual attention computation is performed using the scaled dot-product attention formula (Vaswani et al., 2017):

$$\text{Attention}(\mathbf{Q}, \mathbf{K}, \mathbf{V}) = \text{softmax}\left(\frac{\mathbf{Q}\mathbf{K}^\top}{\sqrt{d_k}}\right)\mathbf{V} \tag{2.1}$$

The dot product $\mathbf{Q}\mathbf{K}^\top$ produces the raw scores, which are scaled by $\sqrt{d_k}$ and passed through a softmax function to produce *attention scores*. These scores determine how much each token attends to every other token in the input sequence, and are the values typically visualized in interpretability tools such as BERTviz (Vig, 2019). Figure 2.6 illustrates the process of calculating attention scores.

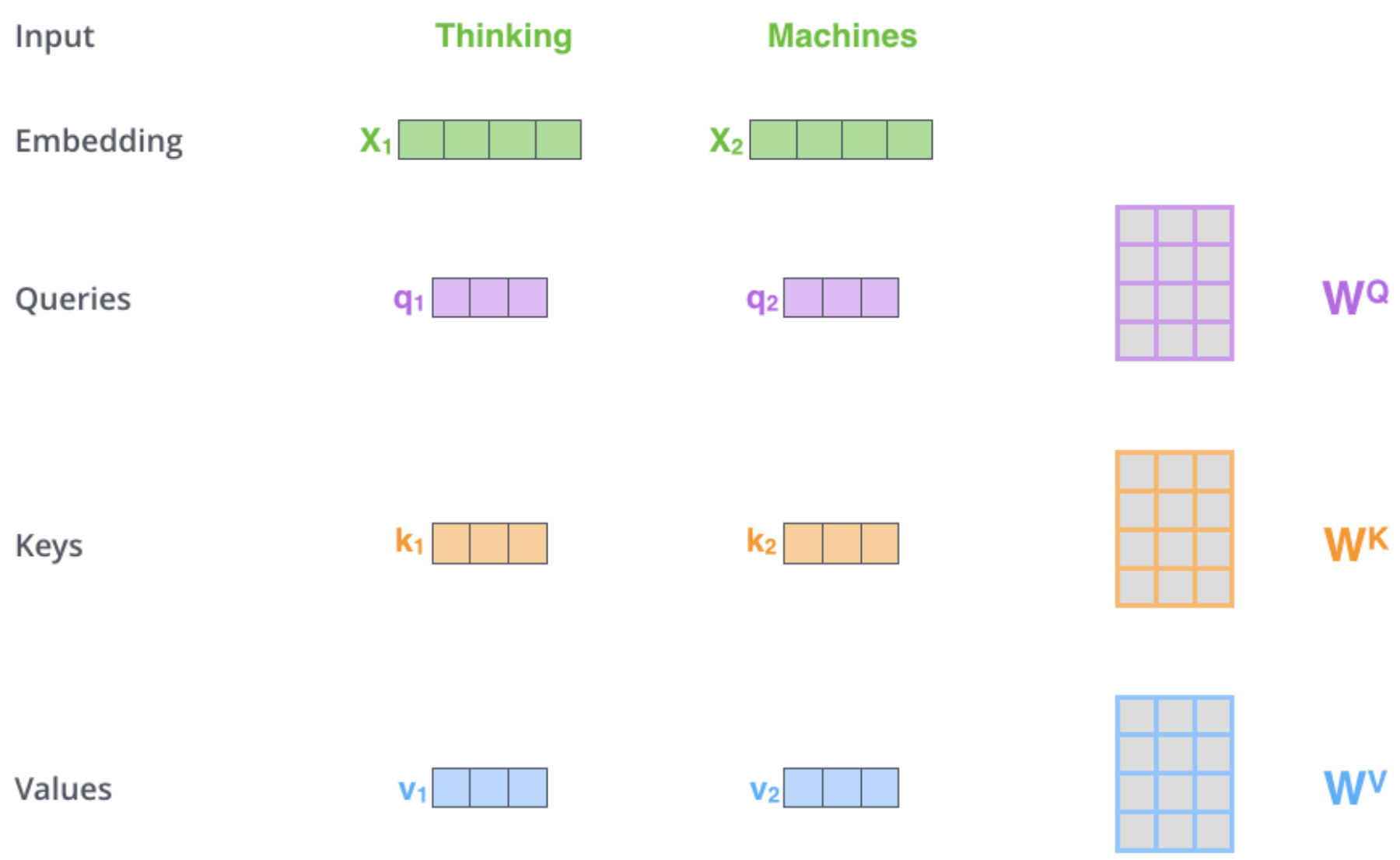

Figure 2.5: Projection of input embeddings into query, key, and value vectors in a single attention head. Each input embedding (e.g., for *Thinking* and *Machines*) is multiplied by learned weight matrices $\mathbf{W}_Q$, $\mathbf{W}_K$, and $\mathbf{W}_V$ to generate the corresponding query, key, and value vectors. Reproduced from (Alammar, 2018).

### 2.1.3 Tokenisation Approaches

Many pre-trained language models also rely on subword tokenisation methods to convert raw text into sequences of input tokens. Two widely adopted approaches are Byte Pair Encoding (BPE) and WordPiece (Mielke et al., 2021). While models like RoBERTa and GPT typically use BPE, BERT (Devlin et al., 2019) and its variants—including AraBERT (Antoun et al., 2020)—use WordPiece. Tokenisers convert input text into a numerical format that can be processed by the model. They consist of a predefined vocabulary of tokens and a mapping from each token to a unique integer ID. As noted earlier, both BERT and AraBERT rely on WordPiece tokenisation. WordPiece builds its vocabulary during pretraining by identifying frequently occurring character sequences in a large corpus and merging them into subword units. This process allows the tokeniser to split rare or complex words into smaller components while preserving common words as whole units. Each resulting token is assigned a unique ID that corresponds to an entry in the tokeniser's vocabulary. These tokens may represent full words or subwords, depending on the tokenisation's outcome. Both the BERT and AraBERT tokenisers include a special [UNK] token, which is used when a word or subword unit is not found in the vocabulary.

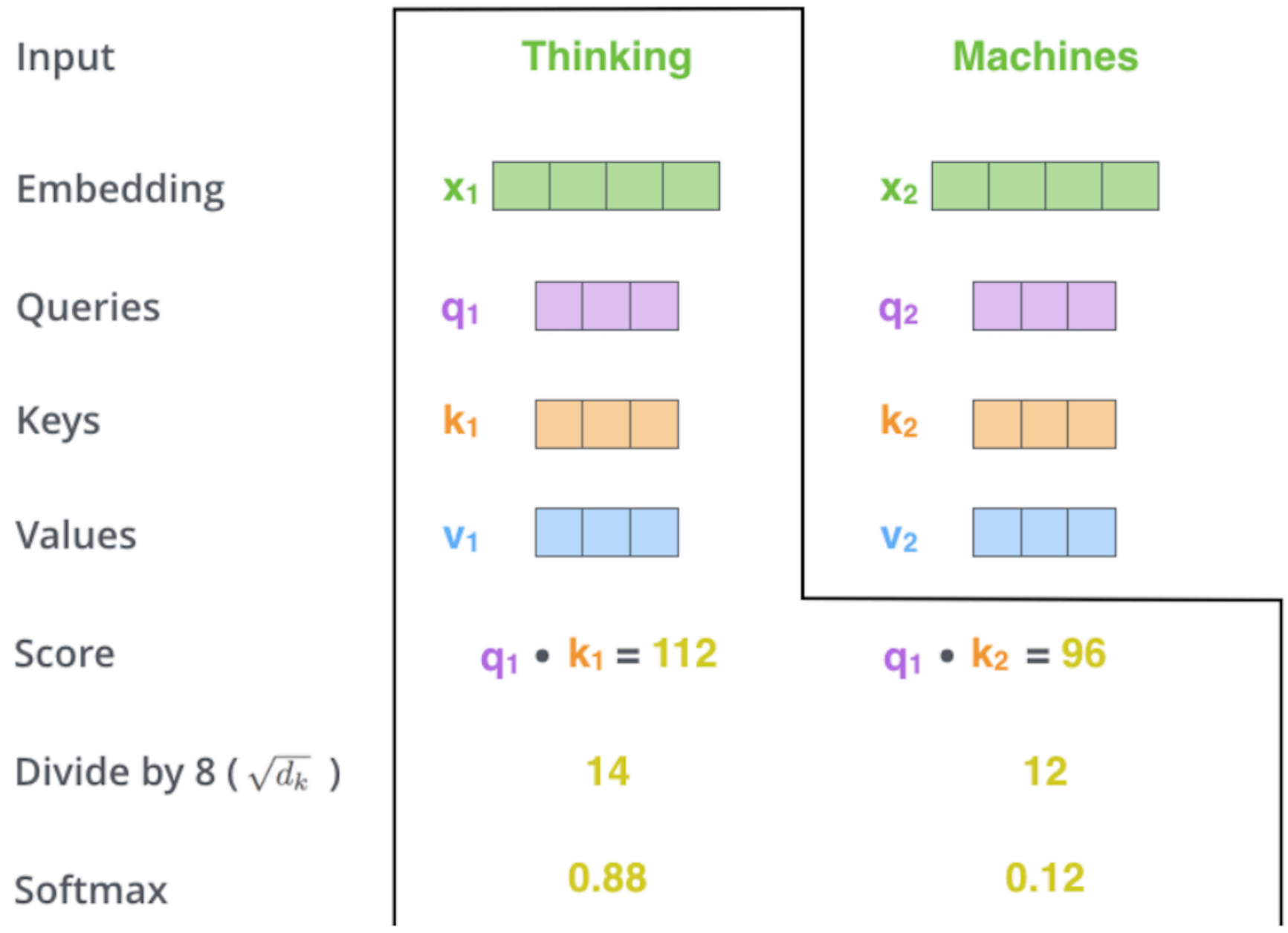

Figure 2.6: Illustration of attention score computation. For each token, attention scores are calculated by taking the dot product between its query vector and all key vectors, scaling the result by $\sqrt{d_k}$, and applying the softmax function. This produces a probability distribution over the input sequence, determining how much attention is paid to each token. Reproduced from (Alammar, 2018).

Although both BERT and AraBERT rely on WordPiece tokenisation, their tokenisation pipelines differ due to language-specific preprocessing. BERT applies tokenisation directly to raw English text (either cased or uncased), using a fixed vocabulary of 30,000 tokens learned during pretraining. AraBERT, on the other hand, performs additional normalization steps before tokenisation—such as removing diacritics, standardising Arabic characters, and inserting spaces around punctuation. Some versions of AraBERT, like v2, also apply segmentation, which splits words into stems and affixes to handle variation in surface forms caused by Arabic morphology (Antoun et al., 2020). However, as mentioned earlier, this thesis uses AraBERTv0.2, which applies the same normalization but does not perform segmentation. The vocabulary in the AraBERTv0.2 tokeniser consists of approximately 64,000 tokens. These differences in preprocessing and vocabulary directly affect how input is tokenised into subword units and can impact downstream performance (Antoun et al., 2020). Further details about the preprocessing pipeline are available in the official AraBERT GitHub repository[4].

While subword tokenisation methods like WordPiece have proven effective for many languages, they can struggle with languages that exhibit complex morphological and orthographic structures. Arabic, in particular, presents a unique set of linguistic challenges that complicate both tokenisation and downstream performance (Alkaoud and Syed, 2020). In the next section, we explore these characteristics in detail and discuss their implications for tokenisation.

## 2.2 Arabic Language

Arabic is a Semitic language and one of the world's most widely spoken languages, with over 300 million native speakers across the world. It can be divided into three main variants: Classical Arabic (CA), used in historical and religious texts; Modern Standard Arabic (MSA), the formal written and spoken standard across most Arab countries; and a wide range of regional dialects, which vary significantly in phonology, vocabulary, and syntax (Farghaly and Shaalan, 2009; Darwish et al., 2021).

This thesis focuses on Modern Standard Arabic (MSA). The remainder of this section provides a brief overview of its linguistic characteristics and the challenges they pose. We begin by examining how the concept of the "word" is defined in Arabic, then explore how these features have been discussed in the literature, and finally briefly consider how Arabic compares to more widely studied languages in NLP.

### 2.2.1 What Is a Word in Arabic?

According to علم النحو (the field of Arabic grammar), Arabic words are classified into three categories: اسم (nouns), which include names of entities, objects, and abstract concepts; فعل (verbs), which denote actions and tense; and حرف (particles), which may refer to a single letter of the alphabet or to what is known as حروف المعاني (particles of meaning), including prepositions (e.g., ”in”) and connectors (e.g., ”and”). For example, the particle و (and) may appear as a letter within a word—such as in واجب (obligatory)—or as a standalone coordinating conjunction, as in وأحمد كان هناك (and Ahmed was there), where it connects two clauses.[5]

[4] https://github.com/aub-mind/arabert

A key part of this system is the use of حركات (diacritics), which serve both phonological and grammatical functions. Diacritics help distinguish between otherwise identical words (e.g., the word كتب can appear as كَتَبَ ”he wrote” vs. كُتِبَ ”it was written”) and indicate voice, case, or tense. Yet in Modern Standard Arabic (MSA), these markings are often omitted in writing, introducing significant ambiguity (Farghaly and Shaalan, 2009; Darwish et al., 2021).

From a morphological perspective, Arabic morphology—known as صرف—relies on a templatic structure in which most words are derived from roots. For example, the root كتب can produce a variety of words such as كَتَبَ (he wrote), يَكْتُبُ (he writes), and كِتَاب (book). These changes reflect differences in meaning, tense, or grammatical role. Words may also change through the addition of prefixes or suffixes that carry syntactic information. For instance, forms like يَكْتُبُونَ (they write) or مُسْلِمُونَ (Muslims) include endings that mark number, gender, or person. This illustrates the interdependence between morphology and syntax in Arabic, where internal word structure conveys semantic roles, and affixes—especially suffixes—often reflect syntactic functions.

### 2.2.2 Arabic in the Literature

Based on the features discussed above, Arabic is widely described in the literature as a morphologically rich language, due to its combination of non-concatenative morphology (word formation through internal changes to the root structure), agglutination (attaching multiple morphemes—e.g., pronouns, prepositions, conjunctions—to a base word), and inflection ( modification of word forms to express grammatical features like number, gender, or case) (Farghaly and Shaalan, 2009; Darwish et al., 2021).

However, as noted by Gutierrez-Vasques et al. (2023), languages like Arabic do not fit into a single morphological category. Instead, Arabic exhibits features from multiple systems—including both agglutinative and fusional morphology—which makes it challenging to apply segmentation strategies, especially those originally developed for Indo-European languages (Gutierrez-Vasques et al., 2023).

[5] Both instances appear orthographically identical, but serve different grammatical roles.

These morphological features make Arabic structurally different from Indo-European languages like English, which typically exhibit more concatenative morphology, clearer word boundaries, and relatively shallow inflectional systems (Mielke et al., 2021). Arabic also allows more flexible word order and syntactic variation, whereas English generally follows a fixed subject-verb-object structure. As a result, models and tokenisation algorithms initially developed for English may struggle to accommodate the linguistic variability present in Arabic (Mielke et al., 2021; Alkaoud and Syed, 2020). Further discussion of how these characteristics affect model performance—particularly in the context of NER—is provided in Section 2.4.2.

## 2.3 Named Entity Recognition (NER)

NER is an NLP task that focuses on identifying and classifying words or phrases in unstructured text into predefined semantic categories such as *Person*, *Location*, and *Organization* (Li et al., 2022; Qu et al., 2023; Pakhale, 2023; Keraghel et al., 2024). NER is a component of information extraction and plays an important role in a range of NLP applications including question answering, text summarization, and machine translation (Li et al., 2022; Keraghel et al., 2024).

Several methods have been proposed for NER over the years. Early systems were rule-based, relying on hand-crafted linguistic rules and domain-specific gazetteers (Li et al., 2022; Keraghel et al., 2024). These were followed by statistical machine learning approaches such as Conditional Random Fields (CRFs) and Support Vector Machines (SVMs), which treated NER as a sequence labelling task using manually engineered features (Pakhale, 2023). To address the limitations of feature dependency, unsupervised and semi-supervised methods were introduced, though they generally achieved lower performance (Li et al., 2022). Deep learning approaches, particularly BiLSTM-CRF architectures, have since become the standard, enabling automatic feature extraction and improved generalisation across datasets (Li et al., 2022; Pakhale, 2023).

Transformer-based models, more recently, have significantly improved NER performance across a range of datasets and languages. BERT has become a widely used baseline for NER tasks due to its ability to produce contextualized token representations (Li et al., 2022; Pakhale, 2023; Keraghel et al., 2024). In this section, we focus on system architectures developed for BERT and its language-specific variants. These architectures typically use a contextual encoder such as BERT or AraBERT to generate token-level representations, followed by an output layer that maps these embeddings to entity tags (Qu et al., 2023). Most approaches adopt one of two configurations: (1) a direct mapping using a linear classification layer—often implemented as a Multi-Layer Perceptron (MLP) with a softmax activation, or (2) a sequence-aware classification layer such as CRF, which models label dependencies. CRF is more commonly used with deep learning approaches such as BiLSTM-CRF, while transformer-based models often rely on a simple linear classifier (Qu et al., 2023). In some cases, hybrid configurations are employed—for example, integrating a BiLSTM between the encoder and output layer (Keraghel et al., 2024).

Additionally, other work has proposed modifying the output layer by decomposing complex labels into components and learning shared embeddings for them, which has been shown to improve generalisation to rare or fine-grained labels (Kato et al., 2020). However, the most widely used architecture remains a direct encoder-to-classifier pipeline using either a linear layer or CRF.

### 2.3.1 NER Datasets

NER models are trained and evaluated on annotated corpora that define named entity spans and categories. For English, a wide range of datasets has been developed, covering various domains and annotation granularities. These include standard benchmarks such as CoNLL-2003 (Tjong Kim Sang and De Meulder, 2003) for news domains, OntoNotes 5.0 for multi-genre applications, and WNUT-2017 for social domains; GENIA and BC5CDR for biomedical text; and Few-NERD and WikiAnn for large-scale or multilingual evaluation (Keraghel et al., 2024). These resources differ in domain coverage, entity type granularity, and tagging schemes, offering a broad foundation for both general-purpose and domain-specific NER.

Arabic NER, by contrast, has fewer publicly available datasets (Qu et al., 2023). Key resources used in Arabic NER research include ANERcorp (Benajiba et al., 2007), a general-domain dataset based on MSA news and web texts; AQMAR (Mohit et al., 2012), which covers diverse domains via Wikipedia articles; and REFLEX and ACE, which are based on formal or broadcast news sources (Qu et al., 2023). Other resources such as CANERCorpus (Salah and Binti Zakaria, 2018) focus on religious texts. Finally, Wojood (Jarrar et al., 2022), while relatively large in scale and specifically designed for nested entity recognition, also supports flat annotation and covers both Modern Standard Arabic (MSA) and dialectal text across multiple domains. Table 2.2 summarises these the datasets for both languages in terms of size, language, and entity types.

In this thesis, we use the CoNLL-2003 (Tjong Kim Sang and De Meulder, 2003) dataset for English, which is an established benchmark and provides four entity types (Person, Location, Organization, Miscellaneous). For Arabic, we use ANERcorp (Benajiba et al., 2007), specifically the standardised CamelLab split released by CAMeL Lab (Obeid et al., 2020), which also provides four entity types, aligning with the CoNLL-2003 annotations. In this standardised version, several minor corrections were applied in collaboration with the dataset's original author (Obeid et al., 2020; AlDuwais et al., 2024). These modifications include:

- Correcting minor tag spelling errors.
- Converting middle and bold periods (e.g., '·', '•') to regular periods ('.').
- Removing blank Unicode characters (e.g., \u200F).
- Adding sentence boundaries after sequences of one or more periods.
- Splitting the dataset sequentially: 5/6 of the words go to training, and the remaining 1/6 to testing. The training split contains 125,102 words and the test split contains 25,008 words.

These two corpora are widely used standard benchmarks, developed around the same period and designed with comparable annotation schemes and entity categories, which makes them suitable for cross-lingual comparison.

### 2.3.2 NER Challenges

Despite substantial progress, NER continues to face several open challenges. One major difficulty is *entity ambiguity*, where named entities can overlap with common nouns or appear in different forms depending on context, complicating classification (Pakhale, 2023). Accurate *boundary detection* also remains problematic, particularly in the case of multi-word entities or fine-grained settings involving nested or overlapping structures (Li et al., 2022). NER systems frequently struggle with *domain shift*, as entity types and surface forms can differ significantly between general-purpose and specialised domains (Keraghel et al., 2024; Pakhale, 2023). Another ongoing challenge is handling *rare and emerging entities*, particularly those not seen during training (Li et al., 2022).

Table 2.2: Overview of NER datasets, including size, language, and number of entity types.

| Dataset | Size (Tokens) | Language | Entity Types |
|---|---|---|---|
| **Arabic Datasets** | | | |
| ANERcorp | 150,000 | MSA | 4 (PER, LOC, ORG, MISC) |
| AQMAR | 74,000 | MSA | Open-ended (core + custom MIS) |
| REFLEX | 200,000 | MSA | 4–6 (varies) |
| ACE (Arabic) | 300,000 | MSA | ∼7 (PER, ORG, GPE, LOC, etc.) |
| CANERCorpus | 258,000 | Classical Arabic | 14 (e.g., REL, EVENT, etc.) |
| Wojood | 550,000 | MSA + Dialect | 21 (nested + flat) |
| **English Datasets** | | | |
| CoNLL-2003 | 301,418 | English | 4 (PER, LOC, ORG, MISC) |
| OntoNotes 5.0 | ∼1,700,000 | English, Chinese, Arabic | 18 |
| WNUT-2017 | 65,124 (train) | English | 6-7 |
| GENIA | 500,000 | English | Biomedical (terms) |
| BC5CDR | 1,500 abstracts | English | 2 (CHEMICAL, DISEASE) |
| Few-NERD | 4,601,223 | English | 8 coarse, 66 fine-grained |
| WikiAnn (PAN-X) | Varies by language | Multilingual (176+ langs) | 3 (LOC, PER, ORG) |

In addition to modelling challenges, NER performance is constrained by issues in data quality and annotation. Datasets frequently contain *annotation errors* and inconsistencies, where the same entity may be labelled differently across or even within corpora (Li et al., 2022). Such inconsistencies contribute to the *glass ceiling effect*, where noise in labels places an upper bound on achievable model performance, regardless of modelling advances (Stanislawek et al., 2019; Reiss et al., 2020). This has been reported even in widely used benchmarks such as CoNLL-2003, which is estimated to contain annotation noise affecting up to 7.5% of its labels (Reiss et al., 2020). Similarly, a recent re-annotation of the ANERcorp dataset identified various tagging inconsistencies and labelling errors, leading to updates in 6.4% of its annotations and the release of CLEANANERCorp, a corrected version designed to improve label quality and evalua-tion reliability for Arabic NER (AlDuwais et al., 2024). Finally, the cost of high-quality annotation—particularly in low-resource languages or specialized domains—further limits the availability of reliable training data (Li et al., 2022).

**Arabic Specific Challenges** While English NER has been extensively studied, research on Arabic remains comparatively limited (Qu et al., 2023). Arabic NER presents unique challenges due to the language's morphological complexity and orthographic characteristics. These challenges are distinct from those encountered in English and other Indo-European languages, as highlighted by Qu et al. (2023). Key challenges include:

- **No Capitalisation:** Unlike English, Arabic lacks capitalisation as an orthographic feature, removing a strong cue for identifying proper nouns (e.g., person or location names).

- **Agglutinative Morphology:** Arabic's rich morphology allows prefixes, suffixes, and clitics (e.g., prepositions, conjunctions) to attach directly to named entities, altering their surface form. For example, the word والبحرين ("and Bahrain") combines the conjunction و ("and") with the location البحرين, complicating boundary detection.

- **Diacritic Absence and Ambiguity:** Modern Arabic texts typically omit short vowels (diacritics), leading to homography—a single written form may represent multiple meanings or grammatical roles depending on context. For instance, the undiacritized word كتب could mean "he wrote" (*kataba*) or "books" (*kutub*).

- **Spelling Variants and Normalization:** Arabic entities often have multiple transcriptional forms (e.g., القاهرة vs. القاهره for "Cairo"), requiring normalization or robust subword representations.

- **Limited Annotated Data:** Arabic NER datasets (e.g., ANERcorp, AQMAR) are smaller than their English equivalents, which limits model training and evaluation.

### 2.3.3 Annotation Schemes

Various names have been used in the literature to refer to the conventions to label entities, including *annotation schemes* (Tjong Kim Sang and De Meulder, 2003; Alshammari and Alanazi, 2021; Keraghel et al., 2024), *tagging schemes* (Zhao et al., 2022), and *segment representations* (Cho et al., 2013). In this thesis, we adopt the term *annotation scheme*. Annotation scheme refers to the convention used to label and segment named entities in a text (Tjong Kim Sang and De Meulder, 2003; Alammar, 2018; Keraghel et al., 2024; Shaalan, 2014). This includes both the label set (e.g., PER for persons, LOC for locations) and the positional encoding format used to denote the structure and span of entities within sequences (Zhao et al., 2022). Commonly used schemes include IO, IOB (or BIO), IOE, IOBES, BI, IE, and BIES (Alshammari and Alanazi, 2021; Keraghel et al., 2024).

Different annotation schemes differ primarily in how they specify entity boundaries and handle structural ambiguities. For example, while the IO scheme assigns only inside (I) and outside (O) labels, it cannot distinguish between consecutive entities of the same type—e.g., in the phrase “Apple and Microsoft”, both entities would be tagged I-ORG, making it unclear whether they represent one entity or two. Other schemes such as IOB, IOE, IOBES, and BIES address this limitation by explicitly marking the start (B), end (E), or both, and in some cases identifying single-token entities (S).

Some schemes also extend the tagging of non-entity words. Advanced formats like BI, IE, and BIES introduce specific tags for non-entity spans (e.g., B-O for the start of a non-entity segment, I-O for inside a non-entity span, and S-O for a single-token non-entity). For example, in the phrase “the quick fox”, where “quick fox” is not an entity, the tokens might be tagged B-O I-O. This provides richer structural context around entities. Table 2.3 summarises the description of each scheme, following the definitions and distinctions described in Alshammari and Alanazi (2021).

Although NER has been widely studied, the role of annotation schemes—particularly in Arabic—has received limited attention (Alshammari and Alanazi, 2021). A number of studies have shown that annotation schemes can significantly influence model performance, introduce varying levels of complexity, and interact differently with language-specific features. For example, Ratinov and Roth (2009) and Cho et al. (2013) show that the choice of encoding scheme can significantly impact system performance. In terms of complexity, Zhao et al. (2022) argue that annotation schemes introduce an additional layer of difficulty, as models are required to learn both the entity type and its position within the span (e.g., beginning, inside, or outside). They state that this positional encoding adds complexity to the model's learning process. Language-specific alignment has also been shown to affect performance. Chen et al. (2024) demonstrate that in Korean, aligning the annotation scheme with the structure of the language leads to performance improvements. For Arabic, Alshammari and Alanazi (2021) similarly show that the choice of annotation scheme influences model performance due to features like complex morphology and the absence of orthographic cues. They recommend choosing schemes that better align with the language's syntactic and morphological structure.

Table 2.3: Overview of common NER annotation schemes and their structural properties.

| Scheme | Description |
|---|---|
| IO | The simplest scheme: each token is assigned either an I (Inside) tag for named entities or an O (Outside) tag for non-entities. Cannot distinguish consecutive entities of the same type. |
| IOB (BIO) | Adds a B (Beginning) tag to mark the start of an entity, followed by I (Inside) for continuation and O (Outside) for non-entity tokens. Adopted in the CoNLL shared tasks. |
| IOE | Uses E (End) to indicate the final token in an entity span, in combination with I (Inside) and O (Outside). |
| IOBES | Extends IOB by explicitly marking single-token entities with S (Single) and multi-token endings with E (End), improving boundary precision. |
| BI | Similar to IOB but applies B-O to mark the beginning of non-entity segments, and I-O for internal non-entity tokens. |
| IE | Based on IOE, but also uses E-O and I-O to annotate the boundaries and interior of non-entity spans. |
| BIES | Combines the structure of IOBES for entities with explicit non-entity encoding: B-O, I-O, and S-O for non-entity segments. |

**Scheme Selection and Challenges** In this thesis, we focus on the IOB tagging scheme, which is one of the most commonly used annotation formats in NER (Zhao et al., 2022). This scheme uses three chunk tags: B-XXX to mark the beginning of an entity, I-XXX to indicate continuation, and O to label non-entity tokens. While widely adopted, IOB actually exists in two main variants: IOB1 and IOB2.

*IOB1*, introduced by Ramshaw and Marcus (1995), uses the B- prefix only when a named entity directly follows another entity of the same type. Otherwise, entities—even those at the start of a sentence or after an O—are labelled with I-XXX. This makes the scheme ambiguous, as it is not always clear whether an I-tag indicates a continuation or a new entity. *IOB2*, by contrast, requires that every entity start with a B- tag, regardless of its position. This removes ambiguity and is now the preferred format in most modern toolkits and datasets. The distinction between the two variants was formally clarified by Tjong Kim Sang and Veenstra (1999), who introduced the IOB1/IOB2 terminology. However, earlier literature often referred to IOB1 simply as IOB or BIO (Tjong Kim Sang and De Meulder, 2003). Table 2.4 provides a side-by-side example comparing the IOB1 and IOB2 tagging schemes. based on the following sentence: "Barack Obama and Michelle Obama visited Paris."

Table 2.4: Example comparison between IOB1 and IOB2 tagging schemes for the sentence *"Barack Obama and Michelle Obama visited Paris."*

| **Token** | **IOB1** | **IOB2** |
|---|---|---|
| Barack | I-PER | B-PER |
| Obama | I-PER | I-PER |
| and | O | O |
| Michelle | B-PER | B-PER |
| Obama | I-PER | I-PER |
| visited | O | O |
| Paris | B-LOC | B-LOC |
| . | O | O |

This terminology inconsistency also exists in recent work. For example, Alshammari and Alanazi (2021) refer to the scheme as IOB and note its association with CoNLL, but their implementation follows the IOB2 convention. Similarly, Keraghel et al. (2024) and Lignos and Kamyab (2020) refer to the scheme as BIO, implicitly assuming IOB2. To avoid confusion, we adopt the following naming convention: IOB1 refers to the original, adjacency-sensitive variant, and IOB2 refers to the version where every entity begins with a B- tag. While the original CoNLL-2003 dataset uses the IOB1 tagging scheme (Tjong Kim Sang and De Meulder, 2003), we rely on the version provided by Hugging Face's datasets library,[6] which adopts the IOB2 format for consistency and compatibility with modern NER tools. In contrast, ANERcorp is referred to as using IOB in both the CamelLab split (Obeid et al., 2020) and CLEANANERCorp version (AlDuwais et al., 2024). However, upon inspection, the actual tagging format in ANERcorp follows the IOB2 scheme.

Some papers do not specify whether they use IOB1 or IOB2 tagging, often referring to their format simply as BIO or IOB. This terminology can be ambiguous, as BIO is sometimes used generically and other times to refer specifically to IOB2. This has implications for evaluation, which are discussed in the following subsection.

### 2.3.4 Evaluation Approaches in NER

NER evaluation is commonly performed at two levels: token-level and entity-level. The token-level approach treats each token independently and computes classification metrics (e.g., precision, recall, F1) without regard to span boundaries. This is often referred to as a relaxed evaluation, and is supported by general-purpose libraries such as *sklearn* (Cuevas Villarmin et al., 2024; Pedregosa et al., 2018).

By contrast, entity-level evaluation focuses exclusively on full entity spans and ignores non-entity (O-labelled) tokens during scoring. A prediction is counted as a true positive only if both the entity type and the span boundaries match exactly. This span-level matching has become the standard approach in NER, particularly following the CoNLL shared tasks (Tjong Kim Sang and De Meulder, 2003; Lignos and Kamyab, 2020).

To illustrate the difference, consider the sentence: "Apple acquired the London office." Suppose the gold annotation labels "Apple" as an ORG and "London office" as a LOC. If a model predicts "Apple" as B-ORG and only "London" as B-LOC, token-level evaluation would count four out of five tokens as correct, including the partially correct "London" tag. However, entity-level evaluation considers only entity spans and ignores O-labelled tokens. In this case, the predicted span "London" does not match the full gold span "London office" and is therefore counted as a false positive. Table 2.5 summarises the difference.

Note that in entity-level evaluation, only entity spans are considered in the metric

[6] https://huggingface.co/datasets/conll2003

calculations—non-entity (O-labelled) tokens are ignored. This requires converting flat token-level predictions into structured spans, a process known as *span decoding*. The conventions that guide this process are defined by the annotation scheme (e.g., BIO, IOB2), which specifies how sequences of B/I/O tags correspond to named entity spans. This interpretation is typically handled by a scoring tool, or *scorer*. In the NER literature, a scorer refers to an evaluation tool or script—rather than a mathematical function—that calculates performance metrics such as precision, recall, and F1 based on entity-level predictions. Scorers determine how to interpret and count true positives, false positives, and false negatives from the predicted label sequences. They often include rule-based mechanisms to validate, interpret, or repair such sequences prior to evaluation.

Table 2.5: Example illustrating the difference between token-level and entity-level evaluation.

| **Token** | Apple | acquired | the | London | office | . |
|---|---|---|---|---|---|---|
| Gold Label | B-ORG | O | O | B-LOC | I-LOC | O |
| Predicted Label | B-ORG | O | O | B-LOC | O | O |
| Token-level Eval. | Correct | Correct | Correct | Correct | Incorrect | Correct |
| Entity-level Gold | ORG | – | – | LOC | LOC | – |
| Entity-level Pred. | ORG | – | – | LOC | – | – |
| Entity-level Eval. | Correct | – | – | Incorrect (partial span) | | – |

Recent work by Cuevas Villarmin et al. (2024) highlights a core issue in NER research: the lack of transparency and standardisation in evaluation methodologies—particularly regarding preprocessing steps such as annotation schemes and metric calculation. These inconsistencies can lead to statistically significant differences in reported performance, making direct comparisons between studies unreliable. A key source of this reproducibility issue lies in how evaluation metrics are calculated when label sequences do not conform to the expected annotation scheme (i.e., contain "improper label sequences"). An *improper label sequence*, as defined by Lignos and Kamyab (2020), is one that violates the structural constraints imposed by the annotation scheme—such as beginning an entity span with an `I-` tag instead of a `B-` under IOB2 conventions. These sequences frequently arise in models that use softmax classifiers over tokens, which make independent predictions and do not model relationships between labels. In contrast, CRF-based decoders enforce structural constraints by learning transition probabilities between tags, making them less prone to producing invalid sequences. As discussed in Section 2.3, the models used in this thesis employ dense output layers with softmax, making the treatment of such sequences a relevant consideration for evaluation.

NER scorers differ across multiple dimensions—not only in how they handle improper sequences, but also in the annotation schemes they support, the range of features they provide, and whether they are implemented as external tools or internal components. Lignos and Kamyab (2020) and Palen-Michel et al. (2021) show that different scorers adopt different strategies to handle such improper sequences, and that these choices can significantly affect evaluation scores. These strategies are known as *repair mechanisms*, and generally fall into two categories: *begin repair* (or CoNLL-style repair), which coerces an unexpected `I-` tag into a valid `B-`, and *strict/discard*, which excludes invalid spans entirely. Table 2.6 illustrates how each strategy handles an improper sequence. Note that the example follows IOB2: the improper sequence (bolded) violates IOB2 rules. Each repair mechanism results in different outcomes—the begin-repair strategy produces a corrected sequence that matches the gold standard (no penalty), while the strict strategy discards the span entirely, lowering the evaluation score.

Table 2.6: Comparison of repair strategies for handling an invalid IOB2 label sequence in NER evaluation.

| **Repair Strategy** | Apple | acquired | the | London | office | . |
|---|---|---|---|---|---|---|
| **Valid** | B-ORG | O | O | B-LOC | I-LOC | O |
| **Invalid** | **I-ORG** | O | O | **I-LOC** | **I-LOC** | O |
| **Begin Repair** | B-ORG | O | O | B-LOC | I-LOC | O |
| **Discard Repair** | O | O | O | O | O | O |

In addition to their differing repair strategies, scorers also vary in how transparent and documented their behaviour is. In practice, scorers fall into two categories: external tools and internal components embedded within NER toolkits. External scorers—such as `conlleval`, `seqeval`, and `SeqScore`—are typically open-source, well-documented, and standalone (Palen-Michel et al., 2021). Their methods for converting token-level predictions into spans, and for handling improper label sequences, are explicitly defined and can be inspected or configured. For example, `seqeval` offers both begin-repair and strict modes, while `SeqScore` allows fine-grained control over repair behaviour. By contrast, internal scorers—such as those in widely used frameworks like

NCRF++ or `Stanza`—are often bundled into larger toolkits and rarely document their decoding logic (Palen-Michel et al., 2021). Their handling of improper sequences is typically hard-coded and not user-configurable, making it difficult to assess how scores are calculated. This lack of transparency further complicates reproducibility and obscures fair comparisons between systems.

Below is a summary of the differences between the most common scorers and the types of repair mechanisms they apply:

- *conlleval*: The original Perl script used in the CoNLL-2003 shared task (Tjong Kim Sang and De Meulder, 2003). It applies a *begin repair* strategy, coercing unexpected `I-` tags to `B-`. It supports only IOB encoding and has limitations in numerical precision.

- *seqeval*: A Python-based reimplementation of `conlleval` that reproduces its results for BIO sequences (Nakayama, 2018). It adds strict mode support, extends compatibility to BIOES/IOBES schemes, and is widely used for its accessibility and reproducibility.

- *SeqScore*: A more advanced scorer offering explicit validation, encoding conversion, configurable repair strategies, and support for a wide range of chunking formats (Palen-Michel et al., 2021). It also outputs repaired sequences and supports aggregation across runs, enabling statistically robust evaluations.

- *NCRF++*: A widely used sequence labelling toolkit with an internal scorer that applies undocumented, stricter interpretations of label sequences (Lignos and Kamyab, 2020). Its behaviour leads to systematically higher F1 scores—especially for softmax-based models—due to silently discarding improper spans or treating them as O. This design complicates evaluation reproducibility.

These differences in scorer behaviour underscore the importance of explicitly reporting the evaluation tool, annotation scheme, and handling of improper sequences. Most studies that address these evaluation concerns—such as scorer inconsistency, annotation scheme compatibility, and improper label handling—focus almost exclusively on English, particularly the CoNLL-2003 dataset. To our knowledge, only one compa-rable analysis has systematically explored these issues in Arabic NER (Alshammari and Alanazi, 2021).

One of the key challenges in NER—particularly for Arabic—is the lack of transparency around evaluation practices, including which scorer is used and how improper label sequences are handled. As noted by Palen-Michel et al. (2021) and Lignos and Kamyab (2020), the absence of these details can significantly affect reproducibility. Some studies refer to using the "CoNLL script" without clarifying which variant or repair strategy was applied (Benajiba et al., 2007; Obeid et al., 2020; AlDuwais et al., 2024). Despite its widespread use, the *conlleval* script exists in multiple implementations—specifically a Perl version[7] and a Python reimplementation[8]. The Perl script supports only the IOB1 format and applies begin repair by default, while the Python version expects IOB2 and requires all input data to be converted accordingly.

Even the official studies that introduced BERT and AraBERT evaluate on NER tasks without explicitly reporting the annotation scheme, evaluation scorer, or repair strategy used (Devlin et al., 2019; Antoun et al., 2020).

In this thesis, we adopt both token-level and entity-level evaluation. Token-level evaluation is performed using *sklearn*, while entity-level evaluation is performed using the *seqeval* package in both strict and non-strict (default) modes. As noted by Lignos and Kamyab (2020), *seqeval* provides a faithful and transparent reimplementation of *conlleval*, with the added benefit of direct integration into Python-based workflows. They recommend it as one of the best available solutions for reproducible NER evaluation.

So far, this chapter has introduced the core components involved in building and evaluating NER systems—covering the background of transformer architectures, the linguistic characteristics of Arabic, the structure of the NER task, and the evaluation practices used to assess system performance. Together, these provide the necessary context for understanding how different factors might influence NER outcomes. In the next section, we turn to the literature to examine how these components have been studied as potential sources of performance variation.

[7] https://www.cnts.ua.ac.be/conll2000/chunking/conlleval.txt
[8] https://github.com/sighsmile/conlleval

## 2.4 Existing Factors Influencing NER Performance

As introduced in Section 1.2, NER performance is shaped by multiple interdependent factors—including model architecture, language characteristics, and training procedures. While these have each been studied in the literature, they are often investigated in isolation, with limited attention to how they interact within end-to-end NER systems. This section is reviewing how these factors have been examined in the literature. Where possible, we highlight how these components are treated in isolation versus in interaction, and identify gaps—especially in Arabic NER—that motivate the diagnostic approach developed in Chapter 3.

### 2.4.1 Model Architecture

The architecture of pretrained language models plays an important role in NER performance. Transformer models are fundamentally composed of two core components: tokenisation algorithms and model weights. These weights are learned during pretraining on large-scale text corpora, which are first tokenised using the model's own subword algorithm. This subsection examines architectural aspects related to both tokenisation and pretraining, with a focus on how they may contribute to the performance discrepancy between Arabic and English.

**Tokenisation Bias and Morphological Mismatch** A key architectural factor that may affect NER performance is tokenisation bias. Although subword tokenisation algorithms do not take into account linguistic information (Kiyono et al., 2023), it's important to recognise that these algorithms might still be designed to suit the language for which they were initially developed (Bender, 2011). According to Ruder et al. (2022), a prevalent bias in many models often relates to morphology. They argue that this bias stems from the fact that these models were often not designed with complex morphology in mind, arguably because they were initially developed primarily for English, a language with relatively limited morphology. This issue is prevalent in widely used tokenisation algorithms such as WordPiece and Byte Pair Encoding (BPE), which have become the de facto standard for Pre-trained Language Models (PLMs) (Mielke et al., 2021). Kiyono et al. (2023) highlighted several problems with these algorithms, particularly their shortcomings in representing linguistic phenomena such as lexical or semantic information. Consequently, the outcomes from these algorithms are not aligned with morphology, resulting in sub-optimal performance in downstream tasks (Bostrom and Durrett, 2020). The impact of such tokenisation on Arabic is illustrated by Shapiro and Duh (2018), who have demonstrated that byte-pair encoding performs sub-optimally for Arabic compared to other languages. Furthermore, experiments conducted by Alkaoud and Syed (2020) revealed that applying BERT tokeniser to Arabic often results in erroneous segments that misrepresent the intended meaning.

**Pretraining Data Size and Quality** In recent years, the size of pretraining data has been identified as a key factor influencing downstream performance in NLP tasks (Baevski et al., 2019; Radford et al., 2019; Liu et al., 2019; Devlin et al., 2019). Zhang et al. (2021) show that scaling data beyond billions of words improves performance in Natural Language Understanding (NLU), with models such as RoBERTa achieving further gains as data size increases (Liu et al., 2019). However, this trend appears to contradict the findings discussed in Chapter 1: AraBERT, despite being pretrained on a significantly larger corpus than BERT (77GB vs. 16GB), achieves lower performance on the NER task. This inconsistency—where larger pretraining data does not translate into better performance—raises the possibility that other factors, such as data quality or tokenisation alignment, may also play an important role.

AraBERT's pre-training data, for instance, encompasses a diverse set of sources, including Arabic Wikipedia dumps, manually scraped web articles and a wide range of news sources from various Arab regions (Antoun et al., 2020). BERT, on the other hand, is predominantly trained on BookCorpus and English Wikipedia (Devlin et al., 2019). Although these details offer information about the models' data sources, directly assessing the quality based on this information alone is challenging. It is also important to note that data quality can be influenced by both the content quality and how this data is represented to the model. We argue that the misalignment of tokenisation algorithms with the morphological characteristics of Arabic can affect data representation quality, hence impacting downstream performance. This perspective aligns with the findings of Bostrom and Durrett (2020), who highlighted the impact of input encoding techniques on the effectiveness of pretrained models in downstream tasks. Further supporting this view, Rust et al. (2021) demonstrated that the effectiveness of NLP models depends not only on the size of pre-training data but also on the choice of high-quality, language-specific tokenisers. Their study showed that replacing the original multilingual tokeniser in mBERT with a specialized monolingual tokeniser led to improvements in downstream performance for almost every task and language, including Arabic NER.

**Interdependence and Research Gaps** So far, we have examined two architectural factors that may help explain the performance gap between Arabic and English in NER systems: tokenisation and pretraining data. Each has been studied individually—for example, Rust et al. (2021) examine the effect of language-specific tokenisers, while other work focuses on pretraining data size, language variant, and task type (Inoue et al., 2021). However, these studies often treat the discussed components in isolation and do not explore how they interact. In particular, tokenisation not only affects how input is segmented, but also how pretraining data is represented to the model. If the tokeniser is misaligned with the linguistic structure of the language, it may distort the input during training, which can degrade the quality of learned representations.

To our knowledge, no study has jointly examined the combined effect of tokenisation and pretraining data quality—especially in the context of morphologically rich languages like Arabic. Moreover, most evaluations rely on surface-level performance metrics such as F1 score, without deeper investigation into how these factors affect internal model behaviour. This highlights an existing gap in the literature and motivates the need for more detailed diagnostic approaches.

In the following section, we examine language-specific factors—particularly those related to Arabic morphology and orthography—that may further contribute to performance differences. We also explore how these linguistic properties interact with architectural components.

### 2.4.2 Linguistic Complexity

As discussed in Section 2.2, Arabic presents a number of linguistic characteristics that distinguish it from languages like English. These include its non-concatenative and ag-glutinative morphology, lack of capitalisation, omission of diacritics in most written contexts, and relatively free word order (Farghaly and Shaalan, 2009; Darwish et al., 2021). These are well-established features of Arabic and are frequently cited as reasons why Arabic presents challenges for NER and NLP more broadly (Farghaly and Shaalan, 2009; Shaalan et al., 2018; Darwish et al., 2021). However, while these characteristics are widely acknowledged, their connection to system performance is rarely examined in depth. In particular, few studies have investigated how they interact with architectural components such as tokenisation or pretraining, or how they contribute to downstream outcomes in a quantifiable and diagnostic way.

Morphologically, Arabic can encode complex syntactic and semantic information within a single word. A root-and-pattern system combined with affixation allows for wide variation in word forms, often compressing what would be full phrases or clauses in English into a single token (Diab, 2008; Darwish et al., 2021). For example, the word سالتمونيها (*sa'altumūnīhā*)—which translates to "you asked me it"—encodes tense, subject (you plural), verb, indirect object ("me"), and direct object ("it") all in a single word (Darwish et al., 2021). This morphological density results in a large number of unique word forms, contributing to data sparsity and vocabulary diversity, which makes learning and generalisation more difficult for NER models (Shaalan, 2014; Diab, 2008).

This complexity also poses challenges for tokenisation algorithms such as WordPiece and BPE (Radford et al., 2019), which often rely on consistent token boundaries. As these techniques were not designed with Arabic's root-and-pattern system in mind, they frequently produce malformed or fragmented subwords that do not align with the language's internal structure (Alkaoud and Syed, 2020; Mielke et al., 2021). These issues can distort entity boundaries and degrade NER performance (Shaalan et al., 2018; Rust et al., 2021).

Orthographic and morphological characteristics in Arabic also introduce several layers of ambiguity. At the orthographic level, the absence of capitalisation removes a major cue used in languages like English, where uppercase letters help signal named entities (Diab, 2008; Farghaly and Shaalan, 2009). Additionally, the frequent omission of diacritics introduces lexical ambiguity: many words share the same consonantal structure but differ in meaning depending on short vowels, which are not typically written (Darwish et al., 2021). These issues complicate word sense disambiguation and boundary detection (Shaalan, 2014; Shaalan et al., 2018; Farghaly and Shaalan, 2009). It also complicates preprocessing pipelines, including tasks like sentence segmentation, which is a required step in NER dataset preparation.

These challenges illustrate that linguistic features such as morphology, orthography, and syntax influence system performance not in isolation, but through their interaction with architectural assumptions and fine-tuning dynamics. However, as with architectural factors, the impact of these interactions is rarely examined in depth. In most cases, evaluations rely on aggregate metrics such as F1 score, which capture overall performance but obscure where and how systems struggle with specific linguistic constructions. This highlights a broader gap in the literature: the limited use of diagnostic methods to trace performance back to linguistic sources of difficulty and understand their interactions with other system components.

### 2.4.3 Training Procedures

Beyond architecture and linguistic complexity, the training procedure plays a critical role in shaping NER performance. This includes the choice of system architecture as well as how datasets are constructed, annotated, and prepared for fine-tuning. As discussed in Section 2.4.2, Arabic's morphological and orthographic features introduce challenges for standard preprocessing steps such as sentence segmentation, which is often a prerequisite for NER dataset construction. These early pipeline decisions have direct implications for both learning and evaluation.

Another important factor in dataset construction is the choice of annotation scheme. Prior work has explored how different schemes—such as BIO, BILOU, or more fine-grained tagsets—affect performance (Ratinov and Roth, 2009; Lee et al., 2023b; Zhu et al., 2020; Tkachenko et al., 2013; Demir and Özgür, 2014; Cho et al., 2013). These studies show that annotation complexity and granularity impact performance, but also that the effectiveness of a scheme depends on how well it aligns with the underlying linguistic structure of the target language (Alshammari and Alanazi, 2021; Chen et al., 2024; Mozharova and Loukachevitch, 2016). However, most of this research compares schemes through F1 scores—assuming that higher scores imply better alignment or reduced complexity—and rarely investigates why certain schemes perform better, or how their alignment with language structure affects system behaviour in practice.

Furthermore, annotation consistency and label noise is another important aspect studied in the literature. Mislabelled or inconsistently applied tags are a known issue in many NER datasets and can significantly limit achievable performance. Reiss et al. (2020) estimate that the CoNLL-2003 test set contains at least 7.5% annotation errors, while Stanislawek et al. (2019) describe a "glass ceiling" effect—where performance plateaus not due to model limitations, but due to inconsistencies in the data. These issues have received limited attention in Arabic benchmarks. Furthermore, inconsistencies may not stem from annotation error—they can also occur due to language complexity and tokenisation impact on the lexical structure, as demonstrated in Chapter 4.

This further supports the argument that system components (model, data, and language) are interconnected, and require an evaluation approach that accounts for these interactions—a gap that this thesis is trying to fill. The next section provides a more detailed discussion of existing evaluation and diagnostic approaches.

## 2.5 Existing Evaluation and Diagnostic Approaches

Existing research on NER evaluation and diagnostics tends to focus on specific components in isolation, rather than examining how language, data, and architectural choices interact across the full system. Broadly, prior studies fall into three categories: data-centric, model-centric, and methodology-driven approaches.

**Data-Centric Approaches** Many NER studies investigate how annotation schemes, tagging formats, or data quality affect NER outcomes. As discussed in Section 2.4.3, these studies typically compare scheme designs (e.g., BIO vs. BILOU) or address issues like label noise and inconsistency (Ratinov and Roth, 2009; Alshammari and Alanazi, 2021; Reiss et al., 2020; Stanislawek et al., 2019). However, this work is usually evaluated using standard metrics, and rarely investigates how such issues interact with other components in the NER pipeline.

**Model-Centric Approaches** Transformer-based NER research typically focuses on examining architectural choices, including tokenisation strategies, pretraining data, and language variants. These have already been covered in Section 2.4.1, with particular attention to tokenisation biases in morphologically rich languages (Rust et al., 2021; Kiyono et al., 2023; Mielke et al., 2021). While these analyses offer valuable insights, they often isolate individual factors and rely primarily on surface-level performance metrics such as F1—overlooking the compounding effects of architectural and linguistic interactions.

**Methodology-Driven Approaches** A growing body of research proposes tools for better understanding model behaviour. These studies are typically methodology-led and evaluate systems using influence-based data attribution (Jain et al., 2022), explanation-guided supervision through entity triggers (Lin et al., 2020; Lee et al., 2023a), or instance-level interpretability via saliency and adversarial perturbations (Wallace et al., 2019). These approaches rely on model outputs—such as prediction scores, gradients, or saliency maps—to interpret individual predictions. While valuable for diagnostic purposes, these methods often focus on a single dimension of evaluation. As noted by Ruder et al. (2022), interpretability work tends to concentrate on English, while multilingual studies typically focus on performance metrics like F1, with little integration between them.

This methodological gap is one that this thesis aims to address by introducing a comprehensive diagnostic approach that combines standard evaluation metrics with interpretability and system-level analysis—integrated through visual analytics. Our proposed framework is designed to analyse how system components interact, identifying where and how failures occur and clarifying the role each component plays in shaping those outcomes. In the following section, we review key concepts in interpretability, and outline how they are used to analyse model behaviour—particularly in transformer-based architectures.

## 2.6 Interpretability

Despite the strong performance of transformer-based models—which have achieved state-of-the-art results across a wide range of NLP tasks—these models remain largely opaque (Sun et al., 2021; Zini and Awad, 2023; Singh et al., 2024). It is often unclear how they arrive at their predictions, what internal representations they form, or how specific inputs influence their outputs. Performance metrics like F1 scores offer limited insight into why models fail, where they struggle, or how particular inputs shape their predictions. The lack of interpretability techniques specifically designed for sequence labelling tasks like NER (Sun et al., 2021)—particularly in morphologically rich and under-explored languages such as Arabic—highlights the need for tools that can offer insight into model behaviour, guide debugging, and support fair cross-lingual evaluation (Lertvittayakumjorn and Toni, 2021).

While interpretability has become a growing area of research in NLP, the terminology itself remains vaguely defined (Sun et al., 2021; Singh et al., 2024). The terms interpretability and explainability are often used interchangeably (Zini and Awad, 2023), yet their meanings are subjective and highly dependent on context (Singh et al., 2024). In this thesis, we adopt the following distinction: interpretability refers to the ability to understand how a model processes inputs and makes decisions, by analysing internal components such as hidden representations, attention patterns, or learned parameters (Saha et al., 2022; Zini and Awad, 2023; Singh et al., 2024). Explainability, on the other hand, refers more broadly to the ability to explain a model's predictions—regardless of how this explanation is produced.

To clarify the difference: if we aim to understand what the model predicted and why, we refer to this as explainability; if we aim to understand how the model internally arrived at that prediction, we refer to this as interpretability. We adopt this distinction throughout the thesis and primarily rely on the term interpretability, as it more accurately reflects the focus of our analysis. In the remainder of this section, we explore the main taxonomies proposed in the literature, outline commonly used interpretability techniques, and highlights current limitations especially relevant for Arabic NER.

### 2.6.1 Interpretability Taxonomies

Several taxonomies have been used to organize interpretability methods in NLP, each highlighting different aspects of model analysis. One taxonomy proposed by Sun et al. (2021) organizes interpretability methods based on the data split used to generate explanations. It defines three categories: (1) training-based methods, which trace model behaviour back to influential training examples; (2) test-based methods, which identify the parts of a test input that most affect the model's output; and (3) hybrid methods, which link training data to test predictions by examining how specific examples influence particular attention patterns or activation signals. Another taxonomy, presented by Zini and Awad (2023), focuses on the level of analysis, distinguishing methods that operate on the input level (e.g., word embeddings), processing level (e.g., internal representations, attention heads), or output level (e.g., prediction decisions).

Another taxonomy is the one proposed by Zhao et al. (2024), which categorizes interpretability techniques by both scope and method type. They distinguish between local explanations (targeting individual predictions) and global explanations (analysing model behaviour across many inputs), and further classify methods into categories such as feature attribution, attention-based explanations, example-based reasoning, and concept-based methods. Lastly, Ferrando et al. (2024) propose a component-oriented taxonomy, organizing interpretability research based on the specific model component being analysed—such as neurons, layers, or attention heads—often with a focus on low-level behaviour, probing, and representation analysis.

While these taxonomies differ in focus—ranging from data sources and levels of analysis to explanation scope and model components—they share several common distinctions. One common theme is the contrast between intrinsic interpretability, where models are designed to be self-explanatory, and post-hoc interpretability, which involves analysing model behaviour after training. Another shared distinction is the distinction between model-specific methods, such as attention-based explanations tailored for transformer architectures, and model-agnostic approaches like LIME, which can be applied across a broad range of models (Zini and Awad, 2023; Sun et al., 2021).

### 2.6.2 Interpretability Techniques

To provide an overview of existing interpretability techniques, we adopt the level-of-analysis taxonomy proposed by Zini and Awad (2023). We adapt this framework in two ways to better suit our focus on transformer-based models. First, we merge the input and processing levels into a single category, since transformer architectures generate contextual embeddings that combine input and intermediate representations. Second, we incorporate an additional category for example-based explanations, which are not explicitly included in the original taxonomy but are widely used in interpretability research. We also include two further distinctions commonly used in the literature: the scope of explanation (local vs. global), and whether the method is model-specific or model-agnostic.

At the input level, global explanations include methods such as classifier-based probing, which investigate whether specific linguistic features—such as part-of-speech tags or syntactic roles—are encoded in the model's representations (Ferrando et al., 2024; Zini and Awad, 2023). Local methods include attention-based explanations, which highlight the tokens that receive high attention weights during prediction (Sun et al., 2021; Braşoveanu and Andonie, 2020). These methods typically rely on visualization to assess what kinds of information are captured across attention heads and layers (Brath et al., 2023; Braşoveanu and Andonie, 2020).

At the output level, feature attribution and saliency-based techniques assign importance scores to individual input tokens with respect to the model's prediction (Sun et al., 2021; Ferrando et al., 2024). These methods offer token-level explanations by estimating how much each part of the input contributes to the output. Attention is also used at this level; however, its reliability as an explanation is debated (Sun et al., 2021; Zini and Awad, 2023; Zhao et al., 2024). For instance, Jain and Wallace (2019) argue that attention weights can be manipulated without affecting the model's output, suggesting they are not a faithful explanation. In contrast, Wiegreffe and Pinter (2019) argue that attention can be more interpretable than general model weights, but its ability to provide full transparency or meaningful explanations remains limited and context-dependent.

Finally, example-based methods explain predictions by referencing influential or similar instances from the training data. These include techniques such as influence functions, which trace predictions back to impactful training samples, and counterfactual explanations, which identify minimal changes to the input that would change the model's output. Together, these three categories provide a practical structure for organizing interpretability methods in transformer-based NLP models (Sun et al., 2021; Zini and Awad, 2023; Zhao et al., 2024; Ferrando et al., 2024).

Recent work has highlighted the growing role of visual analytics in interpretability research for NLP. These approaches combine interactive visualization with human reasoning to support tasks such as model understanding, bias detection, and debugging. Unlike static explanation methods—which generate fixed outputs such as saliency scores or token highlights—visual analytics enables users to engage dynamically with model representations, allowing for iterative exploration, hypothesis testing, and refinement (Brath et al., 2023).

Brath et al. (2023) provide a broad survey of visualization in NLP, focusing particularly on transformer-based models. They propose a framework in which visualization supports various stages of model development—from inspecting training data, to understanding internal representations, to analysing model outputs. The paper highlights a range of visualization techniques—including saliency maps, attention graphs, bar charts, dimensionality reduction plots, and heatmaps—and notes that visualization can serve both expert model developers and domain specialists with limited technical background. Tools like LIT and Seq2Seq-Vis illustrate how interactive interfaces can be used to investigate and compare model behaviour across different tasks. Importantly, the paper highlights the value of layer-wise analysis, which examines how linguistic features and semantic information are distributed across different layers of a transformer. This type of analysis is especially useful for identifying where and how specific information—such as part-of-speech or named entities—is encoded in the model.

Lertvittayakumjorn and Toni (2021) introduce the concept of Explanation-Based Human Debugging (EBHD), where humans use model explanations to identify and correct errors in NLP systems. They define a three-step workflow: (1) provide explanations to the user, (2) collect user feedback, and (3) update the model. EBHD is particularly useful when training data introduces artifacts or biases that cause unreliable model behaviour. The paper categorizes EBHD techniques by explanation scope (local vs. global), method (e.g., saliency or example-based), and the form of user interaction.

Braşoveanu and Andonie (2020) offer a focused review of tools and visualization strategies designed for transformer models. They describe how attention weights, token-level importance, and layer-wise activations can be visualized to inspect model behaviour. Tools like BertViz and exBERT allow researchers to inspect specific attention patterns and compare model decisions. However, they note that most tools are designed for English and classification tasks, making them less suitable for multilingual or sequence labelling settings.

### 2.6.3 Interpretability Challenges

Across the interpretability literature, several limitations remain persistent. Most techniques are developed and evaluated in the context of English and focus predominantly on classification tasks, with limited attention to NER or Arabic (Sun et al., 2021; Braşoveanu and Andonie, 2020). Evaluating interpretability methods is also an open challenge. Because there is no formal definition of interpretability, there are no widely accepted standards for how to evaluate it (Zhao et al., 2024; Lertvittayakumjorn and Toni, 2021). Most approaches rely on human-centred assessments, based on how useful or understandable the explanations seem. As a result, how model behaviour is interpreted often depends on the task, the user, and the design of the tool being used.

These challenges motivate the need for a multi-dimensional, language-aware framework—integrating evaluation metrics, interpretability techniques, and system-level diagnostics through visual analytics—as introduced in the following chapter.

# Methodology 3

The previous chapters introduced the research problem, outlined the motivations for this work, and reviewed background literature across five dimensions: transformer architectures, the NER task and its datasets, linguistic and structural challenges in Arabic, potential factors influencing performance, and the limitations of existing evaluation and interpretability techniques. We also introduced **DeformAr**, a visual analytics and diagnostic framework designed to support comparative evaluation of transformer-based NER systems. DeformAr combines standard metrics with component-level behavioural analyses and interpretability methods, integrated through an interactive dashboard, to diagnose how model and data components jointly shape system behaviour.

This chapter presents the core methodology of the thesis. It addresses **RQ0**, which focuses on how NER systems can be decomposed into model- and data-level components to support interpretable, cross-lingual evaluation. The answer to this question has both a *conceptual aspect*—the design of a framework that explains system-level performance outcomes in terms of underlying component behaviour—and a *technical aspect*—the implementation of the system pipeline, experimental setup, and dashboard that integrate this design. To assess the effectiveness of the framework, the thesis adopts a **comparative case study** between Arabic and English, enabling both depth of analysis and a practical demonstration of the framework.

The chapter is organised as follows. Section 3.1 provides a system-level overview of DeformAr. Section 3.2 outlines the experimental setup and motivates the case-study design (single dataset and single model). Section 3.3 details the data-extraction work-flow and resulting outputs. Section 3.4 introduces the dashboard which supports both **cross-component analysis** and **behavioural analysis**. Section 3.5 positions DeformAr relative to existing tools and methods. Section 3.6 established how the proposed methodology addresses **RQ0**, defining both its analytical scope and its role in supporting the subsequent research questions. Finally, Section 3.7 outlines the methodological contributions of this chapter.

## 3.1 DeformAr Overview

**DeformAr** is designed to support comparative evaluation of NER systems by decomposing them into two core components: the **data** and the **model**. Each component is further divided into subcomponents that capture different aspects of system behaviour and their interdependencies. The **data component** consists of the input vocabulary and the entity annotations. These subcomponents capture both the linguistic characteristics of the language and the structure of its labelled datasets. The **model component** includes the encoder and the output layer. The encoder generates contextualised token representations, while the output layer maps these representations to predicted entity labels by learning from the annotations. This conceptual design enables **DeformAr** to analyse NER system behaviour both in isolation and in interaction. It examines the properties of individual subcomponents (e.g., vocabulary, encoder) as well as the interactions between them, in order to understand how they shape overall performance and contribute to performance discrepancies.

To facilitate this analysis, **DeformAr** is organised into two main phases. The first is the **data extraction phase** (Section 3.3), where the NER pipeline is defined and the decomposition into data and model components is implemented. This phase prepares and processes both components by applying language-specific preprocessing, fine-tuning the NER models, extracting outputs from multiple levels of the pipeline, and generating structured outputs to support the dashboard analysis. The second is the **dashboard phase** (Section 3.4), which supports both cross-component and behavioural analysis. Cross-component analysis quantifies differences between Arabic and English using metrics and distributional summaries across data and model subcomponents, while behavioural analysis combines interpretability views—such as representations and attention— with token-level behaviour metrics (e.g., uncertainty and ambiguity) and instance-level (example-based) analysis, enabling examination of how subcomponent behaviours interconnect and vary across languages.

## 3.2 Experimental Setup and Case Study Design

To implement and assess the framework described above, this thesis adopts a comparative case study between Arabic and English. The goal is to investigate how data and model components interact across languages under a controlled setup that keeps the architecture, training procedure, and dataset consistent. This design enables a deeper analysis of how component-level behaviours jointly shape performance, isolating interpretability and diagnostic differences rather than experimental variation. This section outlines the system configuration, the selected datasets, and the methodological justification for focusing on a single model and a single dataset per language.

**Case Study Scope** To provide an in-depth analysis, this thesis adopts a comparative case study using a single dataset and a single model for both English and Arabic. This design prioritises depth over breadth: expanding to multiple datasets or models would risk reducing the analysis to surface-level comparisons, while the aim is to examine in detail how component-level behaviours jointly shape system outcomes. Basing the case study on established benchmarks and widely adopted models ensures both reproducibility and comparability with prior work, while allowing the framework to demonstrate its diagnostic capabilities in detail.

**System Configuration** The NER system is based on transformer encoders with a token-classification output layer. Specifically, we use **BERT-base-cased** (Devlin et al., 2019) for English and **AraBERTv02-base** (Antoun et al., 2020) for Arabic. These models are widely adopted in NER evaluations, chosen for their architectural similarity and effectiveness (Devlin et al., 2019; Li et al., 2022; Pakhale, 2023; Jarrar et al., 2023). Both models are pretrained on large monolingual corpora and fine-tuned for token classification. The output layer is a linear classifier (single-layer MLP) that maps contextualised token representations to the encoded entity labels. This configuration reflects standard practice in BERT-based NER systems, as discussed in Section 2.3.

**Datasets** For Arabic, we use **ANERcorp** (Benajiba et al., 2007), a newswire corpus that has been widely regarded as the standard Arabic benchmark (Obeid et al., 2020; AlDuwais et al., 2024). For English, we use the **CoNLL-2003** dataset (Tjong Kim Sang and De Meulder, 2003), which is the canonical benchmark in NER research. Both datasets are well suited for cross-lingual comparison because they define the same four entity types: PER (Person), LOC (Location), ORG (Organization), and MISC (Miscellaneous). In addition, both adopt the IOB2 annotation scheme, ensuring structural alignment for direct comparison.

The next section introduces the data extraction phase, where data and model components are prepared, fine-tuned, and transformed into structured outputs used by both cross-component and behavioural analysis.

## 3.3 Data Extraction Phase

The data extraction phase is the core of the **DeformAr** methodology. It performs the main processing steps and produces the structured outputs that the dashboard relies on for both cross-component analysis and behavioural analysis. This phase is therefore the main focus of this chapter.

As outlined in Section 3.1, DeformAr decomposes NER systems into data and model components, which can be examined both in isolation and in interaction. The extraction phase executes this design by defining the NER pipeline, applying preprocessing and fine-tuning, and systematically extracting outputs from multiple levels of the system. These outputs are then fed into the dashboard, where component behaviours and their interactions can be analysed to understand how they shape cross-lingual performance. Figure 3.1 illustrates the core components of DeformAr for a single language, highlighting how the data and model subcomponents relate to one another. The next subsection provides an overview of the key processing stages applied to each language during this phase.

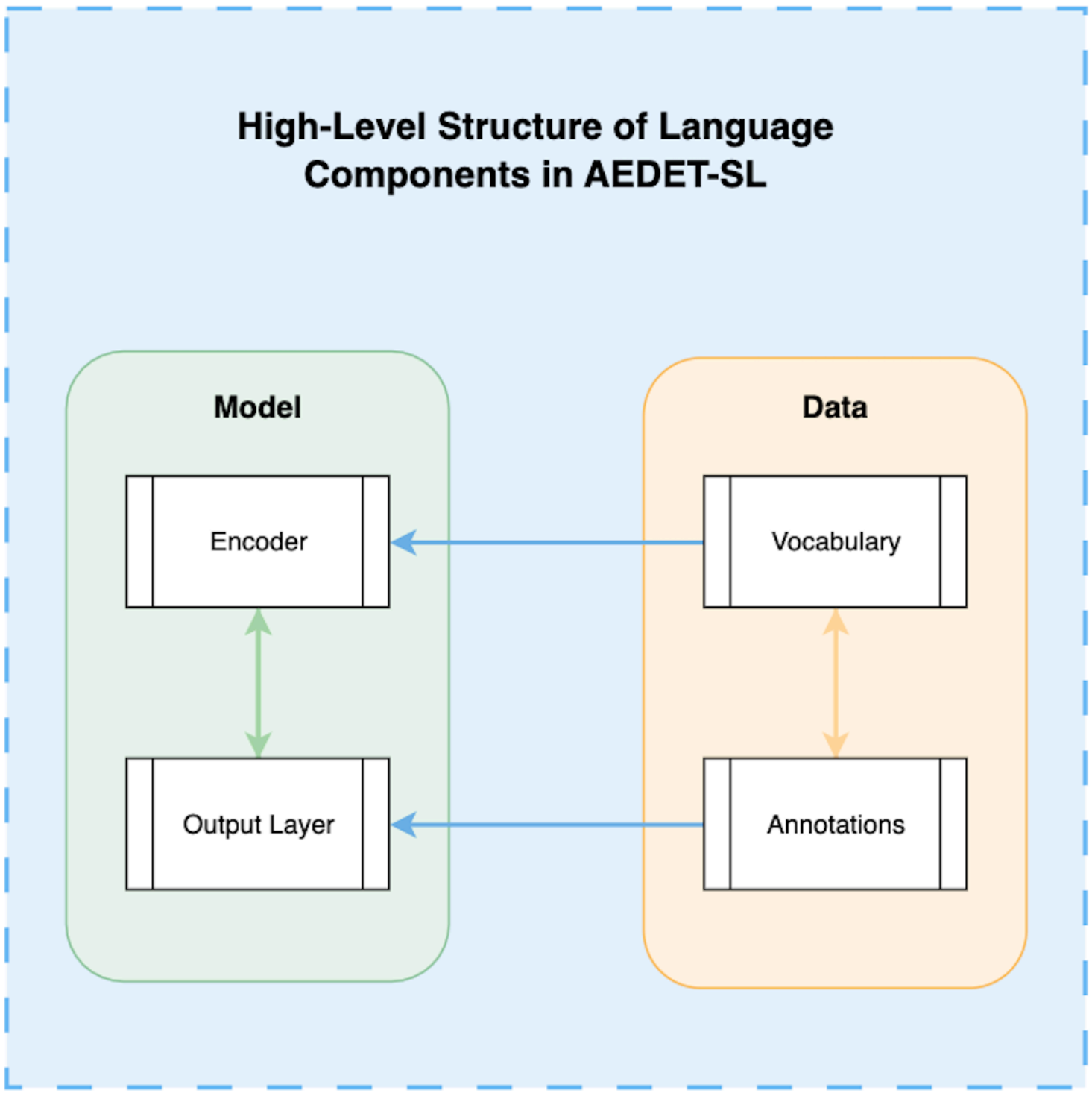

Figure 3.1: High-level structure of language components in DeformAr. **Orange arrows** and **green arrows** represent within-component interaction, while **blue arrows** represents cross-component interaction between data and model subcomponents.

### 3.3.1 Overview

The data extraction phase serves two purposes: it prepares the data and model components for processing, and it produces the structured outputs used across both cross-component and behavioural analysis. Each language passes through the following steps:

- **Data preparation**: Raw datasets are loaded from various formats and converted into a consistent structure. This includes organising the data into sentences and aligning labels into a unified format that facilitates further processing.

- **NER system architecture**: We adopt a pretrained BERT encoder followed by a classification layer as the architecture for the NER model. This setup is shared across all languages and experiments.

- **Preprocessing and tokenisation**: Language-specific preprocessing (e.g., normalisation for Arabic) is applied, followed by subword tokenisation using the WordPiece tokeniser. This step also handles label alignment, padding, truncation, and sentence boundary formatting.

- **Model fine-tuning**: The NER model is trained on the preprocessed data using the same architecture and hyperparameter configuration across languages. This step also includes evaluation, where predictions, confidence scores, and loss values are generated for later analysis.

- **Output extraction**: We extract a range of outputs from both the data and model components. These include the fine-tuned model, tokenised inputs, gold and predicted labels, prediction confidences, attention weights, hidden states, and alignment metadata. These outputs support both metric-based and interpretability-based analysis.

- **Output processing**: Extracted outputs are further processed into a unified analysis format. This includes dimensionality reduction using UMAP, clustering, calculating various metrics across components, and consolidating all outputs into a single structured output.

### 3.3.2 Data Preparation

The data preparation step involves loading and structuring the datasets into a unified format. As described in Section 3.2, we utilise two datasets: ANERCorp for Arabic and CoNLL-2003 for English. This section summarises how these datasets are prepared for further processing and fine-tuning.

NER datasets are typically structured as sequences of words, grouped into sentences. Each word is annotated with a label that indicates whether it is inside or outside a named entity, along with the entity type according to the annotation scheme. As described by Tjong Kim Sang and De Meulder (2003), each word appears on a separate line, and sentences are separated by empty lines. A simplified example is shown in Table 3.1.

ANERCorp is provided as plain text files, with separate files for training and testing. It uses the same word-per-line format described above, where each word is paired with its NER label, according to the IOB2 format. Sentence boundaries are determined using full stops. We utilise the split provided by Obeid et al. (2020), where 5/6 of the words are used for training and 1/6 for testing[1]. For CoNLL-2003, we use the version provided by the HuggingFace datasets library,[2] which provides the data in a structured format, split into training, validation, and test sets. The data is organised into sentences, with each word annotated for named entity recognition, part-of-speech, and syntactic chunking. In our setup, we extract only the NER annotations. These are provided as integer indices and mapped to standard IOB2 entity tags. Note that each dataset supports 9 classes (B/I-PER, B/I-ORG, B/I-LOC, B/I-MISC) plus the O tag for non-entity tokens.

Table 3.1: Hypothetical example of sentence-level NER annotation format in Arabic and English. Each word is labelled with its corresponding entity tag according to the IOB2 scheme.

| Arabic Token | Label (AR) | English Token | Label (EN) |
|---|---|---|---|
| أحمد | B-PER | Ahmed | B-PER |
| ذهب | O | went | O |
| إلى | O | to | O |
| المدرسة | O | school | O |
| . | O | . | O |

Both formats are supported in our library through a unified loading process. We convert each dataset into a dictionary-based format that stores the data split and the sentences it contains. Within each split, we track the sentence ID, the list of words, and their corresponding entity labels. This unified format streamlines both processing and fine-tuning. Details of the preprocessing and tokenisation strategy are described in Section 3.3.4.

As shown in Figure 3.1, the data component is consolidated into two key subcomponents: vocabulary and annotations. The relationship between these subcomponents is bidirectional. For instance, annotation patterns can increase the ambiguity of certain words by assigning multiple entity types to the same form. Conversely, certain properties of the vocabulary — such as the presence of rare words or long entity spans — may make some entity types harder to learn during training. In the next section, we examine the NER system architecture — the component that processes this data and learns to map vocabulary to entity labels through contextual representation.

[1] https://camel.abudhabi.nyu.edu/anercorp/
[2] https://huggingface.co/datasets/eriktks/conll2003

### 3.3.3 System Architecture

The NER system follows a standard BERT-based design, as described in Section 3.2. It consists of a BERT encoder followed by a linear classification layer. The input sentence is first encoded through BERT, producing contextualised token representations. We use the last hidden states from the encoder to represent each token — a common setup introduced by Devlin et al. (2019)[3]. These representations are passed through a dropout layer and then into the classification layer, which produces raw prediction scores (logits) over the nine labels for each token.

These scores are used to compute the classification loss against the gold labels, as demonstrated in Figure 3.2. The model is also configured to output attention weights and hidden states to support later interpretability analysis.

As shown in Figure 3.1, the model component is composed of two subcomponents: the encoder and the output layer. The encoder transforms input tokens into contextualised representations, while the output layer learns to map these representations to predicted labels. This interaction is bidirectional: the output layer learns to distinguish between tokens based on the encoder's representations, while the encoder itself is updated through supervised learning from the annotations to better align those representations with the target labels. In the next section, we describe how data is preprocessed and prepared for fine-tuning.

### 3.3.4 Preprocessing and Tokenisation

As mentioned earlier, each dataset is formatted as a sequence of words grouped into sentences, which are passed to the tokenisation module. Each word is tokenised using the WordPiece tokeniser for both Arabic and English. However, as discussed in Section 2.1.3, AraBERTv02 applies an additional language-specific preprocessing step before tokenisation, which normalises the input text[4]. For English, no additional preprocessing is required.

[3] This is also the default behaviour in HuggingFace's token classification implementation. We follow this approach to avoid additional pooling or abstraction approaches that may affect interpretability, and to align with standard NER fine-tuning practices.

[4] If the preprocessor is skipped, many valid Arabic words may be incorrectly mapped to the [UNK] token.

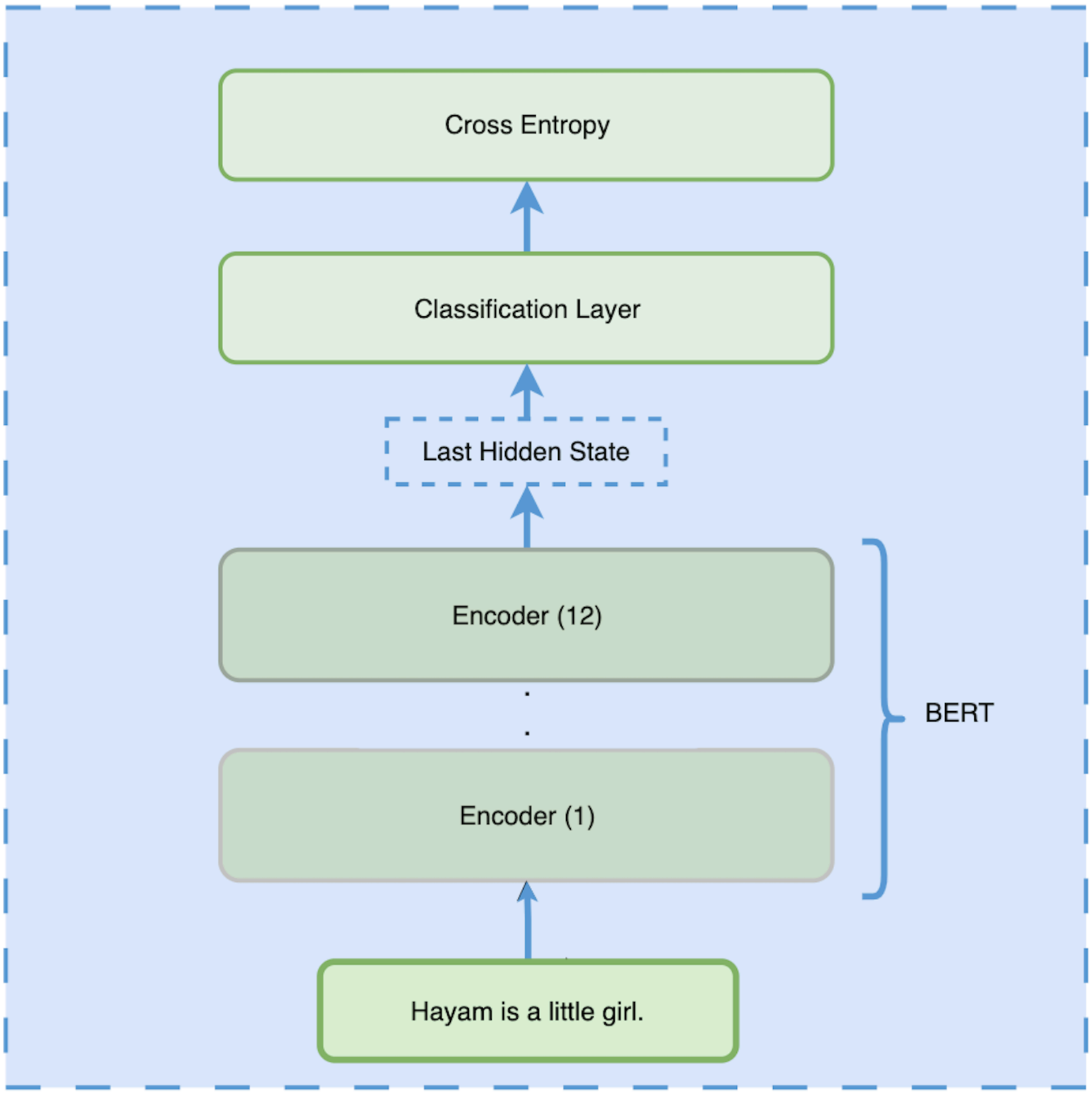

Figure 3.2: Architecture of the NER system used in this thesis. The input sentence is processed through a BERT encoder. The last hidden states are passed through a dropout layer and then to a linear classification layer. The output consists of token-level prediction scores over the NER label set.

An important consideration at this stage is how words are represented when tokenised into subwords. A single word can be split into multiple subwords, which requires aligning the entity labels accordingly. We follow the standard approach proposed by Devlin et al. (2019) and adopted by Antoun et al. (2020), where the entity label is assigned to the first subword only. All remaining subwords are assigned a special ignore index (-100), which marks them to be excluded from later training steps. The handling of these ignored tokens during fine tuning computation is discussed in Section 3.3.5.

After tokenisation, sentence boundaries are preserved, and each sequence is wrapped with special tokens: [CLS] at the beginning and [SEP] at the end. Sequences are truncated to a maximum of 256 tokens and padded if shorter. This process introduces two types of tokens: *core tokens*, which contribute to the model's predictions (i.e., the first subword of each word, or the word itself if it is not split), and *ignored tokens*, which include [CLS], [SEP], padding, and all other subwords beyond the first.

Our framework also supports alternative strategies, such as using the last subword or assigning labels to all subwords. However, we adopt the first subword strategy as the default, as it introduces the smallest changes to the structure of the data (see Figure 3.3). Other approaches — such as assigning labels to all subwords — can distort label distributions[5].

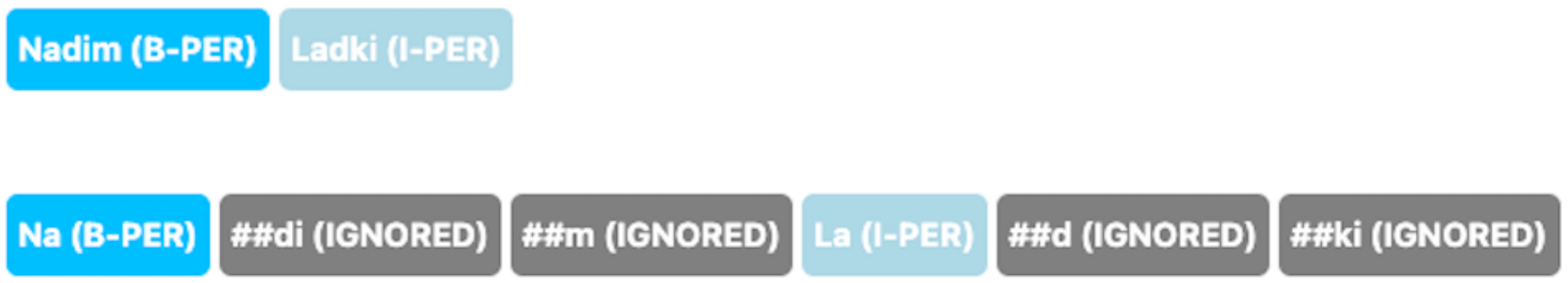

Figure 3.3: Example of tokenisation using the first subword strategy. The word *Nadim* is split into three subwords: *Na*, *##di*, and *##m*. Only the first subword (*Na*) is assigned the entity label (*B-PER*); the remaining subwords are marked as *IGNORED* and excluded from later training steps. The same process is applied to *Ladki*.

As shown in Figure 3.4, the preprocessing step expands the data and model components. On the data side, the vocabulary is unpacked into two elements: raw words and core tokens, while the annotations are specified as true labels. On the model side, the encoder is divided into two subcomponents: tokenisation and representation, which highlights how the model processes the input and generates contextualised representations. A new interaction is also introduced: tokenisation maps words into core tokens, which are then passed to the representation layer. It's important to note that while only the first subword of each word is associated with an entity label, the full tokenised sequence — including ignored subwords and special tokens — is still processed by the representation subcomponent. Likewise, the output layer generates logits for the entire sequence, including both core and ignored tokens.

5 This is the default behaviour in HuggingFace's token classification pipeline, where each subword is assigned the I-TAG corresponding to the original entity label.

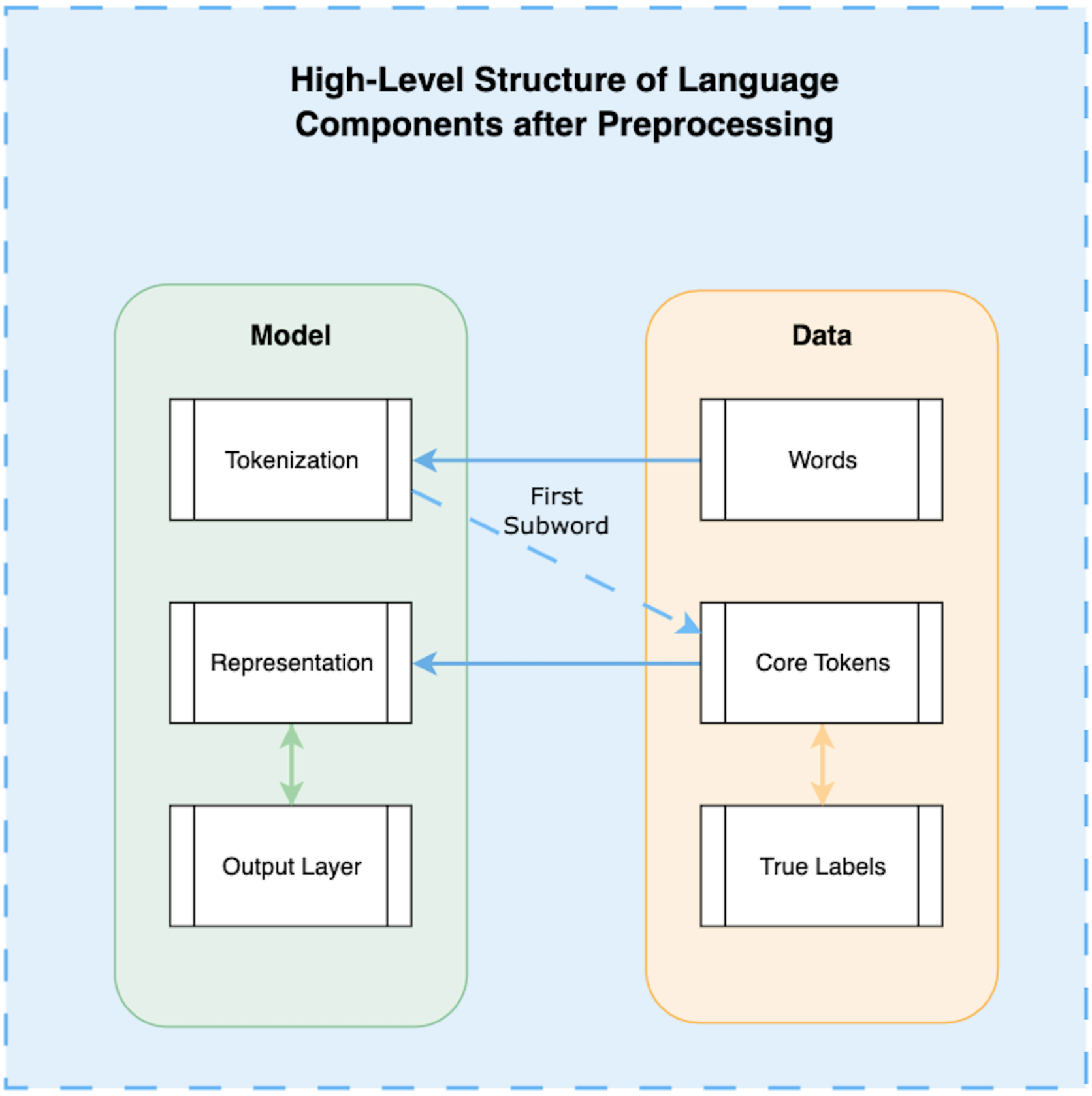

Figure 3.4: High-level structure of language components after preprocessing in DeformAr. **Orange** and **green arrows** indicate within-component interaction, while **blue arrows** represent cross-component interaction. The **dashed blue arrow** highlights the new interaction introduced by the tokenisation step.

### 3.3.5 Model Fine Tuning

This section describes the fine-tuning procedure used to train the NER models on Arabic and English datasets. As discussed in Section 3.2, we use AraBERTv02-base for Arabic and BERT-base for English. Both models are fine-tuned using the same architecture, training procedure, and evaluation approach to ensure consistency across languages.

**Hyperparameter configuration** The models are fine-tuned using the AdamW optimiser with a learning rate of `5e-5`, a batch size of `16`, and a total of 4 epochs. A linear learning rate scheduler with warm-up is applied, with a warm-up ratio of `0.1` to stabilise early training. All model parameters are trained, with the exception of LayerNorm and bias terms, which are frozen[6]. Dropout with a rate of `0.1` is applied before the classification layer. Gradient accumulation is set to `1`, and gradient clipping is applied with a maximum norm of `1.0` to prevent unstable parameter updates during training. The same configuration is used for both Arabic and English models to ensure comparability across languages.

[6] Following common practice in fine-tuning to improve stability and prevent over-fitting.

**Fine Tuning Procedure** During fine-tuning, each tokenised input sentence is passed through the encoder, which produces contextualised representations for every token in the sequence. These representations are fed into the output layer, which generates a set of logits — raw prediction scores — over the nine entity labels for each token. These logits are then compared to the corresponding true labels to compute the loss, which is used to update the model parameters through backpropagation.

However, due to subword tokenisation, not every token in the sequence is aligned with a true label. As discussed in Section 3.3.4, only the first subword of each tokenised word (i.e., the core token) is assigned the true label associated with the original word. All remaining subwords, along with special tokens such as `[CLS]` and `[SEP]`, are assigned a special ignore index (`-100`) and excluded from the loss calculation. This index is recognised by PyTorch's cross-entropy loss function, which ignores these tokens during training.

**Evaluation Strategy** During evaluation, the raw prediction scores (logits) produced by the model are passed through a softmax function to obtain probability distributions over the nine entity labels. The label with the highest probability is selected as the predicted tag for each token.

We evaluate the model using two standard approaches: **token-level** and **entity-level** evaluation. For both, we compute precision, recall, F1-score, and provide a full classification report. As discussed in Section 2.3.4, we use `scikit-learn` for token-level evaluation and `seqeval` for entity-level evaluation. Within `seqeval`, we report both strict and non-strict modes: the strict evaluation requires exact span and type matches according to the IOB2 scheme with discard repairing mechanism for improper sequences. The non-strict variant (default) applies a beginning-tag repair mechanism according to the IOB1 format.

As shown in Figure 3.5, the fine-tuning step introduces two additional subcomponents: a loss function on the model side and predicted labels on the data side. The encoder output (i.e., the contextual representations) is passed to the output layer, which generates token-level logits over the label set. These logits serve two purposes: (1) they are compared against the true labels to compute the training loss, and (2) they are used to derive predicted labels via a maximum probability operation (i.e., `argmax`). The training loss is computed using cross-entropy, which compares the logits of core tokens (those aligned with true labels) against their true label. Subwords and special tokens with the ignore index (`-100`) are excluded. The loss value is then used to update the weights of the encoder and output layer via backpropagation. This introduces a new internal interaction (gradient update) within the model. On the evaluation side, the predicted labels are compared to the true labels to assess performance using standard metrics. It is worth noting that although the model processes all tokens in the sequence, gradient updates are computed based on core tokens only. The result of this process is a fine tuned model and a set of intermediate outputs from both the data and model components. In the next step, we extract these outputs to support evaluation and analysis across the DeformAr framework.

### 3.3.6 Output Extraction

After the fine-tuning process, **DeformAr** extracts various outputs from both the model and data components. These outputs are selected to provide broad coverage across the different subcomponents of the system that were introduced during the earlier steps of the data extraction phase. While these outputs are not exhaustive of all possible interactions and relationships, they are sufficient to study the internal behaviour of each component in isolation, as well as their interactions across languages.

In this step, we consolidate the subcomponents introduced during fine-tuning. For each subcomponent, we identify a set of outputs that help us understand its internal behaviour and how it interacts with other parts of the system. Most outputs are extracted during inference on the test set and are selected based on their relevance to the three dimensions of **DeformAr**: model-level behaviour, data-level characteristics, and cross-component interactions.

Figure 3.6 provides an overview of the extracted outputs, grouped by component and subcomponent. On the model side, we extract outputs from the encoder (e.g., hid-den states, attention weights), the output layer (e.g., prediction probabilities), and the loss function. On the data side, we extract vocabulary information (e.g., core tokens), annotation information (e.g., true and predicted labels), and tokenisation metadata. Together, these outputs form the basis for both cross-component and behavioural analysis.

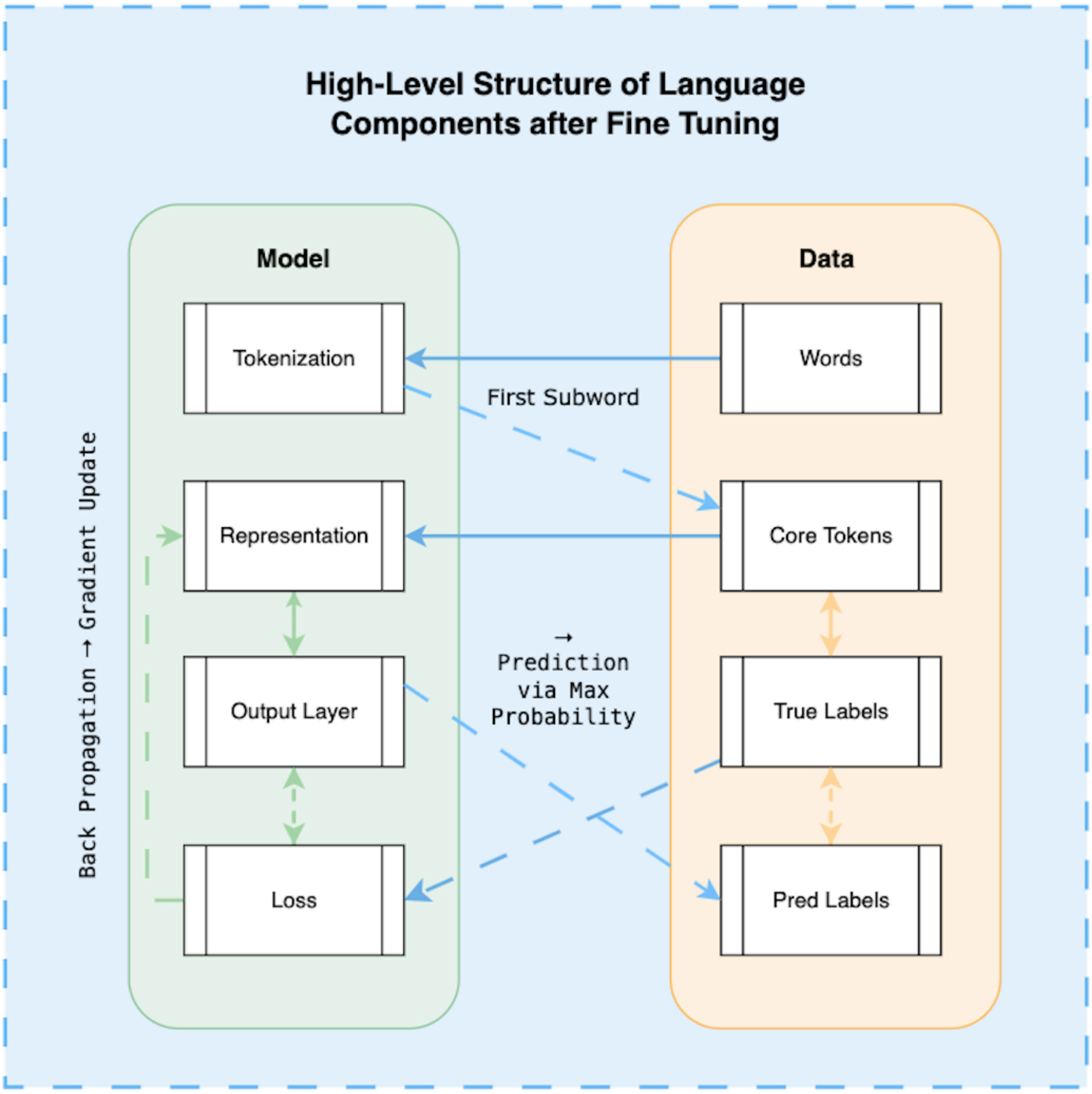

Figure 3.5: High-level structure of language components during fine-tuning in DeformAr. **Orange** and **green arrows** represent within-component interaction, while **blue arrows** represent cross-component interaction. Dashed arrows indicate new interaction introduced by the fine tuning step, such as the generation of predicted labels via `argmax`, and weight updates from the loss function via backpropagation.

**Encoder Outputs** The first set of model outputs comes from the encoder. These include the last hidden states from the last layer (layer 12) of the transformer, as well as the attention scores and attention weights associated with each attention head. All three outputs are analysed in the behavioural analysis stage to support interpretability-driven exploration. As discussed in Section 3.3.5, the loss function updates the representations of the encoder through backpropagation. The **attention weights** are one of the parameters that get updated during this process. We extract these weights from both the pretrained and fine-tuned models in order to compare how they change after training, and to understand how these changes differ across languages. The **attention scores** represent how each token attends to other tokens in the sequence and are commonly used in interpretability research. As mentioned in Section 2.1.2, these scores are derived from the attention weights, so we extract them to investigate how attention patterns shift after fine-tuning. These scores are also extracted for both the pretrained and fine-tuned models across languages. Finally, we extract the **last hidden states**, which are the contextualised representations for each token in the input sequence. These representations serve as the input to the output layer and directly influence the model's predictions. Their behaviour changes during training as a result of updates to the attention weights and the contextual patterns learned by the model. We also extract them for both the pretrained and fine-tuned models to understand model behaviour across languages and how it changes after fine-tuning.

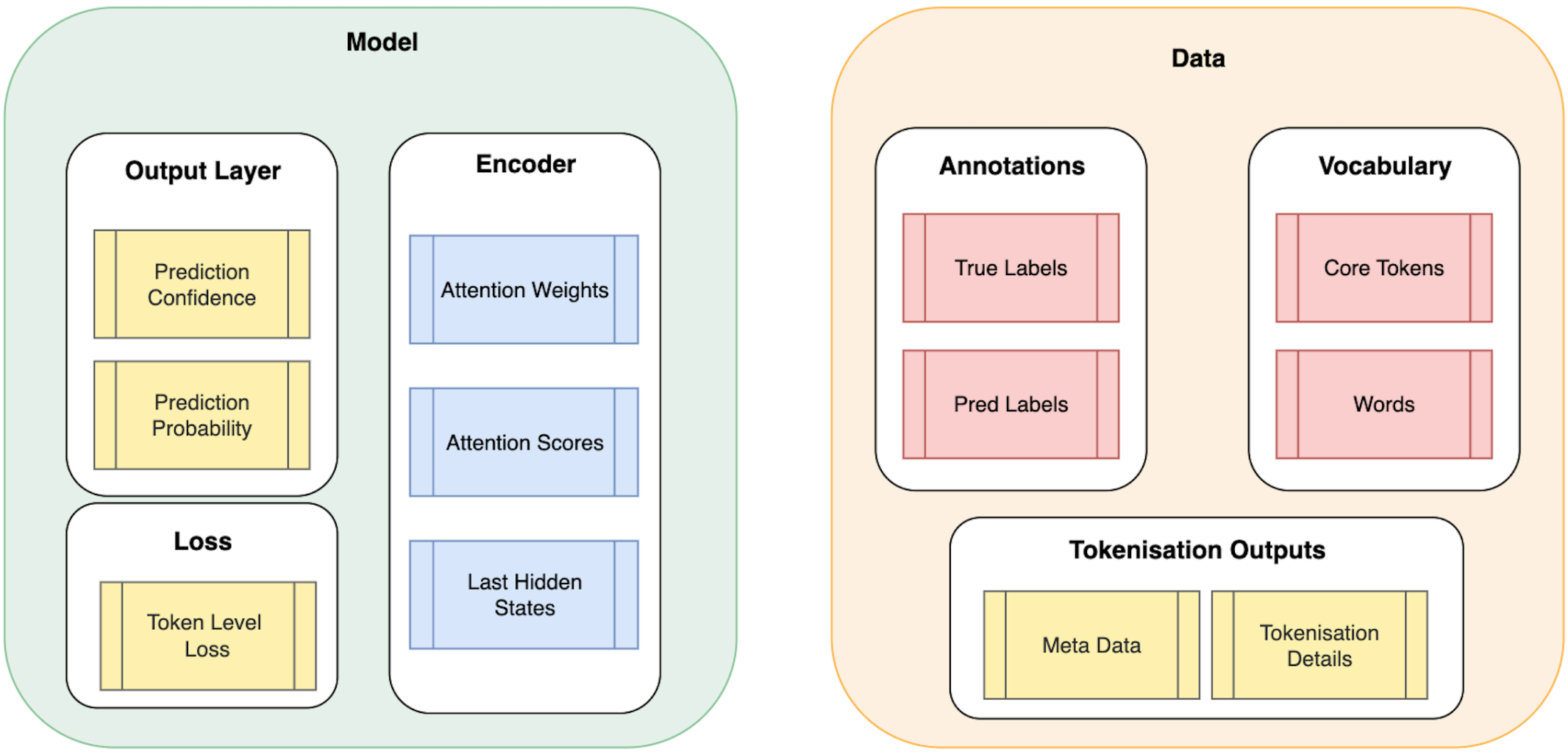

Figure 3.6: Overview of extracted outputs in **DeformAr**, grouped by component and subcomponent. Colours indicate the main type of analysis each output supports: red for cross-component analysis, blue for behavioural analysis, and yellow for outputs shared across both.

**Output Layer** The second set of model outputs comes from the output layer. These include the softmax-normalised probabilities and the prediction confidence. Both are used to help understand the behaviour of the output layer. The **probabilities** are calculated by applying the softmax function to the raw prediction scores (logits), producing a normalised distribution over the nine labels for each token. These probabilities are used to study model confidence and uncertainty, particularly for ambiguous or low-certainty predictions. The **prediction confidence** is defined as the maximum probability from the predicted distribution for each token. This value helps us understand how confident the model is in its final decision.

**Loss Function** The final model output we extract is the **token-level loss**. This value represents the error between the predicted label distribution and the true label for each core token. It helps identify difficult or uncertain instances where the model struggles to align its predictions with the annotated labels. It also helps us understand how the loss behaviour differs across languages.

**Vocabulary Outputs** The first set of data outputs comes from the vocabulary subcomponent. These include the **original words** and the **core tokens** of the test data. These outputs help us understand how the model processes the input text at the lexical level, and how tokenisation affects the linguistic characteristics of the data across languages. By examining these outputs, we can analyse how often certain word forms are tokenised, how frequently they appear, and how these patterns vary across languages. These differences can affect how reliably the model learns to recognise entity types.

**Annotation Outputs** The second set of data outputs comes from the annotation subcomponent. These include the **true labels** assigned to each core token, the **predicted labels** generated by the model, and the **alignment metadata** used to map between words, tokens, and labels. These outputs allow us to evaluate model performance, analyse errors, and compare annotation behaviour across languages. Both sets of annotations are used in the cross-component analysis to produce standard evaluation metrics such as precision and F1-score.

**Generic Outputs** In addition to component-specific outputs, DeformAr extracts a number of general-purpose outputs such as **sentence IDs** and **tokenisation details**, including the full WordPiece sequences for each word. These are primarily used for navigation and filtering purposes within the analysis and dashboard interface.

Not all extracted outputs are passed directly to the dashboard. Certain outputs—such as attention weights, attention scores, and hidden states—require further processing before they can be used for analysis or visualisation. In the next section, we describe how these outputs are transformed and consolidated into a unified format that supports both the cross-component and behavioural analysis stages of DeformAr.

### 3.3.7 Output Processing

The output processing phase serves two main purposes. First, it applies further processing to selected extracted outputs in order to generate additional information required for analysis. Second, it consolidates all relevant outputs into a unified format for use in the dashboard. In the following subsections, we briefly describe how each output is processed and summarise the key hyperparameters used. Further details about how these derived metrics are calculated, interpreted, and analysed are provided in the relevant analysis chapters.

**Hidden States** The last hidden states from the encoder are high-dimensional vectors representing contextualised token representations. To make these representations more interpretable, we apply dimensionality reduction using UMAP (McInnes et al., 2020). UMAP is a non-linear projection technique that preserves both global and local structure in a lower-dimensional space. It is applied separately to hidden states from the pretrained and fine-tuned models to allow comparison before and after training. We use the following UMAP hyperparameters: `n_neighbors=15`, `min_dist=0.1`, `metric="cosine"`, and `n_components=2`. The output consists of two-dimensional coordinates, which are later used in the behavioural analysis stage to examine representational behaviour across languages and entity types. Further details about the analysis and how these coordinates are used are provided in Chapter 5.

In addition to UMAP, we apply K-Means clustering to the hidden states from the fine-tuned model. K-Means is a centroid-based algorithm that partitions the representation space into a fixed number of clusters. We use `k` values of `3`, `4`, and `9`, which correspond to different levels of interpretation (e.g., boundary-level, entity-level, and token-level clusters). Clustering is performed using the `k-means++` initialisation method with

n_init=10, and the representations are L2-normalised prior to clustering. The resulting outputs include: (1) clustering assignments for each token, (2) centroid similarity scores for selected k, and (3) standard clustering metrics such as V-Measure. These are described in more detail in Section 5.6.

Finally, we compute the silhouette score for the hidden states, using both the true labels and the predicted labels as label sets. Silhouette analysis is based on cosine similarity and provides a measure of how well-separated the representations are with respect to each label set. Further details on how silhouette scores are calculated are provided in Section 4.4.1. Figure 3.7 presents the processing steps applied to the last hidden states, including dimensionality reduction, clustering, and silhouette evaluation.

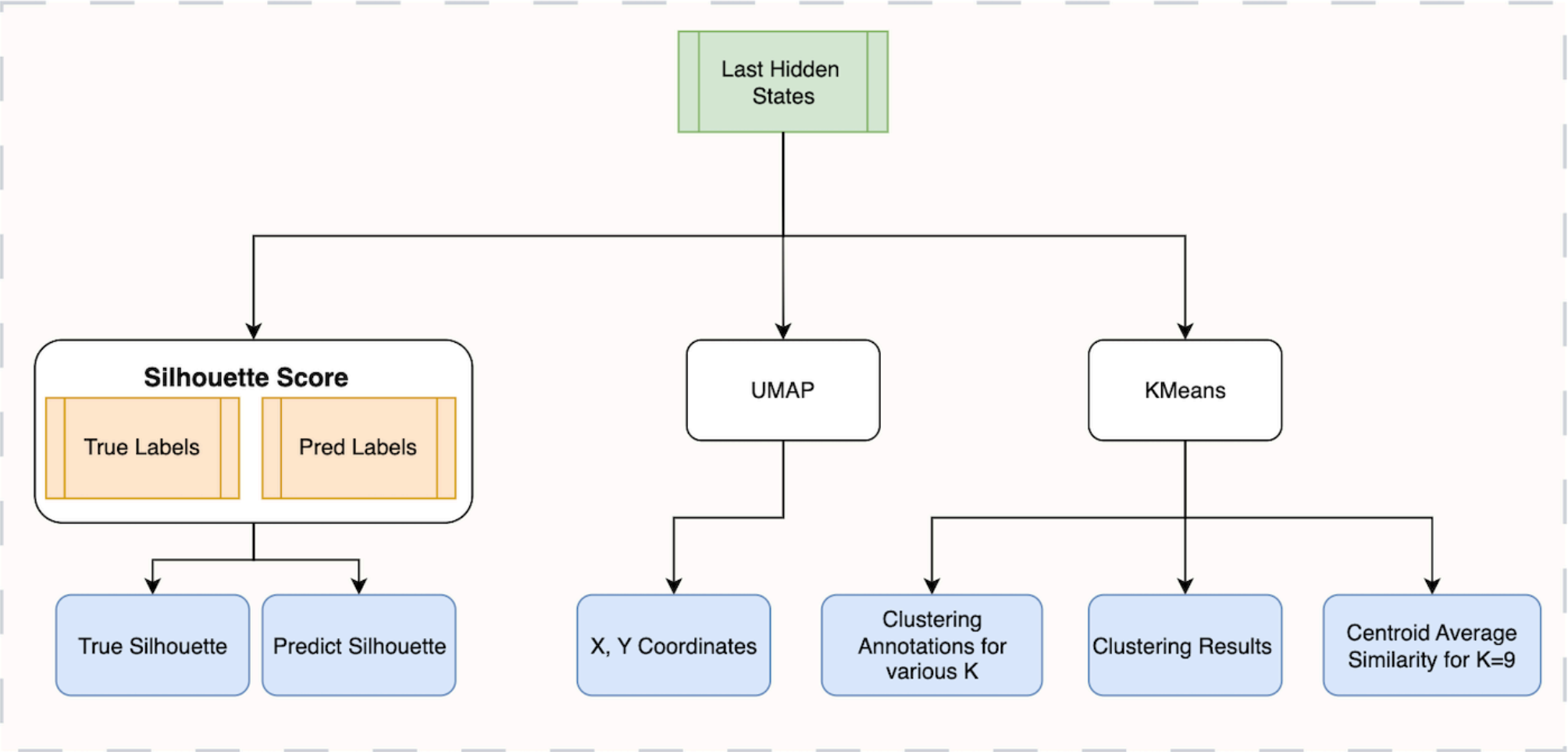

Figure 3.7: Overview of the processing applied to the last hidden states. UMAP and K-Means are used to produce dimensionality-reduced coordinates and clustering results, while silhouette scores are computed using true and predicted labels. Green indicates outputs extracted from the model; blue indicates outputs used in behavioural analysis stages.

**Similarity and Uncertainty Measures** In addition to the high-dimensional representations described above, we apply two types of simple post-processing using cosine similarity and entropy. For both attention weights and attention scores, we calculate cosine similarity between the outputs of the pretrained and fine-tuned models. This produces two similarity measures—**attention weight similarity** and **attention score similarity**—which allow us to track how each attention mechanism shifts after training. For the prediction probabilities, we calculate the entropy of the softmax-normalised output for each token. This provides a measure of uncertainty in the model's predic-tions. Further details on how these measures are calculated are provided in Section 5.2 and 4.5.1 respectively. Figure 3.8 presents the processing applied to attention weights, attention scores, and prediction probabilities, where cosine similarity and entropy are used to produce interpretable numeric outputs.

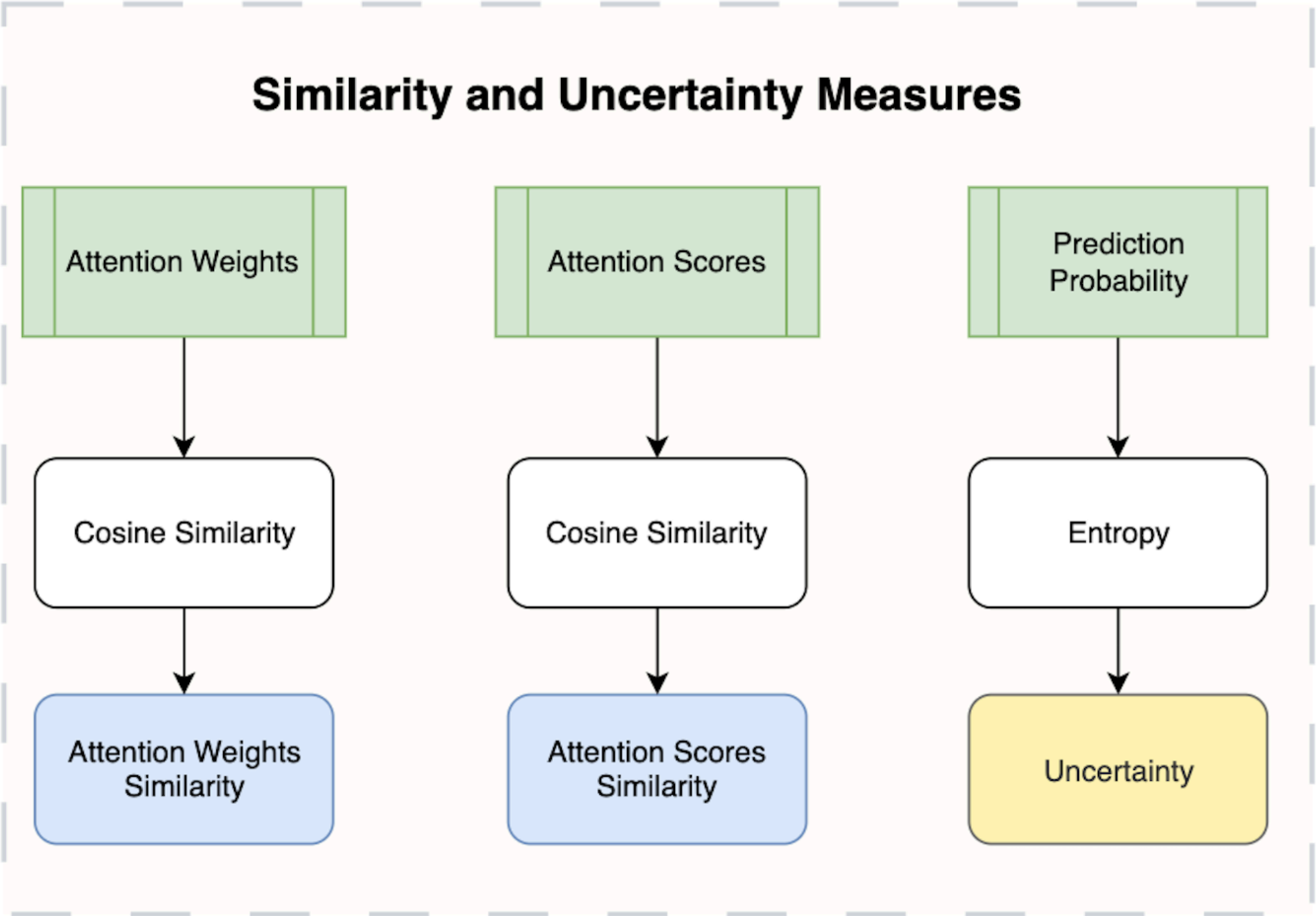

Figure 3.8: Overview of the processing applied to attention weights, attention scores, and prediction probabilities. Green indicates model-derived outputs; blue indicates outputs used in behavioural analysis, while yellow indicates outputs used in both cross-component and behavioural analysis.

**Evaluation Metrics** To evaluate model performance, we compare the predicted and true labels using standard classification metrics. We compute token-level metrics using `scikit-learn` and entity-level metrics using `seqeval`. These include precision, recall, F1-score, and a full classification report. In addition, we generate confusion matrices and apply error analysis techniques to identify common misclassifications. All evaluation results are presented and discussed in the cross-component analysis in Section 4.5. Figure 3.9 presents the processing used to evaluate model predictions, including standard classification metrics, confusion matrices, and error analysis techniques.

**Vocabulary-Based Processing** As discussed in Section 3.3.4, there is a strong relationship between the structure of the vocabulary and the annotation patterns it receives.

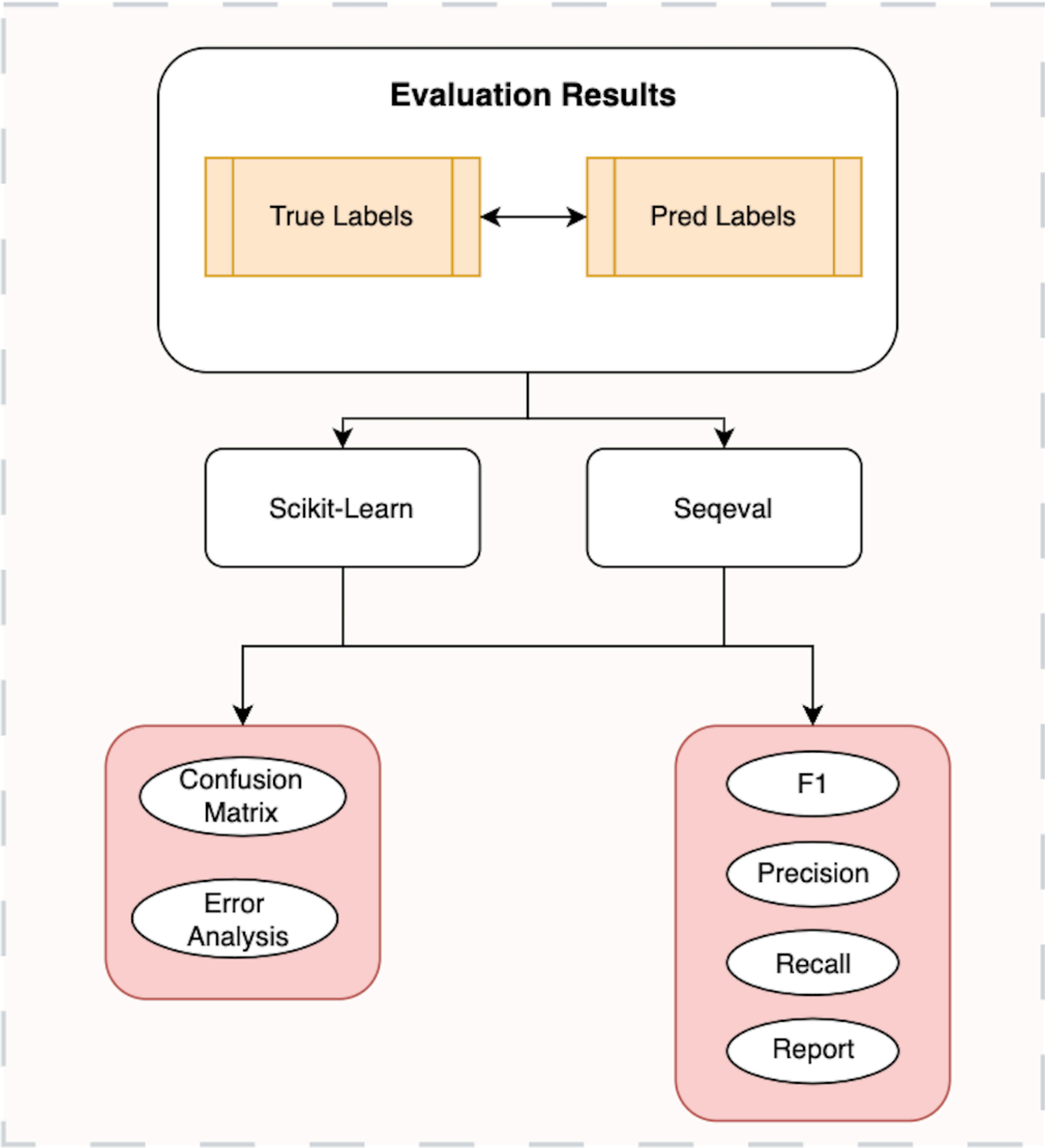

Figure 3.9: Overview of the processing used to evaluate model predictions. Red indicates outputs used in quantitative analysis.

To capture this interaction, we introduce an additional analysis step that connects the test set to the training set. For each word (i.e., pre-tokenised form) and each core token (i.e., the word itself or its first subword after tokenisation), we identify all of its occurrences in the training data, then calculate two metrics: **ambiguity** and **consistency**. Ambiguity captures how many different annotation types the token received in the training split. Consistency measures whether the annotation assigned to the token in the test set aligns with the annotations it received in the training set. These metrics are used to explore the interaction between vocabulary structure and annotation patterns. Figure 3.10 presents the calculation of ambiguity and consistency metrics.

The final step in the output processing phase is to consolidate all processed components into a unified output format, ready for use in DeformAr dashboard. This in-cludes the following elements: the fine-tuned model itself, evaluation results (including classification reports and confusion matrices), averaged metrics such as similarity and uncertainty scores, and analysis data that combines annotations, predictions, and derived metrics into a single structured representation. All of this data is persisted and subsequently loaded into the dashboard. In the next section, we describe how these consolidated outputs are used to support interactive visualisations and analysis through the dashboard interface.

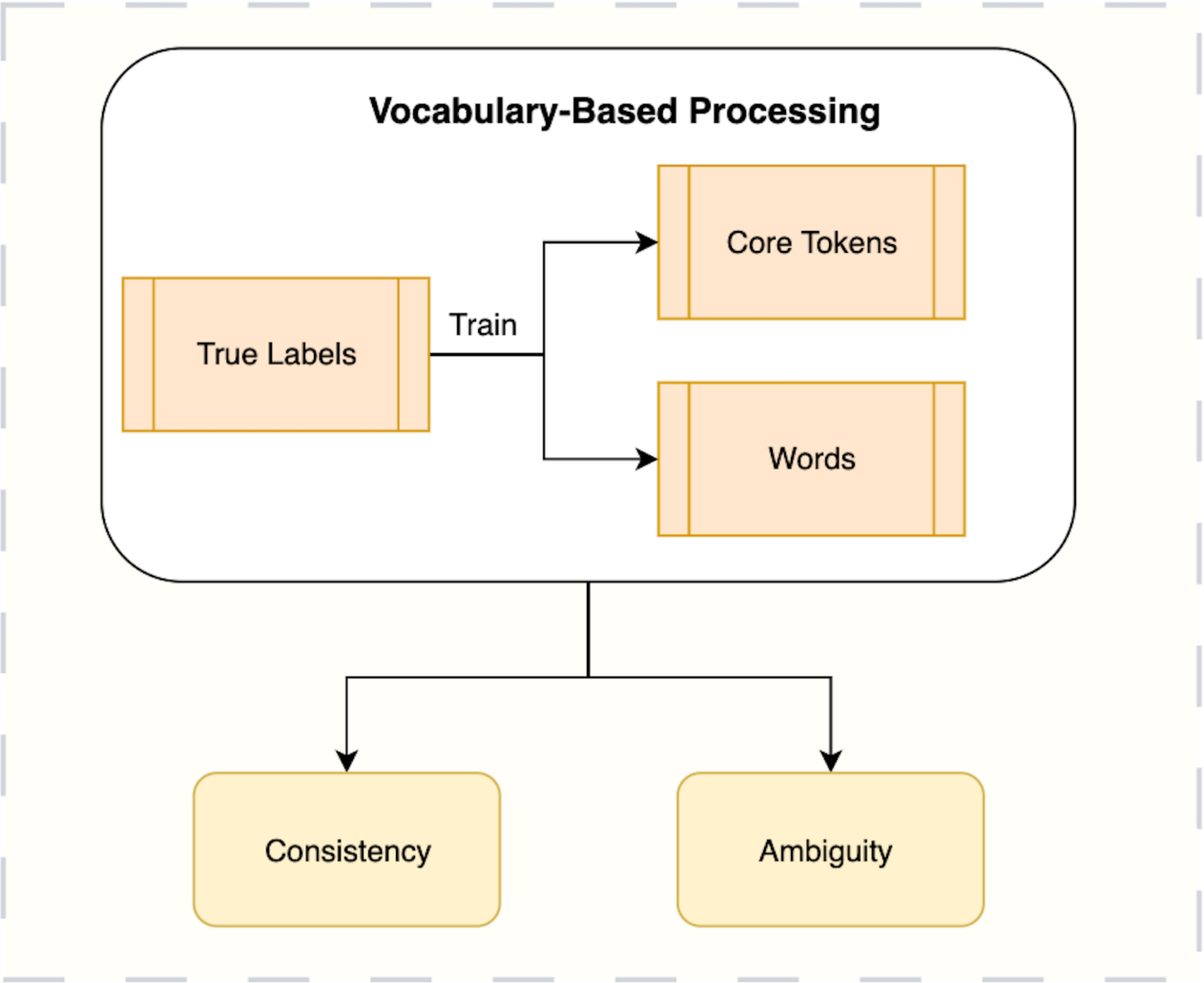

Figure 3.10: Overview of the processing used to analyse vocabulary-based interactions. Ambiguity and consistency are computed by linking each word or core token in the test set to its annotation instances in the training set. Orange indicates data-derived components; yellow indicates outputs shared across both cross-component and behavioural analysis.

## 3.4 Dashboard Phase

The DeformAr dashboard is the primary interface for exploring model and data behaviour across Arabic and English. It supports both cross-component and behavioural analysis through interactive visualisations, implemented using **Plotly Dash**. Building on the outputs extracted and processed in the previous phase, the dashboard allows us to inspect, compare, and analyse components of the NER pipeline across different levels of granularity — from global performance trends to individual token-level behaviours. The following sections describe the structure and design of the dashboard followed by an outline of the different types of analyses it supports.

### 3.4.1 Dashboard Overview

The DeformAr dashboard provides an interactive interface for exploring the outputs of the data extraction phase. It supports different modes of analysis by organising the content into four main tabs: a **Loading tab**, a **Cross-Component tab**, a **Behavioural Analysis tab**, and an **Instance-Level tab**. Each tab focuses on a specific aspect of the analysis and presents the data in a way that supports comparison across languages, data, and model components.

The data used in the dashboard is prepared and loaded by the **Data Manager**, a configuration-based module that manages how different outputs are structured and connected to the appropriate tab. The dashboard receives three main categories of output from the data extraction phase. The first is analysis data, which includes token-level predictions, tokenisation details, UMAP projections, clustering assignments, and derived metrics such as ambiguity and uncertainty. The second category includes a set of matrices used for similarity-based comparisons, such as attention similarity and centroid similarity. The final category contains evaluation results, including classification reports, confusion matrices, and error analysis.

Each output is integrated into the corresponding tab of the dashboard where it is most applicable. For instance, UMAP projections and clustering assignments are presented in the behavioural analysis tab, while classification reports and metric distributions are shown in the cross-component tab. Figure 3.11 illustrates how the processed outputs are organised and passed into the dashboard interface.

This design supports a gradual analysis workflow, allowing us to begin with high-level performance summaries and then explore more detailed outputs—such as attention patterns, uncertainty scores, and token-level behaviour—within a unified and consistent interface.

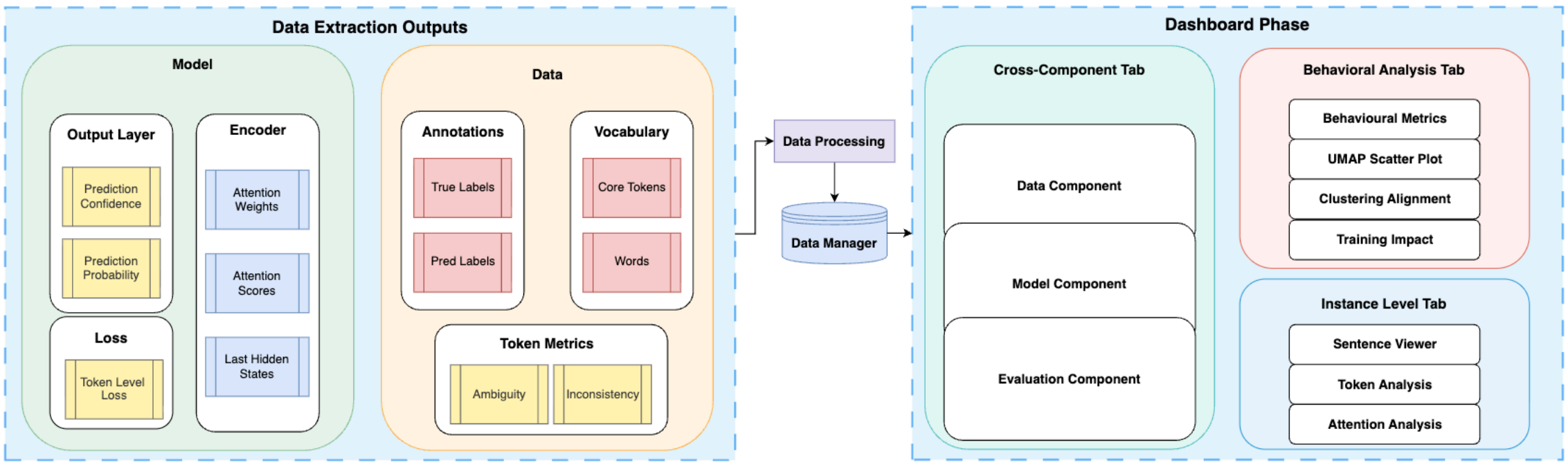

Figure 3.11: Overview of the dashboard structure in DeformAr. Outputs from the processing phase are grouped into analysis data, similarity matrices, and evaluation metrics. These are loaded through the Data Manager and displayed across the four main dashboard tabs.

### 3.4.2 Cross-Component Tab Overview

The Cross-Component Tab supports metric-based evaluation by providing high-level comparisons across model and data components. Its main purpose is to quantify discrepancies between Arabic and English using a diverse set of interpretable metrics. These metrics offer a quantifiable cross-component summary of system behaviour at both broad and fine-grained levels. For example, overall F1 performance enables broad comparisons between languages, while detailed error patterns provide more fine-grained insights.

The analyses displayed in this tab come from two sources: precomputed outputs from the extraction phase (e.g., confidence scores, ambiguity metrics, prediction probabilities), and dynamically generated summaries within the dashboard (e.g., label distributions, error breakdowns). Visualisations are structured to support side-by-side comparisons across languages and components.

This tab is organised into three analytical sections:

- **Data Component Section** — explores dataset-level factors such as annotation distributions, ambiguity, and consistency;
- **Model Component Section** — examines fine-tuning outcomes and prediction behaviour using metrics such as loss and confidence;
- **Evaluation Component Section** — reports token- and entity-level performance using precision, recall, and F1-score.

Together, these sections provide an interpretable overview of how each component contributes to performance differences across languages. Further details of these analytical sections are provided in Section 4.2.

The Cross-Component Tab is focused on **isolated analysis**. In practice, isolation refers to the examination of a single metric, rather than isolating a specific component. Each metric is analysed independently, even if it is measuring an interaction between components. For instance, ambiguity is derived from the relationship between vocabulary and annotation patterns, but is analysed here as a standalone metric. In contrast, interaction-based analysis—where multiple metrics are studied together—is the focus of the Behavioural Analysis Tab.

The tab follows a **top-down analysis** workflow. We begin with high-level summaries, such as overall model performance, and then drill down into more detailed aspects of system behaviour. The metrics in this tab serve two main purposes. First, they provide a summary of differences across components, forming a numerical characterisation of each language. Second, they help guide deeper investigation in the Behavioural Analysis Tab by identifying areas of potential difference for further visual and contextual analysis. The next section introduces the Behavioural Analysis Tab, which complements the metric-based summaries presented here by supporting interpretability-based analysis.

### 3.4.3 Behavioural Analysis Tabs Overview

The Behavioural Analysis Tabs support interpretability-based analysis by providing interactive visualisations of the model's internal representations and their relationship to various metrics calculated throughout the pipeline. It complements the Cross-Component Tab by enabling users to explore contextualised token representations, attention patterns, and other internal dynamics that help explain the quantitative patterns observed earlier.

This phase is implemented through two interactive tabs:

- **Behavioural Analysis Tab** — provides global, token-level exploration of representation patterns and behavioural metrics (e.g., loss, confidence, ambiguity);

- **Instance-Level Tab** — enables detailed inspection of individual sentences, entities, and prediction contexts, linking visual patterns to concrete examples.

Further details of these interactive tabs and their layout are provided in Section 5.3.

As discussed in Section 2.6.2, several interpretability techniques have been proposed in the literature. We adopt a combination of these, including: global explanation through UMAP projections of hidden states, local explanation via attention head analysis, and instance-level exploration inspired by example-based explanation and influence functions. Together, these techniques allow us to examine the model from different perspectives, uncovering how specific components behave and helping to interpret insights identified in the Cross-Component Tab.

In addition to interpretability techniques, the Behavioural Analysis Tab incorporates several numerical variables extracted during the data extraction phase, particularly those calculated at the token level—such as ambiguity, tokenisation rate, and silhouette score. These variables are viewed alongside 2D representation visualisations to ensure that interpretations are not based solely on UMAP structure. For example, regions of overlap in the UMAP view that align with low silhouette scores provide stronger evidence of poorly-separated clusters, offering both visual and numerical support. These metrics help support the visual interpretation, as they are calculated in the original high-dimensional space, while UMAP provides a visual summary in two dimensions.

This integration allows users to connect model behaviours with data properties across components. Users can filter by entity type, inspect where high-loss or low-confidence tokens appear in the representation space, and correlate these behaviours with factors such as annotation ambiguity, tokenisation patterns, and other structural properties. These techniques are connected through interactive visual analytics. Users can highlight a group of data points in one view (e.g., UMAP) and immediately see how they behave in others (e.g., loss, ambiguity). This interaction helps users explore relationships between the model, data, and predictions in a more integrated manner.

## 3.5 Related Work

This section positions the DeformAr framework within existing approaches to training, evaluation, and analysis of NER systems and visual analytics. Rather than comparing directly, we highlight key methodological choices made during the framework's design and how prior work relate to these decisions.

Several tools separate model processing from interpretability, including InterpreT (Lal et al., 2021) and T3-Vis (Li et al., 2021). InterpreT incorporates an offline phase where hidden states and attention matrices are extracted and processed using t-SNE projections. These extracted elements, referred to as collateral, are designed to support specific visualisation components such as token embedding exploration or attention behaviour summaries. T3-Vis, by contrast, focuses on training and fine-tuning dynamics, with a back-end that dynamically extracts hidden states, head importance scores, and attention values during or after fine-tuning. Extraction in T3-Vis is organised around high-level user tasks such as identifying mislabelled examples or monitoring training progress. In both cases, the extraction phases are linked to the needs of analysis or visualisation, rather than formalising a general framework for structured model and data extraction. In developing DeformAr, we adopt a system-first approach: defining explicit data and model components, structuring the extraction phase independently of any particular analysis task, and consolidating outputs into a reusable, extensible format. This design allows DeformAr to support both cross-component and behavioural analysis across languages and components. It is also worth noting that both tools support English-language classification tasks, and do not explicitly address token-level tasks such as NER.

T-NER (Ushio and Camacho-Collados, 2021) is a Python library designed to facilitate the training and evaluation of Transformer-based NER models, with a particular focus on cross-domain and cross-lingual generalisation. The system provides utilities for model fine-tuning, prediction, and evaluation across several datasets unified into a common format. While T-NER supports multilingual training and evaluation, it does not include a structured extraction phase, or perform detailed preprocessing across languages. DeformAr, by contrast, is focused on systematic analysis and introduces a structured extraction methodology that models the NER system as interacting components, supporting both cross-component and behavioural analysis.

In the extraction phase, we adopt first-subword alignment for mapping entity labels to subword-tokenised sequences. While alternative strategies, such as aggregating representations across all subwords, have been explored (Ács et al., 2021), the first-subword approach remains the most used in standard NER pipelines. Notably, Ács et al. (2021) report that first-subword alignment achieves the best results for NER tasks, particularly on benchmarks such as CoNLL-2003. In developing DeformAr, we therefore adopt the first-subword strategy to maintain consistency with prior work, ensure compatibility with standard evaluation and token classification frameworks, and avoid introducing aggregation that could complicate interpretability.

Finally, while most NER systems treat labels such as B-PER and I-LOC as individual units, alternative approaches have proposed to decompose labels into their span and type components. For example, Kato et al. (2020) introduce a method for learning embeddings of label components separately, showing that this improves performance on fine-grained NER tasks, particularly for low-frequency entity types. In developing DeformAr, we retain the standard flat label representation, following the conventions established in the CoNLL-2003 benchmark and HuggingFace's token classification pipeline. This decision ensures compatibility with standard evaluation approaches.

## 3.6 Discussion

This section discusses how the proposed methodology addresses **RQ0** and establishes the foundation for the remaining research questions. RQ0 concerns how NER systems can be decomposed into interpretable model- and data-level components to enable diagnostic, cross-lingual evaluation. The answer to this question is divided into two complementary aspects: a *conceptual* layer that defines the decomposition strategy, and a *technical* layer that implements it through the DeformAr extraction pipeline.

### 3.6.1 RQ0 Summary: Conceptual Aspect

At the conceptual level, DeformAr defines an interpretable decomposition of NER systems into two interacting components — *data* and *model*. This structure enables system-atic analysis of how linguistic and structural factors in the data interact with representational and behavioural dynamics within the model.

The framework examines different layers of the system pipeline, observing how both data and model components evolve through each stage. It identifies key subcomponents that capture these transformations and defines diagnostic metrics to quantify the resulting behavioural and structural changes. By examining these subcomponents jointly, DeformAr provides a basis for tracing how local variations — such as tokenisation, annotation, or contextual shifts — influence overall system behaviour. It also provides an extensible structure that allows new subcomponents and additional metrics to be incorporated as the analytical scope expands.

### 3.6.2 RQ0 Summary: Technical Aspect

At the technical level, DeformAr implements this conceptual design through an integrated yet modular pipeline that extracts, processes, and consolidates diagnostic outputs from both the data and model components. The implementation spans the entire NER workflow — from data preparation and model fine-tuning to output extraction and processing — ensuring that all relevant subcomponents are systematically represented.

The pipeline produces a unified set of diagnostic artefacts, including vocabulary statistics, annotation distributions, attention weights, hidden-state representations, and prediction confidences. These artefacts capture multiple layers of system behaviour and are pre-computed in a consistent format to support both cross-component and behavioural analysis.

Importantly, each module within the pipeline can be executed independently, allowing targeted re-computation or focused analysis without rerunning the full workflow. This modular structure makes DeformAr both extensible and reproducible: new metrics, subcomponents, or languages can be incorporated seamlessly while maintaining compatibility with the broader extraction and processing phases described in Sections 3.3.1–3.3.7.

### 3.6.3 Analytical Focus and Relation to Research Questions

This section establishes the analytical focus of the proposed methodology, clarifying what the DeformAr framework is designed to capture and how it supports the broader research questions of this thesis.

**Analytical Focus** As outlined in Section 2.4, various factors influence NER performance, including architectural variation, pretraining scale, and linguistic complexity. However, not all of these can be directly examined within DeformAr. The framework focuses on factors that are observable and measurable within the supervised NER pipeline — from preprocessing and tokenisation to prediction and evaluation. These include linguistic and structural challenges specific to Arabic and English, annotation-related effects such as ambiguity and consistency, dataset design characteristics, entity-type behaviour, and training dynamics reflected in loss, prediction patterns, and representational structure.

In particular, factors such as model size or the scale of pretraining data fall outside the scope of this work, since the models are not retrained from scratch nor are their underlying architectures modified. Instead, DeformAr targets the components of NER behaviour that can be empirically analysed and compared across languages, providing a controlled setting in which diagnostic differences can be interpreted.

**Relation to Research Questions** The framework integrates a range of diagnostic outputs extracted from both the data and model components. These outputs support two complementary modes of analysis: the **Cross-Component Tab**, which enables structured, side-by-side comparisons across languages and components, and the **Behavioural Analysis Tab**, which allows interactive inspection of internal model behaviour and its relation to data and model subcomponents. The **DeformAr** dashboard provides the methodological basis for addressing the remaining research questions, with each tab corresponding to a distinct analytical focus:

- **RQ1–RQ4** quantify the discrepancies between Arabic and English NER systems across data, model and evaluation components, and their interactions — addressed through the **Cross-Component Tab**.

- **RQ5** investigates how interpretability and visual analytics can be used to explain and visualise these differences — addressed through the **Behavioural Analysis Tab**.

Together, these questions define the analytical scope of the thesis. DeformAr enables their investigation through its combination of metric-based cross-component analysis and interactive visual exploration. The following two chapters apply this methodology in practice — first through quantitative cross-component analysis (Chapter 4), and then through interpretative behavioural analysis (Chapter 5).

## 3.7 Contributions

This chapter introduced the methodological foundations of the **DeformAr** framework, which defines how NER systems can be decomposed, extracted, and analysed across model and data components. The key contributions of this chapter are summarised below.

**First: A structured, system-level extraction phase.**
DeformAr formalises the extraction and processing of diagnostic artefacts as an independent, system-level procedure, decoupled from any single evaluation or interpretability method. This enables comprehensive and reproducible analysis of model and data behaviour across languages and components.

**Second: A multilingual and cross-lingual diagnostic framework.**
The framework applies consistent extraction, fine-tuning, and evaluation procedures to both Arabic and English NER systems, addressing a key limitation of prior interpretability tools that focus primarily on English. To our knowledge, DeformAr represents the first comprehensive interpretability framework explicitly designed for Arabic NER.

**Third: Integration of cross-component evaluation and behavioural analysis.**
DeformAr unifies quantitive cross-component evaluation and behavioural analysis within a single extensible dashboard, allowing seamless transition between metric-based comparisons and interpretability-driven exploration. This integration provides a comprehensive view of model behaviour across components, layers, and languages.

**Fourth: Component and subcomponent modelling for diagnostic decomposition.**
Each language is represented as two high-level components—data and model—further divided into subcomponents such as vocabulary, annotations, encoder representations, and output predictions. Extraction strategies are guided by the relationships between these subcomponents, enabling both isolated inspection and cross-component interaction analysis. This structure supports fine-grained investigation of how component behaviours influence one another across training, fine-tuning, and evaluation.

**Fifth: Extensibility beyond NER and across tasks.**
While this case study focuses on Arabic–English NER, the framework's decomposition strategy and pipeline design are task-agnostic. The same approach can be applied to other sequence-labelling tasks such as part-of-speech tagging or morphological analysis. Moreover, the framework's modular structure allows additional subcomponents, metrics, or interactive visualisations to be incorporated as needed.

Together, these contributions provide a structured and extensible approach for investigating cross-linguistic discrepancies in supervised NER performance and for applying interpretability techniques in a multilingual, token-level analysis. In doing so, this chapter addresses **RQ0** by providing both the conceptual decomposition and the technical implementation required to support the analyses developed in later chapters. The next chapter applies these methods in practice, beginning with the cross-component evaluation of NER systems across Arabic and English.

# Cross-Component Analysis 4

## 4.1 Introduction

This chapter represents the first phase of the DeformAr dashboard, corresponding to the Cross-Component Analysis Tab, and addresses Research Questions RQ1 to RQ4. Its goal is to investigate and quantify the performance discrepancy between Arabic and English NER systems by systematically examining dataset characteristics, model behaviours, and evaluation metrics, as well as their interactions. While existing work often reports NER performance using overall scores such as F1, these aggregate metrics alone cannot explain why some languages perform worse than others. By jointly analysing data, model, and evaluation components, this chapter provides a structured quantitative foundation for understanding the factors that contribute to the observed cross-lingual performance gap. As outlined in Chapter 3, this analysis is conducted as a comparative case study applying DeformAr to Arabic and English NER, using one model and one dataset per language to enable focused diagnostic comparison.

The analysis is organised into three main sections. The first focuses on the data component (RQ1), which examines characteristics of the input data—such as lexical diversity, annotation distribution, and the effects of tokenisation. The second is the model component (RQ2), which focuses on the representation space, output layer, and loss function, comparing prediction confidence, loss, and silhouette scores to understand and quantify model behaviour and its interaction with the data. The third is the evaluation component (RQ3), comparing model predictions against true labels for Arabic and English using token-level and entity-level standard evaluation metrics, prediction outcomes, and error analysis.

Finally, we address RQ4 by providing a cross-component analysis that connects data characteristics, model behaviours, and evaluation metrics, building a structured narrative to explain performance discrepancies across languages at the entity-span level. By systematically analysing each of these components—starting with data characteristics and model behaviours, and gradually narrowing the focus to how these are reflected in evaluation metrics—this chapter builds a detailed, evidence-based understanding of the performance gap between Arabic and English NER. These findings provide direction for the next chapter, which explores RQ5 through interpretability and behavioural analysis.

Building on the system design introduced in Chapter 3, this chapter implements the first analytical phase of the DeformAr dashboard — the **Cross-Component Analysis Tab**. The tab is organised into three diagnostic components — **Data**, **Model**, and **Evaluation** — each corresponding to one of the analytical stages outlined above. Within each component, the dashboard integrates precomputed outputs from the extraction phase (e.g., ambiguity scores, classification metrics) with dynamically generated summaries (e.g., label distributions, correlation patterns) to support side-by-side cross-lingual comparison. These analyses are presented through interactive visualisations such as bar charts, heatmaps, and correlation matrices, providing a quantitative foundation for the behavioural and interpretability analysis explored in Chapter 5. Further details about the tab structure and its modular layout are provided in Section 4.2.

**Chapter Outline** The remainder of this chapter is organised as follows:

- **Section 4.2 — Tab Layout**
  Describes the architecture of the Cross-Component Analysis Tab, including the modular configuration of its sections, control panels, and visualisation canvases. This section explains how the tab facilitate the analyses that follow.

- **Section 4.3 — Data Component**
  Addresses RQ1 by examining the structure of the input data, including annotation patterns, tokenisation effects, and ambiguity metrics. The analysis focuses on the **vocabulary** subcomponent and its interactions with both **annotations** and the **tokenisation** subcomponents.

- **Section 4.4 — Model Component**
  Addresses RQ2 by analysing model behaviour through prediction confidence, loss, and silhouette score. This section focuses on the **output layer** and **representation** subcomponents of the model.

- **Section 4.5 — Evaluation Component**
  Addresses RQ3 by comparing true and predicted labels to quantify performance discrepancies between Arabic and English. This section focuses on the interaction between the **annotation** (true labels) subcomponent of the data and the **output layer** (predicted labels) subcomponent of the model.

- **Section 4.7 — Discussion**
  Summarises key findings from the cross-component analysis and identifies areas of interest for deeper exploration through interpretability and behavioural analysis in Chapter 5.

## 4.2 Tab Layout

The **Cross-Component Analysis Tab** represents the first analytical tab within the DeformAr dashboard phase. It provides a dynamic interface for examining how the **data**, **model**, and **evaluation** components interact. Figure 4.1 illustrates the overall architecture of the tab, showing the three analytical components—Data, Model, and Evaluation.

Each section of the tab follows the same internal layout structure, dynamically generated from a configuration file that defines three key elements: the *Section Title*, the *User Controls*, and the *Canvas*. This modular design ensures consistency and flexibility across all analytical components. Figure 4.2 illustrates this layout, showing how the control panel, title area, and canvas are generated within the Data section.

The **User Controls** panel allows users to select the analytical focus (e.g., structural analysis, ambiguity, or evaluation metrics) and the visualisation plot to display. Once a configuration is selected, pressing the *Render Canvas* button dynamically generates the corresponding plots and tables within the Canvas area.

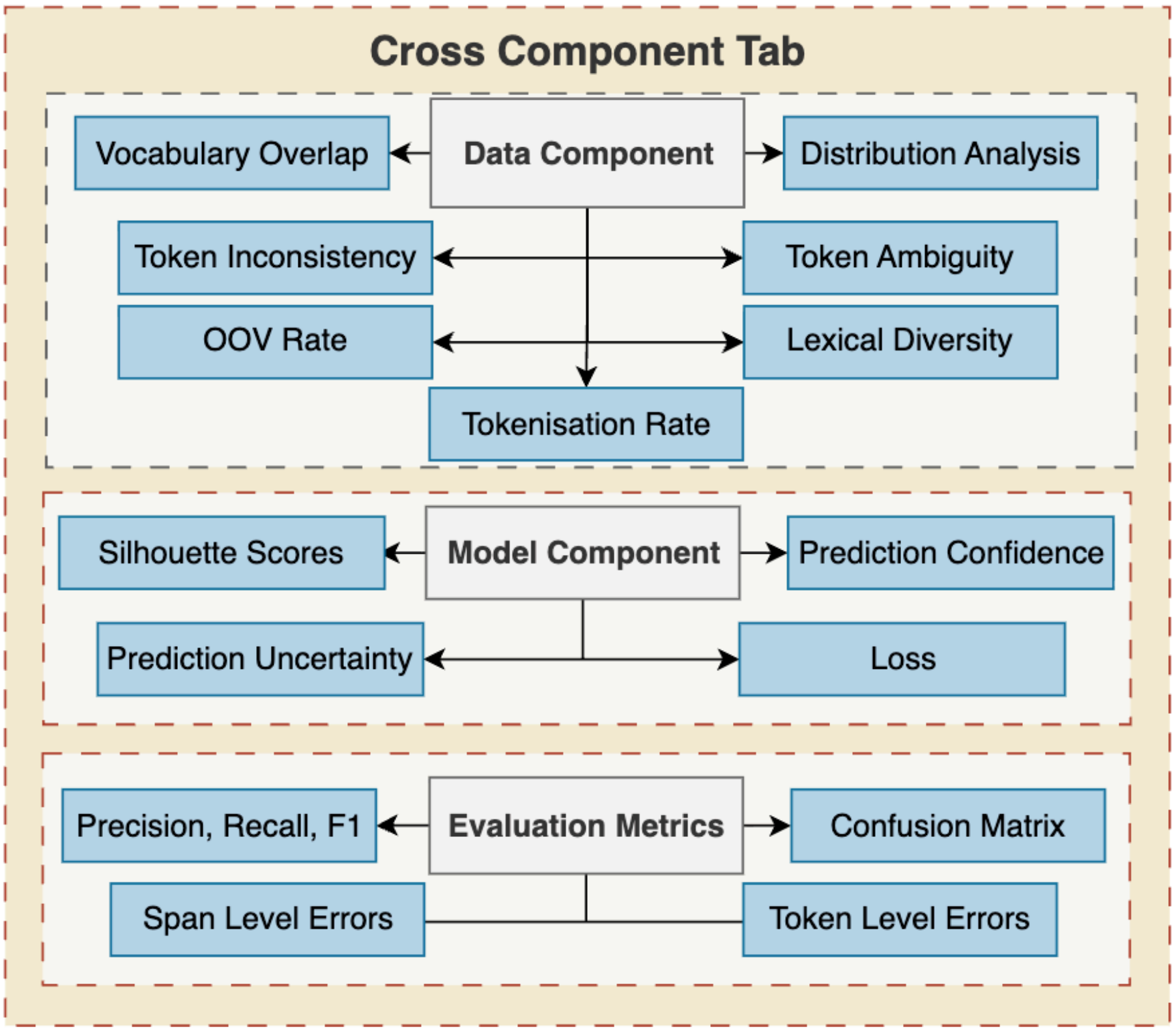

Figure 4.1: Architecture of the Cross-Component Analysis Tab, illustrating the data, model, and evaluation components and their analytical submodules. White boxes indicate the individual sections within the tab, and arrows represent analytical dependencies across components.

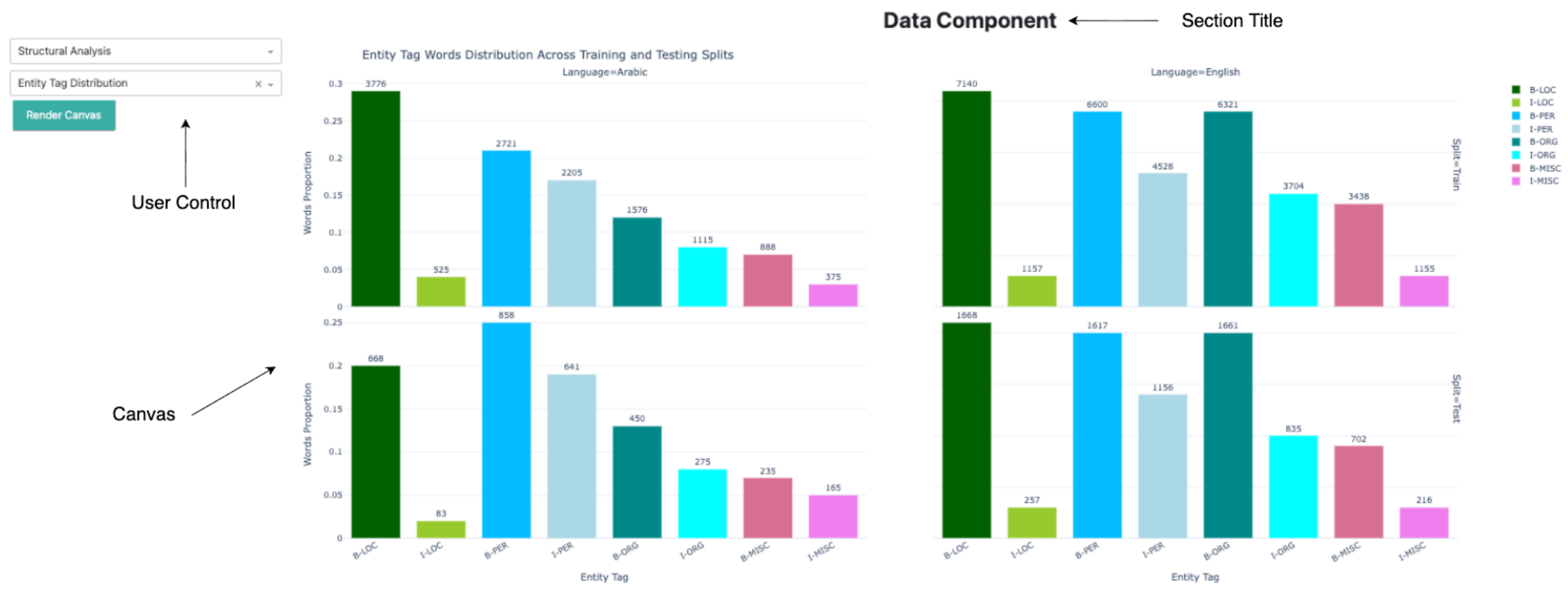

Figure 4.2: Layout of the Data Section in the Cross-Component Analysis Tab, showing the relationship between the *User Controls*, *Section Title*, and *Canvas* areas. Each component (Data, Model, Evaluation) provides its own analytical options through the control panel.

The **Canvas** acts as the main analytical surface, rendering both pre-defined and dynamic computations. Depending on user selection, the Canvas displays bar charts, heatmaps, or correlation plots representing the chosen analytical layer. All visualisations are automatically updated when parameters in the control panel are modified.

Together, the **User Controls**, **Section Titles**, and **Canvas** define the modular layout of the Cross-Component Analysis Tab. This tab provides structured analysis for each component. The specific analytical outputs and results for each component are discussed in detail in the following sections of this chapter.

## 4.3 Data-Component

This section addresses **Research Question 1 (RQ1)** by analyzing the characteristics of the ANERCorp (Arabic) and CoNLL-2003 (English) datasets that may influence the performance of NER models. We focus on the interaction between vocabulary and gold-standard annotations across training and testing splits, while also considering tokenization's effect on this interaction.

Our analysis is structured around three key perspectives: **structural characteristics**, **lexical characteristics**, and **token-level behavioural characteristics**. The first two dimensions examine dataset-level properties, while the third focuses on token-level variability in annotation patterns and tokenization. Together, these perspectives offer complementary insights into the underlying properties of the datasets and how they might influence NER model performance.

**Structural characteristics** relate to the size and distribution of tokens, annotations, and entity spans within the data. **Lexical characteristics** explore vocabulary-related behaviour such as lexical diversity, entity tag overlap, and out-of-vocabulary (OOV) rates. **Token-level behavioural characteristics**, on the other hand, capture annotation consistency, ambiguity, and tokenization rate at the level of individual tokens. While lexical and structural metrics describe dataset trends, behavioural metrics serve as token-level indicators that help establish connections between dataset properties and model behaviour.

We also examine how **tokenization** affects both the lexical structure and annotation behaviour at the token level. Tokenization transforms word representations into core tokens, which can impact lexical diversity, introduce new token types, and changes in how annotations are distributed across tokens. These effects are captured through our **token-level behavioural metrics**, including tokenization rate, annotation ambiguity, and consistency. By analyzing structural, lexical, and behavioural properties—before and after tokenization—we aim to identify dataset characteristics that may help explain differences in NER model performance across languages.

To address RQ1, we consider the following aspects: **a)** What structural discrepancies exist between the datasets, including differences in token-level and entity-level annotation distributions, as well as word and token counts across training and testing splits; **b)** How the annotation scheme (e.g., IOB1 vs. IOB2) affects the structure and distribution of entity spans across splits; **c)** What differences exist in lexical behaviour across languages, including lexical diversity, entity tag overlap, and out-of-vocabulary (OOV) rates; **d)** How tokenization transforms lexical and structural properties of the data, particularly in relation to subword tokenisation and tag alignment; and **e)** What token-level behavioural metrics—such as annotation ambiguity, label consistency, and tokenization rate—can reveal about the interaction between structural and lexical characteristics and their implications for model performance.

**Extracted Outputs** Figure 4.3 illustrates the outputs extracted during the data extraction phase that support this analysis. All outputs are derived from the interaction between the vocabulary and gold-standard annotations, analysed both before and after tokenization. These outputs directly correspond to the aspects outlined above. The **count-based outputs** support the analysis of structural characteristics (aspects **a** and **b**), including token-level and entity-level annotation distributions, as well as word and core-token counts across training and testing splits. The **type-based outputs** reflect lexical characteristics (aspects **c** and **d**), such as lexical diversity, entity tag overlap, and out-of-vocabulary (OOV) rates. While both structural and lexical metrics are dynamically computed during the dashboard-based analysis, certain **token-level behavioural metrics**—including **tokenization rate**, **ambiguity**, and **annotation consistency**—are pre-computed during the data extraction phase to support aspect **e**. Together, these outputs provide a multi-level view of dataset properties and how they may affect NER model performance across languages.

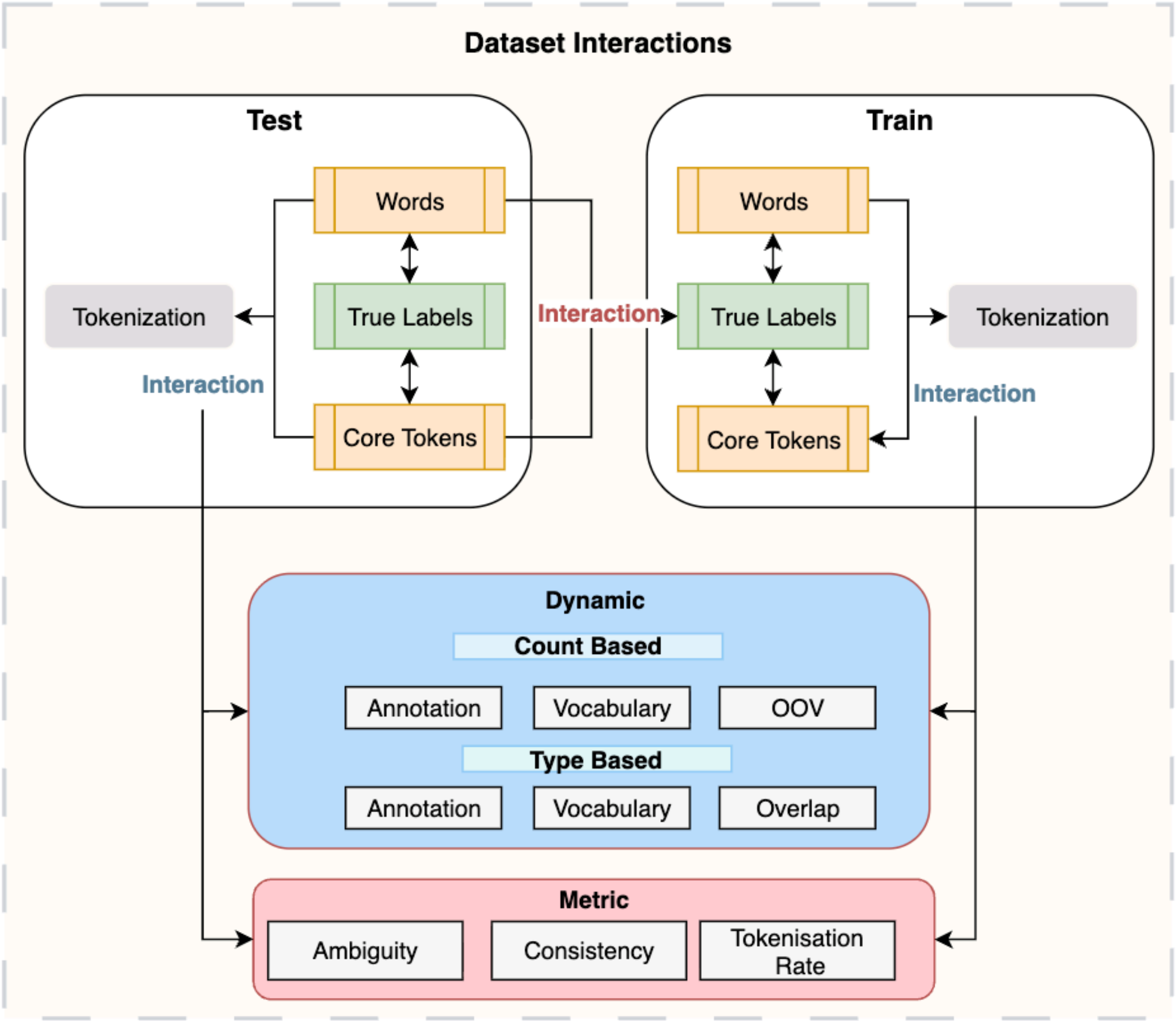

Figure 4.3: Dataset Interaction Pipeline for NER Analysis. This diagram illustrates the flow of data through tokenization and the extraction of **count-based** (structural) and **type-based** (lexical) features from the training and testing splits. These features are used to compute **dynamic metrics** (e.g., lexical diversity, OOV) within the dashboard, and **pre-computed token-level behavioural metrics** (e.g., ambiguity, consistency, and tokenization rate), which are used to analyse dataset characteristics and support downstream analysis.

This section follows a top-down approach to analyzing dataset characteristics—starting with high-level structural patterns and narrowing to fine-grained lexical and token-level behaviours. We begin with a conceptual and technical overview of the metrics used to quantify structural, lexical, and behavioural aspects of the data. These are then applied across datasets to extract quantifiable insights. The analysis is presented in three stages: structural analysis, lexical analysis, and token-level behavioural analysis, with the impact of tokenization considered throughout each stage.

**Section Outline** This section is structured as follows:

- **Metric Definitions and Analysis Approach:** Overview of the structural, lexical, and behavioural metrics used to characterize the datasets, along with how each metric is computed and applied in the analysis.
- **Structural Analysis:** Comparison of token- and entity-level annotation distributions and counts across splits and annotation schemes.
- **Lexical Analysis:** Examination of vocabulary-based metrics such as lexical diversity, entity tag overlap, and OOV, before and after tokenization.
- **Token-Level Behavioural Analysis:** Exploration of annotation patterns through token-level metrics such as tokenization rate, ambiguity, and consistency, highlighting their role in understanding dataset variability and affecting model evaluation.

### 4.3.1 Metric Definitions and Analysis Approach

This section introduces the key metrics used to analyse structural, lexical, and token-level behavioural characteristics of the NER datasets, and outlines how these metrics are applied to support the dataset-level analysis. Our analysis considers annotations at both the token and entity levels and includes metrics that quantify annotation patterns and the impact of tokenization. The metrics presented here are the basis for the structural, lexical, and behavioural analyses that follow.

**Annotation Structure** To assess the structural properties of the datasets, we examine the distribution of entity types at both the token and entity levels. At the token level, each word is annotated with an entity tag (e.g., B-LOC, I-PER), capturing its position and type. From these annotations, entity-level spans are constructed by applying an annotation schemes (e.g., IOB1, IOB2), grouping sequences of tagged tokens into a span. As discussed in Section 2.3.3, the annotation scheme plays an important role in how entity spans are detected and therefore influences their distribution. We measure these aspects across both training and testing splits.

**Lexical Structure** As described in Figure 4.3, vocabulary metrics are calculated based on two representations: *words* (pre-tokenization) and *core tokens* (post-tokenization), which combine words and their first subword tokens (see Section 3.3.4). Across both representations, we compute three key lexical metrics: lexical diversity, overlap, and out-of-vocabulary (OOV) rate. Before defining these metrics, it is important to clarify that a *token type* is the set of unique tokens in a dataset, irrespective of frequency. Here, "token" refers either to a full word (pre-tokenization) or a core token (post-tokenization), depending on the tokeniser's outcome. These lexical metrics are calculated at two levels of granularity. First, we calculate an overall value across the entire dataset to provide a high-level view of lexical structure. Second, we break down each metric by entity tag within the token-level annotations, across both training and testing splits. This enable us to assess how lexical behaviour varies across entity tags and between splits, offering a more fine-grained understanding of annotation patterns. **Lexical Diversity** captures the richness and variability of a dataset's vocabulary. It is measured as the ratio of unique token types to the total number of tokens:

$$\text{Lexical Diversity} = \frac{\#\ \text{Unique Types}}{\#\ \text{Total Tokens}} \tag{4.1}$$

A higher lexical diversity indicates a broader range of vocabulary usage, which can reflect more varied linguistic structures or domain coverage. **Overlap** measures the proportion of token types that are assigned to multiple entity tags across the dataset. It provides insight into potential annotation ambiguity or inconsistency where the same token type appears under different entity labels (e.g., tagged as both B-LOC and B-ORG). High overlap may introduce confusion during model training and evaluation. **OOV rate** measures the proportion of token types in the test set that do not appear in the training set. It captures the relationship between the tokens across splits. This metric reflects the dataset's generalization capacity: a high OOV rate suggests that many tokens encountered during testing were unseen during training, potentially affecting model performance.

**Token-Level Behavioural Metrics** In addition to structural and lexical metrics, we introduce a set of **token-level behavioural metrics** that serve as diagnostic indicators of annotation patterns and tokenization effects. These include: **(1) Consistency** — the alignment of token labels across training and test splits; **(2) Ambiguity** — annotation uncertainty quantified via entropy; and **(3) Tokenization Rate** — the degree to which words are split into subwords during preprocessing.

Whereas lexical metrics capture broad dataset properties, behavioural metrics operate at the level of individual tokens. This granularity allows them to link dataset characteristics with model outcomes: in this chapter, they support the cross-component analysis by aggregating token-level scores for ambiguity, inconsistency, and tokenisation, and reporting the mean score for each entity tag (e.g. B-LOC, I-PER). In Chapter 5, they are revisited in the behavioural analysis stage to draw more detailed connections between data properties and model behaviour at the token level.

**Tokenisation Rate** Tokenisation rate measures the extent to which individual words are segmented into subword units by the tokenizer. For each word, it is defined as the number of tokens produced by the tokenizer. If a word remains whole, its tokenisation rate is 1 (i.e., one word equals one token). If the word is split into subwords, the tokenisation rate increases accordingly. At the dataset level, we calculate the average tokenisation rate as the ratio between the total number of tokens and the total number of words:

$$\text{Tokenisation Rate} = \frac{\#\text{Total Tokens}}{\#\text{Total Words}} \tag{4.2}$$

This metric provides insight into the tokenisation granularity of the dataset, reflecting how frequently words are tokenised. Higher tokenisation rates may change the lexical structure of the data and can affect model performance—particularly in tasks like NER. Further discussion of these implications is provided in the tokenisation impact analysis.

**Token Ambiguity** Token ambiguity captures the uncertainty in how individual tokens are annotated across different contexts. In this analysis, ambiguity is quantified using Shannon entropy, which measures the unpredictability of the distribution of entity labels assigned to a token. A high entropy value indicates that a token has been annotated with a diverse set of entity tags, suggesting either inherent linguistic ambiguity or annotation inconsistency. Token ambiguity measures the relationship between each token in the test split and its annotation patterns in the training split. For each token in the test set, we locate all instances of that token in the training data and examine their associated annotations. Ambiguity is computed only for tokens that appear in the training set, while tokens that are OOV in the test split are assigned a placeholder entropy score of $-1$, since no prior annotation distribution exists for them. For example, consider the token `European`, which appears 22 times in the ANERCorp training split—15 times labelled as `I-ORG` and 7 times as `O`. The probabilities for each tag are computed as:

$$P(\text{I-ORG}) = \frac{15}{22} \approx 0.68, \quad P(\text{O}) = \frac{7}{22} \approx 0.32$$

These probabilities are then used in Shannon's entropy formula to calculate annotation uncertainty:

$$H(X) = -\sum_{i=1}^{n} P(x_i) \log P(x_i) \tag{4.3}$$

where $P(x_i)$ represents the probability of each entity tag $x_i$ from a set of $n$ tags. The entropy contributions for each tag are:

$$H(\text{I-ORG}) = -P(\text{I-ORG}) \log(P(\text{I-ORG})), \quad H(\text{O}) = -P(\text{O}) \log(P(\text{O}))$$

Summing these gives the overall **token entropy**:

$$H(\text{European}) = 0.376 + 0.525 \approx 0.901$$

This entropy score reflects annotation ambiguity—higher values indicate greater variability in how a token has been labelled across its training instances. Whereas earlier we used the term token to refer to both words (pre-tokenisation) and core tokens (post-tokenisation), here we distinguish between the two. We compute entropy at both levels, referring to the word-level measure as *Word Ambiguity* and the core-token-level measure as *Token Ambiguity*. This distinction allows us to examine how tokenisation influences annotation stability.

**Token Consistency** Token consistency captures how consistently a token's annotation in the test split aligns with its labelling in the training split. For each token in the test set, we examine its annotation patterns in the training set. If the token does not appear in the training split, its consistency and inconsistency ratios are set to zero. For tokens found in both splits, we compute: (1) the total occurrences in the training set, (2) the consistency count, the number of times the token was labelled with the same entity tag as in the test instance, and (3) the inconsistency count, the number of times the token received a different tag in training. Normalising by the total number of training occurrences yields the consistency ratio and inconsistency ratio, which proportionally capture the alignment between training and test annotations.

For example, the token بن ("the son of") appears 114 times in the ANERCorp training set: 103 times as I-PER, 8 as B-PER, and once each as I-ORG, O, and I-MISC. If this token is labelled as I-PER in a test instance, its consistency ratio is $\frac{103}{114} \approx 0.90$ and its inconsistency ratio is $\frac{11}{114} \approx 0.10$. Conversely, if it appears as B-PER in the test, the consistency and inconsistency ratios are $\frac{8}{114}$ and $\frac{106}{114}$, respectively.

These metrics highlight alignment patterns between training and test annotations. High consistency suggests that a token tends to appear in stable contexts and is reliably annotated with the same tag, whereas high inconsistency may indicate either contextual variation or inconsistent annotation practices. We compute these metrics for both words and core tokens, referring to them as *Word Consistency* and *Token Consistency*, respectively. This distinction allows us to examine how tokenization influences annotation stability across splits.

Although presented individually, these behavioural metrics are closely related. A high tokenisation rate may lead to increased annotation ambiguity, especially if the resulting core tokens show greater entropy. Similarly, the interaction between ambiguity and consistency can help diagnose the source of variability: if tokens display both high ambiguity and high inconsistency, this may reflects annotation inconsistency rather than genuine contextual variation. These relationships help us better understand how structural and lexical factors interact, and how tokenisation shapes these dynamics, as discussed further in Section 4.7.

### 4.3.2 Structural Analysis

To address aspects (a) and (b) of RQ1—*(a) what structural discrepancies exist between the datasets, and (b) how annotation schemes affect entity span distributions*—we compare the size and annotation distributions of the Arabic (ANERCorp) and English (CoNLL-2003) datasets. We study structural differences such as the number of tokens and the distribution of entity tags and spans, as outlined below.

We analyse the size of each dataset and assess how tokenization affects it. This includes comparing word-level and core-token-level distributions. We then examine the distribution of token-level and entity-level annotations using distribution-based analysis across training and test splits. To understand the influence of annotation schemes, we also analyse span-level distributions and how they are represented in each dataset. While tokenization does not affect the raw annotation counts (e.g., the number of B-LOC tags remains unchanged), it can influence other structural aspects, as shown in Section 4.3.3.

**Dataset Size Discrepancy** To examine structural differences in dataset size, we compare the total word count and total number of named entity (NE) words in ANERCorp (Arabic) and CoNLL-2003 (English). As shown in Table 4.1, CoNLL-2003 is significantly larger, containing over 250,000 words compared to 150,000 in ANERCorp. It also includes more NE words, with over 42,000 compared to just 16,500 in ANERCorp. In proportion, named entities make up 16.86% of the CoNLL-2003 corpus, compared to just 11.03% in ANERCorp. These discrepancies suggest that the English dataset provides more training in terms of both overall size and entity-specific examples.

Table 4.1: Total Word Counts, Named Entity Word Counts, and NE Proportions for High-level Dataset Size Comparison.

| Dataset | Total Words | NE Words | NE Proportion |
|---|---|---|---|
| ANERCorp | 150,110 | 16,556 | 11.03% |
| CoNLL-2003 | 250,056 | 42,155 | 16.86% |

To complement the overall dataset size analysis, we examine the training and testing splits of ANERCorp and CoNLL-2003 to assess whether the same trends persist across splits. As shown in Table 4.2, CoNLL-2003 remains larger in both splits, with over 203,000 words in training and 46,000 in testing, compared to 125,000 and 25,000 in ANERCorp, respectively. The English dataset also contains more named entity (NE) words in each split—over 34,000 in training and 8,000 in testing—compared to 13,000 and 3,300 in ANERCorp. This supports the earlier observation that CoNLL-2003 offers more overall and entity-specific coverage during training and evaluation.

Table 4.2: Word and Named Entity Word Counts across Training and Testing Splits.

| Metric | ANERCorp | | CoNLL-2003 | |
|---|---|---|---|---|
| | **Train** | **Test** | **Train** | **Test** |
| Total Words | 125,102 | 25,008 | 203,621 | 46,435 |
| NE Words | 13,181 | 3,375 | 34,043 | 8,112 |
| NE Proportion | 10.54% | 13.50% | 16.72% | 17.47% |

**Tokenisation Impact On Dataset Size** To understand how preprocessing and tokenisation affect dataset size, we examine changes in vocabulary counts across training and test splits for ANERCorp and CoNLL-2003. Tokenisation can change dataset size by splitting words into subwords or removing words during preprocessing. These changes may affect the number of tokens used during model training, even if entity-level annotations remain structurally consistent.

Table 4.3 summarises the difference between the total number of words (pre-tokenisation) and tokens (post-tokenisation). In CoNLL-2003, the total number of words and tokens remains unchanged, indicating that tokenisation does not affect dataset size. In contrast, ANERCorp experiences a slight reduction in token count due to preprocessing: 443 words are removed in the training split, and 17 in the test split. All removed words belong to the non-entity (”O”) class and therefore do not impact the distribution of entity annotations.

Table 4.3: Comparison of Word and Token Counts across ANERCorp and CoNLL-2003 for Training and Testing Splits. Words refer to pre-tokenisation units; tokens refer to post-tokenisation results.

| Metric | ANERCorp | | CoNLL-2003 | |
|---|---|---|---|---|
| | **Train** | **Test** | **Train** | **Test** |
| Total Words | 125,102 | 25,008 | 203,621 | 46,435 |
| Total Tokens | 124,659 | 24,991 | 203,621 | 46,435 |

In summary, our analysis of dataset size revealed that CoNLL-2003 is consistently larger than ANERCorp, both in total word count and in the number of NE words. This discrepancy holds across training and testing splits, suggesting that the English dataset provides a higher volume of training instances overall, as well as richer entity-specific content. While tokenisation does not alter CoNLL-2003, ANERCorp experiences a small reduction in token count due to preprocessing. However, all removed words belong to the non-entity ("O") class, meaning the structure of the entity annotations remains intact.

We now turn to annotation-specific structural differences. Our analysis begins at the *token level*, where we examine how entity tags are distributed across the datasets. This is followed by an *entity-level* analysis, where we explore span-based annotation characteristics. In addition to comparing annotation distributions, we also investigate how annotation schemes influence the structure and representation of entity spans across the two datasets.

**Token-Level Annotation Distribution** To examine how named entity tags are distributed within the datasets, we analyse token-level annotation proportions across training and testing splits. While previous analysis showed that entity words constitute 11.03% of ANERCorp and 16.86% of CoNLL-2003, this analysis focuses on the breakdown of the entity tags themselves.

Figure 4.4 presents the distribution of entity tags, normalised as proportions of total entity words. Absolute counts are also shown above each bar to allow interpretation from both relative and raw perspectives. The most frequent tags in both datasets are B-PER and B-LOC, followed by I-PER and B-ORG, although the ordering differs between datasets. For example, in ANERCorp, I-PER tags occur more frequently than B-ORG tags, whereas in CoNLL-2003, B-ORG tags are more frequent than I-PER.

Overall, entity tags in CoNLL-2003 are more evenly distributed, with comparable frequencies across B- and I- tags. In contrast, ANERCorp shows more variation, with certain tags like I-PER being proportionally dominant, especially in the test split. This suggests that person names in Arabic are more often multi-token entities. Conversely, the lower frequency of I-LOC in both datasets implies that location names tend to be single-token. These structural differences, especially in span length and segmentation,

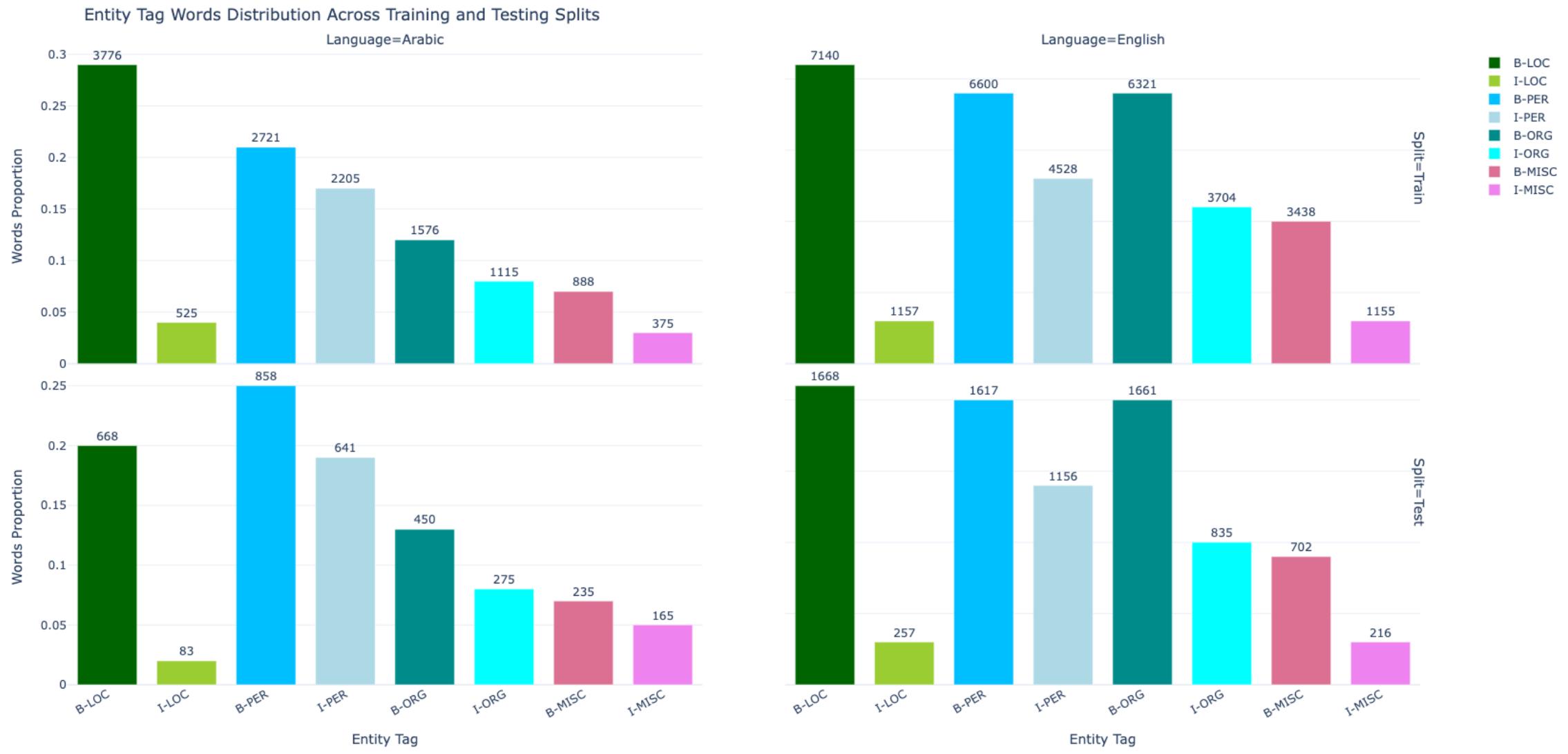

Figure 4.4: Distribution of entity tags across training and testing splits in Arabic (ANERCorp) and English (CoNLL-2003). Bars represent the proportion of each entity tag relative to the total number of named entity tags in the respective split. Absolute tag counts are displayed above each bar to provide counts alongside normalised proportions.

are explored further in entity level analysis.

**Entity-Level Annotation Distribution** We now shift to investigate the distribution and complexity of entity spans across both datasets under two annotation schemes: IOB1 and IOB2. As discussed in Section 2.3.4, these schemes are applied following the `Seqeval` evaluation logic, which defines how spans are reconstructed from token-level annotations. IOB1 follows a repair mechanism, where an `I-` tag without a preceding `B-` is treated as a continuation of the previous entity. In contrast, IOB2 follows a discard mechanism, ignoring such improper sequences entirely. This span-based analysis allows us to assess the influence of annotation conventions and their alignment with the datasets, directly addressing aspect (b) of RQ1.

Figure 4.5 presents entity-level span distributions in ANERCorp and CoNLL-2003 across training and testing splits, under two annotation schemes: IOB1 and IOB2. Both absolute counts and normalized proportions are shown for each entity type (LOC, ORG, PER, MISC). Presenting both absolute and proportional values enables comparison across datasets of different sizes. Two key findings emerge from this analysis.

First, the annotation scheme has a different impact across datasets. CoNLL-2003 remains unaffected when switching from IOB1 to IOB2, showing no change in entity span counts. In contrast, ANERCorp exhibits a small but notable reduction in total entity spans under IOB2. This reduction stems from annotation inconsistencies where some entity spans do not align with the IOB2 transition rules. A total of 126 entity spans are discarded from ANERCorp due to violations of IOB2 constraints, with 84 instances removed from the test split and 42 from the training split. Table 4.4 provides a detailed breakdown of the discarded spans by entity tag across both splits.

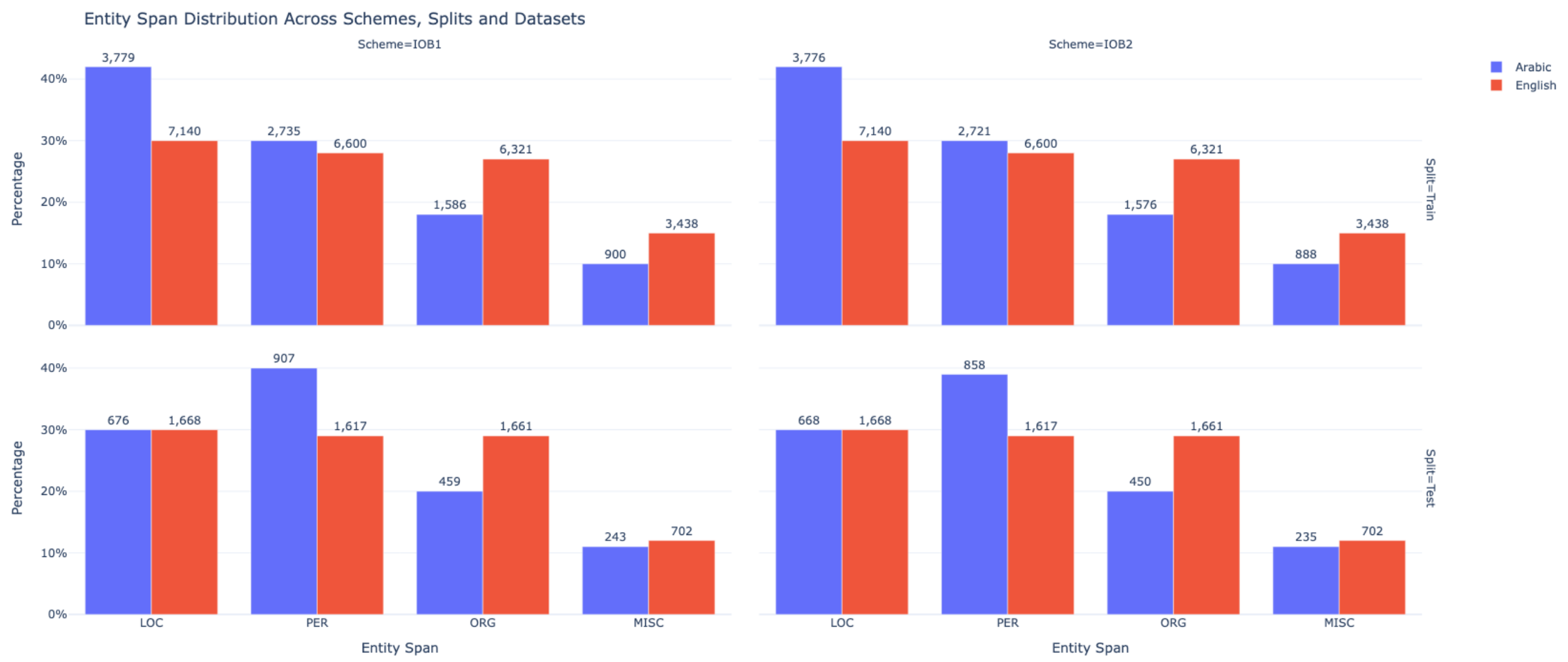

Figure 4.5: Entity span distributions across annotation schemes (IOB1 and IOB2) for Arabic (ANERCorp) and English (CoNLL-2003). Bars reflect both absolute counts and normalised proportions of entity spans in the training and test splits.

Table 4.4: Number of Entity Spans Discarded in ANERCorp under IOB2 Due to Misaligned `I-` Tags

| **Entity Type** | **Train Split** | **Test Split** |
|---|---|---|
| I-PER | 15 | 57 |
| I-MISC | 13 | 10 |
| I-ORG | 11 | 9 |
| I-LOC | 3 | 8 |
| **Total** | **42** | **84** |

Second, in terms of entity span distribution, CoNLL-2003 exhibits a relatively balanced distribution across LOC, ORG, and PER entities, with MISC being the least frequent. ANERCorp, by contrast, shows notable variation between splits: the training split is dominated by LOC spans, while the test split contains a higher proportion of PER spans. This shift in dominant entity types between training and test sets suggests potential differences in content or domain focus. Additionally, the higher frequency of `B-LOC` over `I-LOC` in ANERCorp—discussed in Figure 4.4—confirms that most LOC spans are single-token entities. In contrast, PER spans tend to be more complex and multi-word.

To assess the complexity of entity spans, we examine their length distributions across entity types, languages, and annotation schemes. We use box-plot style visualisations to capture both central tendency and variability. Figure 4.6 presents span length distributions under both annotation schemes (IOB1 and IOB2), separated by train and test splits. In these plots, the dots represent the mean span length, the upper whisker indicates the maximum observed span, and the lower bound corresponds to the minimum span, which is always a single token. This design allows us to capture both typical span lengths and the full observed range of entity complexity.

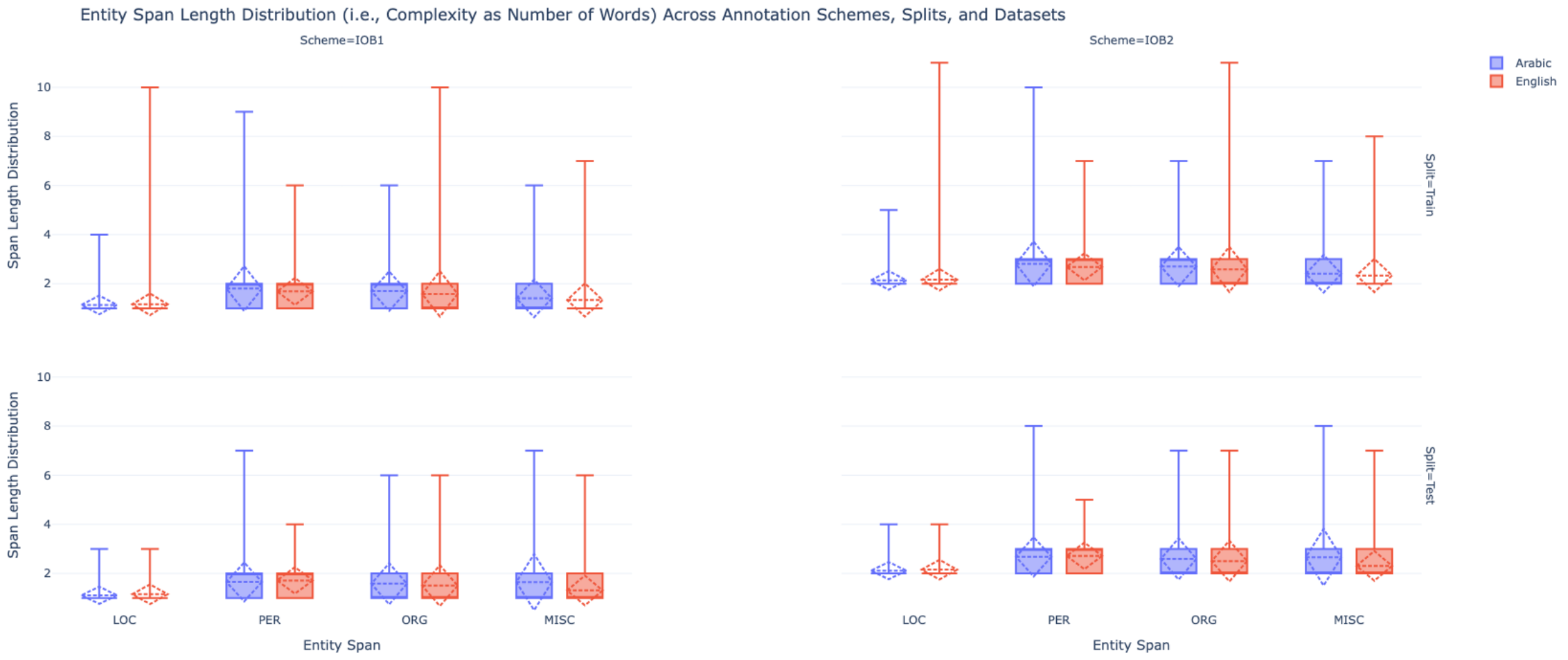

Figure 4.6: Entity span length distributions across annotation schemes (IOB1 and IOB2) in Arabic (ANERCorp) and English (CoNLL-2003). The dots represents the mean span length, the upper whisker indicates the maximum observed span, and the lower bound corresponds to the minimum span (always one token). This visualisation captures both typical span lengths and the full observed range of entity complexity across entity types in train and test splits.

Across both annotation schemes, LOC entities consistently exhibit the shortest spans, almost always a single token and rarely exceeding three tokens. By contrast, PER entities display not only higher mean lengths but also the longest maximum spans, confirming that person names often span multiple tokens. In Arabic (ANERCorp), PER spans reach up to nine tokens in the training split and seven in the test set, whereas in English (CoNLL-2003) the maxima are slightly shorter, at six and four tokens respectively. ORG and `MISC` entities fall between these extremes, with Arabic tending to produce slightly longer organisational spans compared to English. The choice of anno-tation scheme (IOB1 vs. IOB2) has minimal impact on the distributions, indicating that span complexity is robust to annotation conventions.

In summary, our entity-level analysis reveals several structural differences between ANERCorp and CoNLL-2003 that are shaped by both annotation schemes and dataset characteristics. ANERCorp exhibits annotation inconsistencies that impact entity span distribution under stricter IOB2 constraints, resulting in 126 spans being discarded across training and test sets. CoNLL-2003, by contrast, remains fully consistent under both IOB1 and IOB2 schemes. Span distribution in CoNLL-2003 is relatively balanced across entity types, while ANERCorp shows a shift in dominant entity types between splits, with LOC spans most frequent in training and PER spans in testing. In terms of complexity, LOC entities are generally short and often single-token, while PER entities are longer and more complex—especially in ANERCorp.

In the next section, we analyse lexical diversity, token overlap, and out-of-vocabulary (OOV) patterns within and across ANERCorp and CoNLL-2003. This lexical analysis uncovers additional dimensions of dataset variation that may influence model generalisation performance.

### 4.3.3 Lexical Analysis

To address aspects (c) and (d) of RQ1—*(c) how lexical behaviour differs across datasets, and (d) how tokenisation affects lexical structure*—we conduct a comparative analysis of lexical diversity, vocabulary overlap, and out-of-vocabulary (OOV) rates within and between Arabic (ANERCorp) and English (CoNLL-2003). Each of these properties is analysed at both the word level (pre-tokenisation) and the token level (post-tokenisation), enabling us to quantify the impact of tokenisation on the vocabulary characteristics of each language.

We adopt a granular approach to lexical analysis, beginning at the dataset level, moving to split-level comparisons, and finally examining entity tag-specific distributions. At each level, we investigate the impact of tokenisation by comparing both word-level (pre-tokenisation) and token-level (post-tokenisation) characteristics. We start by examining type distributions—that is, how vocabulary is distributed across datasets, splits, and entity tags—before introducing lexical diversity metrics such as the Type-to-Word Ratio (TWR) and Type-to-Token Ratio (TTR). We then explore out-of-vocabulary (OOV) rates to quantify the extent of unseen lexical content across splits and datasets. Finally, we analyse vocabulary overlap across entity tags to identify potential annotation ambiguities. This structured analysis enables us to assess the effects of linguistic variation and tokenisation across different levels of granularity, providing a foundation for interpreting how dataset structure impacts performance differences between Arabic and English NER tasks in later sections

**High-Level Type Analysis** To assess broad lexical characteristics, we compare the number of unique word types and unique NE word types across ANERCorp and CoNLL-2003. Table 4.5 reports the total and unique word counts, as well as the number of NE words and unique NE words for each dataset. While CoNLL-2003 is larger overall in terms of total word and NE word count, ANERCorp contains a higher number of unique word types (32,714 vs. 27,316). This suggests that ANERCorp's vocabulary is more diverse, even though it is drawn from a smaller total corpus.

Table 4.5: Total and Unique Word Counts across Datasets.

| Dataset | Total Words | Unique Words | NE Words | Unique NE Words |
|---|---|---|---|---|
| ANERCorp | 150,110 | 32,714 | 16,556 | 5,026 |
| CoNLL-2003 | 250,056 | 27,316 | 42,155 | 9,819 |

Building on the overall analysis, we explore the individual dataset splits—training and testing—to assess whether the previously observed patterns hold. Table 4.6 presents key word and type statistics for the training and testing splits of ANERCorp and CoNLL-2003.

Table 4.6: Comparison of Word and Type Counts Across Training and Testing Splits for ANERCorp and CoNLL-2003.

| Metric | ANERCorp | | CoNLL-2003 | |
|---|---|---|---|---|
| | Train | Test | Train | Test |
| Total Words | 125,102 | 25,008 | 203,621 | 46,435 |
| Unique Words | 29,252 | 9,075 | 23,623 | 9,488 |
| NE Words | 13,181 | 3,375 | 34,043 | 8,112 |
| Unique NE Words | 4,069 | 1,603 | 8,287 | 3,129 |

Table 4.6 provides a comparative analysis of vocabulary and type-based statistics across the training and testing splits of ANERCorp and CoNLL-2003. While CoNLL-2003 contains more NE words and unique NE types in both splits, ANERCorp shows higher proportions of unique types relative to its overall word count—especially in the training split. For instance, ANERCorp's training set has more unique word types (29,252) than CoNLL-2003 (23,623), despite being significantly smaller in size. While CoNLL-2003 includes more named entity words, ANERCorp exhibits higher lexical variety relative to its size, likely due to its broader linguistic and topical coverage.

**Tokenisation Impact on High-level Type Characteristics** To investigate the impact of tokenisation, we first compare word-level and token-level counts across the training and testing splits of ANERCorp and CoNLL-2003. As previously noted in Section 3.3.4, BERT-based models operate on subword tokenisation, which segments words into smaller subword units. In this analysis, we represent each word by its first subword.

Table 4.7: Comparison of Word and Token Statistics across ANERCorp and CoNLL-2003 for Training and Testing Splits. Words refer to pre-tokenisation units; tokens refer to post-tokenisation units.

| **Metric** | **ANERCorp** | | **CoNLL-2003** | |
|---|---|---|---|---|
| | **Train** | **Test** | **Train** | **Test** |
| Total Words | 125,102 | 25,008 | 203,621 | 46,435 |
| Total Tokens | 124,659 | 24,991 | 203,621 | 46,435 |
| Unique Words | 29,252 | 9,075 | 23,623 | 9,488 |
| Unique Tokens | 23,010 | 8,455 | 11,243 | 5,883 |

As shown in Table 4.7, tokenisation leads to a reduction in the number of unique types. In ANERCorp, unique token types decrease by 21% in training and 7% in testing. In CoNLL-2003, the reduction is more substantial—52% in training and 38% in testing—suggesting a higher degree of tokenisation in English. This likely reflects differences in tokeniser behaviour and language morphology: Arabic's morphologically complex words are often preserved as larger units, whereas English words are more frequently decomposed into smaller, repetitive subword units.

Although total token counts remain almost unchanged relative to word counts, the observed reductions in unique types highlight how tokenisation affects type-level metrics more significantly. This high-level comparison reveals key lexical differences between ANERCorp and CoNLL-2003. While CoNLL-2003 contains more total and named entity words, ANERCorp consistently exhibits higher type diversity relative to its size—particularly in the training split. These differences appear at both the word and token levels, though tokenisation reduces the number of unique types more noticeably in English. This shift highlights the importance of examining both word-level and token-level vocabulary characteristics when assessing dataset complexity. Next, we investigate how word and token types are distributed across entity tags.

**Word Type Distribution Across Entity Tags** Building on our previous analyses of word distributions across entity tags, we now turn to the distribution of word types across entity tags. This allows us to explore how vocabulary diversity maps onto annotation structure. While word-level analysis reflects the frequency of individual terms, type-level analysis highlights how many distinct words are used within each entity tag, regardless of frequency. Figure 4.7 presents this distribution of word types per entity tag across training and test splits for both ANERCorp and CoNLL-2003.

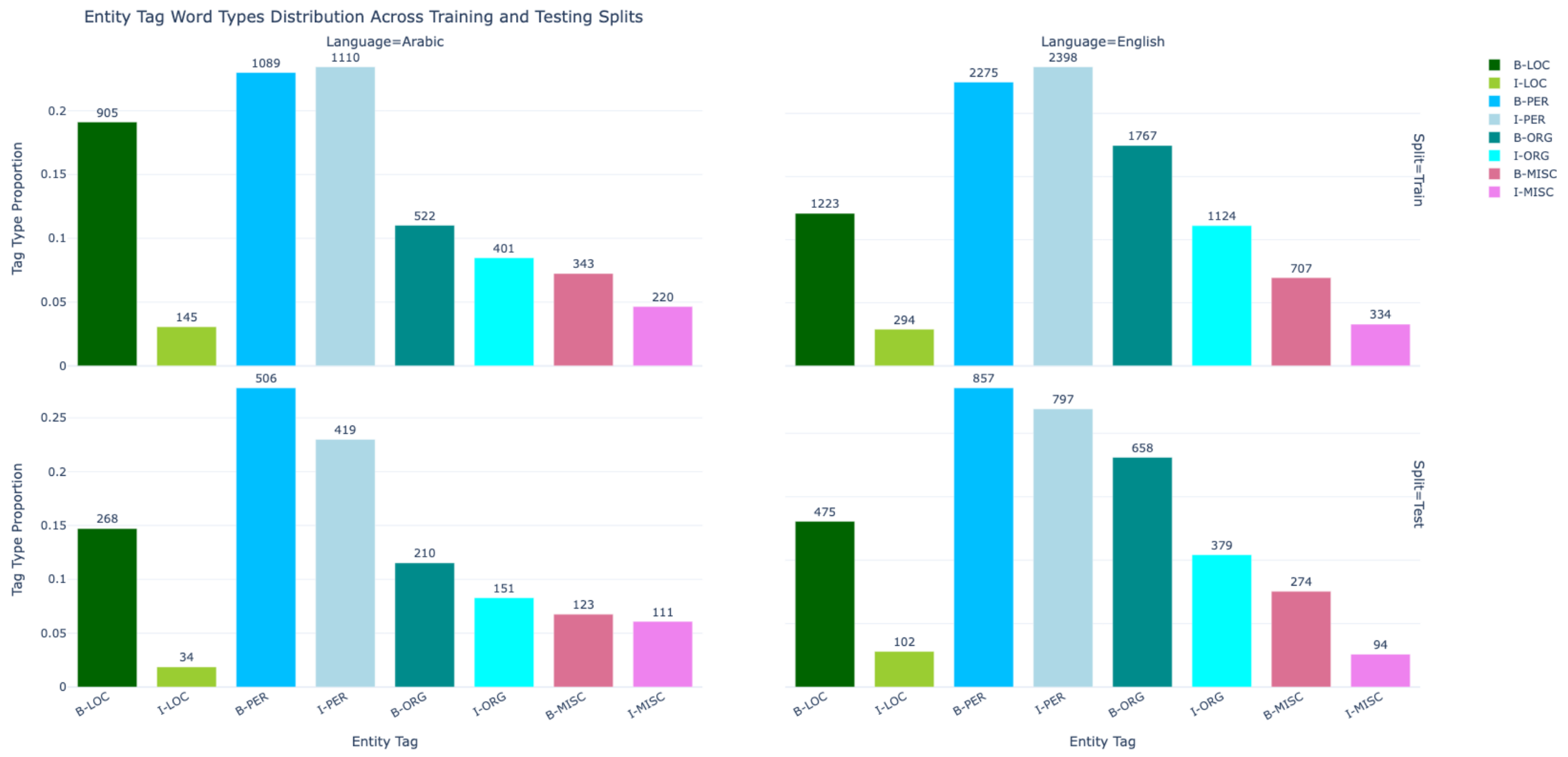

Figure 4.7: Distribution of word types per entity tag across training and test splits in Arabic (ANERCorp) and English (CoNLL-2003). Bars represent the percentage of unique word types assigned to each tag, with absolute counts shown above each bar.

Several patterns emerge. LOC tags (B-LOC, I-LOC) show relatively lower type counts compared to their word frequency, particularly in ANERCorp. This suggests a small set of recurring location names (e.g., country names), potentially repeated across many contexts. In contrast, tags like I-PER exhibit higher diversity in CoNLL-2003, particularly in the test split, suggesting more varied vocabulary for person entities.

To assess the impact of tokenisation on this distribution, we now compare token-level type distributions across entity tags. Note that tokens refer to either the word itself or the first subword if the word is tokenised, following the BERT-based tokenisation scheme described in Section 3.3.4.

As shown in Figure 4.8, ANERCorp retains a similar distribution of token types across entity tags when compared to the word-level analysis (Figure 4.4). The rank order of tags remains largely unchanged, and while the total number of types decreases, proportional relationships between tags are preserved. This suggests that tokenisation has a relatively moderate effect on the type distribution structure in Arabic.

By contrast, CoNLL-2003 shows a larger reduction in token type counts across all entity tags, especially in the training split. B-PER becomes the most frequent tag, appearing more often than tags like B-ORG and I-PER. The small decrease in I-LOC—from 294 to 263—suggests that location entities in English are less affected by tokenisation, possibly because they tend to be shorter or simpler in form.

Overall, while both datasets show a reduction in token type after tokenisation, the impact on CoNLL-2003 is more. This reflects the higher tokenisation rate in English compared to Arabic. However, in both ANERCorp and CoNLL-2003, the relative proportions and rank order of entity tags remain stable. This indicates that tokenisation affects the number of distinct types but does not significantly alter the overall annotation structure.

#### 4.3.3.1 Lexical Diversity

Lexical diversity, as discussed in Section 4.4.1, quantifies the variety of vocabulary in a dataset. It is measured as the ratio between the number of unique word types (pre-tokenisation) or token types (post-tokenisation) and the total number of words or tokens. These metrics provide insight into how repetitive or varied a dataset is, which can influence model learning and generalisation.

We begin our analysis at the dataset level, examining overall lexical diversity across ANERCorp and CoNLL-2003. We then break this down by training and testing splits, and finally explore diversity at the entity tag level. At each stage, we also investigate how tokenisation affects lexical diversity by comparing word-level and token-level patterns in both languages.

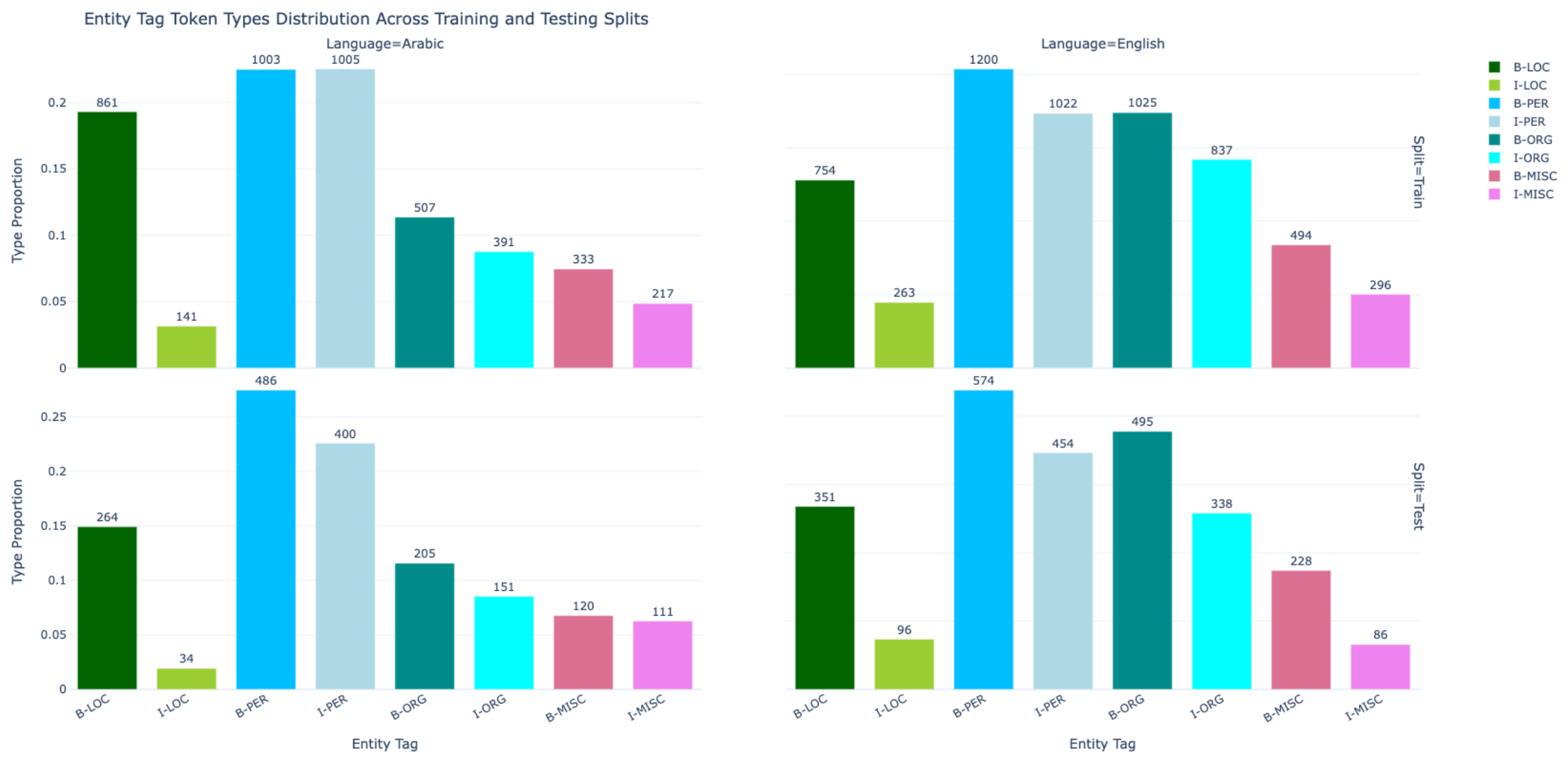

Figure 4.8: Distribution of token types per entity tag across training and test splits in Arabic (ANERCorp) and English (CoNLL-2003). Bars represent the percentage of unique token types assigned to each tag, with absolute counts shown above each bar.

**High-Level Lexical Diversity** Table 4.8 shows the TWR (Type-to-Word Ratio) and TEWR (Type-to-Entity-Word Ratio), which provide high-level indicators of lexical diversity across ANERCorp and CoNLL-2003. TWR captures general vocabulary variety, while TEWR focuses on diversity among named entities. ANERCorp exhibits higher variability on both metrics—21.79% vs. 10.92% for TWR and 30.36% vs. 23.29% for TEWR—indicating higher lexical diversity in both general vocabulary and named entities.

Table 4.8: Proportional metrics for lexical diversity (TWR), proportion of named entity words (NE Proportion), and named entity diversity (TEWR).

| Dataset | TWR | NE Proportion | TEWR |
|---|---|---|---|
| ANERCorp | 21.79% | 11.03% | 30.36% |
| CoNLL-2003 | 10.92% | 16.86% | 23.29% |

We now turn to the training and testing splits of each dataset to examine whether the lexical diversity patterns observed at the dataset level hold. Table 4.9 presents lexical diversity for both ANERCorp and CoNLL-2003 across splits.

Table 4.9 compares lexical variability across the training and testing splits for ANERCorp and CoNLL-2003. In both datasets, lexical diversity increases notably from training to test, as reflected in higher TWR and TEWR values. For instance, in ANERCorp, TWR rises from 23.38% to 36.29% and TEWR from 30.87% to 47.50%. A similar trend is observed in CoNLL-2003, with TWR increasing from 11.60% to 20.43% and TEWR from 24.34% to 38.57%. While both languages exhibit higher diversity in their test sets, ANERCorp consistently maintains higher lexical diversity than CoNLL-2003 across both training and test splits.

Table 4.9: Lexical diversity metrics across training and testing splits for ANERCorp and CoNLL-2003, showing general vocabulary variation (TWR) and named entity diversity (TEWR).

| Metric | ANERCorp | | CoNLL-2003 | |
|---|---|---|---|---|
| | Train | Test | Train | Test |
| Unique Words | 29,252 | 9,075 | 23,623 | 9,488 |
| TWR (%) | **23.38** | **36.29** | 11.60 | 20.43 |
| Unique NE Words | 4,069 | 1,603 | 8,287 | 3,129 |
| TEWR (%) | **30.87** | **47.50** | 24.34 | 38.57 |

The observed discrepancy in lexical diversity between ANERCorp and CoNLL-2003 can be potentially attributed to both data sourcing strategies and linguistic structure. ANERCorp was compiled from 316 articles drawn from a diverse range of news outlets across the Arab world. This includes publications from different countries, regions, and editorial styles, intentionally designed to capture broad linguistic and cultural variation. As a result, the dataset encompasses a wide range of vocabulary—including regional expressions, topic-specific terms, and morphologically rich word forms—which contributes to its high Type-to-Word Ratio (21.79%).

In contrast, CoNLL-2003 is constructed from a single source—Reuters English newswire—over a fixed one-year period. This potentially results in a more homogenous corpus, both stylistically and lexically. Its vocabulary reflects the relatively standardized and repetitive style of formal, international news reporting, which may contribute to the lower TWR of 10.92%.

Together, these factors—multi sources, regional and topical diversity, and linguistic morphology—can potentially explain the higher lexical diversity observed in ANERCorp relative to CoNLL-2003. These properties may affect not only the complexity of the learning process but also a model's ability to generalize across diverse or morphologically rich input, especially when working with Arabic.

**Entity Tag Lexical Diversity** We examine lexical diversity within entity tags to understand how varied the vocabulary is within each class. Two tags might appear equally frequent but differ in how many unique words they contain. Figure 4.9 shows the Type-to-Word Ratio (TWR) for each entity tag across Arabic (ANERCorp) and English (CoNLL-2003).

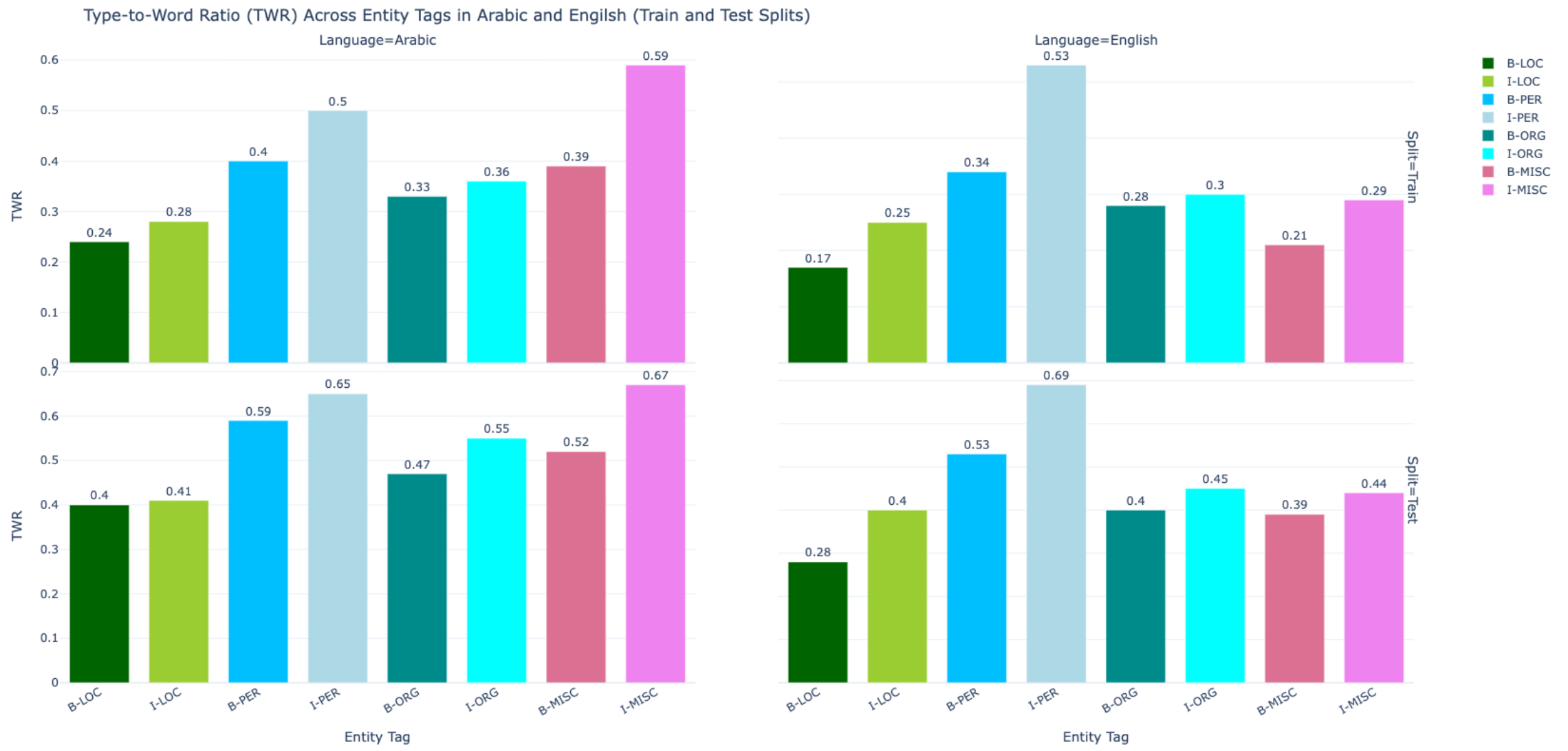

Figure 4.9: Type-to-Word Ratio (TWR) across entity tags in Arabic (ANERCorp) and English (CoNLL-2003), shown separately for training and test splits.

ANERCorp shows higher lexical diversity for almost all tags compared to CoNLL-2003, except for I-PER. Tags like I-MISC and I-ORG in ANERCorp have especially high TWR values, indicating a wider variety of terms. In the test split, I-LOC also shows more diversity in ANERCorp. The most diverse tag in ANERCorp is I-MISC, despite having fewer total words. In CoNLL-2003, the most diverse tag is I-PER. Across both datasets, B-LOC consistently shows the lowest diversity.

Lexical diversity can influence model performance in different ways. Higher diversity—especially in training data—can support better generalization by exposing models to a wider set of word forms. However, it may also increase learning complexity. On the other hand, lower diversity can simplify training due to repetition, but may limit a model's ability to generalize and can introduce ambiguity when similar terms are inconsistently labelled across contexts.

To better understand how entity types are distributed within tags, we analyse the standard deviation of word type frequencies for each tag. Figure 4.10 illustrates how evenly—or unevenly—word types are distributed within entity tags, based on frequency.

A higher standard deviation indicates that a few word types dominate the tag, while a lower standard deviation suggests a more even spread.

For each entity tag (e.g., B-LOC), we count how many times each unique word type appears with that tag and then measure how much these counts vary. For example, if B-LOC contains 1,000 tokens but only 200 unique word types, and some words such as "United," "Arabia," and "France" occur dozens of times while most others appear only once or twice, the variation between frequent and rare types produces a high standard deviation. In contrast, if all word types occur with roughly similar frequency, the standard deviation would be low. This measure therefore captures lexical concentration and imbalance within each entity class, highlighting which tags are dominated by a small subset of frequent words.

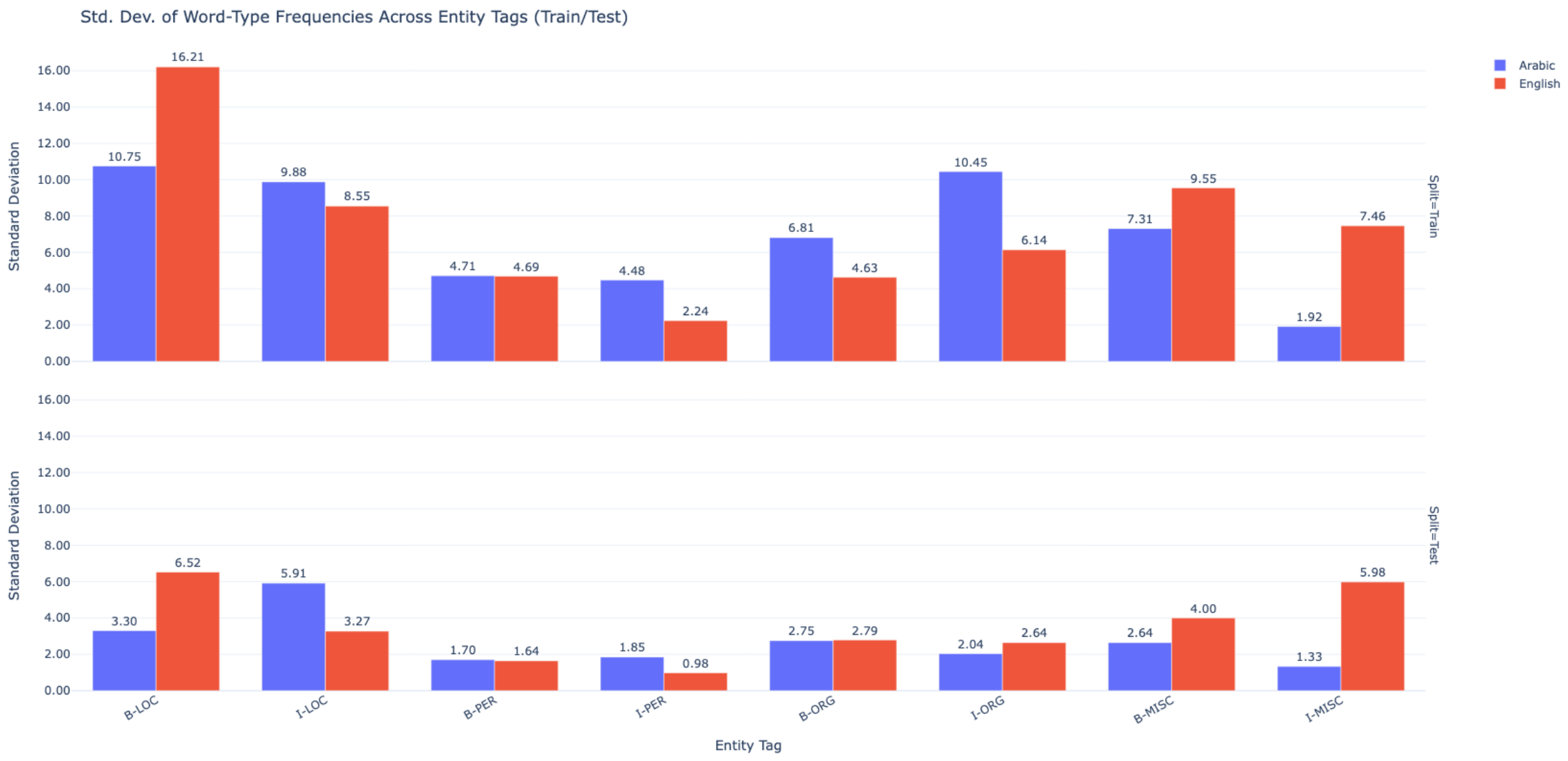

Figure 4.10: Standard deviation of word type frequencies across entity tags in Arabic (ANERCorp) and English (CoNLL-2003), shown separately for training and test splits.

As shown in Figure 4.10, the B-LOC tag consistently exhibits the highest standard deviation across both datasets. This is especially evident in CoNLL-2003, where a small number of word types—such as "U.S.", which appears 309 times out of 1,223 instances—dominate the distribution. Such skewed patterns may help models memorize frequent examples but can affect generalization. In contrast, tags like B-MISC and I-MISC in ANERCorp show high diversity alongside relatively low standard deviation, indicating a broad vocabulary where most types appear only a few times. CoNLL-2003 displays a much less balanced structure in these tags, with fewer distinct types demonstrated by a higher standard deviation value.

**Tokenisation Impact on Entity Tag Lexical Diversity** Shifting from word-level to token-level analysis, we evaluate how subword tokenisation affects lexical diversity across entity tags. We use the Type-to-Token Ratio (TTR) to measure variability, calculated as the ratio of unique token types to the total number of tokens within each tag. Figure 4.11 presents TTR values across both training and test splits.

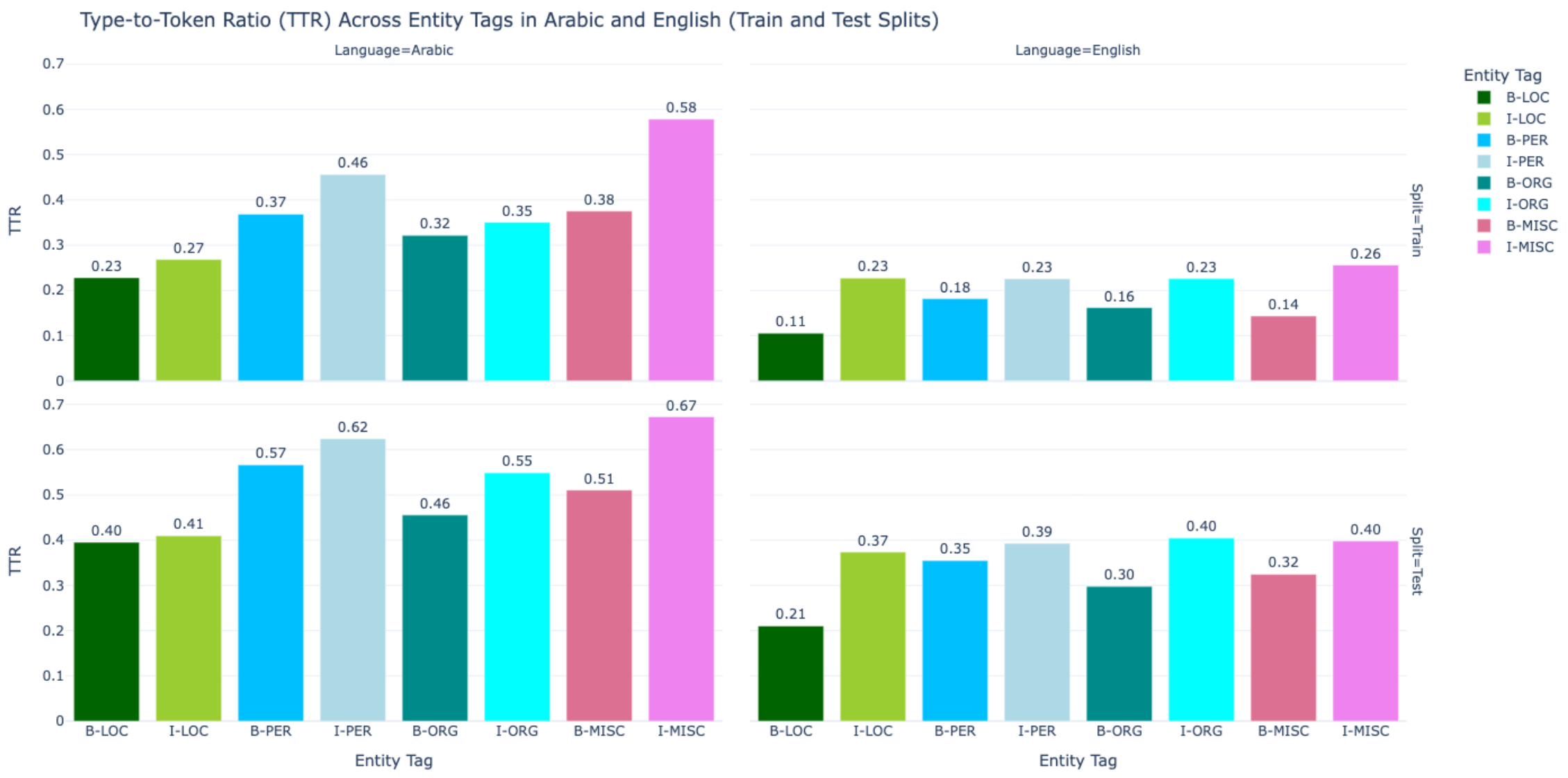

Figure 4.11: Type-to-Token Ratio (TTR) across entity tags in Arabic (ANERCorp) and English (CoNLL-2003), shown separately for training and test splits.

Tokenisation reduces lexical diversity in both datasets, particularly in the training splits. The reduction is more noticeable in CoNLL-2003, where TTR values drop across all tags. ANERCorp, by contrast, retains higher TTR across both splits, indicating higher variability even after tokenisation.

To assess whether this change also affects frequency balance, Figure 4.12 shows the standard deviation of token frequencies within each tag. This reflects how evenly token types are distributed across their occurrences.

Three observations emerge: First, standard deviation increases across all tags and splits, confirming that tokenisation leads to more skewed frequency distributions. Second, ANERCorp shows only moderate increases, suggesting tokenisation has limited effect on type balance. Third, CoNLL-2003 exhibits stronger skewness—especially in B-LOC, B-PER, and B-ORG—indicating that fewer types dominate after tokenisation.

These shifts suggest that tokenisation reduces vocabulary diversity in both datasets, but in English the reduction is more severe, creating overlapping subword units that increase type frequency concentration and lead to more uneven annotation distributions.

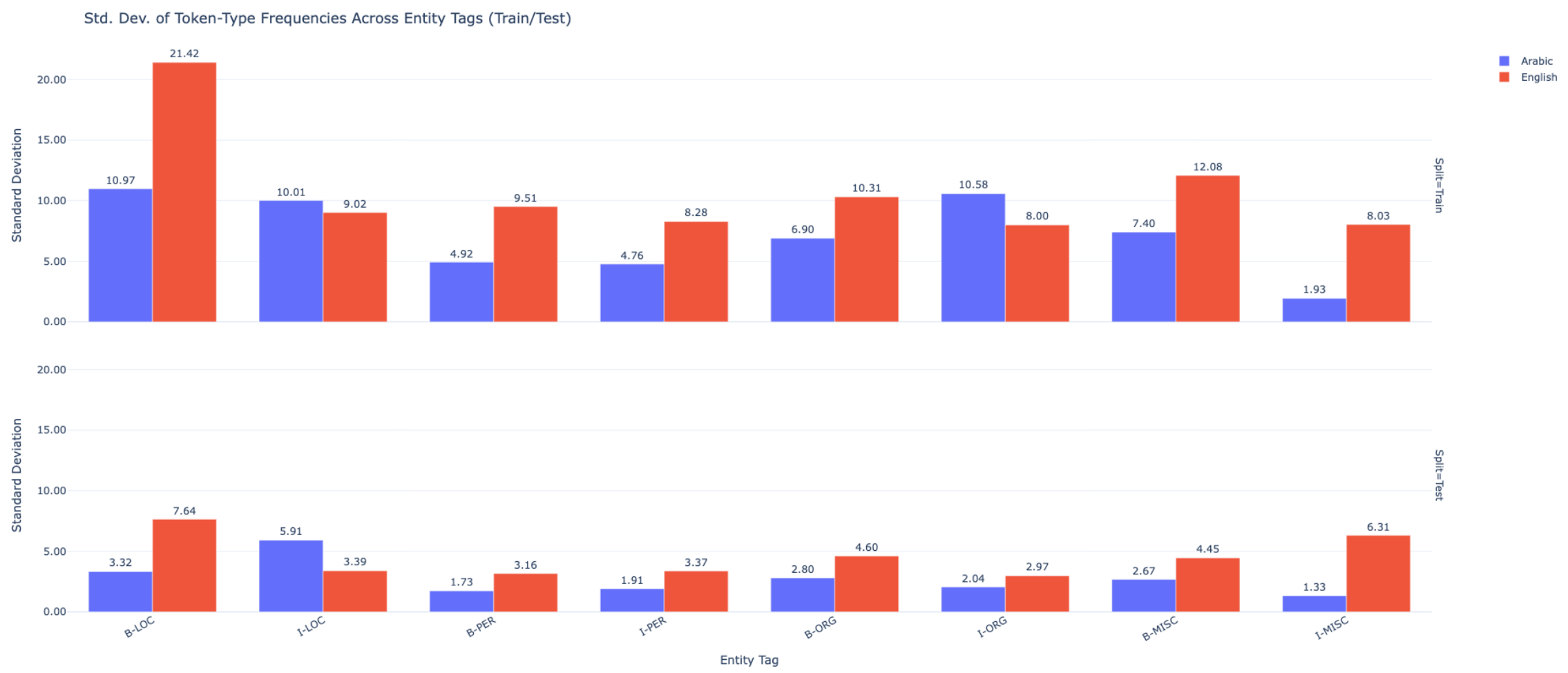

Figure 4.12: Standard deviation of token type frequencies across entity tags in Arabic (ANERCorp) and English (CoNLL-2003), shown separately for training and test splits.

In summary, our analysis of lexical diversity reveals distinct patterns between ANERCorp and CoNLL-2003. ANERCorp consistently exhibits higher overall lexical diversity and higher variability within most entity tags compared to CoNLL-2003, likely due to its diverse data sources and richer morphology. Tags like `I-MISC` in ANERCorp exhibited particularly rich and balanced vocabularies, while tags such as `B-LOC` in CoNLL-2003 were dominated by a few frequent terms, leading to skewed distributions. While lexical diversity generally increases from training to test splits in both datasets, tokenisation leads to a reduction in this diversity and an increase in the skewness of token frequency distributions across entity tags. Notably, the impact of tokenisation on reducing diversity and increasing frequency imbalances appears more pronounced in CoNLL-2003 (English) than in ANERCorp (Arabic). In English, tokenisation tends to produce more repetition by collapsing different words into common subword forms, increasing the dominance of a few token types. In contrast, this repetition effect is more moderate in Arabic, particularly for tags like `I-MISC`, which retain a wide and balanced token distribution even after tokenisation. These differences in lexical richness and the effect of tokenisation may have implications for how NER models learn and generalise on these two languages.

#### 4.3.3.2 Out-of-Vocabulary (OOV)

To better understand the relationship between training and test splits, we analyse Out-of-Vocabulary (OOV) rates—defined as the proportion of unique word types in the test set that are not present in the training set. This metric provides insight into how well the training data captures the vocabulary encountered during evaluation. Table 4.10 presents the OOV counts and rates for both datasets.

Table 4.10: OOV analysis for ANERCorp and CoNLL-2003. "OOV Count" indicates the number of unique word types in the test split not found in the training split. "OOV Rate" expresses this as a proportion of all unique test word types.

| Dataset | OOV Count | OOV Rate |
|---|---|---|
| ANERCorp | 3,462 | 0.3815 |
| CoNLL-2003 | 3,693 | 0.3892 |

Despite ANERCorp's higher lexical diversity in both training and test splits, its OOV rate (38.15%) is nearly identical to that of CoNLL-2003 (38.92%). This suggests that while ANERCorp offers broader vocabulary coverage, its test set also introduces a high number of new terms—likely due to its linguistic complexity and diverse topical content. CoNLL-2003, by contrast, has a more focused and repetitive vocabulary, which might be expected to yield lower OOV, but the rate remains slightly higher, potentially due to a less diverse training vocabulary.

**Entity Tag-Level OOV** While overall OOV rates provide a general indication of vocabulary coverage, they do not capture differences between specific entity tags. Certain tags—such as person names—often exhibit higher OOV rates due to their inherent variability, whereas others—like location names—tend to be more stable and repetitive. To analyse these variations, Figure 4.13 presents the OOV rate per entity tag in the test split for both datasets.

ANERCorp exhibits higher OOV rates across most entity tags, especially for `I-MISC`, `B-MISC`, and `I-ORG`, reflecting a broader range of unique or rare terms. In contrast, CoNLL-2003 shows lower OOV rates for most tags but maintains relatively high OOV in `B-PER` and `I-PER`. These differences are important for NER performance, as tags with higher OOV rates typically present more challenges for model generalization to unseen data.

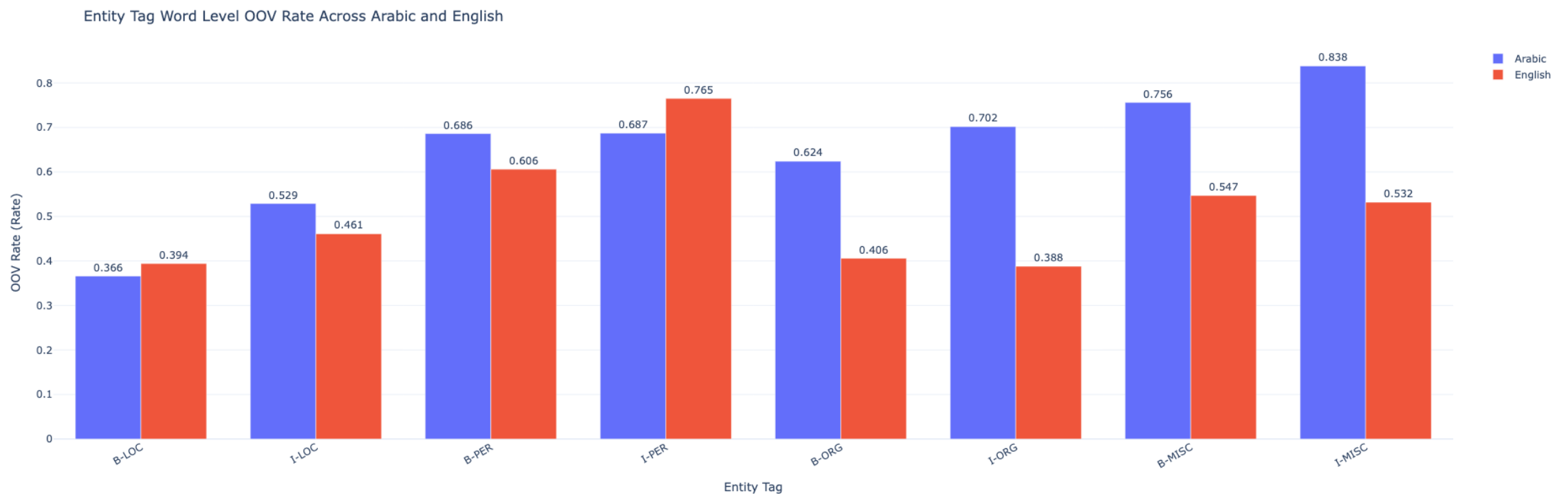

Figure 4.13: Word-level out-of-vocabulary (OOV) rates across entity tags in the test split of Arabic (ANERCorp) and English (CoNLL-2003).

**Token-Level OOV and Tokenisation Effects** Tokenisation helps reduce OOV by segmenting unseen words into known subword units. To explore this, we compute token-level OOV rates for each entity tag, shown in Figure 4.14.

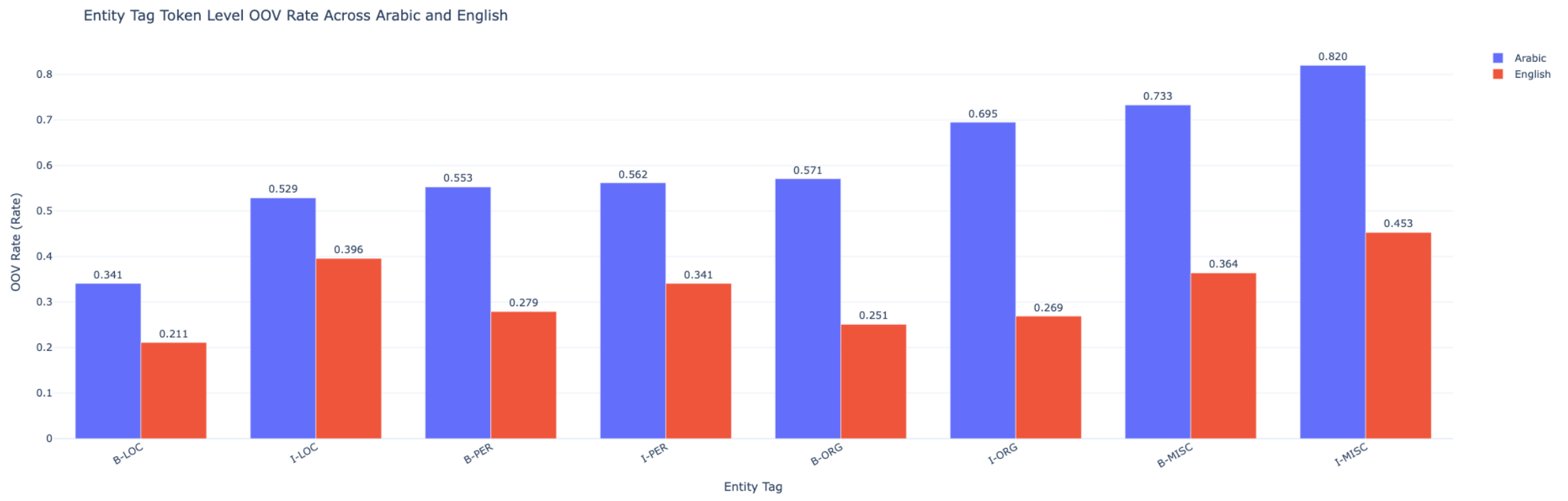

Figure 4.14: Token-level out-of-vocabulary (OOV) rates across entity tags in the test split of Arabic (ANERCorp) and English (CoNLL-2003).

Tokenisation generally lowers OOV rates by breaking down unseen words into familiar subword units. Figure 4.14 illustrates token-level OOV for each entity tag. CoNLL-2003 shows a more substantial OOV reduction post-tokenisation, particularly for tags like B-LOC and I-PER, possibly because the vocabulary associated with these entities has more predictable subword patterns. ANERCorp's OOV decreases are less significant, aligning with its lower tokenisation rates. Notably, I-MISC maintains high OOV in both datasets even after tokenisation, suggesting subword tokenisation struggles with the vocabulary within this diverse category.

In summary, OOV analysis shows that Arabic (ANERCorp) and English (CoNLL-2003) exhibit broadly comparable overall OOV rates at the dataset level, despite differences in dataset size and lexical diversity. At the entity-tag level, the largest cross-language differences occur in categories such as `I-MISC` and PER, which consistently show the highest OOV rates but with different distributions across Arabic and English. Tokenisation reduces OOV in both datasets, but the effect is much stronger in English, where words are decomposed into subword units. This increases training–test overlap, leading to greater repetition and improved coverage. In Arabic, tokenisation also reduces OOV, though more moderately, and morphologically complex or rare tags such as `I-MISC` remain challenging. Together, these patterns highlight how language morphology and tokeniser behaviour jointly shape vocabulary coverage. We next examine vocabulary overlap across entity tags to further explore these effects.

#### 4.3.3.3 Vocabulary Overlap Across Entity Tags

Understanding how word and token types are distributed across entity tags helps reveal annotation inconsistencies and areas where models might confuse entity boundaries or categories. In this section, we analyse vocabulary overlap across entity tags at both the word and token levels, considering training and test splits separately.

**Entity Tags Overlap — Training Split** As depicted in Figure 4.15, analysis of word-type overlap in the training splits of both datasets reveals several consistent patterns: (i) overlap between the O tag and named entities, (ii) cross-entity type overlap (e.g., B-LOC with B-ORG), and (iii) boundary overlap (e.g., B-PER with I-PER). The latter is especially pronounced for PER-related tags in both ANERCorp and CoNLL-2003, though its magnitude differs—CoNLL-2003 shows substantially higher counts (e.g., around 724 overlaps for PER boundaries compared to 283 in ANERCorp). More generally, the dominant overlap types vary between the datasets: in CoNLL-2003, cross-entity type overlaps such as B-ORG with B-LOC are more prominent, whereas in ANERCorp, the O tag overlaps extensively with nearly all other entity tags. Mixed overlaps, where both boundary and entity-type inconsistencies occur, are also present in both datasets.

We now turn to token overlap. Figure 4.16 illustrates the impact of tokenisation on entity-tag overlap in the training splits. In both datasets, the degree of overlap increases once words are decomposed into subword units. In ANERCorp, this reinforces the existing dominance of the O tag overlapping with multiple entity tags, while also amplifying cross-entity overlaps. In CoNLL-2003, the effect is even stronger: tokenisation increases overlap across most entity types, with cross-entity overlaps such as B-ORG with B-LOC becoming more prominent, and new patterns emerging where B-ORG, B-PER, and I-PER show greater co-occurrence with the O tag. These increases reflect the behaviour of subword tokenisers, which often produce shared initial subwords, thereby introducing artificial overlap between otherwise distinct entities.

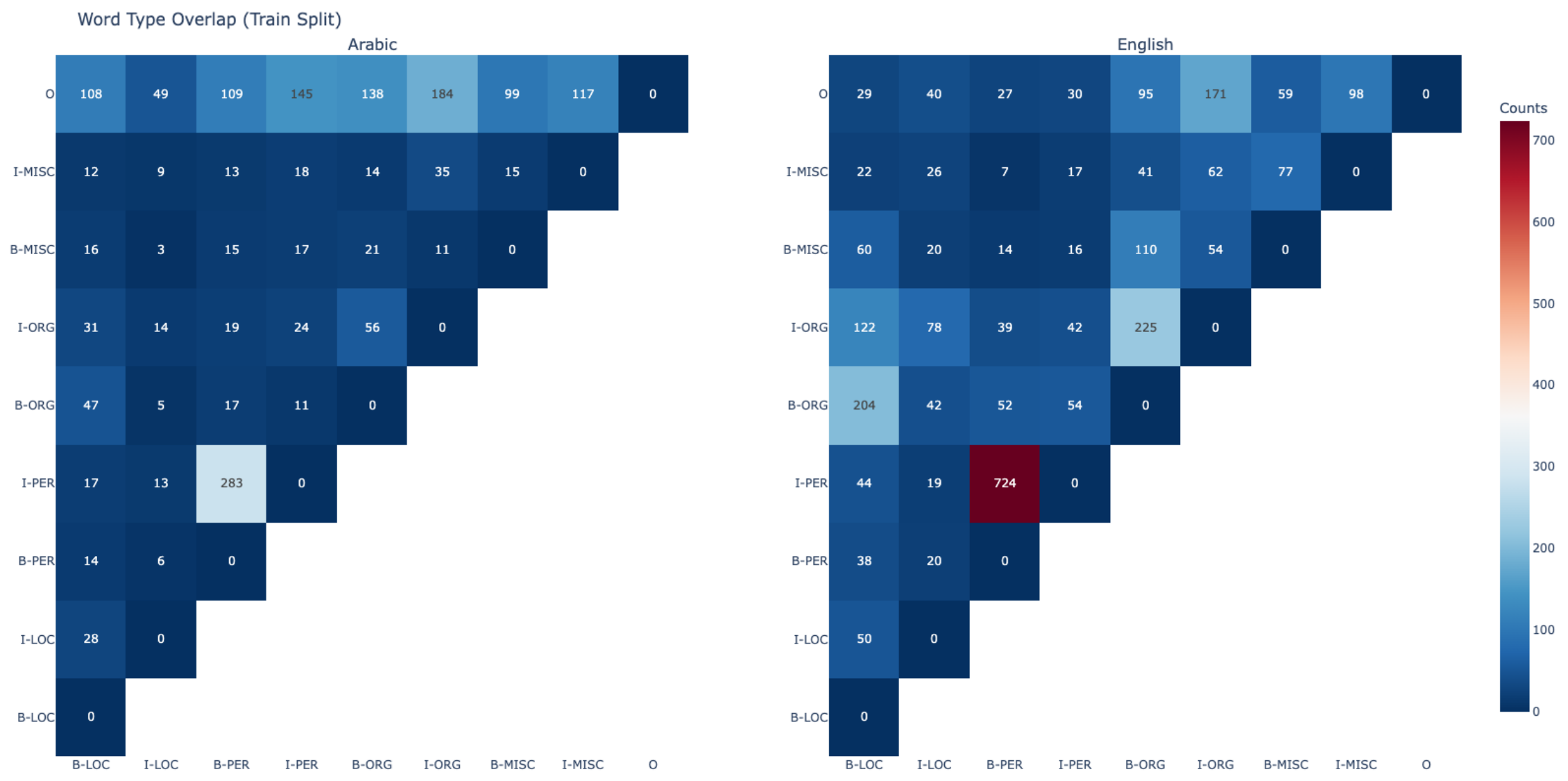

Figure 4.15: Word-type overlap across entity tags in the training split of Arabic (ANERCorp) and English (CoNLL-2003). Diagonal values are zeroed out to focus on inter-tag overlap.

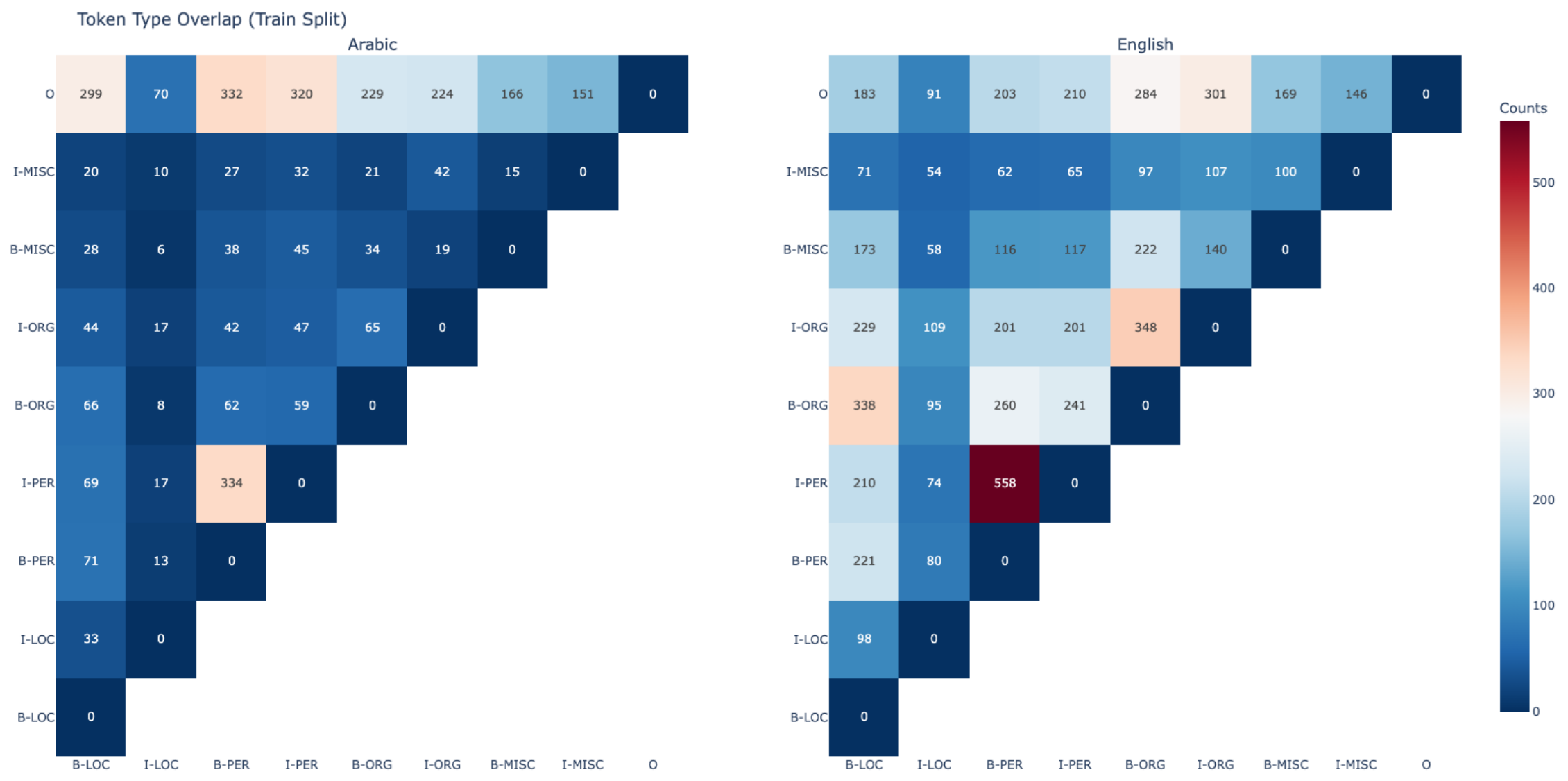

Figure 4.16: Token-type overlap across entity tags in the training split of Arabic (ANERCorp) and English (CoNLL-2003). Diagonal values are zeroed out to focus on inter-tag overlap.

**Entity Tags Overlap — Test Split** The word type overlap patterns observed in the testing splits, as depicted in Figure 4.17, are very similar to those found in the training data for both ANERCorp and CoNLL-2003. These include overlaps between the O tag and named entities, cross-entity type overlaps such as between B-LOC and B-ORG, and boundary overlaps, notably high within PER-related tags in both datasets. For ANERCorp, the most dominant patterns remain boundary overlap, particularly within PER, and overlap with the O tag, with less frequent cross-entity type overlap. CoNLL-2003 also shows significant PER and ORG overlaps, alongside a more notable overlap between B-ORG and B-LOC.

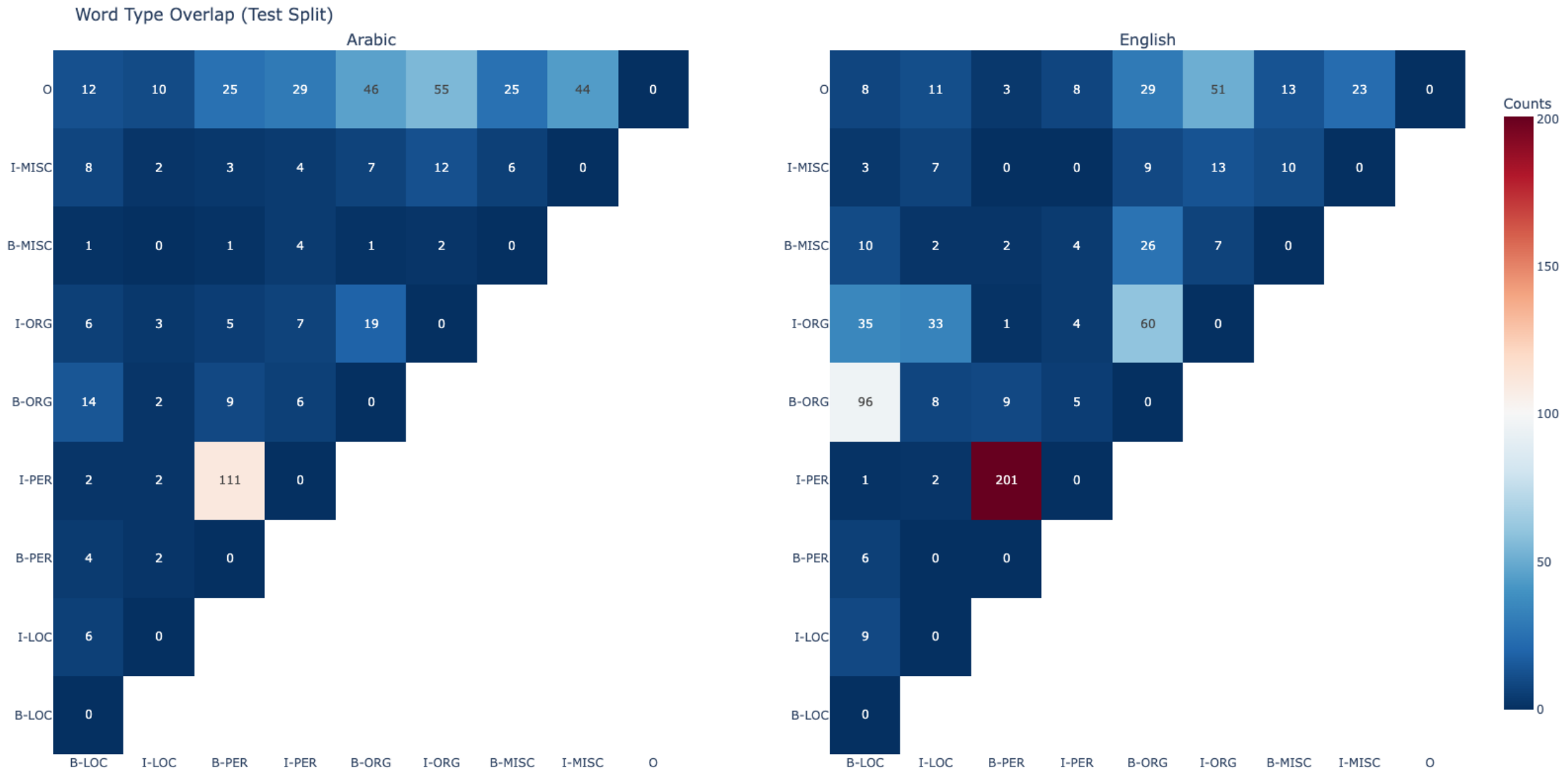

Figure 4.17: Word-type overlap across entity tags in the test split of Arabic (ANERCorp) and English (CoNLL-2003). Diagonal values are zeroed out to focus on inter-tag overlap.

Figure 4.18 presents the token type overlap in the testing splits. The impact of tokenization on the test splits is generally similar to the trends seen in the training data, leading to an increase in overlap due to the creation of shared first subwords. However, the extent of this increase differs between the datasets. In ANERCorp, the degree of token overlap in the test split, while still present, shows a lower overall increase compared to the training split, with the overlap concentrated mainly around the boundaries between PER entity tags and the O tag. Conversely, CoNLL-2003 maintains a relatively consistent level and distribution of token overlap across both its training and testing splits.

In summary, our analysis reveals distinct patterns of vocabulary overlap across entity tags in ANERCorp and CoNLL-2003 at both word and token levels, and across training and testing splits. Boundary overlap, particularly within PER entities, and overlap with the O tag are dominant in ANERCorp, while cross-entity type and boundary overlap is more notable in CoNLL-2003. Subword tokenization generally increases the degree of overlap in both datasets due to shared first subwords. However, the distribution of this increased overlap differs: in CoNLL-2003, it tends to spread across various entity tags, whereas in ANERCorp, especially in the training data, it reinforces the O tag and boundary overlaps. In the testing split, CoNLL-2003 maintains a consistent spread of token overlap, while in ANERCorp, it becomes more concentrated on O tag and boundary overlaps. These varying patterns of annotation ambiguity have potential implications for NER model training and performance in each language.

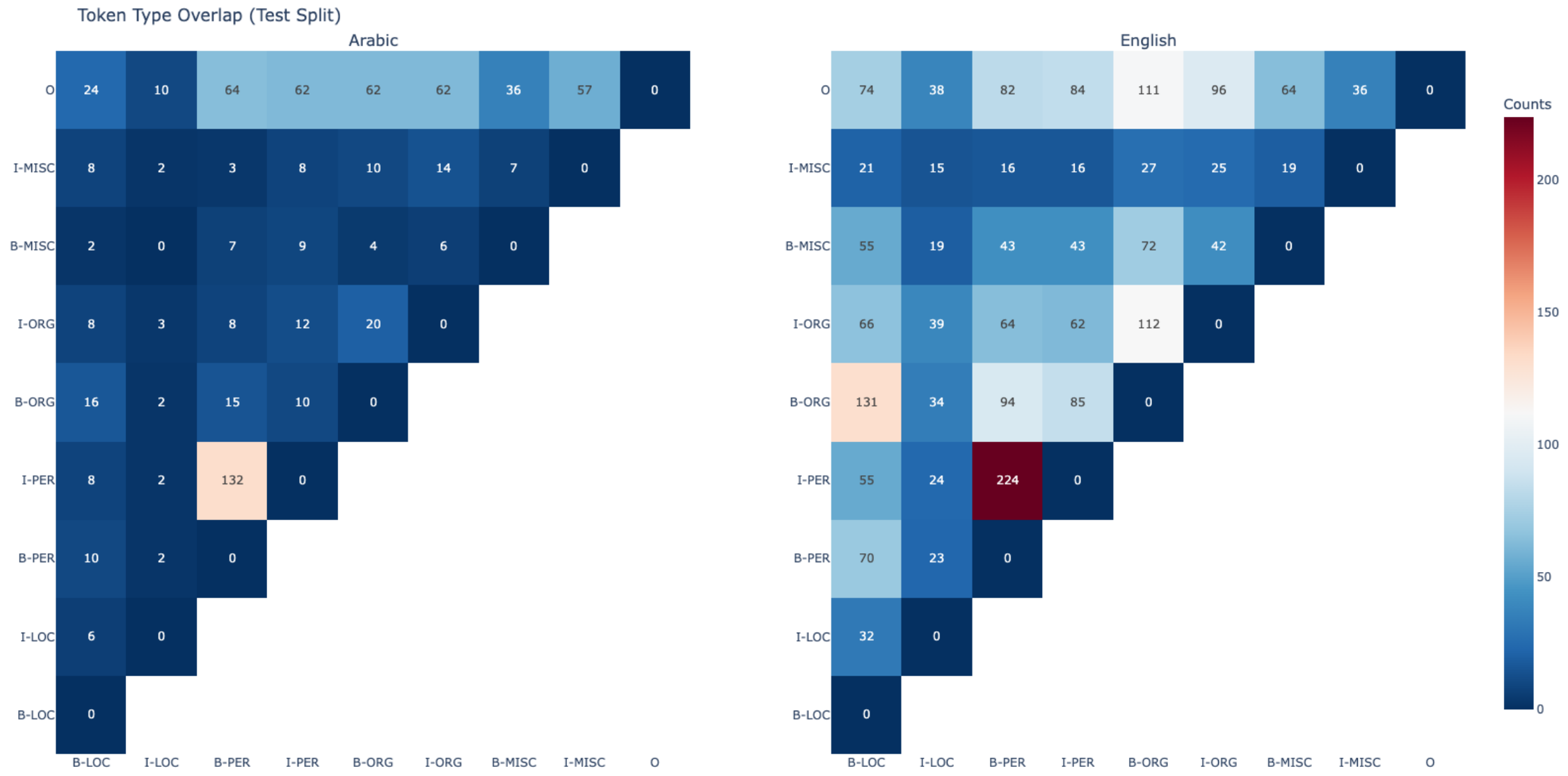

Figure 4.18: Token-type overlap across entity tags in the test split of Arabic (ANERCorp) and English (CoNLL-2003). Diagonal values are zeroed out to focus on inter-tag overlap.

### 4.3.4 Token Level Behaviour

The previous sections examined lexical and structural properties of the datasets. In this section, we shift to a more granular perspective by analysing individual tokens. Each token is assigned a score that captures a specific behavioural property, and these scores can be aggregated to provide an overall view across entity tags. This token-level perspective allows us to link dataset characteristics more directly to model behaviour, since both are expressed at the level of tokens. To address aspect (e) of RQ1, this section introduces three token-level behavioural metrics—ambiguity, consistency, and tokenisation—which are analysed here and revisited in Chapter 5.

The overlap in the datasets, at both word and token levels, can occur due to few reasons: linguistic ambiguity of the words, the effects of subword tokenization, or inconsistencies in annotation. Overlap in the training data can confuse the model during learning, while overlap in the test data can complicate inference and affect the model's ability to accurately distinguish between entity tags. To quantify these aspects at the token level, we developed three key metrics. Tokenization Rate measures the degree to which individual words are segmented into subword tokens. This metric provides insight into the granularity of the tokenization process as it applies to each word in the dataset. Ambiguity focuses on the interaction between the training and test sets. It quantifies the annotation uncertainty associated with a specific token in the test set by examining the distribution of its annotations within the training data. This allows us to identify tokens that exhibit variable labelling in the training set, potentially indicating ambiguity. Consistency assesses the alignment between the annotation of a token in the test set and its annotation patterns in the training set. This metric helps determine how reliably a token is annotated with the same entity tag across both splits. Together, these token-level behavioural metrics allow us to examine the impact of tokenization and annotation overlap on individual data points.

**Tokenisation Rate** Starting with tokenisation, we examine tokenization rates at the entity tag level. Tokenization rate is defined as the average number of subword tokens produced per word. Figure 4.19 presents mean tokenization rates (± standard deviation) per entity tag for Arabic and English.

In English, the entity tag with the highest average tokenization rate is `I-PER` (around 2.34), with `B-ORG` also showing a relatively high rate (around 2.12). The lowest tokenization rate for English entities is observed for `B-LOC` (around 1.78). The `O` tag in English has a rate of approximately 1.25.

In Arabic, the entity tags generally exhibit lower tokenization rates. The highest rates are for `B-PER` and `I-PER` (both around 1.44-1.49), which are still lower than their English counterparts. The lowest rates in Arabic are for `B-LOC` (around 1.16) and I-ORG (around 1.12). The O tag in Arabic has a rate of approximately 1.07.

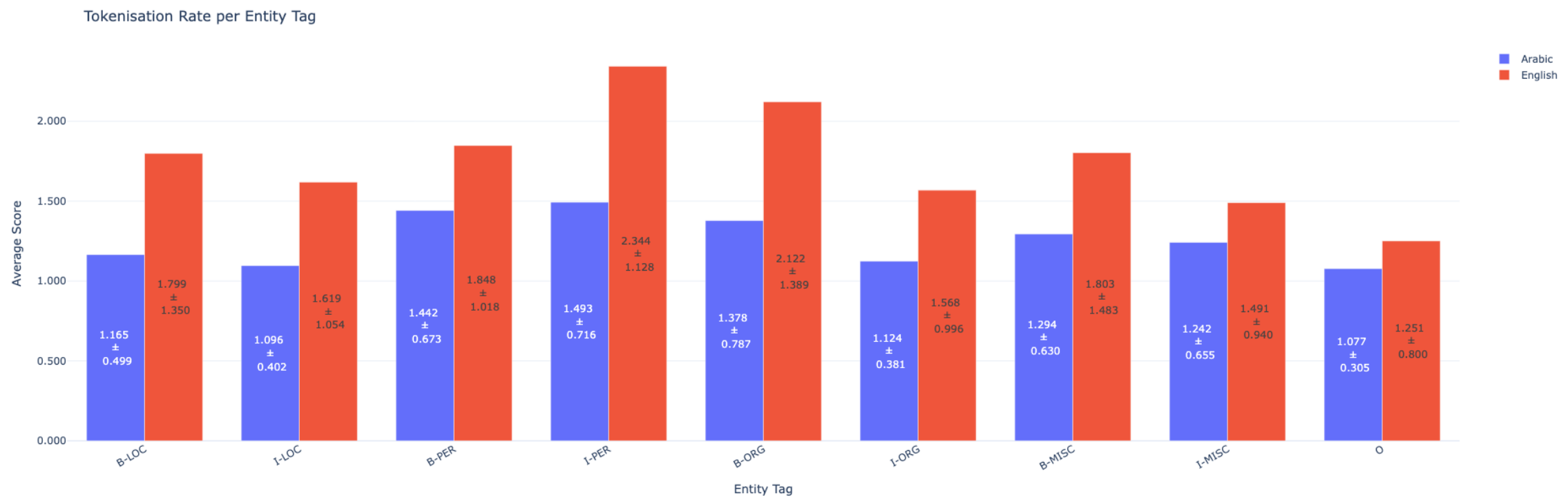

Figure 4.19: Tokenization Rate Per Entity Tag in Arabic and English. Values represent the average number of subword units per word (± std) within each entity tag.

The variability in tokenisation rate distribution appears larger for English entity tags compared to Arabic, suggesting higher inconsistency in the number of tokens per word within the same entity tag in English. For example, I-PER and B-ORG in English show relatively large standard deviations. By contrast, Arabic entity tags display more stable tokenisation rates within each tag.

In summary, the analysis of tokenization rates reveals a consistent trend of higher rates in English compared to Arabic across all entity tags, with higher variability observed within English tags. The most notable differences are seen in I-PER and B-ORG. These varying tokenization patterns likely contribute to differences in how NER models process and learn from the two languages.

**Ambiguity Analysis** Switching to ambiguity analysis we examine annotation uncertainty across entity tags. Ambiguity is computed as the normalized entropy of a token's label distribution in the training data, averaged across all tokens within each tag. Higher entropy indicates higher uncertainty in how tokens are annotated, which can indicate potential challenges for model learning. Figure 4.20 presents mean ambiguity scores (± standard deviation) for each entity tag, comparing Arabic and English at both the word and token levels.

For **Arabic** (top row), ambiguity is relatively balanced between the word and token levels, with tokenisation producing only small increases. Tags such as I-LOC, B-PER, and I-ORG display relatively high ambiguity, with I-LOC showing one of the highest mean values (0.556 ± 0.286 at the word level). By contrast, O tokens remain highly stable, exhibiting minimal ambiguity across both representations.

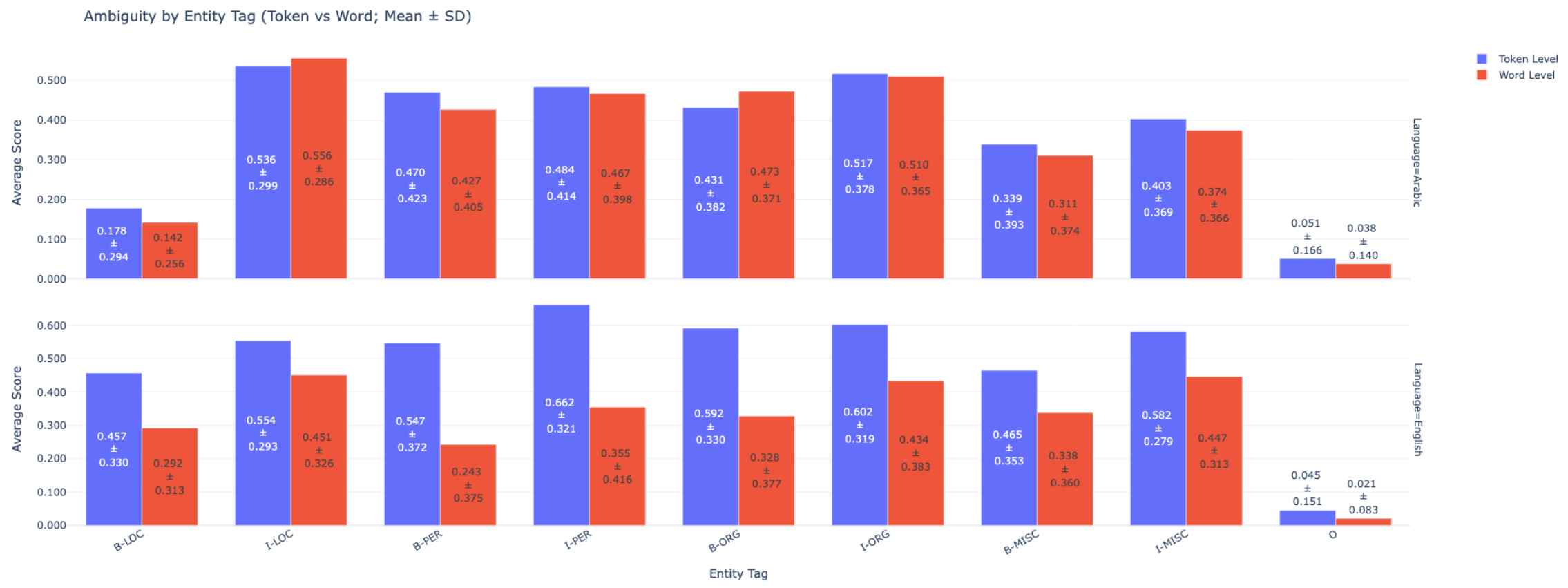

Figure 4.20: Entity-Tag Ambiguity Comparison at the Word and Token Levels for Arabic and English. Mean normalized entropy (± std) is shown per entity tag.

For **English** (bottom row), tokenisation lead to a higher increase in ambiguity. Continuation tags such as I-PER and I-MISC show more increase at the token level compared to the word level, suggesting that subword segmentation introduces greater uncertainty. Beginning tags (B-ORG, B-LOC) also show increased token-level ambiguity relative to their Arabic counterparts, highlighting the stronger effect of subword overlap in English. Similarly, the O tag shows the lowest ambiguity, confirming the stability of non-entity annotation.

Across both languages, continuation tags (I-) generally show higher ambiguity than beginning tags (B-), reflecting greater uncertainty in entity boundaries. Overall, these patterns indicate that tokenisation amplifies annotation variability more in English than in Arabic, particularly for semantically broad or morphologically complex categories like I-MISC, I-PER, and I-LOC.

**Consistency Analysis** Consistency measures the stability of annotations across token occurrences, while inconsistency reflects variability in how tokens are labelled. Since the two are complementary (i.e., one minus the other), they are presented side by side for clarity. Figure 4.21 compares consistency (blue) and inconsistency (red) ratios by entity tag for Arabic (top row) and English (bottom row).

For **Arabic**, inconsistency is higher for the majority of tags, particularly I-MISC (0.936 ± 0.176), I-LOC, and I-ORG, indicating substantial annotation inconsistency within these categories. Beginning tags such as B-PER and B-MISC also show increased inconsistency, suggesting less stable annotation practices for both inner and boundary positions. By contrast, B-LOC and O tokens exhibit strong consistency (0.848 ± 0.301 and 0.977 ± 0.103, respectively), pointing to stable labelling for locations and non-entity tokens.

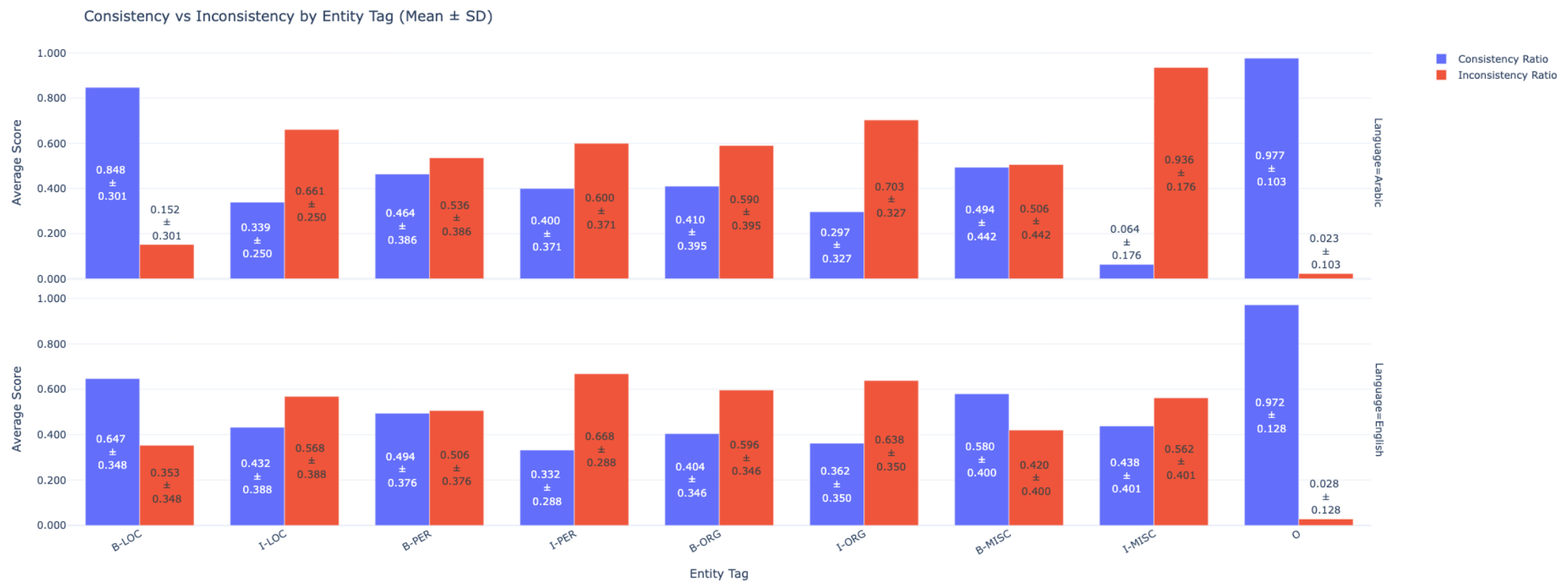

Figure 4.21: Consistency (blue) and inconsistency (red) ratios by entity tag for Arabic (top row) and English (bottom row). Bars show mean values with standard deviations.

For **English**, consistency is more evenly distributed across tags. While inconsistency still appears in continuation tags such as I-PER and I-ORG, the inconsistency ratios are generally lower than in Arabic. The O tag similarly shows near-perfect consistency (0.972 ± 0.128), confirming reliable annotation of non-entities. Compared to Arabic, English annotations appear more stable across a wider range of entity types, with fewer extreme imbalances between tags.

Overall, these results suggest that Arabic annotation stability is concentrated in specific categories such as locations and non-entities, whereas English maintains broader cross-tag consistency. This points to greater annotation challenges in Arabic, particularly for complex or ambiguous categories like I-MISC and I-ORG.

In summary, our token-level behavioural analysis reveals distinct cross-linguistic and cross-tag patterns in how data is tokenized and annotated in Arabic and English. English exhibits higher tokenization rates across all entity tags, along with higher variability, particularly in I-PER and B-ORG, suggesting that English entity tokens are more frequently split into multiple subword units. This, in turn, amplifies annotation ambiguity, especially at the token level, where English shows increased uncertainty in labels for I-MISC, I-PER, and B-LOC. Arabic, while generally exhibiting lower tokenization rates and less increase in ambiguity post-tokenization, shows higher annotation inconsistency for most tags—most notably I-MISC, I-LOC, and I-ORG—suggesting less stable annotation practices tags that are harder to label or appear within multi-word entities. English annotations, by contrast, are more consistent across a broader range of entity tags.

Taken together, these results indicate that subword tokenization and annotation variability interact differently across the two languages, with Arabic exhibiting higher instability in tag assignment and English experiencing higher tokenisation rate and ambiguity. Understanding the interaction between these token-level behaviours helps understand where models may struggle, particularly in handling ambiguous or inconsistently labelled entities, as explored further in Chapter 5.

### 4.3.5 Answer to RQ1: Data-Component

This section addressed **RQ1**, focusing on the dataset-level characteristics that might explain differences in NER system performance between Arabic (ANERCorp) and English (CoNLL-2003). We systematically examined these properties across three main dimensions: structural characteristics, lexical characteristics, and token-level behavioural metrics. The following summarises the key findings for each of these aspects.

**Aspect a) Structural Differences Between Datasets** CoNLL-2003 is significantly larger than ANERCorp, containing more words overall and a higher proportion of named entities. Entity span distributions are more balanced in CoNLL-2003, while ANERCorp shows stronger variations between training and test splits, particularly a shift from location (LOC) dominance in training to person (PER) dominance in testing. These structural discrepancies suggest that CoNLL-2003 offers a more stable annotation for model learning.

**Aspect b) Impact of Annotation Schemes on Span Structures** Annotation scheme effects are notable in ANERCorp due to gold-standard inconsistencies, but negligible in CoNLL-2003. ANERCorp exhibits annotation inconsistencies that violate IOB2 scheme, leading to the removal of entity spans under Seqeval's discard mechanism. Switching from IOB1 to IOB2 leads to the loss of 126 entity spans in ANERCorp, mainly due to invalid I-tag transitions. These inconsistencies highlight the need for careful checking of annotation guidelines when using multilingual datasets.

**Aspect c) Lexical Behaviour Across Languages** ANERCorp exhibits higher lexical diversity than CoNLL-2003, both at the overall dataset level and within entity tags. This higher diversity stems from the broader range of data sources and linguistic variability in Arabic. However, this variability also correlates with higher rates of unseen vocabulary (OOV) across entity tags like I-MISC and I-ORG. In contrast, CoNLL-2003 has lower diversity but more repetition, particularly within dominant tags like B-LOC. These lexical differences imply that models trained on Arabic data need to generalise across a wider variety of forms.

**Aspect d) Tokenisation Effects on Lexical and Structural Patterns** The impact of tokenisation is stronger in English than in Arabic, fragmenting words into subwords and reducing the number of unique types. Subword tokenisation increases overlap between entity tags, particularly in English, where entity boundaries become more blurred. While tokenisation helps reduce OOV rates in both languages, the reduction is more substantial in CoNLL-2003, leading to a more repetitive and concentrated subword vocabulary. In ANERCorp, tokenisation has a smaller effect: it reduces the vocabulary less, allowing more lexical diversity to be preserved. However, it also preserves more linguistic ambiguity, as many rare or complex forms remain ambiguous after tokenisation.

**Aspect e) Token-Level Behavioural Metrics and Annotation Variability** Token-level behavioural metrics reveal that English exhibits higher tokenisation rates and higher ambiguity after tokenisation, particularly for continuation tags like `I-PER` and `I-MISC`. Arabic, while less affected by tokenisation, shows higher annotation inconsistency across most tags, especially `I-MISC`, `I-LOC`, and `I-ORG`. This suggests that Arabic data is more prone to annotation instability, particularly for complex or ambiguous entity types. Although English data has higher tokenisation rate, it remains more consistent in its annotations. In contrast, Arabic data, despite lower tokenisation rates, shows higher inconsistency in how tokens are labelled. This difference means that Arabic models are more likely to struggle with unstable annotations.

**Conclusion** Together, these findings show that differences in NER model performance across Arabic and English are linked to structural, lexical, and behavioural properties of the datasets. In Arabic, structural inconsistencies and higher lexical diversity introduce annotation challenges that can affect model learning stability, while in English, higher tokenisation rates lead to higher ambiguity and overlap. Understanding these data-level patterns, and how they connect to model behaviour, offers a clearer explanation of the performance discrepancies, as further discussed in Chapter 5. In the next section, we analyse model behaviour, exploring model-specific insights that can help explain and contextualise the performance differences observed between Arabic and English.

## 4.4 Model-Component

Building on the token-level behavioural metrics introduced in Section 4.3, this section addresses **Research Question 2 (RQ2)** by examining model behaviour at the token level with respect to internal subcomponents. The aim is to understand how differences between Arabic and English NER can be explained not only by properties of the data, but also by the way models represent and process tokens. This connection between data characteristics and model behaviour provides a basis for explaining variation in NER performance across languages.

We examine model behaviour from three perspectives: **prediction confidence and uncertainty**, **token-level loss at inference time**, and **representation space separability**. Together, these measures quantify how confidently the model makes predictions, how errors are distributed, and how well different entity types are separated in the internal representation space.

To address RQ2, we consider the following aspects: **a)** what patterns in loss values reveal about model stability and errors during inference; **b)** how prediction confidence and uncertainty vary across languages and entity tags; **c)** how well the learned token representations separate entity types across languages, as assessed by silhouette scores.

**Extracted Outputs** Figure 4.22 illustrates the outputs extracted during the data extraction phase to support this analysis. These outputs include predicted probabilities (for confidence analysis), token-level loss values, and hidden representations from the last hidden states (for representation space analysis). All outputs are precomputed during model evaluation and dynamically analysed within the dashboard.

Unlike earlier sections that examined dataset properties from the top down, this section focuses directly on the **entity-tag level**. The analysis is organised into three stages—loss analysis, prediction confidence and distribution, and representation space analysis—each providing fine-grained insights into model behaviour across languages.

**Section Outline** This section is structured as follows:

- **Metric Definitions and Analysis Approach:** Description of the model-specific metrics and how they are derived from internal subcomponents.
- **Loss Analysis:** Evaluation of token-level loss values to identify patterns of stability and error distribution.
- **Prediction Confidence and Distribution:** Analysis of probability distributions and confidence scores across entity types and languages.
- **Representation Space Analysis:** Examination of entity separability in the hidden space using silhouette scores.

### 4.4.1 Metric Definitions and Analysis Approach

This section introduces the key metrics used to analyse model behaviour at the token level. The analysis focuses on outputs extracted from the model at inference time, including loss values, predicted probabilities, and hidden representations. All metrics are computed over the **test set tokens** and are **broken down by entity tag**. The metrics presented here form the basis for the subsequent behavioural analyses: loss analysis, prediction confidence analysis, and representation space analysis.

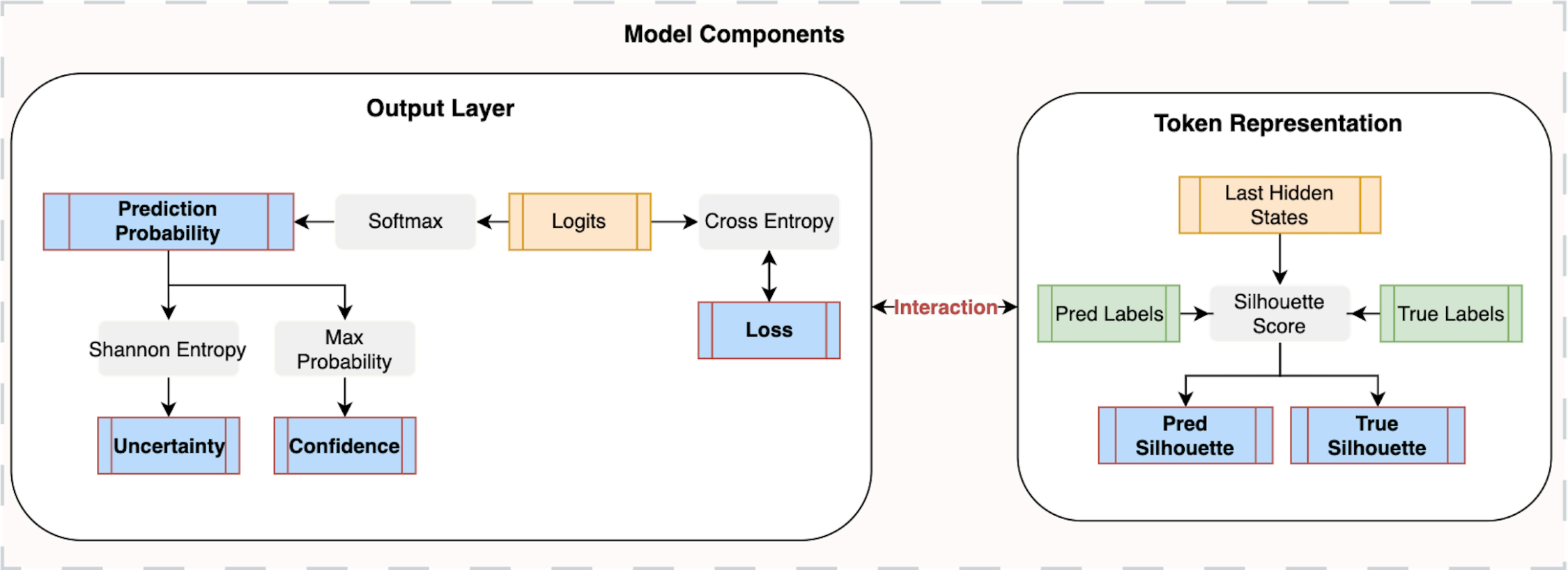

Figure 4.22: Model analysis pipeline in DeformAr. At inference time, the framework extracts predicted probabilities, token-level loss values, and hidden representations, which are then analysed to evaluate confidence, error distributions, and representation space separability.

**Loss Metric** Loss values quantify how far the model's predictions deviate from the correct labels. In this analysis, we focus on **cross-entropy loss**, a widely used metric for classification tasks such as NER. Minimising loss during training improves the model's ability to generalise to unseen data.

Cross-entropy loss measures the difference between the predicted probability distribution and the true labels, the loss for a single prediction is calculated as:

$$\text{Loss} = -\sum_{i=1}^{C} y_i \log(p_i) \tag{4.4}$$

where $C$ is the number of classes, $y_i$ is a binary indicator of whether class $i$ is the correct label, and $p_i$ is the predicted probability for class $i$. This formulation penalises the model more when it assigns low probability to the correct label and produces smaller values when the correct label is assigned high probability (e.g., $p_i = 0.9$).

In this analysis, we compute the **per-token loss** for each token individually. This allows us to identify specific tokens and entity types where the model exhibits higher uncertainty or makes more prediction errors.

**Relation to Prediction Confidence** While loss and prediction confidence are related, they capture different aspects of model behaviour. **Loss** measures the discrepancy between the predicted probability distribution and the true label, taking into account whether the prediction is correct. **Prediction confidence**, in contrast, simply reflects how sure the model is about its top prediction, regardless of whether that prediction is right or wrong.

For example, a model could be highly confident about an incorrect prediction, leading to a high loss. Alternatively, a model could be uncertain (low confidence) but still assign the correct label the highest probability, resulting in a lower loss. Thus, loss provides a supervised indicator tied to correctness, whereas confidence reflects the model's certainty. In the next paragraphs, we define how prediction confidence and probability distributions are computed and how they are used to characterise model behaviour.

**Prediction Confidence and Probability Distribution** This section introduces three prediction-level metrics derived from the model's outputs at inference time: **token confidence**, **prediction uncertainty**, and the **prediction probability distribution**. Each of these captures different aspects of model behaviour and complements the loss analysis.

Following the final output layer (typically a Multi-Layer Perceptron), the model produces raw unnormalised scores called *logits*. Each logit reflects the model's raw confidence for a particular entity tag, but these scores are not directly interpretable as probabilities. To transform them into a probability distribution across all possible tags, the logits are passed through a softmax function.

$$P(\text{tag}_i \mid \text{token}) = \frac{e^{\text{logit}_i}}{\sum_j e^{\text{logit}_j}} \tag{4.5}$$

Here, $P(\text{tag}_i \mid \text{token})$ denotes the probability of assigning tag $i$ to the token, ensuring that all probabilities are positive and sum to one. The softmax function transform each logit and normalises the result so that the output values lie between 0 and 1 and sum to 1 across all classes, producing a valid probability distribution. This probability distribution forms the basis for the following prediction metrics.

**Prediction Probability Distribution** The prediction probability distribution provides a full view of how the model allocates probability across all entity tags for a given token. Rather than focusing only on the most probable tag or the overall confidence, the full distribution across all entity tags (nine tags) is provided for each token.

The prediction probability distribution is particularly useful for identifying whether uncertainty arises from confusion between a few closely competing classes or from a lack of distinction across many classes.

**Prediction Uncertainty** Prediction uncertainty captures the overall spread of the probability distribution across all possible entity tags. It is quantified using Shannon entropy, as previously introduced in Equation 4.3. To allow comparison across different numbers of classes, we compute the **normalised prediction uncertainty**:

$$\text{Normalised Uncertainty} = \frac{H(X)}{\log(C)} \tag{4.6}$$

where $C$ is the number of possible entity tags. Normalised uncertainty ranges from 0 (complete certainty) to 1 (maximum uncertainty, where probabilities are uniformly distributed).

Unlike token confidence, which focuses only on the top prediction, prediction uncertainty considers the entire distribution to assess how confidently the model differentiates between classes.

**Token Confidence** Token confidence quantifies the model's certainty about its top prediction for each token. It is defined as the highest probability assigned to any tag:

$$\text{Token Confidence} = \max\left(P(\text{tag}_i \mid \text{token})\right) \tag{4.7}$$

A higher token confidence score indicates that the model is highly certain about its chosen label, while a lower score suggests greater uncertainty.

**Silhouette Scores** The silhouette score is a metric used to evaluate how well data points are grouped, measuring both *compactness* within clusters and *separation* between clusters. Although traditionally applied in clustering analysis, in this thesis the silhouette score is used to assess how well the contextualized embeddings produced after fine-tuning distinguish between different NE tags.

Ideally, tokens labelled as B-LOC should have embeddings that are close to other B-LOC tokens (high compactness) and well separated from unrelated tags such as O (high separation). By computing silhouette scores, we quantify how effectively the model's learned representations capture these distinctions.

In this analysis, silhouette scores are computed as **per-token scores** using two sets of class labels: (1) **True label silhouette**, where the gold-standard entity tags assigned in the dataset are used; and (2) **Predicted label silhouette**, where the model's predicted entity tags at inference time are used.

For a given token $i$, the silhouette score is computed as:

$$S(i) = \frac{b(i) - a(i)}{\max(a(i), b(i))} \tag{4.8}$$

where $a(i)$ represents the average distance between token $i$ and all other tokens within the same class, capturing the compactness of the cluster. $b(i)$ represents the average distance between token $i$ and the nearest neighbouring class, capturing the separation between clusters.

This dual computation allows us to measure both how well the learned representations align with the gold-standard (true) labels, and how well they align with the model's own predictions.

The silhouette score ranges from $-1$ to 1: - A score near 1 indicates that a token's embedding is well clustered with others of the same tag and clearly separated from different tags. - A score near 0 suggests that the token lies close to the boundary between classes, reflecting ambiguous representation. - A negative score means the token is closer to embeddings from another class than its own, indicating poor separation and potential misclassification.

In the following sections, we first analyse token-level loss to characterise model errors, then examine prediction confidence and uncertainty to evaluate the reliability of predictions, and finally use silhouette scores to assess how well the learned representation space separates different entity tags.

### 4.4.2 Loss Analysis

To address aspect (a) of RQ2—*what patterns in loss values reveal about model stability and errors during inference*—we analyse the loss values produced by the model at inference time, focusing on differences across entity tags and languages.

Loss is traditionally used during training for model optimization, but it also serves as a useful error metric during inference. By comparing predicted outputs to true labels, loss measures prediction errors and provides insights into a model's weaknesses. A high loss indicates a large disparity between predictions and true labels, while a low loss suggests fewer errors and better model performance.

As discussed in Section 4.4.1, loss relies on the logits to generate its values. In this section, we examine the outputs resulting from the interaction between the loss function and the outputs of the classification layer.

Figure 4.23 presents the average loss values broken down by entity tag. Arabic exhibits higher loss than English across most entity tags, except for O and B-LOC. In Arabic, the highest loss is observed for I-MISC, followed by B-MISC, I-ORG, and B-ORG—consistent with performance metrics showing lower F1 scores for these tags. For English, the highest losses are similarly seen in I-MISC and B-MISC, while other tags show relatively low loss values.

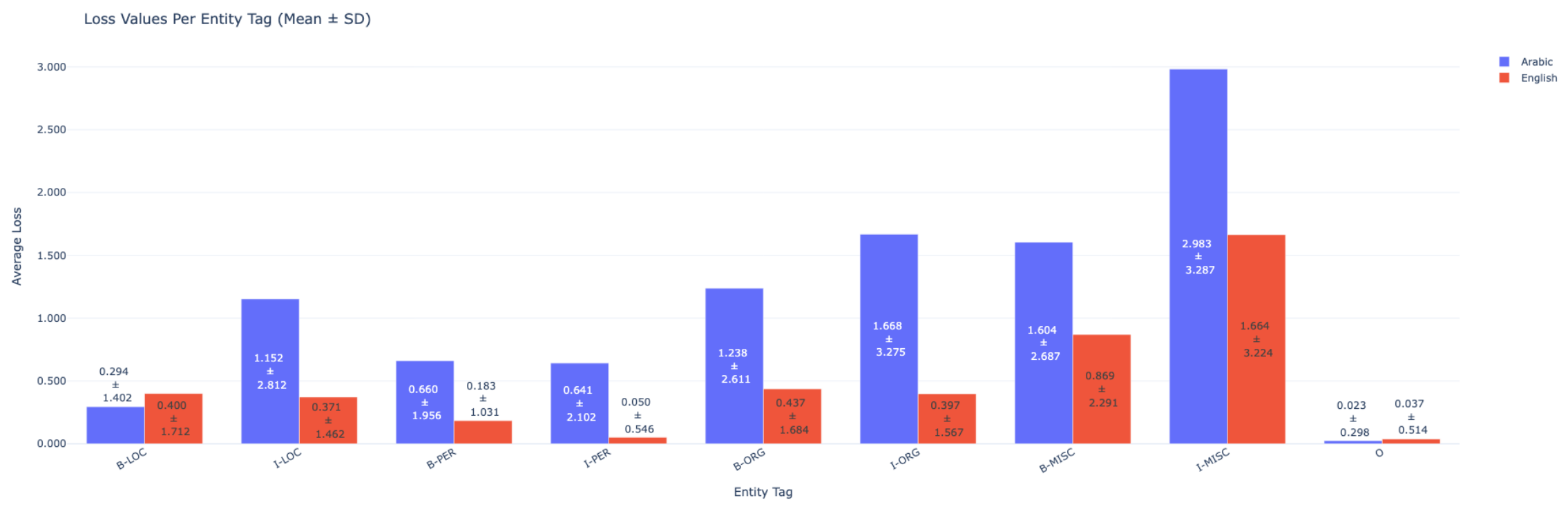

Figure 4.23: Entity-Tag Loss Analysis for Arabic and English models. This figure presents the average loss scores per entity tag for both languages.

It is important to note that loss penalises incorrect predictions more when confidence is high. Even when the model predicts correctly but with low confidence, the loss value still increases, as the probability assigned to the correct class is not high enough. Therefore, loss is directly influenced by the confidence distribution.

Interestingly, while B-LOC in the English model shows higher loss compared to other English tags, it still achieves a slightly higher F1-score than Arabic (as discussed in Section 4.5). This suggests that the English model is confidently making certain types of errors on B-LOC tokens. To investigate this, we examine the prediction probability distributions across entity tags. For each token, the model assigns a probability to each of the nine possible tags; we then average these values separately for correct and incorrect predictions. As shown in Figure 4.24, the English model tends to assign higher probabilities to B-LOC—even in incorrect predictions—than to correct ones. This supports the interpretation that confident misclassifications drive the higher loss values observed for B-LOC, despite its stronger overall F1 performance.

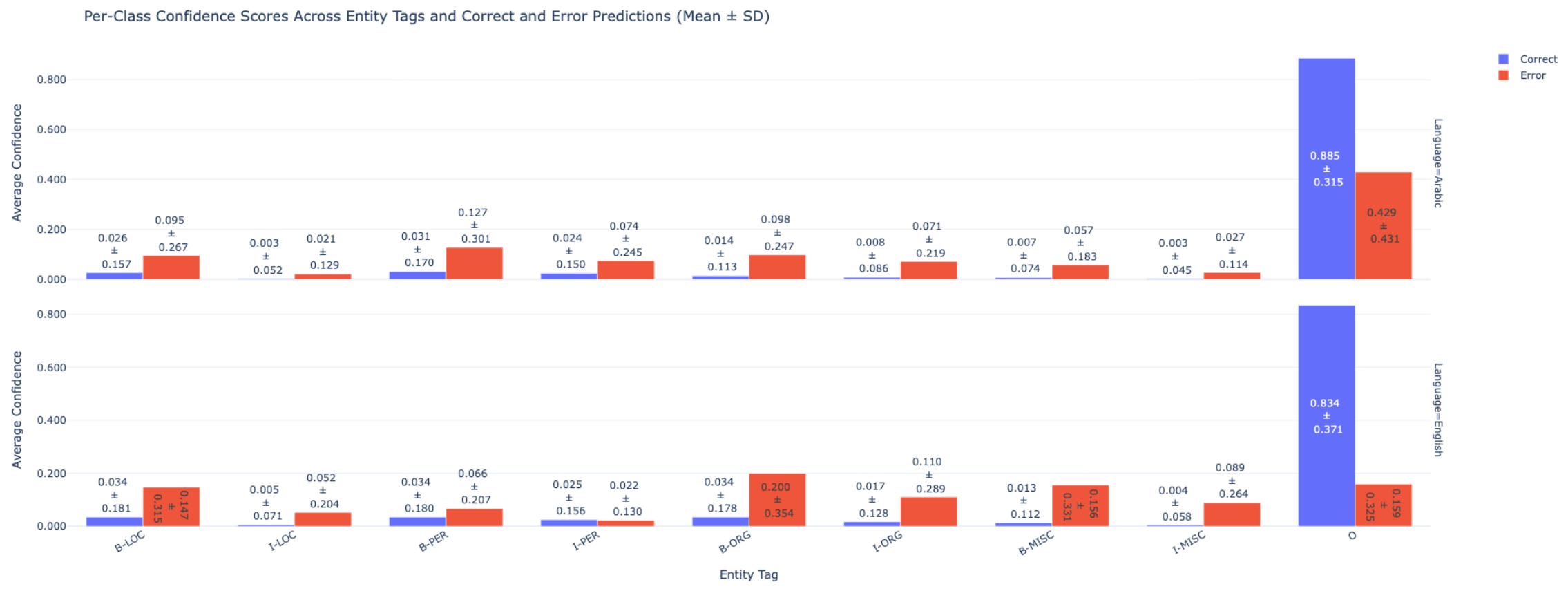

Figure 4.24: Confidence scores across entity tags in correct and incorrect predictions. Bars show the average probability assigned to each entity tag across all tokens.

These relationships between loss and prediction confidence are explored further in Chapter 5. In the next section, we examine confidence scores and probability distributions across entity tags.

### 4.4.3 Prediction Confidence Analysis

To address aspect (b) of RQ2—*how prediction confidence and uncertainty vary across languages and entity tags*—we analyse the model's prediction distributions and how they relate to both uncertainty and confidence. Specifically, we examine the entropy-based uncertainty within this distribution, and the model's confidence in its top prediction.

**Prediction Uncertainty** Figure 4.25 shows average normalised uncertainty scores across entity tags, with correct and incorrect predictions distinguished by bar colour.

For **Arabic** (top row), uncertainty is relatively high even for correct predictions, especially in I-MISC (0.242) and B-MISC (0.140). This indicates that the Arabic model often predicts correctly while remaining uncertain. For incorrect predictions, uncertainty is lower in some cases (e.g., B-LOC, I-MISC), suggesting that the model can be confidently wrong—an indicator of mis-calibration.

For **English** (bottom row), correct predictions generally show low uncertainty (close to zero), reflecting more confident correct outputs. In contrast, incorrect predictions show substantially higher uncertainty across most tags, particularly I-LOC (0.250), I-PER (0.240), and B-PER (0.160). This suggests the English model tends to be uncertain when making errors, a sign of better calibration.

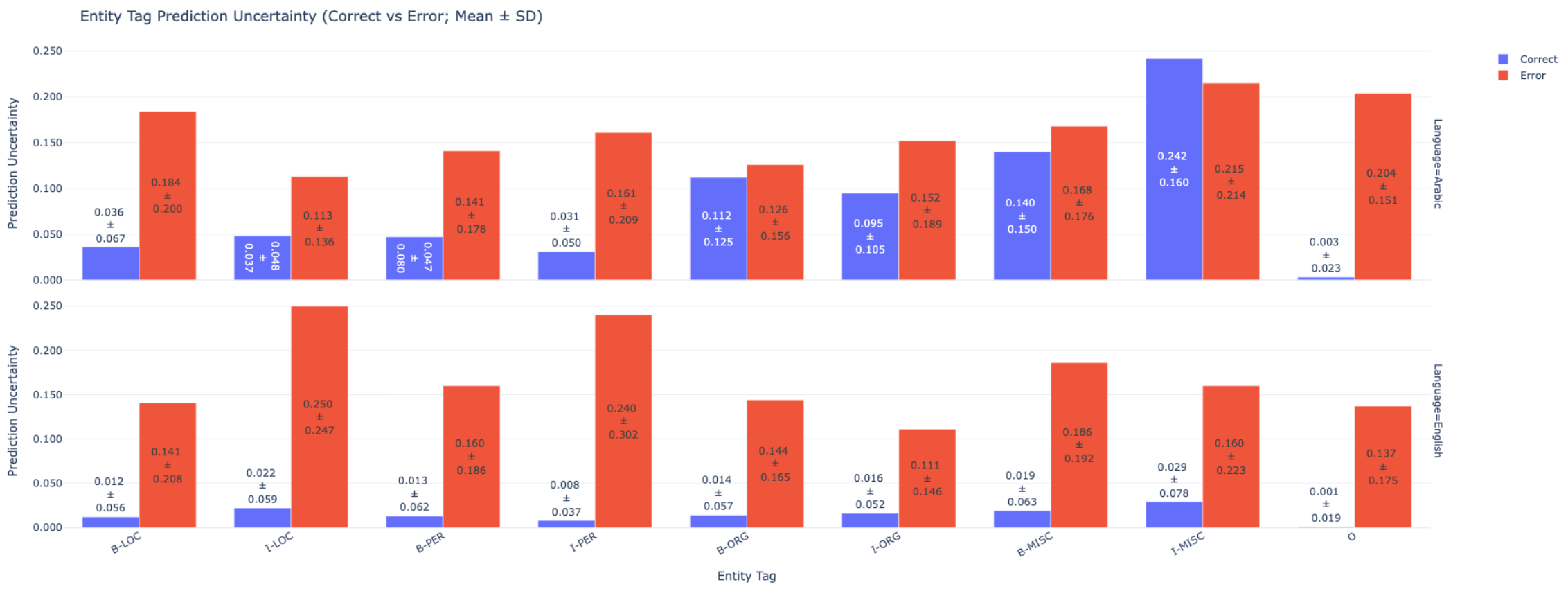

Figure 4.25: Prediction uncertainty (normalised entropy) by entity tag for Arabic (top row) and English (bottom row). Correct and incorrect predictions are shown in blue and red, respectively.

This pattern reinforces earlier observations about model calibration: the Arabic model tends to be less confident even in correct predictions, and more confident in incorrect ones. This imbalance suggests that the Arabic model's behaviour may be less well-calibrated. Moreover, the lower entropy values in its incorrect predictions indicate that the model's output distributions are often dominated by a single (incorrect) class, rather than being uncertain across multiple classes. This pattern reflects systematic overconfidence and a tendency to make high-certainty errors.

*Model calibration* refers to how well a model's predicted probabilities reflect the actual likelihood of being correct. A well-calibrated model assigns high confidence when it is likely to be correct and low confidence when it is likely to be wrong. Poor calibration, by contrast, can lead to overconfident errors or uncertain correct predictions, both of which can reduce trust in the model's outputs.

**Prediction Confidence** We next analyse **token-level confidence**—the maximum probability assigned to a tag during inference, reflecting how confident the model is about its top prediction (see Section 4.4.1). Figure 4.26 shows average token confidence across entity tags, with correct and incorrect predictions distinguished by colour. Across both languages and for most entity tags, the model shows higher average confidence when it predicts correctly. This aligns with expectations and reflects the model's ability to recognise when it is likely to be right.

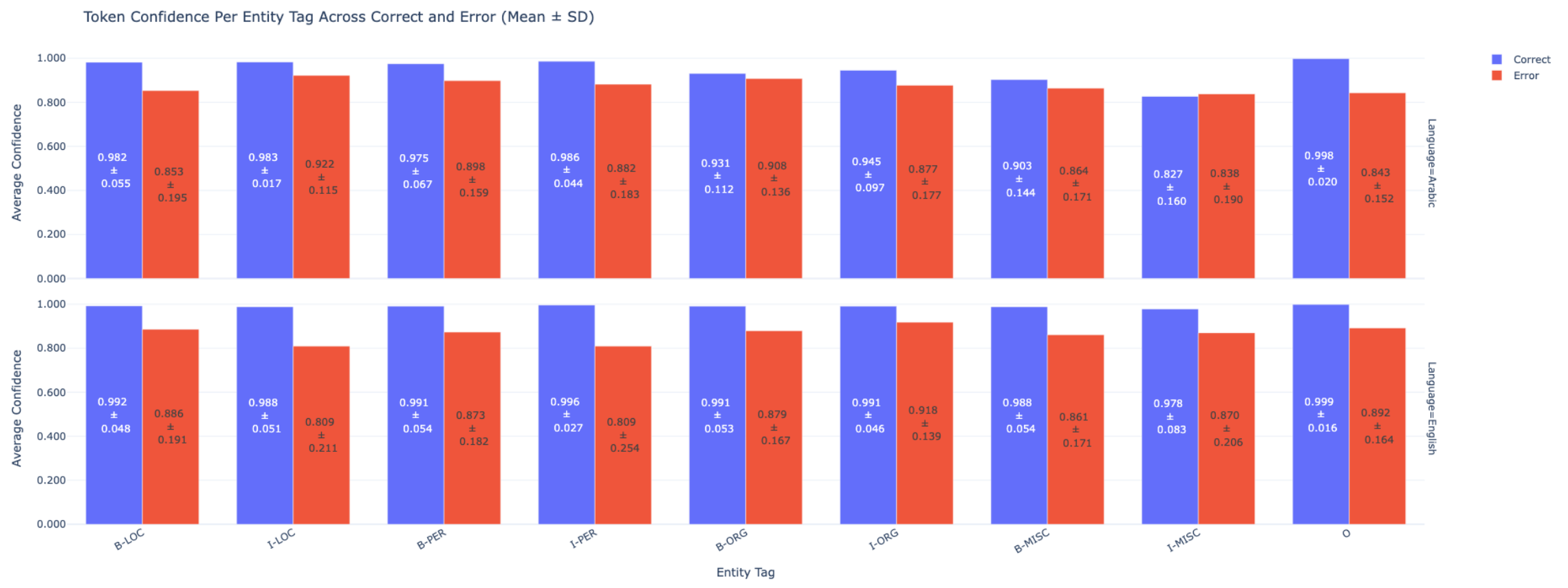

Figure 4.26: Token confidence by entity tag for Arabic (top row) and English (bottom row). Correct and incorrect predictions are shown in blue and red, respectively.

For **correct predictions**, the O tag exhibits the highest confidence in both languages, which is expected given its high frequency and generally lower ambiguity. Among the named entities, English shows consistently higher confidence than Arabic, especially for the B-MISC, B-ORG, and I-ORG tags. This indicates that the English model is more confident in its entity recognition across categories.

For **incorrect predictions**, confidence levels drop slightly, but many tags still show surprisingly high scores. In Arabic, entity tags such as B-PER, I-PER, B-ORG, and I-LOC continue to display high confidence, even when predictions are wrong. This suggests that the model is often confident about misclassifications—an indication of potential overconfidence. In comparison, the English model also shows high confidence in incorrect predictions and, tends to be more confident than the Arabic. For example, in tags like B-MISC, I-MISC, O, and I-ORG, English exhibits noticeably higher average confidence despite being wrong.

To further investigate confidence in misclassifications, Figure 4.27 presents the **mis-**

**classification confidence heatmap**. Each cell shows the *total token confidence* assigned when the model incorrectly predicted a specific tag for a token with a different true label.

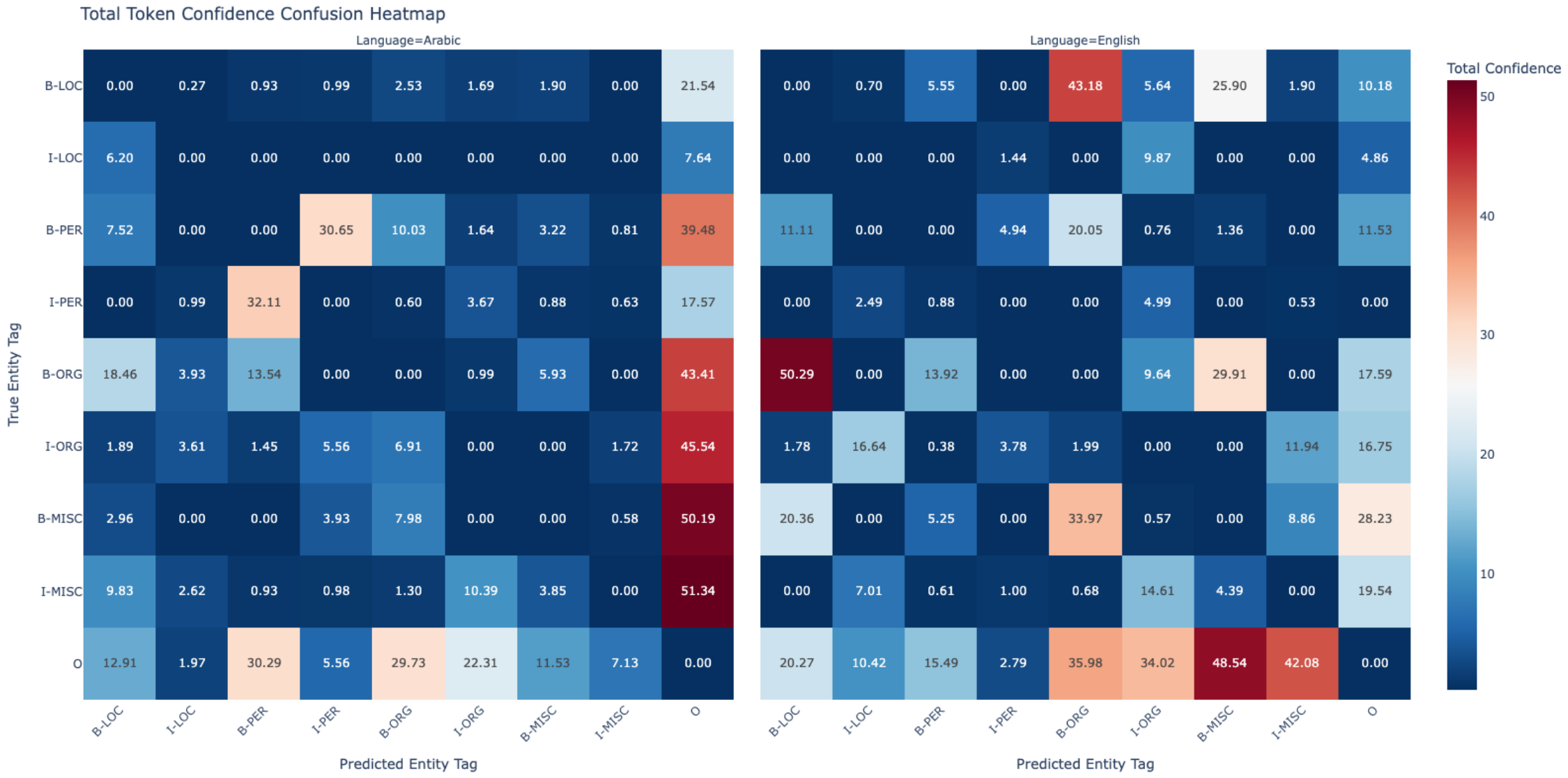

Figure 4.27: Misclassification confidence heatmap for Arabic and English models. Each cell shows the cumulative confidence assigned when tokens with a given true entity label (rows) were misclassified as another tag (columns). Darker shades indicate higher confidence in incorrect predictions.

In Arabic, the O tag is frequently misclassified with high confidence, particularly as B-MISC, I-MISC, B-ORG and I-ORG. There is also a tendency to confuse B-PER and I-PER, often with high confidence in both directions. These findings indicate that the Arabic model is prone to overconfident misclassifications, especially for exclusion-type errors—where entities are missed or labelled as non-entities—as well as boundary errors. In contrast, the English model exhibits a different pattern of high-confidence errors. Notably, it tends to confuse B-ORG and B-LOC, and frequently misclassifies O tokens as entity tags, suggesting more inclusion-type errors.

Together, the token confidence analysis and the heatmap highlight systematic differences in misclassification behaviour. Both models demonstrate high confidence in many of their incorrect predictions, but the nature of these errors differs by language. These trends help lay out the foundation for the error analysis presented in Section 4.5.

In the next section, we switch from output-layer metrics to examine the underlying **token representations** themselves, by analysing how well different entity types are separated based on **true labels** and **predicted labels** using silhouette scores.

### 4.4.4 Representation Space Analysis

To address aspect (c) of RQ2—*how well the learned token representations separate entity types across languages, as assessed by silhouette scores*—we analyse the separation of token representations produced by the model after fine-tuning. Using the silhouette score metric, we quantify how separable the internal representations are for different entity tags, computed separately with respect to the **true labels** and the **predicted labels**. Figure 4.28 presents the average silhouette scores for each entity tag, computed separately based on true and predicted labels.

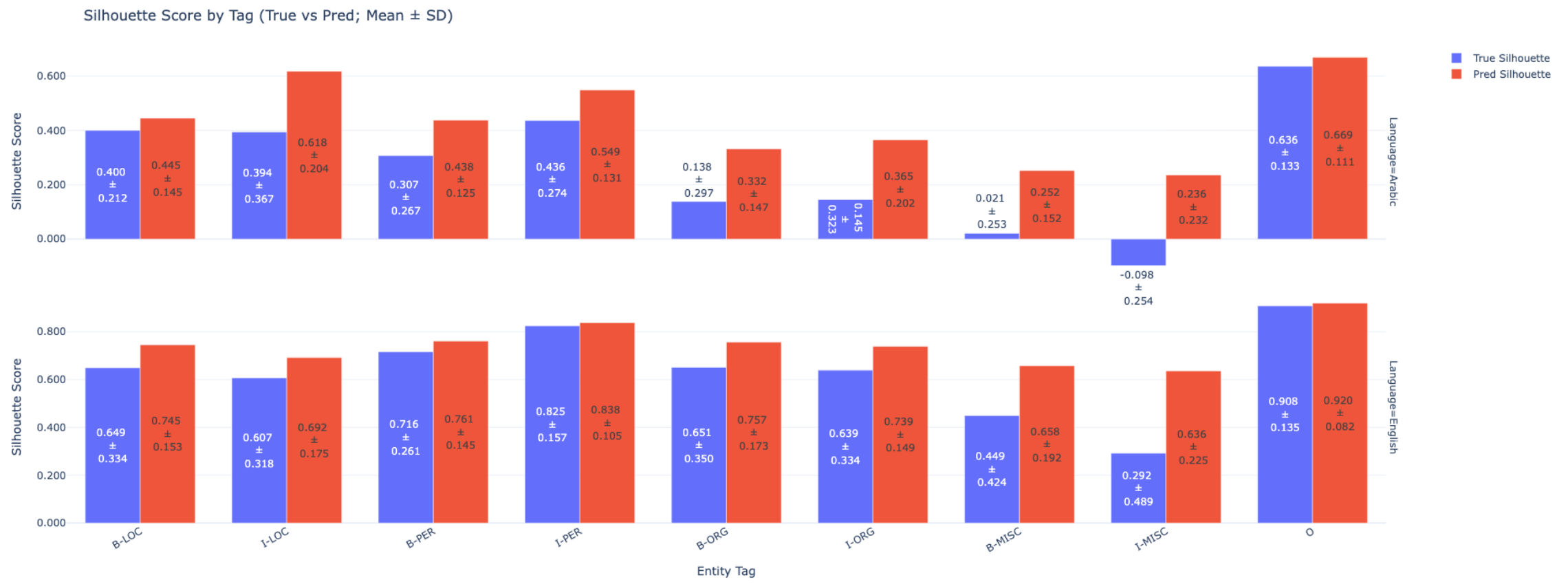

Figure 4.28: Entity-Tag Silhouette Score Analysis. This figure presents the average silhouette scores per entity tag for Arabic and English models.

Across both true and predicted labels, the English model consistently achieves higher silhouette scores than the Arabic model. This suggests that entity representations in English are more separated within the internal embedding space.

Focusing on **true label silhouette scores**, the O tag achieves the highest score in both languages, reflecting its dominance. Arabic shows relatively low separation, with `I-MISC` exhibiting a negative silhouette score, indicating that tokens labelled as I-MISC often cluster near non-I-MISC tokens. In English, the highest-scoring named entities are `I-PER` and `B-PER`, showing strong separability for person-related tags.

When comparing B- and I- chunks, the English model shows that `I-PER` and `I-ORG` are often better separated than their B- counterparts. In Arabic, this pattern is less consistent: `B-LOC` achieves higher separation than `I-LOC`, whereas the opposite is true for PER and ORG.

Turning to **predicted label silhouette scores**, we observe that the English model achieves higher scores across most entity tags. In both languages, predicted silhouette scores tend to be slightly higher than true label scores, indicating that the model's internal representation aligns better with its own predictions than with ground-truth labels. This behaviour could potentially be a sign of over-fitting, where the model's internal structure is more influenced by its own learned predictions than by the true underlying entity distinctions.

While higher silhouette scores indicate more coherent and well-separated internal representations, they do not necessarily imply correct predictions. A model may form tightly clustered representations even around incorrect or systematically misclassified labels. Although the embedding space shows organised clustering, the model's internal representations may still be **misaligned** with the true entity boundaries. These aspects—particularly how errors and inconsistencies affect internal clustering—are investigated further in Chapter 5.

Overall, these findings reveal that the English model produces more structured and well-separated internal representations of entity types, while the Arabic model's representation space is less partitioned, particularly for more ambiguous or under-represented entity categories.

### 4.4.5 Answer to RQ2: Model-Component

This section addressed **RQ2**, focusing on model-level behavioural patterns that explain performance differences between Arabic and English NER models. We examined model behaviour across three main aspects: token-level loss, prediction confidence and uncertainty, and internal representation structure.

**Aspect a) Patterns in Loss Values** Arabic models exhibit higher loss values across most entity tags compared to English, particularly for complex tags like `I-MISC` and `I-ORG`. Loss values reveal that Arabic models make more uncertain or overconfident errors, while English models, although not error-free, show lower average loss per tag. These patterns indicate greater prediction instability in Arabic models.

**Aspect b) Prediction Confidence and Uncertainty Patterns** Arabic models display higher uncertainty even in correct predictions and lower uncertainty in many incorrect ones, suggesting mis-calibration: they are less confident when right and more confident when wrong. English models show slightly better calibration but still demonstrate high-confidence errors. Confidence heatmaps reveal systematic misclassification patterns, particularly around common tags like `O` and `B-LOC`.

**Aspect c) Representation Space Separability** Internal representation analysis shows that English models form more structured and better-separated clusters across entity types, as measured by silhouette scores. Arabic models produce less distinct clusters, particularly for ambiguous categories such as `I-MISC` and `I-ORG`. Interestingly, clusters based on predicted labels often show higher separability than those based on true labels, suggesting that the model may be organising representations around its own learned decision patterns rather than reflecting the underlying gold-standard annotations.

**Conclusion** Together, these findings reveal that Arabic models are less stable across loss, confidence, and representation metrics. English models demonstrate better internal separation and slightly better calibration, although both models show patterns of systematic overconfidence. In Chapter 5, we further explore these model-level behavioural insights and examine the interactions between the different behavioural metrics.

## 4.5 Evaluation Component

This section addresses **Research Question 3 (RQ3)** by analysing the interaction between the gold-standard annotations (true labels) and the predicted labels produced by the model. This represents an interaction between the **annotation** subcomponent of the data and the **output layer** subcomponent of the model.

We examine this alignment at two levels. The **token level** evaluates the model's ability to correctly classify individual tokens according to flat BIO-style annotations. The **entity level**, on the other hand, evaluates the model's ability to correctly identify full entity spans and their types. Together, these perspectives offer complementary insights into how well predicted labels align with the gold-standard annotations, and how these alignments differ between Arabic and English.

To address RQ3 across both levels, we consider the following aspects: **a)** how well true and predicted labels are aligned across languages; **b)** what strengths and weaknesses exist in each language's performance; **c)** what types of errors affect performance at each level and how they differ across languages; **d)** for entity-level evaluation specifically, how the annotation scheme and scorer repair mechanisms (e.g., IOB1 vs. IOB2) influence span-level predictions and associated error types; and **e)** for token-level evaluation specifically, how performance varies in relation to label frequency, examined through a correlation-based analysis. These aspects are supported by three key outputs extracted during the data extraction phase, discussed next.

**Extracted Outputs** Figure 4.29 illustrates the outputs extracted during the data extraction phase that support this analysis. All outputs are derived from the comparison between true and predicted labels within the evaluation pipeline. These outputs map directly onto the aspects outlined above: the **classification report** provides the key evaluation metrics (precision, recall, and F1) that quantify overall alignment between true and predicted labels (aspect a); the **prediction outcomes** highlights strengths and weaknesses in model behaviour across entity classes by showing where predictions are correct or misclassified (aspect b); and the **error analysis** captures the types of errors made by the model across both levels (aspects c and d). All of these outputs are precomputed during the data extraction phase. The correlation analysis used for (aspect e) is calculated dynamically within the dashboard interface, based on the classification report outputs. Together, these components allow us to evaluate model performance at both the token and entity levels in a structured and interpretable manner.

This section adopts a top-down approach, beginning with high-level evaluation metrics and gradually narrowing the focus to more detailed analyses. We first outline the theoretical background of each metric and how it is calculated, then describe the evaluation methodology across token- and entity-level perspectives. The results are presented in a structured sequence—performance metrics, prediction outcomes, error analysis, and correlation analysis—each providing fine-grained insight into model performance across languages and evaluation levels.

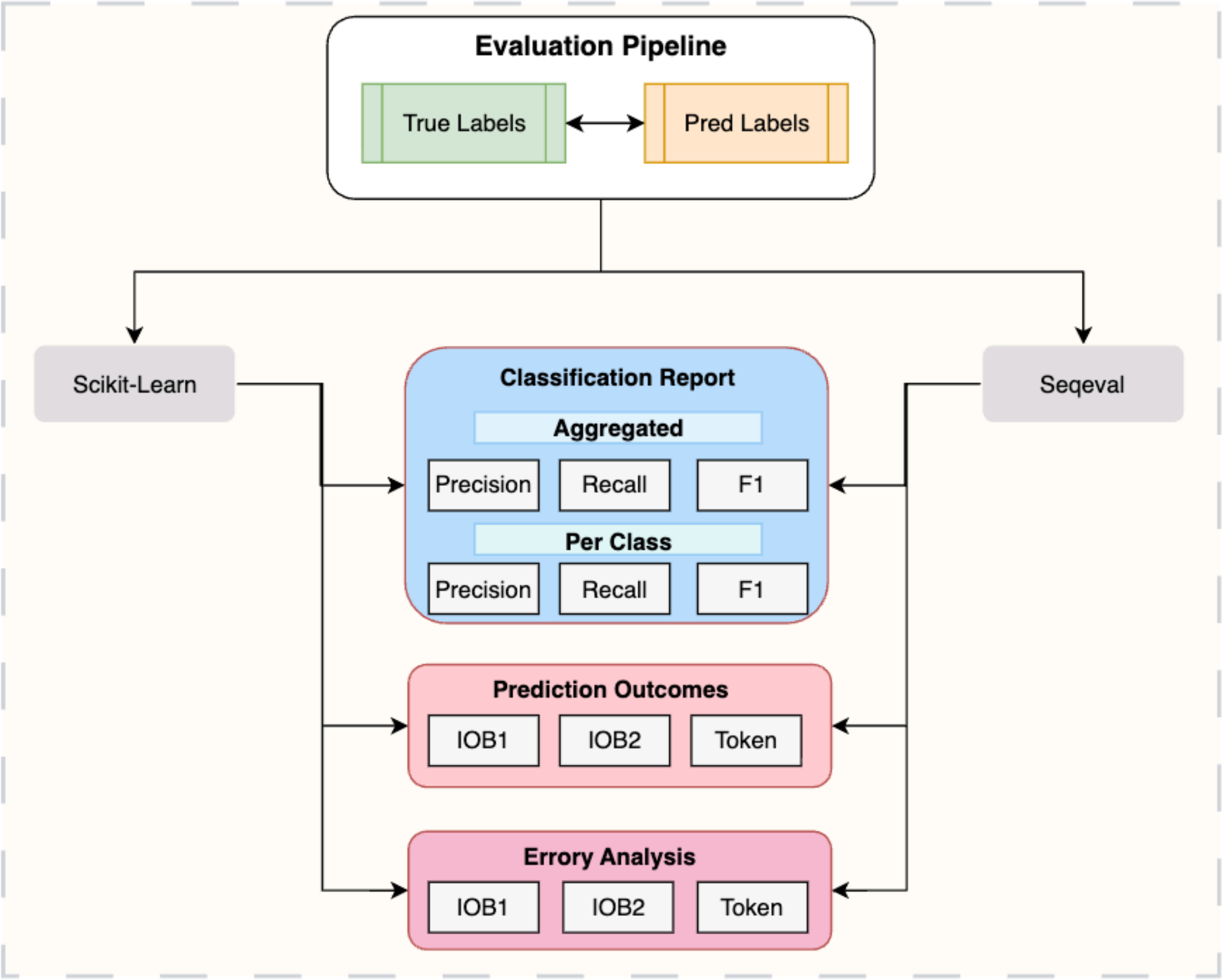

Figure 4.29: Evaluation pipeline in DeformAr. True and predicted labels are passed to two evaluation tools — `Scikit-learn` for token-level evaluation and `Seqeval` for entity-level evaluation. Outputs include classification reports (aggregated and per class), prediction outcomes, and error analysis across both IOB1 and IOB2 schemes.

**Section Outline** This section is structured as follows:

- **Theoretical Background:** Definitions of evaluation metrics at both token and entity levels.
- **Evaluation Approach:** Methodological details of how these metrics are applied across evaluation levels.
- **Results and Analysis:** Cross-component analysis of Arabic and English models, structured by output type (performance metrics, error analysis, etc.).

### 4.5.1 Theoretical Background

This section outlines the key evaluation metrics used in this section, focusing on the two standard approaches to NER evaluation — token-level and entity-level — and the metrics derived from them. As noted in Section 2.3.4, entity-level metrics serve as the main indicator of system performance, while token-level evaluation provides finer-grained insights independent of annotation scheme rules. We also introduce correlation-based methods to examine the relationship between label frequency and model performance.

**Performance Metrics** NER performance evaluation is calculated based on the relationship between true labels and model predictions, which can be organised into four outcomes: True Positives (TP), False Positives (FP), False Negatives (FN), and True Negatives (TN). These are the standard confusion matrix components, forming the basis for calculating standard evaluation metrics such as precision, recall, and F1-score. Understanding these outcomes is important for interpreting model performance, as they provide insights into the types of errors a model may produce and its ability to correctly predict entities. Both entity-level and token-level evaluation rely on these outcomes.

Below, we define each outcome and explain its relevance to both evaluation levels. To maintain clarity throughout this chapter, we refer to the token-level labels as entity tags (e.g., `B-LOC`, `I-PER`), and to the entity-level predictions as entity spans (e.g., a multi-token sequence like New York City labelled as LOC).

**True positives** refer to instances where the model's predictions correctly match the true labels. At the token level, this means that each token's predicted **entity tag** must match the corresponding true tag, including both the entity type (e.g., PER, LOC) and the correct boundary marker (e.g., `B-`, `I-`). However, token-level evaluation treats each token independently and does not require the full entity span to be predicted correctly. Whereas at the entity level, a prediction is considered correct only if the model identifies the entire **entity span** with the correct type; for example, the full sequence *New York City* must be predicted as a single LOC span with no missing or incorrectly labelled tokens.

**False positives** occur when the model incorrectly predicts a positive instance that does not correspond to any true label. For instance, tagging *Paris* as `B-PER` instead of B-LOC, or assigning an entity tag like B-LOC to a non-entity token such as *"the"*. **False negatives**, on the other hand, occur when the model fails to predict a positive instance. This may happen when *Paris* is labelled as O instead of B-LOC, or when the model misses part of an entity span—for example, predicting only *New York* instead of the full span *New York City*.

In NER, positive instances include tokens tagged with entity types such as LOC, ORG, MISC, or PER, using the B- and I- boundary format. Tokens labelled as O (outside any named entity) are considered negative. **True negatives** are tokens that are correctly predicted as O. While true negatives are not applicable at the span level—since entity-level evaluation only concerns full named entities—they are considered at the tag level to assess the model's ability to distinguish entities from non-entities. Table 4.11 provides examples of how each prediction outcome is applied at both token and entity levels.

Table 4.11: Examples of prediction outcomes in token-level and entity-level NER evaluation.

| Outcome | Token-Level Example | Entity-Level Example |
|---|---|---|
| **True Positive (TP)** | *Paris* tagged as B-LOC and correctly predicted as B-LOC | *New York City* correctly predicted as a single LOC span |
| **False Positive (FP)** | *Paris* is tagged as O but predicted as B-PER | *the Paris office* predicted as a LOC span but the gold span is only Paris |
| **False Negative (FN)** | *Paris* is tagged as B-LOC but predicted as O | *New York City* is labelled as a LOC entity, but model predicts only *New York* |
| **True Negative (TN)** | *the* tagged as O and correctly predicted as O | Not applicable (entity-level evaluation focuses only on predicted entities) |

Based on these outcomes, we define the key evaluation metrics used in NER: precision, recall, and F1-score. Precision measures the proportion of true positives out of all positive predictions made by the model, and is defined by the formula:

$$\text{Precision} = \frac{\text{TP}}{\text{TP} + \text{FP}} \tag{4.9}$$

This metric indicates how many of the predicted positive instances are actually positive. A high precision indicates that the model makes few false positive errors, suggest-ing high reliability in its positive predictions. Recall measures the proportion of true positive predictions among all actual positive instances in the dataset. It is calculated by the formula:

$$\text{Recall} = \frac{\text{TP}}{\text{TP} + \text{FN}} \tag{4.10}$$

This metric reflects how many of the actual positive instances the model successfully identifies. A high recall indicates that the model effectively captures most of the positive instances, minimising the number of false negatives. F1-score is the harmonic mean of precision and recall, providing a single metric that balances the trade-off between the two. It is calculated by the formula:

$$\text{F1-score} = 2 \cdot \frac{\text{Precision} \cdot \text{Recall}}{\text{Precision} + \text{Recall}} \tag{4.11}$$

A high F1-score indicates that the model performs well in minimising both false positives and false negatives, balancing between precision and recall. In NER evaluation, precision, recall, and F1-score are calculated for each class individually. To provide an overall measure of model performance, these metrics are typically aggregated using one of three standard averaging methods:

Macro Averaging calculates the metrics for each class independently and then computes their unweighted average. This approach ensures that all classes contribute equally to the overall performance. The formula is:

$$\text{Macro Average (Metric)} = \frac{1}{N} \sum_{i=1}^{N} \text{Metric}(i) \tag{4.12}$$

where $N$ is the number of classes, and Metric $(i)$ refers to the metric (e.g., precision, recall, F1 score) for the $i$-th class.

Micro Averaging sums the true positives, false positives, and false negatives across all classes, then calculates precision, recall, and F1-score based on these totals. This approach is useful for imbalanced classes, as it gives more weight to frequent classes, providing a performance measure that reflects the class distribution. The formula is:

$$\text{Micro Average (Metric)} = \frac{\sum_i \text{TP}_i}{\sum_i (\text{TP}_i + \text{FP}_i)} \tag{4.13}$$

where $\text{TP}_i$ and $\text{FP}_i$ are the true positives and false positives for the $i$-th class, respectively.

Finally, the weighted average, which is less commonly used, is similar to the macro average but weights each class's contribution by the number of true instances (support) for that class. This method provides a performance measure that reflects the relative frequency of each class within the dataset. The weighted average is calculated using the following formula:

$$\text{Weighted Average (Metric)} = \frac{\sum_i (\text{Support}_i \cdot \text{Metric}_i)}{\sum_i \text{Support}_i} \tag{4.14}$$

Here, $\text{Support}_i$ represents the number of true instances for the $i$-th class, and $\text{Metric}_i$ could be precision, recall, or F1-score for that class. This method of averaging is useful for imbalanced classes, as it gives more weight to classes with more instances. Further details about how these metrics are applied and analysed are provided in Section 4.5.2.

**Correlation Analysis** In addition to standard evaluation metrics, correlation coefficients such as Pearson and Spearman are used at the token level to examine the relationship between tag frequency (i.e., support) and model performance. These correlations help identify whether frequent entity tags tend to achieve better performance scores, such as precision or F1.

**Pearson correlation** measures the strength of a linear relationship between two variables. It is calculated by standardising the covariance of the variables:

$$r = \frac{\text{Cov}(X, Y)}{\sigma_X \cdot \sigma_Y} \tag{4.15}$$

where $\text{Cov}(X, Y)$ is the covariance between variables $X$ and $Y$, and $\sigma_X$ and $\sigma_Y$ are their standard deviations. Covariance itself is computed as:

$$\text{Cov}(X, Y) = \frac{\sum (X_i - \bar{X})(Y_i - \bar{Y})}{n} \tag{4.16}$$

Pearson correlation standardises the relationship between variables, resulting in a value between -1 and 1. A value close to 1 indicates a strong positive linear relationship, while a value near -1 indicates a strong negative one. When the variances of the variables are similar, Pearson correlation provides a more reliable estimate. However, if one variable has much higher variance than the other, the correlation may appear weaker or be more sensitive to noise.

**Spearman correlation**, by contrast, assesses the monotonic relationship between two variables based on their rank order. It is calculated using:

$$\rho = 1 - \frac{6 \sum d_i^2}{n(n^2 - 1)} \tag{4.17}$$

where $d_i$ represents the difference in ranks between corresponding values. Spearman is less sensitive to outliers and more robust to skewed distributions, which makes it particularly useful in cases of class imbalance. For example, when a few entity tags are very frequent and others are rare, Spearman correlation can still capture rank-based trends even if the underlying values are unevenly distributed.

Together, Pearson and Spearman correlations offer complementary views: the former captures proportional (linear) relationships, while the latter focuses on rank-based trends. In this chapter, both are used to explore whether more frequent entity tags are associated with higher performance.

In the next section, we outline how these evaluation metrics are applied across different outputs and levels of analysis.

### 4.5.2 Evaluation Approach

This section provides the technical details of our evaluation approach. The analysis is organised into four components: performance metrics, prediction outcomes analysis, error analysis, and correlation analysis. For the first three, we report results at both the token and entity levels, while the correlation analysis is applied only at the token level. The models under evaluation are `AraBERTv02`, fine-tuned on ANERCorp (Arabic), and BERT-`base`, fine-tuned on CoNLL-2003 (English), as described in Chapter 3. The details of how each of these four components is computed and applied in our analysis are outlined below.

**Terminology** For clarity and consistency throughout this chapter, we distinguish between two evaluation modes supported by `seqeval` (see Section 2.3.3). The strict mode applies discard repairs to invalid sequences and enforces exact span boundaries; this requires explicitly specifying the annotation scheme, and we adopt the IOB2 format. The default mode, applies begin repairs and by default follows the IOB1 scheme. We refer to the strict mode as *Strict* and the default mode as *Non-Strict* and we refer to the annotation schemes as IOB1 and IOB2. We also refer to token-level labels as **entity tags** and to entity-level labels as **entity spans** throughout this section. *Note:* In some parts of this section, we use the terms "IOB1" and "IOB2" as shorthand to refer not only to the annotation schemes but also to the associated evaluation modes (Non-Strict and Strict, respectively). When used in this way, "IOB1" refers to the combination of the IOB1 scheme with Non-Strict evaluation, and "IOB2" refers to the IOB2 scheme with Strict evaluation.

Finally, we use the term **prediction outcomes** to refer to the four standard evaluation components — true positives (TP), false positives (FP), false negatives (FN), and true negatives (TN). At the entity level, where true negatives are not applicable, we restrict outcomes to TP, FP, and FN.

**Performance Metrics** In the performance metrics section (based on the classification report), we present two types of analysis: aggregated and per-class. As shown in Figure 4.29, the aggregated view provides an overall summary of system performance, while the per-class view offers a breakdown by individual tags. Although the classification report outputs precision, recall, and F1 for each tag, in the aggregated view we focus on the overall scores.

We adopt different averaging strategies depending on the evaluation level. For token-level evaluation, we use **macro-averaging**. This is because token-level false positives and false negatives are symmetrical: each misclassification contributes one FP for the predicted tag and one FN for the true tag. For example, predicting B-PER instead of B-LOC increases the FP count for B-PER and the FN count for B-LOC. This symmetry causes token-level micro-averaged precision, recall, and F1 to converge to the same

value — equivalent to overall accuracy. However, accuracy is often misleading in the presence of class imbalance, where the O tag dominates. To mitigate this, we adopt macro-averaging, which treats each class equally and provides a more balanced view of model performance across all tags.

For entity-level evaluation, we instead use micro-averaging. Unlike token-level evaluation, entity-level predictions are assessed at the span level, introducing asymmetry between false positives and false negatives. A span prediction is only counted as correct if both its boundaries and entity type are exactly correct. For example, if the true entity is New York City and the model predicts New York as (LOC), this is considered one false positive for the predicted entity span ("New York") but one false negative for the entire true entity span ("New York City"). This illustrates the asymmetry: a single boundary error produces two distinct mistakes at the span level. Since the O tag is excluded from entity-level evaluation, micro-averaging provides a balanced measure of overall system correctness.

**Predictions Outcomes Analysis** To identify strengths and weaknesses in model predictions, we analyse the standard prediction outcomes: true positives (TP), false positives (FP), false negatives (FN), and true negatives (TN), to examine how predicted labels align with gold-standard labels. At the token level, we rely on the standard `scikit-learn` implementation to compute these outcomes over entity tags (e.g., `B-`PER, `I-`LOC). At the entity level, however, standard tools such as `seqeval` do not provide these outcomes, as they focus only on reporting precision, recall, and F1. To address this, we implement a custom module that computes span-level counts of true positives, false positives, and false negatives under both IOB1 and IOB2 schemes. These span-level counts provide a consistent basis for error analysis at the span level.

**Error Analysis** To better understand the differences in error patterns across languages, we classify errors at both the token and entity levels. At the token level, errors are identified through direct tag mismatches (e.g., predicting `B-`LOC instead of `B-`PER)—this can be computed straightforwardly using standard sequence alignment. At the entity level, however, evaluating errors is more complex due to the need to align predicted and true spans. To address this, we extend `seqeval` with functionality to extract position-aligned spans and classify errors into a taxonomy of distinct categories at both the token and entity levels, enabling structured comparison across languages and evaluation modes.

While `seqeval` provides entity-level metrics (e.g., precision, recall, F1), it does not natively support span alignment or structured error categorisation. As part of our contribution, we extend `seqeval` with functionality to extract position-aligned spans and classify errors into four categories—a feature not provided by the original `seqeval` tool. We define the following error categories for our span-level analysis: a **boundary error** occurs when the predicted entity type is correct, but the span boundaries are incorrect; an **entity error** occurs when the boundaries are correct but the entity type is incorrect; an **entity and boundary error** occurs when both the type and the span boundaries are incorrect; and an **O error** occurs when the model predicts an entity where the true label is O, or fails to detect an entity that should have been labelled. Consider the following example sentence:

*"Yesterday, John Smith travelled to New York."*

This sentence is annotated with entity tags as follows:

`["O", "B-PER", "I-PER", "O", "B-LOC", "I-LOC"]`

`seqeval` processes these annotations by grouping consecutive entity tags and extracting entity spans according to the annotation scheme. In IOB2, each entity is extracted as:

$$(\text{sentence_id}, \text{entity_type}, \text{start_index}, \text{end_index})$$

while in IOB1, the extraction format is:

$$(\text{entity_type}, \text{start_index}, \text{end_index})$$

Table 4.12: Comparison of Seqeval Entity Extraction in IOB1 and IOB2 schemes (Strict and Non-Strict modes)

| Annotation Scheme | Extracted Entities |
|---|---|
| IOB1 | [('PER', 1, 2), ('LOC', 4, 5)] |
| IOB2 | [(0, 'PER', 1, 3), (0, 'LOC', 4, 6)] |

Building on this understanding, we classify prediction errors by comparing true and predicted spans extracted from `seqeval`. For example, if the true span is (1, 3, "LOC") but the model predicts (1, 2, "LOC"), the type is correct but the span is too short—this is a boundary error. If both the type and span are incorrect (e.g., predicted as (1, 2, "PER")), the error is categorised as an entity and boundary error. If the span is correct but the predicted type differs (e.g., predicted as (1, 3, "PER")), it is an entity error. Finally, if the model predicts an entity in a position where the true label is O, or fails to predict an entity span altogether, the error is counted as an O error.

These structured error categories provide a more interpretable breakdown of span-level mistakes, complementing the aggregate precision, recall, and F1 metrics normally reported by `seqeval`. To our knowledge, standard tools like seqeval do not support this form of structured, span-aligned error analysis as part of its standard output.

These structured error categories provide a more interpretable breakdown of span-level mistakes, complementing the aggregate precision, recall, and F1 metrics normally reported by `seqeval`. To our knowledge, existing tools such as `seqeval` do not provide this functionality; our extension therefore contributes the first structured, span-aligned error analysis for NER evaluation.

**Correlation Analysis** At the token level, we calculate Pearson and Spearman correlation coefficients to examine the relationship between entity tag frequency (support) and model performance (precision and F1 score). These correlations are computed between precision and recall for each entity tag and the support of that tag in the training and testing datasets. This allows us to assess whether the frequency of an entity tag influences how well the models perform on it. In addition, we provide a detailed examination of the correlation outcomes, highlighting both their interpretive value and their limitations.

### 4.5.3 Performance Metrics

To address aspect (a) of RQ3—*how well true and predicted labels are aligned across languages*—we analyse system performance using standard evaluation metrics: precision, recall, and F1-score. These metrics are computed at both the token and entity levels. The analysis is presented in two parts: **aggregated evaluation**, which provides a high-level comparison of overall performance in Arabic and English, and **per-class evaluation**, which examines performance across individual entity tags and spans.

**Aggregated Evaluation** Tables 4.13 and 4.14 summarise the aggregated evaluation metrics for AraBERTv02 and BERT under IOB1[1]. While we report results separately for each model, some general patterns emerge across them. The results show that BERT consistently achieves higher performance than AraBERTv02 across all metrics and evaluation levels. However, small differences in precision-recall balance across levels and languages offer insights that require further investigation.

Table 4.13: Aggregated NER Performance Metrics for AraBERTv02 (Arabic) across token and entity levels (IOB1 mode). Entity-level metrics use micro-averaging, while token-level metrics use macro-averaging.

| Metric | Token-Level | Entity-Level |
|---|---|---|
| *Precision* | 0.8649 | 0.8483 |
| *Recall* | 0.7934 | 0.8178 |
| *F1-score* | 0.8205 | 0.8327 |

For AraBERTv02 (Table 4.13), precision is consistently higher than recall at both levels, suggesting that the model prioritises precision over coverage—resulting in fewer false positives but more false negatives. The precision–recall gap is larger at the token level than at the entity level, a pattern we investigate further in the level-specific analysis.

Table 4.14: Aggregated NER Performance Metrics for BERT (English) across token and entity levels (IOB1 mode). Entity-level metrics use micro-averaging, while token-level metrics use macro-averaging.

| Metric | Token-Level | Entity-Level |
|---|---|---|
| *Precision* | 0.8945 | 0.9058 |
| *Recall* | 0.9156 | 0.9193 |
| *F1-score* | 0.9047 | 0.9125 |

For BERT (Table 4.14), recall slightly exceeds precision, indicating that the model captures more entities overall. The precision–recall gap is also slightly larger at the token level than at the entity level, but the difference is much smaller than for AraBERTv02. We investigate this pattern further in the level-specific analysis.

[1] We include IOB1 results first because it is the default mode used in many NER studies; see Section 2.3.3.

To evaluate the impact of evaluation mode, we report entity-level precision, recall, and F1-score under both Non-Strict (IOB1) and Strict (IOB2) settings. Tables 4.15 and 4.16 summarise the results for AraBERTv02 and BERT, respectively. Across both models, small differences appear between IOB1 and IOB2, highlighting the effect of stricter span requirements on performance.

Table 4.15: Entity-level NER Performance for AraBERTv02 (Arabic) under IOB1 (Non-Strict) and IOB2 (Strict) evaluation.

| **Metric** | **IOB1 (Non-Strict)** | **IOB2 (Strict)** |
|---|---|---|
| *Precision* | 0.8483 | 0.8476 |
| *Recall* | 0.8178 | 0.8299 |
| *F1-score* | 0.8327 | 0.8387 |

For AraBERTv02 (Table 4.15), recall improves under IOB2, with a corresponding increase in F1. This is expected since the dataset annotations follow the IOB2 format. However, precision drops slightly, suggesting that stricter evaluation may introduce some over-predictions.

Table 4.16: Entity-level NER Performance for BERT (English) under IOB1 (Non-Strict) and IOB2 (Strict) evaluation.

| **Metric** | **IOB1 (Non-Strict)** | **IOB2 (Strict)** |
|---|---|---|
| *Precision* | 0.9058 | 0.9126 |
| *Recall* | 0.9193 | 0.9184 |
| *F1-score* | 0.9125 | 0.9155 |

For BERT (Table 4.16), precision increases slightly under IOB2, while recall decreases slightly. Overall, this results in a small improvement in F1, indicating stable performance under stricter evaluation mode.

Overall, these results show that evaluation mode can affect performance, even at the aggregate level. IOB2 generally shifts the balance between precision and recall, but in opposite directions for the two models: AraBERTv02 gains recall at the cost of precision, while BERT gains precision at the cost of recall. In both cases, the changes are of similar magnitude, leading to small but consistent improvements in F1. Despite these differences, the broader trend remains unchanged: BERT achieves stronger performance than AraBERTv02 under both evaluation settings. We next turn to the entity-level analysis to explore how each model handles specific categories such as PER, LOC, and ORG.

**Dataset Size and Controlled Comparison** One potential confounding factor in comparing AraBERTv02 and BERT is the difference in dataset sizes between ANERCorp (Arabic) and CoNLL-2003 (English). As shown in Section 4.3.2, CoNLL-2003 contains 250k tokens with 42k NE tokens, compared to 150k tokens with 16.5k NE tokens in ANERCorp. This size discrepancy raises the concern that BERT's stronger performance might simply reflect greater exposure to entity examples during training.

To probe this issue, we constructed a reduced version of CoNLL-2003 (DeCoNLL) by downsampling to approximate ANERCorp's size and entity distribution (see Table 4.17). This yielded 93k total tokens with 11.6k NE tokens, comparable in scale to ANERCorp.

Table 4.17: Comparison of dataset sizes for ANERCorp, CoNLL-2003, and downsampled DeCoNLL.

| Dataset | Total Tokens | NE Tokens | NE Proportion |
|---|---|---|---|
| ANERCorp | 150,110 | 16,556 | 11.0% |
| CoNLL-2003 | 250,056 | 42,155 | 16.9% |
| DeCoNLL | 93,214 | 11,676 | 12.5% |

We then evaluated BERT on DeCoNLL under the same setup. At the token level, macro F1 was 0.8903, closely matching the result on the full CoNLL-2003 dataset (0.905). At the entity level, micro F1 was 0.911, slightly lower than the full CoNLL result of 0.913. Note that entity-level evaluation here was reported under IOB1 (Non-Strict). These outcomes indicate that dataset size alone does not account for the observed performance gap. However, we emphasise that this controlled experiment does not remove—or rule out—the impact of dataset size, nor other structural differences such as class imbalance, entity-type distribution, or annotation conventions. Moreover, relying solely on aggregate F1 scores risks masking the very error patterns and annotation effects this thesis aims to uncover[2]. While F1 provides a quick sanity check, it is not sufficient for explaining why performance differs. A more systematic study of these dimensions, including varying fine-tuning data size, distribution, and annotation quality, is left for future work, as discussed in Chapter 6. Full classification reports of the token-level and span-level results for DeCoNLL are provided in Appendix A, allowing inspection of per-class precision, recall, and F1.

[2] Both ANERCorp and CoNLL-2003 contain known annotation inconsistencies (Rücker and Akbik, 2023; Liu and Ritter, 2023; Reiss et al., 2020). Downsampling may inadvertently sample or avoid such noise, meaning the results here should be interpreted as indicative rather than definitive.

**Entity-Level Analysis** Figure 4.30 reports entity-level F1-scores by span type (LOC, MISC, ORG, PER) for AraBERTv02 (Arabic) and BERT (English) under both IOB1 and IOB2 schemes. This view highlights how the choice of scheme affects span-level performance within each language.

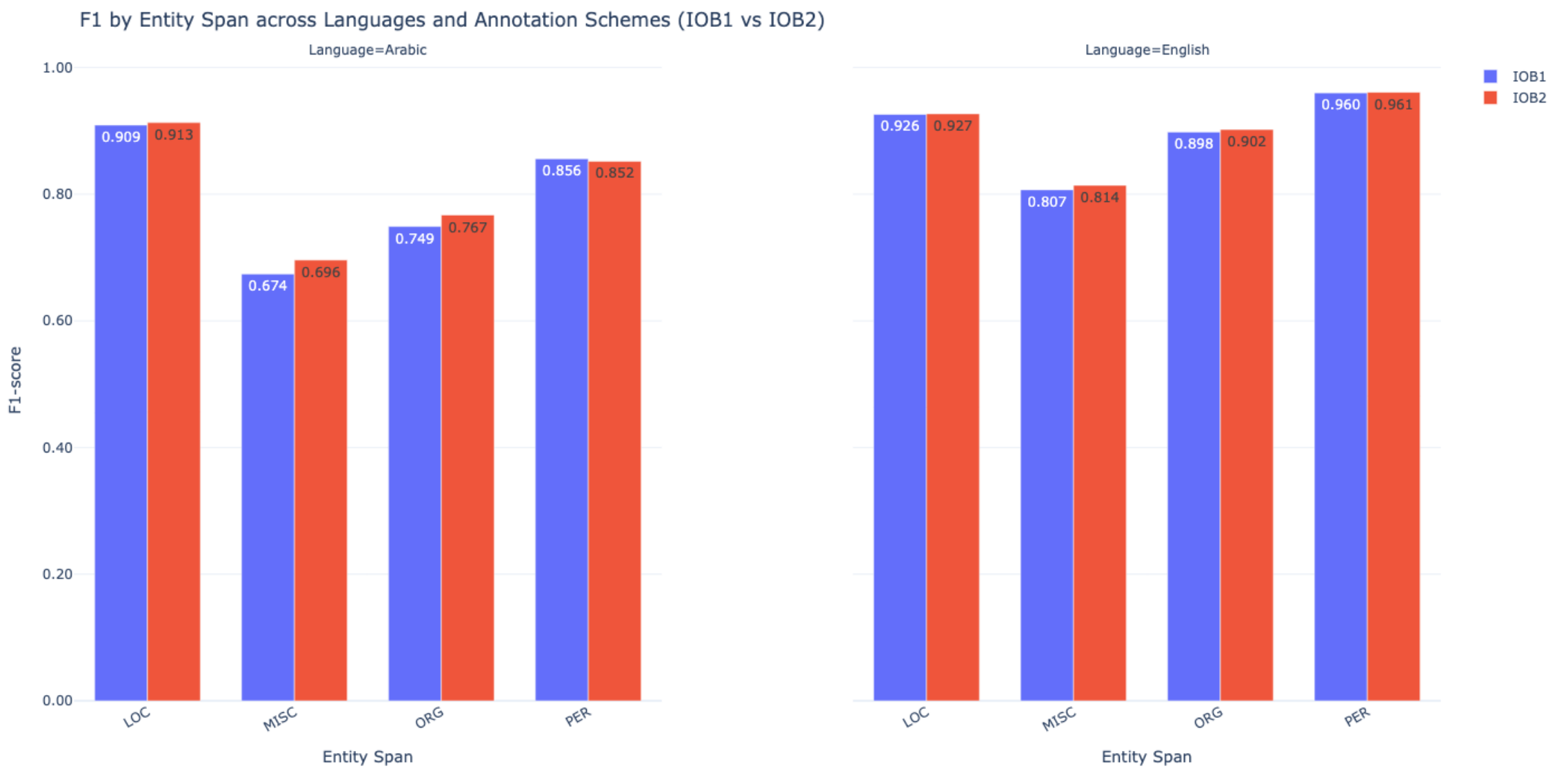

Figure 4.30: Entity-level F1-scores by span type (LOC, MISC, ORG, PER) for AraBERTv02 (Arabic) and BERT (English), under both **IOB1** and **IOB2** annotation schemes.

For Arabic, F1 increases under IOB2 for *MISC* (0.674 → 0.696) and *ORG* (0.749 → 0.767), with a smaller gain for *LOC* (0.909 → 0.913). *PER* shows a slight decrease (0.856 → 0.852). Overall, IOB2 yields modest improvements for the more difficult spans (MISC and ORG), with minimal change elsewhere.

For English, IOB2 produces small F1 increases across most spans: *LOC* (0.926 → 0.927), *MISC* (0.807 → 0.814), *ORG* (0.898 → 0.902), and *PER* (0.960 → 0.961). The effects are minor but consistently positive.

Across both languages, the ranking of entity spans by F1 remains consistent: PER and LOC are the strongest categories, while MISC remains the most challenging. The differences are smallest for LOC, where scores are relatively close, and largest for ORG and MISC, particularly under IOB1, indicating that these categories are more difficult for AraBERTv02.

While the F1-score provides a useful overall measure of performance, it does not explain the underlying trends observed earlier. In particular, it does not reveal why AraBERTv02 often shows higher precision or why BERT achieves better recall, nor does it show how the annotation scheme shifts the balance between the two. To address this, we next break down precision and recall by entity span and annotation scheme.

Figure 4.31 reports precision and recall by entity span (LOC, MISC, ORG, PER) for AraBERTv02 (Arabic) and BERT (English) under both IOB1 and IOB2 schemes. This breakdown highlights how evaluation mode shifts the balance between precision and recall across languages.

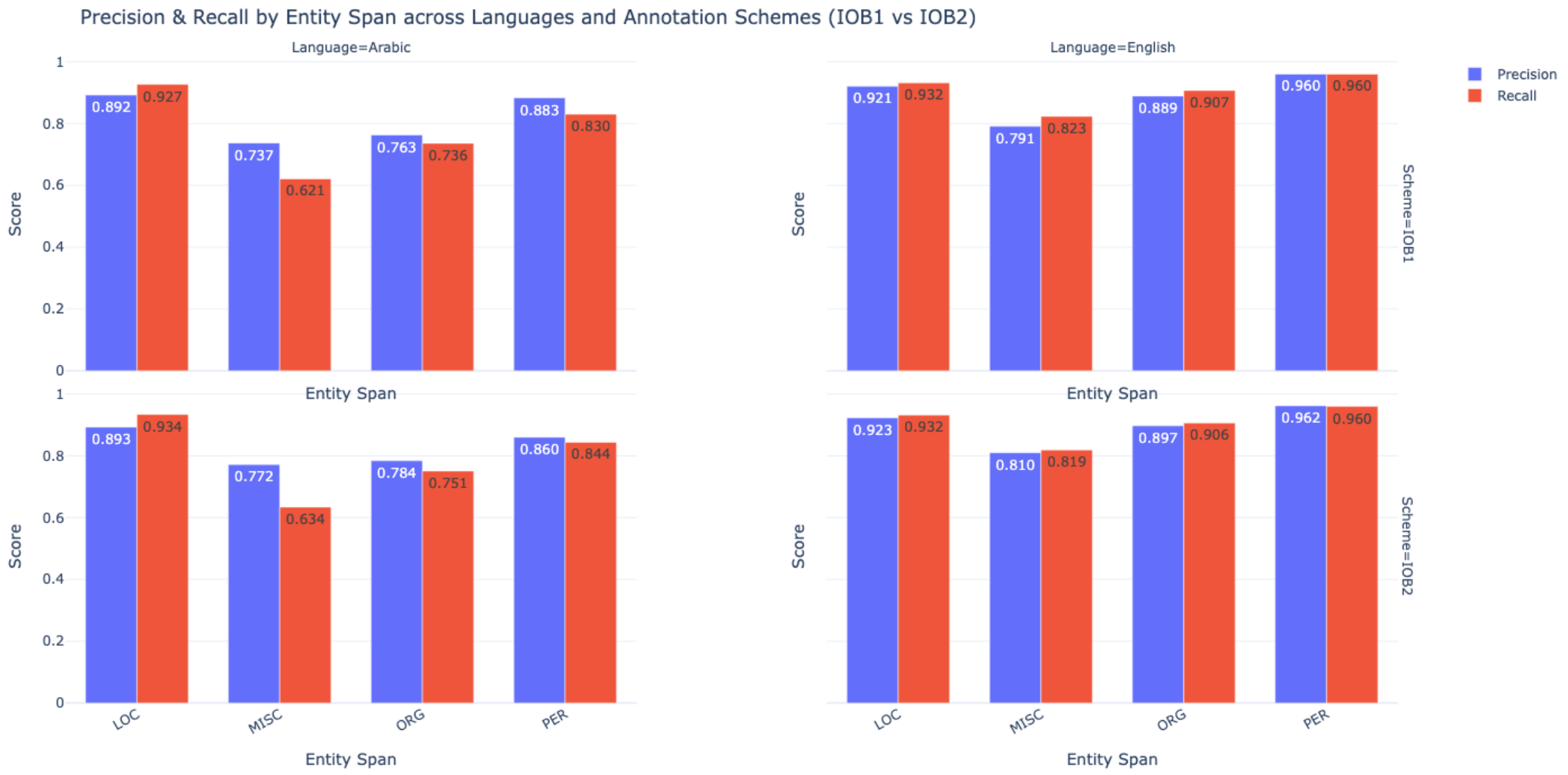

Figure 4.31: Precision and Recall by entity span (LOC, MISC, ORG, PER) for AraBERTv02 (Arabic) and BERT (English), under both **IOB1** and **IOB2** annotation schemes.

For Arabic, AraBERTv02 shows higher precision than recall under IOB1 across most spans, especially MISC (0.737 vs 0.621) and PER (0.883 vs 0.830). LOC is the exception, where recall is slightly higher (0.927 vs 0.892). Under IOB2, recall improves in LOC (0.934) and increases alongside precision in MISC and ORG, though PER shows a small drop in precision (0.860 $\rightarrow$ 0.844). Overall, IOB2 reduces AraBERTv02's precision–recall gap, particularly for the harder spans.

For English, BERT displays more balanced precision and recall. Under IOB1, recall is slightly higher than precision for LOC and ORG, while the two are nearly equal for PER (0.960 vs 0.960). IOB2 produces minor improvements in precision across all spans, with recall remaining stable. This stability contrasts with AraBERTv02's shifts, showing that BERT maintains a steadier balance under both schemes.

These trends help explain the aggregate patterns observed earlier in Tables 4.13 and 4.14, where AraBERTv02 showed a wider precision–recall gap than BERT. AraBERTv02 tends to achieve higher precision but lower recall across most spans, while BERT maintains a more balanced behaviour.

To examine these gaps more closely, we also consider the relative prevalence of entity spans. As shown in Figure 4.32, PER and LOC are the most frequent categories in both datasets, while MISC and ORG are less common.

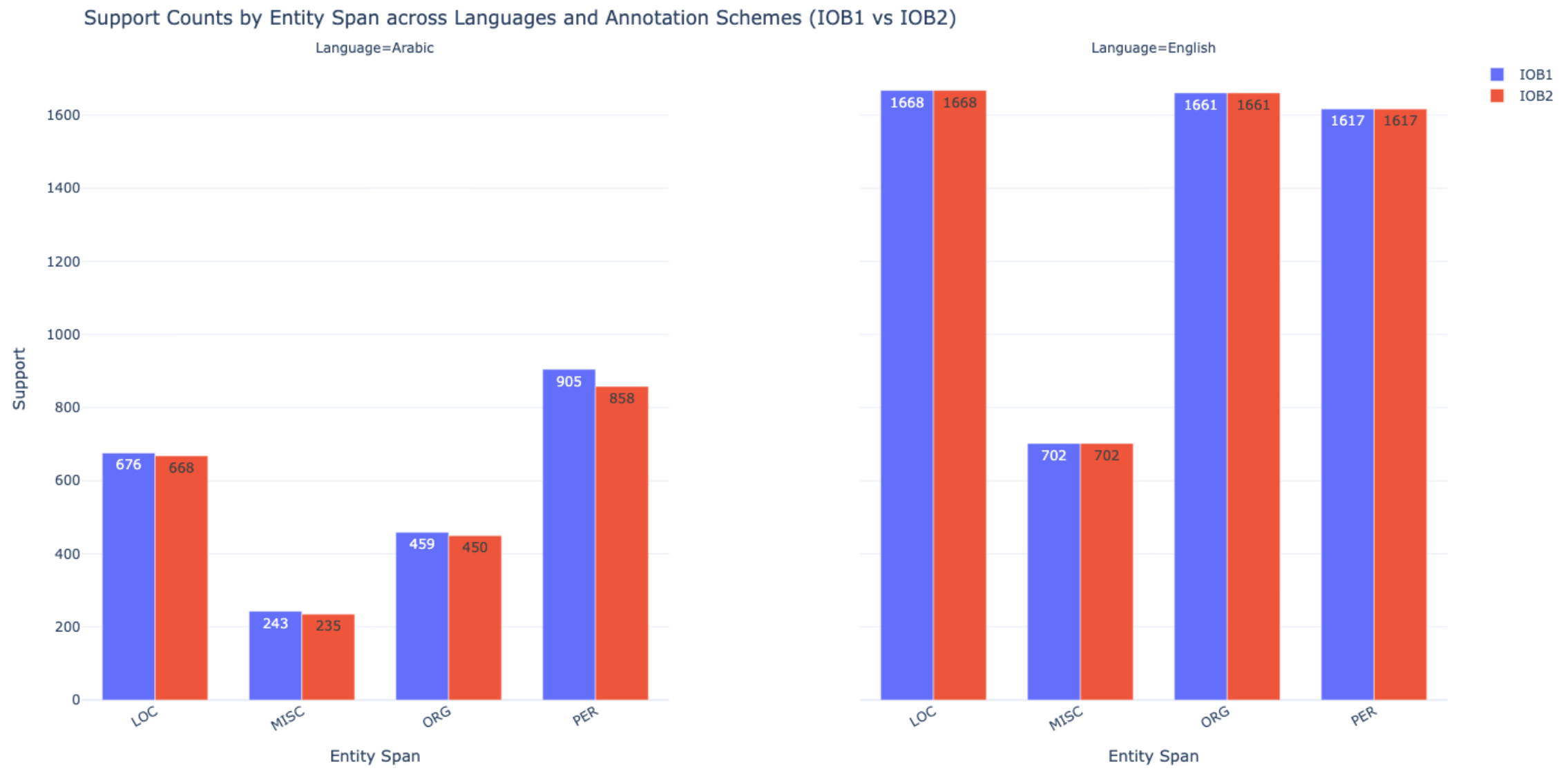

Figure 4.32: Support counts (number of examples) per entity span (LOC, MISC, ORG, PER) for AraBERTv02 (Arabic) and BERT (English), under both **IOB1** and **IOB2** annotation schemes.

For AraBERTv02, PER's recall is consistently lower than its precision under both schemes (IOB1: 0.830 vs. 0.883; IOB2: 0.844 vs. 0.860). Because PER is frequent, these false negatives have a large impact on the overall micro-averaged recall. MISC and ORG also show weaker recall (IOB1: 0.621 and 0.736; IOB2: 0.634 and 0.751), further widening the gap. Together, PER, ORG, and MISC account for roughly 70% of spans in the test set, so their recall behaviour strongly shapes AraBERTv02's aggregate results.

For BERT, precision increases slightly under IOB2, while recall is stable or only marginally lower for some spans (e.g., MISC 0.823 → 0.819, ORG 0.907 → 0.906). The relatively balanced precision–recall profile across high-frequency spans (LOC, ORG, PER) helps explain why BERT's aggregate performance is more stable.

**Token-Level Analysis** In this section, we analyse model performance at the token level. Unlike entity-level evaluation, which depends on how spans are extracted and aligned under different annotation schemes, token-level analysis is based directly on the raw tags predicted by the model. This provides a clearer view of how well the models learn to assign specific tags such as B-LOC or I-PER, independent of span-level boundaries. Because models are trained to predict tags one token at a time, token-level metrics also align more directly with the training objective.

We first examine the distribution of entity tags in the test split. Figure 4.33 shows the support counts for AraBERTv02 and BERT. Across both models, B-tags are generally more frequent than I-tags, with the notable exception of I-PER, which is relatively common. For AraBERTv02, B-LOC and B-PER are the most frequent tags, while I-LOC appears much less frequently than I-PER. For BERT, B-LOC, B-ORG, and B-PER dominate, while tags like I-LOC, I-MISC, and I-ORG are comparatively under-represented. Overall, BERT has higher support across most tags, reflecting differences in dataset size and balance.

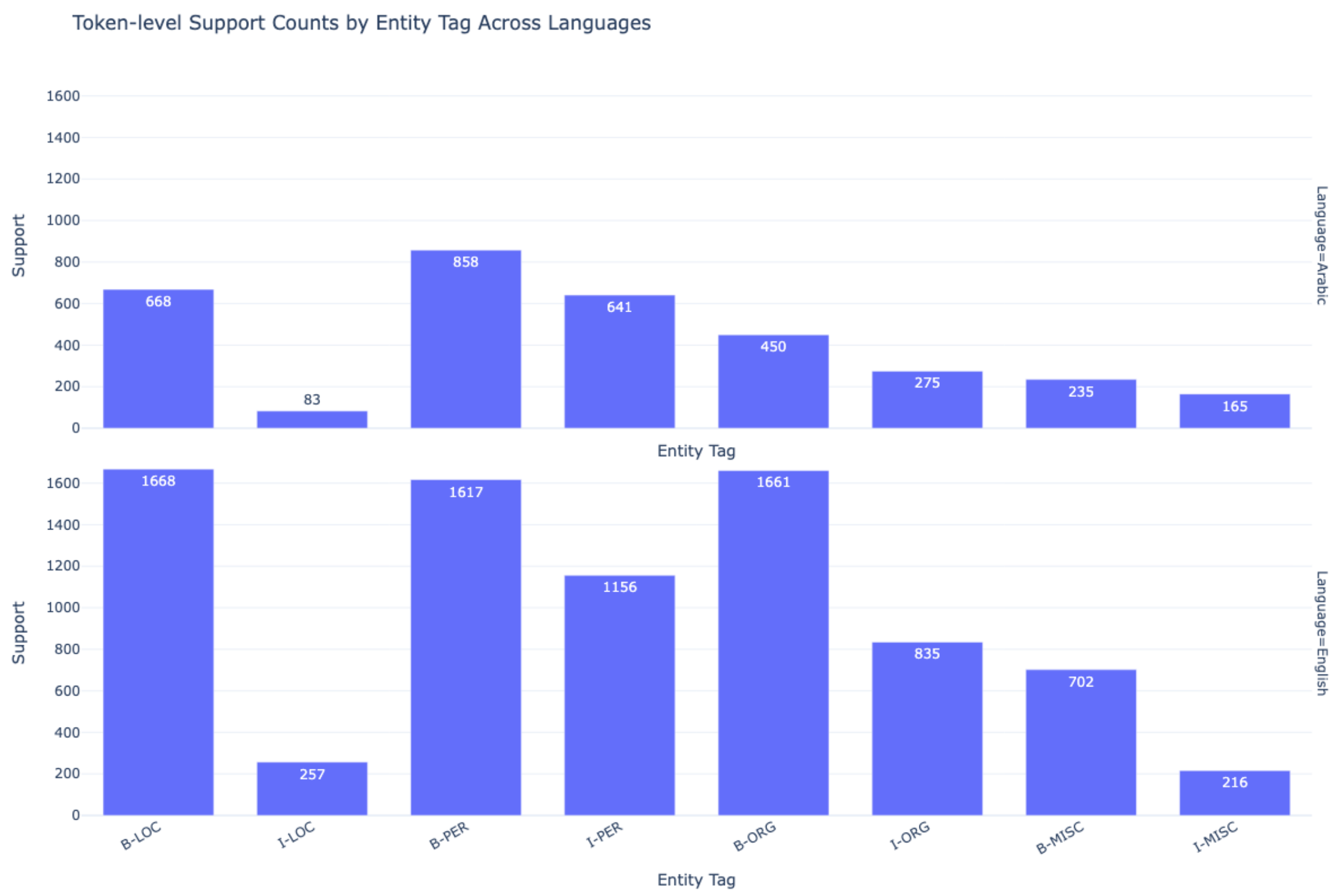

Figure 4.33: Support counts (number of examples) per entity tag in the test split for AraBERTv02 (Arabic) and BERT (English).

We first examine the F1-score across entity tags. Figure 4.34 summarises the F1-scores by entity tag. For Arabic, AraBERTv02 performance is strongest for B-LOC (0.926) and I-PER (0.909), but weaker for MISC tags (B-MISC: 0.743, I-MISC: 0.548), again highlighting the difficulty of this category. For English, BERT achieves consis-tently high F1 across tags, with especially strong results for PER and LOC (both above 0.96), while I-MISC remains the weakest (0.716).

These patterns reinforce two trends observed earlier: (1) AraBERTv02 tends to achieve lower recall in several categories, especially for minority tags like MISC, which drives its wider precision–recall gap. (2) BERT maintains a more stable precision–recall balance, aided by higher support across tags and stronger recall for frequent categories like PER and LOC.

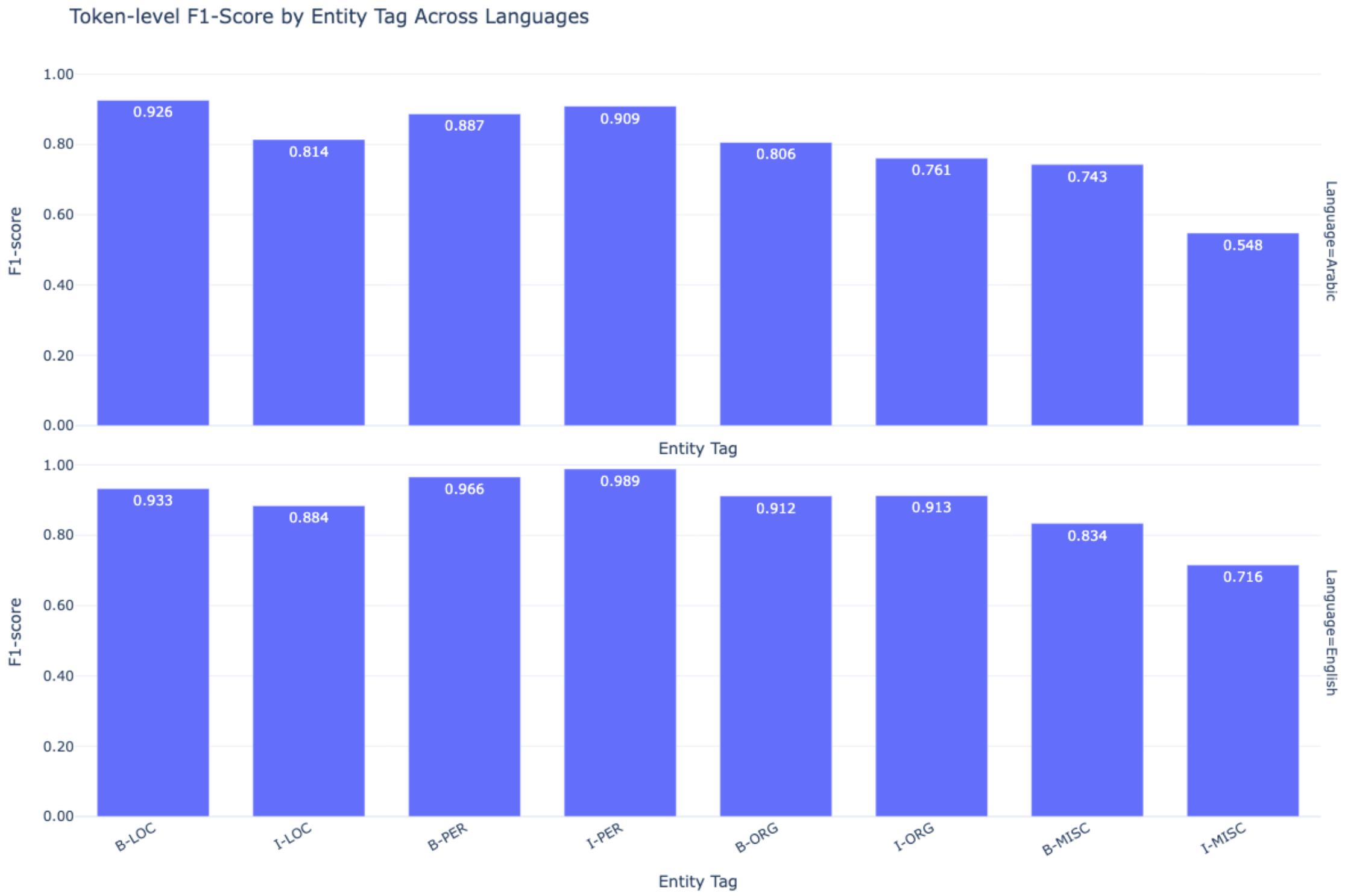

Figure 4.34: Token-level F1-scores by entity tag for AraBERTv02 (Arabic) and BERT (English).

Both models show B-LOC and B-PER as the most frequent tags in the test set. However, BERT also shows a higher count of B-ORG tokens compared to AraBERTv02, making B-ORG one of its more dominant tags—unlike in AraBERTv02, where it is less frequent. For higher-frequency tags such as B/I-LOC and B/I-PER, the performance gap is relatively small, though BERT consistently achieves higher scores. The gap widens for I-tokens, especially I-PER and I-LOC. In the case of B/I-ORG, where BERT has higher support, the performance difference is already noticeable for B-tokens and grows further for I-tokens. This suggests that lower-frequency entity tags are more error-prone, particularly for AraBERTv02.

To better understand the precision–recall gap observed earlier—particularly at the token level, where the gap is wider for both models—we next break down precision and recall by entity tag. Figure 4.35 shows the trade-off between precision and recall for AraBERTv02 and BERT.

For AraBERTv02, dominant tags such as B/I-LOC and B/I-PER show relatively balanced scores, though the pattern varies. B-LOC and I-LOC show higher recall than precision, while B-PER and I-PER show the opposite. The gap is widest for B-LOC, suggesting stronger coverage but weaker reliability. As tag frequency decreases, the gap widens further, particularly due to reduced recall. This is most evident in B/I-MISC, where B-MISC (0.824 vs. 0.677) and especially I-MISC (0.819 vs. 0.412) show strong precision but poor recall. I-ORG follows a similar but less severe trend, with lower recall than precision.

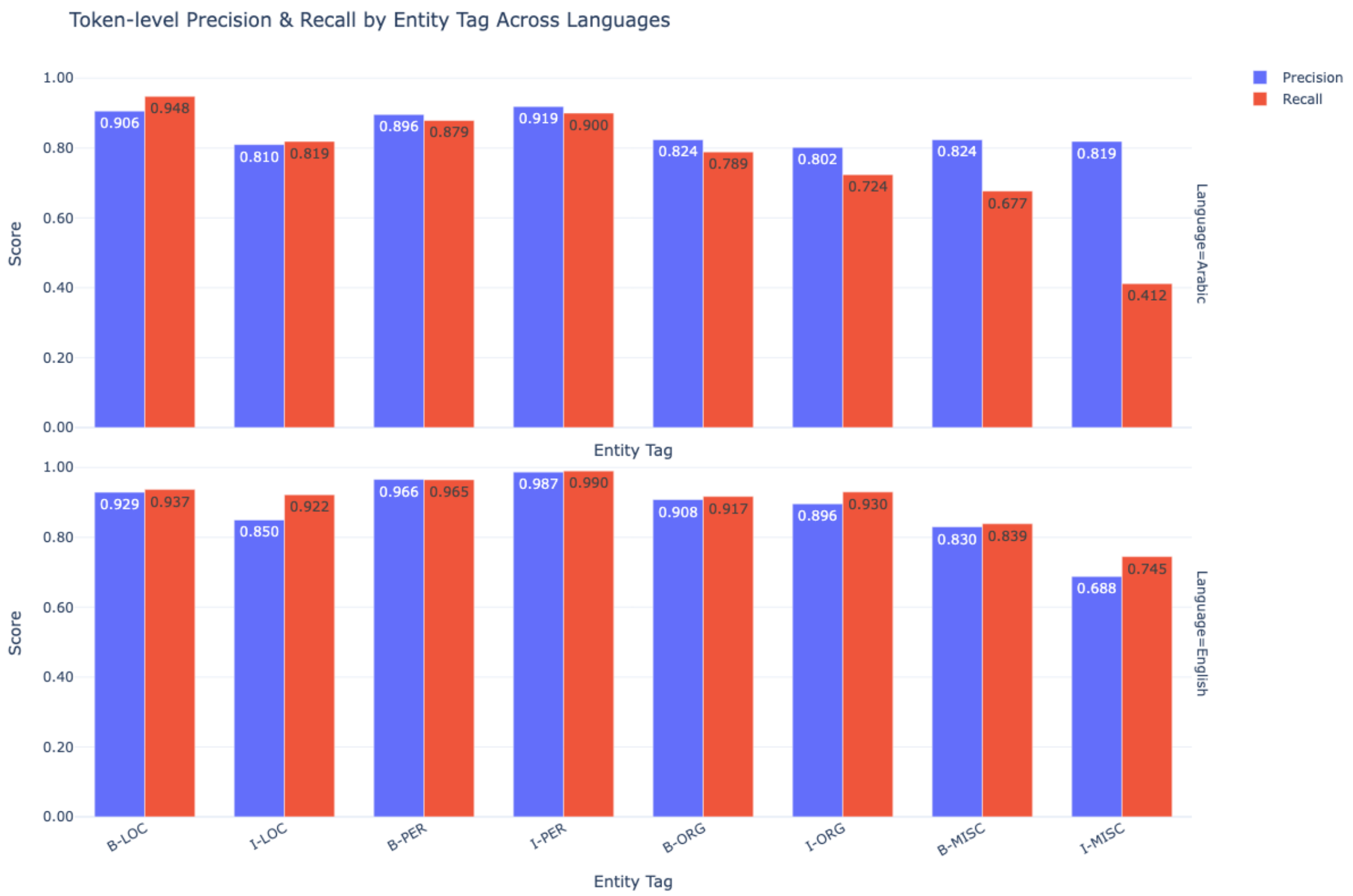

Figure 4.35: Precision and Recall by token-level entity tags (B-LOC, I-LOC, B-PER, etc.) for AraBERTv02 (Arabic) and BERT (English).

For BERT, precision and recall are generally balanced. For dominant tags like B-LOC, B-PER, and I-PER, the two are almost identical (e.g., B-LOC 0.937 vs. 0.929, I-PER 0.990 vs. 0.987). This indicates that BERT maintains strong coverage without introducing many false positives. For rarer I-tags, performance is more variable. I-LOC shows a wider gap with lower precision (0.850 vs. 0.922), while I-ORG (0.930 vs. 0.896) performs relatively well, showing that I-tags are not uniformly weak in BERT. For B/I-MISC, B-MISC (0.839 vs. 0.830) is balanced, but I-MISC again shows lower scores and a wider gap (0.745 vs. 0.688).

Across both models, some general trends emerge. Both AraBERTv02 and BERT achieve strong precision and recall for the "O" tag, indicating reliable separation of entities from non-entities. However, AraBERTv02 shows a clearer pattern where lower-frequency tags correspond to larger precision–recall gaps, mostly due to reduced recall. BERT, by contrast, shows more consistent behaviour across tags, even when frequency is low. These observations highlight the potential influence of dataset frequency on model performance—an aspect explored further in Section 4.5.6.

This helps explain why the token-level precision–recall gap in the aggregated evaluation appears larger than at the entity level. Unlike entity-level metrics, which are micro-averaged, token-level evaluation uses macro-averaging, giving equal weight to all tags regardless of frequency. As a result, weak performance on rare tags like I-MISC and I-ORG—where AraBERTv02 particularly struggles—contributes disproportionately to the overall metric. BERT's stronger balance across both frequent and rare tags allows it to maintain narrower gaps and more stable aggregate performance.

### 4.5.4 Prediction Outcomes Analysis

To address aspect (b) of RQ3—*what strengths and weaknesses exist in each language's performance*—this section presents an analysis of prediction outcomes. As introduced in Section 4.5.1, prediction outcomes are the core components of evaluation metrics (TP, FP, FN, and TN), providing a breakdown of how predicted labels align or conflict with gold-standard labels. Analysing these outcomes helps us move beyond aggregate scores to understand the nature and distribution of model predictions and errors.

The analysis is organised into two parts: first, entity-level outcomes, which capture errors in complete span predictions under different annotation schemes; and second, token-level outcomes, which provide a more fine-grained view of individual tag assignments. We begin with the entity-level analysis.

**Entity-Level Prediction Outcomes** To better understand the precision–recall trade-offs observed in AraBERTv02 and BERT, we analyse prediction outcomes (true positives, false positives, and false negatives) at the span level. The outcomes are normalised relative to the total number of predictions per entity span, ensuring comparability across categories with different frequencies[3]. Figure 4.36 shows the proportional breakdown of TP, FP, and FN across entity spans and annotation schemes.

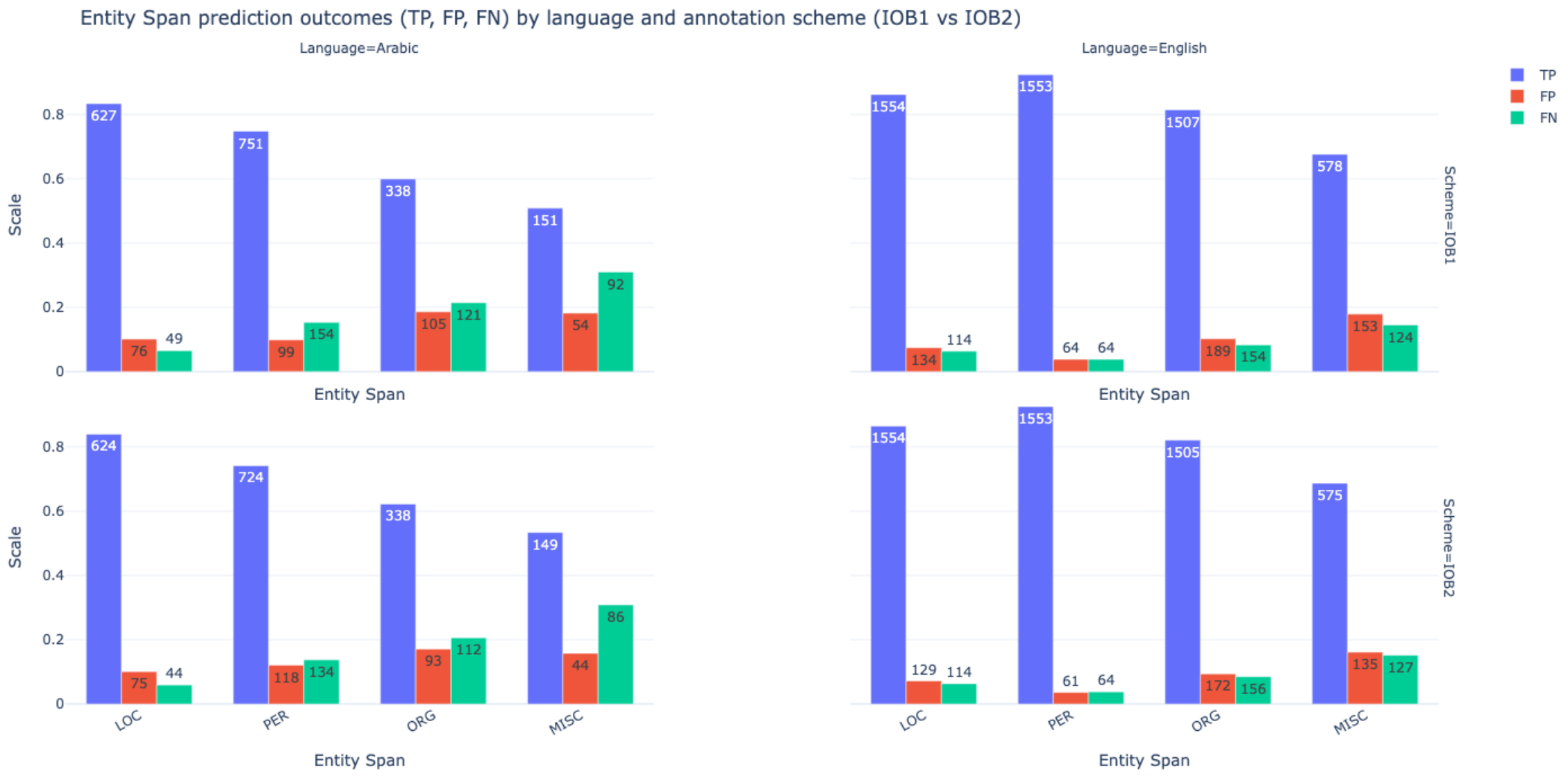

Figure 4.36: Breakdown of prediction outcomes (TP, FP, FN) for AraBERTv02 (Arabic) and BERT (English), across entity spans (LOC, MISC, ORG, PER) and annotation schemes (IOB1 and IOB2).

Under IOB1, several patterns emerge. For AraBERTv02, LOC and PER show high numbers of true positives but diverge in error type: PER produces more false negatives, lowering recall, while LOC produces more false positives, lowering precision. These differences explain why PER achieves higher precision but lower recall, while LOC shows the opposite—illustrating how the distribution of error types directly shapes precision–recall trade-offs.

The size of the test set also influences how sensitive performance metrics are to errors. Smaller test sets with fewer spans—such as MISC and ORG in AraBERTv02—amplify the impact of false positives and false negatives. As shown earlier in Figure 4.32, support varies substantially across entity spans, and this variation directly affects the stability of precision and recall.

[3] We plot normalised outcome: TP-proportion $= \frac{TP}{TP+FP+FN}$, FP-proportion $= \frac{FP}{TP+FP+FN}$, FN-proportion $= \frac{FN}{TP+FP+FN}$. For example, for LOC in AraBERTv02 under IOB1, TP-share $= \frac{627}{627+76+49} \approx 0.834$.

For AraBERTv02, the MISC span includes only 243 examples. With 54 false positives and 92 false negatives, these errors heavily reduce its precision (0.737) and recall (0.621). By contrast, BERT handles a much larger MISC set with 702 examples. Although it produces more absolute errors—124 false positives and 153 false negatives—these represent a smaller proportion of the total, resulting in higher precision (0.791) and recall (0.823). This demonstrates how larger test sets help mitigate error variability, leading to more stable metrics.

Taken together, these patterns show the importance of accounting for test set size and entity span distribution when comparing model performance. While BERT consistently outperforms AraBERTv02 across most spans, AraBERTv02 performs strongly on high-frequency spans such as LOC and PER under both annotation schemes. This suggests that although AraBERTv02 struggles on lower-frequency spans like MISC, it remains competitive on dominant spans. In multilingual NER evaluation, these imbalances must be considered to avoid drawing biased conclusions.

To understand the effect of the annotation scheme, we compare how error distributions change between IOB1 and IOB2. While the overall impact is small, the scheme influences the balance of false positives (FP) and false negatives (FN) across models and spans. Figure 4.37 presents a heatmap of FP and FN counts for AraBERTv02 and BERT under both schemes.

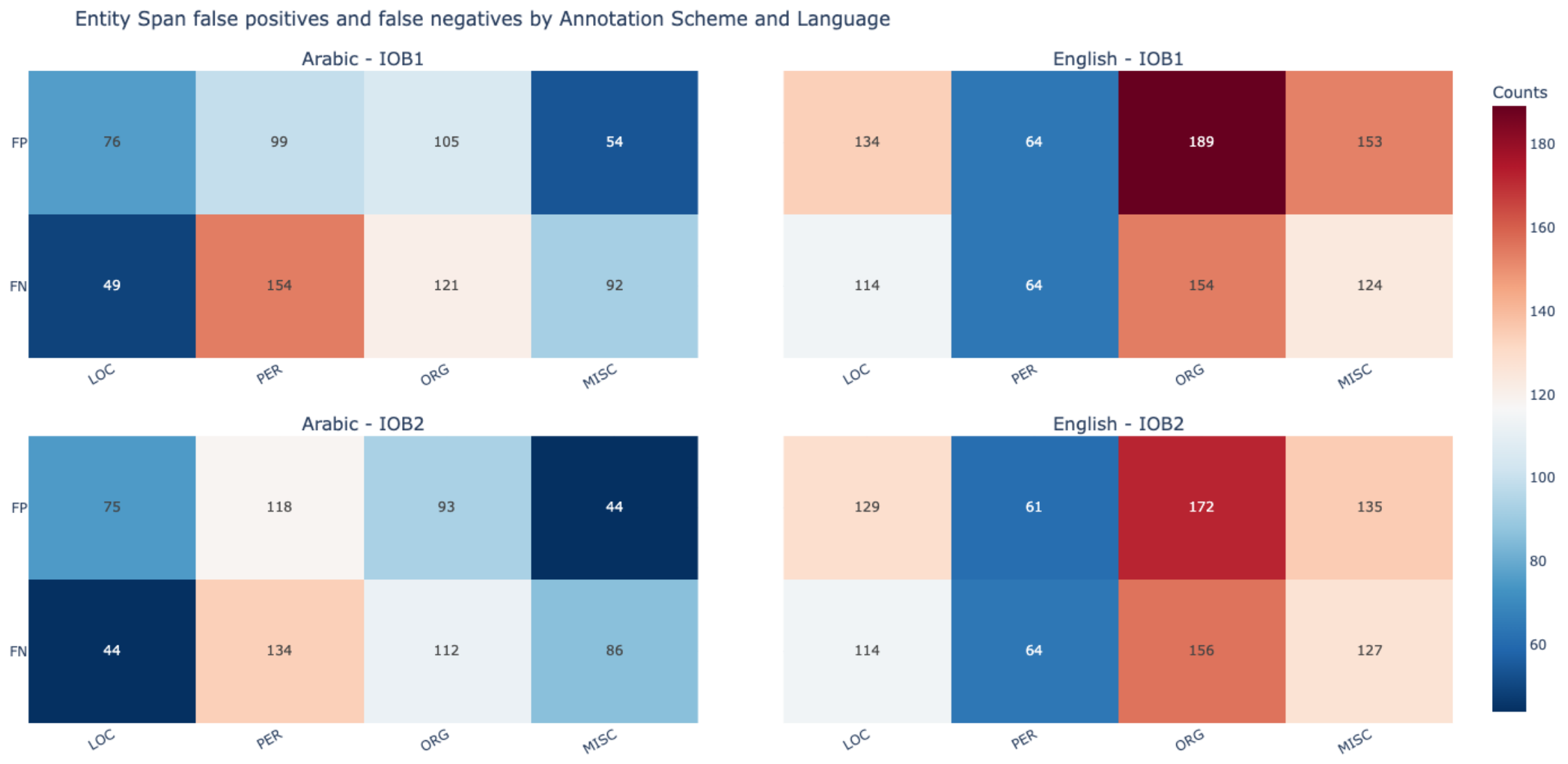

Figure 4.37: False positives (FP) and false negatives (FN) by entity span (LOC, PER, ORG, MISC) for AraBERTv02 (Arabic) and BERT (English), separated by annotation scheme (IOB1 vs IOB2).

For BERT, switching from IOB1 to IOB2 has little effect. In MISC, IOB2 slightly reduces false positives but increases false negatives, leading to the minor precision–recall shifts seen earlier. LOC and PER show minimal changes. Similar trends are seen across other spans, with LOC and PER showing minimal variation between schemes.

AraBERTv02, however, is more sensitive to scheme choice. Under IOB2, MISC and ORG both see reductions in FP and FN, contributing to their performance gains. PER shows the opposite pattern: FP decrease but FN increase, meaning the model predicts fewer spans overall and misses more correct ones. This explains why PER's recall drops slightly while its precision rises under IOB2, as noted earlier. Overall, the stricter span rules in IOB2 reduce errors for AraBERTv02 in less frequent spans, while having little effect on high-frequency spans like LOC and PER.

These shifts stem from the evaluation logic. IOB1 applies a repair mechanism, treating some invalid sequences (e.g., spans starting with `I-`) as valid, while IOB2 applies a discard mechanism, ignoring spans without a proper `B-` start. As a result, some instances are no longer counted as valid predictions, reducing the TP count or causing reallocation between FP and FN.

A drop in TP occurs when spans previously accepted under IOB1 become invalid under IOB2. For instance, if the true label is `B-LOC I-LOC` but the model predicts `I-LOC I-LOC`, the span is valid under IOB1 (repaired) but ignored under IOB2 due to stricter rules. This change reclassifies a former TP as FN. Similarly, invalid sequences in the *true labels* also impact TP. For example, if the true label for "Google" is `I-ORG` followed by `O`, the span is ignored under IOB2—even if the model correctly predicts `I-ORG`. If the model instead predicts `B-ORG`, this is counted as a FP due to the mismatch with the discarded gold label. On the other hand, if the model predicts `O`, the instance is ignored entirely.

In short, shifts in FP and FN arise both from changes in TP and from the constraints imposed by the annotation scheme. Switching schemes does not necessarily introduce new errors; rather, it changes how errors are counted and whether they are visible. Some cases are discarded under stricter rules, while others are reclassified. Recognising these shifts is essential when comparing evaluation results across annotation schemes, since the scoring logic itself shapes how model behaviour is interpreted. We now turn to the token-level confusion outcomes to examine how individual entity tag predictions contribute to model strengths and weaknesses.

**Token-Level Prediction Outcomes** At the token level, prediction outcomes provide a finer-grained view of strengths and weaknesses across individual entity tags. Figure 4.38 presents the breakdown of true positives (TP), false positives (FP), and false negatives (FN) for AraBERTv02 and BERT.

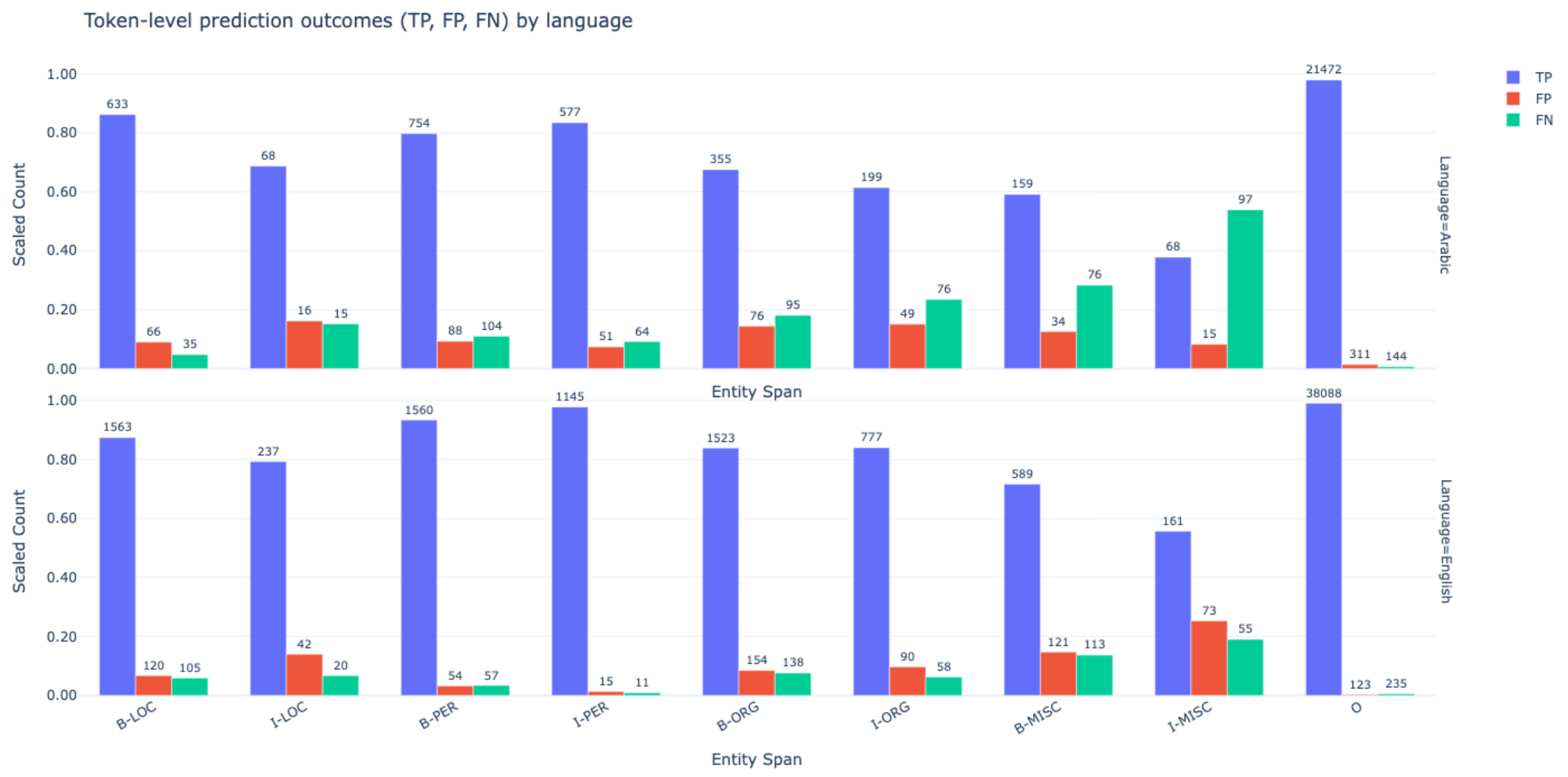

Figure 4.38: Token-level prediction outcomes (TP, FP, FN) for AraBERTv02 (Arabic) and BERT (English), reported across entity tags. Counts are scaled to highlight the relative balance of outcomes within each tag.

As discussed in Section 4.5.3, high-support tags such as O, B-LOC, B-PER, and I-PER dominate the outcome distribution. For these tags, large numbers of true positives dilute the effect of FP and FN, which helps explain their consistently strong F1-scores. For example, AraBERTv02 achieves 633 true positives for B-LOC but only 66 false positives and 35 false negatives, resulting in a strong F1 (0.926).

In contrast, lower-frequency tags such as I-MISC, B-MISC, and I-LOC are more sensitive to errors. AraBERTv02 shows 68 true positives but 15 FP and 144 FN for I-MISC, leading to a recall collapse (0.412). False negatives are the dominant error type in AraBERTv02 across most tags—except for B-LOC, I-LOC, and O—indicating that the model frequently misses entity tokens altogether, which drives its lower recall.

BERT shows a different pattern. Its larger number of examples reduces the proportional impact of errors, and FP and FN counts are more balanced across tags. For exam-ple, `I-MISC` still produces many false negatives, but with 161 TP against 73 FP and 55 FN, the effect is less extreme than in AraBERTv02. Similarly, for B-ORG, BERT records 1,523 TP with 154 FP and 138 FN, a much more balanced distribution than AraBERTv02 (355 TP, 76 FP, 95 FN). This balance explains BERT's narrower precision–recall gap and its stronger macro scores.

For the O tag, the error patterns differ between the two models. AraBERTv02 produces more false negatives, often mislabelling entities as non-entities (exclusion errors), while BERT produces more false positives, sometimes misclassifying non-entities as entities (inclusion errors).

### 4.5.5 Error Analysis

This section addresses aspect (c) of RQ3—*what types of errors affect performance at each level and how they differ across languages*—by analysing structured error categories derived from misalignments between true and predicted labels. We analyse errors at both the token and entity levels. At the token level, errors are captured directly through standard tag prediction outcomes (e.g., predicting B-LOC instead of B-PER). At the entity level, we extend `seqeval` with custom functionality to classify errors into a taxonomy of four types:

- **Boundary error**: Correct entity type but incorrect span boundaries.
- **Entity error**: Correct boundaries but incorrect entity type.
- **Entity and boundary error**: Both type and boundaries are incorrect.
- **O error**: False positives (non-entities predicted as entities) or false negatives (entities predicted as non-entities).

Table 4.18 provides span-based examples for each of the four error types in the taxonomy.

**Overview of Error Types in Entity-Level Evaluation** To understand the nature and distribution of entity-level errors in AraBERTv02 and BERT, we first analyse how the four error types are distributed across false positives (FP) and false negatives (FN) under both IOB1 and IOB2 annotation schemes. This provides a high-level view of the dominant error patterns before turning to their distribution across specific entity spans (LOC, MISC, ORG, PER).

Table 4.18: Examples of the four span-level error types defined in the error taxonomy.

| Error Type | Span-Based Example |
|---|---|
| Boundary Error | True: *[New York City]*$_{\text{LOC}}$; Predicted: *[New York]*$_{\text{LOC}}$ |
| Entity Error | True: *[Apple]*$_{\text{ORG}}$; Predicted: *[Apple]*$_{\text{MISC}}$ |
| Entity and Boundary Error | True: *[New York City]*$_{\text{LOC}}$; Predicted: *[New York]*$_{\text{PER}}$ |
| O Error | True: *the (O)*; Predicted: *[the]*$_{\text{LOC}}$ |

Figure 4.39 illustrates the distribution of error types within FP and FN for both models and annotation schemes. It is important to note that **O errors** differ in interpretation depending on context. Within FP, they represent *inclusion errors*, where the model incorrectly predicts an entity for a span that is truly labelled as O (non-entity). Within FN, they represent *exclusion errors*, where a true entity is misclassified as O, meaning the model fails to recognise an entity span that should have been detected.

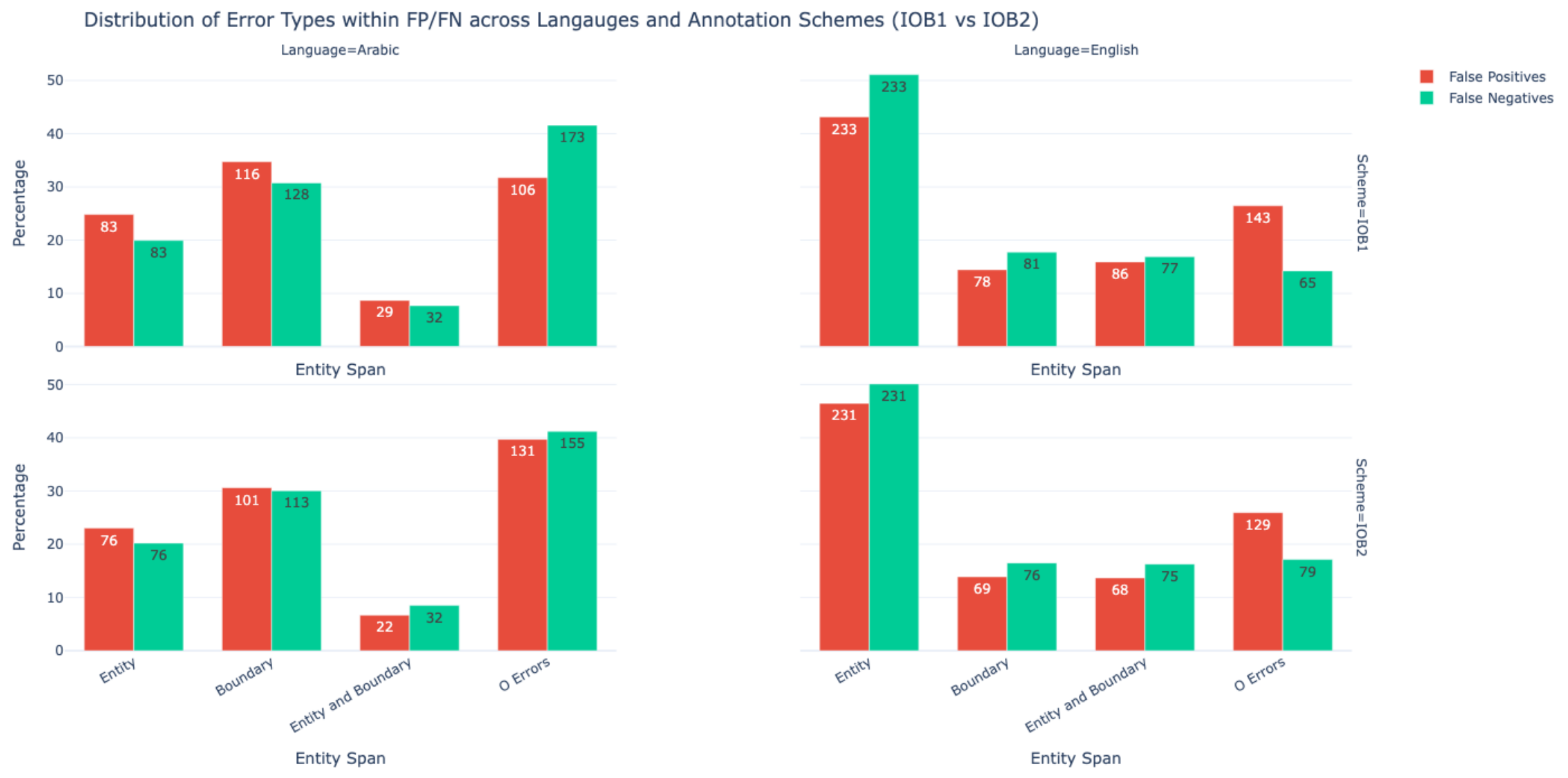

Figure 4.39: Distribution of span-level error types within false positives (FP) and false negatives (FN) for AraBERTv02 (Arabic) and BERT (English), categorised by annotation scheme (IOB1 vs IOB2).

**Error Patterns under IOB1** Under the IOB1 scheme (top row of Figure 4.39), AraBERTv02 and BERT show distinct error distributions.

For AraBERTv02, the largest proportion of false positives (red bars) arises from **boundary errors**, followed by **O errors** (inclusion) and then **entity errors**. In contrast, the dominant source of false negatives (green bars) is **O errors** (exclusion), followed by boundary errors, with entity errors being least common. This suggests that when AraBERTv02 misses entities (FN), it is mainly because true entities are misclassified as non-entities, whereas its false positives typically stem from boundary misplacements, with some additional inclusion errors.

BERT exhibits a different error distribution under the IOB1 scheme (top right). **Entity errors** are the most frequent error type in both false positives (red bars) and false negatives (green bars). **Boundary errors** and combined **Entity and Boundary errors** are the least common for BERT in both FP and FN. Furthermore, **O errors** (inclusion errors) are more prevalent in BERT's FP than **O errors** (exclusion errors) in its FN. This indicates that BERT's main challenge lies in correctly classifying entity types, leading to both incorrect predictions and missed entities. BERT also shows a tendency to over-predict entities by misclassifying non-entity tokens (inclusion errors) more often than it fails to detect true entities (exclusion errors).

In summary, under the IOB1 scheme, the main error patterns differ between the two models. AraBERTv02 is most affected by its tendency to misclassify true entities as non-entities (**O errors** as exclusion errors in FN) and by inaccuracies in span segmentation (**Boundary errors** in FP). BERT, by contrast, is most affected by misclassifying entity types, with **Entity errors** being the most frequent in both FP and FN. While AraBERTv02 struggles more with completely missing entities, BERT is more prone to assigning incorrect types. Both models, however, share a common tendency to over-predict entities, with **O errors** as inclusion errors in FP.

**Error Patterns under IOB2** Under the IOB2 scheme (bottom row of Figure 4.39), both models show distinct shifts in their error distributions.

For AraBERTv02, most error counts decrease in both false positives (red bars) and false negatives (green bars). The main exceptions are **Entity and Boundary errors** in FN, which remain at a similar level, and **O errors** (inclusion) in FP, which increase. This rise in inclusion errors suggests that under the stricter IOB2 constraints, AraBERTv02 more often misclassifies non-entity tokens as entities. At the same time, reductions in boundary and exclusion errors indicate that stricter span rules mitigate some of the segmentation and detection mistakes observed under IOB1.

For BERT, the overall reduction in error counts when moving to IOB2 is smaller than for AraBERTv02. **Entity errors** show the least improvement, remaining the dominant error type in both FP and FN. In addition, **O errors** (exclusion) increase within FN, meaning BERT misses more true entities under IOB2 than under IOB1. This reflects the effect of stricter evaluation mode, which occasionally discard spans that would have been counted under IOB1's repair mechanism.

In summary, IOB2 affects the two models differently. AraBERTv02 benefits more overall, reducing boundary and exclusion errors but at the cost of more inclusion errors. BERT, by contrast, shows only small reductions in error counts and a rise in exclusion errors, highlighting its continued difficulty with entity-type classification. Despite these differences, both models demonstrate that stricter annotation rules shift the balance of FP and FN without fundamentally altering the dominant error types observed under IOB1.

**Why FPs and FNs Change Between Schemes** As shown earlier, switching between annotation schemes (IOB1 vs. IOB2) and evaluation modes (Non-Strict vs. Strict) affects how errors are distributed across false positives (FP) and false negatives (FN). Here we explain why these shifts occur by examining how predicted labels and gold-standard annotations are reinterpreted under each setting.

For AraBERTv02, the rise in **FPs (inclusion errors)** under IOB2 can be traced to inconsistencies between gold-standard labels in IOB1 and IOB2. In Seqeval's Non-Strict mode (IOB1), the *repair* mechanism accepts spans that begin with an `I-` tag without a preceding `B-`. Under IOB2, however, the *discard* mechanism is applied: spans not explicitly starting with `B-` are dropped or relabelled as `O`. If AraBERTv02 predicts these tokens as entities (often semantically correct given the context), the predictions are now mismatched against the stricter IOB2-aligned gold standard and therefore counted as false positives.

Table 4.19 illustrates this behaviour with a simple example. The gold label sequence begins with `I-PER`, which is valid under IOB1 but discarded under IOB2. While AraBERTv02 correctly predicts B-PER, the evaluation outcome flips: counted as a true positive under IOB1 but as a false positive under IOB2.

Table 4.19: Example of Increased False Positives under IOB2 due to Gold Standard Change

| **Annotation Scheme** | **Token Sequence and Labels** |
|---|---|
| Gold Standard<br>Prediction | *I*-PER O O O O O O<br>*B*-PER O O O O O O |
| Evaluation (IOB1) | *Hayam* → PER (True Positive) |
| Evaluation (IOB2) | *Hayam* (Predicted: PER, True: O) → False Positive |

As shown in Table 4.19, the gold standard starts the person entity "Hayam" with an I-PER tag. While this does not strictly conform to IOB1 guidelines, Seqeval's repairing mechanism treats it as a valid PER span. AraBERTv02 correctly predicts B-PER, which is counted as a True Positive under IOB1. However, under IOB2 the gold label is discarded to O due to its strict beginning-tag requirement. Consequently, AraBERTv02's B-PER prediction is reclassified as a False Positive because the gold standard no longer marks "Hayam" as a person entity. Importantly, the model did not necessarily make an error—in fact, predicting "Hayam" as a person is semantically correct—but discrepancies in annotation schemes still propagate into the evaluation metrics. This limitation is revisited in Chapter 5.

Conversely, the increase in FN (exclusion errors) for BERT under IOB2 stems from cases where its predictions contained boundary errors that were previously repaired by Seqeval's mechanism. Under IOB1, even when predicted entity spans did not exactly match the gold standard boundaries, Seqeval could repair them, resulting in fewer false negatives. Under IOB2's strict rules, however, any boundary mismatch is treated as a failure to detect the entity. As a result, predictions that were previously counted as correct under IOB1 are now considered false negatives under IOB2. Table 4.20 illustrates this with the example: "Visiting New York City is fun."

As shown in Table 4.20, the gold standard correctly labels "New York City" as a LOC entity starting with B-LOC. BERT's prediction begins with I-LOC, which violates IOB2's strict requirement for entities to start with B-. Under IOB1's repair mechanism, the span is accepted and scored as a true positive. Under IOB2, however, the span is dis-carded, causing the model's prediction to be counted as a false negative. This illustrates how IOB1's repairing mechanisms can mask boundary errors that IOB2 exposes.

Table 4.20: Example of increased false negatives for BERT under IOB2 due to stricter span requirements.

| Annotation Scheme | Token Sequence and Labels |
|---|---|
| Gold Standard | *B*-LOC New *I*-LOC York *I*-LOC City |
| BERT Prediction | *I*-LOC New *I*-LOC York *I*-LOC City |
| Evaluation (IOB1) | *New York City* → LOC (True Positive) |
| Evaluation (IOB2) | *New York City* (Predicted: LOC span discarded) → False Negative |

It is important to note that this behaviour does not represent a limitation of `seqeval`. Rather, it reflects a design choice, well established in the NER literature and adopted by many evaluation tools as discussed in Section 2.3.4. Specifically, non-strict evaluation applies a *repair* mechanism, while strict evaluation applies a *discard* mechanism. These mechanics are widely recognised in NER evaluation practice Tjong Kim Sang and De Meulder (2003); Cuevas Villarmin et al. (2024); Palen-Michel et al. (2021); Lignos and Kamyab (2020), and have been discussed in the context of reproducibility and scoring consistency. Their role is not to introduce errors but to define how sequences are interpreted during scoring. Hence, observed differences between IOB1 and IOB2 results should be understood as consequences of the evaluation scheme, rather than flaws in the chosen tool.

#### 4.5.5.1 Span-Level Entity Error Patterns

Building on the overall distribution of error types, we now examine how these errors are distributed across specific entity spans (LOC, ORG, PER, MISC). This span-level analysis offers a more detailed view of model behaviour, showing which entity categories are most affected by each error type. To achieve this, we break down FP and FN outcomes separately and classify the error types within each, enabling us to pinpoint where AraBERTv02 and BERT diverge most under the two annotation schemes (IOB1 vs. IOB2).

**False Positive Error Types By Entity Span** To compare FP errors across models and annotation schemes, we examine how error types are distributed across entity spans. FP errors, occurring when models predict entities that do not exist in the gold standard, are categorised into Boundary, Entity, Entity and Boundary, and Inclusion errors as described in Section 4.5.5. Figure 4.40 presents this distribution as a heatmap.

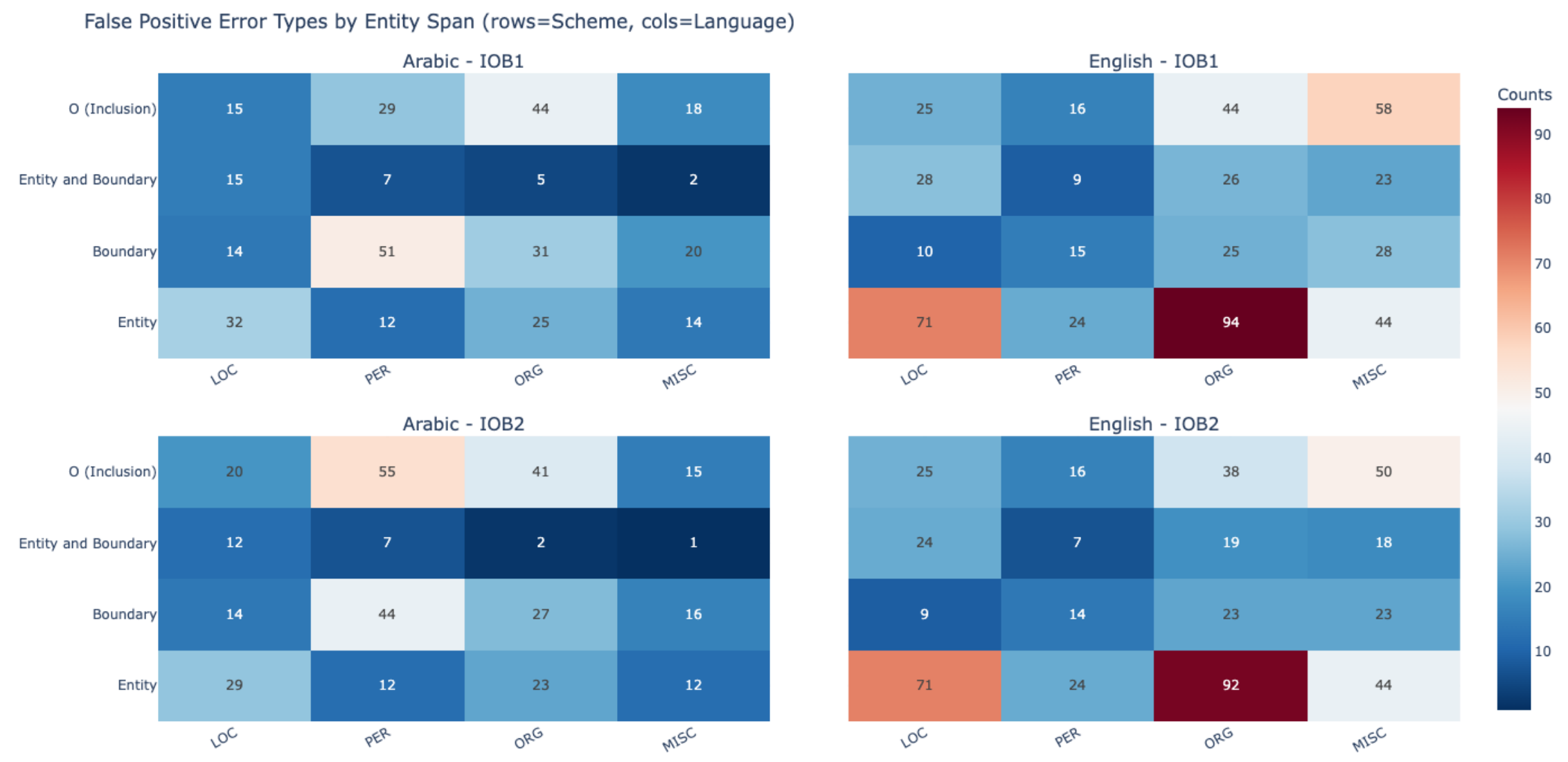

Figure 4.40: False Positive error type heatmap by entity span, for AraBERTv02 (Arabic) and BERT (English), categorised by annotation scheme (IOB1 vs. IOB2).

For AraBERTv02 (IOB1), Boundary errors are most prevalent, particularly in PER (51) and ORG (31). Entity errors occur most often in LOC (32) and ORG (25). Entity and Boundary errors are rare, with LOC the most affected. Inclusion errors, where non-entity words are falsely classified as entities, are common in ORG (44) and PER (29), suggesting over-prediction in these spans.

For BERT (IOB1), Entity errors dominate across spans, especially in ORG (94), LOC (71), and MISC (44). Inclusion errors are also substantial, particularly in MISC (58) and ORG (44). Entity and Boundary errors appear more frequently in BERT than AraBERTv02, especially in ORG (26) and MISC (23). By contrast, Boundary errors are the least common, with MISC (28) and ORG (25) most affected.

Under IOB1, the main difference is that AraBERTv02 shows more Boundary errors (notably in PER and ORG), while BERT is dominated by Entity errors (ORG, LOC, MISC). PER is a major source of FP in AraBERTv02 but less so in BERT. Both models frequently over-predict ORG entities, reflected in high Inclusion error rates.

Under IOB2, FP errors generally decrease for both models. For AraBERTv02, PER Inclusion errors increase, suggesting stricter rules make PER more prone to over-prediction.

For BERT, Entity errors remain similar to IOB1, but Inclusion and Boundary errors decrease in MISC and ORG, suggesting improved boundary handling.

To illustrate these shifts, we examine concrete examples from AraBERTv02 and BERT predictions. For AraBERTv02 (Example 209), which shows how boundary errors can be reinterpreted under IOB2, the gold label sequence is O → I-PER, while the model predicts B-PER → I-PER. Under IOB1, this is scored as a **Boundary Error** since the predicted span begins one token too early. Under IOB2, however, the transition O → I-PER in the gold labels is invalid and discarded, leaving the first token as O. The prediction still forms a PER entity, so the misalignment is now treated as an **Inclusion Error**. This explains why inclusion errors increase while boundary errors decrease for AraBERTv02 under IOB2, particularly for PER. Table 4.21 demonstrates this example.

For BERT (Sentence 1700), the gold label is O, but the model predicts I-ORG. Under IOB1, this prediction is repaired into a valid entity span and scored as an **Inclusion Error**. Under IOB2, the invalid I-ORG tag is discarded, leaving both prediction and gold as O, so the error disappears. This contributes to the reduction in inclusion errors observed for BERT under IOB2.

Table 4.21: Example 209: Span reinterpretation under IOB1 vs. IOB2. AraBERTv02 prediction shifts from a Boundary Error to an Inclusion Error.

| **Annotation Scheme** | **Token Sequence and Labels** |
|---|---|
| Gold Standard (IOB1) | O I-PER |
| Prediction | B-PER I-PER |
| Evaluation (IOB1) | Prediction: (1, 2, PER), True: (2, 2, PER) → **Boundary Error** |
| Gold Standard (IOB2) | O O (invalid I-tag discarded) |
| Prediction | B-PER I-PER |
| Evaluation (IOB2) | Prediction: (1, 2, PER), True: no span → **Inclusion Error** |

In summary, shifts in error patterns arise not from new mistakes but from how the evaluation scorer, under different annotation schemes, interprets predictions and gold labels. For AraBERTv02, the changes mainly reflect inconsistencies in the gold standard that are discarded under IOB2's stricter rules, reclassifying what were boundary errors into inclusion errors. For BERT, the differences arise from incorrect predictions that IOB1's repairing mechanism had previously masked: under IOB2 these boundary mismatches are exposed, increasing false negatives. Thus, for AraBERTv02, gold standard quality plays a more prominent role in shaping error counts, while for BERT, stricter evaluation rules more directly reveal weaknesses in its predictions. These dynamics highlight that error shifts are driven by the interaction between evaluation scheme, scorer logic, and data quality, rather than by changes in model behaviour itself[4].

**Entity Misclassification in False Positives** Since entity errors form a major component of false positives—particularly for BERT—we now examine how often models confuse one entity type for another. This analysis focuses on span-level misclassifications (e.g., predicting ORG instead of LOC) and compares patterns across languages and annotation schemes. Figure 4.41 presents a heatmap of these cross-entity confusions within false positives.

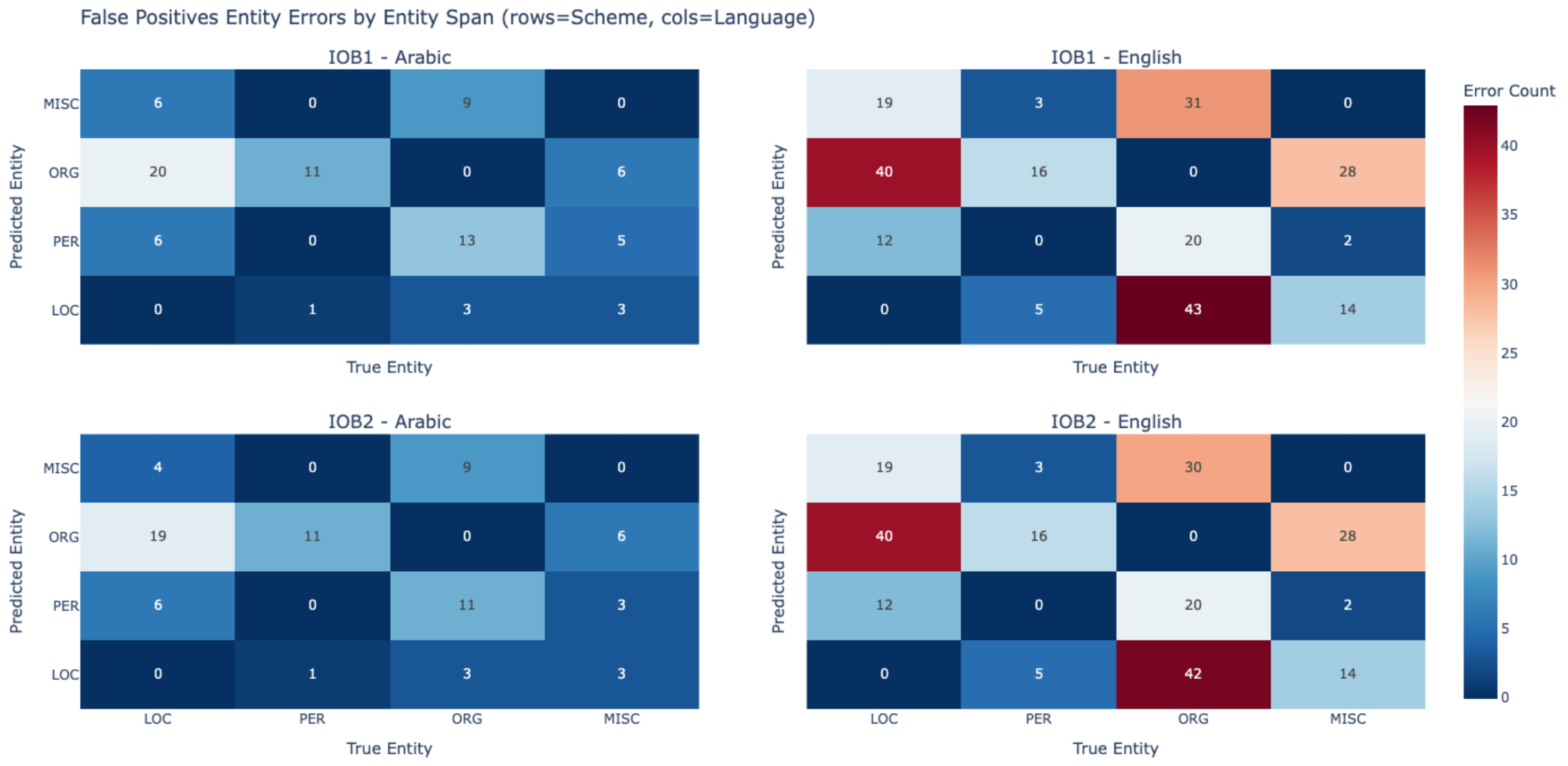

Figure 4.41: Distribution of Entity Errors in False Positives Across Models and Annotation Schemes.

In AraBERTv02, the most frequent misclassifications occur between LOC and ORG (e.g., LOC $\rightarrow$ ORG), with additional confusion between ORG and PER (ORG $\leftrightarrow$ PER). ORG is also often predicted as MISC. Switching from IOB1 to IOB2 has little effect on these patterns, indicating that AraBERTv02's entity-type errors are largely stable across schemes.

For BERT, the influence of the annotation scheme is similarly small, but the misclassification patterns are clearer. ORG is the most error-prone category, frequently confused with LOC, MISC, and PER (ORG → LOC, ORG → MISC, ORG → PER). LOC is also misclassified as ORG and MISC, while MISC is often confused with both ORG and LOC. These recurring error pairs reflect semantic or contextual overlap between entity categories and are examined further in Section 4.3.

[4] This discussion concerns error count changes resulting from the annotation scheme shift, not overall performance differences between the models.

**False Negative Error Types by Entity Span** We now turn to FN and apply the same span-level analysis used for false positives. Here, we examine which entity spans are most commonly missed by the models and how these errors are distributed across different entity types. As before, we consider the four error categories, with O errors in this context referring to **exclusion errors**. Figure 4.42 presents the error heatmap for both models and annotation schemes.

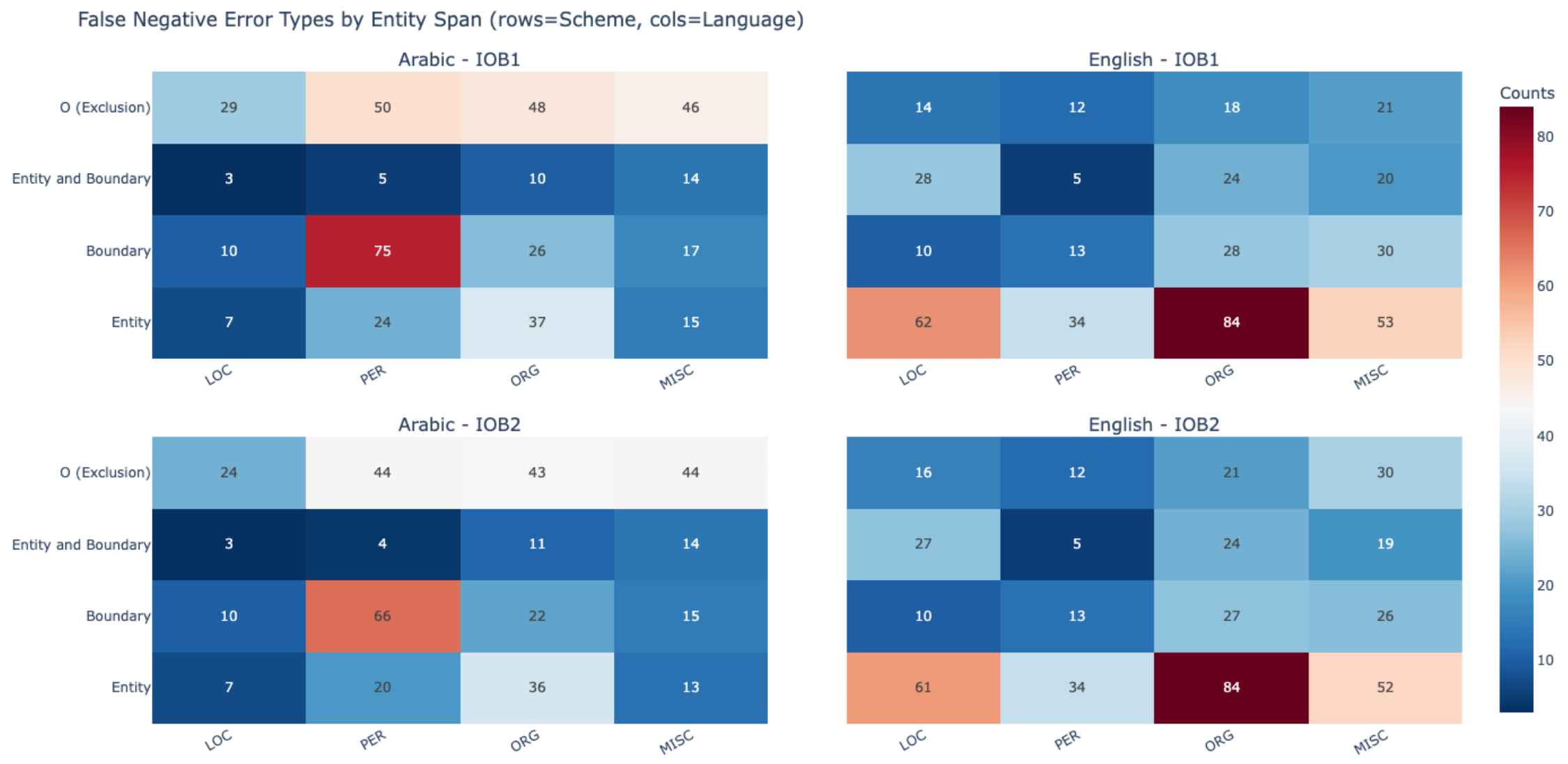

Figure 4.42: False negative error types by entity span, across AraBERTv02 (Arabic) and BERT (English), categorised by annotation scheme (IOB1 vs. IOB2).

For AraBERTv02, under the IOB1 scheme, the most prevalent false negatives are exclusion errors, where true entities are misclassified as non-entities. Boundary errors also contribute to FN, particularly for PER. Additionally, entity errors occur, with ORG entities frequently misclassified (e.g., 37 instances). Under IOB2, exclusion and boundary errors for PER decrease, however, the overall reduction in FN errors for AraBERTv02 is smaller compared to the reduction observed in FP errors.

For BERT under IOB1, **entity errors** dominate the FN distribution, especially for LOC, MISC, and ORG, while PER is least affected. Boundary errors contribute mainly for MISC and ORG, but exclusion errors are less prominent than in AraBERTv02. Under IOB2, the distribution remains broadly similar, but exclusion errors increase, suggesting that stricter rules cause some entities recognised under IOB1 to be missed.

These shifts can be illustrated with two examples. In AraBERTv02 (Example 36), the true sequence is `B-ORG I-ORG I-ORG` while the model predicts `O I-ORG O`. Under IOB1, this partial span is repaired, producing a boundary error (FN). Under IOB2, however, the invalid `O` → `I-ORG` transition in the gold standard is discarded, leaving only O labels, and the model's prediction is reclassified as an exclusion error. In BERT (Example 207), the true sequence is `B-PER I-PER I-PER I-PER` while the model predicts `B-ORG I-PER I-PER I-PER`. Under IOB1, this is an entity+boundary error (FN). Under IOB2, the stricter span rules reclassify the output as a false positive, since the prediction is now treated as an invalid ORG span rather than a partial PER.

In summary, AraBERTv02 tends to benefit from IOB2 through reduced exclusion and boundary errors in FN, while BERT shows less consistent changes, with exclusion errors becoming more common in some spans. As with the FP analysis, AraBERTv02's shifts are strongly influenced by gold standard inconsistencies that IOB2 discards, whereas BERT's shifts mainly reflect stricter evaluation rules exposing errors that IOB1 would have repaired.

**Entity Misclassification in False Negatives** Extending the previous analysis of entity errors in false positives, we now examine how entity confusion occurs in false negatives. Figure 4.43 shows how true entities are misclassified into other types when missed by the models, across both annotation schemes.

For AraBERTv02, the FN misclassification pattern differs from FP. ORG entities are most frequently misclassified as LOC (ORG ⟶ LOC) and PER (ORG ⟶ PER). PER entities are also commonly misclassified as ORG (PER ⟶ ORG), while MISC entities tend to be misclassified as ORG (MISC ⟶ ORG). Switching to IOB2 has minimal impact on these trends, suggesting that AraBERTv02's entity misclassifications persist regardless of the scheme.

For BERT, similar confusion patterns emerge. ORG is the entity most often misclassified, particularly as LOC (ORG ⟶ LOC), MISC (ORG ⟶ MISC), and PER (ORG ⟶ PER). LOC is frequently misclassified as ORG (LOC ⟶ ORG) and MISC (LOC ⟶ MISC), while MISC is most often misclassified as ORG (MISC ⟶ ORG) and LOC (MISC ⟶ LOC).

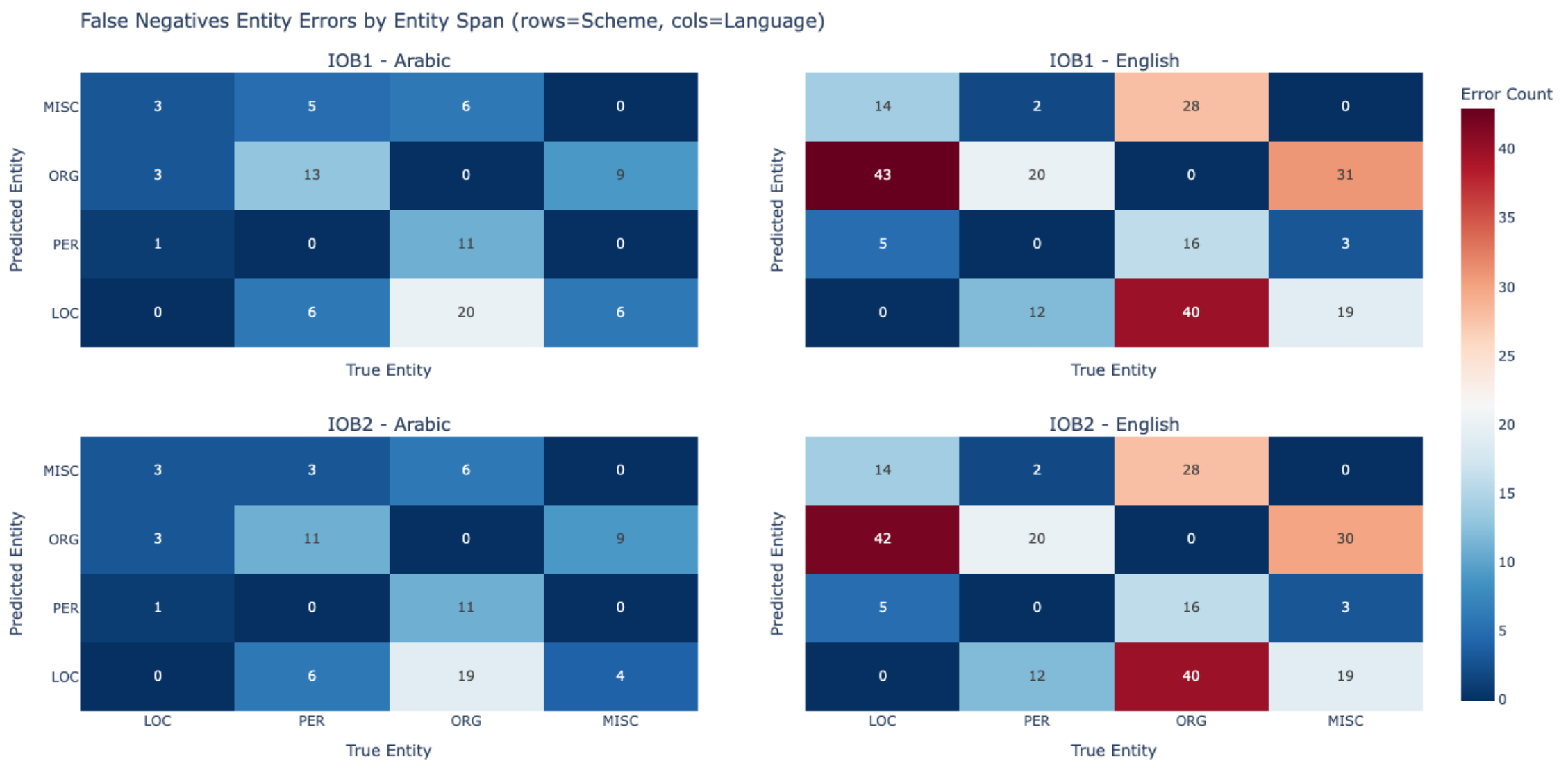

Figure 4.43: Distribution of entity errors in false negatives across AraBERTv02 (Arabic) and BERT (English), categorised by annotation scheme (IOB1 vs. IOB2).

**Token-Level Error Analysis** Having examined error patterns at the span level, we now shift focus to the token level. At this level, each tag encodes both entity type (PER, ORG, LOC, MISC) and position within the span (B-, I-, O). This makes it possible to identify boundary and type confusions directly, without the complexities introduced by span-level aggregation or annotation scheme variations.

Figure 4.44 presents the token-level confusion matrix. Here, false negatives for a true label (row) occur when it is misclassified as another label (column). For example, if the true label is B-LOC but the model predicts O, the error contributes to FN for B-LOC. Conversely, false positives for a predicted label arise when another true label (e.g., O) is misclassified as that entity tag (e.g., B-PER).

While the error analysis approach differ at this level, the errors can still be grouped into the same four categories used in the span analysis: **Entity Errors** (correct chunk, wrong entity type), **Boundary Errors** (correct type, wrong boundaries), **Entity and Boundary Errors** (both wrong), and **O Errors** (inclusion or exclusion of entity tags). Colour intensity in the heatmap indicates misclassification frequency.

Several patterns can be identified in AraBERTv02's misclassifications. The model frequently misclassifies entity tags as O, leading to exclusion errors and contributing to higher false positives for the O tag. This aligns with AraBERTv02's precision-oriented approach. For high-support entity tags like B-LOC, the dominant error is misclassification as O, indicating false negatives. For B-PER and I-PER, boundary errors (misclassification between them) are common. Low-support tags (B-MISC, I-MISC) are mostly misclassified as O, suggesting potential over-fitting or OOV issues, discussed in Section 4.3. Additionally, AraBERTv02 struggles to distinguish B-ORG and I-MISC from B-LOC.

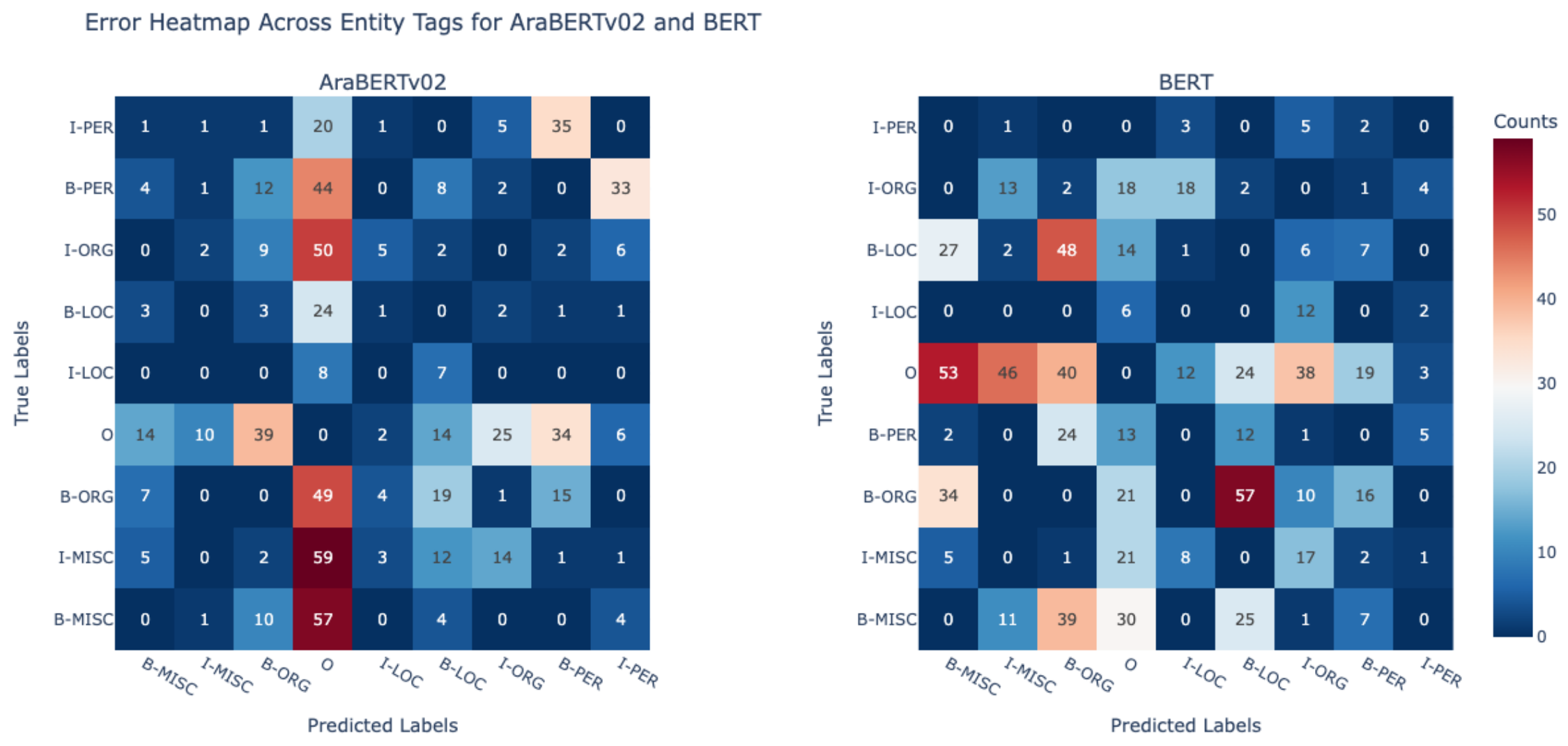

Figure 4.44: Token-level confusion heatmap showing misclassification patterns across entity tags for AraBERTv02 (Arabic) and BERT (English). Colour intensity reflects the frequency of errors.

In BERT, error patterns are more distributed. A key trend is the frequent prediction of O tags as entities (inclusion errors), the opposite of AraBERTv02's behaviour. Entity type errors are also dominant. Similar confusion occurs between B-MISC and B-ORG. While exclusion errors are present in BERT, they are less frequent than entity type errors. For the low-support tag I-MISC, confusion with I-LOC and I-ORG is consistent, suggesting entity type confusion remains a key issue. Boundary errors between B-PER and I-PER are minimal in contrast to AraBERTv02. Across most entity tags in BERT, entity type confusion is the main error, with boundary errors being less common.

In summary, the token-level confusion matrix reveals distinct error patterns in both models. AraBERTv02 shows more boundary errors (e.g., `B-PER` $\leftrightarrow$ `I-PER`) and a tendency towards exclusion errors (entity tags $\rightarrow$ O), reflecting a precision focus. Conversely, BERT exhibits more entity type errors (e.g., `B-MISC` $\leftrightarrow$ `B-ORG`) and inclusion errors (O → entity tags), suggesting a recall-focus. Tags with low support present challenges for both models, while high-support tags show different error types.

### 4.5.6 Support–Performance Correlation Analysis

Finally, to address aspect (e) of RQ3—*how performance varies in relation to label frequency at the token level*—we examine the relationship between entity tag frequency (support) and model performance (precision and recall). Correlations are calculated using both Pearson and Spearman coefficients for AraBERTv02 and BERT across the training and test splits. Pearson correlation captures linear associations, while Spearman correlation captures rank-based trends. By considering support in both training and test sets, we assess how exposure during training and representation in evaluation data influence token-level performance across tags.

Figure 4.45 shows the distribution of entity tag support in the training split, highlighting differences compared to the test set (e.g., I-LOC vs. I-MISC in AraBERTv02, B-PER vs. B-ORG and I-MISC vs. I-LOC in BERT). These variations provide context for interpreting the correlation patterns.

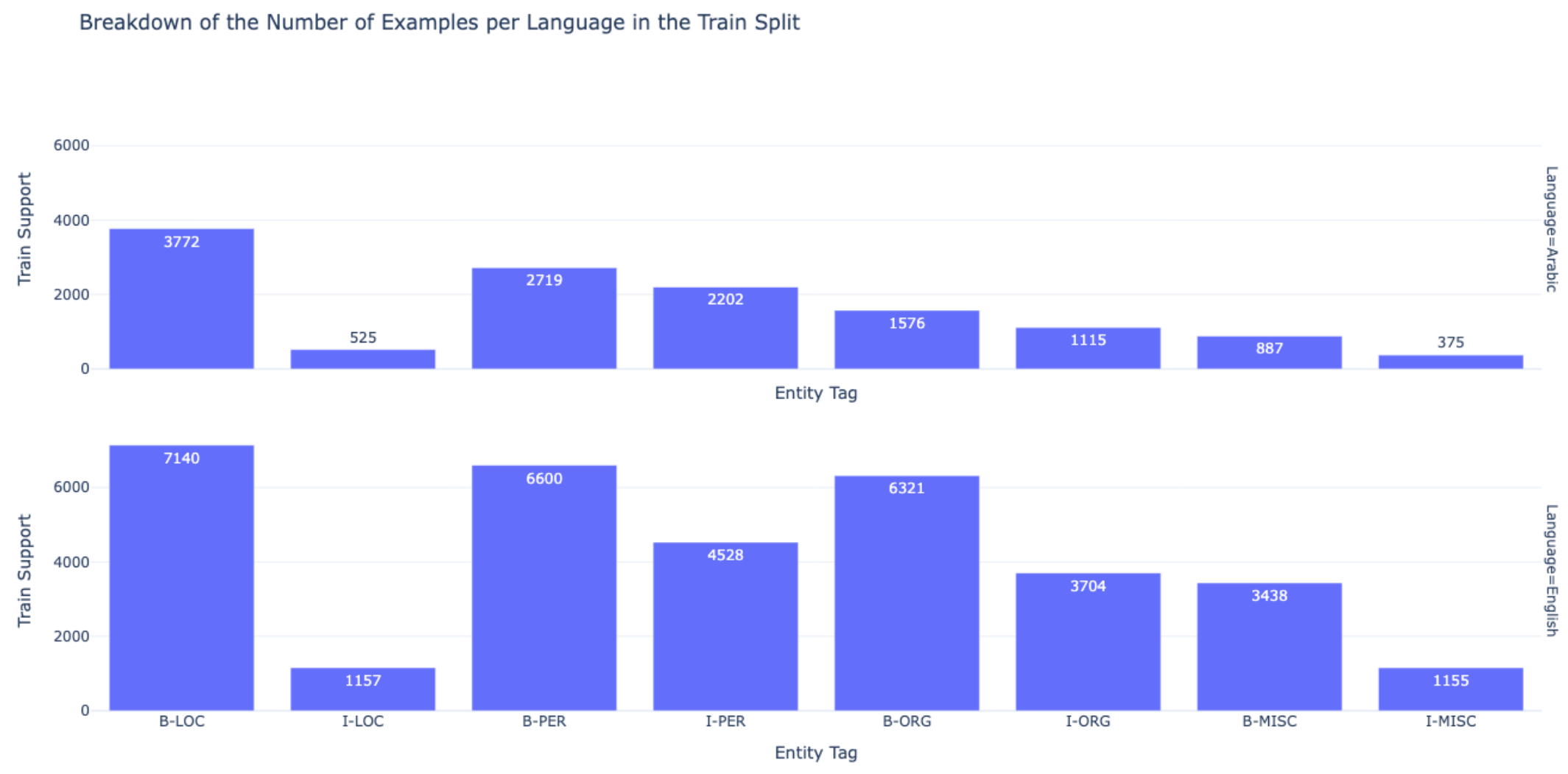

Figure 4.45: Breakdown of the number of examples (support) per entity tag in the training split for AraBERTv02 (Arabic) and BERT (English).

These variations in support distribution are important for interpreting the correlation results in Figure 4.46, where we examine the relationship between tag frequency (support) and performance (precision and recall). The "O" tag is excluded from the anal-ysis because of its high frequency, which would dominate the correlation and obscure trends for entity tags.

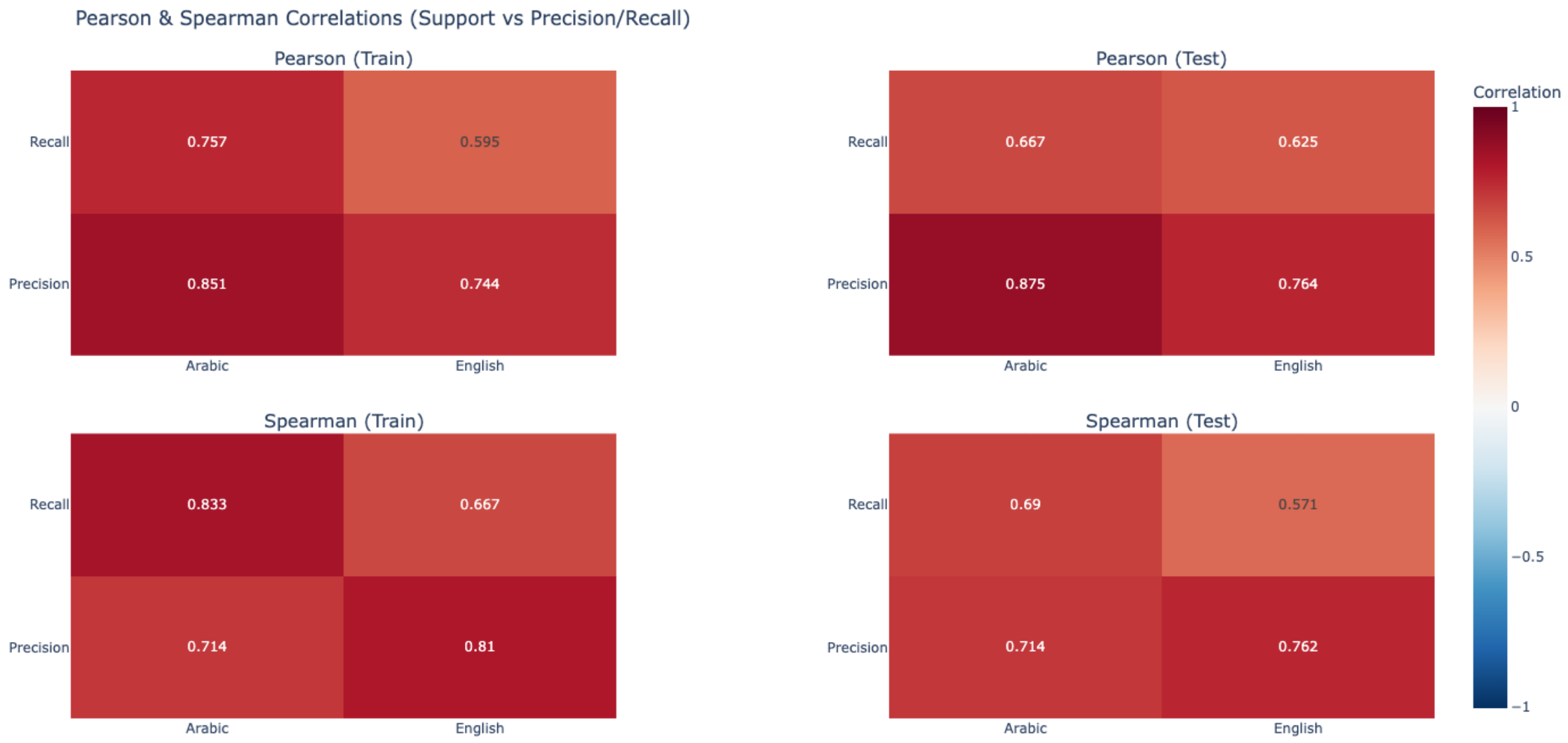

Figure 4.46: Pearson and Spearman correlations between support (number of examples) and performance metrics (precision and recall) for AraBERTv02 (Arabic) and BERT (English). The "O" tag is excluded to avoid imbalance.

The correlation analysis reveals distinct model behaviours. For Pearson correlation, AraBERTv02 consistently shows a stronger positive linear relationship between support and both precision and recall than BERT. Precision correlates more strongly with support than recall in both models. From train to test, precision correlations strengthen for both models, while recall correlations diverge—decreasing in AraBERTv02 but slightly increasing in BERT.

For Spearman correlation, AraBERTv02 in the training split shows stronger alignment between support rank and recall rank, whereas BERT shows stronger alignment with precision rank. In the test split, AraBERTv02's recall correlation drops sharply, while its precision correlation remains stable. For BERT, both precision and recall correlations weaken from train to test, suggesting greater sensitivity to distributional changes.

Overall, AraBERTv02 shows a stronger association between tag frequency and performance than BERT, but these relationships are not equally stable across splits. In particular, AraBERTv02's recall appears more sensitive to changes in data distribution, whereas BERT's weaker correlations suggest that its performance is less directly influenced by frequency. These results address RQ3(e) by highlighting how training exposure and tag frequency influence model behaviour differently across languages. The follow-ing paragraphs explore the specifics of Pearson and Spearman correlations to better understand these differences.

**Pearson Correlation** Following the overall correlation overview, we now explore the two correlation types separately, starting with Pearson correlation, which captures linear relationships between entity tag frequency (support) and performance metrics (precision and recall). This relationship depends on two statistical components: covariance, which measures how two variables vary together, and the product of their standard deviations, which normalises for scale. Even if two variables trend in the same direction, their correlation weakens when changes in one are not proportional to changes in the other. This makes Pearson particularly sensitive to differences in how support varies across the training and testing splits.

Figure 4.46 (top row) shows that AraBERTv02 exhibits a stronger Pearson correlation with both precision and recall across both splits compared to BERT, especially for precision. To explain these trends, we begin by analysing AraBERTv02's behaviour in more detail using scatter plots that visualise the alignment and spread of support and performance across entity tags.

**Pearson Correlation: AraBERTv02** Figure 4.47 shows how support and performance metrics align for AraBERTv02 across the train and test splits. Each point represents an entity tag, with dotted lines marking the mean values of support and performance. In the training split, the tags cluster along a clear diagonal: tags with above-average precision generally also have above-average support, and vice versa. This directional alignment produces high covariance. However, the spread of support counts is large (standard deviation = 1178.1), so the relationship is not consistently proportional—some tags show extreme support values with only small precision differences. As a result, the linear correlation is weakened despite the consistent trend.

In contrast, the test split shows a more compressed distribution of support values (standard deviation = 276.9). The same general alignment between support and precision holds, with a few exceptions such as B-ORG, which lies above the mean in support but below the mean in precision, contributing negative covariance. Nevertheless, because most tags deviate less and remain tightly clustered, the reduced variance strengthens the proportionality of the relationship, yielding a slightly higher Pearson correlation in the test split compared to training.

Recall shows similar trends. In training, tags with high support tend to have high recall, showing a strong positive relationship. However, this relationship is weaker in the test split. Some tags, like I-LOC and B-ORG, have high recall despite low support, which weakens the overall positive relationship. This, along with inconsistent changes in recall for different tags, results in a lower correlation in the test set compared to training.

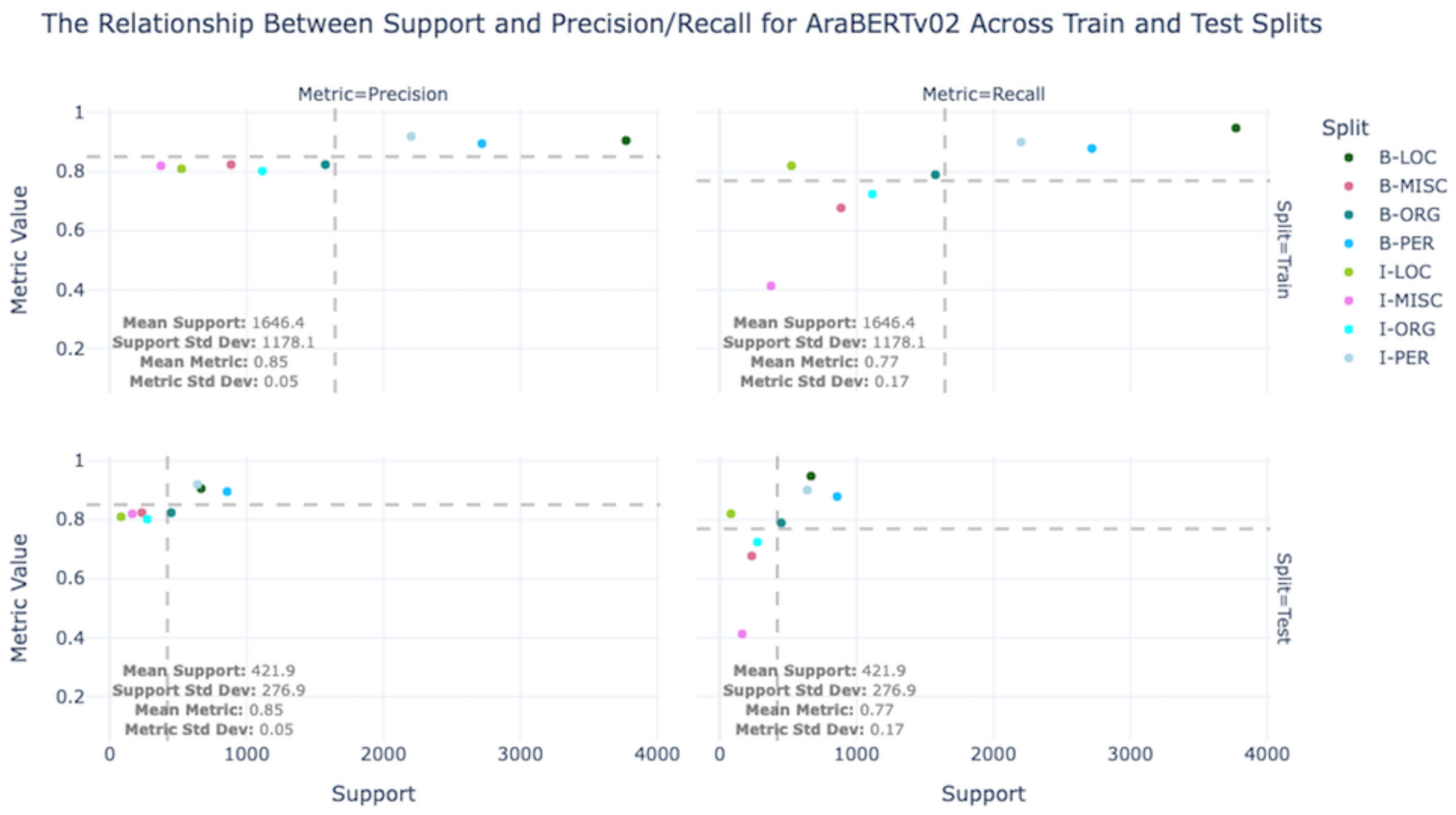

Figure 4.47: Scatter plots illustrating the relationship between support and performance metrics for AraBERTv02 across train and test splits. Dashed lines represent the mean values.

Table 4.22 summarizes the statistical components behind these observations. Covariance is higher in training, especially for recall, but due to the larger standard deviation of support, the normalized Pearson correlation is lower. In the test set, reduced variance strengthens the correlation values, even with small negative contributions from specific tags.

To summarize, AraBERTv02's Pearson correlations indicate a strong positive relationship between support and performance in both splits. Precision benefits more in the test set due to the tighter distribution of support, which enhances proportionality. Recall, on the other hand, shows higher correlation in the training split, where large increases in support align with recall changes. However, this alignment weakens in the test set as variation in recall becomes less proportional to support.

Table 4.22: Breakdown of covariance, standard deviation product, and Pearson correlation between support and performance metrics (precision and recall) for AraBERTv02 across the train and test splits.

| Split | Covariance | | Std Dev Product | | Pearson Correlation | |
|---|---|---|---|---|---|---|
| | Precision | Recall | Precision | Recall | Precision | Recall |
| Train | 48.10 | 151.51 | 56.54 | 200.16 | 0.85 | 0.76 |
| Test | 11.62 | 31.39 | 13.29 | 47.05 | 0.87 | 0.67 |

**Pearson Correlation: BERT** Figure 4.48 shows how support relates to performance for BERT in both training and testing. Similar to AraBERTv02, tags with higher support generally have higher precision and recall, showing a positive relationship. However, this relationship is not as strong in BERT, especially because there's more variation in support during training. Tags like I-ORG and B-LOC have large support changes without consistent precision changes, which weakens the correlation.

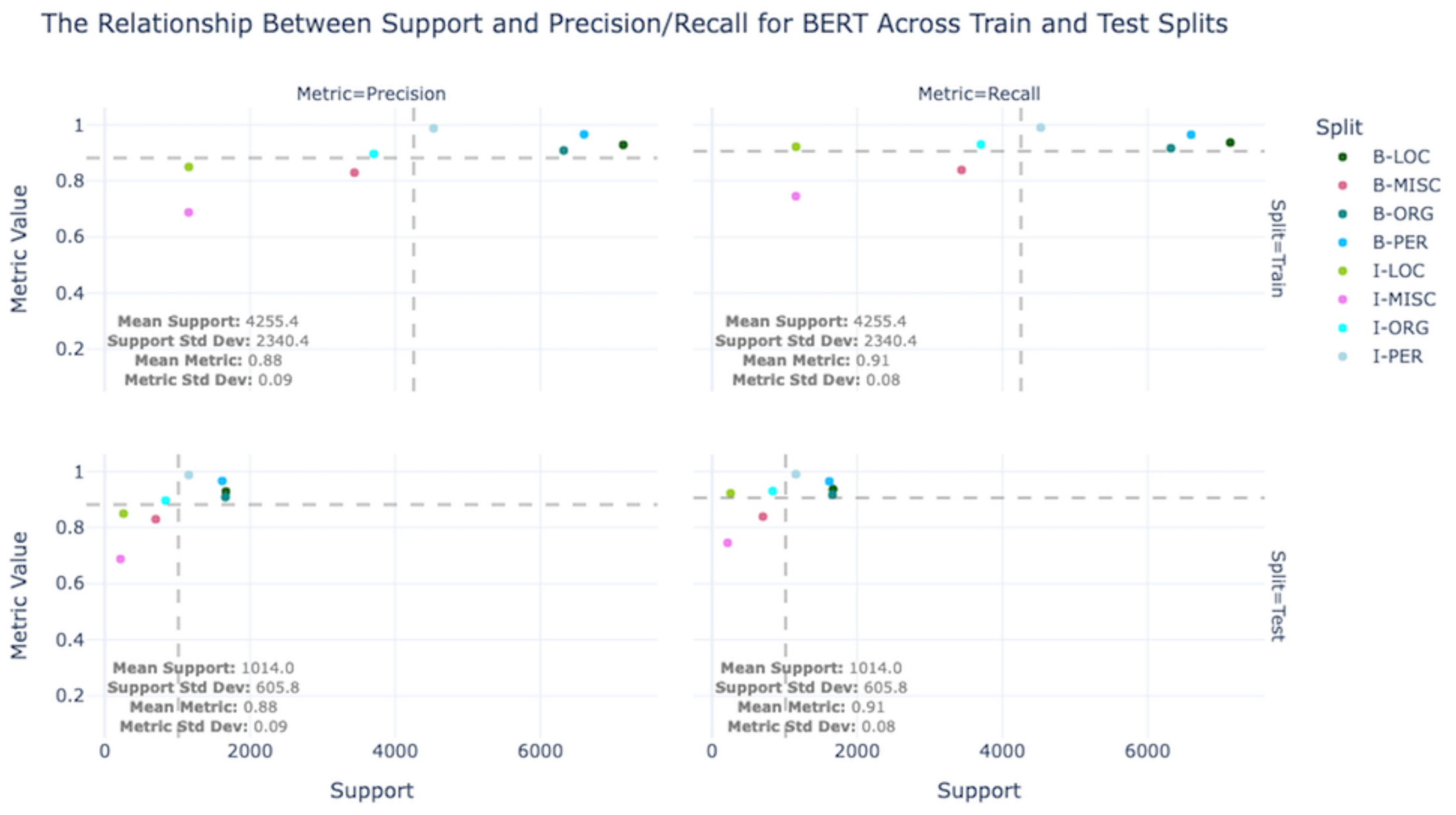

Figure 4.48: Scatter plots illustrating the relationship between support and performance metrics for BERT model across train and test splits. Dashed lines represent the mean values.

In testing, the variation in support is smaller, and precision values remain consistently high across most tags. This more compact distribution improves the proportion-ality of the relationship, making the trend appear more linear. However, certain tags, such as `I-LOC` and `I-ORG`, still introduce minor negative deviations. Despite these outliers, the overall precision correlation increases from training to testing, driven by the reduced variance in support.

Recall shows a different pattern for BERT. In training, recall values vary substantially and do not align closely with support—tags such as `I-LOC` and `I-ORG` even show negative associations, lowering the overall correlation. In testing, these variations are smaller, and recall correlation improves slightly. However, compared to AraBERTv02, the relationship remains weaker, indicating that BERT's recall is less sensitive to tag frequency.

Table 4.23 summarises the underlying statistics behind BERT's correlation scores. Precision shows moderately strong and stable correlations in both splits, reflecting clearer alignment with support. Recall improves from training to testing, but overall remains less correlated due to inconsistent trends across entity tags.

Table 4.23: Breakdown of covariance, standard deviation product, and Pearson correlation between support and performance metrics (precision and recall) for BERT.

| **Split** | **Covariance** | | **Std Dev Product** | | **Pearson Correlation** | |
|---|---|---|---|---|---|---|
| | **Precision** | **Recall** | **Precision** | **Recall** | **Precision** | **Recall** |
| Train | 164.68 | 108.84 | 221.49 | 183.04 | 0.74 | 0.59 |
| Test | 43.80 | 29.60 | 57.33 | 47.38 | 0.76 | 0.62 |

In summary, BERT's Pearson correlations show that precision is more consistent than recall, while overall performance is less closely tied to tag support compared to AraBERTv02. Variance in support distribution plays an important role: in the training split, very high or very low support values lead to inconsistencies in how performance aligns with frequency, weakening the correlation.

While both models exhibit positive correlations between support and performance, these patterns should be interpreted with caution. A high Pearson correlation indicates directional alignment and proportional scaling, but it does not necessarily reflect model quality or generalisability. For AraBERTv02, the tighter support distribution in the test set increases precision correlation, even though certain tags deviate from the overall trend. For BERT, recall shows less predictable variation with support, which weakens the correlation. These results suggest that correlation analysis is valuable for revealing broad trends but can oversimplify the complex relationships at the tag level. This highlights the importance of supplementing correlation analysis with fine-grained error and prediction outcomes analysis to fully understand model behaviour.

**Spearman Correlation** While Pearson correlation highlights linear alignment and proportionality, it can be sensitive to outliers and imbalanced distributions. To complement this, we also examine Spearman correlation, which captures monotonic relationships based on rank order rather than absolute values. By focusing on how support ranks align with performance ranks, Spearman can mitigate some of the variance-driven effects observed in Pearson, offering a more distribution-agnostic view of the relationship.

Figure 4.46 (bottom row) shows that Spearman correlation trends differ from Pearson. AraBERTv02 exhibits a higher rank alignment between support and recall in the training split, while BERT shows stronger precision rank alignment. In the test split, AraBERTv02's recall rank alignment weakens considerably, whereas BERT's precision alignment also declines, highlighting different sensitivities to changes in data distribution. To better understand what drives these differences, we analyse squared rank differences (SRD) for each entity tag. SRD quantifies how much the rank order of support diverges from the rank order of precision or recall. Larger SRD values indicate greater misalignment, contributing to lower Spearman correlations.

Figure 4.49 presents the squared rank differences (SRDs) for AraBERTv02, showing how support ranks diverge from precision and recall ranks. In both train and test splits, most tags show minimal misalignment (SRD = 0–1), but a few contribute disproportionately. In the train split, recall is most misaligned by `I-LOC` (SRD = 9), while precision is most affected by `I-ORG` (SRD = 9) and several tags with SRD = 4. In the test split, recall misalignment remains high for `I-LOC` and increases slightly for `B-PER`. Precision misalignment continues to be led by `I-ORG`, with smaller but consistent SRDs from tags like `B-LOC` and `I-PER`. Overall, misalignment is stronger in recall than precision, explaining the drop in Spearman correlation for recall across splits.

Figure 4.50 shows the SRDs for BERT. While most tags fall within SRD = 0–1, several exhibit large rank misalignments, particularly for recall. In the training split, re-call misalignment is dominated by B-ORG and I-PER (both SRD = 9), while precision misalignment is led by I-PER. In the test split, B-ORG (SRD = 16) and I-PER once again show the largest recall misalignment, with precision misalignment still highest for I-PER. These rank shifts weaken BERT's Spearman correlation for recall more than precision and explain why recall correlations are less stable across splits.

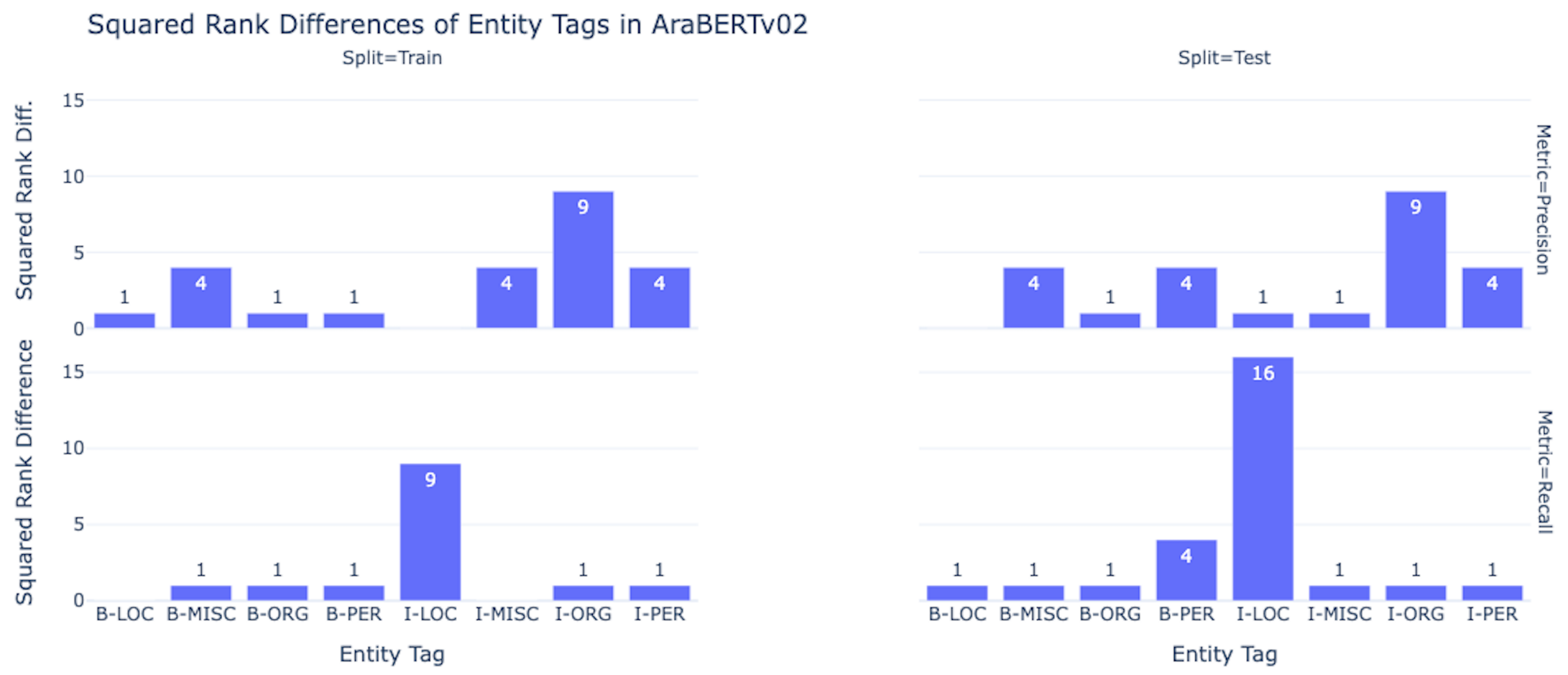

Figure 4.49: Squared rank differences (SRDs) for AraBERTv02 across train and test splits, showing how support ranks diverge from precision and recall ranks.

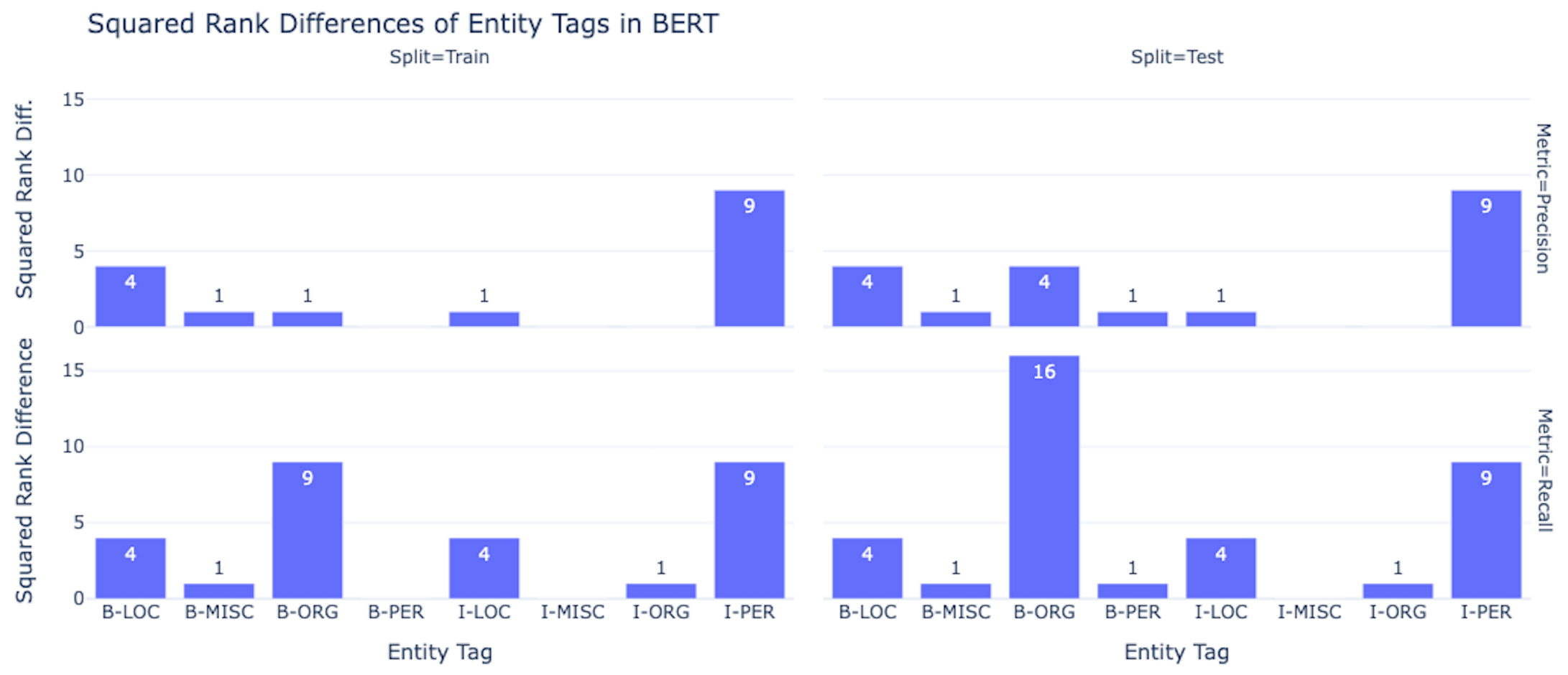

Figure 4.50: Squared rank differences (SRDs) for BERT across train and test splits, illustrating how support ranks diverge from precision and recall ranks.

In summary, while both models show stronger rank alignment for precision than recall, AraBERTv02's misalignments remain more stable across splits, whereas BERT exhibits sharper rank shifts—especially in recall. This supports the earlier conclusion that Spearman correlation, while more robust to scale differences than Pearson, highlights specific tag-level inconsistencies. At the same time, its rank-based design provides an abstract view: it identifies misalignment but does not directly reveal the underlying sources of error. This reinforces the need to supplement correlation analysis with detailed error and prediction outcome analyses to fully understand model behaviour.

### 4.5.7 Answer to RQ3: Evaluation-Component

This section addressed RQ3, examining performance differences between Arabic and English NER. We evaluated AraBERTv02 and BERT and found that BERT consistently outperformed AraBERTv02 at both the token and entity levels. This performance gap cannot be explained by a single factor but instead reflects a combination of evaluation-related influences. We systematically examined these factors across five key aspects: alignment between true and predicted labels, language-specific strengths and weaknesses, the nature and distribution of errors, the impact of annotation schemes and evaluation modes, and the relationship between label frequency and model performance. The following paragraphs summarise the key findings for each aspect.

**Aspect a) Alignment Between True and Predicted Labels** Across both token and entity-level evaluations, BERT shows higher recall but slightly lower precision compared to AraBERTv02, which exhibits the opposite trend. This difference is more noticeable when using macro-averaging, especially for less frequent entity tags. The size and distribution of the test set also influence the observed gap, as smaller representation of some entities can amplify the impact of errors on the overall scores. Notably, the precision–recall gap increases for lower-frequency entity tags across both evaluation levels, while higher-frequency tags like PER and LOC show more similar performance between the models. This increases the sensitivity of precision and recall metrics, making performance appear less stable—particularly in AraBERTv02. It is important to note that while both models are trained and evaluated within their respective language datasets, the comparison here does not aim to evaluate the languages themselves. Instead, it highlights important factors to consider when evaluating NER models in multilingual settings, particularly the influence of class imbalance and evaluation strategies that need to be carefully considered when assessing NER performance.

**Aspect b) Language-Specific Strengths and Weaknesses** Prediction outcomes analysis shows that BERT handles all entity types more evenly, with lower variation in er-ror rates across tags. This suggests more stable and more generalisable performance. AraBERTv02, on the other hand, performs well on high-frequency entities like LOC and PER but struggles with lower-frequency tags such as MISC and ORG. These differences highlight AraBERTv02's greater sensitivity to class imbalance, while BERT's stronger performance on less frequent tags suggests better coverage across entity types.

**Aspect c) Nature and Distribution of Errors** AraBERTv02 shows a higher tendency toward exclusion errors and boundary errors, especially for the PER entity. These errors often result from confusion between entity spans and the O tag, suggesting possible overfitting to non-entity tokens. This aligns with its precision-focused behaviour, where the model avoids predicting entities unless very confident. In contrast, BERT's errors are mainly entity-type errors, particularly confusion between LOC and ORG, with a strong influence from inclusion errors—especially involving the MISC tag. This reflects BERT's recall-focused strategy, where it predicts more entities, sometimes at the cost of type accuracy. These differences are consistently observed across both token- and entity-level analyses and highlight how each model handles classification and generalisation differently.

**Aspect d) Influence of Annotation Schemes and Evaluation Mode** The impact of annotation schemes operates at two levels: how the scorer handles label sequences, and how well predictions align with the scheme used in the dataset. From the scorer perspective, we showed that Seqeval's repair and discard mechanisms affect how predictions are interpreted under IOB1 (Non-Strict) and IOB2 (Strict). For example, in BERT, some entity errors disappear or emerge depending on whether invalid spans are repaired or discarded, revealing that certain model mistakes are either hidden or exposed depending on the scheme.

AraBERTv02 is also affected by these mechanisms, but the more prominent issue was misalignment with the gold standard itself. Unlike BERT, where switching between IOB1 and IOB2 mainly affected how predictions were interpreted by the scorer, AraBERTv02 showed broader shifts in performance—at the aggregate level, per-class level, and in error breakdowns—when evaluated under IOB2. This pointed to inconsistencies in the gold-standard annotations. Specifically, we found that the dataset mixes

IOB1 and IOB2 conventions, with some entities beginning with invalid I-tags. To our knowledge, this issue has not been addressed in the literature. The literature typically focuses on the complexity of the schemes themselves or model's ability to learn them (see Section 4.6). These inconsistencies significantly affect evaluation outcomes—particularly under stricter schemes like IOB2—and highlight the importance of checking annotation consistency in multilingual NER datasets.

In Section 4.3, we further explored how prominent this issue is within the dataset. In Chapter 5, we also investigate AraBERTv02's predictions to examine whether there are consistent patterns indicating the model's difficulty in aligning with the annotation scheme at the prediction level.

**Aspect e) Relationship Between Label Frequency and Performance** Correlation analysis showed a positive relationship between performance and tag frequency, with AraBERTv02 exhibiting stronger correlations—especially between precision and support—compared to BERT. This effect was more evident in the test set, indicating that AraBERTv02's performance reflects the entity frequency patterns in the test data. BERT's correlations were weaker, especially for recall, indicating less consistent dependence on tag frequency. Spearman correlation confirmed these patterns through rank-level alignment, particularly for precision in both models. However, we also highlighted limitations of correlation metrics: they do not explain why certain tags deviate from expected trends and may oversimplify model behaviour. These findings emphasise that while correlation analysis is useful for identifying broad trends, it must be complemented with detailed error and prediction outcome analyses to fully understand NER model performance.

**Conclusion** Together, these findings show that evaluating NER performance requires more than just comparing F1-scores. Factors such as class imbalance, test set composition, annotation consistency, and evaluation strategy all influence how model outputs are interpreted. In the following sections, we situate these findings within the broader literature and discuss their implications for understanding multilingual NER performance.

## 4.6 Related Work

This section reviews prior research related to evaluation methodologies, diagnostic analysis, annotation quality, and annotation scheme literature in NER. We highlight how these efforts inform, but also differ from, the structured and systematic approach adopted in this thesis.

**Evaluation Pipelines and Leader-boards** Standard evaluation pipelines for NER models often rely on leaderboards, where systems are ranked based on aggregated metrics like F1-score. While such leaderboards have driven progress, they typically provide limited transparency into evaluation practices, scoring mechanisms, and model behaviour. In particular, they often rely on a single aggregate number, which, as we demonstrate in this thesis, can be misleading or insufficiently informative—especially when comparing performance across languages.

**Beyond Aggregate Metrics: Diagnostic Evaluation Efforts** To address the limitations of simple aggregate metrics, several efforts have introduced more diagnostic evaluation approaches. ExplainaBoard (Liu et al., 2021) partitions examples into attribute-based buckets (e.g., by entity length or frequency) and analyses performance variations across these buckets. While this provides useful insight into how input characteristics relate to performance, it represents only one aspect of system behaviour. In DeformAr, we go beyond attribute correlation by systematically extracting outputs from multiple components and modelling interactions between data, model, and evaluation mechanisms.

Other approaches, such as the label error detection method by Wang and Mueller (2022), prioritize sentences for manual review based on model confidence scores. Although useful for identifying annotation errors, these methods depend on model calibration and extensive fine-tuning, and focus only on detecting label noise without broader analysis of model behaviour or evaluation reliability.

Finally, the CheckList framework (Ribeiro et al., 2020) introduces structured behavioural testing of NLP models through controlled perturbations designed to assess specific linguistic capabilities, such as robustness to named entity changes or negation.

While CheckList provides valuable insights into model reliability across tasks, it primarily focuses on task-agnostic capability testing rather than structured evaluation of NER systems. Although it incorporates named entity manipulations in its tests, it does not directly evaluate NER. In contrast, DeformAr focuses specifically on structured analysis of NER system behaviour, connecting model predictions, annotation quality, and evaluation processes across multiple languages.

**Annotation Quality** Efforts to address errors and inconsistencies in NER datasets have gained increasing attention in recent years. An earlier initiative by Reiss et al. (2020) applied a semi-supervised method to detect incorrect labels in CoNLL-2003, combining ensembles of NER models with manual review to correct over 1300 errors. Their work highlights the importance of benchmark quality and demonstrates that dataset corrections can significantly affect evaluation scores. More recent efforts, such as Clean-CoNLL (Rücker and Akbik, 2023) and Clean-ANERCorp (AlDuwais et al., 2024), involve larger-scale manual corrections, assisted by automatic techniques like entity linking and model-based flagging of suspicious annotations. While these initiatives improve dataset quality substantially, they rely on extensive manual validation, making them resource-intensive.

At the time of conducting this study, the cleaned version of ANERCorp had only recently been released. Therefore, we chose to work with the original CoNLL-2003 and ANERCorp datasets to maintain comparability with prior benchmarks, and to assess what structured analysis could reveal about annotation consistency without relying on external corrections. Future work could incorporate these cleaned versions to further validate and extend our findings. In contrast to corpus-correction efforts, DeformAr operates directly on original datasets, systematically extracting span-level errors, annotation inconsistencies, and evaluation artifacts to analyse their impact on model behaviour in practical cross-lingual settings.

In terms of span-level error analysis, Ortmann (2022) introduced a taxonomy that distinguishes labelling errors, boundary errors, and combined label-boundary errors, aiming to provide fairer evaluation by avoiding double penalties for near-correct spans. Their work assumes flat span matching and proposes a new scoring algorithm, while DeformAr takes a complementary path: rather than introducing a new scoring standard, we extend the widely-used seqeval evaluation to systematically extract structured span-level prediction outcomes (FP, FN, TP, TN) and error taxonomy, preserving compatibility with standard metrics while enabling detailed diagnostic analysis.

Together, these studies reflect the growing emphasis on moving beyond simple F1-scores toward more nuanced understanding of model behaviour and data quality. DeformAr builds on these insights, but focuses specifically on structured error extraction and multilingual, cross-component analysis without requiring dataset relabelling or re-training.

**Annotation Scheme** Several studies have investigated how annotation schemes influence NER performance from different perspectives. Some research focuses on the complexity and rigidity of annotation schemes. For example, Zhao et al. (2022) propose a dynamic entity recognition framework that avoids rigid boundary constraints, arguing that fixed annotation guidelines can limit model learning, particularly in ambiguous or multi-span scenarios. This highlights the tension between annotation complexity and model adaptability.

Other work focuses on the alignment between annotation schemes and language-specific characteristics. Alshammari and Alanazi (2021) examine how different annotation schemes impact NER system performance across domains and languages, showing that misalignment between annotation guidelines and linguistic structures can degrade accuracy. Similarly, (Luoma et al., 2021) and Tkachenko et al. (2013) explore annotation efforts for morphologically rich languages like Finnish and Estonian, demonstrating how careful adaptation of annotation schemes is necessary to support robust multilingual or cross-lingual evaluation. These studies collectively emphasise that annotation design choices have downstream effects on NER system performance and comparability.

In contrast, DeformAr shifts the focus from training effects to evaluation effects. We investigate how inconsistencies between annotated data and annotation schemes, such as violations of the IOB2 standard in ANERCorp, impact evaluation reliability. To our knowledge, this form of annotation consistency checking has received little attention, particularly in multilingual contexts. Furthermore, we show empirically how scorer behaviour—specifically, discard versus repair mechanisms—can interact with these inconsistencies, changing observed error distributions and model performance metrics.

This highlights the need for more systematic diagnostic frameworks that jointly consider annotation quality and evaluation protocol effects, especially in under-resourced and multilingual settings.

**Scorer Design and Evaluation Consistency** Recent work has highlighted how differences in scorer design and label repair strategies can impact NER evaluation and reproducibility. Lignos and Kamyab (2020) first demonstrated that internal scorers—such as the one in NCRF++—can introduce undocumented repairs for invalid label sequences, resulting in artificially inflated F1 scores compared to external scorers. However, they primarily focused on flagging the issue without providing a systematic solution.

Building on these concerns, Palen-Michel et al. (2021) introduced **SeqScore**, a comprehensive framework and software package to ensure reproducibility in NER evaluation. SeqScore formalises several key guidelines: (i) reporting the chunk encoding scheme, (ii) using an external scorer, (iii) being explicit about repair strategies (begin repair, discard repair, or none), (iv) validating that gold annotations comply with the expected encoding, and (v) reporting statistical distributions of results rather than only best scores. Through empirical experiments, they show that even minor differences in scorer repair mechanisms can cause statistically significant changes in reported F1, especially in low-resource settings.

While these papers emphasize that evaluation inconsistencies can affect reproducibility and comparability across systems, their primary focus remains on the aggregate metric (overall F1), and their experiments aggregate across datasets or settings. In contrast, DeformAr extends this line of work by systematically quantifying and analysing how different scorer repair mechanisms, such as `begin` vs `discard` strategies, impact not just the final metrics but also the underlying error types, span-level distributions, and annotation inconsistencies. DeformAr thus provides a more granular, component-level analysis that supports deeper diagnosis.

**Summary** Across evaluation pipelines, diagnostic analysis, annotation scheme design, and scorer behaviour, prior work has progressively highlighted important weaknesses in standard NER evaluation practices. However, no existing approach systematically provides cross-component analysis, where the system is divided into data, model, and evaluation components, each studied in depth and then integrated to build a more comprehensive profile of performance. DeformAr addresses this gap by combining structured data characteristics, model behaviour, error extraction, annotation quality analysis, and scorer behaviour investigation within a unified multilingual evaluation framework. In the following section, we synthesise these findings into explicit answers to the research questions and highlight the key contributions of this chapter.

## 4.7 Discussion

This section synthesises the findings from the analyses conducted in this chapter. The focus is on how data characteristics, model behaviours, and evaluation outcomes interact to explain performance discrepancies between Arabic and English NER models.

The discussion is organised into three parts. First, we summarise the answers to each research question. This chapter systematically addressed four research questions through a quantitative analysis of data, model, and evaluation components. Here, "quantitative" refers to the use of measurable, component-based metrics and diagnostics designed to capture system behaviour. Each research question was mapped to a specific component of the framework—data, model, and evaluation—with RQ4 synthesising across all three. Within each question, the analysis was divided into aspects, allowing fine-grained answers that reflect different dimensions of system performance. The following subsections provide the consolidated answers to each research question.

Second, we connect these findings to the broader contributing factors outlined in Section 1.2. This mapping highlights which factors are empirically supported by the analysis and what they imply for understanding language-specific challenges and model behaviour.

Finally, we outline how these insights shape the next stage of the thesis. In particular, we identify areas where more fine-grained investigation is required—such as token overlap, tokenisation behaviour, calibration issues, and entity-specific complexities. These are explored in greater depth through interactive visualisation and error inspection in Chapter 5.

In summary, this section addresses three overarching questions:

1. What key findings emerge from answering RQ1–RQ3, and how do cross-component interactions (RQ4) influence performance across languages?

2. Which contributing factors are revealed through the quantitative analysis, and what are their implications?

3. How do these insights guide the subsequent qualitative exploration?

### 4.7.1 RQ1 Summary: Data-Component

RQ1 asked how dataset characteristics help explain differences in Arabic and English NER performance. The detailed results, including structural, lexical, and behavioural analyses, are presented in Section 4.3.5. In summary, the analysis showed that CoNLL-2003 provides models with more favourable learning conditions: it is larger, structurally more balanced, consistently annotated, and dominated by repeated surface forms, which make generalisation easier. ANERCorp, in contrast, is smaller, more uneven across splits, and contains annotation inconsistencies that reduce stability. Lexical differences also play an important role: Arabic exhibits greater type diversity and higher OOV rates, while English benefits from more concentrated frequency distributions. Tokenisation compounds these contrasts—reducing OOV and reinforcing redundancy in English, while in Arabic it preserves diversity and ambiguity, leaving learning more unstable. Together, these findings indicate that Arabic models must cope with a noisier, less redundant environment, whereas English models learn from a simpler, more repetitive signal. This helps explain why data-level characteristics are a critical driver of the observed cross-lingual performance gap.

### 4.7.2 RQ2 Summary: Model-Component

RQ2 asked how model-level behaviours contribute to performance differences between Arabic and English NER. The detailed results, including loss, confidence, and representational analyses, are presented in Section 4.4.5. In summary, the analysis showed that Arabic models tend to be less stable across key behavioural metrics: they exhibit higher loss values, weaker calibration (less confident when correct, more confident when wrong), and less distinct representational separability, particularly for complex tags such as `I-MISC` and `I-ORG`. English models, while not error-free, generally show lower loss, better calibration, and more structured internal clusters. These patterns may potentially reflect the structural and lexical challenges identified in RQ1, suggesting that model behaviour and data properties are closely related. This connection is explored further in Chapter 5, where we link data-level characteristics and model-level behaviours through token-level behavioural metrics, supported by fine-grained behavioural analysis and interpretability methods.

### 4.7.3 RQ3 Summary: Evaluation-Component

RQ3 asked how evaluation-related factors contribute to performance differences between Arabic and English NER. The detailed results, covering precision–recall balance, error taxonomies, and scheme effects, are presented in Section 4.5.7. In summary, the analysis showed that BERT consistently outperformed AraBERTv02 at both token and entity levels, but the nature of this gap differed across models. AraBERTv02 adopted a more conservative strategy, favouring precision over recall, while BERT was more recall-oriented and achieved a better overall balance. This trade-off was especially marked in Arabic, where the wider precision–recall gap made performance appear less stable.

Error analyses revealed systematic contrasts between the models. AraBERTv02 was dominated by exclusion and boundary errors, especially in PER spans, reflecting its tendency to miss entities unless highly confident. BERT, in contrast, was dominated by span type confusions and inclusion errors, particularly in LOC, ORG, and MISC, reflecting its recall-driven strategy. Annotation scheme experiments showed that AraBERTv02 was more sensitive to scoring settings (IOB1 vs. IOB2), largely due to inconsistencies in the Arabic gold standard, while BERT's scores were comparatively stable and shifted mainly due to different interpretations of predictions. Correlation analysis confirmed that frequency patterns appear related to performance, but also highlighted that some tags diverge from frequency-driven expectations, and that distributional differences can strongly influence correlation values—making them appear higher or lower than their true explanatory value. This underscores the limits of correlation as a diagnostic tool.

Together, these findings demonstrate that the performance gap between BERT and AraBERTv02 arises from complex interactions between model biases and data-related factors. BERT's higher performance reflects its more stable and generalisable behaviour: a recall-oriented strategy that handles entity types more evenly, even if it introduces more type confusions and inclusion errors. AraBERTv02's weaker performance, by contrast, is closely tied to its sensitivity to data quality issues—particularly class imbalance and annotation inconsistencies—which amplify its precision-oriented bias and lead to more false negatives and boundary errors. These patterns highlight that evaluation outcomes cannot be explained by data or model factors in isolation; rather, they emerge from their interaction. This reinforces why aggregate F1-scores are insufficient: the underlying discrepancies become more clear when precision–recall trade-offs, error distributions, and scheme effects are examined directly. In the next subsection, we provide a cross-component analysis (RQ4), integrating data, model, and evaluation components to explain language performance discrepancies.

### 4.7.4 RQ4 Summary: Cross-Component Analysis

RQ4 asked how data characteristics, model behaviours, and evaluation outcomes interact to shape performance discrepancies across languages. Whereas RQ1–RQ3 analysed each component in isolation, RQ4 synthesises across them to explain why some entity types are consistently easier or harder to model. We focus on two spans—LOC and MISC—as illustrative cases, representing one of the most stable and one of the most challenging categories, respectively.

**LOC Spans** For LOC spans (`B-LOC`, `I-LOC`), the performance gap between Arabic and English is relatively small. Both models achieve strong F1, with the most common error being confusion with ORG, reflecting the semantic proximity of these categories.

Data-level factors help explain this stability. LOC entities often occur as single-token spans and are dominated by frequent surface forms, especially in English. This setup likely simplifies the learning task: the model sees many repeated surface forms, reducing the complexity of generalisation. In both languages, however, overlap between LOC and ORG tokens introduces ambiguity, explaining why entity-type confusions between LOC and ORG are the most frequent error type observed for this category.

At the model level, LOC tokens are generally easier to learn than other entity types.

Across both languages, they show lower loss values and higher confidence in correct predictions, though Arabic I-LOC tokens register slightly higher loss than English, reflecting greater uncertainty tied to OOV rates and annotation inconsistencies in ANERCorp. Calibration patterns also diverge: when errors occur, the Arabic model tends to be uncertain, assigning lower confidence to incorrect LOC predictions, while the English model often makes overconfident mistakes. Representation analysis further supports this contrast. LOC spans form the most compact and well-separated clusters among all entity types in both languages, highlighting their relative stability, although some confusion with ORG remains, likely due to the semantic closeness of the two categories rather than poor embedding separability.

Overall, LOC spans show how data and model factors reinforce each other to produce stable performance in both languages. Their short span length and frequent repetition reduce data complexity, making them easier to learn and producing high performance across languages.

**MISC Spans** For MISC spans (B-MISC, I-MISC), performance gaps are among the largest observed. English maintains moderate scores, while Arabic performs substantially worse, particularly for I-MISC tokens where recall is very low.

Error patterns for MISC spans are more heterogeneous compared to LOC. AraBERTv02 is dominated by exclusion errors, where MISC entities are frequently missed or defaulted to O. This reflects its precision-oriented behaviour, suggesting an over-reliance on the non-entity class and difficulty representing rare, diverse MISC examples. In contrast, BERT is dominated by inclusion errors, where non-MISC tokens are incorrectly predicted as B-MISC or I-MISC. This aligns with its recall-oriented behaviour, where the model over-predicts O tokens as entities.

Data characteristics amplify these challenges. MISC is the rarest entity type, with especially sparse representation in Arabic, making performance highly sensitive to individual errors. Arabic MISC spans are also structurally longer and more diverse, with higher OOV rates and greater lexical overlap with O. Unlike English, where a few frequent MISC types dominate, Arabic MISC is more evenly distributed across many rare types, leaving the model with fewer redundant examples to learn from. This flatter distribution, combined with annotation inconsistencies—particularly in I-tokens—further introduce instability during training and contributes to the model's difficulty in generalising.

Turning to model behaviour, MISC spans show higher loss and lower confidence than other entity types, with Arabic `I-MISC` tokens showing the highest loss values overall. In English, the effect is less severe, though MISC spans remain harder than other categories. Calibration patterns also reveal important differences. In Arabic, MISC errors are often made with high confidence, showing poor calibration, while even correctly predicted `I-MISC` tokens are marked by greater uncertainty than in English. This points to unstable internal representations in Arabic, a pattern reinforced by representation space analysis. MISC tokens form the least compact and least separable clusters in both languages, making them especially prone to confusion across entity boundaries and aligning with the heterogeneous misclassification patterns observed in evaluation.

Overall, MISC spans illustrate how high sparsity, lexical variability, overlap, and inconsistent annotations compound with model calibration weaknesses, shows why MISC is challenging and how this challenge differs across language.

In this section, we illustrated how a layered and fine-grained entity-based analysis can systematically build an empirical narrative of the interactions across evaluation, data, and model components. By focusing on two entity types—LOC and MISC—we demonstrated how different sources of difficulty emerge and how they interact across languages. This approach not only provides evidence-based insights into the factors contributing to the performance discrepancies discussed in Section 1.2, but also guides the direction of further analysis outlined in Section 4.7.6. In particular, it highlights the need for fine grained behavioural analysis, directing the deeper, instance-level investigations in Chapter 5.

### 4.7.5 Insights into Performance Discrepancies: An Empirical Study of Contributing Factors

This section discusses the main contributing factors outlined in Section 1.2 and assesses how the cross-component analysis has helped reveal their effects. As introduced earlier, these factors are grouped into three areas: (1) model architecture—specifically tokenisation, (2) language complexity, and (3) training procedure. We review each in turn, highlighting which aspects are empirically supported by the findings of this chapter.

**Model Architecture (Tokenisation Effects):** Although we do not directly compare different model architectures or pretraining corpora, we examine tokenisation as an important architectural component. The analysis reveals that the tokenisation approach significantly shapes the lexical structure of the data. We observe that English tokenisation is highly fertile, producing more subword units and increasing the number of examples per type, especially for complex spans like MISC. In contrast, Arabic tokenisation is less fertile, sometimes increasing ambiguity or reducing the benefit of subword splits, depending on entity type. These tokenisation patterns interact with language morphology, increasing differences in lexical diversity, type distribution, and token overlap across languages.

Overall, this chapter provides empirical evidence that tokenisation changes the lexical structure differently across languages, influencing the frequency of observed patterns and affecting how easily the model can learn generalisable representations.

**Language Complexity:** The findings also reflect how the linguistic properties impact model behaviour. Arabic shows higher OOV rates, more even type distributions, and greater lexical diversity for complex spans like MISC. These properties, is potentially associated with Arabic's morphological richness and the data sources. English, while also facing challenges (e.g., higher ambiguity due to tokenisation), benefits from more skewed distributions and repeated surface forms, particularly in easier spans like LOC. This shows how language-specific complexity can interact with tokenisation and dataset construction to create uneven modelling challenges.

**Training Procedure:** Training procedure factors are revealed across three dimensions: annotation scheme, evaluation strategy, and annotation quality.

**Annotation Scheme:** The annotation scheme impacts performance differently across languages, particularly in how models handle boundary errors (notably between `B-` and `I-` tokens). We also observe label inconsistencies in the Arabic gold-standard annotations, especially for `I-MISC` tokens—an issue that, to our knowledge, has not been discussed in prior work (see Section 4.6).

**Evaluation Strategy:** We demonstrate that averaging methods and scoring approaches (e.g., micro-averaged vs. entity-specific F1) can substantially alter performance comparisons. This is particularly impactful for rare entities like MISC, where different aggregation choices can lead to divergent interpretations. These findings highlight the importance of transparent evaluation reporting in leader boards, which often rely on average F1 scores without entity-level breakdowns.

**Annotation Quality and Noise:** High label inconsistency, sparse data coverage, and ambiguity—especially for complex spans like MISC—adversely affect model calibration, increase prediction uncertainty, and weaken representation separability. In contrast, the stronger stability observed for LOC spans, which have simpler and more consistent annotations, reinforces the importance of annotation quality (see Section 4.6). Although annotation issues are present in both languages, they are more pronounced in Arabic, where inconsistencies and sparsity has more significant impact across all components of the framework—data, model, and evaluation.

After examining the main contributing factors and how they influence data characteristics, model behaviour, and evaluation outcomes, we now turn to the next step. The insights from this cross-component analysis not only help explain the observed discrepancies but also provide guidance for the behavioural analysis phase, where these patterns are explored in greater depth. This transition is developed in the next section.

### 4.7.6 Bridging Cross-Component and Behavioural Analysis: Motivations and Next Steps

The cross-component analysis revealed important patterns in model behaviour, data characteristics, and evaluation outcomes across languages. However, several open questions remain that cannot be resolved by aggregate metrics alone. This section outlines the main focus areas that emerged from the cross-component findings and motivates the transition to behavioural analysis in the next chapter.

As demonstrated in Section 4.7.4, data, model, and evaluation components often interact in complex ways to influence performance. Some metrics capture standalone behaviours (e.g., loss reflecting classification difficulty), while others capture interactions (e.g., token overlap shaped by data properties, tokenisation, or annotation inconsisten-cies). Moreover, these components influence each other: for example, loss may rise because of overlap, while overlap itself may be driven simultaneously by linguistic diversity, tokenisation effects, or inconsistent labels. Untangling such interactions requires closer inspection than aggregate results allow.

The cross-component analysis has also helped establish a more focused direction for the next stage. By identifying the most problematic areas—such as tokenisation rate, annotation scheme misalignments, and cross-linguistic variation in model behaviour—we now have empirical evidence to support a targeted, fine-grained investigation. Rather than approaching the data based on assumptions, the behavioural analysis phase is guided directly by these empirical insights.

Through interactive visualisation and token-level metrics, the behavioural analysis approach will reinforce earlier explanations and provide deeper insight into the sources of discrepancy. By linking model predictions, token properties, and internal representations, we aim to uncover underlying causes of instability across languages.

Specifically, three hypotheses emerge from the cross-component analysis that will be tested in the next chapter:

1. **Tokenisation distortion:** In Arabic, tokenisation introduces structural shifts that alter type distributions and increase ambiguity, without providing the redundancy benefits observed in English.

2. **Scheme misalignment:** Misalignments between gold and predicted IOB structures—especially in Arabic I-tokens—create unstable learning dynamics and boundary errors.

3. **Language-specific error taxonomy:** Arabic exhibits distinct error profiles (e.g., exclusion errors and boundary instability) that are not mirrored in English to the same extent, and therefore require closer inspection.

## 4.8 Contributions

This section summarises the main contributions of the chapter. Building on the cross-component analysis developed in this work, the chapter extends existing NER evalua-tion and diagnostic practices and introduces new tools and insights for analysing cross-linguistic model behaviour. The key contributions are outlined below.

**First: Fine-grained empirical comparison between Arabic and English NER systems.**

Unlike previous studies that mainly compared final performance scores, this work systematically quantifies differences across data properties, model behaviours, and internal representations.

**Second: Introduction of new evaluation utilities.**

- A span-based confusion matrix to extract errors at the span level (boundary errors, entity-type confusion, false positives/negatives).
- A structured error types categorising span-based errors into entity-type errors, boundary errors, exclusion errors, and inclusion errors.

**Third: Empirical analysis of evaluation scoring settings.**

We demonstrate how different evaluation modes in `seqeval` (strict vs. default) impact model performance and change the distribution of errors at the span level.

**Fourth: Empirical analysis of aggregation effects.**

We show how aggregation strategies (micro vs. macro averaging) interact with dataset size and label distribution, affecting the interpretation of language comparisons and hiding underlying performance gaps.

**Fifth: Identification of a previously undocumented annotation issue in Arabic datasets.**

We highlight misalignment between gold-standard annotations and the IOB2 tagging scheme in ANERCorp, a problem not previously discussed, which can impact performance and introduce misleading evaluation outcomes.

**Sixth: Introduction of new token-level behavioural metrics.**

- Token confidence and uncertainty.
- Per-token loss.

- Representation separability (using silhouette scores).
- Token overlap, ambiguity, and inconsistency.

These metrics connect dataset properties to model behaviours, helping to explain the interaction between different components. Note that while some of these individual metrics are well-known, their combined use to systematically analyse cross-component interactions is, to our knowledge, novel.

**Seventh: Demonstrating the impact of tokenisation on lexical structure.**

We show how tokenisation affects type distribution, ambiguity, and token overlap differently across languages, impacting model generalisation and error patterns.

**Eighth: Quantifying performance discrepancies through cross-component analysis.**

We move beyond reporting simple F1 gaps, showing how factors like data structure, tokenisation, annotation noise, model calibration, and representation quality interact to create performance differences.

**Ninth: Highlighting limitations of correlation-based analysis.**

We show that relying only on correlations between data features and performance can hide complex underlying patterns, reinforcing the need for fine-grained qualitative inspection.

Together, these contributions not only strengthen the quantitative analysis of NER systems but also provide the foundation for the next chapter. Building on these findings, we shift to a fine-grained qualitative investigation to deepen our understanding of model errors, data challenges, and interpretability across languages.

# Behavioural Analysis 5

## 5.1 Introduction

This chapter continues the analysis by building on the cross-component evaluation presented in Chapter 4. While the cross-component analysis quantified discrepancies between Arabic and English NER models and revealed where they emerged, it primarily examined components in isolation using aggregate metrics. It did not directly explore how dataset properties and model behaviours interact at the token or instance level. As established in Chapter 3, this analysis continues the comparative case study design, applying DeformAr to Arabic and English NER to examine how behavioural differences emerge across languages and to provide deeper diagnostic explanation of system behaviour.

This chapter adopts a behavioural analysis approach to address this issue. It combines interpretability techniques, token-level behavioural metrics, and instance-level analysis to interactively connect model behaviours to dataset characteristics at the token level. The analysis begins with global interpretability techniques that reveal broad patterns in the representational space and gradually dives into token-level metrics and instance-level analyses to explain how these patterns occur. By integrating these complementary perspectives, the chapter aims to deepen our understanding of model behaviour and performance discrepancies across languages.

This chapter has two goals. **First**, it introduces a multi-layered exploration pipeline that links global representations, local attention patterns, and instance-based analyses to the token-level metrics introduced earlier. This integrated approach allows us to identify high-level patterns, interpret them through numerical metrics, and examine individual examples to provide deeper explanations of model behaviour in each language. **Second**, the chapter connects the behavioural analysis findings back to the broader contributing factors outlined in Chapter 1, such as tokenisation issues, language complexity, dataset design, and model biases. Together, this approach addresses Research Question 5 (RQ5) while also complementing and enriching the findings related to RQ1–RQ4.

**Chapter Outline** Following the approach taken in the previous chapter, we divide RQ5 into three main sections. In each section, we define key aspects that structure the behavioural analysis investigation and guide our analysis. Each aspect is introduced at the beginning of its corresponding section to specify the questions and focus areas addressed. The three sections are: **Training Impact**, **Patterns and Anomalies in Global Representation Space**, and **Clustering Alignment with Annotation Structures**. The remainder of this chapter is organised as follows:

- **Section 5.2 — Technical Background**
  Introduces the interpretability techniques, visualisation methods, and token-level behavioural metrics that support the behavioural analysis approach.
- **Section 5.3 — User Interface**
  Provides an overview of the interactive tabs developed for the behavioural analysis phase and explains how it is used.
- **Section 5.4 — Training Impact**
  Investigates how fine-tuning changes internal model parameters, particularly attention weights and patterns, and examines differences across languages. We focus on: (a) how model parameters, specifically attention weights and patterns, change after fine-tuning in each language; (b) what differences in internal model behaviour emerge across languages after fine-tuning, and how they motivate the need for language-specific interpretability analysis.
- **Section 5.5 — Patterns and Anomalies**
  Explores global structures in the model's learned representations, combining token-level metrics and instance-level analysis. We focus on: (a) what patterns and anomalies emerge in the representation space post fine-tuning; (b) how these patterns relate to model errors; (c) how token-level metrics (e.g., loss, confidence, silhouette) help explain representation anomalies. We also address three Arabic-specific questions: (d) what Arabic-specific error patterns can be identified, and how they affect representation structure, model confidence, and prediction consistency? (e) despite English showing higher tokenisation rate and ambiguity, why does Arabic perform worse — and what does the tokenisation impact look like in Arabic specifically? (f) how do annotation scheme misalignments emerge in Arabic, and how do they differ across gold-standard and predicted labels?

- **Section 5.6 — Clustering Alignment with Annotation Structure**
  Examines how clustering algorithms applied to learned representations align with annotation structures. We focus on: (a) what patterns emerge from clustering, and to what extent they reflect entity tags or span patterns; (b) how clustering alignment differs across languages and how it can be measured.

- **Section 5.7 — Related Work**
  Reviews related approaches and existing literature on interpretability techniques and behavioural analysis, providing context for this chapter.

- **Section 5.8 — Discussion**
  Summarises the answers to Research Question 5 (RQ5), synthesises the key findings from the behavioural analysis, and reflects on their broader implications. before integrating them into the overall chapter conclusions.

- **Section 5.9 — Contributions**
  Summarises the main contributions of the behavioural analysis phase.

## 5.2 Technical Background

This section provides an overview of the technical concepts, techniques, and analytical approaches that underpin the behavioural analysis phase. It establishes the background necessary to understand the approach adopted to address RQ5, supported by the interactive tabs described in Section 5.3.

As discussed in Section 3.4.3, the behavioural analysis phase is implemented through two interactive tabs: the **Behavioural Analysis Tab** and the **Instance-Level Tab**. The Behavioural Analysis Tab focuses on global model behaviour, enabling exploration of representation patterns and token-level metrics, while the Instance-Level Tab supports detailed examination of specific sentences, entities, and prediction contexts. Together, these tabs facilitate the interpretability approach introduced in this chapter. Further details of these analytical tabs are provided in Section 5.3.

We begin by briefly introducing the interpretability techniques adopted in this chapter. Following this overview, we organise the background according to the main sections outlined in the chapter: training impact, representation patterns, and clustering alignment. Within each section, we define the key concepts, methods, and metrics that provide the foundation for addressing RQ5.

- The subcomponents analysed (e.g., attention weights, hidden representations, prediction confidence);
- The outputs extracted during the data extraction phase (e.g., attention weight distributions, loss values, 2D UMAP);
- The methods used to analyse and visualise these outputs (e.g., interactive visualisations, token-level metric comparisons, instance-level analysis).

### 5.2.1 Interpretability and Interactive Visualisation

As discussed in Chapter 3, we adopt a combination of interpretability techniques supported by interactive visualisation to carry out the behavioural analysis approach. These techniques include global representation space analysis using UMAP projections, local attention score analysis, and a modified instance-based analysis derived from example-based explanation methods. All techniques are implemented and integrated within the interactive tabs introduced in Section 5.3.

In addition to these interpretability methods, we utilise a range of visual analytics techniques, including scatter plots, heatmaps, and bar charts. These visualisations are designed to dynamically interact with one another based on user selections, enabling a layered and responsive analysis. For example, selecting a region in a scatter plot can highlight corresponding patterns in other visualisations, allowing users to explore relationships across multiple levels of model behaviour and dataset characteristics simultaneously.

Together, these interpretability techniques and interactive visualisations provide the basis for the interactive tabs used throughout the behavioural analysis phase.

### 5.2.2 Attention Weights and Attention Scores

In this section, we analyse how the internal attention mechanisms change before and after fine-tuning by examining two key outputs: attention scores and attention weights. As discussed in Chapter 2, attention scores are calculated based on the interaction between token embeddings and attention weights, capturing how strongly each token attends to others in the sequence. Figure 5.1 presents an illustrative example of such attention patterns.

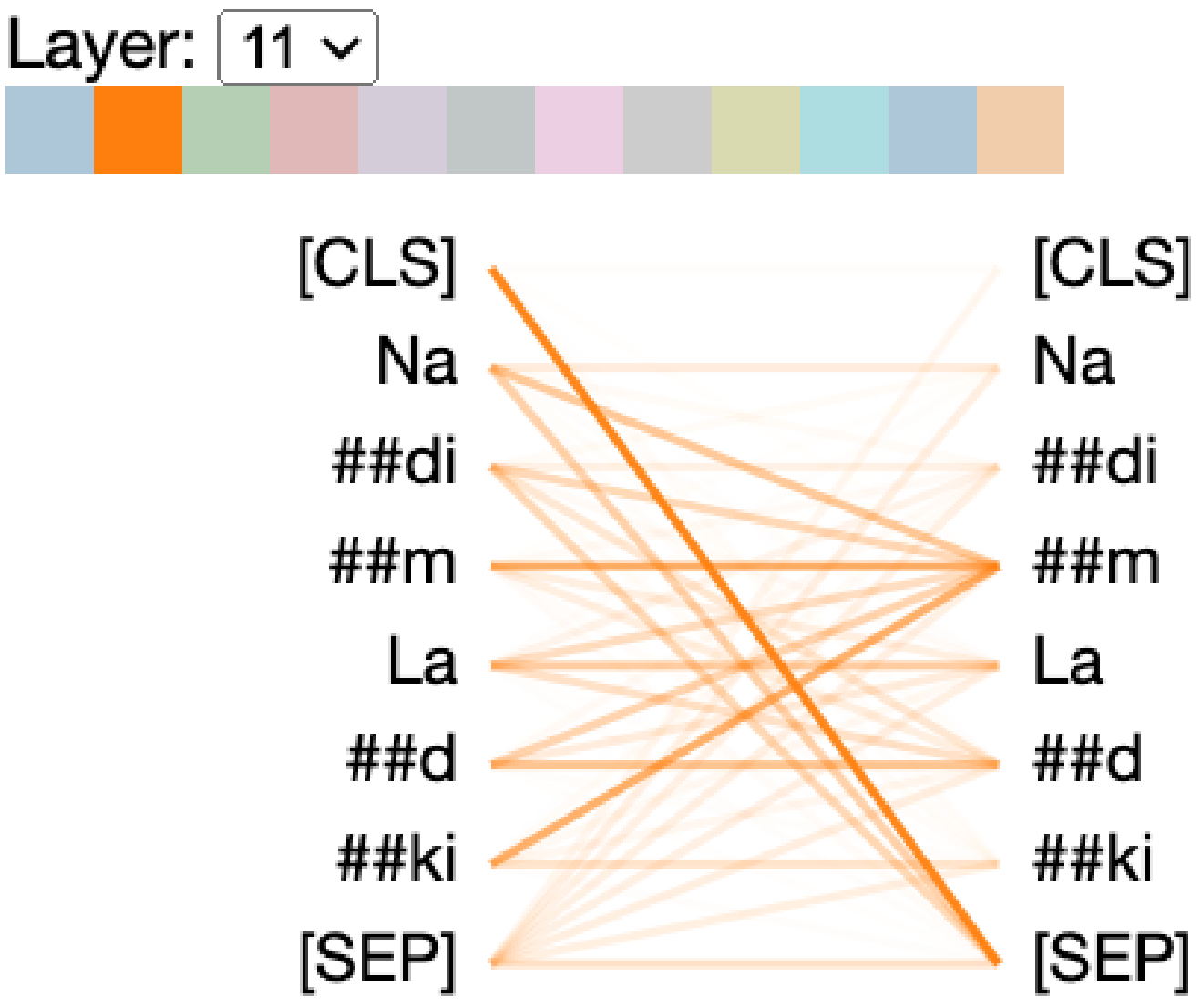

Figure 5.1: Example of an attention pattern showing how tokens attend to each other within a sequence, using Sentence 2 from the CoNLL test set.

To support the training impact analysis, we extract the following outputs during the data extraction phase:

- **Attention Score Similarity Heatmap:** Displays the cosine similarity between attention score matrices generated by the pretrained and fine-tuned models, com-puted for each attention head across all layers and averaged over the full dataset. Specifically, for each instance in the test split, we extract the attention score matrices from both the pretrained and fine-tuned models. For each attention head, we flatten the attention matrix into a vector, and compute the cosine similarity between the corresponding vectors.

- **Attention Weight Similarity Heatmap:** Shows the cosine similarity between the learned attention parameters ($W_Q$, $W_K$, and $W_V$), concatenated and compared between the pretrained and fine-tuned models. In this case, for each head, the learned weight matrices (of size $64 \times 64$) are flattened into vectors, and the cosine similarity is computed between the corresponding pretrained and fine-tuned vectors, providing a direct view of parameter updates.

These outputs allow us to assess how the attention patterns and underlying parameters evolve through the fine-tuning process across different languages.

### 5.2.3 Patterns and Anomalies

In this section, we investigate patterns and anomalies in the global representation space by analysing the last hidden states of tokens after fine-tuning. Each token's representation is extracted and projected into two dimensions using UMAP, allowing us to visualise the global structure through scatter plots.

Alongside the hidden states, we extract a range of token-level behavioural metrics introduced in Chapter 4, including (1) Loss, (2) Prediction metrics (Token Confidence, Token Uncertainty), (3) Token ambiguity, (4) Annotation inconsistency, (5) Tokenisation rate, and (6) Representation separability (True Silhouette, Predicted Silhouette).

The tabs enable interactive exploration of these scatter plots, allowing users to visually identify patterns and anomalies. We also examine correlations between the behavioural metrics themselves to understand how different token properties interact within the observed patterns. For example, a strong correlation between token ambiguity and annotation inconsistency within a specific cluster can empirically explain the behaviour of the tokens in that pattern. It is important to note that these correlations are used as indicators of relationships, not as evidence of causality (based on our findings in Section 4.5.6).

In addition to acting as explanatory indicators, the behavioural metrics are also used as interactive filters within the tabs. By selecting specific ranges of metric values (e.g., high ambiguity, high inconsistency), users can isolate subsets of tokens and observe how these properties correspond to patterns in the global representation space. As shown in Section 5.5.3, these filters allow us to explore specific hypotheses raised during the cross-component analysis. For example, by focusing on tokens with a high tokenisation rate, we can investigate whether the observed ambiguity increases from tokenisation artefacts or from semantic overlap, and how these factors impact model confidence and the structure of representations.

In addition to numerical behavioural metrics, we extract several categorical variables for each token. These variables provide additional structure for analysis and enable interactive filtering and grouping within the tabs. The extracted categorical columns include:

- Annotation variables: True Labels, Predicted Labels, Error Type, Confusion Components, Agreements;
- Entity scheme variables: True Entities, Strict True Entities, True Aligned Scheme, Predicted Entities, Strict Predicted Entities, Predicted Aligned Scheme;
- Clustering variables: K=3 Clusters, K=4 Clusters, K=9 Clusters, Boundary Clusters, Entity Clusters, Token Clusters;
- Vocabulary status: In-Vocabulary or Out-Of-Vocabulary (OOV) token classification.

These categorical variables allow users to explore not only scalar relationships, but also higher-level structural patterns such as cluster membership, annotation inconsistencies, and vocabulary status within the global representation space.

Through this approach, behavioural metrics act as a bridge between model behaviour and dataset characteristics, while interactive visualisation bridges the gap between cross-component analysis and interpretability techniques. This layered exploration enables us to identify global patterns, understand how they are reflected in behavioural metrics, and guide deeper instance-level investigations.

### 5.2.4 Clustering Alignment

In addition to visualising the embedding space, we apply clustering analysis to assess whether the model's learned representations are structured in a way that corresponds to named entity categories. NER annotations typically encode two pieces of information, as described by Zhao et al. (2022): the entity type (e.g., *PER*, *LOC*) and the token's position within the entity span (e.g., *B* for beginning, *I* for inside). Prior studies have shown that models can internally learn this structure, organising representations based on both type and position.

In this section, we examine the extent to which such representational organisation emerges in our models by treating it as a clustering-quality problem. Specifically, we focus on the last hidden states of tokens after fine-tuning as the subcomponent of interest, from which we extract two main outputs:

- The 2D UMAP projections of the token representations;
- The clustering assignments generated by applying KMeans clustering with different numbers of clusters ($k = 3$, $k = 4$, and $k = 9$).

The rationale for choosing these values of $k$ is as follows:

- $k = 3$: Chunk-level grouping, separating entity tokens from non-entity tokens;
- $k = 4$: High-level entity-type grouping, covering major categories such as *PER*, *LOC*, *ORG*;
- $k = 9$: Token entity-tag grouping, capturing both type and position information (e.g., *B-PER*, *I-ORG*).

To evaluate clustering quality, we use three standard extrinsic metrics: homogeneity, completeness, and V-measure. Homogeneity measures whether each cluster contains only tokens from a single class, while completeness checks whether all tokens from a given class are assigned to the same cluster. V-measure is the harmonic mean of these two scores, providing a balanced assessment of cluster-label alignment. All three metrics range from 0 to 1, with higher scores indicating stronger alignment with the gold-standard labels.

By comparing these clustering scores both within and across languages, we examine how well the model's internal representations capture the entity structure. This analysis provides insight into the nature of the groupings formed in the decision boundary and whether underlying representation organisation differs across languages.

Having introduced the technical background and analysis components, we now present the user interface of the interactive tabs developed to support the behavioural analysis phase.

## 5.3 User Interface

The interactive tabs provides the environment for exploring the outputs and metrics extracted during the behavioural analysis. The Behavioural Analysis tab serves as the primary interface for exploration, offering visualisations of the representation space, token-level behavioural metrics, clustering patterns, and training impact outputs. The Instance-Level View tab enables detailed inspection of individual sentences, including model predictions, token-level behaviours, and attention patterns. Together, these two tabs support both global exploration and local, instance-level analysis. The following subsections describe the system layout, key visualisations, and interaction mechanisms.

### 5.3.1 Tab Layout

The interactive tabs follows a modular architecture, consisting of three main modules: the Data Manager, User Controls, and Analytical Visualisations. The Data Manager handles the loading, processing, and management of all preprocessed data. It integrates extracted features into a unified structure and feeds this structured data to the visual components. This data is described in more detail in Section 3.3.7. The User Controls module includes all interactive elements—such as dropdowns, buttons, and selection interfaces—that allow users to filter, adjust, and explore the visualisations dynamically. The Ana-lytical Visualisations module consists of various visual outputs, including scatter plots, heatmaps, and attention maps. These components update based on user selections and are used to display patterns and insights derived from the data. As discussed in the previous section, the behavioural analysis phase is divided into two main tabs.

**The Behavioural Analysis Tab** is the core interface of the behavioural analysis and consists of five main sections. Figure 5.2 presents an architectural diagram of the components within this tab.

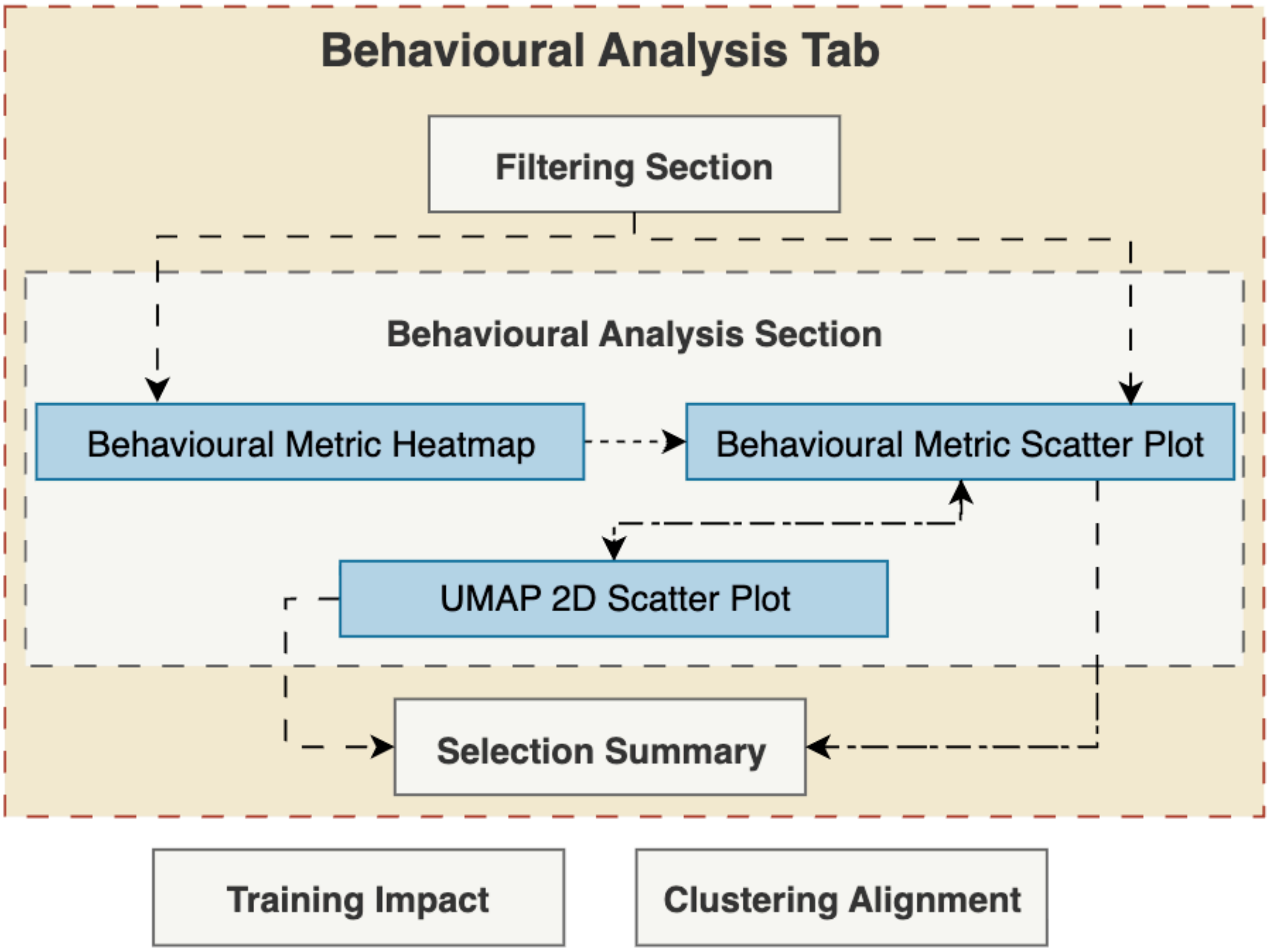

Figure 5.2: Architecture of the Behavioural Analysis tab. The tab is divided into modular sections: a *Filtering Section* (top), a *Behavioural Analysis Section* including the behavioural metric heatmap, scatter plot, and UMAP 2D scatter plot (centre), and a *Selection Summary* (bottom). Training Impact and Clustering Alignment modules are accessible as complementary analyses. Solid arrows indicate data flow, while dashed arrows represent interactive links between components.

The first section is the **Filtering Section**, which allows users to filter and subset the dataset based on various criteria. It consists of dropdowns for filtering based on categorical variables, and a data table that supports filtering based on metadata (such as words and core tokens) as well as behavioural metrics (such as loss, token ambiguity, and others).

The second section is the **Behavioural Analysis Section**, which contains two main components: metric analysis and representation space visualisation. The metric analysis includes a correlation heatmap of all behavioural metrics and an interactive scatter plot that compares any two selected metrics. The heatmap is interactive—selecting a cell (e.g., the correlation between tokenisation rate and token ambiguity) updates the scatter plot to reflect the selected metric pair. The representation space visualisation displays a scatter plot of the 2D UMAP projections of the model's last hidden states. All these components are controlled through dropdowns and buttons, with further details about the interactions provided in Section 5.5.

The third section is the **Selection Summary**, which presents a breakdown of the selected data points—whether chosen from the metric scatter plot or the representation space visualisation. It offers summary statistics for the selected subset of the data.

The fourth section is the **Clustering Alignment Section**, which presents a table of clustering results—including V-measure, homogeneity, and completeness scores—across different values of $K$. It also includes the average centroid similarity matrix, which reflects how closely tokens from each entity class align with the centroids identified by KMeans clustering in the representation space.

Finally, the fifth section is the **Training Impact Section**, which shows how attention behaviour has changed as a result of fine-tuning. This section includes two heatmaps. The first is the Attention Score Similarity heatmap, which measures the similarity between attention matrices produced by the pretrained and fine-tuned models across all heads and layers, averaged over all sentences. The second is the Attention Weight Similarity heatmap, which calculates the cosine similarity between the concatenated attention weights ($W_Q$, $W_K$, $W_V$) from each head before and after fine-tuning. Together, these visualisations provide insight into how the attention mechanism evolves during training.

**The Instance-Level View tab** is designed to provide detailed, sentence-level analysis. It is divided into three main sections: the *Sentence Viewer*, the *Token Analysis* section, and the *Attention Analysis* section. Figure 5.3 presents an architectural diagram of the components within this tab.

The **Sentence Viewer Section** includes a dropdown menu that allows users to select a sentence from the dataset. Once selected, the interface displays several views of that sentence, including the original sentence (before tokenisation), the tokenised version with gold-standard entity tags (Ground Truth), the tokenised version with predicted tags (Prediction), and a Mistakes view highlighting misclassified tokens. Errors are colour-coded based on the type of mistake (e.g., false positives or false negatives) using prediction outcome labels. These visualisations are shown at both the token level—where individual tokens are labelled with entity tags—and the span level, where complete entity spans are displayed independently of the original sentence and coloured by entity type.

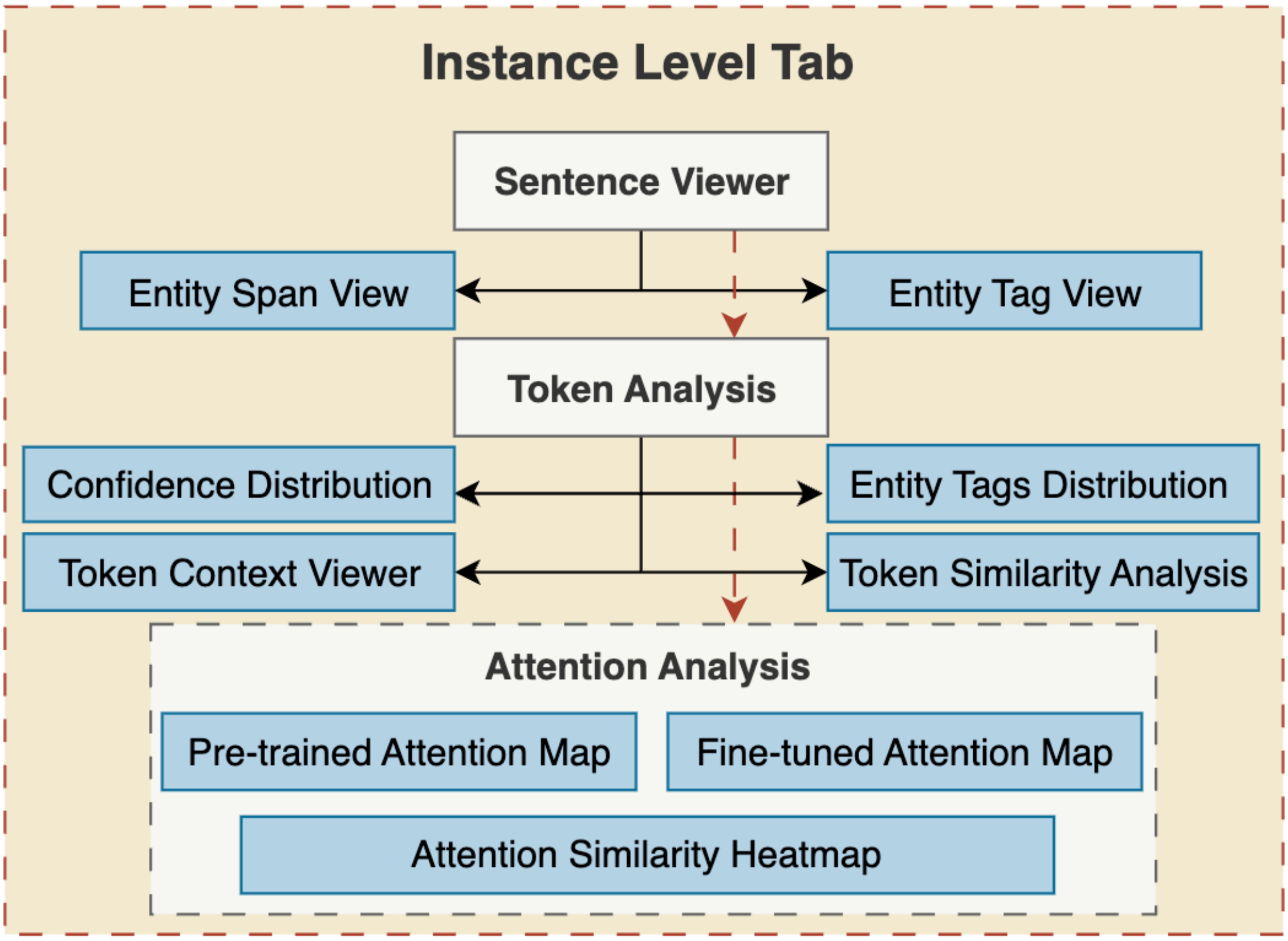

Figure 5.3: Architecture of the Instance-Level View tab. Solid black arrows indicate standard data flow, while dashed arrows represent interactive links between components. Red arrows highlight selection-driven interactions, where user selections in the Sentence Viewer propagate to update the corresponding visualisations in the Token Analysis and Attention Analysis sections.

The **Token Analysis section** allows for fine-grained exploration of tokens within the selected sentence. It includes a prediction confidence visualisation—a bar chart showing the model's confidence across all entity tags for the selected token—and a label distribution bar chart that illustrates how the token was annotated in both the training and test splits. Another key component is the token similarity analysis, which calculates similarity scores between the selected token and all other instances of the same token across the dataset. These scores are based on cosine similarity between the fine-tuned model's representations of the token across both splits. The results are displayed as both a bar chart and a table. In addition, this section enables users to inspect the contexts in which other instances of the token appear. For example, if the token "United" occurs multiple times in the dataset, and the instance in sentence 5 of the test split has the highest similarity score, the user can select that instance to view its surrounding context. The full sentence is then shown with the selected token highlighted, supporting interactive comparison.

The final section is the **Attention Analysis**. It includes two visual panels showing the attention matrices for the selected sentence—one from the pretrained model and one from the fine-tuned version. These visualisations, generated using BERTviz, display token-to-token attention distributions across all heads and layers. In addition, an attention score similarity matrix is presented, measuring the cosine similarity between the attention distributions of the pretrained and fine-tuned models for the selected sentence. This analysis allows users to examine how fine-tuning affects the model's attention patterns on individual sentences.

In the following sections, we explore the Behavioural Analysis tab and the Instance-Level View tab in detail.

### 5.3.2 Behavioural Analysis Tab

This tab serves as the main interface for exploring model behaviour and dataset characteristics. It integrates multiple views—including metric visualisations, representation space projections, clustering outcomes, and training impact visualisation—allowing users to investigate structural and performance-related patterns across different components of the system.

**Filtering Section** The Filtering Section enables targeted analysis by allowing users to subset the dataset based on selected categorical or behavioural metrics. This supports the investigation of specific patterns in the data—for example, isolating tokens with high loss values, particular entity tags, or specific error types. Two filtering mechanisms are provided. The first relies on dropdowns, which allow users to select a categorical variable and one or more values to filter by. The second mechanism is an interactive data table, which enables direct filtering using string matching or numerical conditions (e.g., selecting tokens with low confidence or high ambiguity). Filtering affects only the behavioural metrics visualisations—specifically, the behavioural metrics correlation heatmap and scatter plot—allowing users to explore how patterns vary across selected subsets. Figure 5.4 demonstrates the layout and interface elements of the Filtering Section.

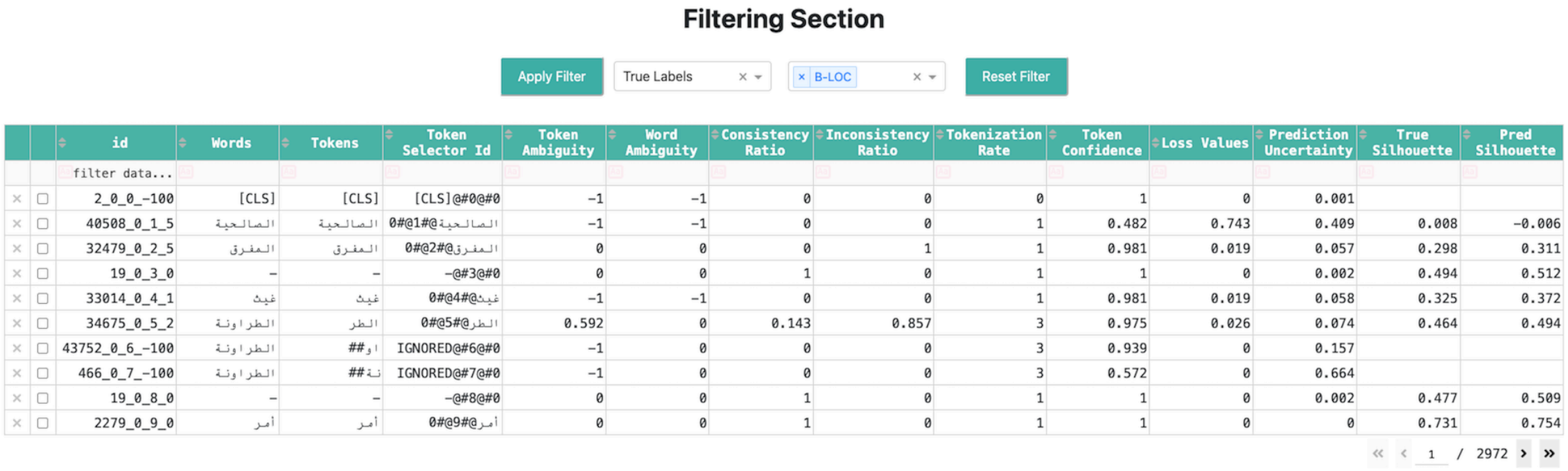

| id | Words | Tokens | Token Selector Id | Token Ambiguity | Word Ambiguity | Consistency Ratio | Inconsistency Ratio | Tokenization Rate | Token Confidence | Loss Values | Prediction Uncertainty | True Silhouette | Pred Silhouette |
|---|---|---|---|---|---|---|---|---|---|---|---|---|---|
| 2_0_0_-100 | [CLS] | [CLS] | [CLS]@#0@#0 | -1 | -1 | 0 | 0 | 0 | 1 | 0 | 0.001 | | |
| 40508_0_1_5 | الصالحية | الصالحية | الصالحية@#1@#0 | -1 | -1 | 0 | 0 | 1 | 0.482 | 0.743 | 0.409 | 0.008 | -0.006 |
| 32479_0_2_5 | المفرق | المفرق | المفرق@#2@#0 | 0 | 0 | 0 | 1 | 1 | 0.981 | 0.019 | 0.057 | 0.298 | 0.311 |
| 19_0_3_0 | - | - | -@#3@#0 | 0 | 0 | 1 | 0 | 1 | 1 | 0 | 0.002 | 0.494 | 0.512 |
| 33014_0_4_1 | غيث | غيث | غيث@#4@#0 | -1 | -1 | 0 | 0 | 1 | 0.981 | 0.019 | 0.058 | 0.325 | 0.372 |
| 34675_0_5_2 | الطراونة | الطر | الطر@#5@#0 | 0.592 | 0 | 0.143 | 0.857 | 3 | 0.975 | 0.026 | 0.074 | 0.464 | 0.494 |
| 43752_0_6_-100 | الطراونة | ##او | IGNORED@#6@#0 | -1 | 0 | 0 | 0 | 3 | 0.939 | 0 | 0.157 | | |
| 466_0_7_-100 | الطراونة | ##نة | IGNORED@#7@#0 | -1 | 0 | 0 | 0 | 3 | 0.572 | 0 | 0.664 | | |
| 19_0_8_0 | - | - | -@#8@#0 | 0 | 0 | 1 | 0 | 1 | 1 | 0 | 0.002 | 0.477 | 0.509 |
| 2279_0_9_0 | أمر | أمر | أمر@#9@#0 | 0 | 0 | 1 | 0 | 1 | 1 | 0 | 0 | 0.731 | 0.754 |

Figure 5.4: User interface of the Filtering Section. The dropdowns allow for filtering by categorical values, while the interactive data table supports both string-based and numerical filtering.

**Behavioural Analysis Section** The Behavioural Analysis Section consists of two interconnected components: the Behavioural Metric Visualisations and the Representation Space Visualisation. These components are designed to help users investigate the relationship between model predictions, internal representations, and various data and model metrics.

At the top of the Behavioural Analysis Section, users can configure the visualisations using a set of dropdowns. Figure 5.5 presents the layout of these user controls.

- **Model Type:** Selects which model's hidden states to visualise in the UMAP view (pretrained or fine-tuned)

- **Colour By (Behavioural Metric View):** Determines how the Behavioural Metric Scatter Plot is coloured, based on any of the categorical variables (e.g., true labels, entity types, or error categories) discussed in Section 5.2.3.
- **Colour By (UMAP Representation View):** Controls the colouring scheme used in the UMAP plot, using categorical variables to highlight different data groupings.
- **Select Numerical Variables:** Specifies which behavioural metrics are included in the Behavioural Metric Correlation Matrix.
- **Coefficient Type:** Sets the type of correlation coefficient (e.g., Pearson or Spearman) used to compute the matrix.

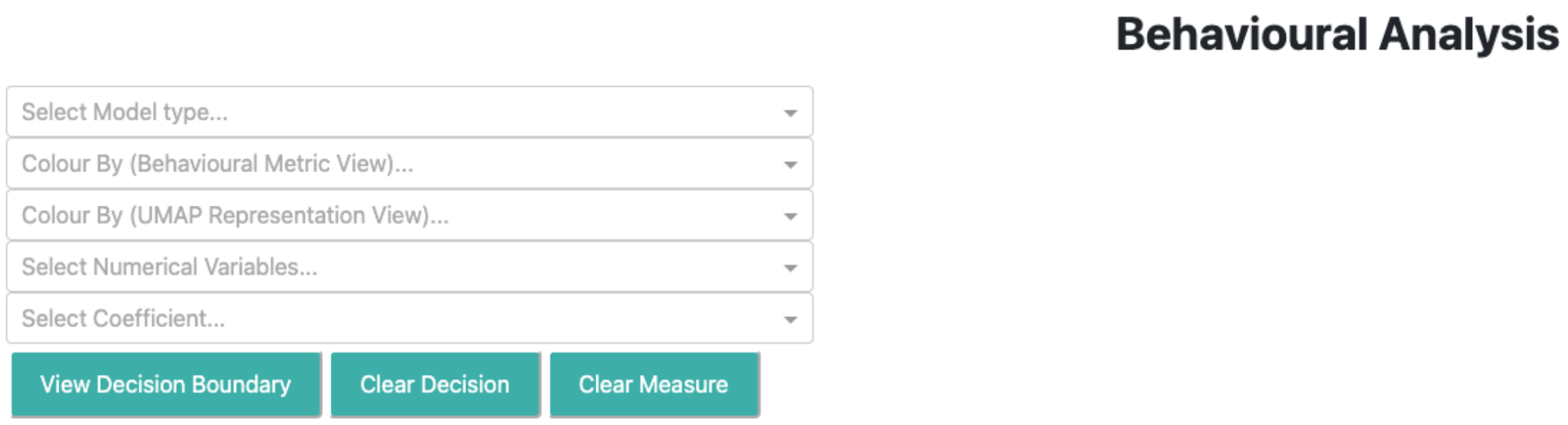

Figure 5.5: User controls for the Behavioural Analysis Section. These dropdowns and selectors allow users to customise the displayed visualisations based on model type, categorical colouring, and selected metrics.

The Behavioural Metric Visualisations include two interactive components. The first is the *Behavioural Metric Correlation Matrix*, which displays the pairwise correlations between all behavioural metrics described in Section 5.2.3. Users can click on any cell in the matrix to select a pair of metrics. This interaction dynamically updates the second component—the *Behavioural Metric Scatter Plot*—which visualises the selected metrics against each other. An additional dropdown, labelled *Colour By (Behavioural Metric View)*, allows users to colour the data points based on a chosen categorical variable, such as True Labels or any of the other variables discussed in Section 5.2.3. Figure 5.6 presents the layout and interactive functionality of these components. It is also worth noting that the scatter plot includes two dashed grey lines that represent the mean values of the selected variables along the $x$- and $y$-axes, respectively. These visual guides help users interpret the correlation structure more intuitively.

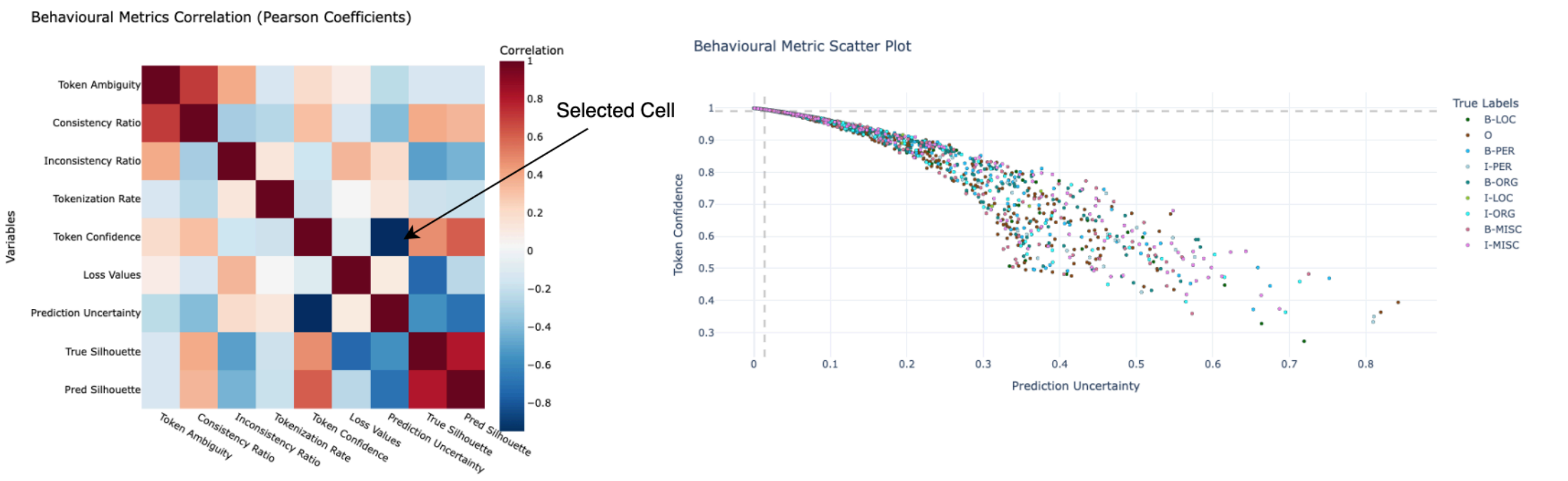

Figure 5.6: Behavioural Metric Visualisations. The left panel shows the correlation matrix between selected behavioural metrics. Clicking on a cell dynamically updates the scatter plot on the right, which visualises the relationship between the selected pair of metrics. Points are coloured based on the selected categorical variable.

The *Representation Space Visualisation* displays a UMAP-projected 2D scatter plot of the model's hidden states. Users can choose between visualising the representations from the pretrained or fine-tuned model using the *Model Type* dropdown. A second dropdown—*Colour By (UMAP Representation View)*—controls the colouring scheme based on a selected categorical variable. This visualisation is linked to the Behavioural Metric Scatter Plot: selected data points in one view are highlighted in the other, allowing users to cross-reference behavioural metrics with structural patterns in the representation space. Two buttons are provided to clear selections in either view. Figure 5.7 shows the interface layout for the Representation Space Visualisation.

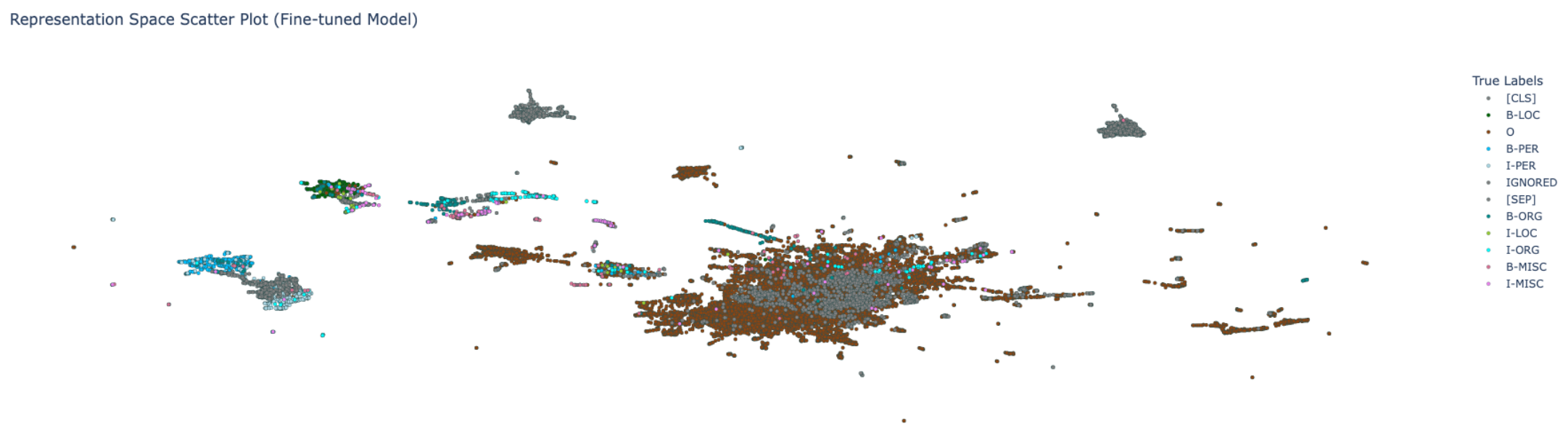

Figure 5.7: Representation Space Visualisation. This UMAP-based scatter plot displays the hidden state representations of tokens, projected into two dimensions. Users can choose between pretrained and fine-tuned views, and colour the data points using a chosen categorical variable.

Together, these visualisations support an integrated analysis of model behaviour and data characteristics. To assist with interpreting specific patterns or data points identified in these views, we provide a dedicated Selection Summary section. This section provides information about the tokens currently selected in either the metric scatter plot or the representation space scatter plot, enabling users to inspect the metadata associated with the selected data points. These summaries are described in the following paragraph.

**Selection Summary Section** Figure 5.8 illustrates the interaction between the Behavioural Metric Scatter Plot (A) and the Representation Space Scatter Plot (B). Users can select a group of tokens in either view using a lasso or box selection. Once selected, the corresponding points are visually highlighted in both plots. In (A), the selected points are highlighted by increasing their marker size, switching their marker shape to diamonds, and adding a black outline. The original colouring—based on the selected categorical variable—is preserved, and the legend appends a "(Selected)" label for the affected categories. In (B), the selected points are similarly highlighted using larger markers with reduced opacity, making them stand out from the surrounding context. Additionally, the selection area is visualised with an approximate bounding box around the selected region, offering spatial context for the user's selection. Users may also interact with the legend to toggle the visibility of individual categories in the scatter plots, helping to reduce visual clutter during exploration. This selection behaviour is bidirectional—selecting points in either plot is reflected in the other—and selections are automatically propagated to the Selection Summary section for further inspection.

The Selection Summary, shown in Figure 5.9, provides both categorical and numerical breakdowns of the selected tokens. The interface includes a bar chart and two tables. A dropdown allows users to choose a categorical variable (e.g., predicted label, chunk type), which determines the colouring and grouping of the bar chart. The y-axis remains fixed to the true labels of the selected data points, while the x-axis displays their distribution over the selected category. Next to the bar chart on the right, there are two summary tables. The first presents the same categorical distribution shown in the bar chart, displayed in tabular form with counts and percentages. The second focuses on the behavioural metrics and provides descriptive statistics—including mean, standard deviation, and percentiles—for all the behavioural metrics. Together, these views offer a structured breakdown of the selected subset, supporting detailed interpretation of the patterns observed in the scatter plots.

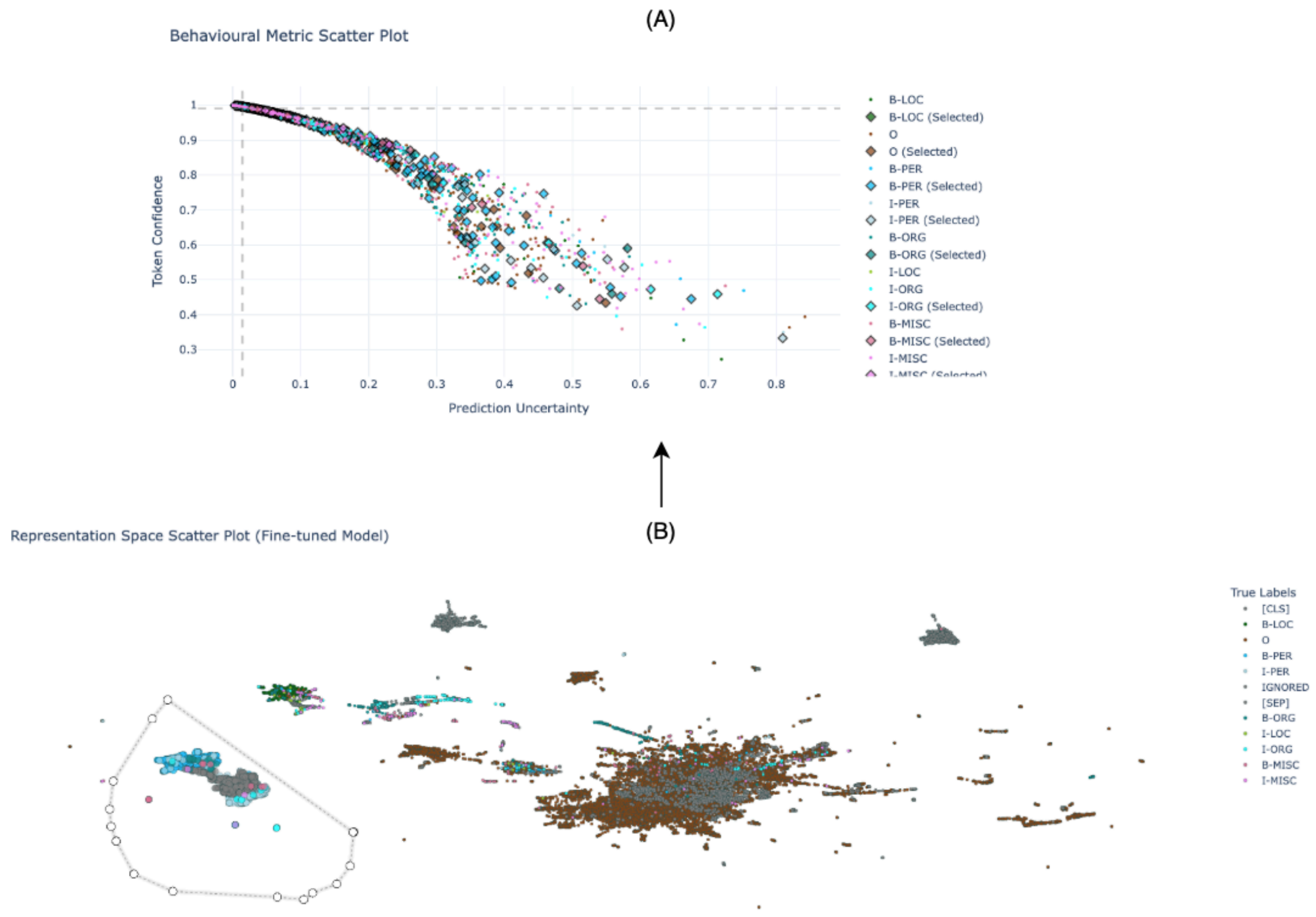

Figure 5.8: Illustration of the selection interaction between the Behavioural Metric Scatter Plot (A) and the Representation Space Scatter Plot (B). Tokens selected in either plot are synchronised across views—highlighted by colour and label changes in (A), and by increased marker size and reduced opacity in (B). This interaction enables cross-referencing of behaviour in metric space and representation space.

**Interaction and Colouring** In addition to selection mechanisms, the tab offers flexible colouring and interactive functionalities that enhance pattern discovery during behavioural analysis. Users can colour the data points based on a chosen categorical variable, such as True Labels or Error Types. The legend associated with each scatter plot allows toggling specific categories on and off; when a category is untoggled, its corresponding points are temporarily removed from the visualisation, and toggling it back restores their display. Furthermore, the tab supports a dual encoding mode, where two categorical variables are used simultaneously: one controls the colour of the points, and the second controls their shape. In this case, points corresponding to the second categorical variable are assigned a diamond marker[1].

The system also includes two types of selection-driven highlighting, which allow users to trace and compare tokens across different views. The first occurs within a sin-gle plot, where selecting points using lasso or box tools increases the marker size and reduces their opacity. This enhances the visibility of patterns within the selected subset, making differences in symbol shape or colour more apparent. The second type involves cross-plot highlighting: when a selection is made in one plot, the corresponding points in the linked plot are emphasised by dimming all non-selected points, increasing the size of selected points, and changing their marker shape into a distinct diamond. This highlighting diamond is visually different from the one used during categorical dual encoding, ensuring that users can easily distinguish between the two. Figure 5.10 illustrates the distinction between category-driven diamonds and selection-driven diamonds. Together, these dynamic colouring and interactive features make it easier to spot structure in the model's behaviour across both behavioural metric space and representation space. These interactive colouring and selection behaviours are implemented specifically within the Representation Space Visualisation and the Behavioural Metric Scatter Plot components.

[1] These are features provided by Plotly and supported in the dashboard.

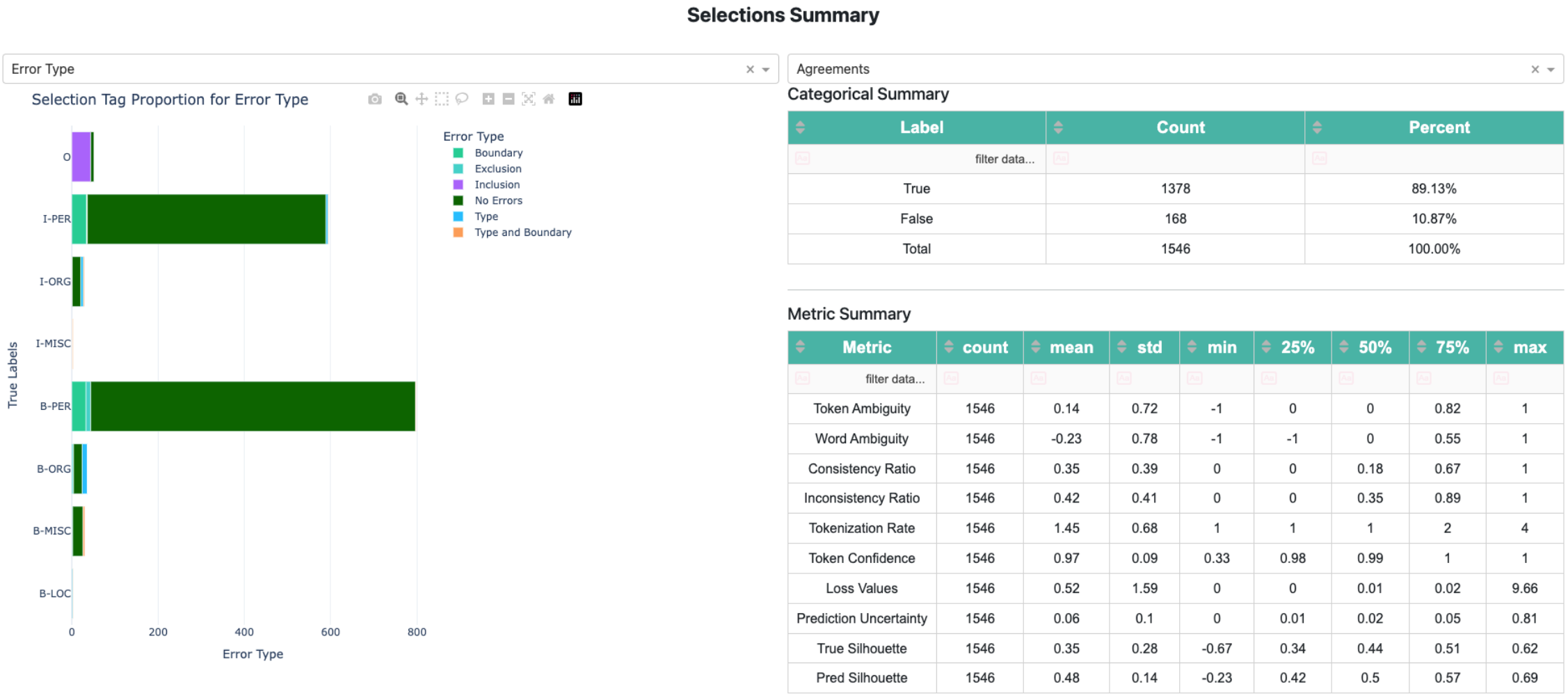

Figure 5.9: User interface of the *Selection Summary* section. On the left, a bar chart shows the distribution of the selected tokens by True Labels (y-axis) and the selected categorical variable (x-axis), in this case *Error Type*. On the right, a categorical summary table displays the corresponding counts and percentages for the selected category (here *Agreements*), and a metric summary table presents descriptive statistics for all behavioural metrics. These views provide a structured overview of the selected data subset.

**Clustering Analysis Section** The Clustering Analysis section is a static component of the tab. Unlike interactive views that respond to user input—such as dropdown selections or region-based scatter plot selection—static components are automatically ren-dered when their tab is activated and do not require further interaction. These views only update when the selected language changes. This behaviour also applies to the Training Impact section. The Clustering Analysis view supports the investigations discussed in Section 5.6, which examine the alignment between the model's internal representations and clusters generated by KMeans. While this analysis is distributed across several components in the tab, this section highlights two key elements:

- **Clustering Metrics Table:** Presents homogeneity, completeness, and V-measure scores for several values of $k$, offering a summary of clustering quality across varying granularities.
- **Centroid Similarity Heatmap:** Reports the average similarity between each discovered centroid and different named entity tags. This helps assess whether centroids are well-aligned with, or positioned near the centre of entity-specific regions in the representation space.

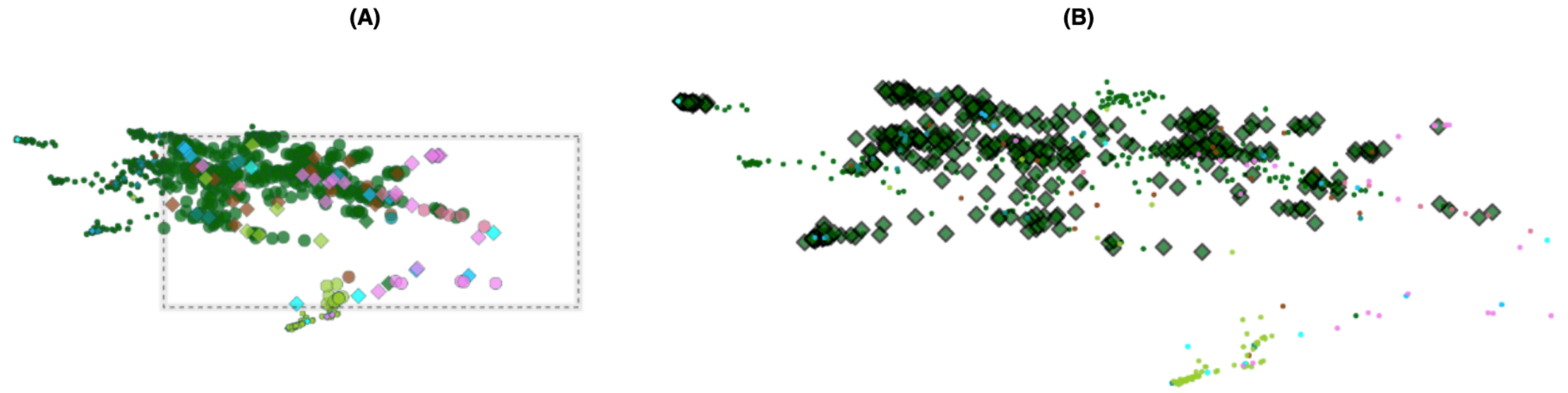

Figure 5.10: Comparison of interactive highlighting modes. (A) shows intra-plot highlighting, where selected points increase in size and reduce opacity within the Behavioural Metric or Representation Space scatter plots. (B) shows cross-plot highlighting, where selected points are emphasised in the linked plot by dimming all non-selected points, increasing the size of selected points, and applying a black border around them. Note that in (B), the diamond markers used for highlighting are visually distinct from those used for dual categorical encoding in (A).

Additionally, in the Behavioural Analysis section, clustering annotations can be selected via the "Colour By" dropdown, enabling users to visually inspect how well the clustering algorithm segments the representation space. For example, the annotation "K=3 (Chunk)" uses only the boundaries of true named entity spans to group tokens, allowing us to examine whether clusters align with known entity groupings.

**Training Impact Section** The Training Impact section is another static component of the tab. Like the Clustering Analysis section, its contents are rendered when the tab is activated and only change when the selected language changes. This section focuses on how the model's attention behaviour changes as a result of fine-tuning and presents two dedicated visualisations. The following visualisations are included:

- **Attention Score Similarity Heatmap:** Displays the cosine similarity between attention matrices generated by the pretrained and fine-tuned models, computed for each attention head across all layers and averaged over the full dataset.
- **Attention Weight Similarity Heatmap:** Shows the similarity between the learned attention parameters ($W_Q$, $W_K$, and $W_V$), concatenated and compared between the pretrained and fine-tuned models. This provides a direct view of how the internal attention parameters are updated during training.

### 5.3.3 Instance-Level View Tab

The Instance-Level View tab supports fine-grained, sentence-level inspection of model behaviour. Unlike the Behavioural Analysis tab, which focuses on token-level empirical and structural patterns, this view provides detailed insight into individual examples. It is organised into three sections: the Sentence Viewer, Token Analysis, and Attention Analysis.

**Sentence Viewer Section** The Sentence Viewer consists of two components: token-level and entity-level annotation views. Both are controlled by a shared dropdown, which determines the sentence to be displayed. This dropdown is also connected to the selection functionality in the Behaviour Analysis tab—if a user selects points in a scatter plot, the dropdown filters to only show sentences containing those selected tokens; otherwise, it defaults to displaying all available sentence IDs.

In the **Token-Level Annotation View** (Figure 5.11), the raw sentence is displayed alongside its tokenised version, the model's predictions, and the ground truth labels. Each token is colour-coded based on its true and predicted label using a consistent colour map shown at the top of the view. Tokens are also annotated with their prediction outcomes such as true positive (TP), false positive (FP), and false negative (FN).

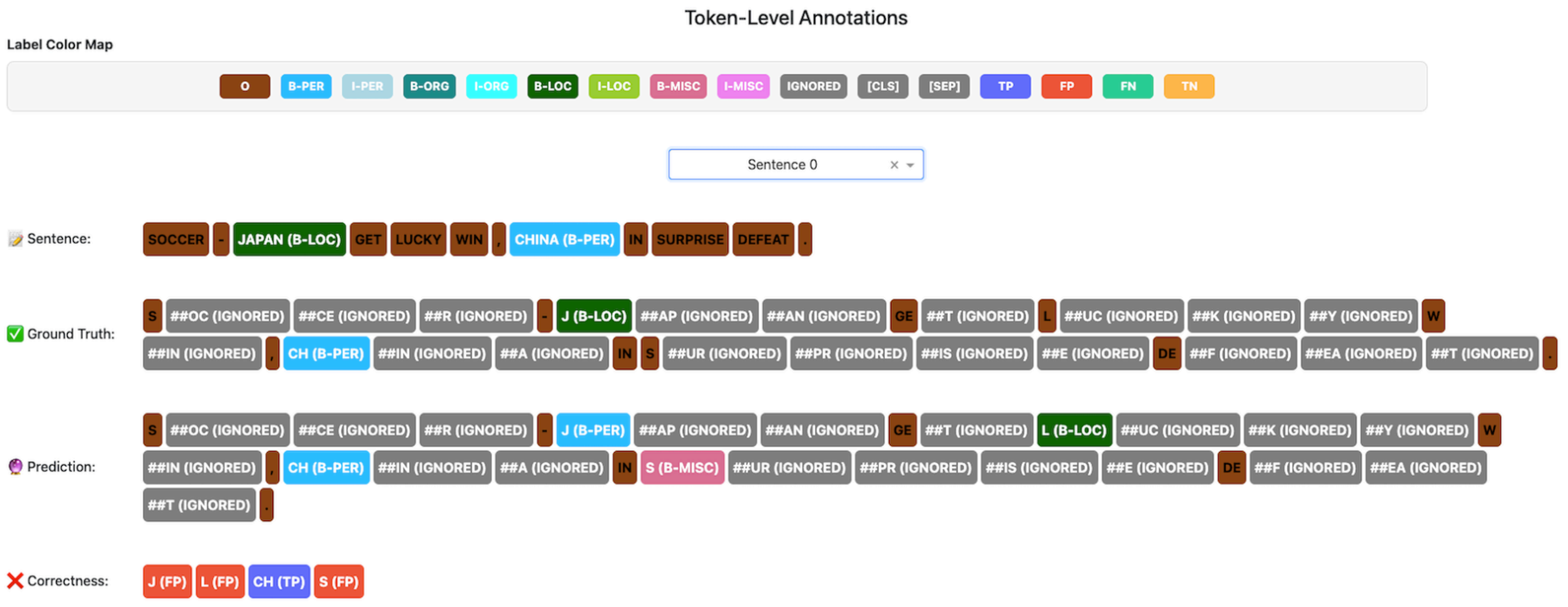

Figure 5.11: This view shows the raw sentence, the tokenised version, ground truth, model predictions, and token-level mistakes. It uses consistent colour coding for entity tags and highlights prediction outcomes with labels such as TP, FP, and FN. Tokens ignored during evaluation are greyed out.

In the **Entity-Level Annotation View** (Figure 5.12), entity spans are grouped and colour-coded by label. The system supports both the IOB1 and IOB2 annotation schemes. These views facilitate the comparison between each scheme. The true and predicted entities are shown in separate lines, and the entity-level mistakes are summarised in the final line. Together, these two views allow for fine-grained inspection of sentence-level predictions, both at the token and entity span level.

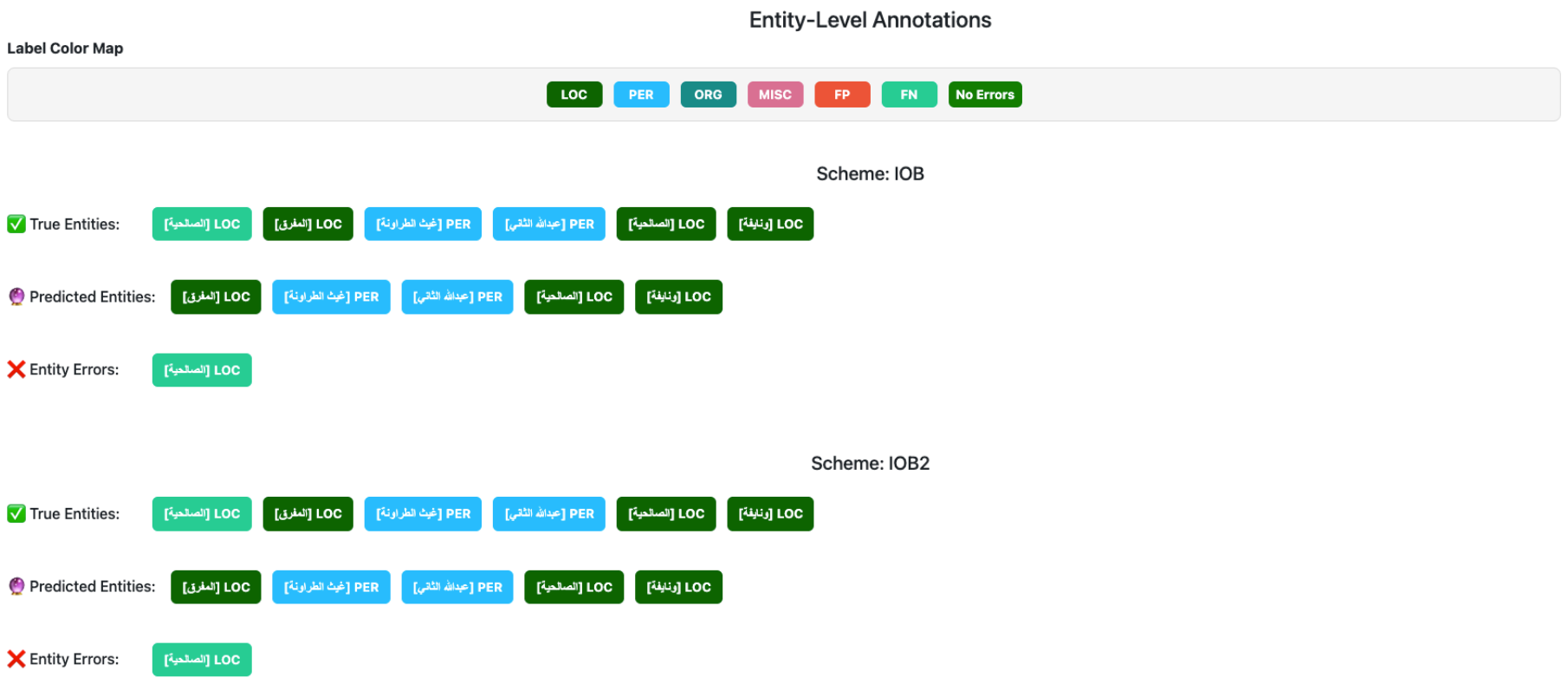

Figure 5.12: Displays predicted and ground truth entity spans, using both IOB1 and IOB2 tagging schemes. Each entity is colour-coded by its type, and entity-level mistakes are listed separately to compare between the errors in each scheme.

**Token Analysis Section** The Token Analysis section provides a detailed view of the behaviour of individual tokens, including their prediction confidence, annotation distribution, contextual representations, and semantic similarity across the dataset. This section is supported by three main components. At the top of the interface, users interact with a set of controls that initialize this section. These include a dropdown menu and two action buttons. The dropdown is populated with all tokens from the sentence selected in the Sentence Viewer. Once a token is selected, two charts are dynamically updated: a confidence score bar chart and a label distribution bar chart. An illustration of this layout is shown in Figure 5.13, which displays the control panel along with the updated visualisations for a selected token.

The Prediction Confidence Chart shows the model's confidence scores across all entity tags for the selected token. This provides insight into the uncertainty of the model's prediction and highlights how probability is distributed among possible tags. The Label Distribution Chart summarizes how the selected token has been annotated across the dataset. It distinguishes between training and test splits, allowing users to investigate whether confidence levels may be influenced by how frequently and consistently a token appears across different splits of the dataset. In the example shown in Figure 5.13, the selected token الصالحية only appears in the test split, making it out-of-vocabulary (OOV). Consequently, the label distribution chart shows only test data.

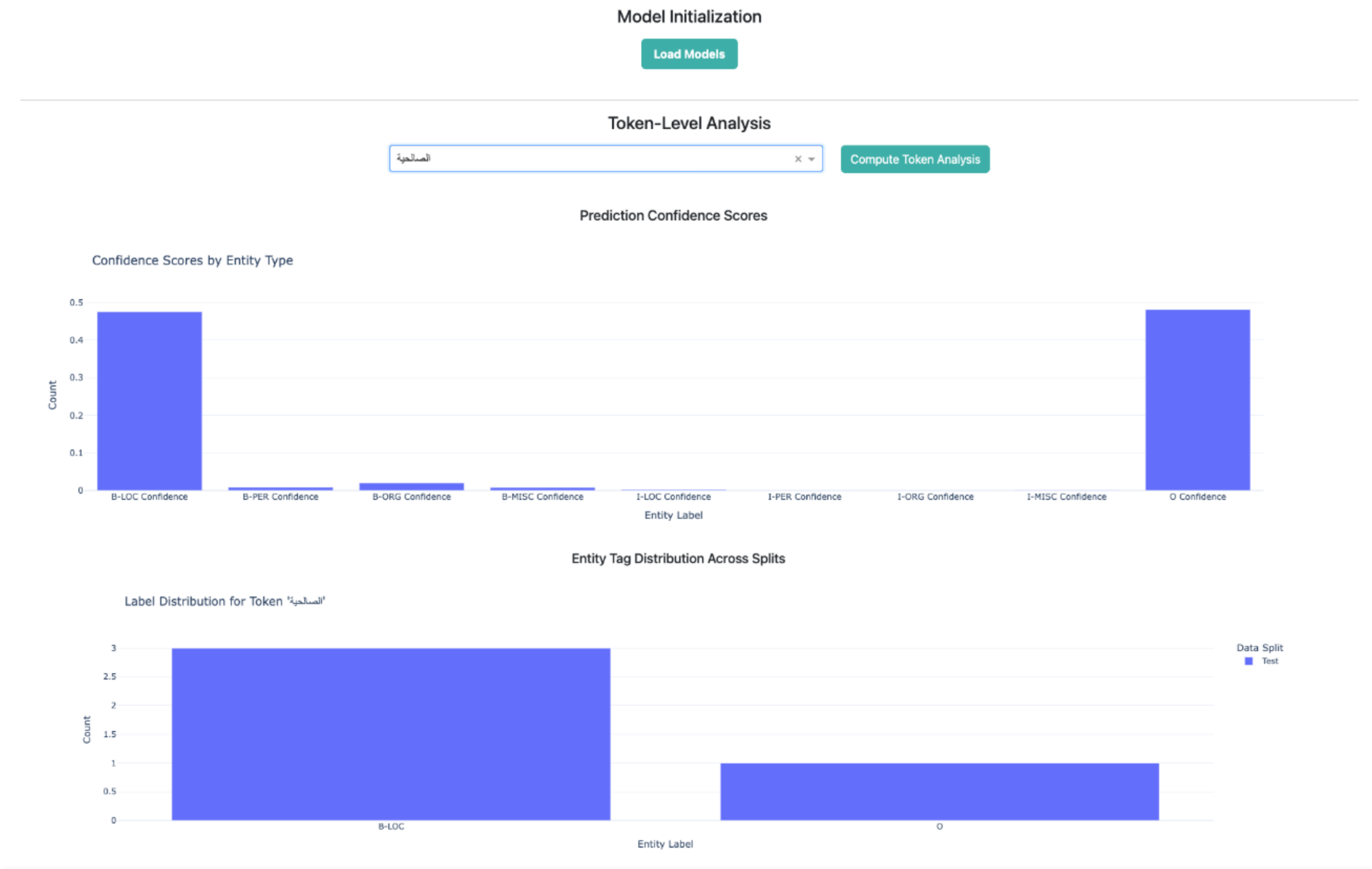

Figure 5.13: Token-Level Analysis for the token الصالحية. Top: confidence distribution across all possible entity tags. Bottom: label distribution across data splits (test only, in this case, as the token is OOV).

The second component is the Token Similarity Analysis, which is triggered by the "Compute Token Analysis" button. This feature retrieves the contextual representations of the selected token from all occurrences in the dataset, then computes pairwise cosine similarity scores between them. The results are grouped by data split and visualised through both a bar chart and a similarity table. As illustrated in Figure 5.14, the token الصالحية appears in three sentences in the test split, and no similar tokens were found in the training split, further confirming its out-of-vocabulary status. The table and chart together provide a summary of which instances are most similar and where they occurred. It is important to note that this component requires model initialization, which can be triggered using either of the action buttons provided in the interface.

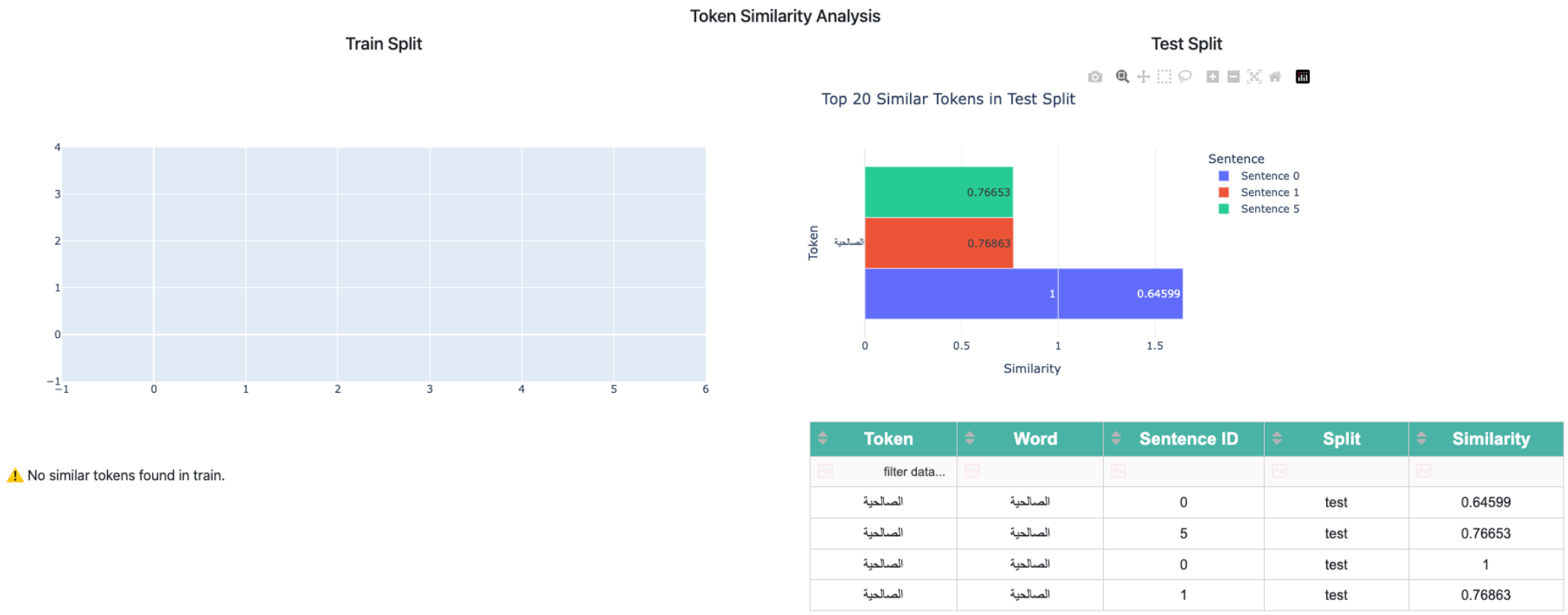

Figure 5.14: Token Similarity Analysis. Right: similarity scores for token instances across the test split. Left: no similar tokens were found in the training split. Bottom: similarity table showing the instances most similar to the selected token.

The final component is the Token Context Viewer, which allows users to examine the sentence-level semantic context in which the selected token appears. For any of the retrieved similar instances, users can view the full sentence with the selected token highlighted. This view supports interactive comparison across contexts and helps users understand how different semantic contexts may influence the token's representation. Figure 5.15 shows an example where the token appears twice in the same sentence.

In summary, the Token Analysis section brings together multiple views—confidence scores, label distribution, semantic similarity, and contextual usage—to enable comprehensive analysis of token-level behaviour. It offers focused insights into how token-level factors—such as annotation frequency and contextual variation—relate to the model's predictions.

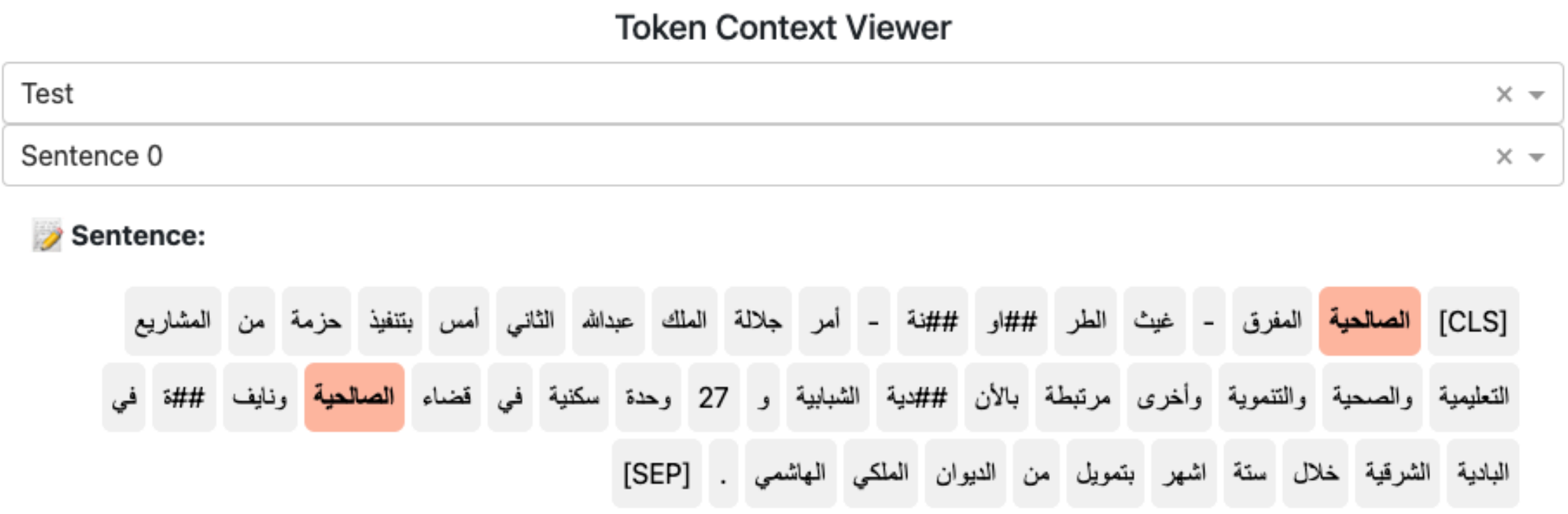

Figure 5.15: Token Context Viewer showing two highlighted instances of the selected token within a single sentence. This view enables fine-grained contextual inspection to aid qualitative analysis.

**Attention Analysis Section** The Attention Analysis section provides an interface for comparing the attention patterns of the pretrained and fine-tuned models on a sentence-level basis. It consists of two main visual components: token-to-token attention maps and an attention score similarity heatmap. Together, these views help users visually inspect how attention distributions change after fine-tuning.

The layout, shown in Figure 5.16, integrates BERTviz (Vig, 2019) to display side-by-side attention maps for a selected sentence from both the pretrained and fine-tuned models. Users can select which layer to display via a dropdown and toggle between Head View and Model View to explore attention behaviour across individual heads, as implemented in BERTviz. The attention patterns are visualised as coloured lines connecting tokens, enabling direct observation of which tokens attend to each other and how this behaviour shifts between pre-training and fine-tuning. While attention maps have recognised limitations as interpretability tools, they remain useful for comparing how attention patterns change before and after fine-tuning.

To complement this, a heatmap view presents the cosine similarity between attention matrices of the pretrained and fine-tuned models. As shown in Figure 5.17, this view provides a higher-level summary of how attention behaviour has changed, aggregated across layers and heads. Lower similarity values (darker regions) indicate more substantial changes due to fine-tuning. This component connects to the broader training impact analysis described in Section 5.4. In the next section, we begin the analysis by examining the impact of fine-tuning on the model.

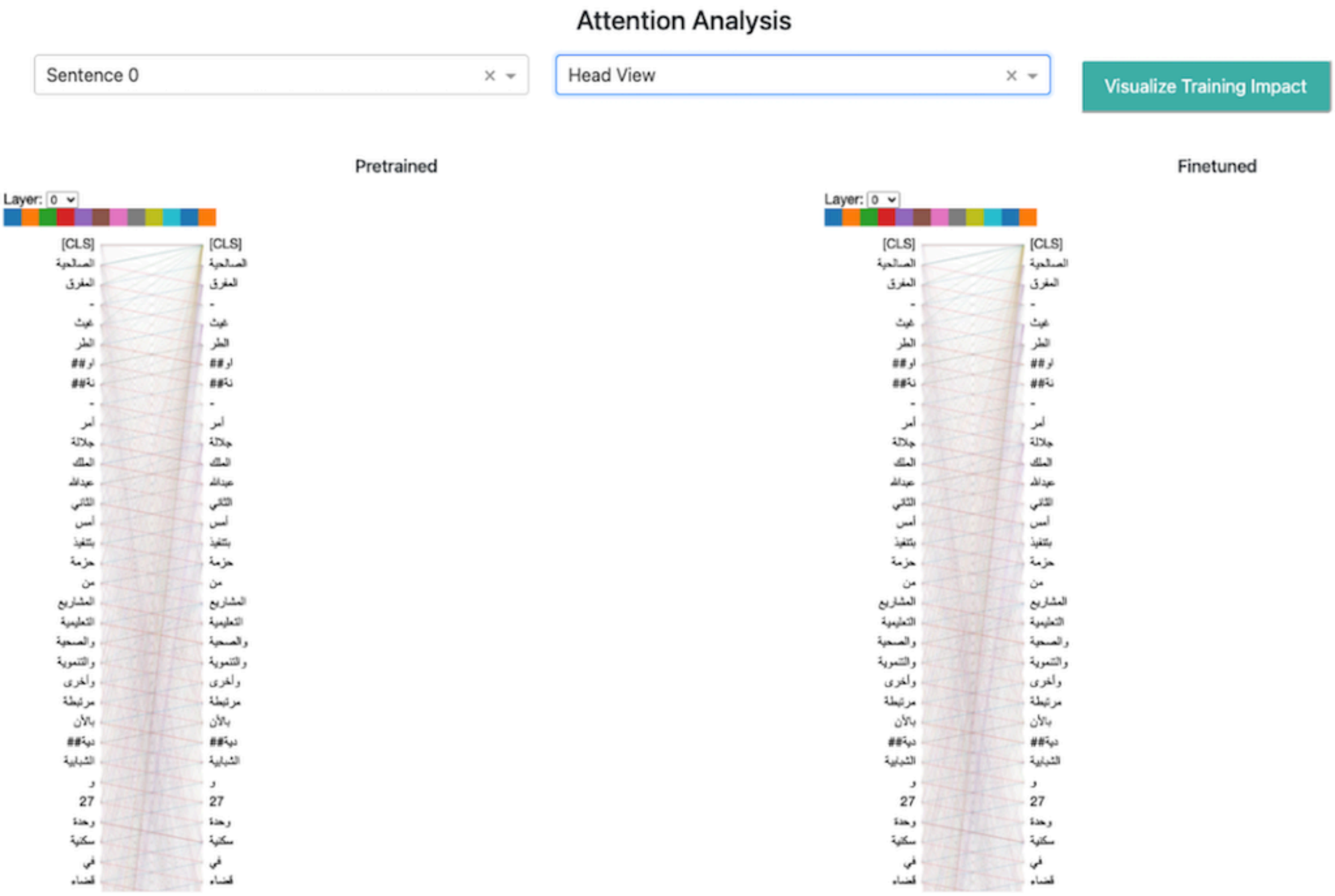

Figure 5.16: Attention Analysis layout using BERTviz. Attention distributions are shown side by side for the pretrained (left) and fine-tuned (right) models. Users can select layers and view the alignment of attention between tokens.

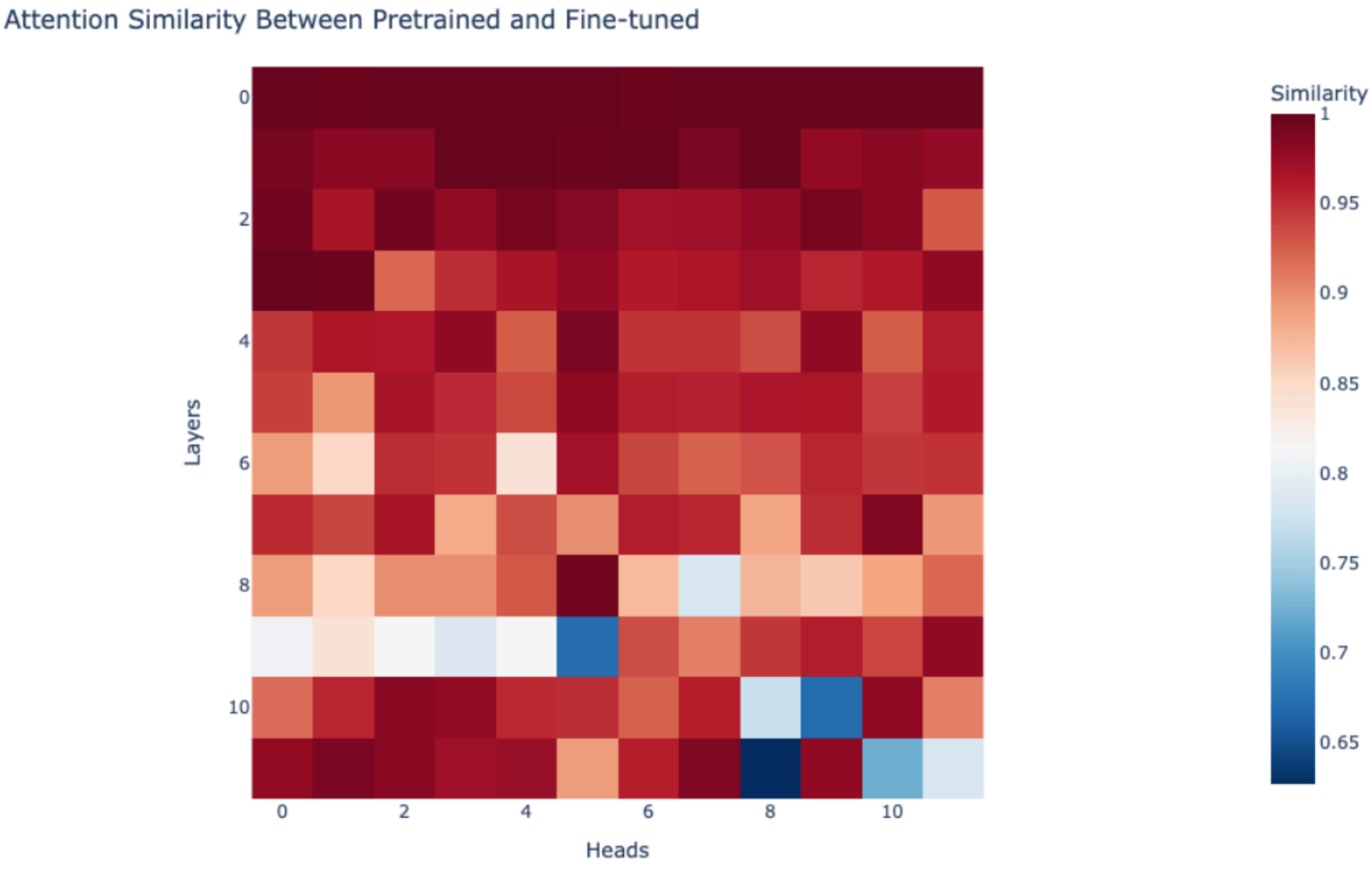

Figure 5.17: Attention Score Similarity Heatmap. Lower similarity (darker regions) indicates greater change due to fine-tuning.

## 5.4 Training Impact

To address aspects (a) and (b) of the *Training Impact component of RQ5—(a) how fine-tuning affects the model's internal attention behaviour and token representations, and (b) how these effects differ across languages*—we investigate how the structure of representations and the internal attention mechanisms evolve before and after fine-tuning. This analysis provides insight into how the model adapts to the NER task and highlights language-specific differences in the nature of these adaptations.

As discussed in Section 2.1.2, BERT builds contextualised token representations by passing input embeddings through multiple layers of self-attention. Fine-tuning modifies these internal structures by updating various sets of parameters, including the token embeddings, the attention weight matrices, and the layer-wise transformation functions.

To analyse the impact of fine-tuning, we begin by examining changes in the global structure of the representation space, using UMAP projections of the last hidden states. This allows us to understand how the structure of token representations shifts after fine-tuning.

Following this, we investigate changes in the internal attention mechanisms. We analyse both the attention scores—capturing how tokens attend to one another—and the learned attention weight parameters (the Q, K, and V matrices) to understand how fine-tuning modifies the model's contextualisation processes.

Finally, we examine how token representations evolve across different layers of the model, from the input embeddings through to the middle and final layers. This layered analysis reveals how fine-tuning gradually changes token-level information throughout the model layers. The following subsections present the results of this analysis.

### 5.4.1 Representation Space Changes

In this section, we examine the global structure of the representation space. We feed the data through the pretrained versions of each model, extract the last hidden state, and apply UMAP to project the high-dimensional representations into two dimensions. Figure 5.18 shows the resulting scatter plots for the pretrained models.

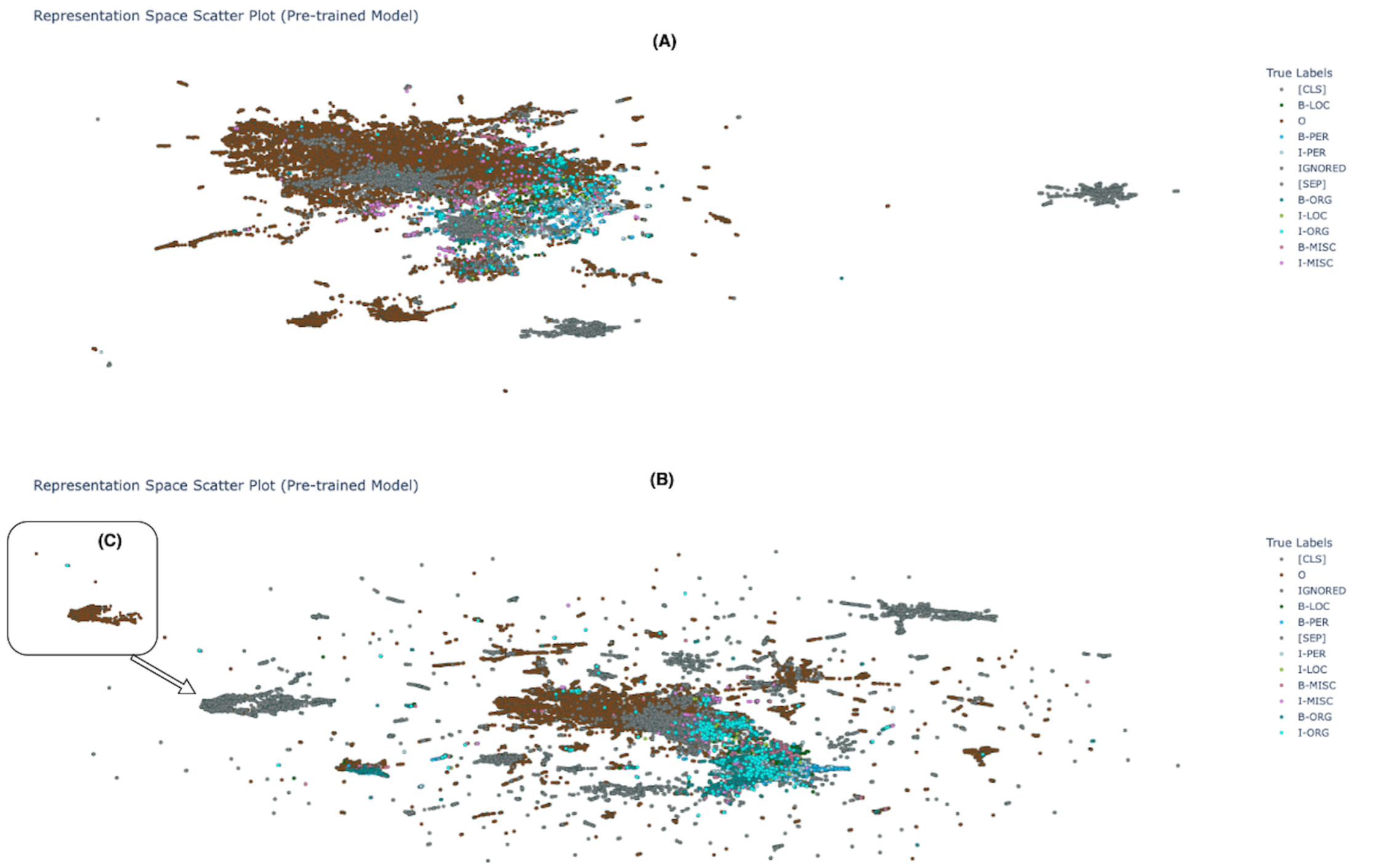

Figure 5.18: UMAP projection of token representations from the pretrained models. (A) AraBERTv02 for Arabic. (B) BERT for English. Colours indicate the token's true entity label. Grey tokens correspond to [CLS], [SEP], and ignored subword tokens. Region (C) highlights a cluster of O-tagged tokens, mostly punctuation and numeric characters.

Special tokens such as [CLS] and [SEP] are clearly marked in grey. IGNORED refers to subword tokens that are not the first subword of a tokenised word; while these subwords are excluded from loss computation, they still contribute to the contextualised representation space, as discussed in Section 3.3.5.

At the pretrained stage, there is no clear separation between entity tags in either language. Most tokens—including those tagged as named entities—are densely packed together with overlapping regions and no obvious structure. This is expected, as the models have not yet been adapted to the NER task.

However, some language-specific patterns can already be observed. In the Arabic model (Figure 5.18 A), the [SEP] tokens form a distinct, isolated group far from the main token distribution. IGNORED tokens tend to concentrate centrally, appearing between dense clusters of O-tagged and entity-tagged tokens. Additionally, we observe small isolated clusters of O-tagged tokens, mainly corresponding to punctuation, numeric tokens, or repeated words such as أنها (annaha). The English model (Figure 5.18 B) shows a slightly more spread-out distribution; however, [CLS] and [SEP] tokens are not as clearly separated. As highlighted in region (C), a small distinct cluster of period characters (.) appears just below the [SEP] tokens. As with Arabic, several O-tagged tokens, mostly punctuation or numbers, form distinct subregions even before fine-tuning.

After fine-tuning, significant structural changes appear in the representation space of both models. As shown in Figure 5.19, the models adapt their internal representations to align more closely with NER annotations. Tokens sharing the same entity labels now tend to form clearer groupings, and the separation between entity types and the O class becomes more pronounced. One notable observation in Arabic is a B-MISC token corresponding to الله (lafz al-Jalalah, the name "Allah"), which appears embedded within the [SEP] token cluster. In English, the previously identified cluster of period characters disappears after fine-tuning, suggesting they have been integrated into the broader O-tagged token space.

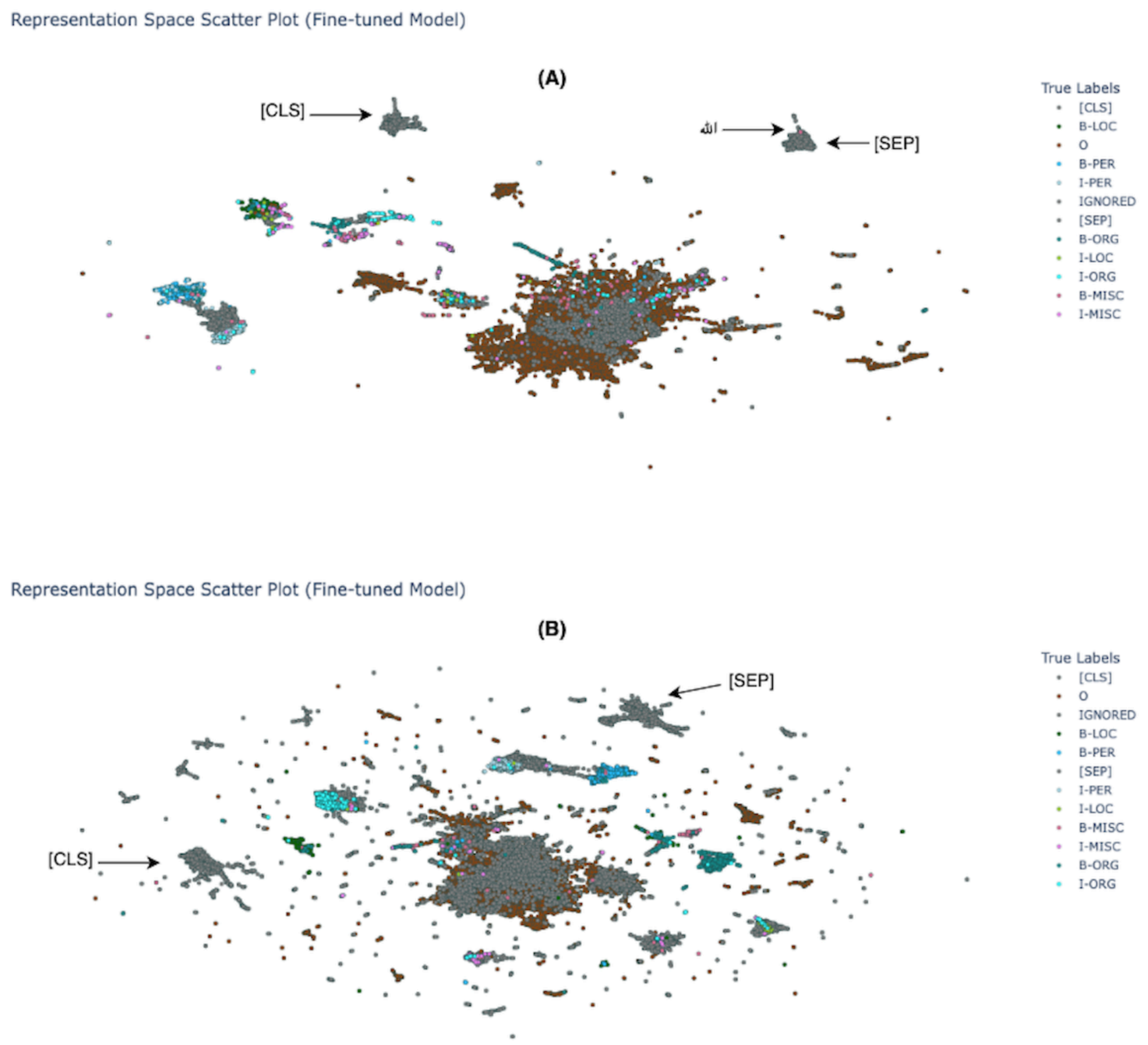

Figure 5.19: UMAP projection of token representations after fine-tuning. (A) Arabic (AraBERTv02). (B) English (BERT). Each point represents the final hidden state of a token, coloured by its gold-standard label. Special tokens such as [CLS], [SEP], and IGNORED are shown in grey.

While both models exhibit improved organisation after fine-tuning, the degree of separation varies across languages. In English (Figure 5.20 B), entity groupings are tighter and boundaries between tags are clearer. In contrast, the Arabic model (Figure 5.20 A) shows more overlap, particularly between ORG, MISC, and O tokens.

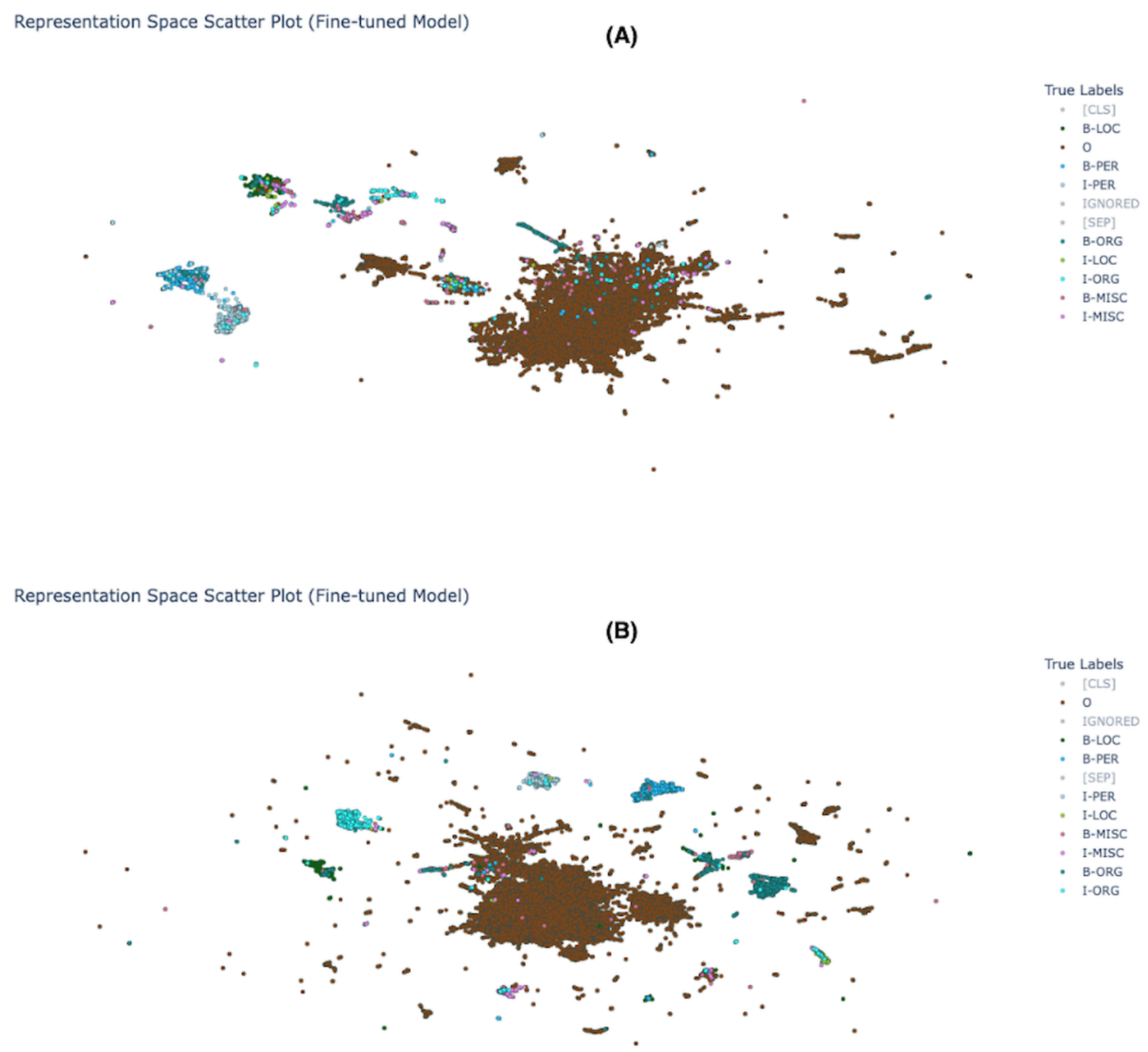

Figure 5.20: UMAP projection of core token representations (excluding [CLS], [SEP], and IGNORED tokens) after fine-tuning. (A) Arabic model. (B) English model. Compared to Arabic, the English model shows tighter, more distinct clusters by entity tag.

We also observe consistent patterns regarding special tokens. Across both languages, [SEP] and [CLS] tokens remain separated from the rest of the distribution. However, in Arabic, they are more tightly grouped and positioned further from core tokens, while in English they remain closer to the main token clusters.

The behaviour of IGNORED tokens also changes after fine-tuning. Despite not directly influencing the loss or output layer, these tokens now align more closely with their associated core tokens. As shown in Figure 5.21, IGNORED tokens are embedded within the corresponding entity regions. This effect is especially visible in English, where tokenisation often produces multiple subwords per word, leading to stacked IGNORED tokens over their core representations.

In summary, fine-tuning causes restructuring of the representation space, improving entity-type separation and revealing language-specific patterns in how tokens are organised. Notably, even IGNORED tokens—those excluded from direct loss computation—show structural changes, aligning with their associated core tokens, suggesting that loss propagation still influences their representations. The next section investigates how these changes correspond to shifts in the model's attention mechanisms.

### 5.4.2 Attention Analysis

Building on the previous section, where we observed substantial changes in the structure of the representation space after fine-tuning, we now investigate how the model's internal attention behaviour evolves during fine-tuning.

As discussed in Section 2.1.2, BERT models construct token representations through stacked self-attention layers, where each attention head computes weighted interactions between tokens. Fine-tuning updates these attention patterns to align the model more closely with the task-specific structure—in this case, NER.

To analyse these changes, we compare attention scores before and after fine-tuning. Specifically, for each attention head, we extract the attention score matrices, flatten them into vectors, and compute the cosine similarity between the pretrained and fine-tuned states. This quantifies how much each attention head's behaviour changes after fine-tuning.

To illustrate the attention mechanism and the analysis process, Figure 5.22 shows an example attention pattern for Sentence 0 in the English test set, using Layer 10, Head 8, extracted from the fine-tuned model. Panel (A) highlights which tokens attend to the selected token "J" (incoming attention), while panel (B) shows which tokens "J" attends to (outgoing attention). Panel (C) displays the full attention score matrix for the sentence.

We compare the pretrained and fine-tuned attention score patterns by computing cosine similarity between corresponding attention heads. For each language, we average the similarity scores across the test split to produce a 12×12 matrix, where each cell reflects the degree of change for a specific head and layer. Darker colours indicate higher similarity (i.e., smaller change), while lighter colours indicate greater shifts in attention behaviour. Figure 5.23 shows the resulting similarity matrices for Arabic and English.

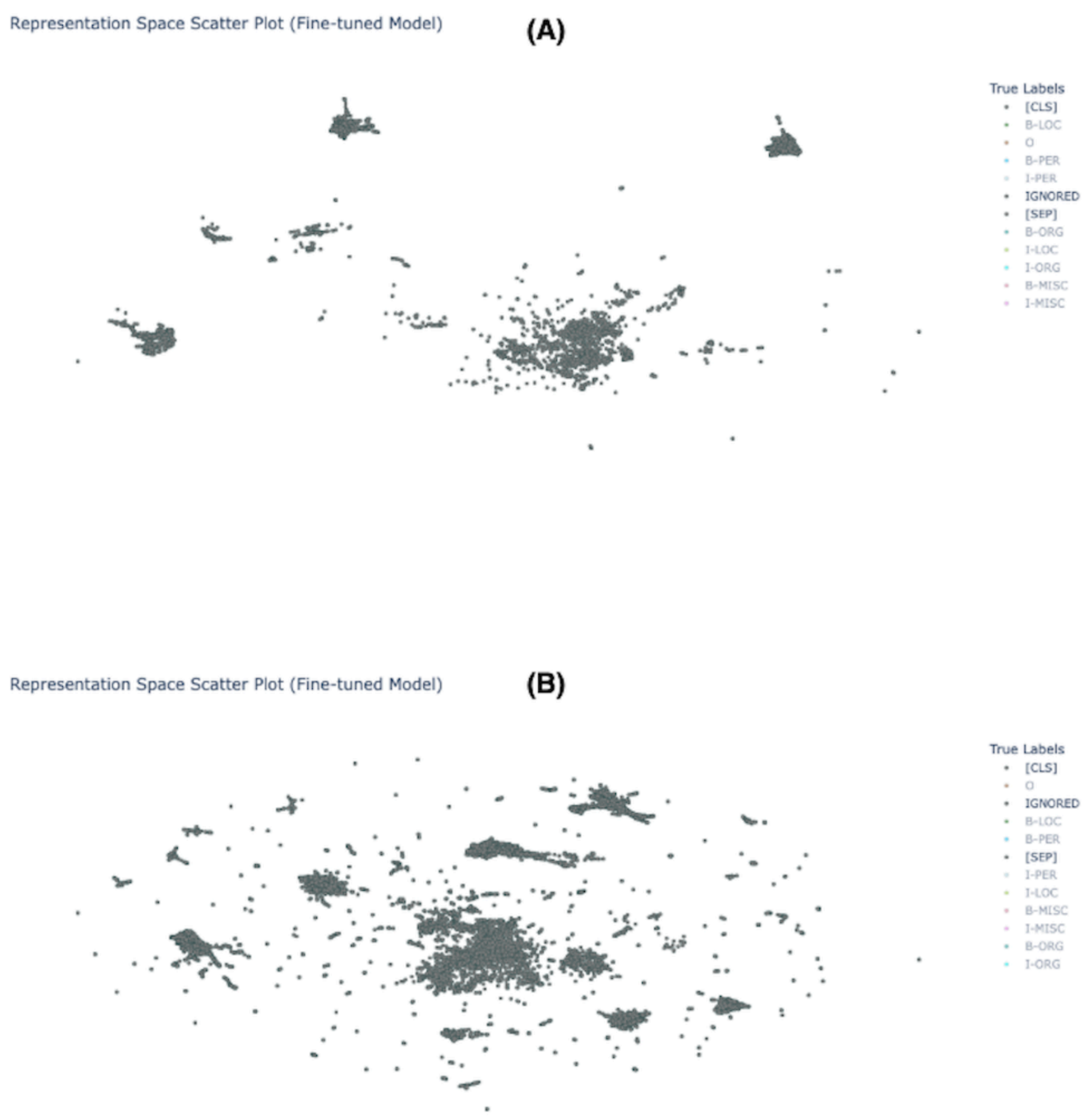

Figure 5.21: UMAP projection of IGNORED tokens (i.e., non-first subword tokens) after fine-tuning. (A) Arabic model. (B) English model. IGNORED tokens align closely with the core entity clusters.

Several patterns emerge from Figure 5.23. In both models, the lower layers (Layers 0–4) stay mostly unchanged, while the higher layers (especially Layers 8–11), closer to the task-specific output layer, show more noticeable differences. In English, many heads in Layers 9–11 show noticeable changes shifts, with similarity scores dropping below 0.8. In contrast, the Arabic model exhibits fewer heads with such low similarity, and the changes are more concentrated within specific heads. Overall, adaptation in English appears more widespread across layers and heads, whereas in Arabic it is more localised.

To understand whether the observed changes correspond to updates in the internal attention parameters, we also compare the model's attention weights (QKV matrices). While the attention score matrices capture how token-to-token focus shifts, the QKV matrices reflect changes in the underlying weights that generates these scores. As before, we flatten the QKV weights for each head and compute the cosine similarity between the pretrained and fine-tuned models. As shown in Figure 5.24, changes in the QKV weights are small for both languages, with most similarity scores remaining high. However, the distribution of these changes differs: in Arabic, the largest shifts are concentrated in the lower layers (especially Layer 0, Head 11), with moderate changes appearing in Layers 4, 7, and 8; in English, the most notable shifts also occur in the lowest layers (Layers 0–1), but more moderate changes are spread across middle layers (Layers 2–7). This suggests that while attention patterns shift significantly, the underlying attention parameters remain relatively stable, with differences in the distribution of changes across languages.

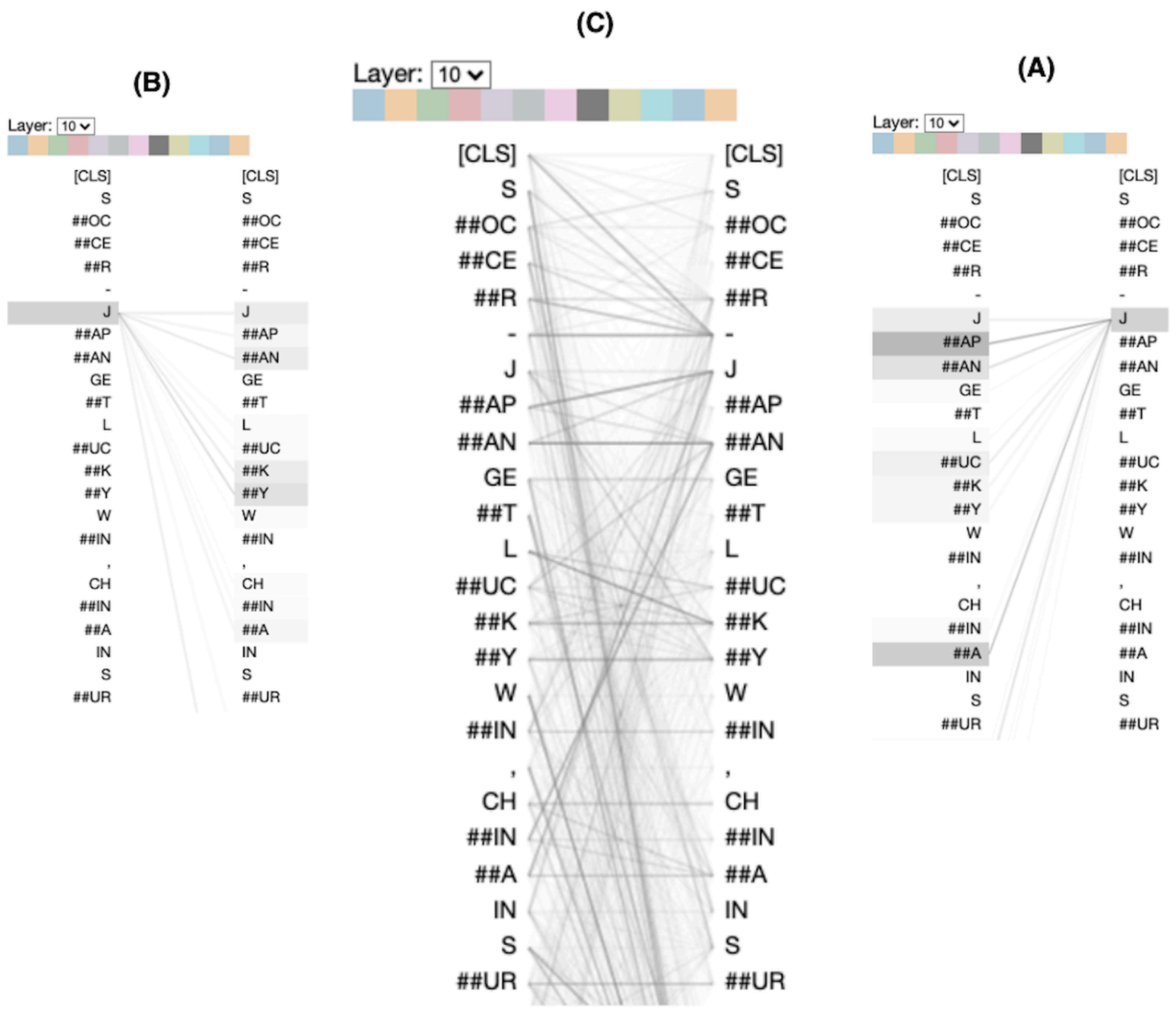

Figure 5.22: Attention score visualisation for Sentence 0 in the English test set at Layer 10, Head 8. (C) shows the full attention matrix; (A) highlights incoming attention to token "J"; (B) shows outgoing attention from "J".

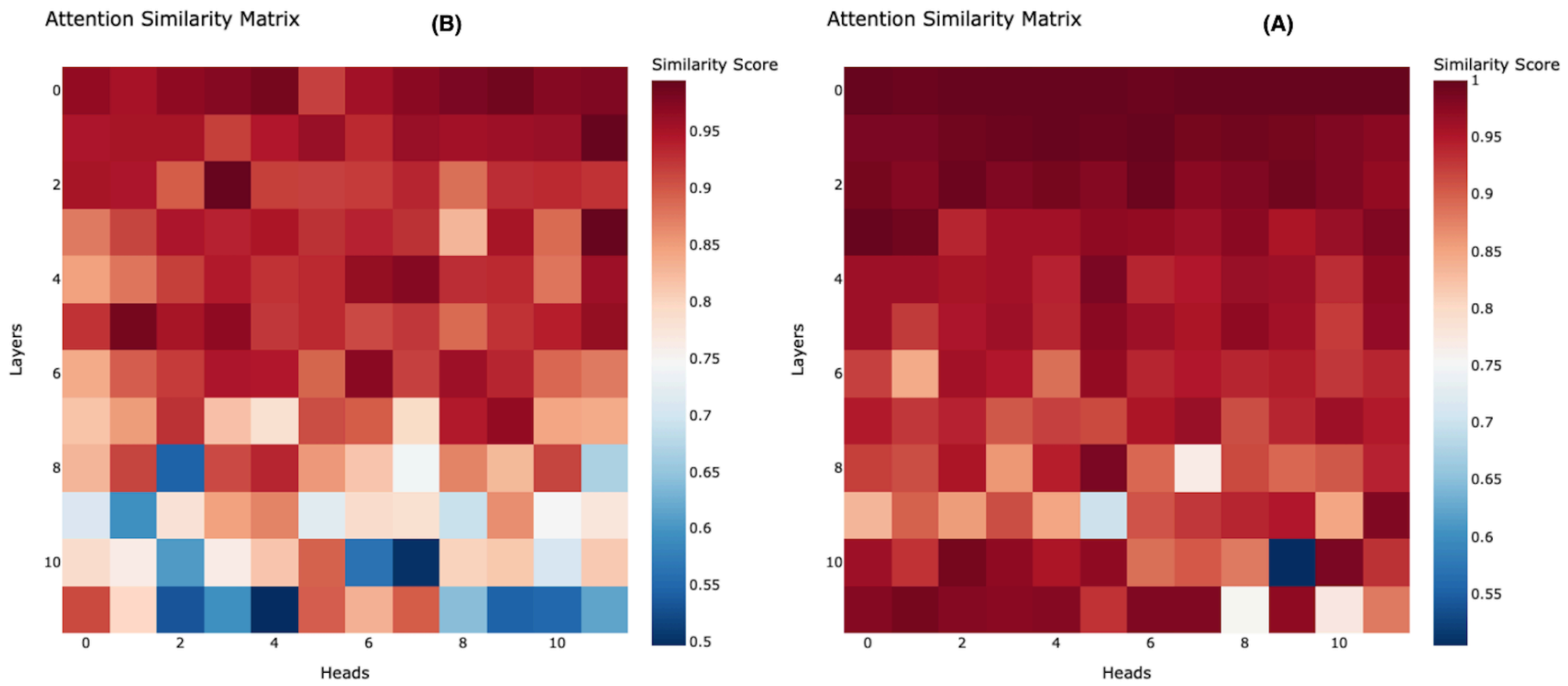

Figure 5.23: Attention similarity matrices for Arabic (A) and English (B). Each cell shows the average similarity between the attention score matrices in the pretrained and fine-tuned models for a specific head and layer.

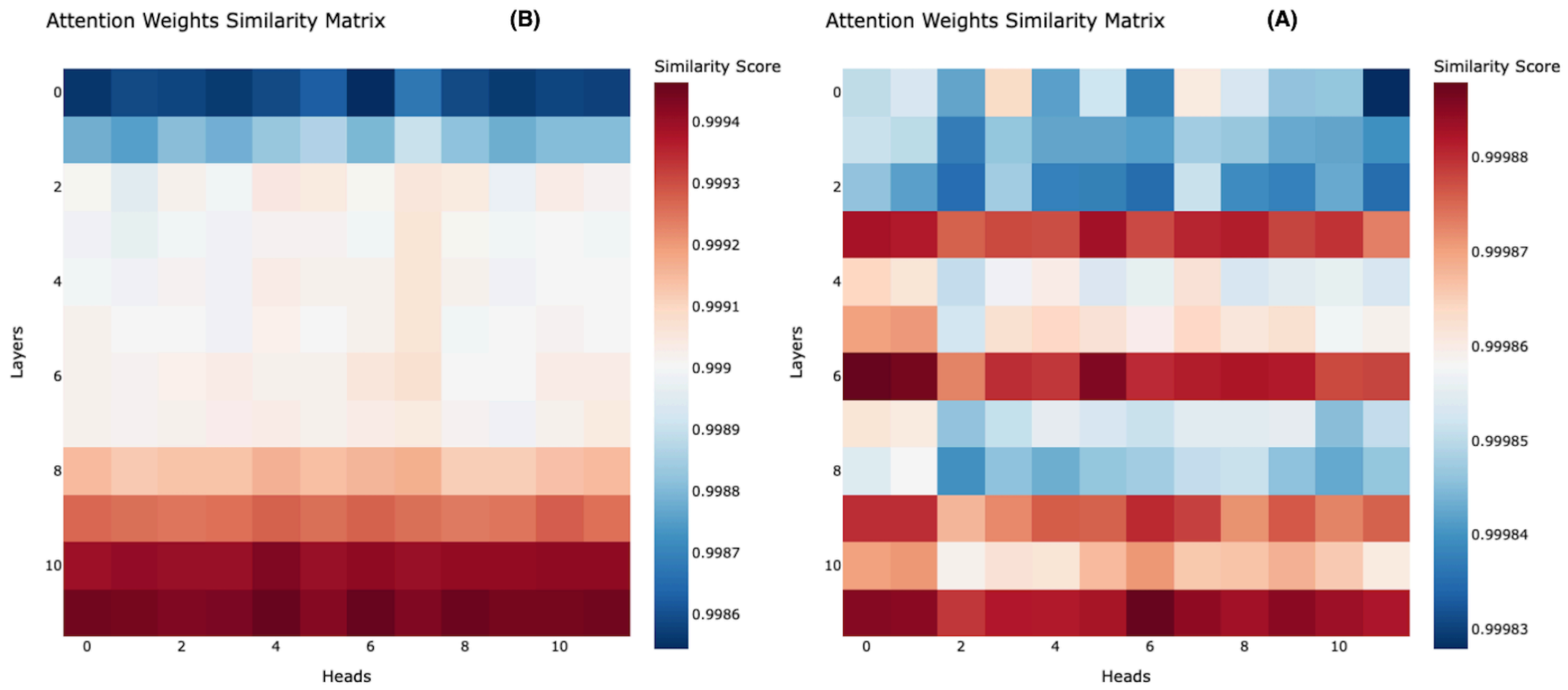

Figure 5.24: QKV weight similarity matrices for Arabic (A) and English (B). Each cell shows the similarity between pretrained and fine-tuned Q, K, and V matrices for a specific head and layer. Darker colours indicate higher similarity (less change), while lighter colours indicate more change in the learned attention parameters.

While the similarity heatmaps identify which heads were most affected by fine-tuning, they do not reveal how the attention behaviour itself changed. They highlight which attention heads were updated but do not show how the attention patterns shifted or what each head learned to attend to.

To better understand these changes, we select the first sentence in the test split for both Arabic and English and visualise their attention score matrices before and after fine-tuning. This helps illustrate how the model's behaviour changes with respect to actual token relationships—for example, which tokens receive more or less attention, whether attention shifts towards named entities, punctuation, or syntactic cues, and how these patterns differ across languages.

Figure 5.25 shows the first sentence from the Arabic and English test sets, along with their tokenised forms, gold-standard labels. These examples are used throughout the following analysis to illustrate how attention patterns and token representations evolve during fine-tuning.

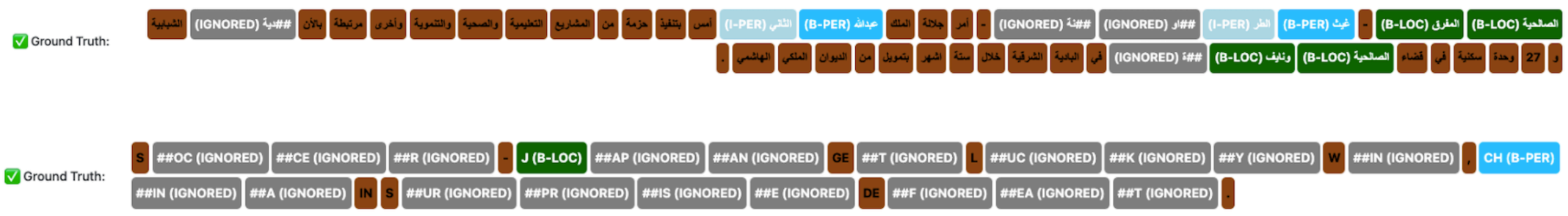

Figure 5.25: The first sentence from the Arabic (top) and English (bottom) test sets used for attention and token representation analysis. Tokens labelled with named entity tags are shown in colour.

Figure 5.26 shows the attention similarity matrices for Sentence 0 in Arabic and English. In Arabic, the most affected head is Layer 11, Head 8, while in English, it is Layer 10, Head 7. Note that these patterns differ from the average similarity matrix shown earlier in Figure 5.23, as sentence-level variations are smoothed out when averaging across the full test split.

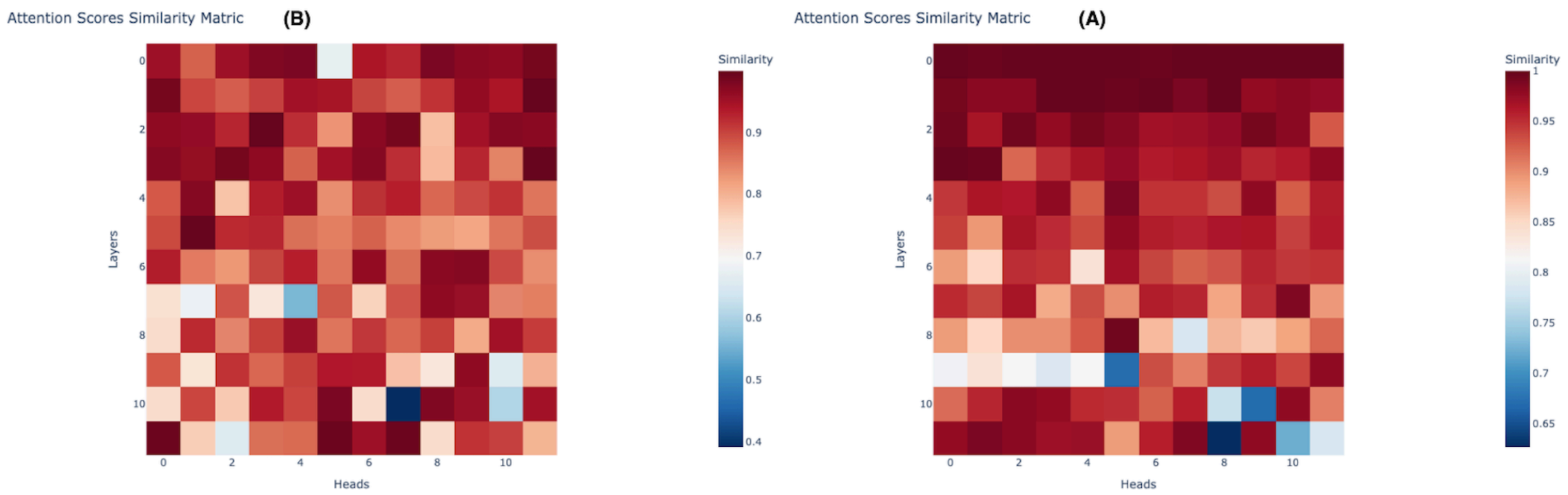

Figure 5.26: Attention similarity matrices for Sentence 0 in Arabic (A) and English (B). Each cell shows the cosine similarity between pretrained and fine-tuned attention score matrices for a given head. Lower similarity indicates stronger changes in attention behaviour for that specific head and layer.

Figure 5.27 shows the attention patterns for the most affected head in Arabic (Layer 11, Head 8). The pretrained model is shown on the left, and the fine-tuned model on the right. For each model, the central plot displays the full attention distribution across the sentence, while the two side plots provide focused views: the left side answers "Who do I attend to?" and the right side answers "Who attends to me?" for the selected token.

After fine-tuning, most tokens direct their attention toward the [SEP] token. This behaviour is known as a no-op pattern, where attention defaults to [SEP] when the model finds no strong focus within the input as discussed in Clark et al. (2019).

We also observe changes in the selected token's behaviour. In the pretrained model, the token attends both to the [SEP] token and to another word, while mainly receiving attention from its own subwords. After fine-tuning, attention becomes more localised: the token attends more strongly to its preceding word, to itself, and to its subword token, while still receiving attention from its subwords.

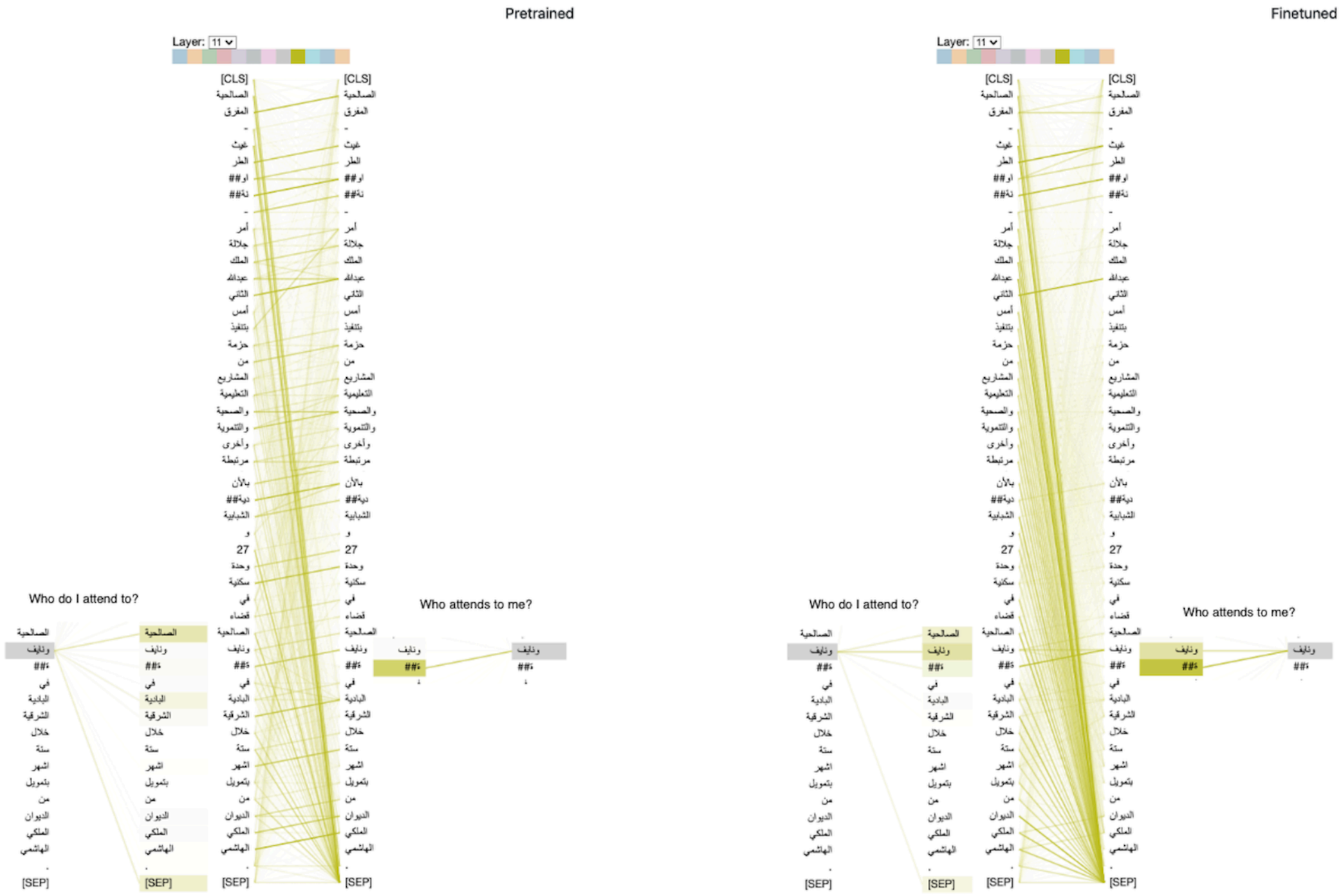

Figure 5.27: Attention patterns for sentence 0 in Arabic at Layer 11, Head 8—the most affected head in the similarity matrix. The pretrained model is on the left and the fine-tuned model is on the right. Each centre plot shows the full attention distribution for the sentence. The side plots highlight attention for a selected token (ونايف)—left side: ”Who do I attend to?”, right side: ”Who attends to me?”.

In English, we observe the opposite trend to Arabic. Figure 5.28 shows the attention patterns for the most affected head in English (Layer 10, Head 7).

In the pretrained model, attention is concentrated mostly on the [SEP] token, and the token “J” receives little attention from other tokens. After fine-tuning, attention becomes more distributed: “J” attends to several other subword tokens and also receives attention from subwords of other words.

This suggests that fine-tuning helps redirect the model’s attention away from generic patterns like [SEP] and toward more token-to-token relationships, a behaviour that is more noticeable in English compared to Arabic.

To further confirm these trends, we use the BERTviz model view to visualise all attention heads at Layer 11 for both Arabic and English, as shown in Figure 5.29. In Arabic, many token-to-token attention disappeared after fine-tuning, with attention focusing on the [SEP] token—except Head 8.

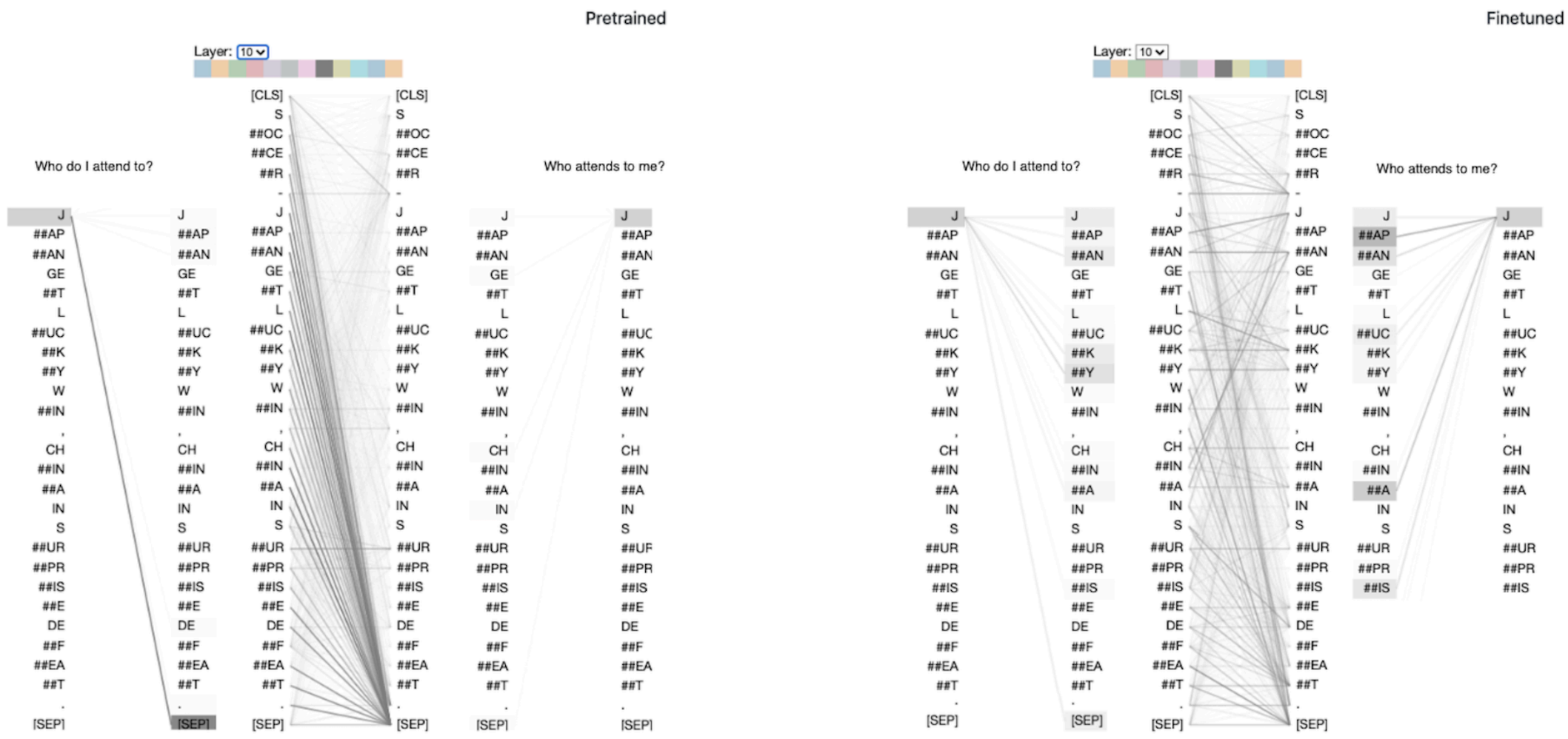

Figure 5.28: Attention patterns for sentence 0 in English at Layer 10, Head 7—the most affected head in the similarity matrix. The pretrained model is on the left and the fine-tuned model is on the right. Center: full attention map; Left: "Who do I attend to?"; Right: "Who attends to me?" for the selected token "J" from the word "JAPAN".

In contrast, while some heads in the English model still retain attention toward [SEP] after fine-tuning, other heads introduce more distributed token-to-token attention, suggesting a richer, task-specific contextualisation across the sentence.

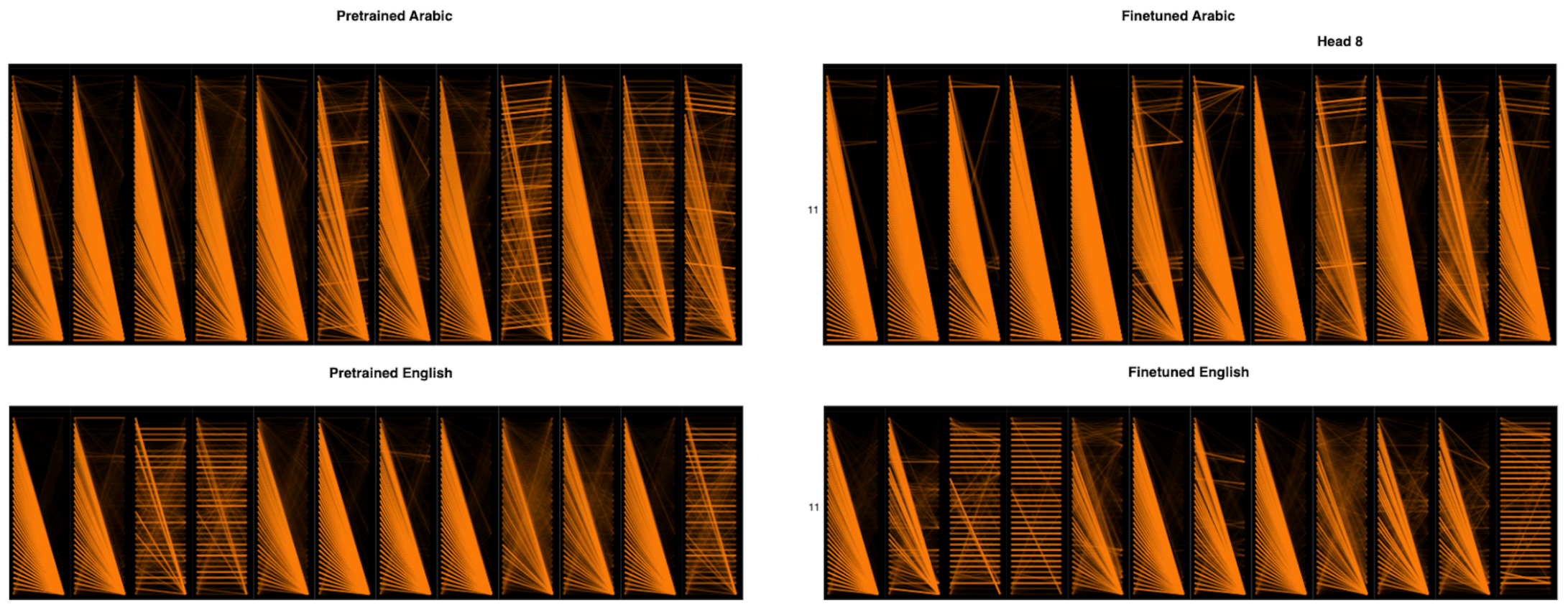

Figure 5.29: Model view of Layer 11 across all heads for Arabic (top) and English (bottom), comparing pretrained (left) and fine-tuned (right) models.

### 5.4.3 Evolution of Token Embeddings

Finally, we examine how token representations evolve across layers by comparing the vectors at three key stages: the input layer (raw word embeddings), an intermediate layer (Layer 6), and the final contextual layer (Layer 11). For each layer, we compute the cosine similarity between the pretrained and fine-tuned token representations to assess how much fine-tuning alters the underlying structure. Figure 5.30 shows the results for the 15 tokens exhibiting the largest changes, separately for Arabic (top) and English (bottom). Each bar represents the similarity score at each stage. For tokens appearing multiple times, we report the average similarity across instances.

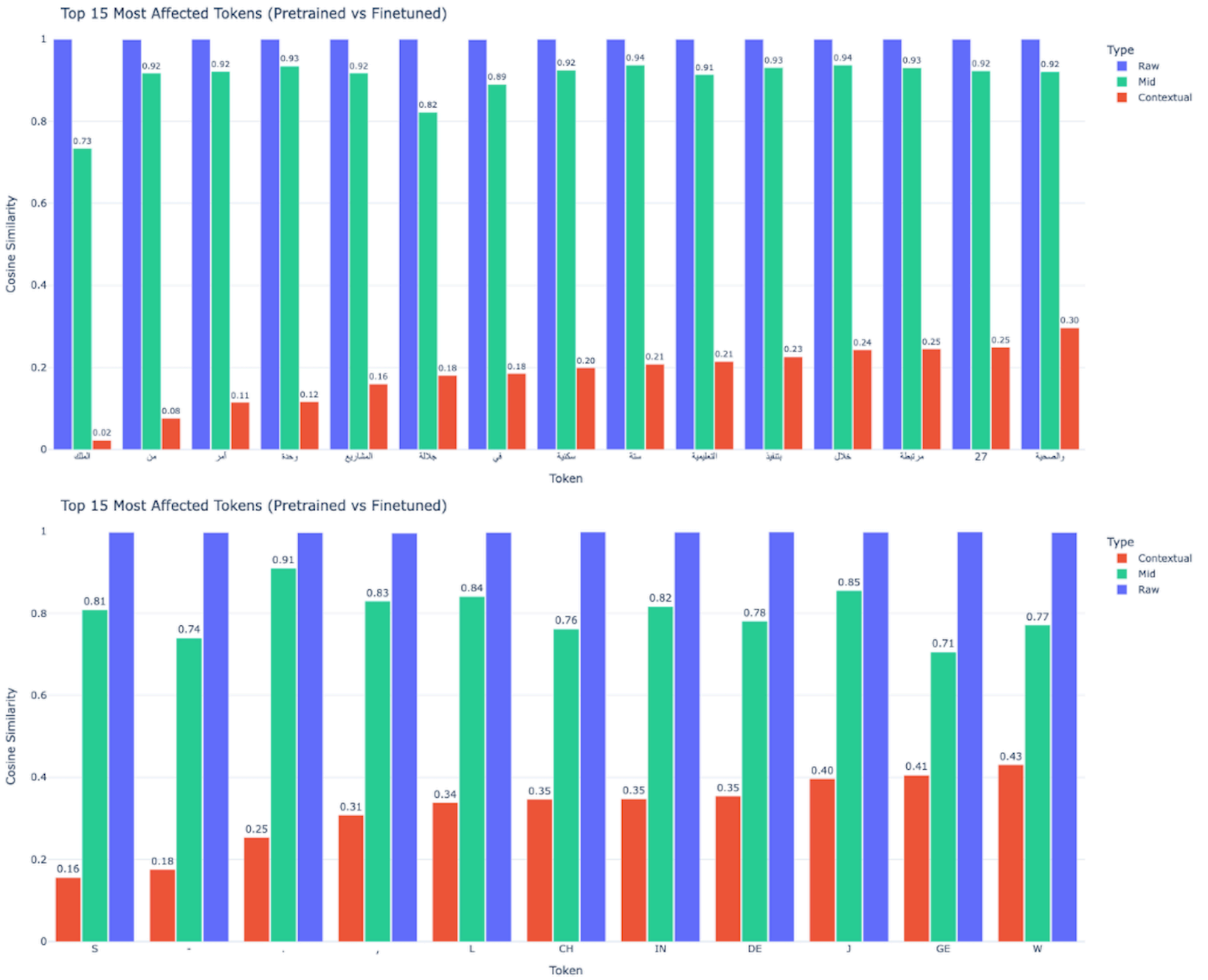

Figure 5.30: Cosine similarity between token representations before and after fine-tuning for the top 15 most affected tokens in Arabic (top) and English (bottom), based on Sentence 0.

Across both languages, the raw input embeddings remain mostly unchanged, with cosine similarities close to 1.0. This is consistent with the earlier observation that fine-tuning minimally updates the model parameters compared to the deeper layers of the model. However, more noticeable changes appear at the intermediate and final layers. In particular, Arabic tokens show larger shifts at the final hidden state (Layer 11), with some similarity scores dropping as low as 0.02. English tokens also experience changes, but their similarity scores generally remain higher overall.

This pattern suggests that fine-tuning has a stronger impact on Arabic token representations at the contextual level. It also supports the idea that the observed changes in attention behaviour result from a combination of both updated attention weights and evolving token representations. Since each layer builds upon the output of the previous one, small changes at lower layers accumulate and increase at higher layers, leading to more divergence in the final contextual representations.

This observation raises an important question: why are these particular tokens the most affected, and why does Arabic appear more affected than English? The degree of representational change depends on several factors: how the token was represented during pretraining, the contexts in which it appeared, and how those contexts diverge in the fine-tuning data. Fine-tuning affects not only the token itself but also its relations with surrounding tokens, with small parameter updates accumulating into shifts in the final contextual representation. While we cannot directly trace how tokens were represented in the pretraining corpus, comparing the most- and least-affected tokens provides useful insights.

To explore this, we examined the tokens with the largest and smallest shifts. In Arabic, least-affected tokens such as والصحيه (”and health related”) are out-of-vocabulary and occur only in the test split, meaning they receive little or no adjustment during fine-tuning. Their semantics are relatively generic (e.g., “health-related”), and the context in which they appear may be similar to the one captured during pretraining, leading to stable representations. By contrast, highly affected tokens such as ملك (“King”) are frequent and appear across both training and test splits, as well as in multiple entity types. This potentially forces fine-tuning to reshape their representations more substantially in order to fit the supervised label structure. Moreover, these tokens often appear in more specific contexts (e.g., references to particular monarchs and countries in a newswire context), which may diverge from their more general pretraining usage.

In English, the pattern differs. Both least- and most-affected tokens occur frequently in both training and test data. Least-affected tokens such as “Win” appear to retain broad, generic semantics that are stable across contexts, meaning their representation potentially remains close to the pretrained form. In contrast, words like “Scorer” undergo larger shifts because their fine-tuning contexts (e.g., domain-specific sports references) potentially diverge more strongly from their broader, pretraining contexts.

These observations and interpretations are not definitive conclusions but informed hypotheses, since we cannot directly observe how these tokens were represented in the pretraining data. They are inferred from their behaviour during fine-tuning. Overall, the findings suggest that representational change is influenced by multiple interacting factors—not only token frequency, but also training exposure, contextual variability, and the degree of alignment between pretraining and fine-tuning semantics. These explanations remain hypothesis-driven, but they illustrate how variation in token usage and context can explain why some tokens shift more than others.

### 5.4.4 Answer to RQ5: Training Impact

This section addressed aspects **(a)** and **(b)** of the training impact component of RQ5, focusing on how fine-tuning alters internal attention behaviour and representational structure in Arabic and English NER models. The key findings are summarised below for each aspect.

**Aspect (a): Effects of Fine-Tuning on Attention Behaviour and Token Representations** Fine-tuning leads to substantial changes in both the structure of token representations and the internal attention behaviour of the model. The representation space becomes more organised, with clearer separation between entity tags, although the degree of separation differs across languages. Attention patterns (relationship between tokens) also change, particularly in the higher layers (near the output layer), with English showing more distributed changes than Arabic. While the learned attention parameters (QKV matrices) remain relatively stable, changes in attention scores indicate significant adjustments in the contextual relationships between tokens. Finally, token representations evolve across layers, with the highest changes occurring at the final contextual layer. Together, these results show that fine-tuning not only changes token-level information but also reorganises how tokens interact internally, through the accumulation of contextual changes.

**Aspect (b): Cross-Linguistic Differences in Fine-Tuning Behaviour** The analysis reveals important differences in how fine-tuning affects Arabic and English models. English models exhibit widespread, task-aligned changes across layers and attention heads, while Arabic models show more localised and sometimes unstable behaviour, including collapsed attention patterns and greater shifts in final token representations. These differences suggest that internal model changes are language-specific, motivating the need for separate interpretability analysis for Arabic and English because trends observed in English cannot be assumed to transfer directly to Arabic.

**Conclusion** One plausible explanation for these differences could be the larger size of English training data, which supports more distributed and robust adaptation. Another possibility is that the Arabic model relies on simple heuristics to solve the task without requiring deep contextualisation, potentially indicating a degree of over-fitting to surface patterns. This further highlights the need to analyse and validate model behaviour separately across languages. Moreover, while attention analyses are common for English, they remain underexplored for Arabic, presenting an important direction for future work (see Section 6.6). Building on these findings, the next section investigates patterns and anomalies in the global representation space of the fine-tuned model.

## 5.5 Patterns and Anomalies

In this section, we examine how patterns and anomalies emerge within the model's learned representation after fine-tuning and how these differ across Arabic and English. We address aspects (a)–(f) of the *Representation component of RQ5*, combining global structural inspection with token-level behavioural metrics to explore the internal organisation of token embeddings and their relationship to model behaviour.

We begin with a high-level exploration of the representation space, identifying broad patterns that characterise model behaviour in both languages. We then turn to a set of targeted investigations focused on Arabic, including the impact of tokenisation, anno-tation scheme misalignments, and error-specific patterns that affect model calibration, prediction confidence, and representation quality. By integrating token-level metrics with instance-level analysis, we uncover patterns and anomalies that deepen our understanding of language-specific discrepancies.

### 5.5.1 High-Level Analysis of Representation Space

In this section, we focus on examining the structure of the representation space within each language independently. For each model, we analyse how tokens are organised after fine-tuning and identify the main structural patterns that emerge. By characterising these patterns, we capture the underlying structure within the Arabic and English models. Finally, we compare the observed structures across the two languages to highlight similarities, differences, and language-specific anomalies.

**Arabic Representation** Figure 5.31 presents the token Representation Space Scatter Plot for the Arabic model after fine-tuning. A clear distinction can be observed between two types of regions: dense regions and scattered regions. The dense regions are mainly composed of entity-tagged tokens or the dominant O class. These regions exhibit different characteristics depending on the entity tag, with some entity spans forming tighter, more compact structures than others (B/I-PER vs B/I-MISC). In contrast, scattered regions are distributed more across the space and are mainly associated with O-tagged tokens. However, there are also isolated scattered regions that belong to specific entity tags.

This suggests that the representation structure is not determined only by membership in an entity tag or main class. Some regions appear to form based on other properties, such as how tokens are used within the sentence, or due to semantic relationships between tokens. These patterns can be summarised as follows (all examples refer to the Arabic representation in Figure 5.31): (1) Region (C) corresponds to punctuation tokens (e.g., periods), forming a small isolated cluster; (2) Region (N) captures numerical tokens, positioned separately from the main dense regions; (3) Region (W) contains semantically similar tokens related to awards (e.g., Nobel Prize, Wimbledon Prize), illustrating local semantic structure; and (4) Region (L) reflects language-specific behaviour, where tokens are grouped together due to spelling variations or errors.

These observations highlight the general structure of the Arabic model's token representations after fine-tuning, revealing a mixture of entity-driven organisation and semantic or linguistic factors shaping the representation space.

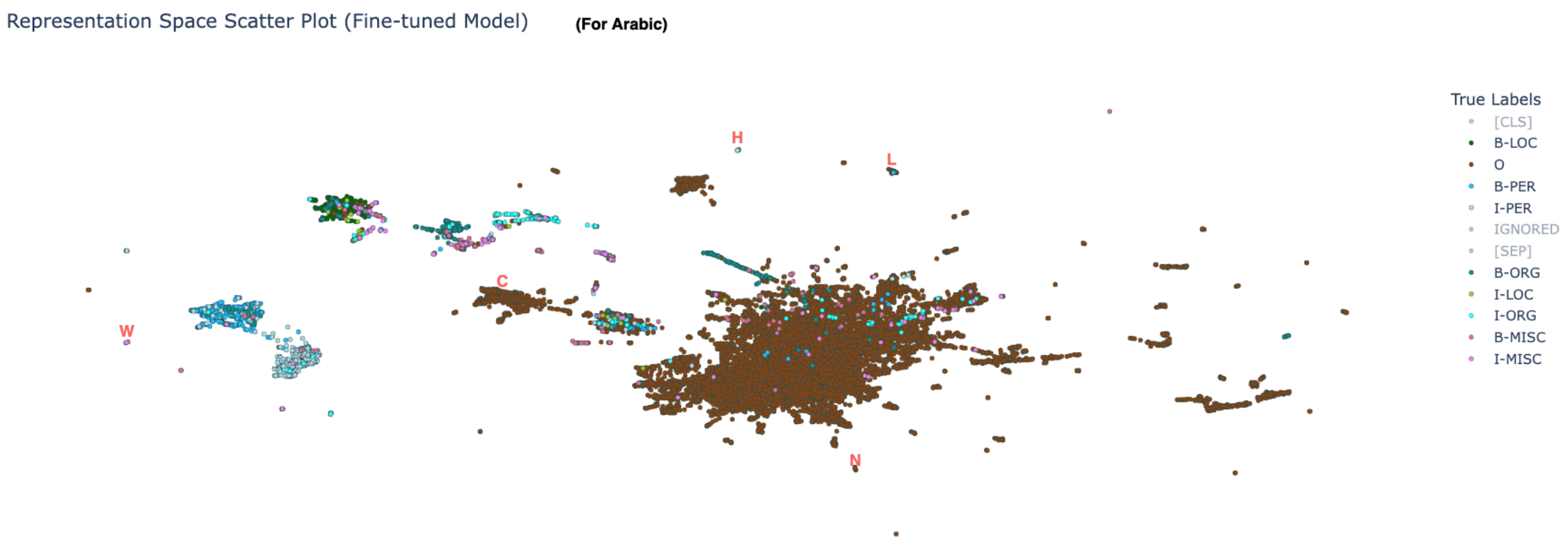

Figure 5.31: Representation Space Scatter Plot for the Arabic model after fine-tuning. The figure shows the token-level organisation after fine-tuning, with several highlighted regions: (W), (C), (N), (L), and (H). These regions mark areas of interest corresponding to different grouping behaviours within the representation space.

**Arabic Spelling Variation Example** To illustrate the spelling variation pattern observed in Figure 5.31 (Region L), we present two examples from that region using the instance level analysis tab. Figure 5.32 presents two sentences containing the words علي (Ali) and على (On). While both words share similar spelling, the word علي (Ali) in Arabic is a proper noun, ending with the letter ي (Yaa) which has two dots underneath. In contrast, the word على (On) ends with the letter ى (Alif Maqsura), which does not have dots. This highlights a common spelling error where both forms are written with dots, causing ambiguity between the two words.

In the examples shown, the first token علي (Ali) is annotated as a B-PER (person name) tag, while the second token على (On) is annotated as O (non-entity). However, due to the spelling variation, both tokens are placed together within the same region (L) in the representation space due to confusion between the two spelling variations. It is also worth noting that this example contains another anomaly: the word علي (Ali) appears abruptly at the beginning of the sentence without clear syntactic structure, a pattern discussed further in Section 5.5.3.

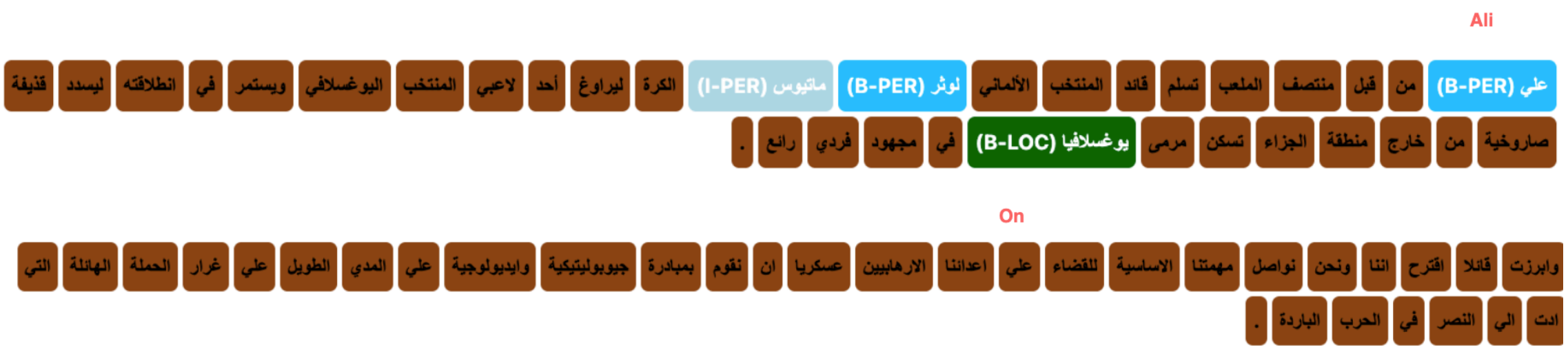

Figure 5.32: Example of Arabic spelling variations captured in the representation space. Highlighted tokens are associated with two words علي (Ali) and the word على (On). The tokens علي (Ali) and على (On) are highlighted in the figure.

**English Representation** Figure 5.33 presents the token Representation Space Scatter Plot for the English model after fine-tuning. As with Arabic, a clear distinction can be observed between two types of regions: dense regions and scattered regions.

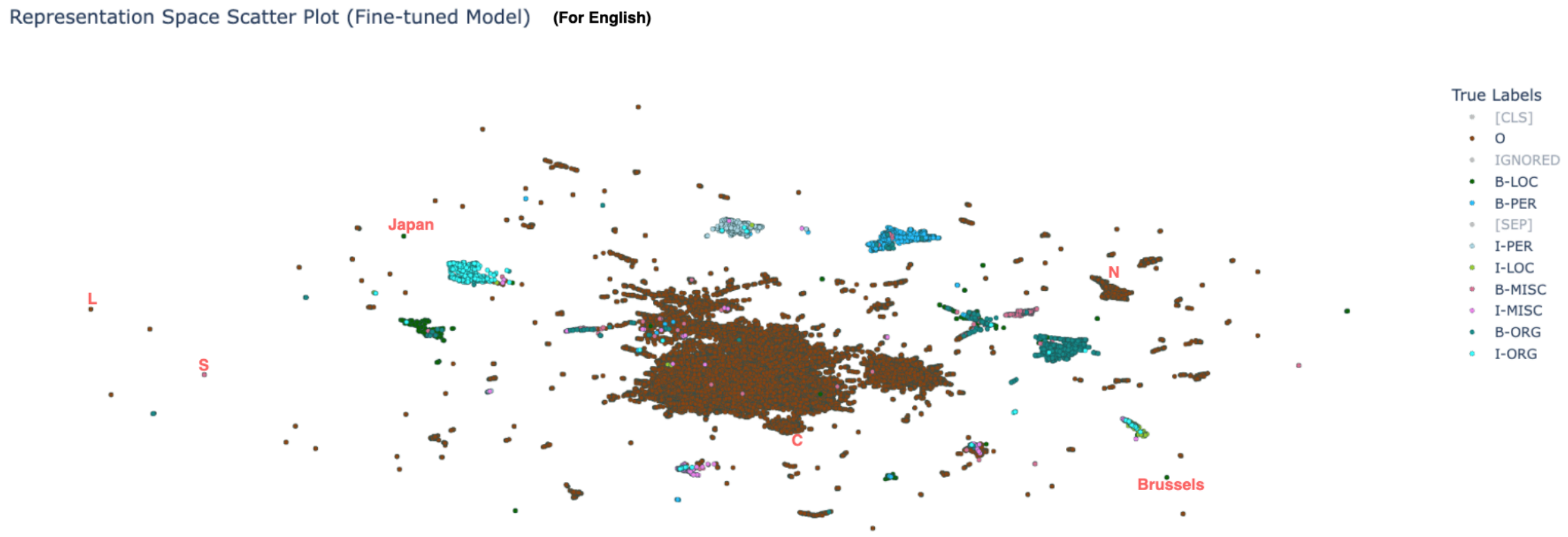

Figure 5.33: Representation Space Scatter Plot for the English model after fine-tuning. The figure shows the token-level organisation after fine-tuning, with several highlighted regions: (C), (N), (S), and (L). These regions mark areas of interest corresponding to different grouping behaviours within the representation space.

The dense regions are mainly composed of entity-tagged tokens or the dominant O class. In English, entity tag regions are generally more separated across entity tags compared to Arabic . In particular, spans corresponding to PER, LOC, and ORG tags form distinct and well-separated regions. Furthermore, the model distinguishes both between entity types (e.g., B-LOC vs. B-ORG) and within entity boundaries (e.g., B-LOC and I-LOC tokens show clear separation).

With respect to the scattered regions, we observe similar patterns to those seen in Arabic. Specifically, (1) Region (C) corresponds to punctuation tokens (periods); (2) Region (N) captures numerical tokens; and (3) Region (S) contains semantically similar tokens, forming small isolated clusters. However, unlike Arabic, we could not find clear language-specific clusters. Instead, most scattered points in English form relatively homogeneous micro-clusters based on clear criteria, such as identical tokens, similar contexts, etc.

We also found an interesting pattern in the English model that explains usage variation: not all tokens belonging to the B-LOC class are embedded within the main B-LOC cluster. In particular, tokens such as "Japan" and "Brussels," highlighted in Figure 5.33, appear in distant scattered regions. We have examined these points and found that these tokens were used differently compared to typical instances of country names — for example, in the case of "Japan," the region corresponds to sentences listing rankings of Japanese players in a tournament as shown in the next example.

**English Different Context Example** Figure 5.34 presents two English sentences that exhibit a consistent pattern. Both sentences appear to be part of a ranked list, starting with a numbered position, followed by the player's name (annotated as B-PER and I-PER), their country (annotated as B-LOC), and then some points. This structured format, repeated across sentences, reflects a list-like structure that was captured by the model during fine-tuning. Such contextual pattern influences the representation of tokens like "Japan" in the embedding space, separating them from other occurrences of the same token used in different contexts.

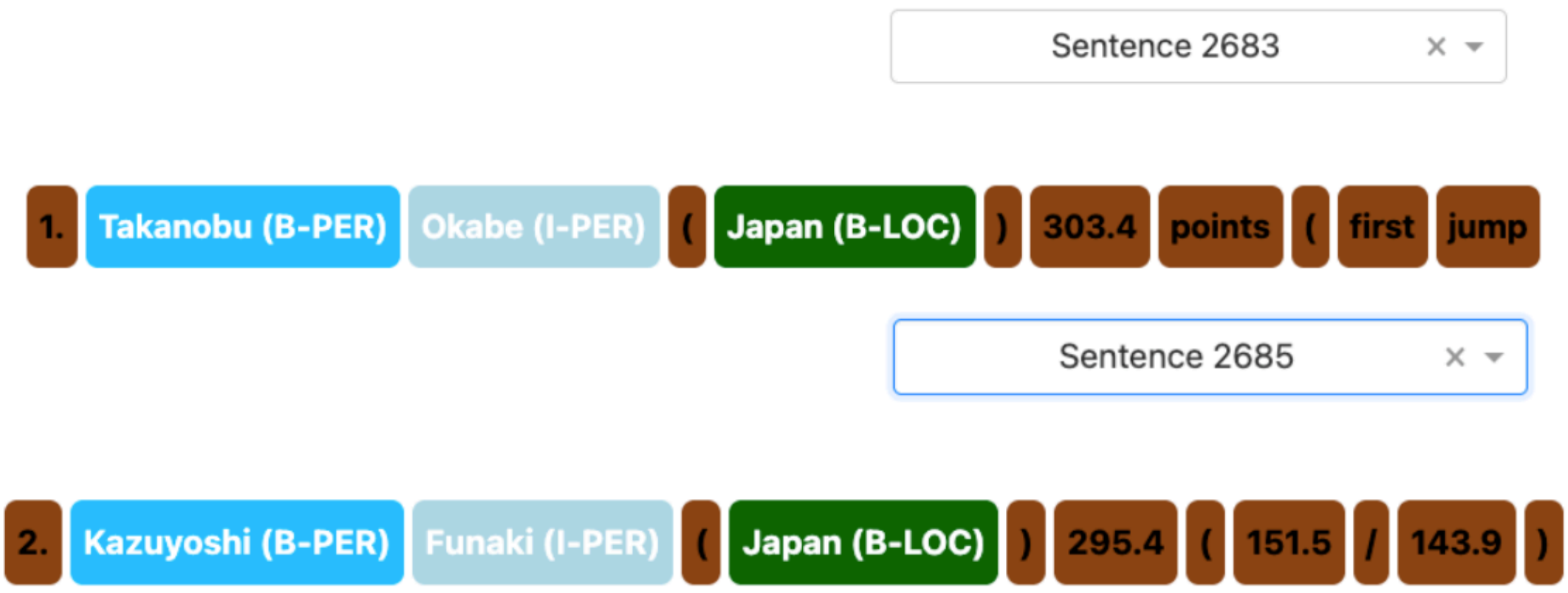

Figure 5.34: Examples of different contextual usage for the token "Japan" in the English dataset. Both sentences follow a structured list-like format, starting with the player's name (B-PER/I-PER), followed by the country (B-LOC), and numerical information information.

It is worth noting that despite shuffling sentences during training, these list-like structures were still captured by the representation space, indicating a strong influence of usage context. This highlights that token structure in the representation space is influenced not only by entity membership but also by broader semantic context and sentence structure. Similar behaviour can also be observed in Arabic — for example, point (H) in Figure 5.31 — although in Arabic, the clusters corresponding to such patterns are often more heterogeneous and are affected by the structural properties of the data.

**Errors and Anomalies Across Languages** Beyond the general dense and scattered regions observed in the representation space, a more detailed pattern emerges within the dense regions themselves. Each dense region mainly consists of a group of tokens belonging to the same entity tag — for example, B-LOC tokens clustering together. However, within these dense clusters, we observe two types of deviations across the two languages.

First, there are errors within the English B-LOC region, where tokens belonging to a different entity tag appear within the wrong cluster — for example, a B-MISC or O token located inside a B-LOC region. These errors are marked as diamonds in Figure 5.35.

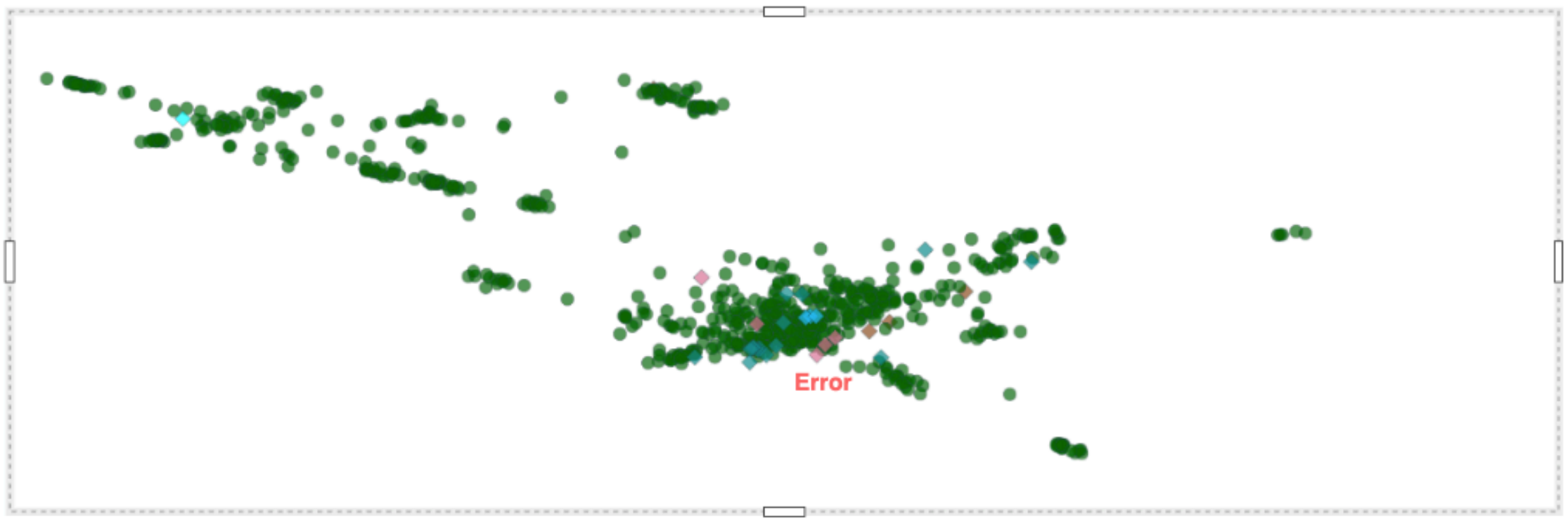

Figure 5.35: Examples of error pattern within B-LOC for the English model. Diamonds indicate errors where tokens belonging to different entity types appear incorrectly within the main B-LOC region. Note that this area is highlighted within the B-LOC region using box selection.

Turning to Arabic, Figure 5.36 shows a similar pattern of tagging errors within the B-LOC region. Arabic also exhibits another anomaly: tokens that fall outside their expected cluster but are still predicted correctly. For example, an O token appears near the B-LOC cluster yet is still classified as O (highlighted as Anomaly a). This suggests that Arabic representations contain structural overlaps that warrant further investigation, as discussed in Section 5.5.3.

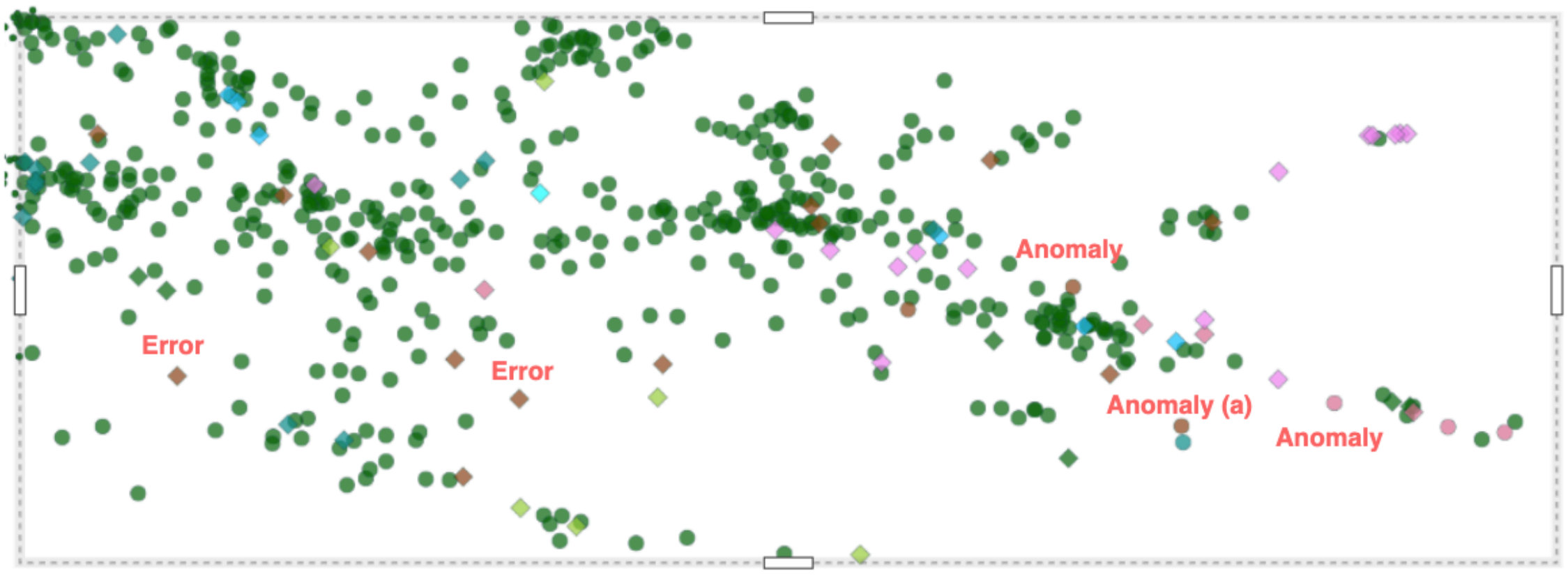

Figure 5.36: Examples of errors and anomalies within B-LOC regions for the Arabic model. Diamonds indicate errors where tokens belonging to different entity types appear incorrectly within the main B-LOC cluster. Circles indicate anomalies where tokens from different entity types are positioned outside their expected cluster but are still correctly classified, reflecting language-specific structural patterns.

This highlights that while both Arabic and English models experience classification errors within dense regions, Arabic representations additionally exhibit more complex boundary behaviours, where cross-tag relationships need to be interpreted in more details.

**Summary** In summary, the analysis of the high-level representation space reveals that not all regions are formed only based on entity membership. Instead, contextual usage, semantic relationships, and linguistic factors also contribute to the structure of tokens within the space.

Across both languages, we observe differences in separability at the span level, the entity type level, and the boundary level (between B- and I- tags), consistent with the trends identified in Chapter 4. In English, entity spans show clearer and more distinct boundaries, while in Arabic, overlaps between types and boundaries are more frequent.

Furthermore, we found evidence of language-specific regions in Arabic, corresponding to linguistic variations and spelling inconsistencies (e.g., regions highlighted in Figure 5.31 - L), which are less prominent in English.

In addition to general structural trends, our analysis of dense regions revealed two types of deviations: tagging errors, which occur in both languages, and language-specific anomalies observed only in Arabic. The latter represent cases where tokens from different entity tags are correctly predicted but positioned within another entity tag group, highlighting more complex cross-tag relationships in Arabic.

To complement the structural analysis, we now turn to a different perspective by examining token-level behaviour using behavioural metrics. This approach allows us to investigate how the patterns observed in the last hidden state representations are reflected across various subcomponents of the model and data.

### 5.5.2 High-Level Analysis of Behavioural Metrics

In this section, we focus on analysing behavioural metrics to provide high-level insights into how different components of the model and data relate to the observed representation structures. We begin by examining the correlations between various behavioural metrics, and then identify interesting patterns to explore throughout this section.

**Correlation Analysis** Starting with the Pearson correlation matrices shown in Figure 5.37, we observe that the overall trends between behavioural metrics are almost identical across Arabic and English. For example, token ambiguity is positively correlated with consistency ratio, while token confidence is negatively correlated with prediction uncertainty. However, subtle differences appear in the degree of these relationships. In most cases, English tends to show stronger correlations than Arabic — for instance, the relationship between loss values and true silhouette scores is stronger in English. We first investigate these cross-lingual distinctions in more detail by focusing on specific relationships between key behavioural metrics.

**Loss Values vs True Silhouette** Figure 5.38 presents the relationship between loss values and true silhouette scores across Arabic and English models. The solid lines represent the average trend for each metric to compare the divergence (see Section 4.5.6).

Both languages exhibit a clear general trend: as silhouette scores increase, loss values decrease, confirming that more compact and separated representations are associated with lower prediction errors. However, subtle differences emerge between the languages. In English, the relationship remains sharper and more focused, with points closely concentrated around the trend line. In Arabic, the relationship becomes noisier: while loss values decrease as silhouette increases, there is higher spread and noise in both metrics. To further understand this divergence, we examine the distribution of errors across the silhouette–loss relationship. Figure 5.39 presents the scatter plot of loss values against true silhouette scores, with points coloured by error type.

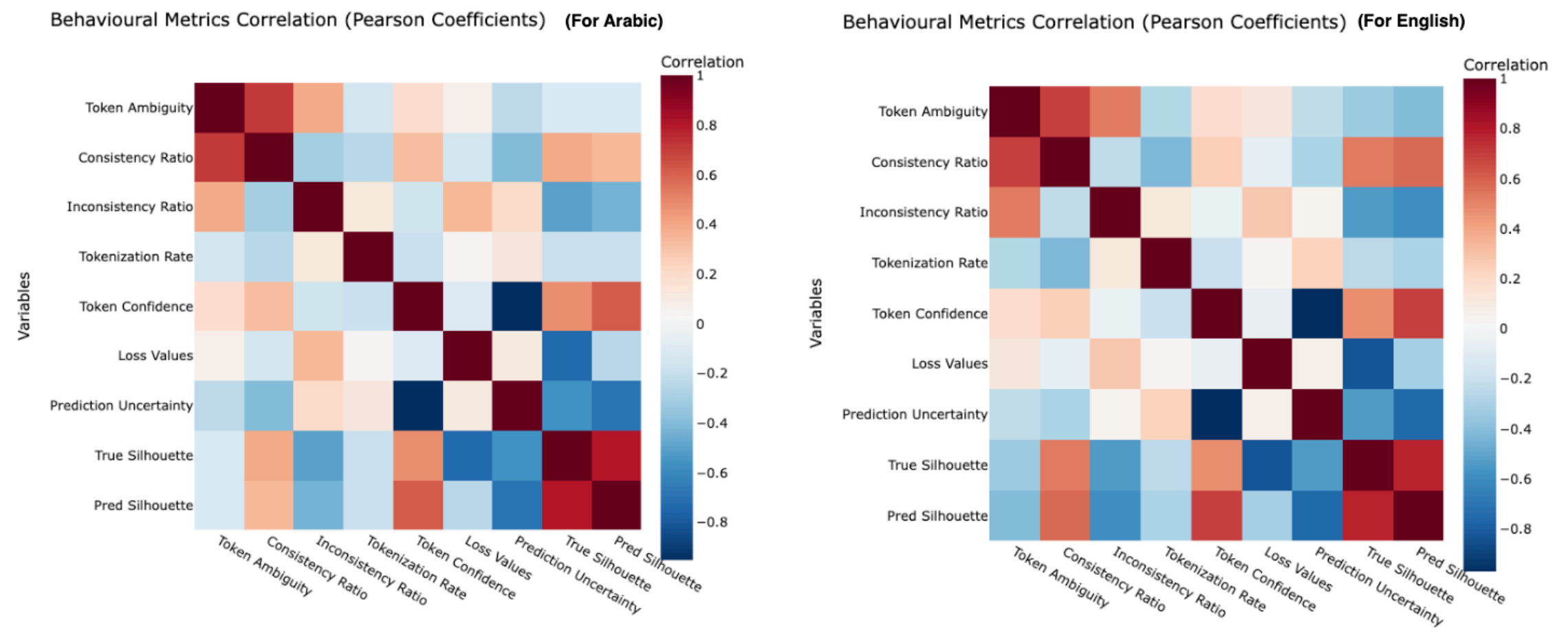

Figure 5.37: Pearson correlation matrices between behavioural metrics for Arabic (left) and English (right).

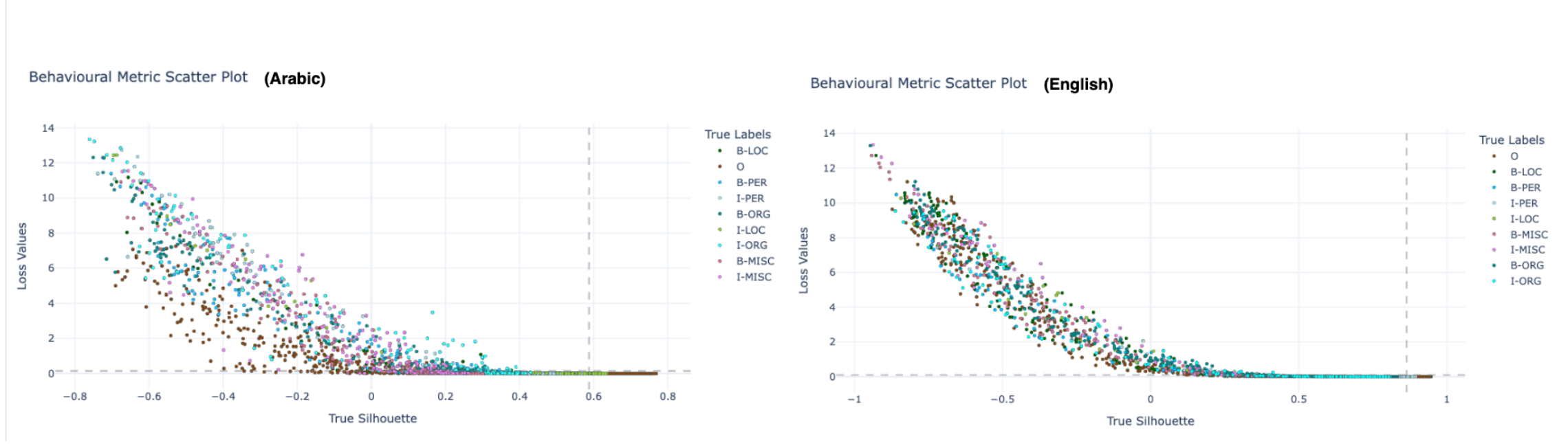

Figure 5.38: Scatter plots showing the relationship between loss values and true silhouette scores for Arabic (left) and English (right).

As shown in Figure 5.39, errors begin to appear more frequently as silhouette scores decrease, particularly below 0.3 in Arabic, whereas in English this trend is delayed until around 0.1. In Arabic, the distribution of error types is more scattered, whereas in English, errors are more concentrated and tightly linked to lower silhouette scores. This indicates that representation structure and prediction correctness become misaligned earlier in Arabic. Errors start to appear at lower silhouette scores, but the corresponding loss values are not always high, indicating that low-loss predictions are not always aligned with compact and separated representations. In contrast, in English, errors are more clearly associated with higher loss and lower silhouette scores.

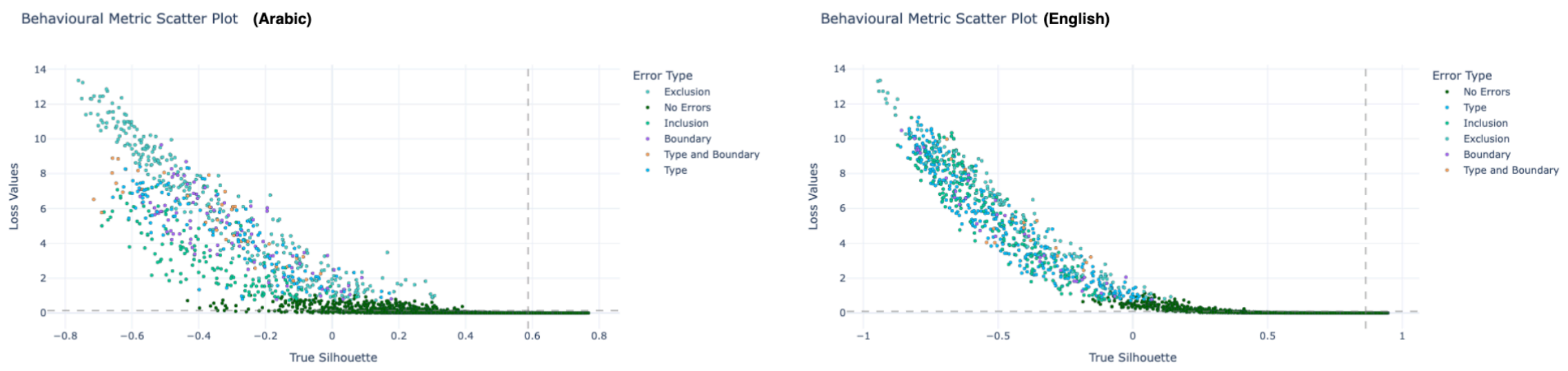

Figure 5.39: Scatter plots showing the relationship between loss values and true silhouette scores, with points coloured by error type, for Arabic (left) and English (right).

These findings suggest that while the global trend between structure and loss holds across languages, the Arabic model exhibits noisier behaviour: structure quality and prediction correctness are less consistently aligned, and this suggest that Arabic contains more irregular error patterns compared to English.

Overall, while the Pearson correlation analysis indicated similar high-level relationships across languages, the detailed investigation of loss values and structure quality reveals that differences in magnitude capture subtle distinctions in language behaviour.

Switching to Spearman correlations, Figure 5.40 presents the Spearman correlation matrices between behavioural metrics for Arabic and English. As with the Pearson correlations, the general trends remain similar across both languages. Behavioural metrics maintain consistent relationships, although some differences in correlation strength persist.

One relationship that was not captured by Pearson correlation but becomes clear with Spearman correlation is between prediction uncertainty and true silhouette scores. While Pearson shows no strong linear association, Spearman reveals a strong negative monotonic trend, with coefficients of about $-0.89$ for Arabic and $-0.71$ for English. This indicates that as clustering compactness and separation improve (higher silhouette scores), prediction uncertainty consistently decreases across both languages. This relationship is examined in more detail in the following section.

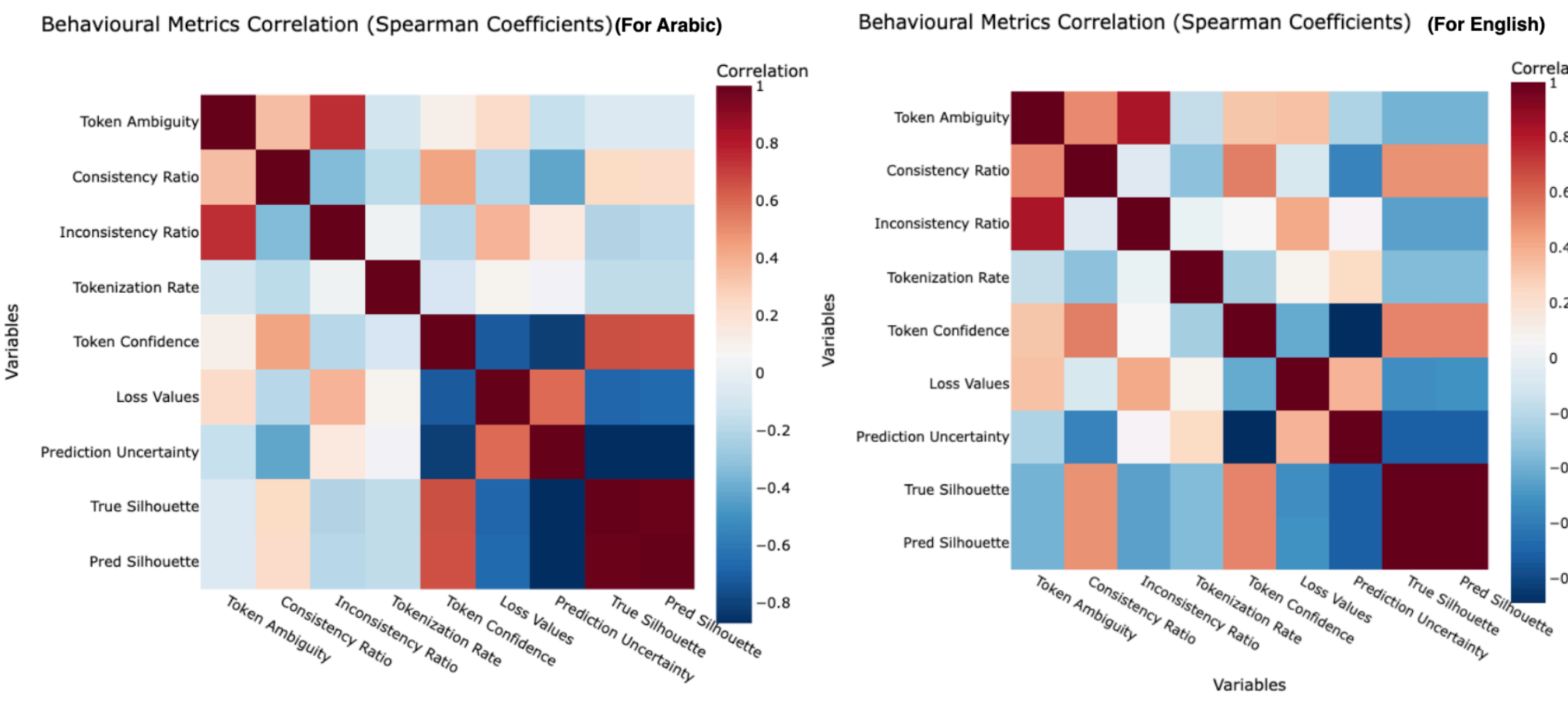

Figure 5.40: Spearman correlation matrices between behavioural metrics for Arabic (left) and English (right).

**Prediction Uncertainty vs True Silhouette** Figure 5.41 shows the relationship between prediction uncertainty and true silhouette scores for Arabic and English, with points coloured by error type. As expected, there is a strong negative relationship: as token representations become more compact and well-separated (higher silhouette scores), prediction uncertainty decreases. The relationship is non-linear, which explains why it appeared weaker in the Pearson correlation analysis.

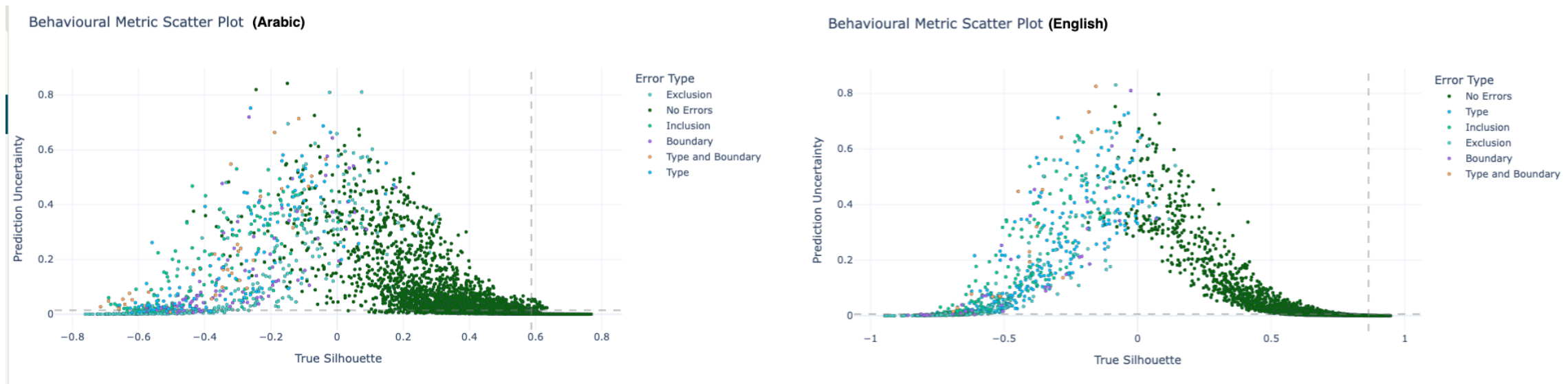

Figure 5.41: Scatter plots showing the relationship between prediction uncertainty and true silhouette scores for Arabic (left) and English (right), coloured by error type. The solid lines represent the average prediction uncertainty at each silhouette value (see Section 4.5.6).

However, same differences emerge between the two languages. In English, the relationship is sharper and more concentrated: tokens with better clustering (higher silhouette) tend to have very low uncertainty, and errors are mainly associated with low silhouette and high uncertainty. In Arabic, while the general trend is similar, the relationship is noisier. Tokens with moderate silhouette scores exhibit higher spread in prediction uncertainty, and errors are more scattered across the silhouette range.

This suggests that while improved clustering generally corresponds to lower uncertainty across languages, Arabic representations show weaker alignment between structure quality and prediction certainty, potentially reflecting model calibration issue (see Chapter 4). While the scatter plot of prediction uncertainty versus true silhouette appears visually sharper and more concentrated in English, the Spearman correlation is stronger in Arabic. This indicates that the ranking between tokens — where higher silhouette scores consistently correspond to lower uncertainty — is more strictly preserved in Arabic than in English. Spearman correlation captures this monotonicity in ordering, even if the absolute spread of points is higher.

These findings indicate that while improved clustering generally corresponds to lower uncertainty across languages, Arabic representations show weaker alignment between structure quality and prediction uncertainty, but stronger preservation of the relative ordering between these two factors.

**Relationship Between True Silhouette and Predicted Silhouette** Another interesting relationship is observed between the true silhouette scores and the predicted silhouette scores. These two metrics show a strong positive correlation across both Arabic and English, both in Pearson and Spearman correlations, although differences in the degree of correlation are observed between the languages. This relationship highlights how the model's internal representation structure (true silhouette) aligns with the structure of its own predicted labels. We discuss this relationship in more details below.

Figure 5.42 presents the relationship between true silhouette scores and predicted silhouette scores, coloured by confusion outcomes (TP, TN, FP, FN). Across both Arabic and English, we observe that TP and TN follow a strong positive trend: as the true silhouette increases, the predicted silhouette also increases, reflecting good alignment between representation structure and prediction outcomes. Similarly, FP and FN show the same trend, where true silhouette decreases as predicted silhouette increases, indicating misalignment in incorrect predictions.

However, there is an interesting distinction between the languages. In Arabic, we observe a longer tail of true negatives (TN) and error points (marked after region A), where tokens exhibit very low predicted silhouette scores despite maintaining relatively higher true silhouette scores. This suggests that in Arabic, there are O tokens and some mispredicted tokens that have a better alignment with true labels but less aligned with prediction labels. This pattern is not as visible in English, where the few TN and error points in the low predicted silhouette region do not show as strong divergence.

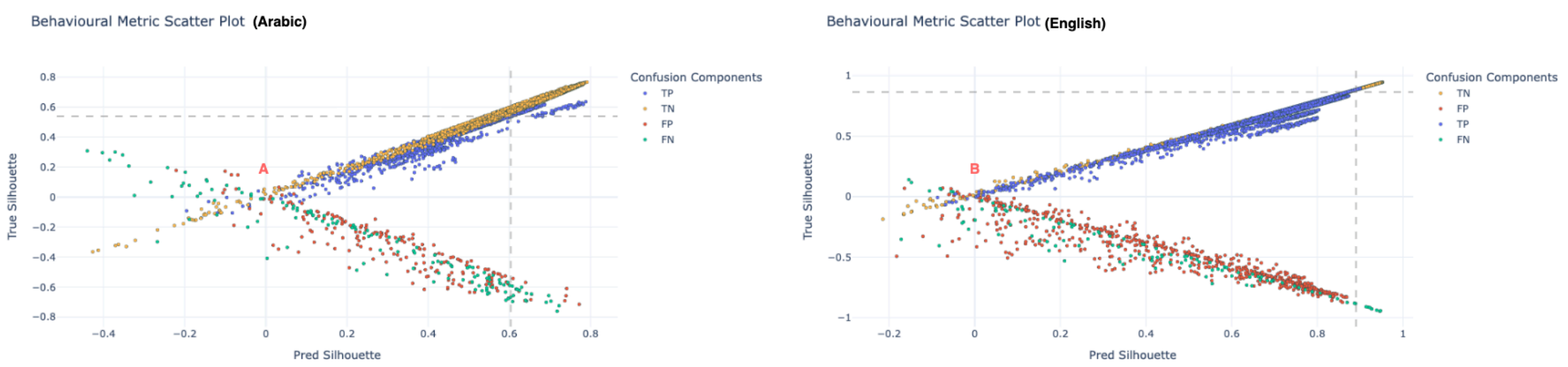

Figure 5.42: Scatter plots showing the relationship between true silhouette scores and predicted silhouette scores for Arabic (left) and English (right), coloured by confusion components (TP: true positive, TN: true negative, FP: false positive, FN: false negative).

The TN points highlighted after region A in Arabic correspond to the anomaly discussed in Figure 5.36, where certain O tokens were located near entity clusters yet still predicted correctly, this pattern is discussed further in Section 5.5.3. In contrast, errors in English tend to behave more regularly, often involving simple misclassifications between O tokens and entity classes. These observations suggest that silhouette scores are valuable not only for interpreting model behaviour, but also for identifying spurious patterns and potential candidates for active learning interventions, as discussed further in Section 6.6.

While the previous analyses revealed strong behavioural trends, not all behavioural metrics exhibit similarly informative relationships. In the following analysis we highlight some aspects associated with these metrics.

**Non Informative Metrics** Some behavioural metrics, while important for understanding data properties, do not exhibit clear or interpretable trends when plotted directly against other metrics. Metrics such as inconsistency ratio and tokenization rate, despite sometimes showing strong correlations with other behavioural metrics, often result in discrete or noisy patterns that are difficult to interpret.

Rather than directly analysing such metrics against representation quality, they are used as filtering criteria — for example, studying how loss values or prediction uncertainty behave for tokens with inconsistency ratios above a certain threshold, or with high tokenization rates. Figure 5.43 presents an example of this, showing the non-informative relationship between inconsistency ratio and true silhouette scores in Arabic.

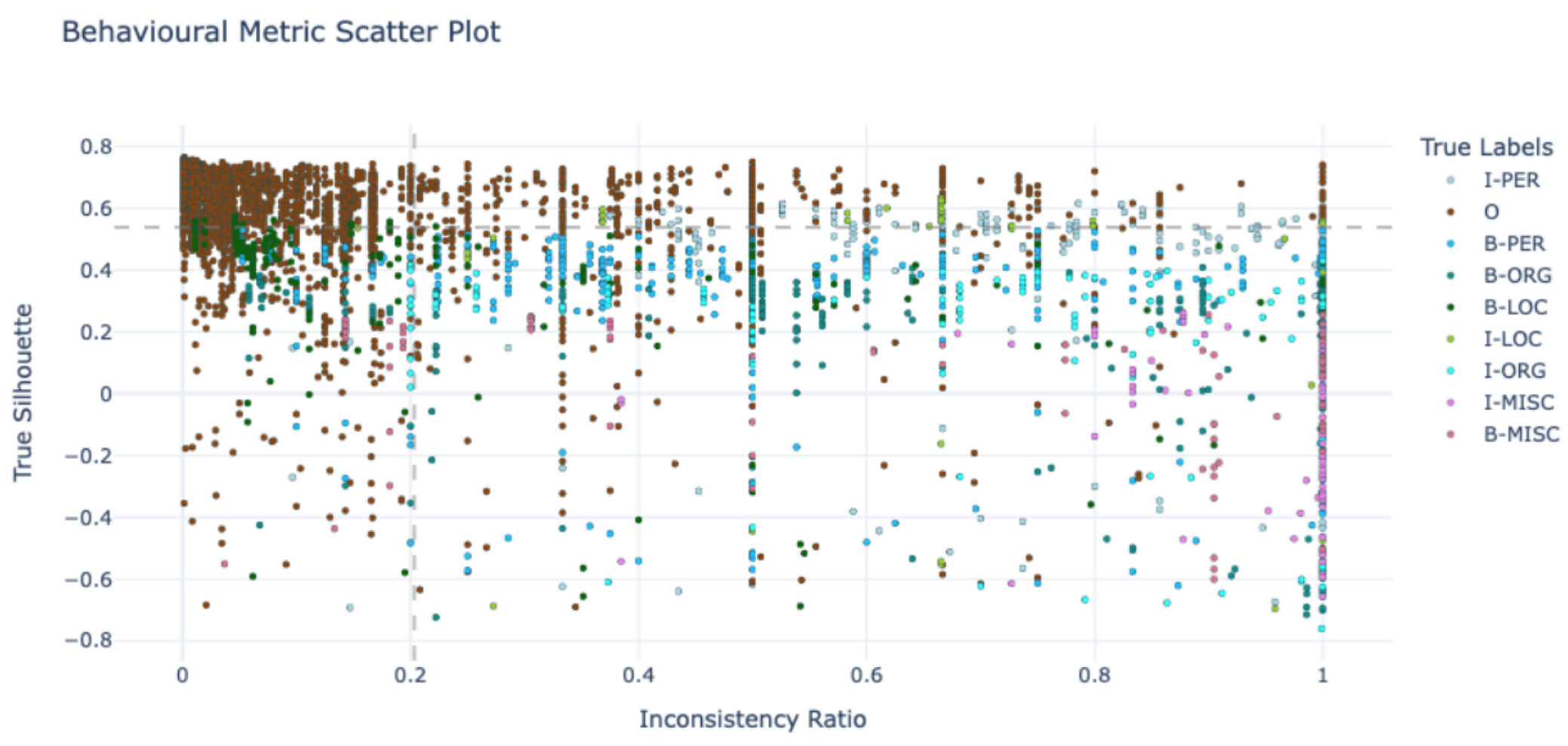

Figure 5.43: Scatter plot showing the relationship between inconsistency ratio and true silhouette scores for Arabic. The English model exhibits a very similar trend; therefore, we only show the Arabic case here for brevity.

Building on the high-level insights into representation structures and token behaviour, we now turn to a deeper analysis of Arabic specific patterns. In the next section, we conduct a targeted exploration of various interesting patterns identified earlier, including error behaviours and structural anomalies.

### 5.5.3 Arabic-Specific Patterns and Anomalies

Building on the global and cross-lingual analysis presented earlier, we now turn to a focused investigation of Arabic-specific representation behaviours. This section explores structural anomalies, systematic errors, and token-level irregularities unique to the Arabic model. These observations build on earlier findings by linking representational patterns to issues in tokenisation, annotation inconsistencies, and error patterns.

As discussed in Section 5.5, we identified two particularly interesting patterns. The first is an error pattern shown in Figure 5.44, where exclusion errors form a distinct subcluster separated from the main dense region. The second is an anomaly, where some O tokens are positioned within the embedding space of different entity tag groups, yet are still correctly predicted as non-entities. In the next section we discuss these patterns in more details.

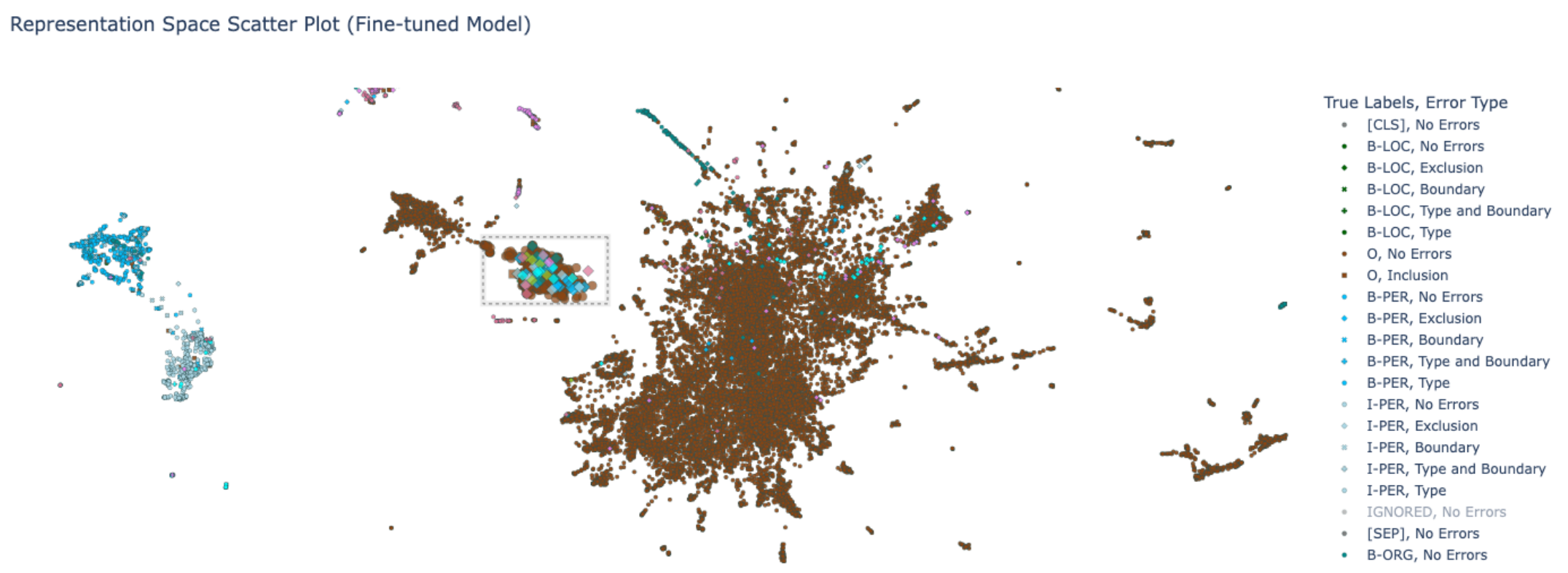

Figure 5.44: Representation space scatter plot for the Arabic model showing O tag tokens and selected entity tokens, coloured by true label and error type. A distinct group of exclusion errors is highlighted by the boxed region, illustrating a localized error pattern separated from the main dense O cluster.

### 5.5.4 Arabic Anomalies

In this section we focus on two distinct Arabic anomalies the first is highlighted in Figure 5.44, the second is (Anomaly a) in Figure 5.36. Examining instances from the highlighted region in Figure 5.44 using the instance-level analysis, we observe a consistent pattern across many examples. Most of the sentences associated with this region start abruptly with random words, without clear semantic or syntactic connections to the rest of the sentence. In these cases, tokens such as country names (e.g., ”Japan”) appear suddenly at the beginning of the sentence, without meaningful contextualisation. These errors are consistent with the ”Sentence Beginning Ambiguity” pattern identified in prior work (AlDuwais et al., 2024). Specifically, they mentioned: ”We noticed an ambiguity in the first words of many sentences where the correct label was not clear.” Our findings similarly show that abrupt or semantically disconnected sentence beginnings contribute to exclusion errors within the Arabic O tag representations.

To further understand how these sentence structures contribute to ambiguity, we examine the same type of example highlighted in their analysis. Figure 5.45 shows Example (1) from the test split. The first word is annotated as B-PER; however, it is not aligned with the semantic context of the sentence that follows, making the correct entity labelling and the position of the word unclear.

Using instance-level analysis, we investigated the history of the sentence under in-vestigation. We performed string-based matching to retrieve the original sentences from the (Benajiba) dataset. This revealed that the token برند ("brand") was originally part of a completely different sentence (shown in Figure 5.47), while the main body of the sentence (shown in Figure 5.46) comes from a separate context. This indicates that during data preparation, the token برند was incorrectly inserted at the beginning of an unrelated sentence, creating a synthetic construction without coherent semantic structure. Such noise in the data likely contributed to the model's confusion and the observed exclusion errors.

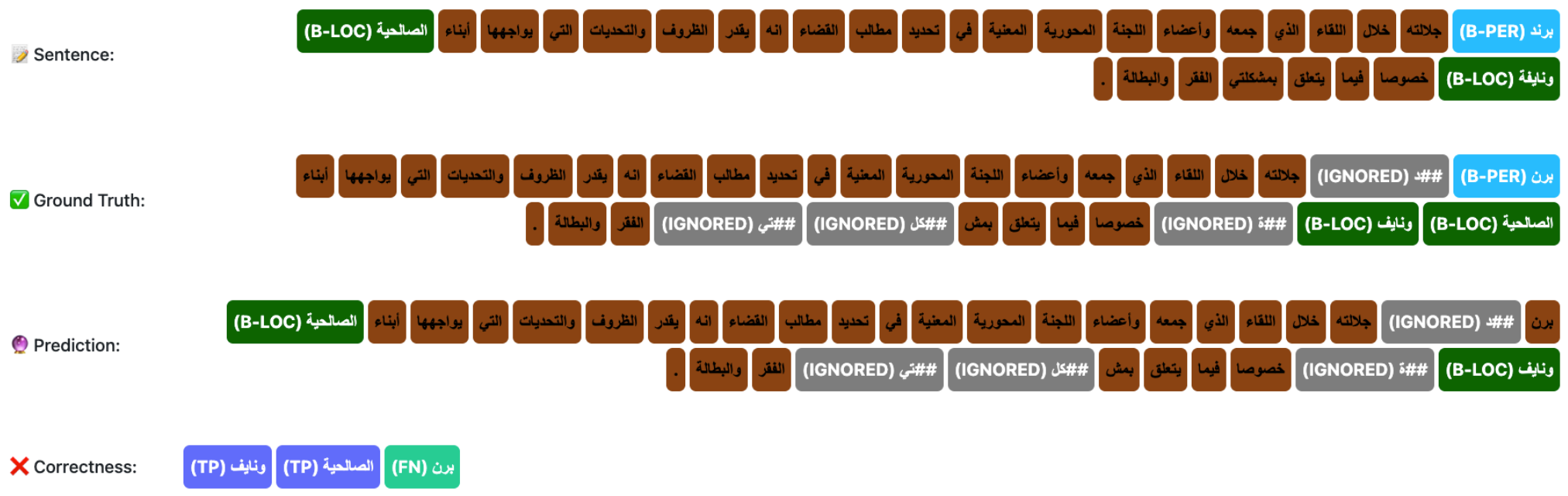

Figure 5.45: The anomalous sentence under investigation, showing an abrupt beginning with the inserted token برند. This construction does not match a natural sentence structure, contributing to model confusion.

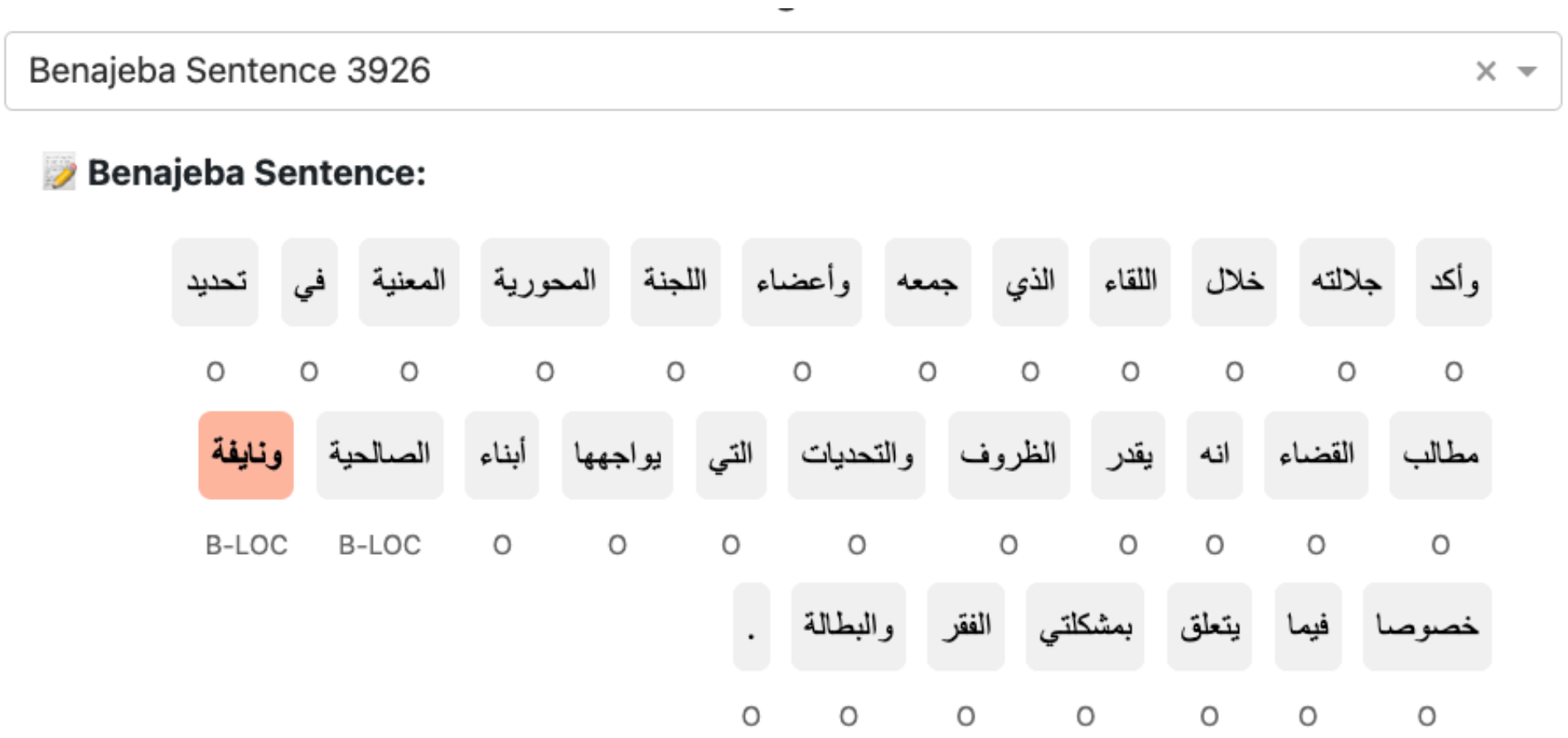

Figure 5.46: Original sentence from the BenaJeba dataset containing the main body of the sentence under investigation.

As mentioned in Section 2.3.1, the ANERCorp dataset was cleaned by the CAMeL Lab, during which some issues were fixed. It is possible that parts of this correction process were performed semi-automatically, which may have introduced systematic errors such as the insertion anomaly observed here. These artifacts appear to have been picked up by the model during training, contributing to the exclusion errors detected in the Arabic O tag representations. Further investigation into the annotation and cleaning processes would be needed to fully understand the source of these anomalies.

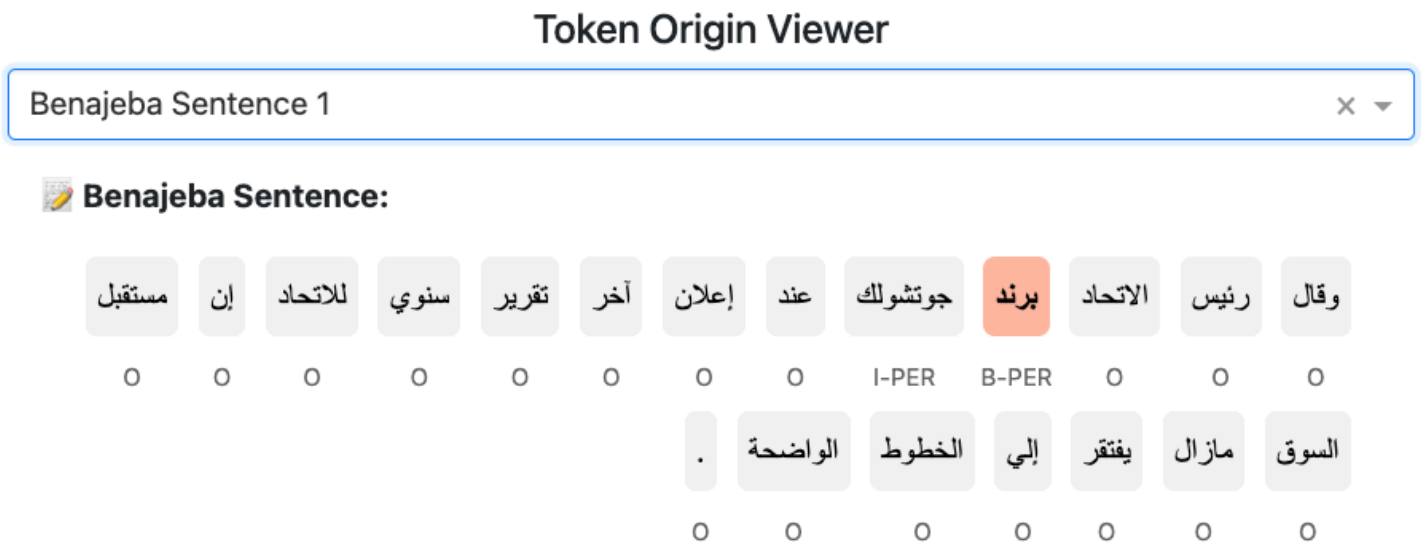

Figure 5.47: Original sentence containing the token برند ("brand"), which was incorrectly inserted into the current sentence.

This observation highlights the importance of transparency in correction mechanisms: while they aim to resolve existing annotation issues, they can inadvertently introduce new errors into the dataset, potentially affecting downstream model behaviour.

To understand the impact of these systematic errors on the behaviour of other model components, we now examine the relationships between key behavioural metrics. These metrics are filtered using our behavioural filtering table to include only exclusion errors associated with the O tag. This allows us to isolate and study this pattern specifically.

**Token Confidence vs Loss** Figure 5.48 presents the relationship between token confidence and loss values for exclusion errors within the Arabic O tag. We observe that token confidence associated with this pattern highlighted by the diamond shape is consistently high — often exceeding 0.9 — while loss values also remain high above 7. This pattern suggests that the model is highly certain in its predictions even when they are incorrect. Since cross-entropy loss penalizes incorrect predictions more when confidence is high, this behaviour suggests that the model is poorly calibrated in these cases. These patterns likely reflect the impact of systematic sentence anomalies, where irregular input structure leads the model to produce confident but incorrect outputs.

**Examining Predicted vs True Silhouette** Figure 5.42 presents the relationship between predicted and true silhouette scores for same pattern. We observe a positive but noisy correlation, with increased variance across both axes.

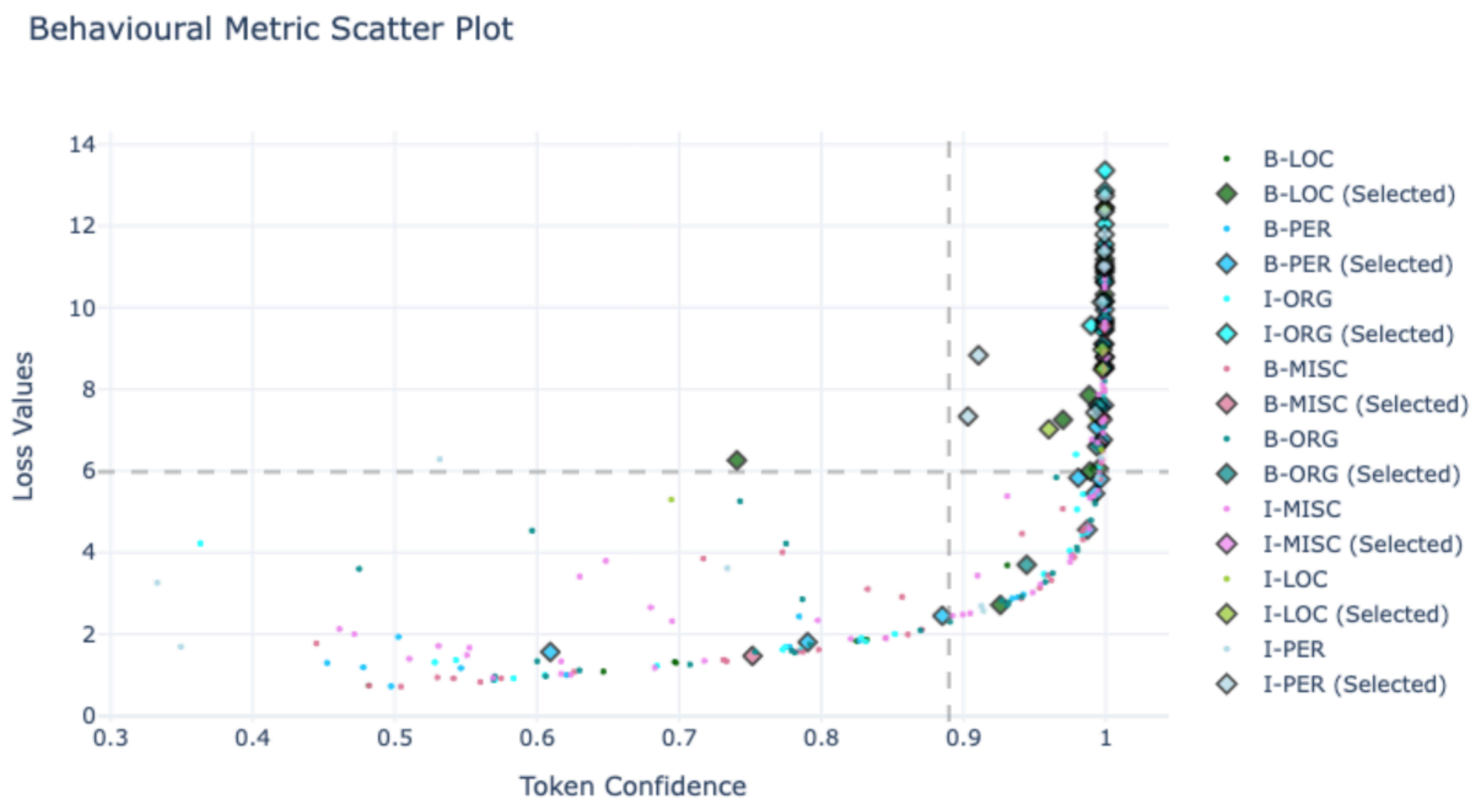

Figure 5.48: Relationship between token confidence and loss values for exclusion errors within the Arabic O tag. Selected tokens are highlighted by diamonds.

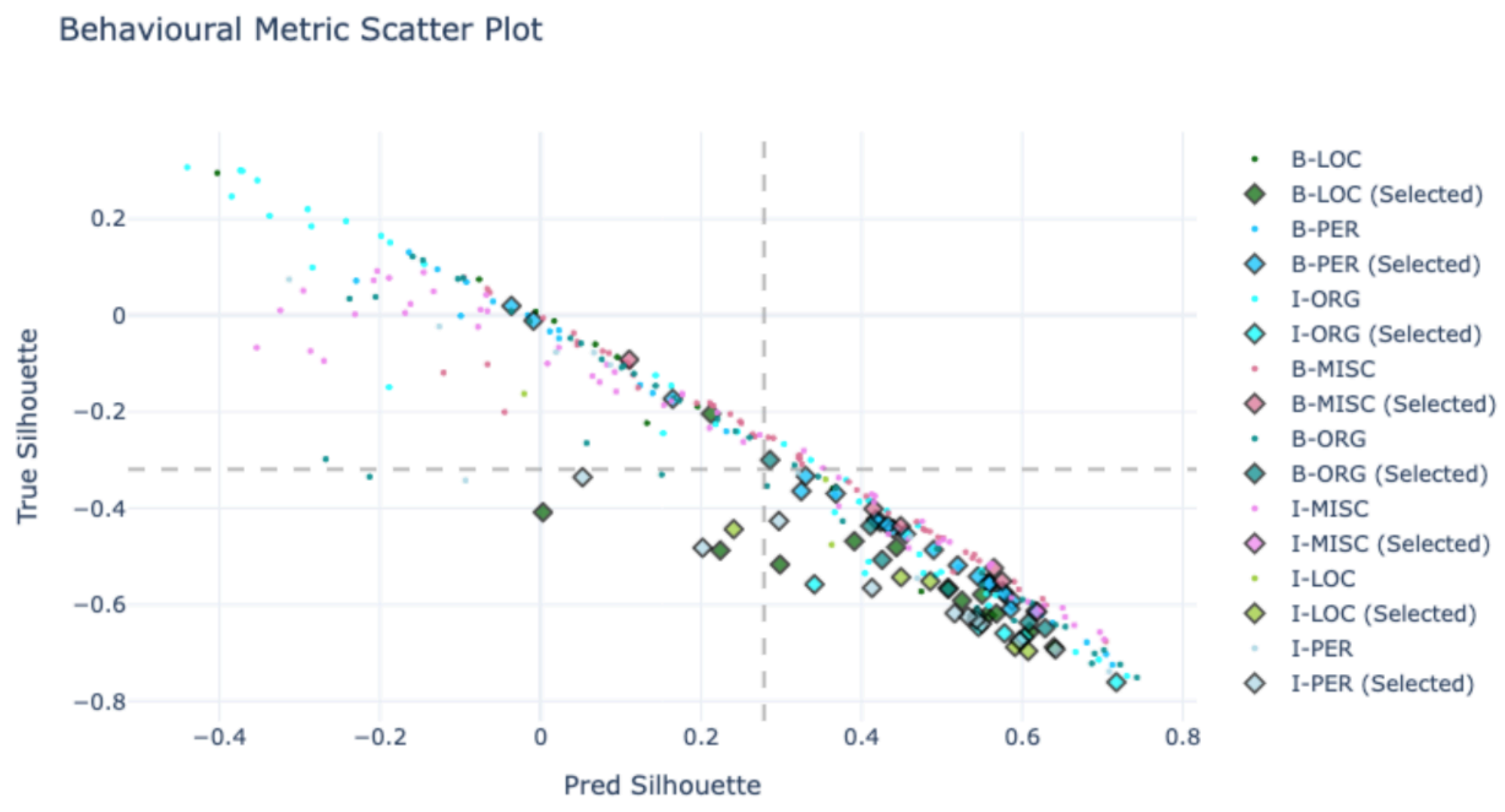

Figure 5.49: Relationship between predicted and true silhouette scores for exclusion errors in the Arabic O tag.

This suggests a mismatch between the internal representation structure and the model's predicted labels, particularly for structurally ambiguous or anomalous tokens. These points deviate from the general trend, exhibiting high predicted silhouette despite low true silhouette, or vice versa. The presence of such points distorts the overall correlation and may explain the larger divergence between Arabic and English silhouette trends observed earlier.

Importantly, this divergence does not appear to result from general label noise, but from systematic structural issues introduced during preprocessing — such as token inser-tions or abrupt sentence constructions. While annotation inconsistencies exist in both languages, the severity and representational impact of these issues appear more notable in Arabic, both on the representation level and other subcomponents such as loss and confidence. This could also indicate that Arabic may suffer from over-fitting to spurious patterns, where the model becomes confident in structurally anomalous inputs.

**Anomaly Region (A)** We now turn our attention to the anomaly pattern previously highlighted in Section 5.5.1. Our analysis revealed two key observations. First, some O tokens appear within the dense regions of other entity tags in the representation space, yet are still predicted correctly as O. This suggests that class membership is not the main indicator of correct predictions in the Arabic model's representation space.

Second, we observed a related behavioural pattern at the metric level: a subset of true negative (TN) tokens in the Arabic O tag show relatively high true silhouette scores but low predicted silhouette scores. This indicates that while these tokens are well-separated from others in the representation space (high true silhouette), the model's predictions fail to capture that structure, further highlighting representational misalignment. To further understand this anomaly, we examine three examples in more detail.

**Example (5)** As highlighted in Figure 5.50, the token البادية (albadiya) is correctly predicted as O by the model, despite being embedded within the B-LOC cluster in the representation space (see Figure 5.36). Semantically, the word refers to a desert region and functions as a location reference in the sentence (see point A), which explains its spatial proximity to the B-LOC cluster. However, according to the dataset annotation, the word is consistently labelled as O, as shown in Figure 5.51.

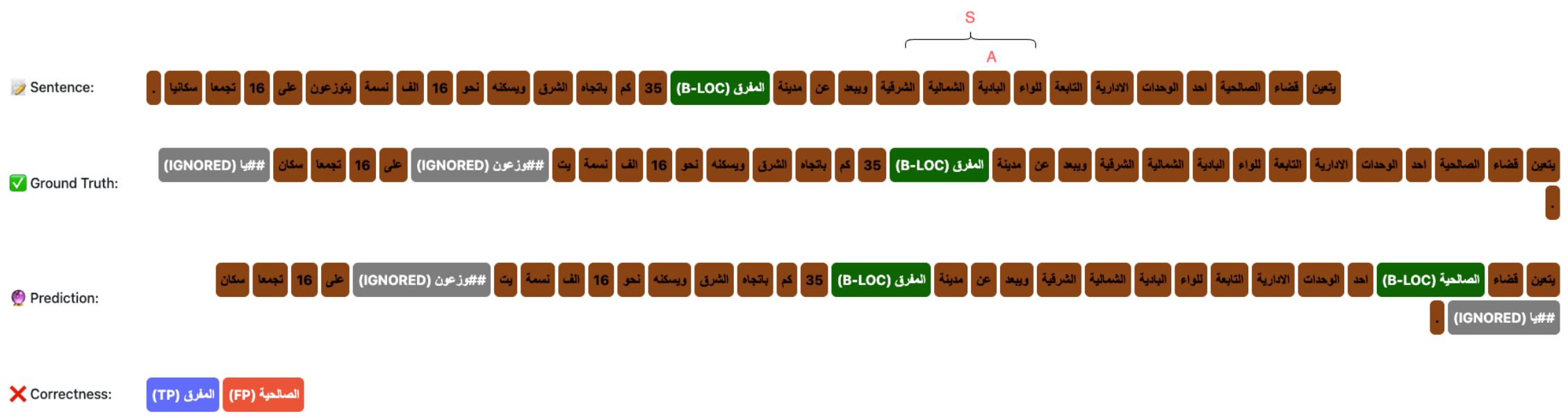

Figure 5.50: Example (5): Anomaly in the Arabic O tag involving the token البادية (albadiya), marked as **A**, which is semantically indicative of a location (desert) and appears within a broader location-related span **S**.

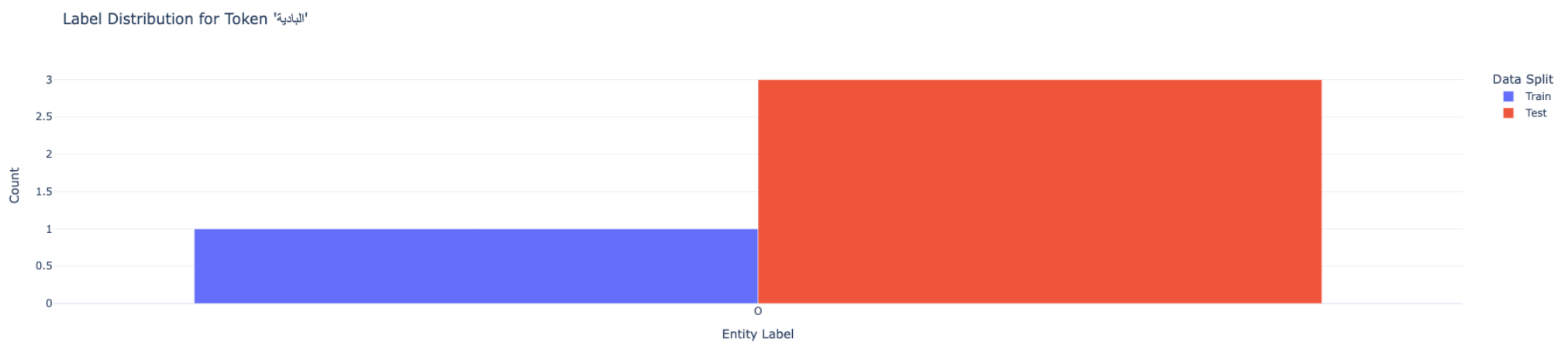

Figure 5.51: Label distribution for the token البادية (albadiya) across the training and test splits. All observed instances are annotated with the O label, reinforcing that the model's semantic grouping is not driven by label supervision alone.

Inspecting the model's prediction probability we observe that although the model favours the O class for this token, the uncertainty is high, suggesting uncertainty in the model output layer, as shown in Figure 5.52. Using another feature of the instance-level analysis tab, we compute the cosine similarity between the token البادية (albadiya) in the example under investigation and its other occurrences across the training and test splits. As shown in Figure 5.53, this word appears once in the training data and three times in the test set. The similarity scores indicate stronger alignment with Example (0) from the test set (0.87), compared to its nearest training example (0.72). Note that the similarity score of 1.0 reflects self-similarity — i.e., the token compared to itself.

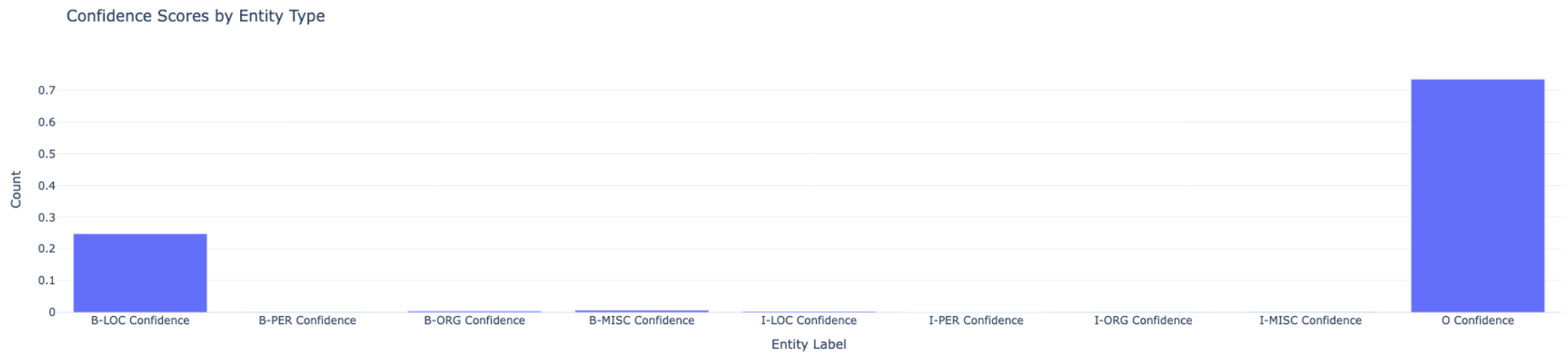

Figure 5.52: Model confidence distribution for the word البادية (albadiya) across all entity labels. Although O receives the highest confidence, the model also assigns noticeable probability to B-LOC, indicating uncertainty and semantic ambiguity.

Examining Sentence (0) from the test split (Figure 5.54) and the training example Sentence (1234) (Figure 5.55), we observe that both contexts use the word البادية (albadiya) to refer to a geographic location. However, in both cases, the token is annotated as O rather than B-LOC, despite its clear geographic reference. This reinforces our earlier observation: the model's prediction of O appears to be driven more by its training exposure to this annotation than by the semantic content of the word itself, which is correctly captured in the representation space as location-like.

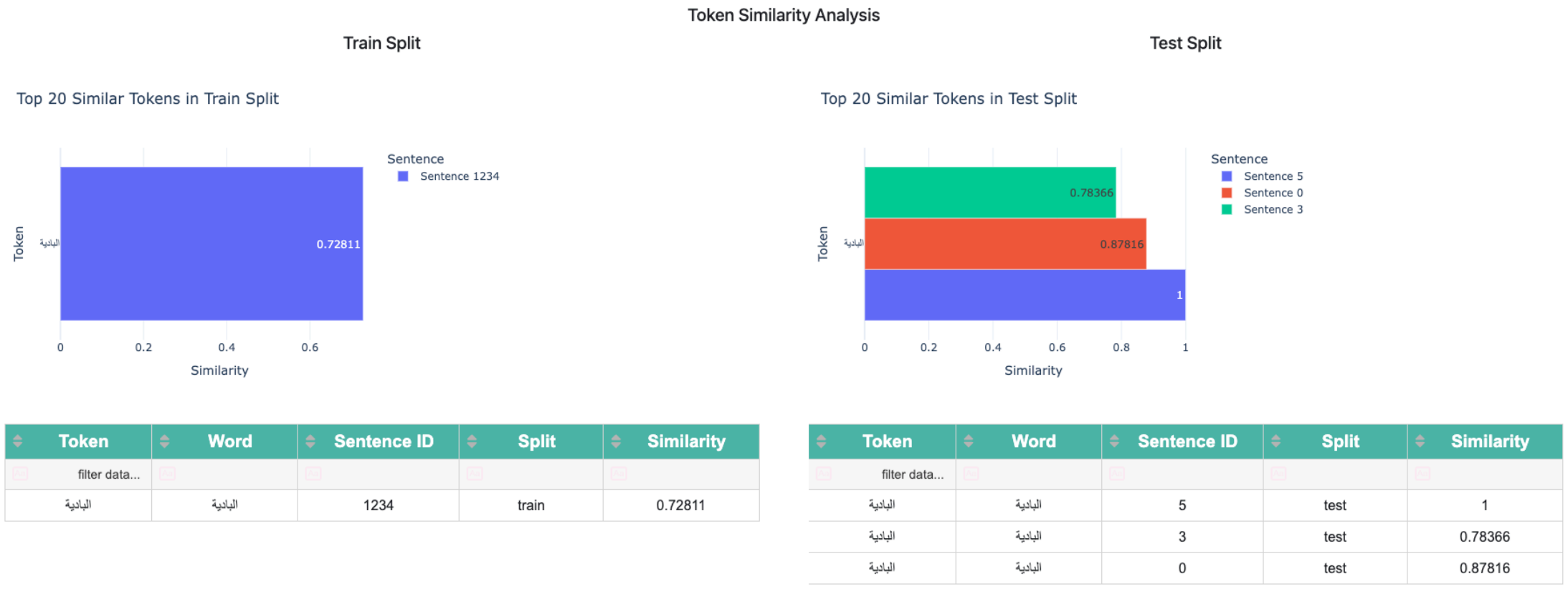

| Token | Word | Sentence ID | Split | Similarity |
|---|---|---|---|---|
| filter data... | | | | |
| البادية | البادية | 1234 | train | 0.72811 |

| Token | Word | Sentence ID | Split | Similarity |
|---|---|---|---|---|
| filter data... | | | | |
| البادية | البادية | 5 | test | 1 |
| البادية | البادية | 3 | test | 0.78366 |
| البادية | البادية | 0 | test | 0.87816 |

Figure 5.53: Cosine similarity between the word البادية (albadiya) in the selected example and other instances in the training and test splits. The word appears once in the training data with 0.73 similarity and three times in the test data with higher similarity (up to 1.0), indicating strong contextual alignment across test examples.

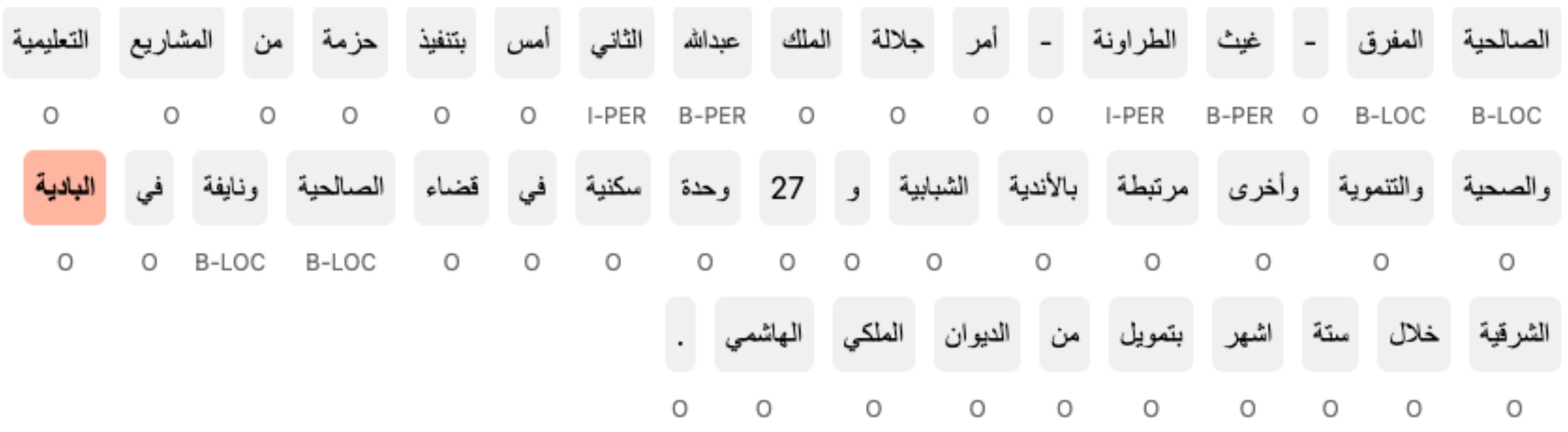

Figure 5.54: Example of the token البادية (albadiya) in Sentence 0 of the test set, used in a geographic context but annotated as O.

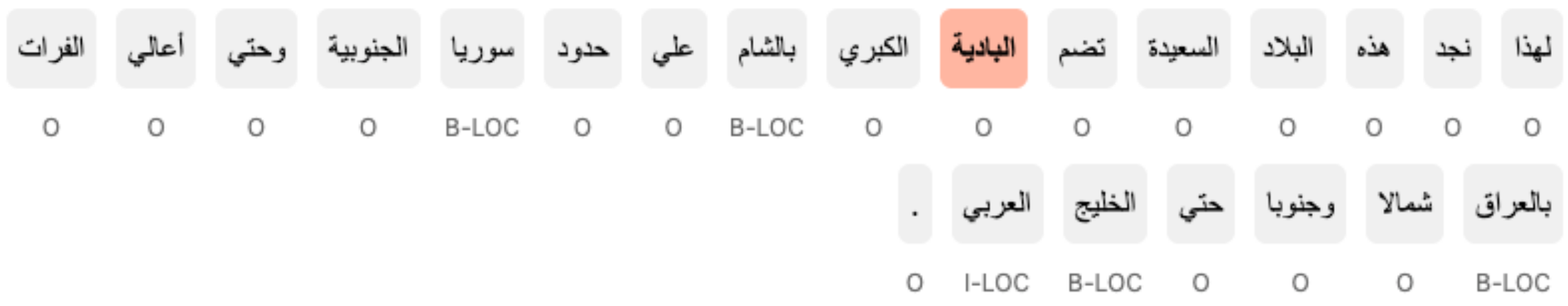

Figure 5.55: Training example (Sentence 1234) showing the token البادية (albadiya) used in a similar context and also annotated as O.

This example illustrates a key discrepancy between prediction behaviour (output layer subcomponent) and representation structure (last hidden state subcomponent) in the Arabic model. While the model correctly predicts the O label for البادية (albadiya), the token is embedded within the B-LOC region in the representation space, aligning with its semantic meaning as a location. This dual alignment — with both the annotation history and the underlying meaning — suggests that the Arabic model has learned to reconcile semantic information with annotation inconsistencies. Such behaviour may explain the increased noise and spread observed in Arabic behavioural metrics compared to English, and may also indicate signs of over-fitting, as the model appears to fit both semantic signals and annotation signals. Furthermore, these findings indicate that the Arabic model develops unique representational heuristics to accommodate language-specific annotation patterns and semantic ambiguity. This highlights the importance of clearer, Arabic-specific annotation guidelines, and the need for further investigation into how fine-tuning affects representations across languages.

### 5.5.5 Tokenisation Impact

In this section, we explore how tokenisation during preprocessing can create misalignments between the model's internal representations and its output predictions. Example (832), shown in Figure 5.56, illustrates two key forms of ambiguity that interact in Arabic: Points A and B highlight the effect of diacritic removal (*tanween*), while Point T highlights ambiguity introduced by subword segmentation during tokenisation. The example demonstrates how both sources of ambiguity can influence model predictions, leading to inconsistent alignment between learned representations and output labels.

**Diacritic-Related Ambiguity** As shown in Figure 5.56, Point A shows the word with *tanween* (diacritic)—إسبانيًا (Spanish, adjective)—and without tanween—إسبانيا; Point B shows the same word in its original sentence position, where it can mean either "Spain" or "Spanish," depending on context. The removal of the *tanween* therefore creates lexical ambiguity: the fully diacritized form إسبانيًا (Spanish, adj.) is unambiguously adjectival, whereas the stripped form إسبانيا may also be interpreted as a location. This ambiguity produces mismatches between annotation and representation. The token frequently appears in the training data as a location due to surface-form matching, reinforcing the B-LOC label in the model's representation. However, in this particular sentence, the context does not clearly indicate a location, leading the model to predict O. This contrast between annotation frequency and contextual usage illustrates how the Arabic model, when uncertain, tends to default to O for ambiguous tokens—especially when they diverge from common usage patterns. It also highlights the critical role of diacritics in disambiguating meaning in Arabic and underscores the need for diacritic-aware tokenisation and modelling strategies.

**Tokenisation-Related Ambiguity** Another source of ambiguity arises from tokenisation artefacts. As highlighted in Figure 5.56 (Point T), the token كات (Kat) was split from the longer word كاتالانية (Catalanian) during subword tokenisation. This subword corresponds to part of the word "Catalonian," but due to the split, it appears only as "Kat." Although the token is annotated as O in the gold standard, the model predicted it as B-LOC.

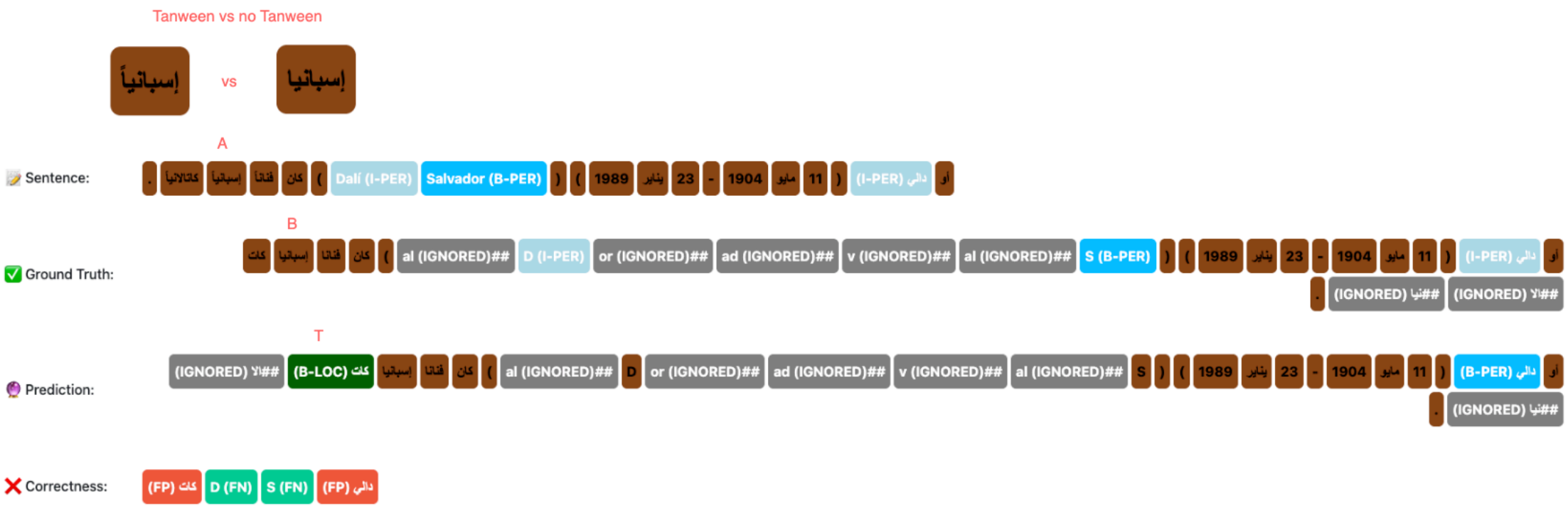

Figure 5.56: Instance-level visualisation of the sentence in Example (832) containing the tokens إسبانيا (Spain) and كات (Kat, from كاتالانية — Catalanian). Point A shows إسبانيًا with tanween (diacritic), while point B shows the same token after diacritic removal (إسبانيا). Point T marks the subword كات (Kat), which was split from the longer word كاتالانية (Catalanian) during tokenisation.

To investigate this decision, we examined the cosine similarity between this token and its other occurrences across the training and test splits. Figure 5.57 presents the similarity results, showing that the most similar instance in the training split is in Sentence 1801. Using the token context viewer, we retrieved this training sentence to examine the token's usage. As shown in Figure 5.58, the token occurs within the phrase "State of Katanga," clearly referring to a location entity. In that context, the token is consistently annotated as B-LOC, reinforcing the model's learned association between this subword and location labels.

In Example (832), however, the same subword occurs in a different context—within the word "Catalonian"—which describes nationality rather than a geographical location. Despite the visual similarity of the subword, the contextual cues do not indicate a location, leading the model to predict O. This contrast shows how the model's prediction is shaped by both contextual similarity and representational memory: when familiar tokens appear in atypical or low-similarity contexts, the model tends to default to O, a behaviour also observed in sentence-initial anomalies.

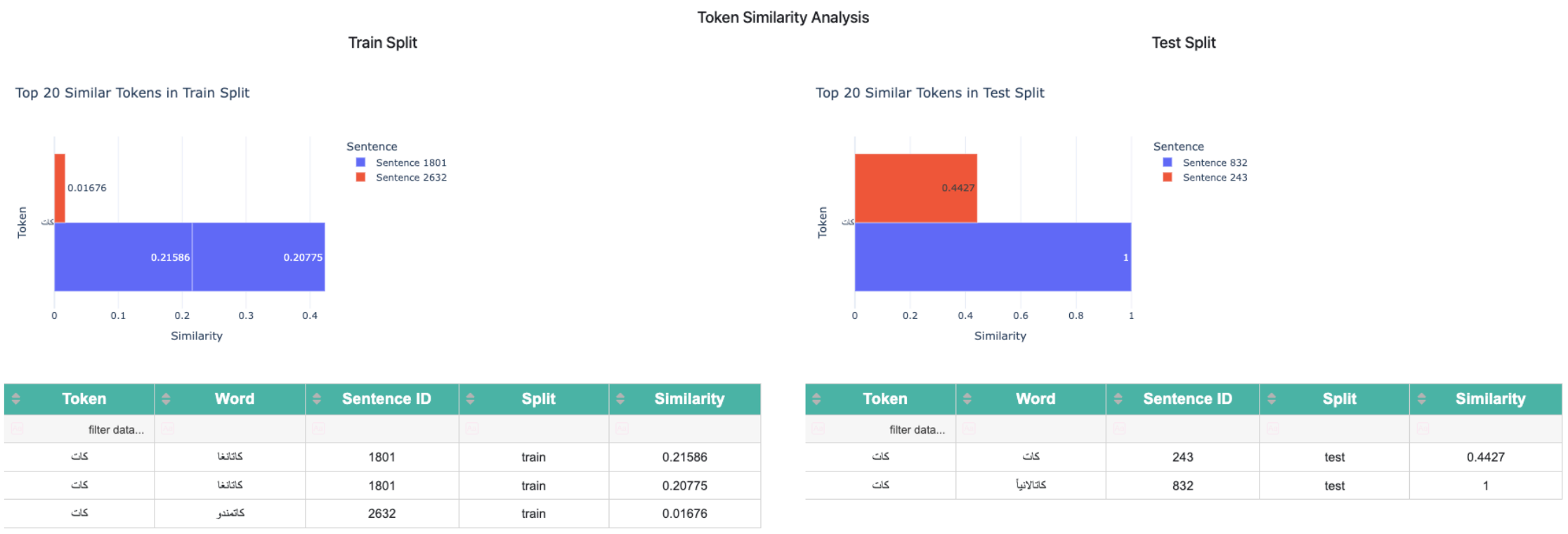

Figure 5.57: Token similarity distribution for the ambiguous token كات (Kat) in Example (832).

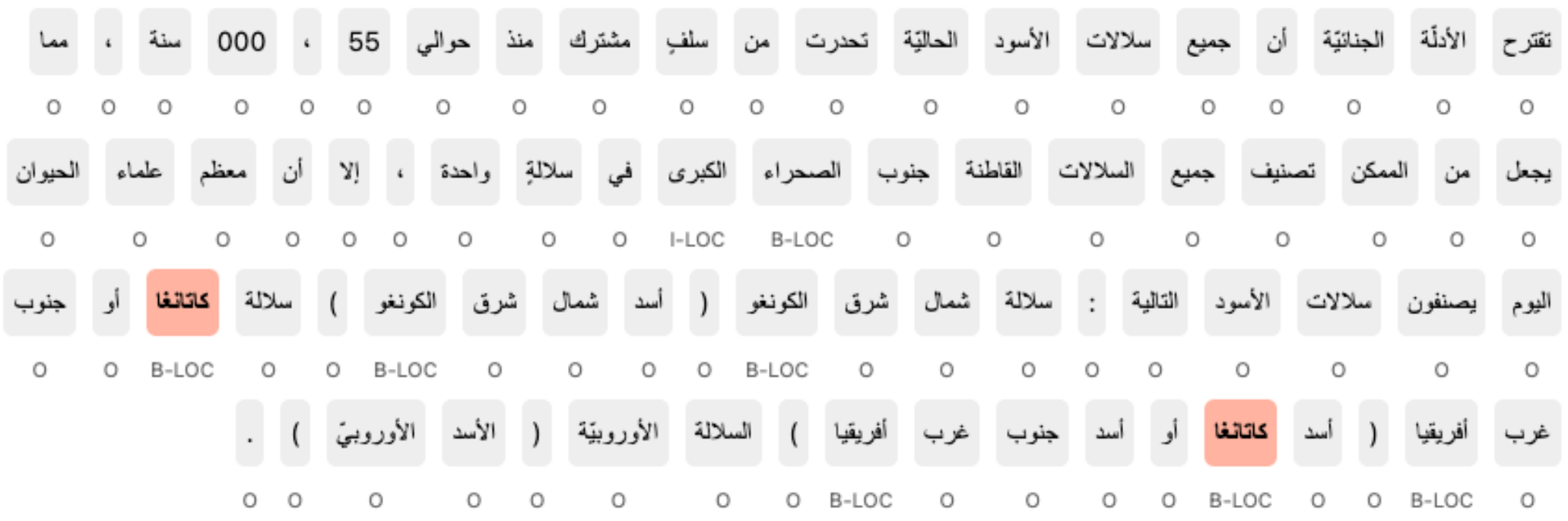

Figure 5.58: Training example showing the most similar occurrence of the token كات (Kat), retrieved from Sentence 1801.

These examples highlight how token-level ambiguity — whether stemming from orthographic variation or subword tokenisation — can cause misalignments between model representations and predictions. This reinforces the importance of diacritic-aware preprocessing and a better handling of foreign name tokenisation in Arabic. Additionally, it suggests that ambiguous or unfamiliar tokens are more likely to be assigned the O label in the model's output layer, even when their internal representation aligns with another class — pointing to a potential form of over-fitting in the output layer's behaviour.

### 5.5.6 Annotation Scheme Misalignment

In this section, we focus on annotation scheme misalignment, examining both gold-standard labels and model predictions. As discussed in Chapter 4, the cross-component analysis revealed inconsistencies in the ground truth annotations. Here, we extend that analysis by investigating whether similar misalignments are present in the model's predicted labels, and how these patterns differ from those found in the gold standard. We begin by identifying where these misaligned instances appear in the representation space and examining whether they align with any of the previously observed structural patterns or errors.

**Ground Truth Misalignment** Figure 5.59 shows the location of ground truth misaligned tokens in the Arabic representation space. We find that many of these tokens cluster within or near the regions previously identified as sentence-start anomalies — a pattern discussed in Section 5.5.1 and further examined in Section 5.5.4. This suggests that some of the annotation scheme misalignments in the gold-standard labels are not isolated errors but are partly driven by systematic annotation errors such as the sentence-beginning anomaly.

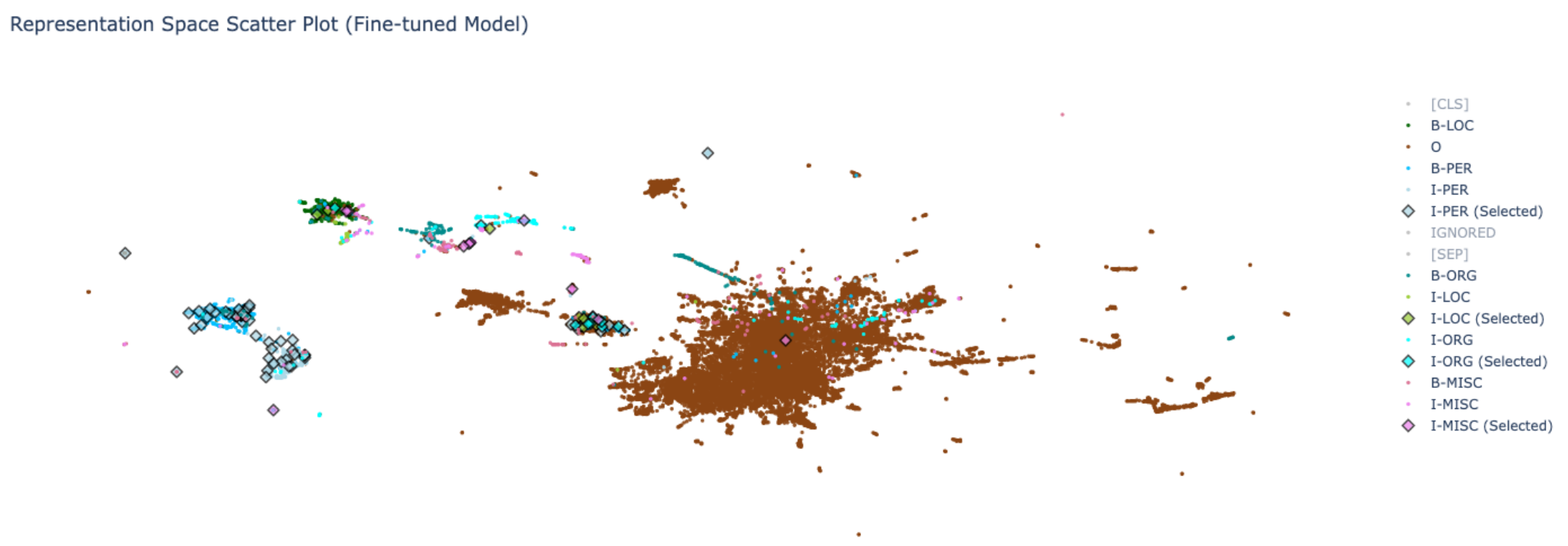

Figure 5.59: Representation space of the Arabic model showing the position of misaligned ground truth tokens (highlighted). These tokens often correspond to previously identified anomalies.

To avoid redundancy, we focus on examples from the PER region and identify two distinct patterns: (1) genuine annotation errors, and (2) tokens affected by the sentence-start anomaly. Figure 5.60 illustrates Example (338) from the PER region, where a token was incorrectly labelled due to abrupt sentence-beginning anomaly. Tracing this token back to its original source in the Benajiba dataset (Figure 5.61) confirms that the sentence begins with an ambiguous and malformed structure. These findings suggest that misalignment in annotation scheme in the ground truth is not isolated to specific region, it is scattered across various entity tags.

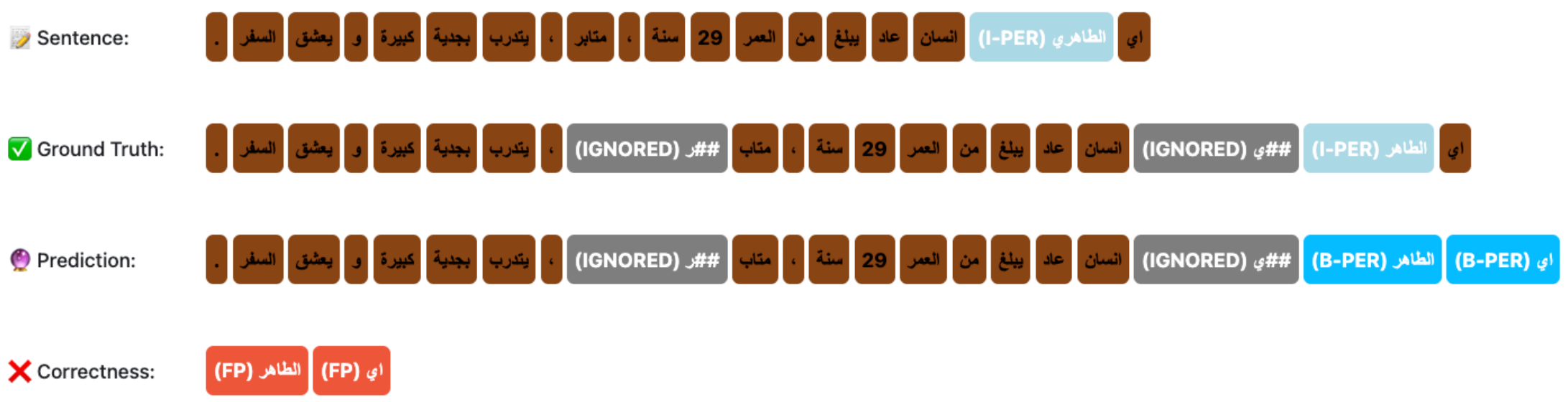

Figure 5.60: Misaligned ground truth example (338) from the PER region.

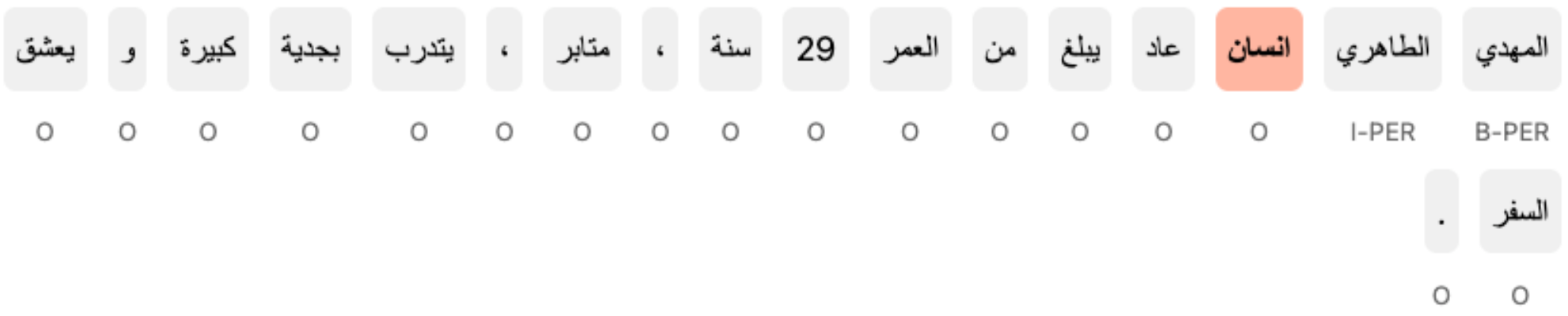

Figure 5.61: Original sentence (ID 4300) from the Benajiba dataset.

In addition to sentence-start anomalies, we also identify direct annotation errors. Figure 5.62 presents example (623) from the LOC region. The token بألمانيا (in Germany) is incorrectly labelled as `I-LOC`, despite being a country name.

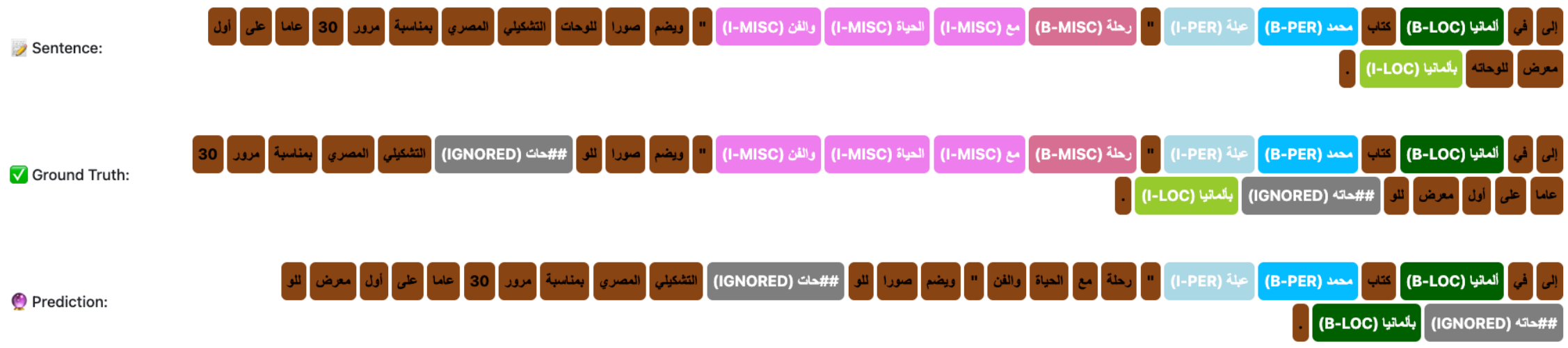

Figure 5.62: Annotation error in example 623: the token بألمانيا is labelled as `I-LOC` in the ground truth, however it should be B-LOC.

This example illustrates how annotation noise can mislead both model training and evaluation. Although the model's prediction diverges from the ground truth, it aligns with the actual semantic structure of the sentence—highlighting the value of behavioural analysis in uncovering annotation inconsistencies that may otherwise be interpreted as model errors.

**Token Confidence vs Prediction Uncertainty** To understand how annotation misalignments affect model behaviour, we investigate the relationship between token con-fidence and prediction uncertainty. Figure 5.63 shows that the model is more confident in its predictions for tokens associated with systematic errors — particularly those tied to the sentence-start anomaly. These tokens appear in the top-left corner of the plot (highlighted box), showing high confidence and low uncertainty. In contrast, tokens affected by annotation inconsistencies (e.g., mislabelled spans) are associated with lower confidence and higher uncertainty. This difference suggests that the model becomes more confident when exposed to systematic error patterns in the training data, while remaining uncertain when faced with inconsistent or ambiguous labels.

**Predicted vs True Silhouette** Another interesting behavioural relationship is observed between predicted and true silhouette scores. As shown in Figure 5.64, we can identify three distinct regions in the scatter plot:

- **Region A:** Represents correctly predicted tokens, where both true and predicted silhouette scores are high. These tokens are mostly coherent and well-aligned between the representation and prediction components.
- **Region C:** Corresponds to tokens exhibiting systematic errors — such as sentence-start anomalies — which have low silhouette scores in both subcomponents. This pattern has been discussed earlier in Section 5.5.4.
- **Region B:** Introduces a novel pattern, where true silhouette scores are low while predicted silhouette scores remain high. This region suggests a divergence between the last hidden state representation and the output layer.

To better understand Region B, we examine Example (705) from the PER region, shown in Figure 5.65. The word indicated by point A is عبد (Abd), which was incorrectly annotated as I-PER in the ground truth but predicted as B-PER by the model. Looking at the training distribution, this token appears more frequently as I-PER (58 instances) but also has a considerable number of B-PER instances (46), suggesting ambiguity at the span-boundary level.

The model's representation component places the token in the I-PER cluster, reflecting its contextual embedding. Meanwhile, the output layer favours the B-PER class, highlighting a divergence between the internal subcomponents. Notably, this sentence also exhibits a sentence-start anomaly, which may not only affect the initial token but seems to also influence subsequent tokens in the sentence. This example illustrates how overlapping span boundaries in Arabic annotations can lead to misalignment be-tween the model's representation and prediction layers.

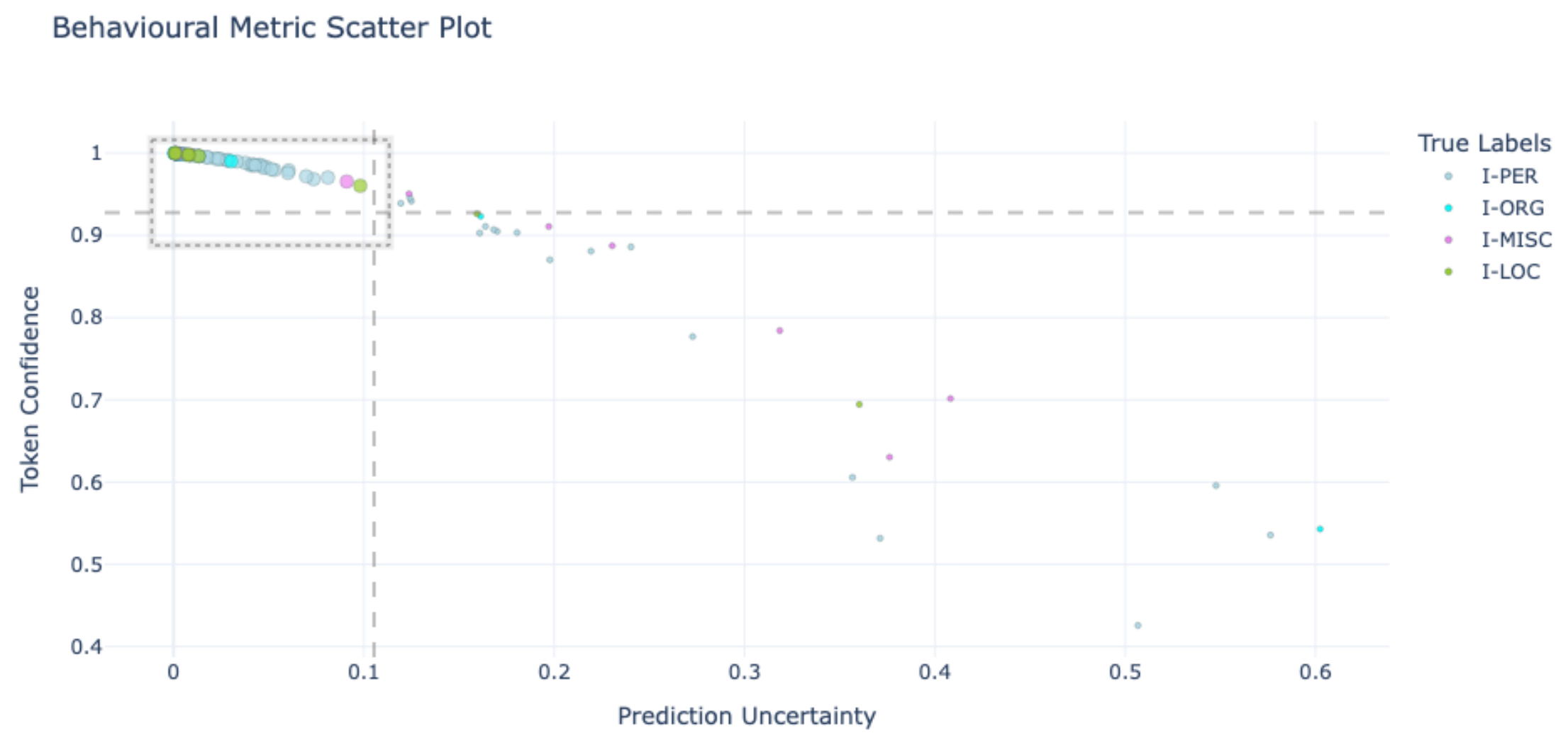

Figure 5.63: Scatter plot showing the relationship between token confidence and prediction uncertainty for misaligned tokens. Highlighted region shows high-confidence systematic errors (e.g., sentence-start anomalies). Remaining points reflect annotation inconsistencies with lower model confidence.

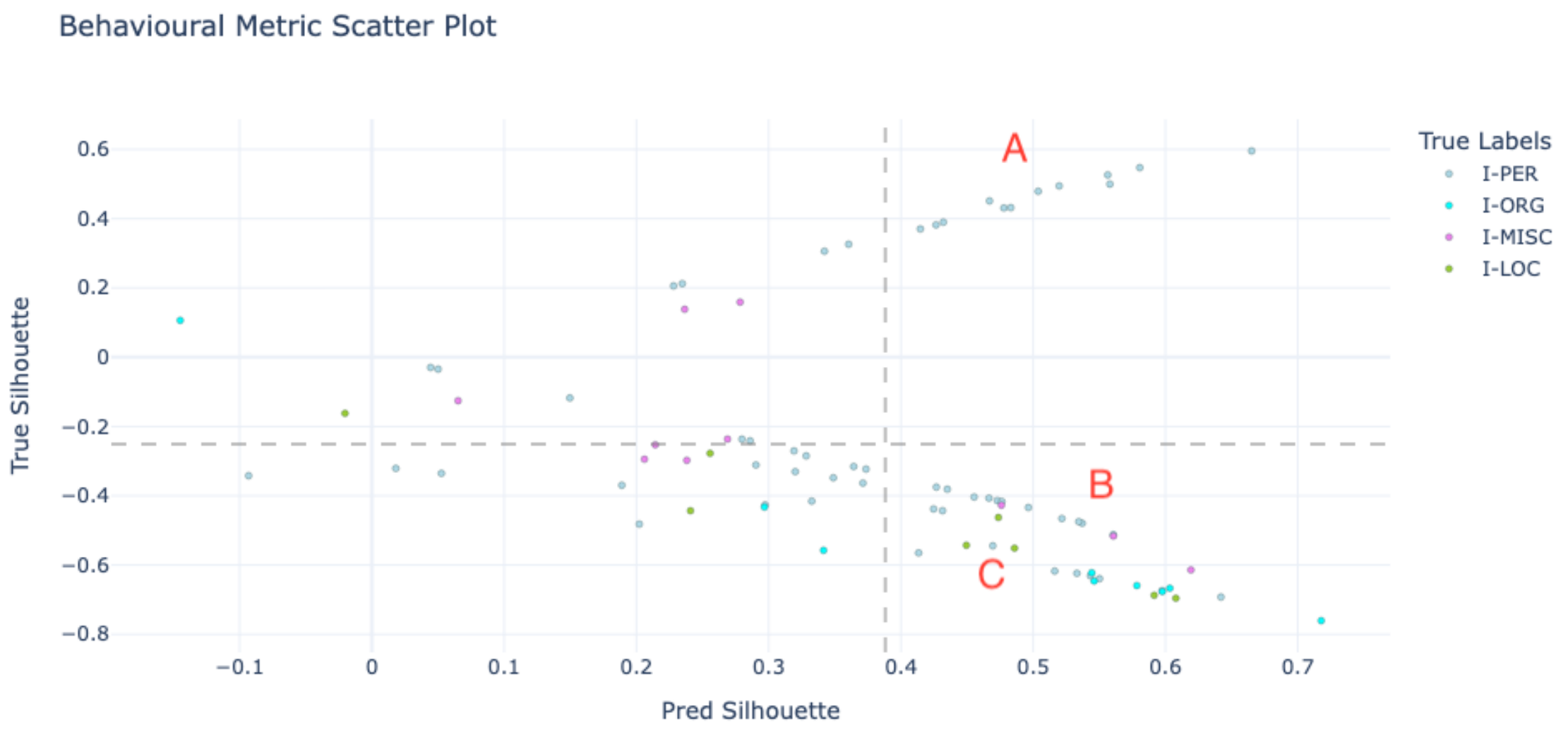

Figure 5.64: Scatter plot showing the relationship between predicted and true silhouette scores for selected misaligned instances. Distinct regions A, B, and C represent correct predictions, subcomponent misalignments, and systematic errors, respectively.

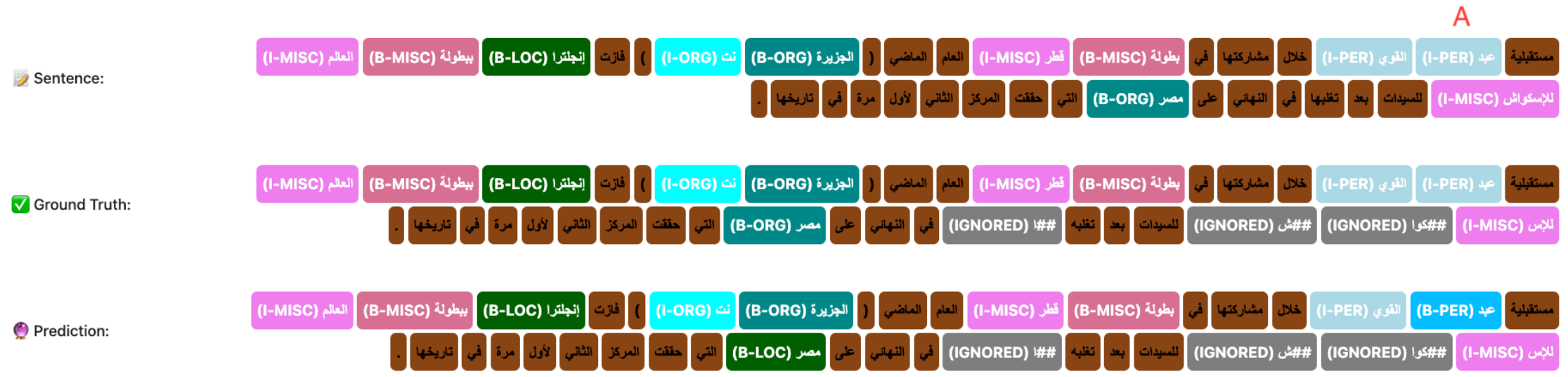

Figure 5.65: Example 705 illustrating a subcomponent misalignment. Token عبد (Abd) was annotated as I-PER but predicted as B-PER, despite its representation aligning more closely with the I-PER cluster.

This example highlights the compounded effects of span-boundary ambiguity and sentence-level noise on model internals. Such misalignments between representation and output suggest that the model organizes its internal representations in a way that reflects the class labels — but due to inconsistent annotations or systematic errors, this internal organization does not always align with the model's prediction decisions.

**Predicted Label Misalignment** We now turn to misalignment patterns observed in the model's predicted labels. These fall into two major categories: entity boundary errors and systemic span misclassifications.

**Partial Entity Predictions** Figure 5.66 presents an example (90) where the model fails to capture the full extent of a named entity span. The predicted entity misses the final token in the true span, resulting in a misalignment even though the model partially recognized the entity. Such errors are common when the entity boundary is long or contains a mix of entity and ambiguous non-entity tokens.

**False Positive False Positive** A second type of misalignment arises when the model makes a reasonable prediction that aligns with the expected annotation scheme (IOB2), but the ground truth label is inconsistent. In the example shown in Figure 5.65, the token عبد is labelled as I-PER in the dataset, even though it appears at the beginning of a person name span. According to the IOB2 scheme, such tokens should be tagged as B-PER, not I-PER. The model correctly predicts B-PER, but the evaluation scorer considers it a false positive because it does not match the provided ground truth.

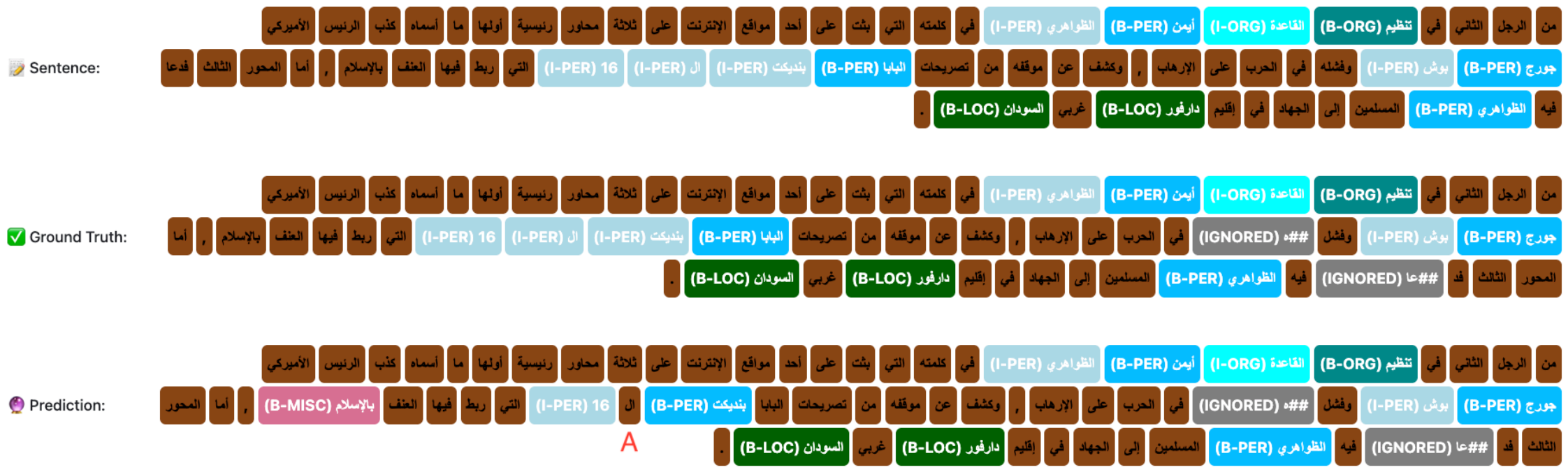

Figure 5.66: Example of a span-level misalignment in prediction. The model fails to capture the full PER entity, leading to a partially correct prediction.

This misalignment stems from annotation inconsistency rather than model error. Under IOB1, the ground truth span would be valid as shown in Figure 5.67 (A), but under IOB2 — which is the evaluation standard — the ground truth is incorrect as shown in Figure 5.67 (B). As a result, even though the model's prediction is correct, it is penalised during evaluation. This case illustrates how annotation noise can lead to what we call a ”false positive false positive”: the scorer reports an error, but that “error” is itself false, since the model's prediction is correct and the ground truth is inconsistent.

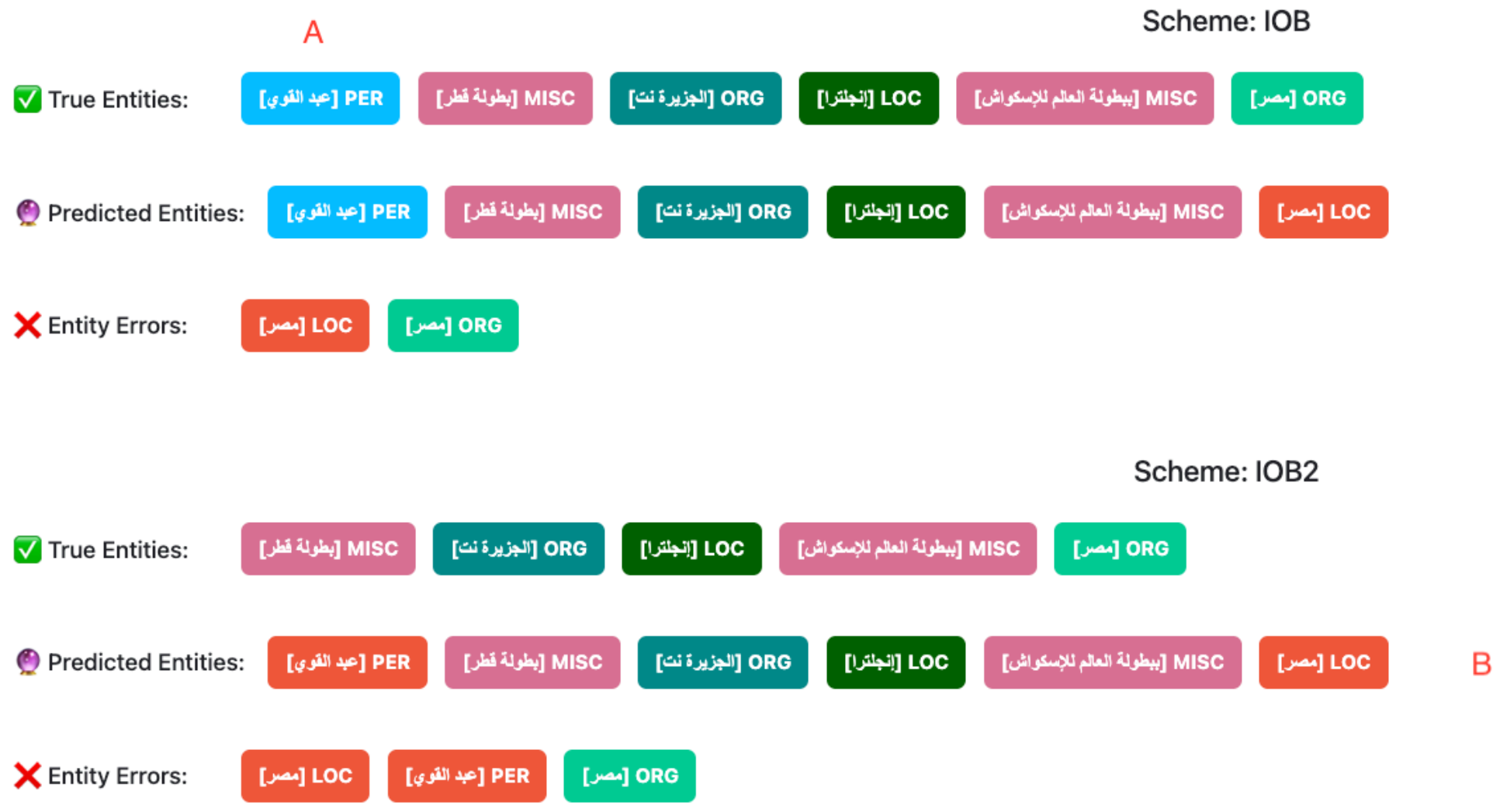

Figure 5.67: Example of misalignment due to incorrect span anchor in prediction. The token عبد is treated as the beginning of a PER entity by the model, causing a mismatch.

Together, these examples show two distinct sources of misalignment. In some cases, the gold annotations cut entity spans too short, leaving out tokens that the model reasonably includes within the entity. In other cases, the gold annotations mark the start of a span incorrectly, labelling what should be a beginning token as a continuation instead.

Both situations produce apparent errors in evaluation, but in reality they reflect inconsistencies in the annotation scheme rather than genuine mistakes made by the model.

### 5.5.7 Answer to RQ5: Patterns and Anomalies

This section addressed aspects **(a)–(f)** of the *Representation* component of RQ5. Below we summarise the main findings for each aspect.

**Aspect (a): Post–fine-tuning patterns in the representation space** We found that both languages developed clearer organisation by entity type after fine-tuning. Frequent classes formed dense regions, while punctuation, numerals, and context-dependent tokens produced scattered regions. English clusters were tighter and showed cleaner B/I separation; Arabic clusters showed greater overlap, particularly among ORG, MISC, and O.

**Aspect (b): Relationship between patterns and model errors** We found that error behaviours differ across Arabic and English in both form and distribution. In English, misclassifications are mostly confined to compact, well-defined regions, often reflecting context-sensitive ambiguities such as LOC vs ORG confusions or unusual contextual usage. In Arabic, errors are more varied and diffuse: they include sentence-initial anomalies, spelling-driven ambiguities (e.g., علي vs. على), and cross-boundary overlaps between B- and I- tags. These differences indicate that while English errors tend to align with local structural weaknesses, Arabic errors reflect a mix of linguistic variation, annotation inconsistencies, and broader context effects.

**Aspect (c): Behavioural metrics as explanations of anomalies** We identified similar correlation patterns between loss, confidence, and silhouette across Arabic and English, but the strength of these relationships differed. In English, the correlations were sharper — for example, loss decreased consistently with higher silhouette — while in Arabic, they were weaker and more scattered, reflecting annotation inconsistencies and sentence-level anomalies. This shows that although both models exhibit comparable tendencies, the Arabic model is less stable in how representational quality maps onto prediction behaviour. Importantly, we found that these behavioural metrics can act as practical indicators for active learning: tokens with high loss but good silhouette, or with high uncertainty in structurally coherent regions, provide useful cues for selecting ambiguous or misaligned examples for re-annotation.

**Aspect (d): Arabic-specific error patterns** We identified two recurrent Arabic phenomena. First, *sentence-start anomalies* (abrupt, malformed onsets) produced exclusion errors and high-confidence mistakes. Second, we observed cross-tag proximity, where some O tokens appeared inside entity regions yet were still predicted correctly. These patterns reflect dataset-specific artefacts and language-specific structures rather than model confusion alone.

**Aspect (e): Tokenisation effects and why Arabic underperforms despite English's higher tokenisation rate** We found that ambiguity from diacritic stripping and subword splits (e.g., first-subword confusion) led to misalignment between internal representations and outputs in Arabic. Ambiguous or unfamiliar forms often defaulted to O at the classifier, even when their embeddings clustered with an entity class. This shows that, although English is more heavily tokenised, Arabic suffers more from orthographic and diacritic ambiguity, as well as foreign-name segmentation, which together degrade calibration and representation–prediction alignment.

**Aspect (f): Annotation-scheme misalignments (gold vs. predicted)** We observed that ground-truth inconsistencies (e.g., IOB2 boundary violations and span trimming) often clustered near previously identified anomaly regions. In several cases, the model's predictions effectively "corrected" these inconsistencies (e.g., predicting B-PER where the gold label was I-PER), producing evaluation artefacts in the form of false positives that are in fact correct—what we call "false-positive false positives." We also identified subcomponent divergences, where the last hidden state aligned with one tag (e.g., I-PER) while the classifier favoured another (e.g., B-PER), reflecting boundary ambiguity and sentence-level noise.

**Conclusion** Overall, the representation space reflects both task structure and dataset artefacts. English shows tighter, task-aligned organisation; Arabic reveals noisier, language-and dataset-specific effects that propagate through calibration, boundaries, and evaluation. These findings motivate the next section on *Clustering Alignment with Annotation Structure*, where we quantify how well unsupervised structure tracks entity schemes across languages.

## 5.6 Clustering Alignment

In this section, we examine how well the clusters formed within the representation spaces of the Arabic and English models align with their gold-standard annotation structures. This analysis provides a complementary perspective on the internal organisation of each model, offering insight into how effectively entity types and span boundaries are separated within the learned representations.

We address aspects (a) and (b) of the *Clustering Alignment component of RQ5*, extending the previous analysis of global patterns by quantifying structural alignment. Specifically, we apply *K-Means* clustering with different values of $k$, each corresponding to a distinct level of annotation granularity:

- $k = 3$: coarse chunk-level grouping (e.g., B-, I-, O);
- $k = 4$: entity-type level grouping (e.g., PER, LOC, ORG, MISC);
- $k = 9$: full token-level labels (e.g., B-LOC, I-PER, B-ORG, etc.).

We begin by computing clustering–annotation alignment scores to evaluate how well the unsupervised clusters correspond to the labelled categories across both languages. We then inspect the representation spaces visually to assess whether the emergent cluster structures are consistent with gold annotations. Finally, we analyse the cosine similarity between the K-Means centroids (for $k = 9$) and the embedding vectors of each entity tag, to identify which entity categories are most distinctly represented.

Together, these analyses extend the exploration of representation structure by linking unsupervised geometric organisation to supervised label consistency, revealing how effectively each model internalises the task structure in Arabic and English.

### 5.6.1 Clustering Alignment Results

We evaluate how well the learned representation clusters align with the gold-standard annotation structure using standard clustering alignment metrics — homogeneity, completeness, and V-measure — across three clustering configurations ($k = 3, 4, 9$). These metrics quantify the degree to which the unsupervised clusters correspond to the annotated entity labels.

Figure 5.68 summarises the results. Across all levels of granularity, English consistently achieves higher alignment scores than Arabic, indicating that the English model's representation space more accurately reflects the labelled entity structure.

The $k = 3$ configuration, corresponding to coarse chunk-level groupings (B, I, O), yields the highest alignment scores for both languages. This suggests that both models capture span boundaries more reliably than finer-grained distinctions, likely due to the relative simplicity of the chunk-level structure. In Arabic, the $k = 4$ configuration (entity-type level) performs better than $k = 9$ (token-level), whereas in English, the $k = 9$ configuration achieves the best alignment. This pattern reinforces earlier findings that English develops clearer token-level organisation, while Arabic struggles with span granularity and label consistency.

### 5.6.2 Chunk-Level (Boundary-Based Clustering)

This level represents the coarsest annotation granularity, corresponding to the boundary tagging scheme: beginning (B), inside (I), and outside (O) a named entity. Figure 5.69 visualises how the model's representation space organises tokens according to these span boundary tags before clustering.

In Arabic, O tokens form a relatively clear and distinct region, but B and I tokens are more intermixed, suggesting weaker separation between entity boundaries. In contrast, the English model shows better separation across all three tags, especially between B and I.

**Clustering Results ($K = 3$)** We then apply K-means clustering with $k = 3$, aiming to measure the alignment of the clustering results with these boundary types, and how

**Arabic Clustering Alignment Results**

| K | Homogeneity | Completeness | V Measure |
|---|---|---|---|
| filter data... | | | |
| k=3 | 0.6376 | 0.6294 | 0.6335 |
| k=4 | 0.6841 | 0.4161 | 0.5175 |
| k=9 | 0.7428 | 0.2895 | 0.4166 |

**English Clustering Alignment Results**

| K | Homogeneity | Completeness | V Measure |
|---|---|---|---|
| filter data... | | | |
| k=3 | 0.8641 | 0.8349 | 0.8493 |
| k=4 | 0.71 | 0.7462 | 0.7277 |
| k=9 | 0.8886 | 0.7253 | 0.7987 |

Figure 5.68: Clustering alignment scores (Homogeneity, Completeness, V-measure) for Arabic and English models at different granularity levels.

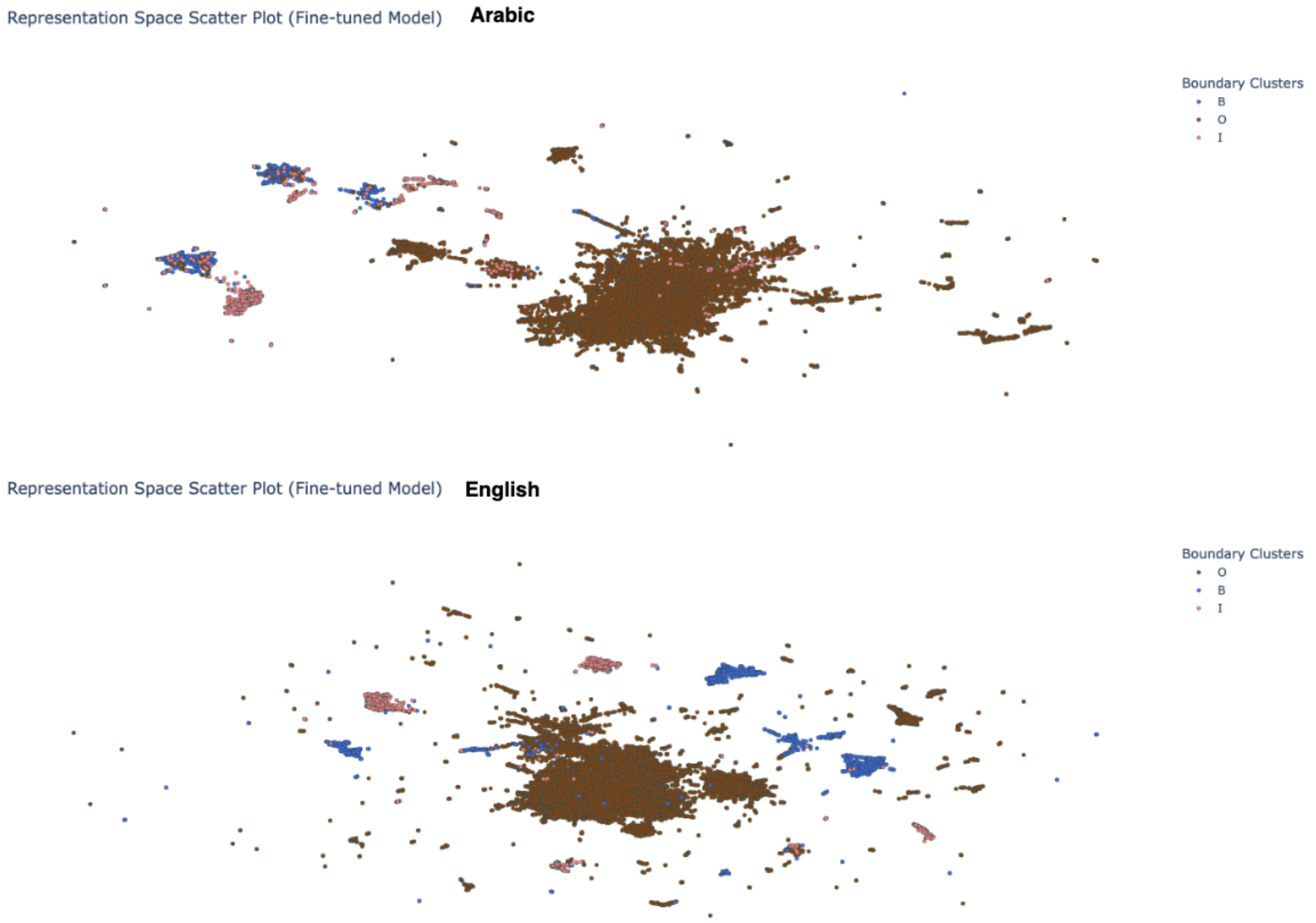

Figure 5.69: Representation space coloured by BIO span boundary tags (B, I, O) for Arabic (top) and English (bottom).

well KMeans can learn them from the internal representations. Figure 5.70 shows the resulting cluster assignments.

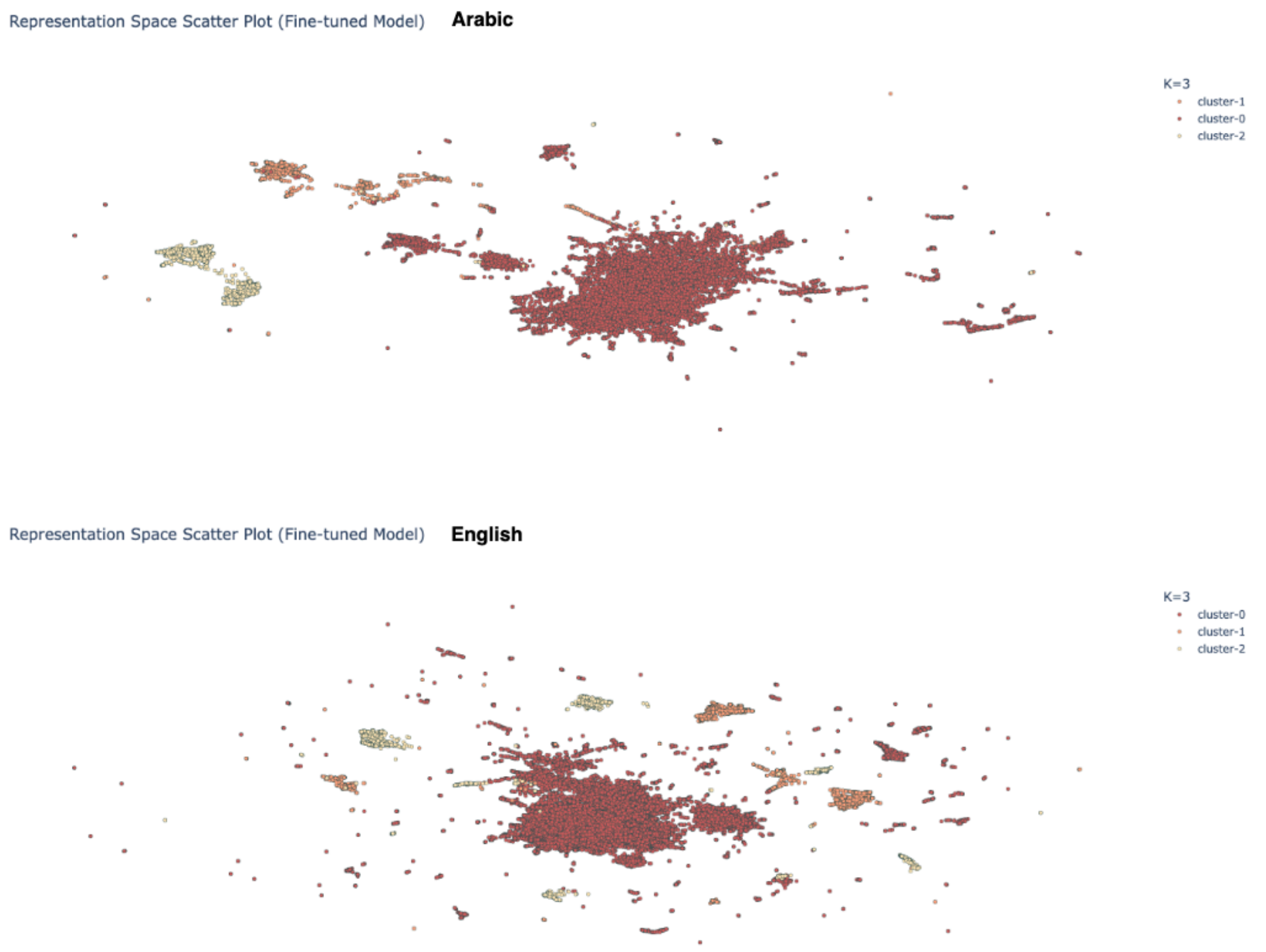

Figure 5.70: Clustering results using $K = 3$.

In Arabic, cluster 0 mainly captures O tokens, while B and I tokens are scattered across the remaining clusters, with no clear separation between them. In contrast, the English model produces cleaner clustering: one cluster corresponds mostly to O, and the other two largely capture B and I tags separately. This clearer structure explains the stronger clustering alignment scores for English at this level.

### 5.6.3 Entity-Level (Type-Based Clustering)

At this level, we assess whether clusters align with named entity types — PER, ORG, LOC, MISC, and O. Figure 5.71 visualises how the model's representation space reflects these entity type labels before clustering.

In Arabic, we observe noticeable overlap between entity types, particularly among MISC, ORG, and LOC, suggesting weaker type discrimination. The O class is somewhat more distinct but still shares boundaries with nearby entities. In contrast, the English representation space shows clearer separation between most entity types, especially for PER and LOC.

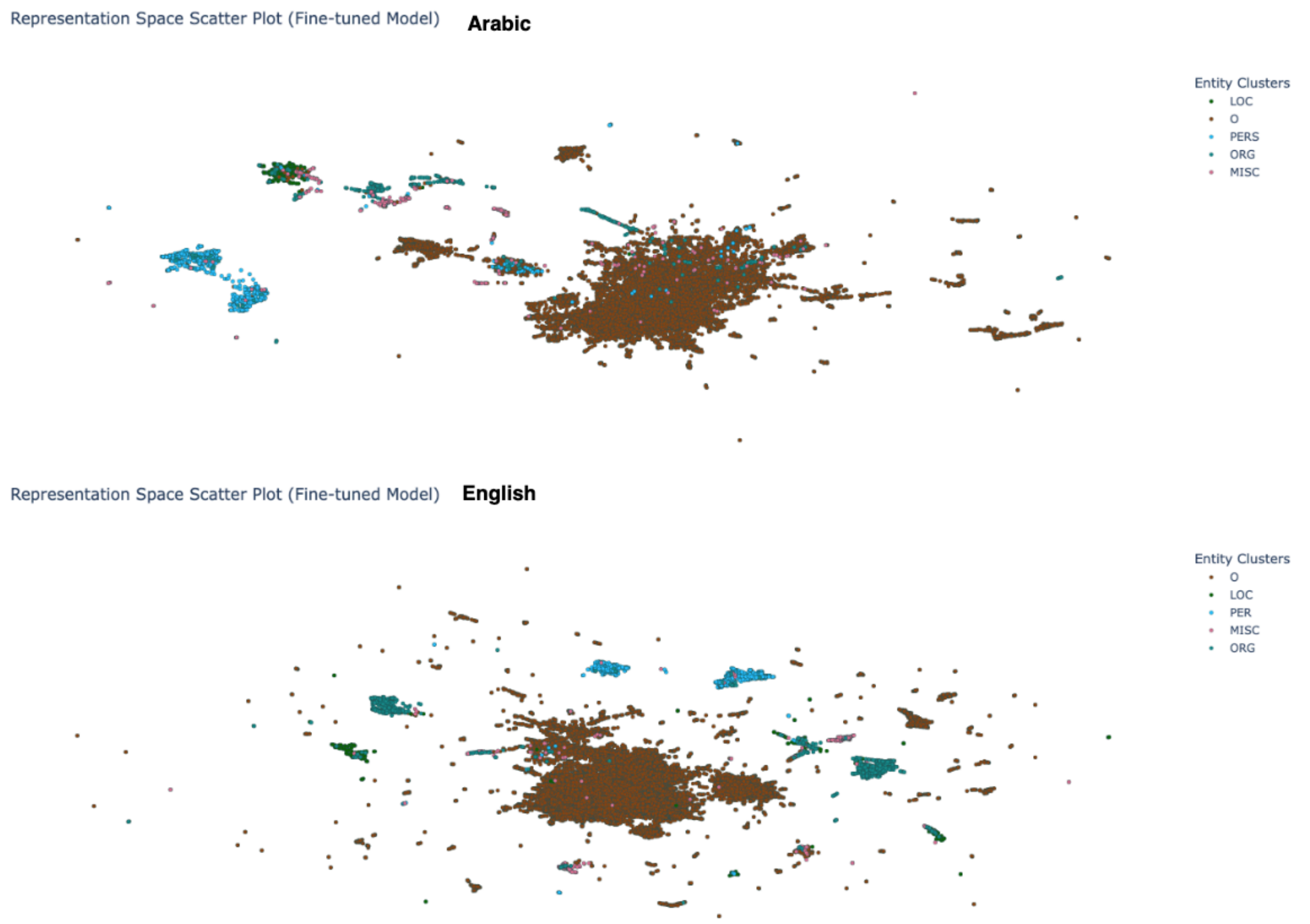

Figure 5.71: Representation space coloured by entity type annotations (PER, ORG, LOC, MISC, O) for Arabic (top) and English (bottom).

**Clustering Results ($K = 4$)** We then apply K-means clustering with $K = 4$, which approximates the main entity types excluding the O category. Presented in Figure 5.72.

In Arabic, PER tokens are mostly captured by cluster 3. However, the O class is split across clusters 0 and 2, and the main entity types — ORG, LOC, and MISC — are all grouped together in cluster 1, indicating poor separation. In English, the clusters are more coherent: each major entity type aligns more clearly with a distinct cluster. Notably, one English cluster isolates a subset of distant ORG tokens, suggesting that the model has learned internal substructure within that class.

### 5.6.4 Token-Level (Combined Boundary and Type Clustering)

Finally, we increase the number of clusters to $k = 9$ to capture both boundary and entity-type distinctions (e.g., B-PER, I-PER, B-LOC, I-LOC, etc.). Figure 5.73 visualises the annotated token-level structure in the representation space before clustering.

In Arabic, token-level tags often overlap, particularly among the B- and I- chunks of ORG, LOC, and MISC. The O class shows more spread, overlapping with other types.

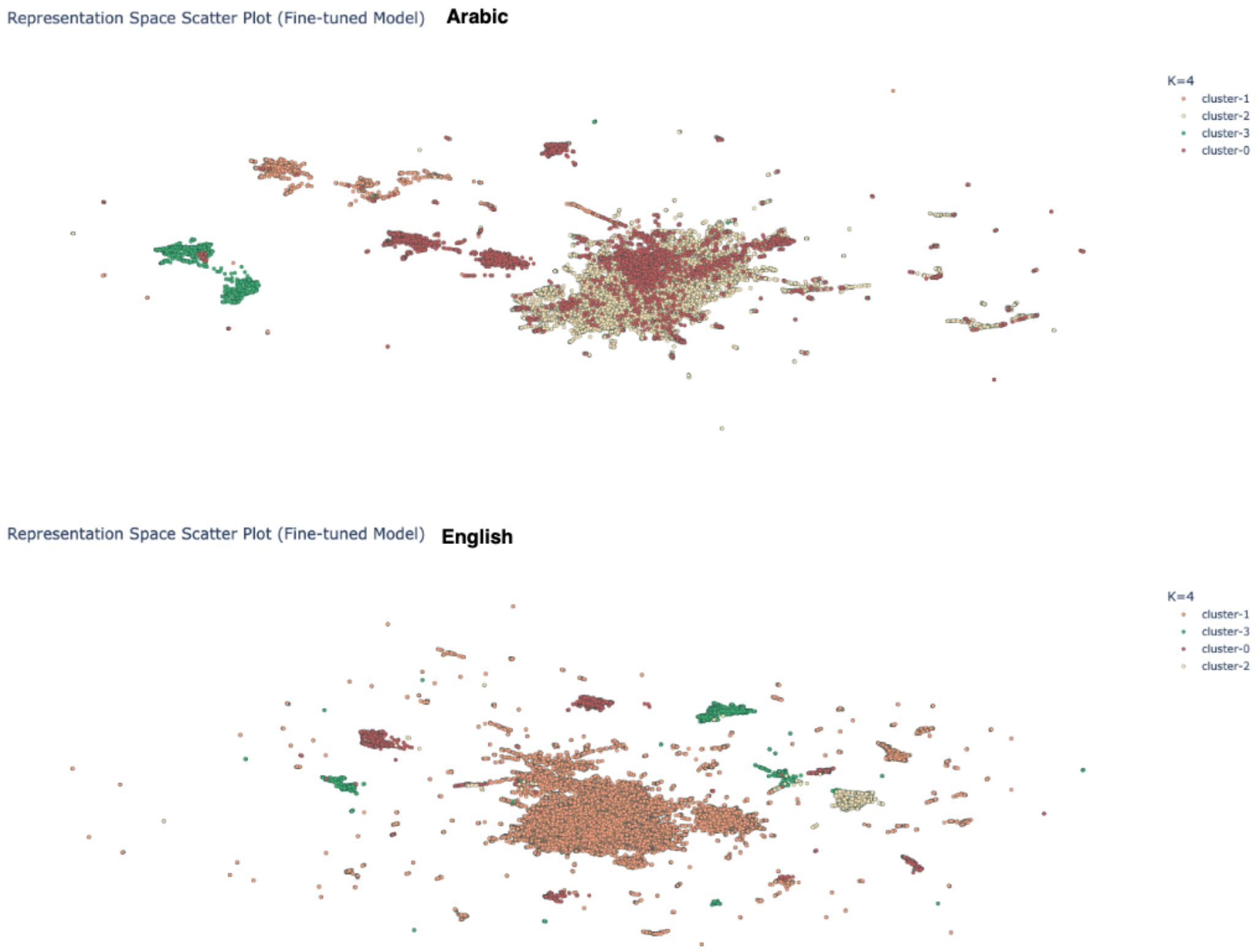

Figure 5.72: Clustering results using $K = 4$.

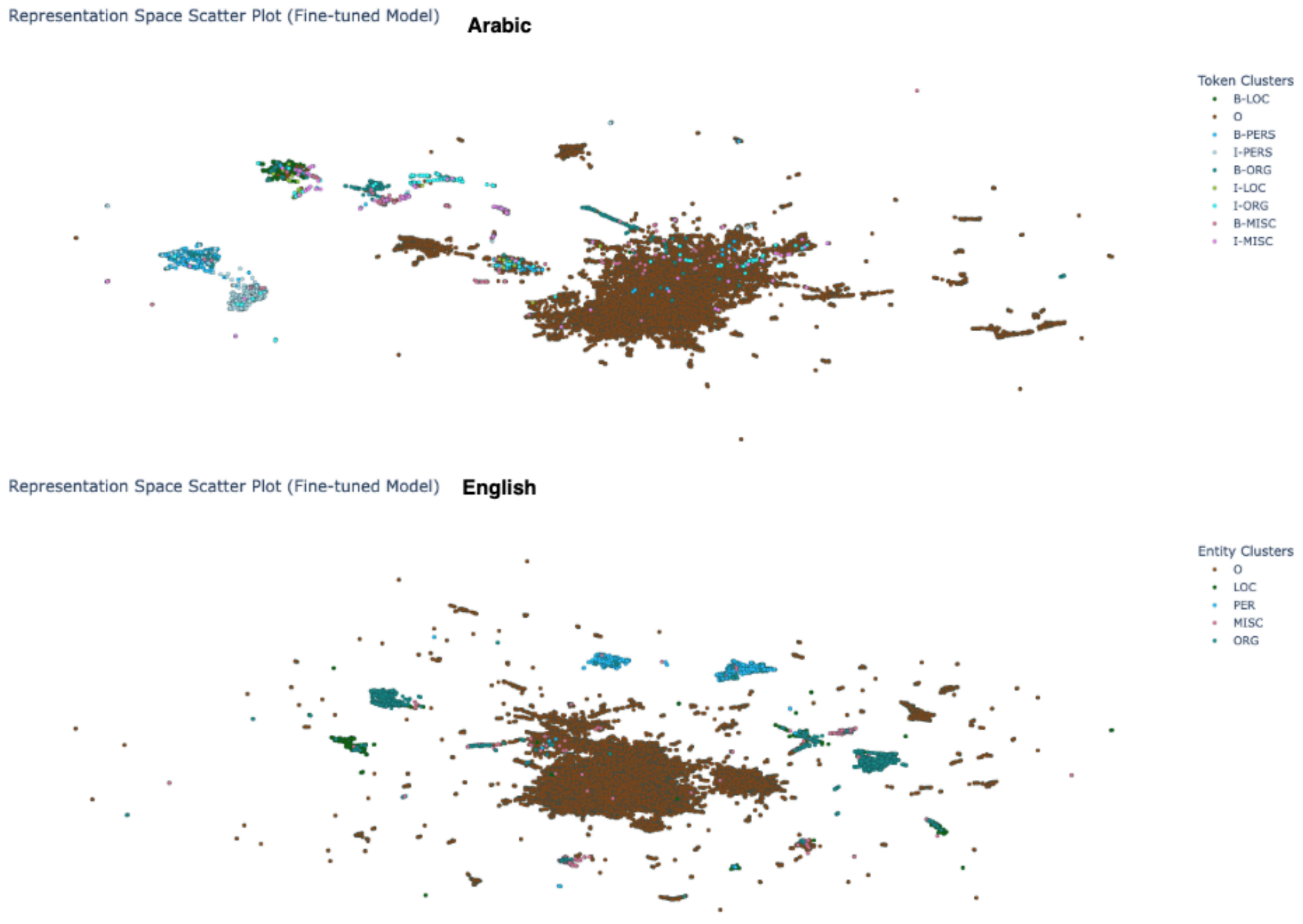

Figure 5.73: Representation space coloured by token-level labels (B-PER, I-LOC, etc.) for Arabic (top) and English (bottom).

English, by contrast, shows more distinct boundaries between token-level labels, especially for PER and LOC, reinforcing the stronger type-boundary structure observed earlier.

**Clustering Results ($K = 9$)** We then apply K-means clustering with $K = 9$, matching the number of token-level tag combinations. Figure 5.74 shows the resulting clusters.

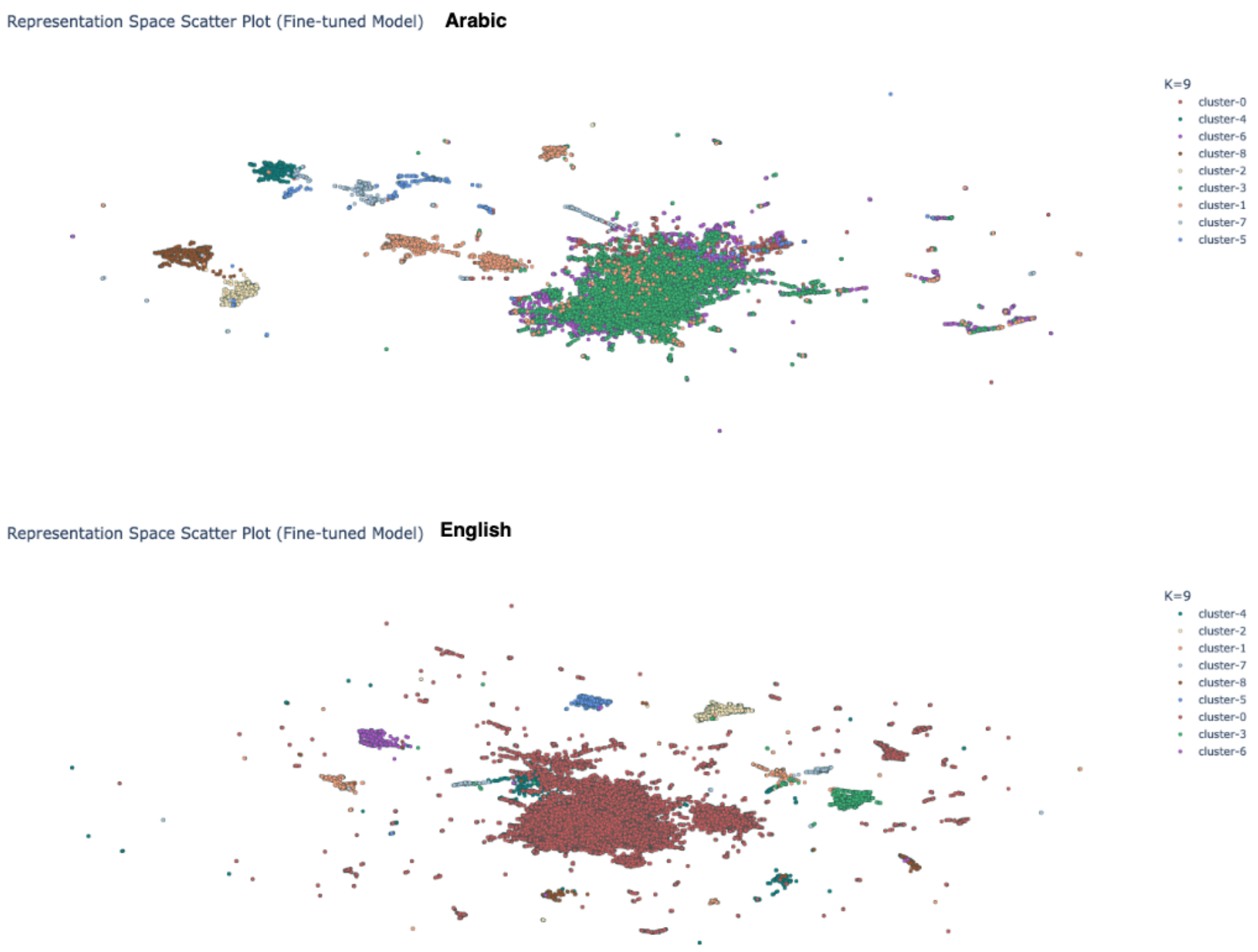

Figure 5.74: Clustering results using $K = 9$.

In Arabic, cluster 2 and cluster 8 primarily correspond to `B-PER` and `I-PER` tokens, while cluster 4 is largely aligned with LOC spans. However, clusters 6, 7, and 1 contain a mix of O tokens and various entity tags, highlighting the noisiness of Arabic representations at this level. Notably, cluster 1 also includes tokens tied to sentence-start anomalies and systemic errors, suggesting that these systematic errors are still captured by the model. English clusters show clearer alignment with token-level annotations. PER, LOC, and ORG tokens form tight, well-separated groups, with minimal overlap between entity classes and O tokens. This explains the higher clustering alignment scores for English under fine-grained configurations.

**Summary** This token-level clustering analysis reinforces previous findings: English models develop more consistent and separable internal structures across entity types and span boundaries, while Arabic models show higher overlap. These differences become more notable at finer annotation granularities, further supporting the hypothesis that Arabic's representational space is affected by noise from annotation inconsistencies and various artefacts.

### 5.6.5 Centroid–Entity Label Alignment Analysis

To further examine the alignment between learned clusters and annotated entity labels at $K = 9$, we compute the average cosine similarity between each cluster centroid and the token representations associated with each entity tag.

For each entity tag (e.g., B-LOC), we calculate the average similarity between its token embeddings and each cluster centroid ($\text{Centroid}_0$, $\text{Centroid}_1$, ..., $\text{Centroid}_8$). This produces a centroid–label alignment matrix, where each cell reflects how closely a given entity tag aligns with a specific cluster in the representation space.

The resulting similarity matrices for Arabic and English are visualised in Figure 5.75 as heatmaps. Rows represent entity tags, columns represent cluster centroids, and colour intensity indicates the degree of alignment based on average cosine similarity.

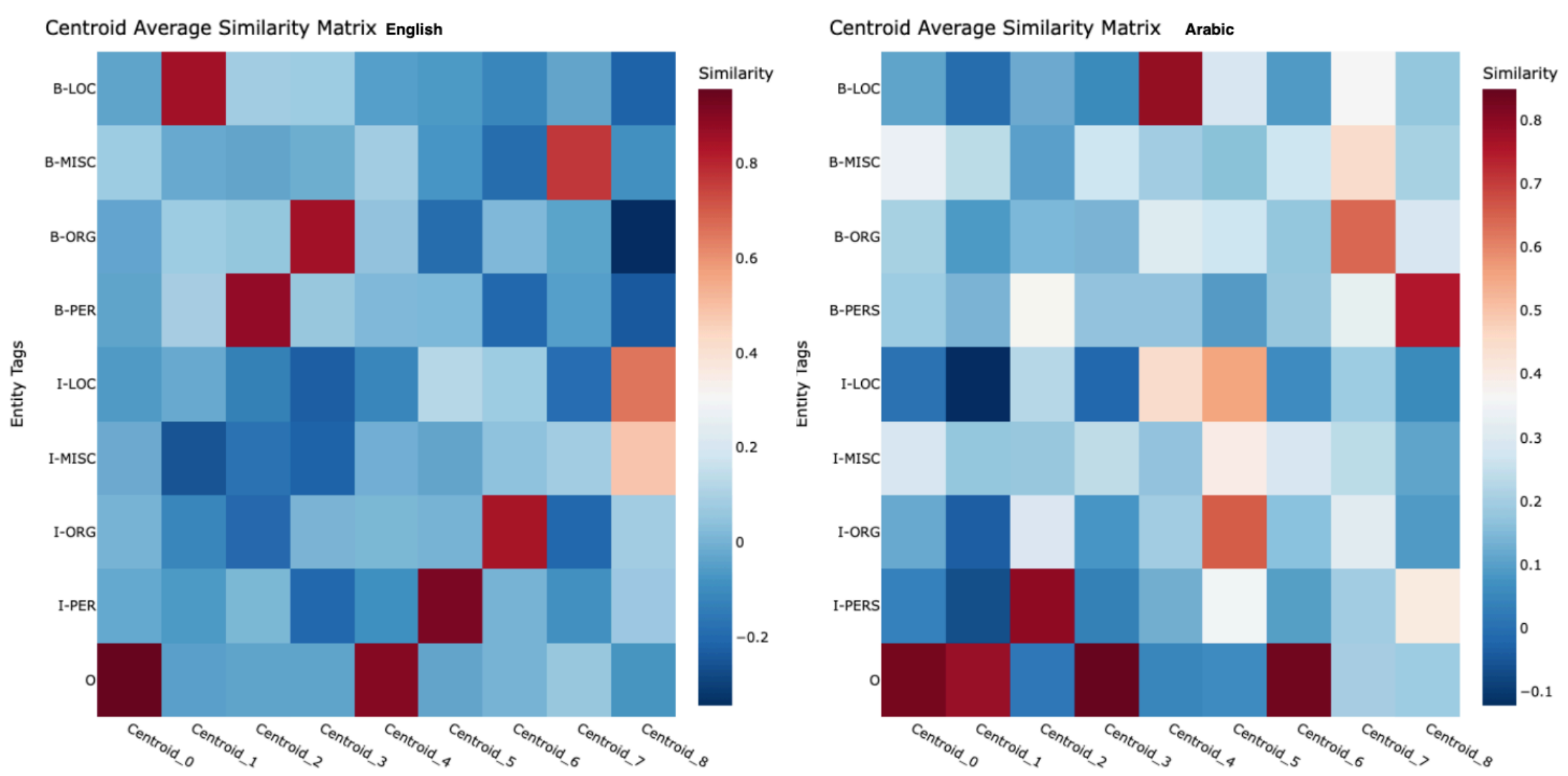

Figure 5.75: Centroid–entity label similarity heatmaps for English (left) and Arabic (right). Rows correspond to entity tags, columns to cluster centroids. Colour intensity indicates average cosine similarity.

As shown in Figure 5.75, English exhibits stronger alignment between clusters and entity labels. We observe clear patterns where certain entity tags are closely associated with specific centroids — for example, B-PER tokens are most similar to $\text{Centroid}_2$, and O tokens align well with $\text{Centroid}_0$ and $\text{Centroid}_3$. This reflects the well-structured clustering seen in English representations.

In contrast, Arabic shows a more diffuse pattern. While some centroids — such as $\text{Centroid}_8$ for B-PER and $\text{Centroid}_5$ for I-LOC — show partial alignment, the similarities are generally weaker and more spread out. The O tag in particular is scattered across several centroids, suggesting less consistency and more noise in the representation space.

These findings support earlier observations: English representations are more tightly aligned with the annotation structure, while Arabic representations remain noisier and more entangled due to annotation inconsistencies and token-level irregularities.

### 5.6.6 Answer to RQ5: Clustering Alignment

This section addressed aspects **(a)** and **(b)** of the *Clustering Alignment* component of RQ5, focusing on how well the internal representation clusters align with annotated entity structures across Arabic and English. Below we summarise the main findings.

**Aspect (a): Emergent clustering patterns and their correspondence to entity structure** We found that unsupervised clustering revealed consistent structural organisation aligned with the task's annotation scheme. At the coarse, boundary-based level ($k = 3$), both models captured the B/I/O boundary distinction, though Arabic showed greater intermixing between B and I tokens. At the entity-type level ($k = 4$), English formed clear clusters for PER, LOC, and ORG, while Arabic displayed overlap between semantically related categories such as ORG and MISC. At the fine-grained token level ($k = 9$), English preserved separable sub-clusters corresponding to both entity type and span boundary, whereas Arabic remained more diffuse, with intermixed boundaries between entity and non-entity regions. These patterns demonstrate that English representations internalise the hierarchical structure of the NER task more coherently, while Arabic embeddings encode a noisier, less differentiated structure.

**Aspect (b): Cross-lingual alignment and quantitative measurement** Across all clustering granularities, quantitative metrics (homogeneity, completeness, and V-measure) confirmed stronger alignment between clusters and gold annotations in English than in Arabic. Cosine-similarity analyses between cluster centroids and entity tags further showed that English centroids correspond closely to specific labels (e.g., B-PER, LOC), whereas Arabic centroids displayed weaker and more overlapping similarities, particularly for O tokens and boundary tags. These differences indicate that Arabic representations, affected by annotation inconsistencies and tokenisation artefacts, do not map cleanly onto the annotation structure, while English models preserve a clearer alignment between the organisation of the representation space and the labelled entities.

**Conclusion** Overall, the clustering alignment analyses show that while both models encode the span and entity structure, the English representations exhibit tighter, more label-consistent organisation. Arabic, by contrast, remains more fragmented and entangled, reflecting language- and data-specific irregularities. This difference reinforces the broader conclusion that English models maintain clearer alignment between internal structure and the annotation scheme, whereas Arabic models display instability driven by linguistic ambiguity, annotation noise, and inconsistent boundary encoding.

In the next section, we outline prior research that informs the analyses presented in this chapter, summarising key approaches to model interpretability and representational analysis. This background contextualises our findings within the broader literature.

## 5.7 Related Work

In this section, we provide an overview of prior work relevant to the behavioural analysis phase of our framework. This phase integrates quantitative behavioural diagnostics with visual analytics and interpretability techniques to produce global explanations of model behaviour. Rather than relying on isolated case studies, it links structural and behavioural metrics to interactive visualisations that reveal systematic patterns within the model's internal representations and outputs. Through this integration, the framework enables both high-level overviews and fine-grained instance analysis, supporting the investigation of cross-lingual variation and behavioural inconsistencies across data and model components in Arabic and English.

We next discuss how each aspect of this behavioural analysis phase—from global explanation techniques and behavioural measurement to local interpretation and error identification—relates to existing approaches in the literature. We highlight where our approach extends or diverges from prior work, particularly in its integration of quantitative diagnostics with interactive exploration. The section is organised as follows:

- **Interpretability Techniques** — situating our framework within broader approaches to model interpretability and representation analysis;
- **Interactive Dashboards** — reviewing prior work on visual analytics tools for NLP and model inspection;
- **Annotation Error Identification** — outlining existing methods for detecting and visualising annotation inconsistencies.

### 5.7.1 Interpretability Techniques

In this section, we investigate various interpretability techniques used in our framework. We start with layer-wise analysis of BERT layer, move to behaviour-driven metrics, and finally explore instance-based analysis using influence functions.

**Layer-Wise Analysis** In this type of analysis the focus is on understanding how Transformer models evolve their representations across layers. One example of this work is the work of Aken et al. (2019), who conducted a layer-wise analysis of BERT's hidden states, particularly in the context of question answering. Rather than relying on attention weights, they examined the hidden states directly, applying dimensionality reduction (e.g., PCA) and clustering methods (e.g., k-means) to visualise and measure changes across layers. Their findings showed that BERT's layers proceed through identifiable phases: early semantic clustering, entity linking, question-fact matching, and final answer extraction. They further applied probing tasks at each layer to track how linguistic information evolves. Notably, they found that fine-tuning has only a mild impact on early linguistic abilities but shapes task-specific processing in later layers.

Our work builds on this idea of layer-wise representational analysis, but extends it in several directions. While Aken et al. (2019) focus primarily on qualitative visualisation and probing for a specific QA task, our framework integrates layer-wise inspection with behavioural evaluation across languages (Arabic and English), examines clustering alignment at multiple annotation granularities (chunk, type, token), and uses behavioural metrics to guide exploration of representation structure.

**Attention Analysis** Another important techniques is the use of attention analysis to understand the impact of fine tuning on the model. Kovaleva et al. (2019) conduct an analysis of BERT's self-attention heads, classifying typical attention patterns (e.g., vertical, diagonal, block structures) and studying how these patterns change with fine-tuning on GLUE tasks. Their findings suggest that BERT exhibits a high degree of over-parameterization, with many heads learning redundant patterns, and that selectively disabling certain heads can even improve performance.

While their work provides valuable insights into attention redundancy and linguistic feature encoding, our analysis differs in several ways. We focus specifically on NER rather than classification or entailment tasks, and we extend the analysis to compare between Arabic and English. Furthermore, rather than analyzing attention patterns directly, we focus on the overall change without going into linguistic details.

**Behavioural Metrics** In terms of behavioural metrics we found the work of Swayamdipta et al. (2020), who track training dynamics of BERT by measuring confidence, variability, and correctness across epochs. These metrics are used to construct dataset cartography maps, which categorise examples as easy, ambiguous, or hard-to-learn, offering insights into how models learn over time. Their work provides a valuable insights on training behaviour, particularly in English classification tasks.

While we inspire from this approach, our framework differs in various ways. First, we develop behavioural metrics that are not tied to training epochs but instead reflect model–data interactions at inference time. Second, our metrics are computed over specific model and data subcomponents (e.g., representation layers, annotation types), allowing for more granular analysis. Third, we apply these metrics in a comparison between two languages in NER, we use them to guide qualitative structural interpretation.

Finally, we integrate these behavioural metrics into interactive visualisations that directly link model behaviour to latent representational space, offering a more integrated diagnostic approach.

**Example-Based Analysis** Recent work by Jain et al. (2022) extends influence functions to structured prediction tasks, such as NER. They propose segment-level influence, which quantifies how specific spans in the training set affect span-level predictions in test instances. Their method approximates influence using gradients of conditional losses, and shows utility in identifying systematic annotation inconsistencies and uncovering training segments responsible for common NER errors.

Inspired by this line of work, we use cosine similarity within the instance level analysis to find semantically similar examples based on selected token representations. While our current system relies on simple cosine similarity, we aim to integrate more complex approaches such as KNN, and NER specific metrics similar to the one used in this paper as discussed in Section 6.6.

**Attribution-Based Explanations** Feature attribution methods aim to identify which input tokens most influence a model's predictions and are widely used in interpretability research, particularly for sequence classification tasks. A recent survey by Zhao et al. (2024) indicate that these techniques are used across a variety of tasks but mainly focuses on classification and generation, with limited discussion of sequence labelling. Ruder et al. (2022) further observe that while feature attribution methods can effectively evaluate interpretability in sequence classification (text classification), they are not directly applicable to sequence labelling tasks such as NER, where each token is associated with an individual label. Among existing work, AllenNLP Interpret Wallace et al. (2019) is the only tool we are aware of that adapts token-level saliency and input reduction methods for structured tasks like NER. Based on these insights, we didn't include attribution methods.

**Interpretability for Arabic Models** To our knowledge the only Arabic interpretability work is the work of Abdelali et al. (2022) which presents the first in-depth interpretability analysis of Arabic transformer models. They apply layer- and neuron-level probing techniques to a set of Arabic and multilingual models, evaluating how linguistic properties such as morphology and dialectal variation are encoded. Their results show that word morphology is captured in lower layers, while dialectal features emerge in higher layers. Neuron analysis reveals polysemy in embeddings and specialization in deeper layers.

Our work complements and extends this line of research in several key ways. First, we focus on structured prediction (NER) rather than POS or dialect ID. Second, we conduct a cross-lingual analysis involving both Arabic and English, exposing structural misalignments between the languages. Third, we focus on task specific distinctions.

### 5.7.2 Interactive Visualisation Tools

Several recent approaches have proposed interactive dashboards to support model interpretability and error analysis. T3-Vis Li et al. (2021) offers a visual analytic framework for exploring hidden states, attention patterns, and training dynamics, based on requirements derived from NLP researchers. LIT Tenney et al. (2020) provides an extensible interface for local explanations, aggregate metrics, and counterfactual generation, across a range of NLP tasks including classification and sequence generation. Similarly, InterpreT Lal et al. (2021) focuses on layer-wise embedding visualisation and attention analysis, applied to tasks like aspect-based sentiment analysis and co-reference resolution.

Most of these frameworks are primarily model-centred and designed for English-only classification or generation tasks, with limited support for structured sequence labelling or multilingual analysis. In contrast, our framework, DeformAr, is built specifically for Named Entity Recognition (NER) and designed to analyse cross-lingual discrepancies between Arabic and English models. Unlike prior dashboards, which are organised around user requirements, DeformAr decomposes the NER pipeline into data and model components and investigates their interactions during fine-tuning and inference.

This design enables deeper subcomponent coverage by linking quantitative metrics (e.g., confidence, loss, consistency) with interpretability views (e.g., representation space, token-level behaviour) through interactive visualisations. Rather than combining a broad set of techniques into a single dashboard, our framework structures the analysis around how representations and behaviours emerge and diverge between languages. It also mitigates the "square-one bias" (Sun et al., 2021) by supporting structured, multilingual, component-level diagnostics instead of focusing solely on model internals. Together, these elements integrate quantitative analysis and behavioural analysis into a unified workflow in which metric-driven insights guide visual inspection and cross-lingual comparison.

**Annotation Error Identification** Several recent works have revisited standard NER benchmarks to identify and correct systematic annotation errors. CLEANANERCorp Al-Duwais et al. (2024) presents a corrected version of the Arabic ANERcorp dataset, using a combination of automatic detection (via CLEANLAB) and manual re-annotation to fix label inconsistencies, typographic errors, and tagging ambiguities. Similarly, Reiss et al. (2020) apply ensemble disagreement methods to identify over 1,300 incorrect labels in the CoNLL-2003 corpus, producing a corrected version with improved evaluation reliability. More recently, CLEANCoNLL Rücker and Akbik (2023) introduces a multi-phase annotation pipeline that combines entity linking, cross-checking with model disagreement, and manual curation to correct 7% of CoNLL-03 labels and reduce annotation error rates dramatically.

Our framework, while not explicitly designed for annotation correction, has revealed similar patterns of systemic label noise — particularly in Arabic. Through behaviour-guided structural analysis, we identify errors previously noted in ANERCorp and extend their interpretation by tracing how these issues affect different model subcomponents, such as representation layers and output structures. Unlike previous approaches that depend on external knowledge bases or manual review, our method integrates behavioural metrics and instance-level analysis to both detect and explain label inconsistencies. This enables more transparent diagnosis and demonstrates how annotation noise contributes to cross-lingual discrepancy. We argue that such structural diagnostics can inform more targeted and standardised annotation correction workflows in future NER corpus development.

### 5.7.3 Summary

In this section, we provided an overview of the interpretability techniques, interactive visualisation tools, and error identification methods that inform our framework. We positioned our approach within the existing literature, highlighting both methodological inspirations and key differences.

These distinctions can be summarised in three main aspects: (1) our framework supports cross-lingual analysis, with particular emphasis on Arabic; (2) it is specifically designed for structured prediction tasks such as NER; and (3) it adopts a component-based view of language, examining interactions between data and model components, and linking quantitative metrics with interactive visualisations.

In the following section, we summarise the key findings uncovered through the behavioural analysis phase and discuss their broader implications.

## 5.8 Discussion

This section synthesises the findings from the behavioural analysis presented in this chapter. It focuses on how token-level behaviours, representational structures, and annotation alignments interact to explain the performance discrepancies observed between Arabic and English NER systems.

Building on the quantitative findings from Chapter 4, this chapter addressed **Research Question 5 (RQ5)**, which examined how data, model, and evaluation components interact through representation and behavioural patterns. Unlike the previous chapter, which analysed these components separately, this chapter adopted a behavioural analysis approach to connect them within a unified interpretive framework.

RQ5 was divided into three main analytical sections: (1) the *impact of fine-tuning* on the evolution of internal representations; (2) the emergence of *patterns and anomalies* within the representation space; and (3) the degree of *clustering alignment* between learned representations and gold-standard annotation structures.

The discussion is organised into three parts. First, we summarise the key findings of each analytical component of RQ5. Second, we integrate these findings to identify the main contributing factors underlying the observed discrepancies in performance between Arabic and English systems—such as tokenisation artefacts, annotation inconsistencies, and cross-component misalignments—and discuss how they collectively shape model behaviour. Finally, we outline the specific contributions of this behavioural analysis phase within the broader diagnostic framework.

### 5.8.1 RQ5 Summary: Training Impact

This section addressed aspects **(a)** and **(b)** of RQ5, examining how fine-tuning alters internal attention behaviour and token representations in Arabic and English NER models. Fine-tuning reorganises the representation space, producing clearer entity-based structure and more task-aligned attention patterns, especially in the upper layers. Although the learned attention parameters remain relatively stable, the changing attention scores indicate shifts in how contextual relationships between tokens are encoded.

Cross-linguistic differences were clear. English models showed distributed, stable adaptation across layers and heads, yielding well-separated representations. Arabic models, in contrast, exhibited more localised adjustments, including collapsed attention heads and larger shifts in final token embeddings. These differences suggest that fine-tuning behaves differently across languages: English develops coherent, task-aligned structure, while Arabic remains more variable and context-dependent.

This asymmetry likely reflects data effects: English benefits from larger, more consistent training data that supports robust adaptation, whereas Arabic's smaller and more heterogeneous corpus promotes heuristic learning and surface-level fitting. These findings underscore the need to interpret fine-tuning behaviour separately across languages and motivated the following analysis of global *Patterns and Anomalies* in the representation space.

### 5.8.2 RQ5 Summary: Patterns and Anomalies

This section addressed aspects **(a)**–**(f)** of RQ5, examining how patterns and anomalies emerge within the fine-tuned representation space and how they differ across Arabic and English. The analysis showed that fine-tuning produced clearer structural organisation in both languages, but English representations were more compact and label-consistent, while Arabic embeddings remained diffuse and overlapping, especially among ORG, MISC, and O tags. Misclassifications in English were largely confined to small, well-defined regions (e.g., LOC–ORG confusions), whereas Arabic errors were more dispersed and reflected sentence-level irregularities, spelling ambiguities, and boundary inconsistencies. Correlation analysis of behavioural metrics (loss, confidence, silhouette) revealed broadly similar trends across languages but weaker, noisier relationships in Arabic, highlighting instability in how representational quality maps onto prediction behaviour.

We also identified Arabic-specific anomalies such as malformed sentence start and cross-tag proximity effects, where O tokens occurred within entity regions but were still predicted correctly. Tokenisation further amplified these issues: diacritic removal and subword splitting introduced ambiguity that led to defaulting to O, even when embeddings aligned with an entity class. Finally, annotation inconsistencies (e.g., IOB2 boundary violations) often coincided with these anomaly regions, with models occasionally "correcting" gold errors—producing false positives that were linguistically accurate.

Overall, the representation space revealed that English models develop tighter, task-aligned structures, while Arabic models remain fragmented by annotation noise, orthographic variation, and tokenisation artefacts. These findings motivateed the next analysis on *Clustering Alignment with Annotation Structure*, where structural correspondence between unsupervised clusters and gold labels is quantified.

### 5.8.3 RQ5 Summary: Clustering Alignment

This section addressed aspects **(a)** and **(b)** of RQ5, examining how well the internal representation clusters align with annotated entity structures across Arabic and English. The analysis showed that unsupervised clustering broadly reflected the task's annotation hierarchy, but with clear cross-lingual differences. At the coarse boundary level ($k = 3$), both models captured the B/I/O distinction, though Arabic showed greater intermixing between B and I. At the entity-type level ($k = 4$), English produced distinct clusters for PER, LOC, and ORG, while Arabic exhibited overlap between semantically related types such as ORG and MISC. At the fine-grained token level ($k = 9$), English maintained separable subclusters for both boundary and type, whereas Arabic remained diffuse and entangled.

Quantitative metrics (homogeneity, completeness, V-measure) confirmed these trends: English achieved higher alignment scores at all granularities, indicating that its internal structure more accurately mirrors the annotation structure. Cosine-similarity analysis between cluster centroids and entity tags further showed clear centroid–label correspondence in English (e.g., B–PER, LOC), but weak, overlapping associations in Arabic, especially for O and boundary tokens.

Overall, both models encode span and entity information, but English representations are more coherent and label-consistent. Arabic representations remain fragmented by linguistic ambiguity, tokenisation artefacts, and annotation noise, leading to weaker structural alignment with gold annotations. These findings reinforce that English models internalise task structure more effectively, while Arabic models exhibit instability rooted in language- and data-specific irregularities.

### 5.8.4 Contributing Factors Identified in This Chapter

Throughout this chapter, we identified several factors that help explain the observed differences in model behaviour between Arabic and English:

- **Representation Structure:** English models develop clearer and more coherent internal representations, with distinct clusters for entity types and boundaries. In contrast, Arabic representations are more fragmented and overlapping, particularly at finer levels of annotation granularity.

- **Systematic Annotation Errors:** While both languages contain annotation inconsistencies, our analysis shows that Arabic errors — especially sentence-start anomalies— have a higher structural impact. These issues distort the representation space and contribute to misalignment between the model's output predictions and its internal representations.

- **Tokenisation Effects and Diacritic Removal:** Arabic preprocessing introduces ambiguity due to subword splitting and the removal of diacritics, which weakens semantic distinctions between tokens. These tokenisation artefacts result in representational and behavioural inconsistencies not commonly seen in English.

- **Over-fitting :** Arabic models exhibit signs of over-fitting to surface-level patterns. Changes in attention and embedding structures remain localised, and behavioural metrics show that the model often makes confident but incorrect predictions — particularly on repeated anomalous inputs — suggesting limited contextual generalisation.

- **Cross-Component Misalignment (Driven by the O Tag):** One of the most prominent issues in Arabic is the misalignment between the model's representation space and its output layer, especially around the O tag. Tokens correctly classified as O are often embedded within entity clusters, reflecting internal conflict across subcomponents. This misalignment is a key driver of prediction inconsistency and noisy metric behaviour.

- **Training Data Differences:** English models benefit from larger and more consistent datasets, supporting more generalisable representations. Arabic, with fewer and noisier examples, leads to tighter reliance on dataset-specific patterns, contributing to the discrepancies observed across both languages.

## 5.9 Contributions

This chapter presented the *behavioural analysis phase* of the DeformAr framework, which focuses on interpreting the internal behaviour of NER models across languages. Extending beyond traditional evaluation scores and correlation-based trends, this phase combines interactive diagnostics, structural inspection, and behavioural metrics to reveal model–data interactions that are not easily captured by standard evaluation methods. The key contributions of this phase are summarised below.

**First: A diagnostic tabs for structured prediction interpretability.**
We develop an interactive tabs designed specifically for structured tasks like NER, enabling model inspection at global, token-level, and instance-level granularity. Unlike generic interpretability tools, these tabs support cross-lingual analysis and integrates behavioural metrics with embedding-space visualisations, forming the core of the behavioural analysis phase.

**Second: Cross-component behavioural analysis of NER behaviour.** Within this phase, we implement a cross-component analysis layer that decomposes NER systems into data and model subcomponents and traces how inconsistencies (e.g., annotation noise, tokenisation errors) propagate through the representation space and affect output behaviour. This provides the visual counterpart to the quantitative diagnostics developed in earlier stages of the framework.

**Third: Structural explanation of Arabic–English performance discrepancies.**

- We show that English models develop clearer, more separable representations, while Arabic representations remain noisier and less structured.
- We trace these discrepancies to specific structural and behavioural anomalies, including annotation inconsistencies and tokenisation artefacts unique to Arabic.

**Fourth: Demonstrating the interpretability value of behavioural metrics.** We show how loss, confidence, uncertainty, and silhouette scores not only reflect prediction behaviour but also help identify structural inconsistencies and systematic errors. Within the behavioural analysis workflow, these metrics act as dynamic diagnostic indicators that guide inspection and support active learning strategies.

**Fifth: A clustering-based evaluation of representation alignment.** We apply K-Means clustering and centroid analysis to quantify how well token-level representations align with annotation structures at multiple granularities. This provides a structural validation mechanism within the behavioural analysis phase, complementing the behavioural insights from earlier components.

**Sixth: Linking annotation noise to model misalignment and overconfidence.** Through instance-level and behavioural visualisations, we show how annotation inconsistencies—especially in Arabic—distort the representation space, induce cross-component misalignment, and produce confidently incorrect predictions. These findings link visual evidence to the data-related issues identified in the cross-component phase.

**Seventh: Discovery and explanation of systemic annotation issues.** The behavioural analysis tools rediscovered known annotation inconsistencies in the Arabic ANERcorp dataset and uncovered new ones. More importantly, they visualise how these issues propagate through model internals and affect representational stability, bridging dataset artefacts with downstream behavioural impact.

**Eighth: Integrating behavioural analysis with quantitative cross-component analysis.**
We introduce a novel interactive mechanism that connects structural representations, behavioural metrics, and instance-level explanations across both model and data components. By integrating visual exploration with behavioural metrics, this phase bridges the gap between behavioural and cross-component analysis, revealing how annotation noise, tokenisation artefacts, and representational inconsistencies jointly contribute to cross-lingual performance discrepancies.

Together, these contributions advance the behavioural analysis phase of the DeformAr framework, providing a practical foundation for interpreting and diagnosing structured prediction models across languages. To our knowledge, this is the first Arabic-specific, component-based interpretability dashboard for structured prediction—a step forward in enabling detailed diagnostic evaluation for under-resourced languages.

# General Discussion and Conclusions 6

In this chapter, we provide a comprehensive synthesis of the work presented in this thesis. We begin by summarising the overall diagnostic framework, revisiting its key motivations, structure, and capabilities. We then discuss the framework's novelty in relation to existing approaches to NER evaluation and interpretability, followed by a concise summary of the research questions and key findings that emerged across the analysis chapters. Building on these findings, we synthesise the main contributing factors identified across the cross-component and visual-analytics analyses, outlining how they jointly explain the observed Arabic–English performance discrepancies. The chapter then turns to the framework's current limitations and possible extensions, before outlining future directions for applying this approach to other languages, NLP tasks, and model architectures. Finally, we conclude with reflections on the broader implications of this research.

## 6.1 DeformAr: A Diagnostic Framework for NER

The DeformAr framework was developed to address three key gaps: (1) the need to understand the performance discrepancy between Arabic and English NER systems, (2) the lack of interpretability tools specifically tailored for Arabic, and (3) the absence of structured, component-based approaches to interpretability in sequence-labelling tasks such as NER. Unlike existing tools that combine multiple interpretability techniques into generic dashboards, DeformAr adopts a comparative, component-oriented perspective.

It contrasts Arabic—a morphologically rich and under-resourced language—with English, which is more widely studied and better resourced. Instead of focusing solely on model outputs, the framework decomposes each language into data and model components and traces how these evolve across the fine-tuning pipeline. This layered design enables the identification of interaction points that explain how performance gaps emerge across languages.

To achieve this, **DeformAr** begins with a **cross-component quantitative analysis** phase that systematically measures discrepancies across the identified components. This provides an empirical foundation for identifying which metrics, components, and interactions are most critical. The results of this phase then guide a targeted **interpretative behavioural analysis** phase, allowing interpretability methods to focus on the most informative areas for explaining cross-lingual differences.

The outcome is a multi-layered diagnostic framework that unifies the cross-component quantitative and interpretative behavioural analysis phases within a single workflow. It enables both focused exploration of individual components and examination of their interactions, effectively bridging the gap between dataset characteristics and model behaviour. The following section discusses the novelty of this approach in relation to existing work.

## 6.2 Where is the Novelty?

The novelty of our framework can be seen across three key dimensions: where we looked, what we chose to examine, and how we examined it.

- **Where we looked:** This work presents the first comprehensive Arabic-specific, component-based interpretability framework for NER. While most existing tools are developed for English and focus on text classification tasks, our approach is designed to uncover structural and behavioural issues in a low-resource setting — a space that remains under-explored.
- **What we chose to examine:** Rather than building a general-purpose interpretability dashboard that combines multiple techniques, we grounded our analysis in a cross-lingual comparison between a well-studied language (English) and a less-explored one (Arabic). We focused on systematically decomposing each system into **data** and **model** subcomponents and tracing how their interactions evolve throughout the fine-tuning pipeline — identifying where performance divergence emerges.

- **How we examined it: DeformAr** integrates a **cross-component quantitative analysis phase** with an **interpretative behavioural analysis phase**. The quantitative layer goes beyond surface metrics, decomposing performance across subcomponents (e.g., span-level error types, tokenisation ambiguity, representation separability), while the interpretative layer leverages behavioural metrics and interactive visualisations to reveal structural patterns and explain model behaviour — an integration not commonly achieved in existing evaluation frameworks.

In summary, the novelty of our approach lies in the comprehensive component-level analysis across both data and model components for Arabic and English. By systematically linking these components through behavioural metrics and interactive interpretability techniques, we uncover relationships that, to our knowledge, have not been previously explored in existing tools. More importantly, we established rich connections between these components — tracing how structural issues in the data (e.g., annotation noise, tokenisation artefacts) propagate through the model's internal representations and influence prediction behaviour. To our knowledge, no prior tool has provided this level of cross-component analysis within a structured prediction setting, particularly for under-resourced languages like Arabic.

## 6.3 Summary of Research Questions and Key Findings

Across all stages of the analysis, **DeformAr** addressed six interrelated research questions (RQ0–RQ5), together forming a coherent diagnostic account of Arabic–English NER evaluation.

**RQ0** established how the DeformAr framework was designed and implemented at both the **conceptual** and **technical** levels. From a conceptual perspective, the NER pipeline was decomposed into three diagnostic components — **data**, **model**, and **evaluation** — each consisting of subcomponents that exhibit distinct behavioural properties while also interacting with one another. This decomposition enables structured component-level analysis and supports systematic diagnosis of where and why performance discrepancies emerge. From a technical perspective, we introduced the dashboard phase, which is divided into two analytical modes: the **Cross-Component Analysis** phase, which performs structured comparison using metric-based summaries across components, and the **Behavioural Analysis** phase, which enables deeper diagnostic investigation using interpretability techniques and token-level exploration. Together, these phases implement DeformAr as an interactive diagnostic framework that connects metric-based evaluation with behavioural explanation.

**RQ1** demonstrated that data characteristics introduce language-specific sources of difficulty in NER. Arabic exhibits higher lexical sparsity and greater annotation ambiguity, particularly for MISC and ORG entity types. Tokenisation affects both languages but in different ways: in English, subword splitting compresses the vocabulary, reducing type diversity and increasing repetition of common subwords. In Arabic, subword tokenisation does not reduce lexical diversity to the same extent—morphological variation continues to generate many unique surface forms, so type–token imbalance and sparsity persist. These structural effects propagate downstream. In both languages, entity tag overlap and ambiguity increase, but the impact is more severe in Arabic, where sparsity and annotation inconsistency lead to greater confusion with the O class and span-boundary. English displays more ambiguity and a more balanced entity distribution. Overall, RQ1 shows that tokenisation behaviour, lexical sparsity, annotation inconsistency, and ambiguity are primary data-level contributors to the cross-lingual performance gap.

**RQ2** examined model-level behaviour and showed that internal representations and prediction dynamics differ significantly across languages, reinforcing the performance gap observed in RQ1. English develops more compact and separable representation spaces after fine-tuning, as reflected by higher silhouette scores and clearer cluster boundaries between entity types. In contrast, Arabic representations exhibit weaker structural organisation due to higher lexical sparsity and morphological variability, which reduce representation coherence and increase class overlap—particularly between entity tags such as MISC, ORG, and O. This representational instability is reflected behaviourally in the model output layer: Arabic shows higher variance in prediction confidence and higher mean loss, especially for low-frequency and morphologically complex entities, while English produces more stable confidence distributions and lower loss. These findings demonstrate that representation structure and predictive stability degrade more severely in Arabic than in English, indicating that model behaviour amplifies structural weaknesses originating in the data.

**RQ3** analysed evaluation behaviour and showed that the choice of evaluation scheme and span-decoding strategy has a direct impact on the reported Arabic–English performance gap. At the token level, the apparent difference between the two languages is larger; however, at the span level this gap narrows due to averaging effects and the influence of certain entity tags, particularly I-MISC, which disproportionately lowers macro-average scores compared to micro-average scores. Changing the annotation scheme also had a measurable effect on performance, especially for Arabic: switching from IOB1 to IOB2 improved precision–recall balance, particularly for diffuse entity types such as MISC and ORG, while this improvement was smaller in English. This is partly due to scorer behaviours (e.g. repair vs. discard strategies in span matching). Repair mechanisms can mask annotation inconsistencies in the Arabic gold standard by forcing invalid label sequences into valid spans, which affects how errors are counted. The scheme change (IOB1⟶IOB2) reduces these inconsistencies, resulting in a measurable improvement in precision–recall balance, particularly for Arabic. Error distributions showed systematic language-specific patterns. In Arabic, boundary and exclusion errors were dominant at the span level, often caused by overlaps with the O class and ambiguous span closure. In contrast, English exhibited more inclusion errors and cross-entity confusion, particularly between semantically related entity types. These findings demonstrate that evaluation metrics are not neutral: they interact with data characteristics and model behaviour in ways that can either mask or exaggerate performance differences. RQ3 therefore shows that part of the observed performance discrepancy is an artefact of evaluation design, reinforcing the need for transparent and component-aware evaluation.

**RQ4** examined how the data, model, and evaluation components interact and how their combined influence explains the Arabic–English performance gap. The findings showed that component effects are not independent: data-level characteristics such as lexical sparsity and annotation ambiguity (RQ1) propagate into model behaviour (RQ2), increasing uncertainty and lowering representation separability, particularly for low-frequency and ambiguous entities. These behaviours, in turn, interact with evaluation design (RQ3), where span-repair mechanisms, averaging strategies, and annotation inconsistencies can mask or amplify errors. RQ4 therefore establishes that the performance gap emerges from cross-component interactions rather than from any single factor, demonstrating the need for component-aware diagnostic evaluation such as that implemented by DeformAr.

**RQ5** used behavioural analysis to explain *why* the differences observed in earlier chapters (RQ1–RQ4) occur by linking data characteristics to model behaviour through interpretability techniques and token-level diagnostics. Fine-tuning analysis showed that representational changes were larger and less stable in Arabic than in English, particularly for sparse or ambiguous tokens, indicating that model adaptation is constrained by data structure. Representation-space analysis revealed systematic patterns and anomalies: overlapping or weakly separated clusters corresponded to high-loss and low-confidence tokens, especially for `MISC` and ORG entities in Arabic. Instance-level inspection confirmed that many Arabic errors originated from annotation ambiguity, tokenisation splits, and systematic annotation errors. Clustering alignment further showed weaker correspondence between learned representations and gold labels in Arabic, indicating inconsistent entity encoding compared to English. Overall, RQ5 demonstrated that the cross-lingual performance gap arises from representational instability and weak semantic alignment in Arabic, driven by data sparsity and annotation inconsistency—insights that only emerged when behavioural analysis was combined with cross-component diagnostics.

Together, these findings demonstrate that **DeformAr** provides a cross-component and cross-lingual diagnostic framework capable of revealing how linguistic, representational, and evaluative factors jointly shape system performance, and how each subcomponent within the system is affected by these factors.

## 6.4 Contributing Factors and Cross-Lingual Insights

The analyses presented in Chapters 4 and 5 identified several interacting factors that collectively explain the performance discrepancies between Arabic and English NER systems. Rather than revisiting each in detail, this section summarises the most influential patterns and situates them within the broader diagnostic interpretation.

Across both analytical phases, three dimensions consistently emerged as central drivers of variation: **(1) tokenisation and model architecture**, **(2) language complexity**, and **(3) training and evaluation practices**. Tokenisation interacts strongly with language morphology—English tokenisation increases redundancy and supports generalisation, while Arabic tokenisation often amplifies ambiguity and sparsity. Linguistic complexity further compounds this gap: Arabic's morphological richness, uneven lexical distributions, and orthographic variation introduce instability in both representation and calibration. Finally, training- and evaluation-related factors—including annotation inconsistency, systematic errors, and the evaluation mode implemented in the scorer—further shape these trends, making Arabic systems appear less stable even when their underlying predictions are linguistically plausible.

Taken together, these interacting factors highlight that the cross-lingual gap is not attributable to a single weakness but to cumulative misalignment across data, model, and evaluation components. English benefits from structural alignment among these components, while Arabic models are challenged by compounding inconsistencies across all three levels. These findings bridge what the framework has achieved with the limitations and future directions that follow.

## 6.5 Limitations

While the DeformAr framework offers a comprehensive cross-component analysis of NER systems, several limitations remain that highlight directions for future refinement:

- **Coverage of contributing factors:** Although the framework examined key factors influencing performance discrepancies, it did not capture all possible sources discussed in the literature—such as dataset design, domain mismatch, and lexical diversity. For example, the relationship between lexical variation and token-level errors could not be fully quantified.

- **Evaluating behavioural analysis insights:** While the behavioural analysis phase uncovered meaningful structural and behavioural patterns, many of these findings are descriptive. Future work should focus on turning these observations into quantifiable features that can be validated and generalised through behavioural metrics.

- **Scalability of instance-level exploration:** The instance-level inspection remains manual and example-driven, limiting scalability and automation. Integrating active learning strategies or efficient search mechanisms could improve the identification of systematic error patterns at scale.

- **Simplified similarity techniques:** The current token similarity retrieval relies on cosine similarity, which may overlook subtle contextual relationships. More advanced alternatives, such as $k$-nearest neighbour retrieval or contextualised embedding matching, could enhance precision and interpretability in instance-level analysis.

## 6.6 Future Directions

The DeformAr framework opens several areas for further development and extension. One direction is to apply the framework to other languages, particularly under-resourced and morphologically rich ones beyond Arabic. Cross-lingual comparisons involving languages such as Persian, Amharic, or Urdu — as well as different Arabic dialects — could help validate the generality of DeformAr's findings and reveal whether similar structural or behavioural discrepancies emerge.

Another promising direction is to apply the analysis to new datasets, including revised or cleaned versions of ANERCorp and CoNLL. Such studies would allow us to evaluate whether ongoing annotation corrections effectively mitigate systemic errors or whether deeper representational inconsistencies persist within the data–model interface.

A further extension concerns model scale and architecture. Applying DeformAr to larger pretrained models or domain-specific variants — such as those explored by Inoue et al. (2021) — could help disentangle the effects of model capacity from data quality. Evaluating different checkpoints of multilingual or domain-tuned transformers would reveal how scaling influences representational stability and cross-lingual robustness.

In terms of linguistic analysis, future work could further investigate the impact of tokenisation by comparing multilingual and monolingual tokenisation strategies, potentially building on the work of Rust et al. (2021) on multilingual tokenisation. Additionally, the framework's behavioural metrics could be expanded to include more subcomponent-level measures, such as attention-based behavioural metrics or attribution-informed uncertainty. These extensions would enable a richer understanding of how tokenisation and attention jointly influence representational stability across languages.

Enhancing the interactivity and visual analytics capabilities of the framework is another important goal. Integrating richer visualisation methods and scalable interaction (e.g., dynamic filtering, progressive sampling) would enable researchers to examine model behaviour at multiple levels of abstraction — from global patterns to instance-level anomalies — and improve the framework's value for both annotation and model-debugging workflows.

**Adapting the framework for LLMs.** Large Language Models (LLMs) introduce new challenges and opportunities for diagnostic evaluation. Unlike NER systems, LLMs are typically generative, instruction-tuned, and capable of multi-step reasoning, which complicates component-based decomposition. Nevertheless, DeformAr's principles can extend naturally to this setting.

- The **data component** can be reframed to analyse instruction–response or prompt–completion pairs, focusing on how linguistic ambiguity, tokenisation, and format consistency influence generative outputs.
- The **model component** can capture layer-wise activations, decoder attention, and intermediate representation shifts, enabling the study of reasoning patterns, calibration, or factual consistency across prompts and languages.
- **Behavioural metrics** such as loss or confidence can be replaced or supplemented

with generative proxies (e.g., entropy, perplexity, response coherence, factuality, or calibration error).

While adapting DeformAr to LLMs would require re-engineering several components — particularly to handle autoregressive decoding and extended context windows — its underlying diagnostic logic remains applicable. Attention-based and representation-space diagnostics, including adapted measures such as silhouette and cluster coherence, could help characterise how LLMs evolve representational and reasoning patterns across layers. By analysing systematic shifts in attention or embedding structure, the framework could help characterise emergent behaviours such as hallucination, over-generalisation, or instability in open-weight models. In this way, DeformAr provides a pathway for interpretable, cross-lingual analysis of how generative models organise information and produce responses, extending diagnostic evaluation beyond token-level prediction tasks.

Finally, DeformAr can be extended to other NLP tasks beyond NER. For sequence-labelling tasks such as part-of-speech tagging, morphological analysis, or syntactic parsing, the existing framework can be applied directly, since the data structure, token-level granularity, and evaluation logic remain the same. For other NLP tasks — including sentiment analysis, machine translation, and question answering — the same conceptual diagnostics apply. By redefining the diagnostic components and behavioural metrics for each setting, the same cross-component structure can be used to trace how input variation or annotation noise influences downstream prediction dynamics.

## 6.7 Conclusions

This thesis introduced **DeformAr**, a cross-lingual diagnostic framework developed to understand and explain the performance gap between Arabic and English Named Entity Recognition (NER) systems. Unlike traditional evaluation approaches that rely solely on aggregate scores, DeformAr provides a multi-layered analysis that examines how data, model, and evaluation components interact to produce observed performance differences.

The framework begins by decomposing each system into its core components and extracting structured diagnostic data. This enables systematic measurement of key factors such as tokenisation behaviour, annotation consistency, and representational quality. The quantitative phase identifies where and how discrepancies arise, while the behavioural analysis phase builds on these results to explore the underlying representational and behavioural patterns in greater depth. Together, these analyses reveal how language-specific data properties interact with model dynamics to shape cross-lingual performance.

Across both components, our findings consistently show that Arabic models face greater challenges stemming from tokenisation artefacts, annotation inconsistencies, and lexical variability, all of which lead to noisier internal representations and less stable decision boundaries. English models, by contrast, develop cleaner, more task-aligned representations that closely mirror gold-standard annotation structures.

DeformAr offers a new perspective on NER evaluation and interpretability, particularly for under-resourced and morphologically rich languages. By linking model performance to underlying data quality and internal representational behaviour, it provides a structured framework for diagnosis, debugging, and future model improvement. To our knowledge, this is the first Arabic-specific interpretability framework that integrates quantitative measurement with interactive, component-level behavioural analysis.

Looking ahead, the framework lays the groundwork for broader extensions — to other languages, NLP tasks, and even generative model settings — where similar diagnostic principles can be applied to trace how structure, behaviour, and data jointly shape system performance.

# Additional Results: DeCoNLL Classification Reports

A

To complement the summary reported in Section 4.5.3, this appendix provides the full token-level and entity-level classification reports for BERT evaluated on the downsampled CoNLL-2003 dataset (DeCoNLL). These reports show detailed per-class performance across entity tags.

Table A.1: Token-level classification report for BERT on DeCoNLL (IOB1 mode).

| Tag | Precision | Recall | F1 | Support |
|---|---|---|---|---|
| LOC | 0.9183 | 0.9476 | 0.9327 | 534 |
| MISC | 0.8643 | 0.7964 | 0.8290 | 280 |
| ORG | 0.8406 | 0.8505 | 0.8455 | 428 |
| PERS | 0.9553 | 0.9637 | 0.9595 | 798 |
| **micro** | 0.9096 | 0.9127 | 0.9112 | 2040 |
| **macro** | 0.8946 | 0.8895 | 0.8917 | 2040 |
| **weighted** | 0.9091 | 0.9127 | 0.9106 | 2040 |

Table A.2: Entity-level classification report for BERT on DeCoNLL (IOB1 mode).

| Tag | Precision | Recall | F1 | Support |
|---|---|---|---|---|
| B-LOC | 0.9287 | 0.9513 | 0.9399 | 534 |
| B-MISC | 0.9012 | 0.8143 | 0.8555 | 280 |
| B-ORG | 0.8832 | 0.8832 | 0.8832 | 428 |
| B-PERS | 0.9614 | 0.9674 | 0.9644 | 798 |
| I-LOC | 0.8571 | 0.8780 | 0.8675 | 82 |
| I-MISC | 0.7907 | 0.5231 | 0.6296 | 65 |
| I-ORG | 0.8938 | 0.8971 | 0.8954 | 272 |
| I-PERS | 0.9741 | 0.9869 | 0.9804 | 457 |
| O | 0.9964 | 0.9978 | 0.9971 | 16117 |
| **macro** | 0.9096 | 0.8777 | 0.8903 | 19033 |
| **weighted** | 0.9858 | 0.9861 | 0.9858 | 19033 |